%% file: ms.tex
\documentclass{article}

\include{preamble}

\usepackage[accepted]{icml2021arxiv}

\icmltitlerunning{Bayesian Quadrature on Riemannian Data Manifolds}

\begin{document}

\twocolumn[
\icmltitle{Bayesian Quadrature on Riemannian Data Manifolds}

\icmlsetsymbol{equal}{*}
\begin{icmlauthorlist}
\icmlauthor{Christian Fr\"ohlich}{unitue}
\icmlauthor{Alexandra Gessner}{equal,unitue,mpiis}
\icmlauthor{Philipp Hennig}{unitue,mpiis}
\icmlauthor{Bernhard Sch\"olkopf}{mpiis}
\icmlauthor{Georgios Arvanitidis}{equal,mpiis}
\end{icmlauthorlist}

\icmlaffiliation{unitue}{University of Tübingen, Germany}
\icmlaffiliation{mpiis}{Max Planck Institute for Intelligent Systems, Tübingen, Germany}

\icmlcorrespondingauthor{Christian Fr\"ohlich}{christian.froehlich@student.uni-tuebingen.de}

% Keywords
\icmlkeywords{Probabilistic Numerics, Riemannian manifolds, Bayesian quadrature}

\vskip 0.3in
]
\printAffiliationsAndNotice{\icmlEqualSupervision}  % leave blank if no need to mention equal contribution

%%%%%%%%%%%%%%%%
%%% ABSTRACT %%% 
%%%%%%%%%%%%%%%%
\begin{abstract}
Riemannian manifolds provide a principled way to model nonlinear geometric structure inherent in data.
A Riemannian metric on said manifolds determines geometry-aware shortest paths and provides the means to define statistical models accordingly.
However, these operations are typically computationally demanding.
To ease this computational burden, we advocate probabilistic numerical methods for Riemannian statistics.
In particular, we focus on Bayesian quadrature (\bq) to numerically compute integrals over normal laws on Riemannian manifolds learned from data.
In this task, each function evaluation relies on the solution of an expensive initial value problem.
We show that by leveraging both prior knowledge and an active exploration scheme, \bq~significantly reduces the number of required evaluations and thus outperforms Monte Carlo methods on a wide range of integration problems.
As a concrete application, we highlight the merits of adopting Riemannian geometry with our proposed framework on a nonlinear dataset from molecular dynamics.
\end{abstract}

%%% INCLUDE CONTENT (mory tidy approach)
\input{sections/01_intro}
\input{sections/02_riemann}
\input{sections/03_land}
\input{sections/04_bq}
\input{sections/05_exp}
\input{sections/06_discussion}
\section*{Acknowledgements} % only in final version
AG acknowledges funding by the European Research Council through ERC StG Action 757275 / PANAMA and support by the International Max Planck Research School for Intelligent Systems (IMPRS-IS).
This work was supported by the German Federal Ministry of Education and Research (BMBF): Tübingen AI Center, FKZ: 01IS18039B, and by the Machine Learning Cluster of Excellence, EXC number 2064/1 – Project number 390727645.
The authors thank Nicholas Kr\"amer, Agustinus Kristiadi, and Dmitry Kobak for useful comments and discussions.

%%% BIBLIOGRAPHY %%%
% \clearpage
\bibliography{bib}
\bibliographystyle{icml2021}

%\pagebreak
\appendix
\input{sections/07_supplementary}

\end{document}

%% file: preamble.tex
\usepackage{amsmath,amsfonts,amssymb, mathtools}
\usepackage{nccmath}
\usepackage{mathrsfs}
\usepackage{lmodern}
\usepackage{nicefrac}
\usepackage{enumitem}
\usepackage{colonequals}
\usepackage{xcolor}
\usepackage{microtype}
\usepackage{graphicx}
\usepackage{subfigure}
\usepackage{booktabs} % for professional tables
\setlist{itemsep=3pt}

\newcommand{\RN}[1]{%
  \textup{\uppercase\expandafter{\romannumeral#1}}%
}

\usepackage{csquotes} 
\usepackage{hyperref}
\usepackage{tikz}
\usetikzlibrary{plotmarks}
\usepackage{pgf}
\usepackage[percent]{overpic}
\newcommand{\bq}{\textsc{bq}}
\newcommand{\mc}{\textsc{mc}}
\newcommand{\wsabi}{\textsc{wsabi}}
\newcommand{\wsabil}{\textsc{wsabi-l}}
\newcommand{\wsabim}{\textsc{wsabi-m}}

\newcommand*{\R}{\mathbb{R}}
\newcommand{\inner}[2]{\langle #1, #2\rangle}

\newcommand{\argmin}{\operatorname*{arg\:min}}

\renewcommand{\d}{\,\operatorname{d}\!}
\newcommand{\diag}{\operatorname{diag}}

\newcommand{\expmap}{\operatorname{Exp}}

\newcommand{\logmap}{\operatorname{Log}}

\newcommand{\bm}[1]{\boldsymbol{\mathsf{#1}}}
\newcommand{\bmx}{\bm{x}}
\newcommand{\bmv}{\bm{v}}
\newcommand{\bmgamma}{\bm{\gamma}}
\newcommand{\N}{\mathcal{N}}
 
\newcommand{\GP}{\mathcal{GP}}

\newcommand{\M}{\mathcal{M}}

%% file: sections/01_intro.tex
\section{Introduction}
\label{sec:introduction}

The tacit assumption of a Euclidean geometry, implying that distances can be measured along straight lines, is inadequate when data follows a nonlinear trend, which is known as the \emph{manifold hypothesis}. As a result, probability distributions based on a flat geometry may poorly model the data and fail to capture its underlying structure. Generalized distributions that account for curvature of the data space have been put forward to alleviate this issue. In particular, \citet{pennec2006} proposed an extension of the normal distribution on Riemannian manifolds such as the sphere.

\begin{figure}[t]
    \centering
    \includegraphics[width=0.47\textwidth]{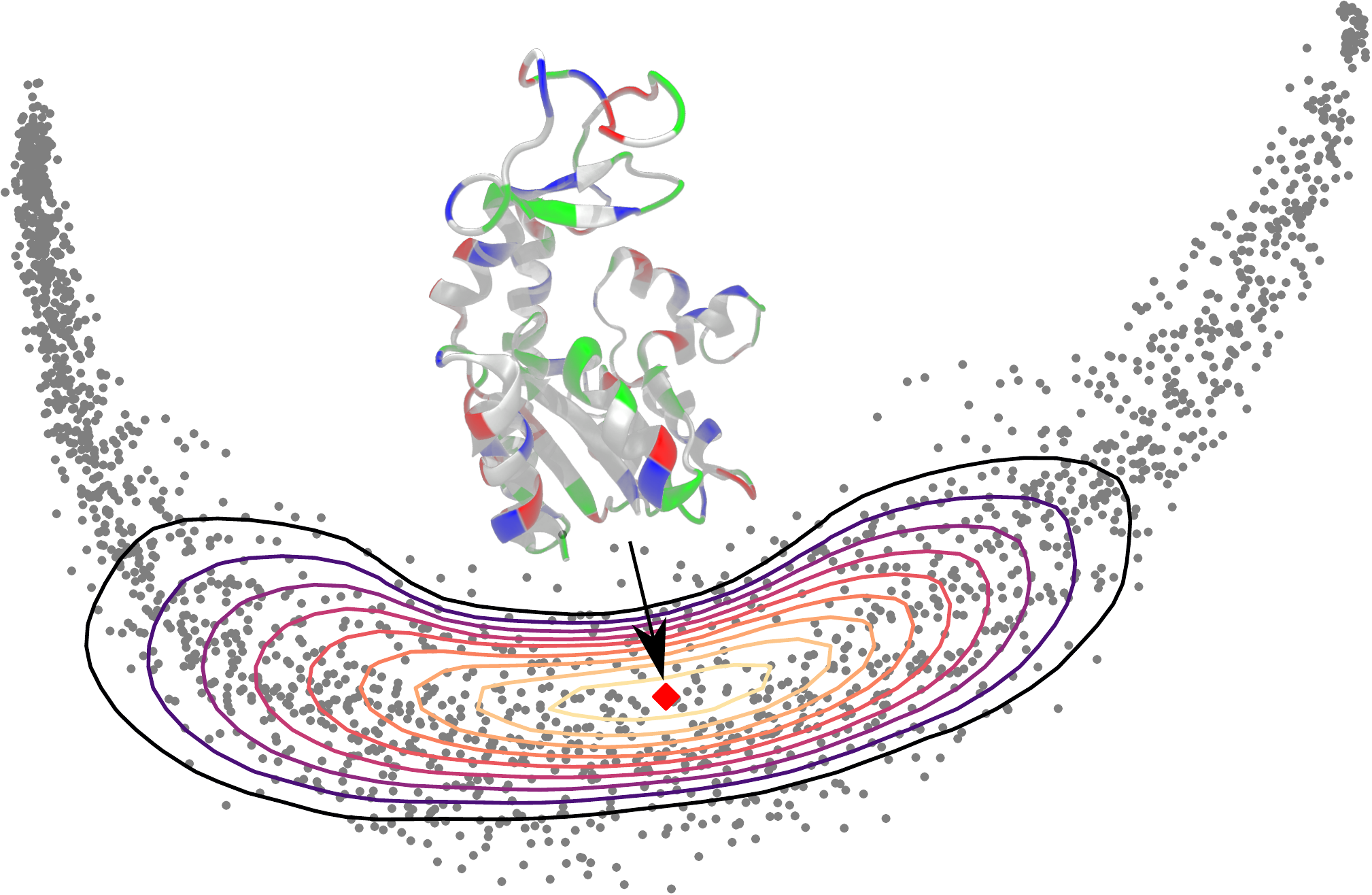}
    \caption{A \textsc{land} on a protein trajectory manifold of the enzyme adenylate kinase. Each datum represents a conformation of the protein, i.e., a spatial arrangement of its atoms. The conformation corresponding to the \textsc{land} mean ({\protect\tikz[red,scale=1.0, rotate around={45:(0,0)}]\pgfuseplotmark{square*};}) is visualized.}
    \label{fig:teaser}
\end{figure}

The key strategy to use such distributions on general data manifolds is by replacing straight lines with continuous shortest paths, known as geodesics, which respect the nonlinear structure of the data. This is achieved by introducing a Riemannian metric in the data space that specifies how distance and volume are distorted locally. 

To this end, \citet{arvanitidis2016land} proposed a maximum likelihood estimation scheme based on a data-induced metric to learn the parameters of a \textit{Locally Adaptive Normal Distribution} (\textsc{land}), illustrated in Fig.~\ref{fig:teaser}. However, it relies on a computationally expensive optimization task that entails the repeated numerical integration of the unnormalized density on the manifold, for which no closed-form solution exists. Hence we are interested in techniques to improve the efficiency of the numerical integration scheme.

Probabilistic numerics \citep{HenOsbGir15,Cockayne2019} frames computations that are subject to noise and approximation error as active inference. A probabilistic numerical routine treats the computer as a data source, acquiring evaluations according to a policy to decrease its uncertainty about the result. Amongst probabilistic numerical methods, we focus on Bayesian quadrature (\bq) to compute intractable integrals on data manifolds. Bayesian quadrature \citep{o1991bayes,briol2019probabilistic} treats numerical integration as an inference problem by constructing posterior measures over integrals given observations, i.e., evaluations of the integrand. Besides providing sound uncertainty estimates, the probabilistic approach permits the inclusion of prior knowledge about properties of the function to be integrated, and leverages active learning schemes for node selection as well as transfer learning schemes, for example when multiple similar integrals have to be jointly estimated \citep{Xi2018, GessnerGM2019}. These properties make \bq~especially relevant in settings where the integrand is expensive to evaluate, and make it a suitable tool for integration on Riemannian data manifolds. 

\paragraph{Contributions} 
\begin{itemize}
    \item The uptake of Riemannian methods in machine learning is principally hindered by prohibitive computational costs. We here address a key aspect of this bottleneck by improving the efficiency of integration on data manifolds.
    \item We customize Bayesian quadrature to curved spaces by exploiting knowledge about their structure. To this end, we introduce a tailored acquisition function that minimizes the number of expensive computations by selecting informative \emph{directions} (instead of single points) on the manifold. Adopting a probabilistic numerical integration scheme enables efficient exploration of the integrand's support under a suitable prior.
    \item We demonstrate accuracy and performance of our approach on synthetic and real-world data manifolds, where we observe speedups by factors of up to $20$. In these examples we focus on the \textsc{land} model, which provides a wide range of numerical integration problems of varying geometry and difficulty. We highlight molecular dynamics as a promising application area for Riemannian machine learning models. The results support the use of probabilistic numerical methods within Riemannian geometry to achieve significant speedup.
\end{itemize}   

%% file: sections/02_riemann.tex
\section{Riemannian Geometry}
\label{sec:riemann}

\begin{figure}
    \centering
    \begin{overpic}[width=0.46\textwidth]{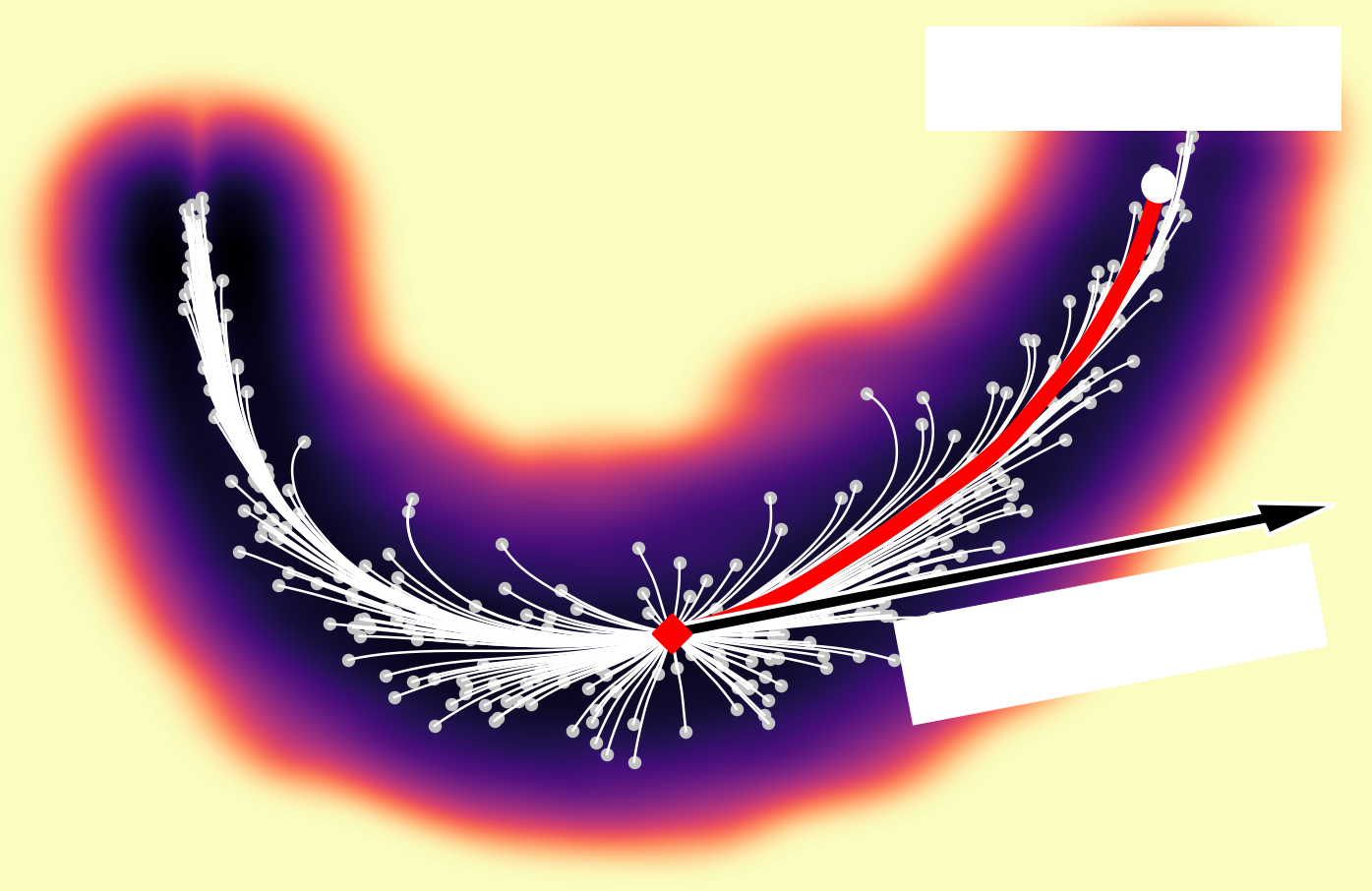}
 \put (3,3) {\large$\displaystyle\mathcal{M}$}
 \put (68.5,58) {\large$\displaystyle\expmap_{\bm{\mu}}(\bm{v}) = \bm{x}$}
 \put (66,15)
 {\rotatebox{10}{\large$\displaystyle\logmap_{\bm{\mu}}(\bm{x}) = \bm{v}$}}
\end{overpic}
    \caption{A protein trajectory manifold. A subset of the geodesics is shown with respect to a fixed point $\bm{\mu}$ ({\protect\tikz[red,scale=1.0, rotate around={45:(0,0)}]\pgfuseplotmark{square*};}). The background is colored according to the volume element $\sqrt{\lvert \bm{M} \rvert}$ (Sec.~\ref{sec:metrics}) on a log scale. We omit a colorbar, since the values are not easily interpreted. Darker color indicates regions with small metric, to which shortest paths are attracted. For one geodesic ({\protect\raisebox{2pt}{\tikz{\draw[red,solid,line width=1.5pt](0,0) -- (2mm,0);}}}), we show the downscaled tangent vector and the $\logmap$ and $\expmap$ operations.}
    \label{fig:manifold}
\end{figure}

In our applied setting, we view $\R^D$ as a \emph{smooth manifold}~$\M$ with a changed notion of distance and volume measurement as compared to the Euclidean case. This view arises from the assumption that data have a general underlying nonlinear structure in $\R^D$ (see Fig.~\ref{fig:manifold}), and thus, the following discussion excludes manifolds with structure known a priori, e.g., spheres and tori. In our case, the \emph{tangent space} $\mathcal{T}_{\bm{\mu}}\mathcal{M}$ at a point $\bm{\mu}\in\M$ is again $\R^D$, but centered at $\bm{\mu}$. This is a vector space that allows to represent points of the manifold as tangent vectors $\bm{v}\in\R^D$. Pictorially, a vector $\bm{v} \in \mathcal{T}_{\bm{\mu}}\mathcal{M}$ is tangential to some curve passing through $\bm{\mu}$. Together, these vectors give a linearized view of the manifold with respect to a base point $\bm{\mu}$. 
A \textit{Riemannian metric} is a positive definite matrix $\bm{M}:\R^D\rightarrow\R_+^{D\times D}$ that varies smoothly across the manifold. Therefore, we can define a local inner product between tangent vectors $\bm{v},\bm{w}\in\mathcal{T}_{\bm{\mu}}\M$ as $\inner{\bm{v}}{\bm{w}}_{\bm{\mu}} = \inner{\bm{v}}{\bm{M}(\bm{\mu}) \bm{w}}$, where $\inner{\cdot}{\cdot}$ is the Euclidean inner product. This inner product makes the smooth manifold a \textit{Riemannian manifold} \citep{carmo1992riemannian,lee:2018}.

A Riemannian metric locally scales the infinitesimal distances and volume. Consider a curve $\bm{\gamma}:[0,1] \rightarrow \M$ with $\bm{\gamma}(0)=\bm{\mu}$ and $\bm{\gamma}(1)=\bm{x}$. The length of this curve on the Riemannian manifold $\M$ is computed as
\begin{equation}
    L(\bm{\gamma}) = \int_{0}^{1} \sqrt{\langle \dot{\bm{\gamma}}(t), \bm{M}(\bm{\gamma}(t))  \dot{\bm{\gamma}}(t) \rangle} \d t,
\end{equation}
where $\dot{\bm{\gamma}}(t)=\frac{\d}{\d t} \bm{\gamma}(t)\in\mathcal{T}_{\bm{\gamma}(t)}\M$ is the velocity of the curve. The $\bm{\gamma}^*$ that minimizes this functional is the shortest path between the points. To overcome the invariance of $L$ under reparametrization of $\bm{\gamma}$, shortest paths can equivalently be defined as minimizers of the energy functional. This enables application of the Euler-Lagrange equations, which result in a system of $2^{\text{nd}}$ order nonlinear ordinary differential equations (\textsc{ode}s). The shortest path is obtained by solving this system as a boundary value problem (\textsc{bvp}) with boundary conditions $\bm{\gamma}(0)=\bm{\mu}$ and $\bm{\gamma}(1)=\bm{x}$. Such a length-minimizing curve is known as \emph{geodesic}.

To perform computations on $\M$ we introduce two operators. The first is the \textit{logarithmic map} $\logmap_{\bm{\mu}}(\bm{x}) = \bm{v}$, which represents a point $\bm{x} \in \M$ as a tangent vector $\bm{v}\in\mathcal{T}_{\bm{\mu}}\M$. This can be seen as the initial velocity of the geodesic that reaches $\bm{x}$ at $t=1$ with starting point $\bm{\mu}$. The inverse operator is the \emph{exponential map    } $\expmap_{\bm{\mu}}(t\cdot \bm{v})=\bm{\gamma}(t)$ that takes an initial velocity $\dot{\bm{\gamma}}(0)=\bm{v}\in \mathcal{T}_{\bm{\mu}}\M$ and generates a unique geodesic with $\bm{\gamma}(0) = \bm{\mu}$ and $\bm{\gamma}(1)=\bm{x}$. Note that $\logmap_{\bm{\mu}}( \expmap_{\bm{\mu}} (\bm{v})) = \bm{v}$, and also, $\|\logmap_{\bm{\mu}}(\bm{x})\|_2 = \|\bm{v}\|_2 = L(\bm{\gamma})$. Computationally, the logarithmic map amounts to solving a \textsc{bvp}, whereas the exponential map corresponds to an initial value problem (\textsc{ivp}). For general data manifolds, analytic solutions of the geodesic equations do not exist, so we rely on specialized approximate numerical solvers for the \textsc{bvp}s; however, finding shortest paths still remains a computationally expensive problem \citep{hennig:aistats:2014,arvanitidis:aistats:2019}. In contrast, the exponential map as an \textsc{ivp} is an easier problem and solutions are significantly more efficient.

We illustrate our applied manifold setting in Fig.~\ref{fig:manifold}, where we show geodesics starting from a point $\bm{\mu}$, as well as the $\logmap$ and $\expmap$ operators between $\bm{\mu}$ and a point $\bm{x}$. 
Note that Figs.~\ref{fig:teaser} to \ref{fig:bq_tangent} are in correspondence, that is, they illustrate different aspects of the same setting. More details on Riemannian geometry and the \textsc{ode} system are in~\ref{app:riemann}.

\subsection{Constructing Riemannian Manifolds from Data}
\label{sec:metrics}

The Riemannian volume element or measure $\d \mathcal{M}(\bm{x}) =  \sqrt{\lvert \bm{M}(\bm{x}) \rvert} \d \bm{x}$ represents the distorted infinitesimal standard Lebesgue measure $\d\bm{x}$. For a meaningful metric, this quantity is small near the data and increases as we move away from them.
Intuitively, this metric behavior pulls shortest paths near the data.

There are broadly two unsupervised approaches to learn such an adaptive metric from data. Given a dataset $\bm{x}_{1:N}$ of $N$ points in $\R^D$, \citet{arvanitidis2016land} proposed a nonparametric metric to model nonlinear data trends as the inverse of a local diagonal covariance matrix with entries
\begin{equation}
    M_{dd}(\bm{x}) = \left(\sum_{n=1}^{N} w_n(\bm{x}) (x_{nd} - x_d)^2 + \rho \right)^{-1},
\end{equation}
where the weights $w_n$ are obtained from an isotropic Gaussian kernel $w_n(\bm{x}) = \exp\left(- \frac{||\bm{x}_n - \bm{x}||^2}{2\sigma^2}\right)$. The lengthscale $\sigma$ determines the curvature of the manifold, i.e., how fast the metric changes. The hyperparameter $\rho$ controls the value of the metric components that is reached far from the data, so the measure there is $\sqrt{\bm{\lvert M \rvert}}=\rho^{-\frac{D}{2}}$. Typically, $\rho$ is set to a small scalar to encourage geodesics to follow the data trend. However, this metric does not scale to higher dimensions due to the curse of dimensionality \citep[Ch.~1.4]{bishop:2006}.

Another approach relies on generative models to capture the geometry of high-dimensional data in a low-dimensional latent space \citep{tosi:UAI:2014,arvanitidis2017latent}. Let a dataset $\bm{y}_{1:N}~\in~\R^{D'}$ with latent representation $\bm{x}_{1:N}~\in~\R^D$ and $D' > D$, such that $\bm{y}_n \approx \bm{g}(\bm{x}_n)$ where $\bm{g}$ is a stochastic function with Jacobian $\bm{J}_{\bm{g}}(\bm{x})\in\R^{D' \times D}$. Then, the \textit{pullback metric} $\bm{M}(\bm{x}) = \mathbb{E}[\bm{J}^\intercal_{\bm{g}}(\bm{x}) \bm{J}_{\bm{g}}(\bm{x})]$ is naturally induced in $\R^D$, which enables the computation of lengths that respect the geometry of the data manifold in $\R^{D'}$. Even though this metric reduces the dimensionality of the problem and can be learned directly from the data by learning $\bm{g}$, it is computationally expensive to use due to the Jacobian.

To mitigate this shortcoming, we propose a surrogate Riemannian metric. Consider a Variational Auto-Encoder (\textsc{vae}) \citep{kingma2014autoencoding, rezende:icml:2014} with encoder
$q_\phi(\bm{x}|\bm{y}) = \mathcal{N}(\bm{x}~|~\bm{\mu}_\phi(\bm{y}), \diag(\bm{\sigma}^2_\phi(\bm{y})))$, decoder $p_\theta(\bm{y}|\bm{x}) = \mathcal{N}(\bm{y}~|~\bm{\mu}_\theta(\bm{x}),\diag(\bm{\sigma}^2_\theta(\bm{x})))$ and prior $p(\bm{x})=\mathcal{N}(\bm{x}~|~\bm{0},\mathbb{I}_D)$, 
with deep neural networks as the functions that parametrize the distributions. Then, the aggregated posterior is
\begin{equation}
    q_\phi(\bm{x}) = \int_{\R^{D'}} q_\phi(\bm{x}~|~\bm{y}) p(\bm{y}) \d\bm{y} \approx \frac{1}{N} \sum_{n=1}^N q_\phi(\bm{x}~|~\bm{y}_n),
\end{equation}
where the integral is approximated from the training data. This is a Gaussian mixture model that assigns non-zero density only near the latent codes of the data. Thus, motivated by \citet{arvanitidis:arxiv:2020} we define a diagonal Riemannian metric in the latent space as
\begin{equation}
    \bm{M}(\bm{x}) = (q_\phi(\bm{x}) + \rho)^{-\frac{2}{D}} \cdot \mathbb{I}_D.
\end{equation}
This metric fulfills the desideratum of modeling the local behavior of the data in the latent space and it is more efficient than the pullback metric. The variance $\sigma^2_\phi(\cdot)$ of the components is typically small, so the metric adapts well to the data, which, however, may result in high curvature.

%% file: sections/03_land.tex
\section{Gaussians on Riemannian Manifolds}
\label{sec:land}

\citet{pennec2006} theoretically derived the \textit{Riemannian normal distribution} as the maximum entropy distribution on a manifold $\mathcal{M}$, given a mean $\bm{\mu}$ and covariance matrix $\bm{\Sigma}$. The density\footnote{The covariance $\bm{\Sigma}_\M$ on $\M$ is not equal to $\bm{\Gamma}_\M^{-1}$, but the covariance on $\mathcal{T}_{\bm{\mu}}\M$ is $\bm{\Sigma}=\bm{\Gamma}^{-1}$. Throughout the paper we take the second perspective. For additional technical details see~\ref{app:riemann}.} on $\mathcal{M}$ is
\begin{equation}
    p(\bm{x}) = \frac{1}{\mathcal{C}(\bm{\mu},\bm{\Sigma})} \exp\left( - \frac{1}{2} \left\langle \logmap_{\bm{\mu}}(\bm{x}), \bm{\Sigma}^{-1} \logmap_{\bm{\mu}}(\bm{x})\right\rangle \right). 
\end{equation}
This is reminiscent of the familiar Euclidean density, but with a Mahalanobis distance based on the nonlinear logarithmic maps. 
Analytic solutions for the normalization constant $\mathcal{C}$ can be given only for certain manifolds that are known a priori, like the sphere, since this requires analytic solutions for
the logarithmic and exponential maps.

\begin{figure}
    \centering
    \begin{overpic}[width=0.46\textwidth]{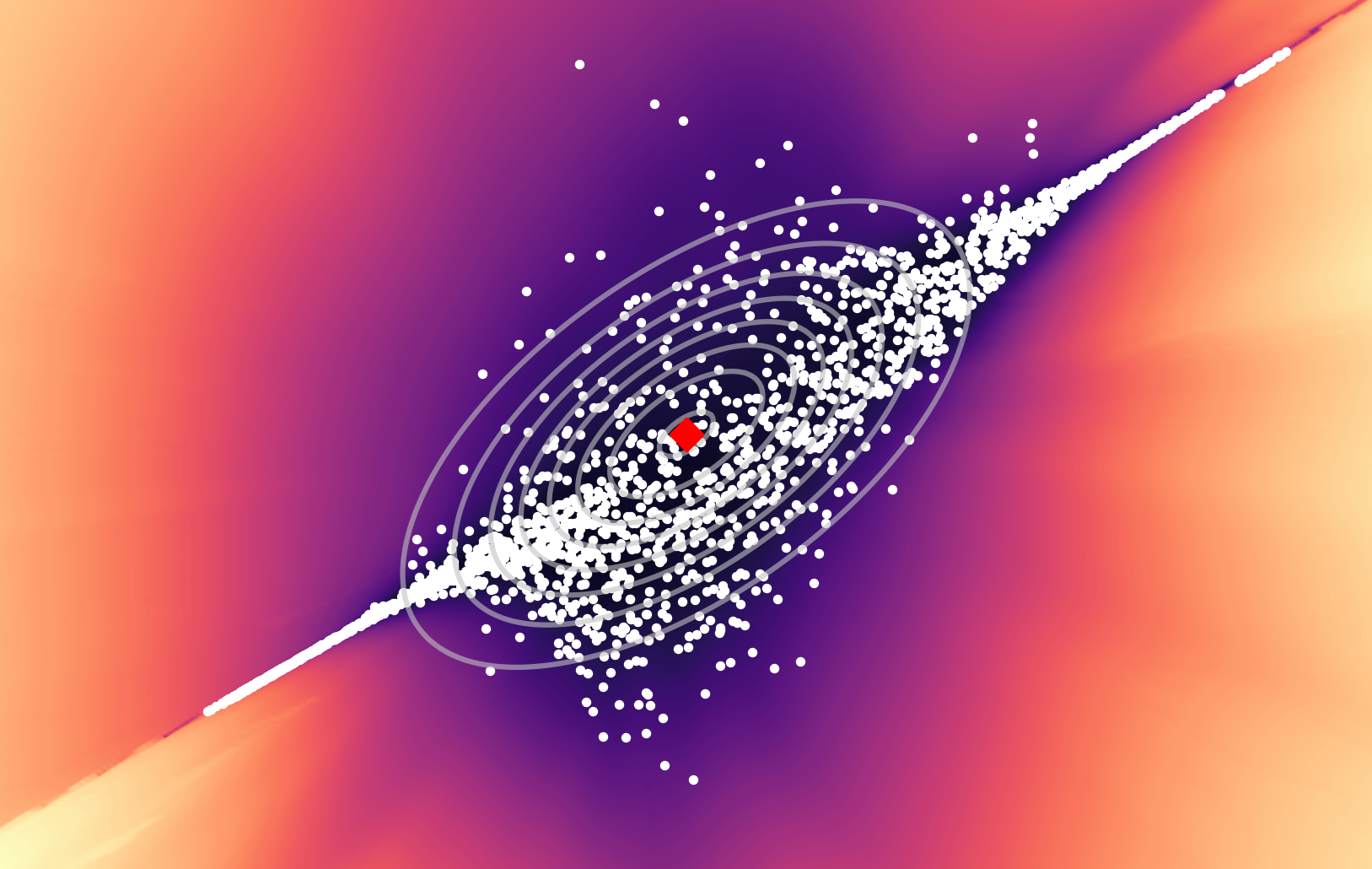}
 \put (3,3) {\large$\displaystyle\mathcal{T}_{\bm{\mu}}\mathcal{M}$}
\end{overpic}
    \caption{The function $g_{\bm{\mu}}$ on the tangent space of the protein trajectory manifold. The origin $\bm{0}$ ({\protect\tikz[red,scale=1.0, rotate around={45:(0,0)}]\pgfuseplotmark{square*};}) corresponds to the point $\bm{\mu}$ on the manifold from which the exponential maps are computed. Contours of the integration measure $\mathcal{N}(\bm{v}; \bm{0}, \bm{\Sigma})$ are in light gray. Logarithmic maps $\logmap_{\bm{\mu}}(\bm{x}_n)$ of the data are scattered in white. The background is colored according to the volume element on a log scale. The color scale is quantitatively unrelated to Fig.~\ref{fig:manifold}.}
    \label{fig:tangentspace}
\end{figure}

\citet{arvanitidis2016land} extended this distribution to general data manifolds (see Fig.~\ref{fig:teaser}) where the Riemannian metric is learned as discussed in Sec.~\ref{sec:metrics}. In this case, $\M=\R^D$ and $\mathcal{T}_{\bm{\mu}}\M = \R^D$, so $\bm{\Sigma}\in\R_+^{D\times D}$. The normalization constant is computed using a na\"ive Monte Carlo scheme as

\begin{align}
\label{eqn:normalization_constant}
        &\int_{\mathcal{M}}  \exp\left( - \frac{1}{2} \langle \logmap_{\bm{\mu}}(\bm{x}), \bm{\Sigma}^{-1} \logmap_{\bm{\mu}}(\bm{x})\rangle \right) \d \mathcal{M}(\bm{x}), \\
    &=\ \sqrt{(2\pi)^D |\bm{\Sigma}|} \int_{\mathcal{T}_{\bm{\mu}}\mathcal{M}} g_{\bm{\mu}} (\bm{v})\, \N (\bm{v}; \bm{0}, \bm{\Sigma}) \d\bm{v} =\mathcal{C}(\bm{\mu},\bm{\Sigma}), \nonumber
\end{align}
where $g_{\bm{\mu}} (\bm{v}) = \sqrt{|\bm{M}(\expmap_{\bm{\mu}}(\bm{v}))|}$ gives the tangent space view on the volume element. An example of $g_{\bm{\mu}}$ is depicted in Fig.~\ref{fig:tangentspace}.
Instead of having to solve \textsc{bvp}s for the logarithmic maps, we now integrate on the Euclidean tangent space, where we solve significantly faster exponential maps (\textsc{ivp}s). 

The normalization constant is needed to estimate maximum likelihood parameters $\bm{\mu}$ and $\bm{\Sigma}$. For this, we use gradient descent in an alternating fashion, keeping $\bm{\mu}$ fixed while optimizing $\bm{\Sigma}$ and vice versa, as detailed in~\ref{app:land}. Note that $\mathcal{C}(\bm{\mu}, \bm{\Sigma})$ acts as a regularizer, keeping $\bm{\mu}$ near the data manifold and penalizing an overestimated $\bm{\Sigma}$. Moreover, the constant enables the definition of a mixture of \textsc{land}s. 

The \mc~estimator for this integral requires the evaluation of a large number of exponential maps and is ignorant about known structure of the integrand. We replace \mc~by \bq~to drastically reduce the number of these costly evaluations needed to retain accuracy. Our foremost goal is to speed up numerical integration on data manifolds since exponential maps are, albeit faster than the \textsc{bvp}s, still relatively slow. The runtime of exponential maps depends on the employed metric (see Sec.~\ref{sec:metrics}) and on other factors such as curvature or curve length.

%% file: sections/04_bq.tex
\section{Bayesian Quadrature}
\label{sec:bq}

Bayesian quadrature (\bq) is a probabilistic approach to integration that performs Bayesian inference on the value of the integral given function evaluations and prior assumptions on the integrand (e.g., smoothness). 
The probabilistic model enables a decision-theoretic approach to the selection of informative evaluation locations. \bq~is thus inherently sample-efficient, making it an excellent choice in settings where function evaluations are costly, as is the case in Eq.~\eqref{eqn:normalization_constant} where evaluations of the integrand rely on exponential maps. The key strategy for the application of \textsc{bq} in a manifold context is to move the integration to the Euclidean tangent space. We review vanilla \textsc{bq} and then apply adaptive \textsc{bq} variants to the integration problem in Eq.~\eqref{eqn:normalization_constant}.

\subsection{Vanilla BQ}
\label{sec:vanilla}
Bayesian quadrature seeks to compute otherwise intractable integrals of the form
\begin{equation}
    \mathcal{C} = \int_{\mathbb{R}^D} f(\bmv) \pi(\bmv) \d \bmv,
\end{equation}
where $f(\bmv) : \mathbb{R}^D \rightarrow \mathbb{R}$ is a function that is typically expensive to evaluate and $\pi(\bmv)$ is a probability measure on $\mathbb{R}^D$. A Gaussian process (\textsc{gp}) prior is assumed on the integrand to obtain a posterior distribution on the integral by conditioning the \textsc{gp} on function evaluations. Such a \textsc{gp} is a distribution over functions with the characterizing property that the joint distribution of a finite number of function values is multivariate normal \citep{williams2006gaussian}. It is parameterized by its mean function $m(\bm{v})$ and covariance function (or kernel) $k(\bmv,\bmv')$, and we write $f \sim \GP(m,k)$.
After observing data $\mathcal{D}$ at input locations $\bm{V}=\bmv_{1:M}$ and evaluations $\bm{f}=f(\bmv)_{1:M}$, the posterior is
\begin{equation}
    \begin{aligned}
        m_{\mathcal{D}} (\bmv) &= m(\bmv) + k(\bmv, \bm{V}) k(\bm{V},\bm{V})^{-1} \left(\bm{f} - m(\bm{V})\right),\\
        k_{\mathcal{D}} (\bm{v}, \bm{v}') &= k(\bm{v}, \bm{v}') - k(\bm{v}, \bm{V}) k(\bm{V},\bm{V})^{-1}  k(\bm{V}, \bm{v}').
    \end{aligned}
\end{equation}

Due to linearity of the integral operator, the random variable $\mathcal{C}$ representing our belief about the integral follows a univariate normal posterior after conditioning on observations, with posterior mean $\mathbb{E}\left[\mathcal{C} \mid \mathcal{D}\right] =~\int m_{\mathcal{D}}(\bmv) \pi(\bmv) \d \bm{v}$ and variance $\mathbb{V}\left[\mathcal{C} \mid \mathcal{D}\right] =~\int k_{\mathcal{D}}(\bmv,\bm{v}') \pi(\bmv) \pi(\bm{v}') \d \bmv \d \bm{v}'$.
For certain choices of $m$, $k$ and $\pi$, these expressions have a closed-form solution \citep{briol2019probabilistic}.

\subsection{Warped BQ}
The integration task we consider is Eq.~\eqref{eqn:normalization_constant}, where the integration measure on the tangent space $\mathcal{T}_{\bm{\mu}}\mathcal{M}$ is Gaussian of the form $\pi(\bm{v}) = \N (\bm{v}; \bm{0}, \bm{\Sigma})$. To encode the known positivity of the integrand $g \coloneqq g_{\bm{\mu}}(\bmv) > 0$, we model its square-root by a Gaussian process. Let $f~=~\sqrt{2(g - \delta)},$ where  $\delta>0$ is a small scalar and $ f \sim \GP (m, k)$. Consequently, $g = \delta + \frac{1}{2} f^2$ is guaranteed to be positive. Assume noise-free observations of $f$ as $\bm{f} = \sqrt{2(g(\bm{V}) - \delta)}$. Inference is done in $f$-space and induces a 
non-Gaussian posterior on $g$.

To overcome non-Gaussianity, \citet{wsabi} proposed an algorithm dubbed \emph{warped sequential active Bayesian integration} (\wsabi). \wsabi~ approximates $g$ by a \textsc{gp} either via a local linearization on the marginal of $g$ at every location $\bm{v}$ (\wsabil) or via moment-matching (\wsabim).
The approximate mean and covariance function of $g$ can be written in terms of the moments of $f$ as
\begin{equation}
    \begin{aligned}
      \tilde{m}_{\mathcal{D}} (\bm{v}) &= \delta + \frac{1}{2} m_{\mathcal{D}} (\bm{v})^2 + \frac{\eta}{2} k_{\mathcal{D}} (\bm{v}, \bm{v}),\\
      \tilde{k}_{\mathcal{D}} (\bm{v}, \bm{v}') &= \frac{\eta}{2}  k_{\mathcal{D}} (\bm{v}, \bm{v}') + m_{\mathcal{D}} (\bm{v}) k_{\mathcal{D}} (\bm{v}, \bm{v}') m_{\mathcal{D}} (\bm{v}'),
    \end{aligned}
\end{equation}
where $\eta=0$ for \wsabil~and $\eta=1$ for \wsabim. For suitable kernels, these expressions permit closed-form integrals for \bq. 
\citet{improvedbq} discuss these and other possible transformations. \citet{NEURIPS2019_165a59f7} showed that the algorithm is consistent for $\delta>0$. 

In the present setting, \wsabi~offers three main advantages over vanilla \bq~and \textsc{mc}: First, it encodes the prior knowledge that the integrand is positive everywhere. 
Second, for metrics learned from data, the volume element typically grows fast and takes on large values away from the data.  This makes modeling $g$ directly by a $\textsc{gp}$ impractical, especially when the kernel encourages smoothness. The square-root transform alleviates this problem by reducing the dynamic range of $f$ compared to that of $g$. 
Finally, \wsabi~yields an active learning scheme that adapts to the integrand in that the utility function explicitly depends on previous function values. \citet{wsabi} proposed \textit{uncertainty sampling}, under which the next location is evaluated at the location of largest uncertainty under the integration measure. The \wsabi~objective is the posterior variance of the unwarped \textsc{gp} scaled with the squared integration measure

\begin{equation}
    u (\bm{v}) = \tilde{k}_{\mathcal{D}}(\bmv,\bmv) \pi(\bm{v})^2,
\end{equation}
which we optimize to sequentially select the next tangent vector for evaluation of the exponential map and thus $g$.

\subsection{Directional Cumulative Acquisition}
In our manifold setting, the acquisition function is defined on the tangent space.
Exponential maps yield intermediate steps on straight lines in the tangent space, as moving along a geodesic does not change the direction of the initial velocity, only the distance traveled. As a result, after solving $\expmap_{\bm{\mu}}(\alpha \cdot \bm{r})$ for a unit vector $\bm{r}$, we get evaluations of the integrand at other locations $\beta \bm{r},\ 0 < \beta < \alpha$, essentially for free.  
Because the \textsc{ivp} is solved already, only $g$ has to be evaluated at the respective location, which is cheap once the exponential map is computed.

This observation motivates rethinking the scheme for sequential design to select good initial directions instead of fixed velocities for the exponential map. We propose to select these initial directions such that the cumulative variance along the direction on the tangent space is maximized, and multiple points along this line with large marginal variance are then selected to reduce the overall variance. The modified acquisition function, i.e., the cumulative variance along a straight line, can be written as
\begin{align}
    \bar{u} (\bm{r}) = \int_0^\infty  u (\beta \bm{r})\ \d\beta.
    \label{eqn:dcv}
\end{align}
The new acquisition policy arises from optimizing $\bar{u}$ for unit tangent vectors $\bm{r}$.
We call the acquisition function from Eq.~\eqref{eqn:dcv} \textit{directional cumulative variance} (\textsc{dcv}). While it does have a closed-form solution, that solution costs $\mathcal{O}(M^4)$ to evaluate in the number $M$ of evaluations chosen for \bq. We resort to numerical integration to compute the objective and its gradient~(\ref{app:dcv}). This is feasible because these are multiple univariate integrals that can efficiently be estimated from the same evaluations. Since $\bm{r}$ is constrained to lie on the unit hypersphere, we employ a manifold optimization algorithm~(\ref{app:dcv}). Once an exponential map is computed, we use the standard \wsabi~objective to sample multiple informative points along the straight line $\alpha \cdot \bm{r}$. For simplicity, we use \textsc{dcv} only in conjunction with \wsabil.

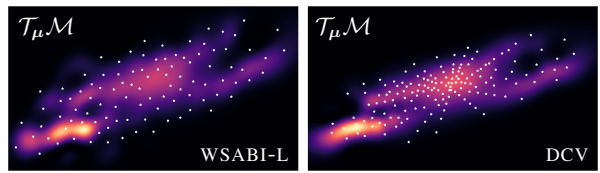
\begin{figure}[t]
    \centering
    \begin{overpic}[width=0.47\textwidth]{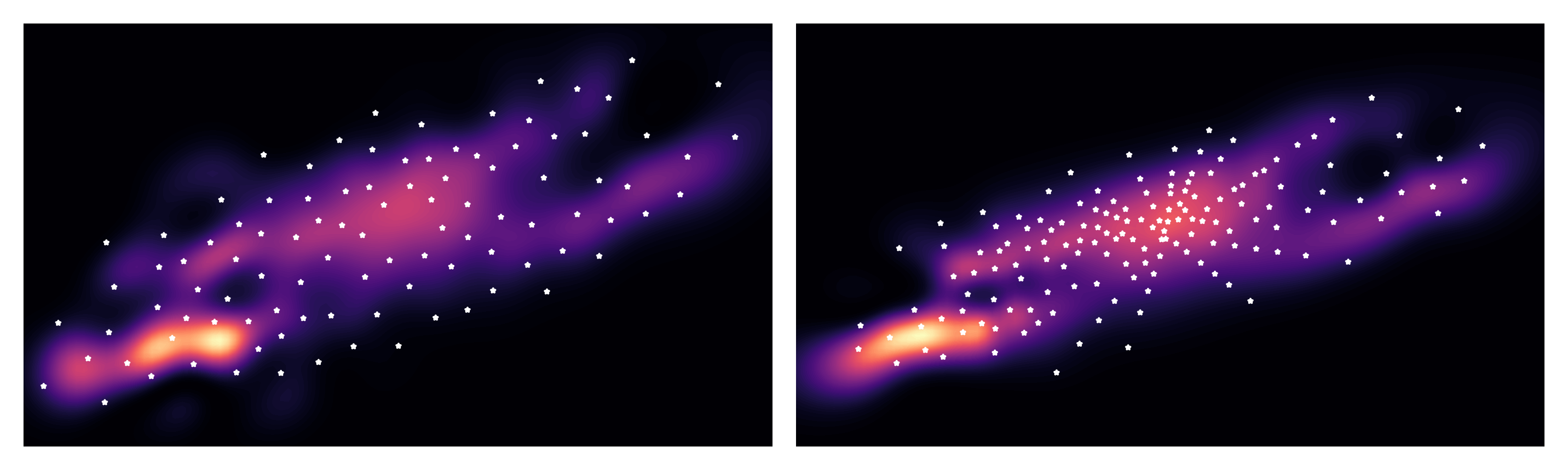}
 \put (3,24) {\small\color{white}$\displaystyle\mathcal{T}_{\bm{\mu}}\mathcal{M}$}
 \put (52,24) {\small\color{white}$\displaystyle\mathcal{T}_{\bm{\mu}}\mathcal{M}$}
 \put (33,3) {\small\color{white}$\displaystyle\textsc{wsabi-l}$} % 33
 \put (90,3) {\small\color{white}$\displaystyle\textsc{dcv}$} % 90
\end{overpic}
    \caption{Posterior over $g_{\bm{\mu}}(\bm{v})\mathcal{N}(\bm{v};\bm{0},\bm{\Sigma})$. We compare \textsc{wsabi-l} using uncertainty sampling (left) and \textsc{dcv} (right). Observed locations are scattered in white. The linear color scale represents the posterior magnitude. 
    }
    \label{fig:bq_tangent}
\end{figure}

Optimizing this acquisition function is costly as it requires posterior mean predictions and predictive gradients of the \textsc{gp} inside the integration. Furthermore, confining observations to lie along straight lines implies that \bq~may cover less space given a fixed number of function evaluations. Therefore, \textsc{dcv} will be useful in settings where exponential maps come at a high computational cost.  
Fig.~\ref{fig:bq_tangent} compares posteriors arising from standard \wsabil~using uncertainty sampling (simply referred to as \wsabil) and \textsc{dcv}. 

\subsection{Choice of Integrand Model}
The known smoothness of the metric 
%tensor
makes the square exponential kernel (\textsc{rbf}) a suitable choice in most cases. 
% However, for high-curvature manifolds, in particular in two dimensions, we found the Mat\'ern-5/2 kernel to be numerically slightly more stable, so we use it throughout instead of the \textsc{rbf}.
% However, for high-curvature manifolds, in particular in two dimensions, we found the Mat\'ern-5/2 kernel to perform slightly better, so we use it throughout instead of the \textsc{rbf}.
The Mat\'ern model makes less strong assumptions on differentiability than the \textsc{rbf} --- \citet[Sec.~1.7]{stein2012interpolation} %explicitly
recommends "use the Mat\'ern model"---and has proven more robust in high-curvature settings.
Depending on the employed Riemannian metric, we can set the constant prior mean of $g$ to the 
known volume element far from the data (Sec.~\ref{sec:metrics}). This amounts to the prior assumption that wherever there are no observations yet, the distance to the data is likely high. 

%% file: sections/05_exp.tex
\begin{figure*}[ht]
\centering     %%% not \center
\hspace{-1.2mm}
\subfigure[\textsc{circle}]{\scalebox{0.3}{\input{fig/boxplot_circle-1000_res.pgf}}}
\hspace{0.8mm}
\subfigure[\textsc{circle}~\oldstylenums{5}\textsc{d}]{\scalebox{0.3}{\input{fig/boxplot_circle-1000-5d_res.pgf}}}
\hspace{0.8mm}
\subfigure[\textsc{mnist}]{\scalebox{0.3}{\input{fig/boxplot_MNIST_res.pgf}}}
\hspace{0.8mm}
\subfigure[\textsc{adk}]{\scalebox{0.3}{\input{fig/boxplot_ADK_res.pgf}}}
\hfill
\caption{Boxplot error comparison (log scale, shared y-axis) of \bq~and \textsc{mc} on whole \textsc{land} fit for different manifolds. For \textsc{mc}, we allocate the runtime of the mean slowest \bq~method. Each box contains 16 independent runs.}
\label{fig:boxplots}
\end{figure*}
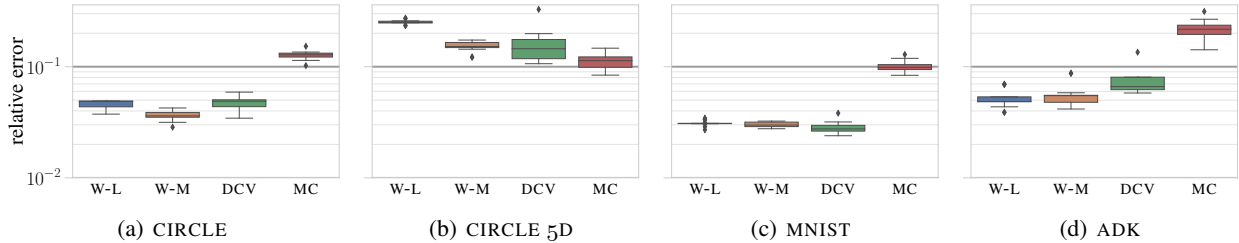  
\begin{figure}
    \centering
    \scalebox{0.52}{\input{fig/runtime_barplot_res.pgf}}
    \caption{Mean runtime comparison (for a single integration) of the \bq~methods. Errorbars indicate 95\% confidence intervals w.r.t the 16 runs on each \textsc{land} fit. 
    The reported runtimes belong to the boxplots in Fig.~\ref{fig:boxplots}.
    }
    \label{fig:runtime_barplot}
\end{figure}
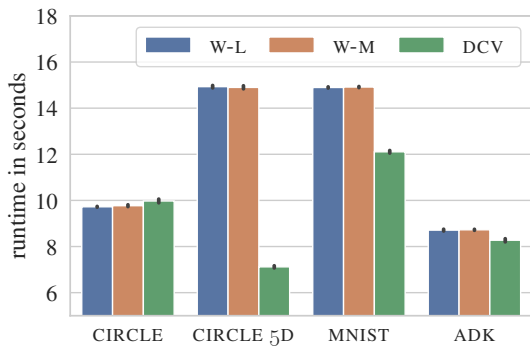

\subsection{Transfer Learning}
The \textsc{land} optimization process requires the sequential computation of one integral per iteration as the parameters of the model get updated in an alternating manner. 
In general, elaborate schemes as in \citet{Xi2018} and \citet{GessnerGM2019} are available to estimate correlated integrals jointly. 
However, our integrand $g_{\bm{\mu}}$ does not depend on the covariance $\bm{\Sigma}$ of the integration measure $\pi$. Therefore, exponential maps remain unaltered when only the covariance $\bm{\Sigma}$ changes from one iteration to the next while the mean $\bm{\mu}$ remains fixed. 
\bq~can thus reuse the observations from the previous iteration and only needs to collect a reduced number of new samples to account for the changed integration measure.
This node reuse enables tremendous runtime savings.

\section{Experiments}
\label{sec:experiments}
We test the methods (\wsabil, \wsabim, \textsc{dcv}) on both synthetic and real-world data manifolds. Our aim is to show that Bayesian quadrature is faster compared to the Monte Carlo baseline, yet retains high accuracy. The experiments focus on the \textsc{land} model to illustrate practical use cases of Riemannian statistics. Furthermore, the iterative optimization process yields a wide range of integration problems (Eq.~\ref{eqn:normalization_constant}) of varying difficulty. In total, our experiments comprise $43{,}920$ \bq~integrations.

For different manifolds, we conduct two kinds of experiments:
First, we fit the \textsc{land} model and record all integration problems arising during the optimization procedure (\ref{app:land}). This allows us to compare the competitors on the whole problem, where \bq~can benefit from node reuse. As the ground truth, we use extensive $\textsc{mc}$ sampling ($S=40{,}000$). We fix the number of acquired samples for \bq~and generate boxplots from the mean errors on the whole \textsc{land} fit for 16 independent runs (Fig.~\ref{fig:boxplots}). 
Due to the alternating update of \textsc{land} parameters during optimization, \textit{either} the integrand \textit{or} the integration measure changes over consecutive iterations. We let \wsabil~and \wsabim~actively collect $80$ in the former and $10$ samples additionally to the reused ones in the latter case; for \textsc{dcv}, we fix $18$ and $2$ exponential maps, respectively, and acquire $6$ points on each straight line. Integration cost for \bq~is thus highly variable over iterations. Allocating a fixed runtime would not be sensible as \bq~benefits from collecting more information after updates to the mean, a time investment that is over-compensated in the more abundant and---due to node reuse---cheap covariance updates. We choose sample numbers so as to allow for sufficient exploration of the space with practical runtime. For \textsc{mc}, we allocate the runtime budget of the mean slowest \bq~method on that particular problem in order to compare accuracy over runtime. Mean runtimes for single integrations, averaged on entire \textsc{land} fits, are shown in Fig.~\ref{fig:runtime_barplot} and mean exponential map runtimes, as computed by \textsc{mc}, are reported in~\ref{app:experiments}.\\

\begin{figure*}[ht]
\centering     
\hspace{-1.7mm}
\subfigure[\textsc{circle}]{\label{fig:conta}\scalebox{0.3}{\input{fig/continuous_circle-1000_res.pgf}}}
\hspace{1.2mm}
\subfigure[\textsc{circle}~\oldstylenums{5}\textsc{d}]{\label{fig:contb}\scalebox{0.3}{\input{fig/continuous_circle-1000-5d_res.pgf}}}
\hspace{1.2mm}
\subfigure[\textsc{mnist}]{\label{fig:contc}\scalebox{0.3}{\input{fig/continuous_MNIST_res.pgf}}}
\hspace{1.2mm}
\subfigure[\textsc{adk}]{\label{fig:contd}\scalebox{0.3}{\input{fig/continuous_adk_res.pgf}}}
\hfill
\caption{Comparison of \bq~and \textsc{mc}
errors against runtime (vertical log scale, shared legend and axes) for different manifolds, on the first integration problem of the respective \textsc{land} fit. Shaded regions indicate $95\%$ confidence intervals w.r.t. the $30$ independent runs (\ref{app:experiments}).
}
\label{fig:continuous_graph}
\end{figure*}
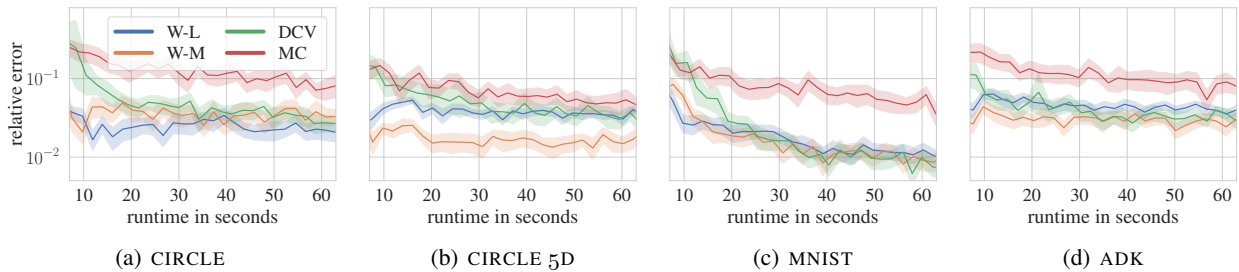

Secondly, we focus on the first integration problem of each \textsc{land} fit in detail and compare the convergence behavior of the different \bq~methods and \mc~over wall clock runtime (Fig.~\ref{fig:continuous_graph}). We use the kernel metric (Sec.~\ref{sec:metrics}) when not otherwise mentioned. All further technical details for reproducibility are in~\ref{app:experiments}. In the plot legends, we abbreviate \textsc{wsabi-l}/\textsc{wsabi-m} with \textsc{w-l}/\textsc{w-m}, respectively. Code is publicly available at \href{https://github.com/froec/BQonRDM}{github.com/froec/BQonRDM}.

\subsection{Synthetic Experiments}
\begin{figure}
    \centering
    \subfigure[\textsc{land}]{\label{fig:acircle}\includegraphics[width=0.17\textwidth]{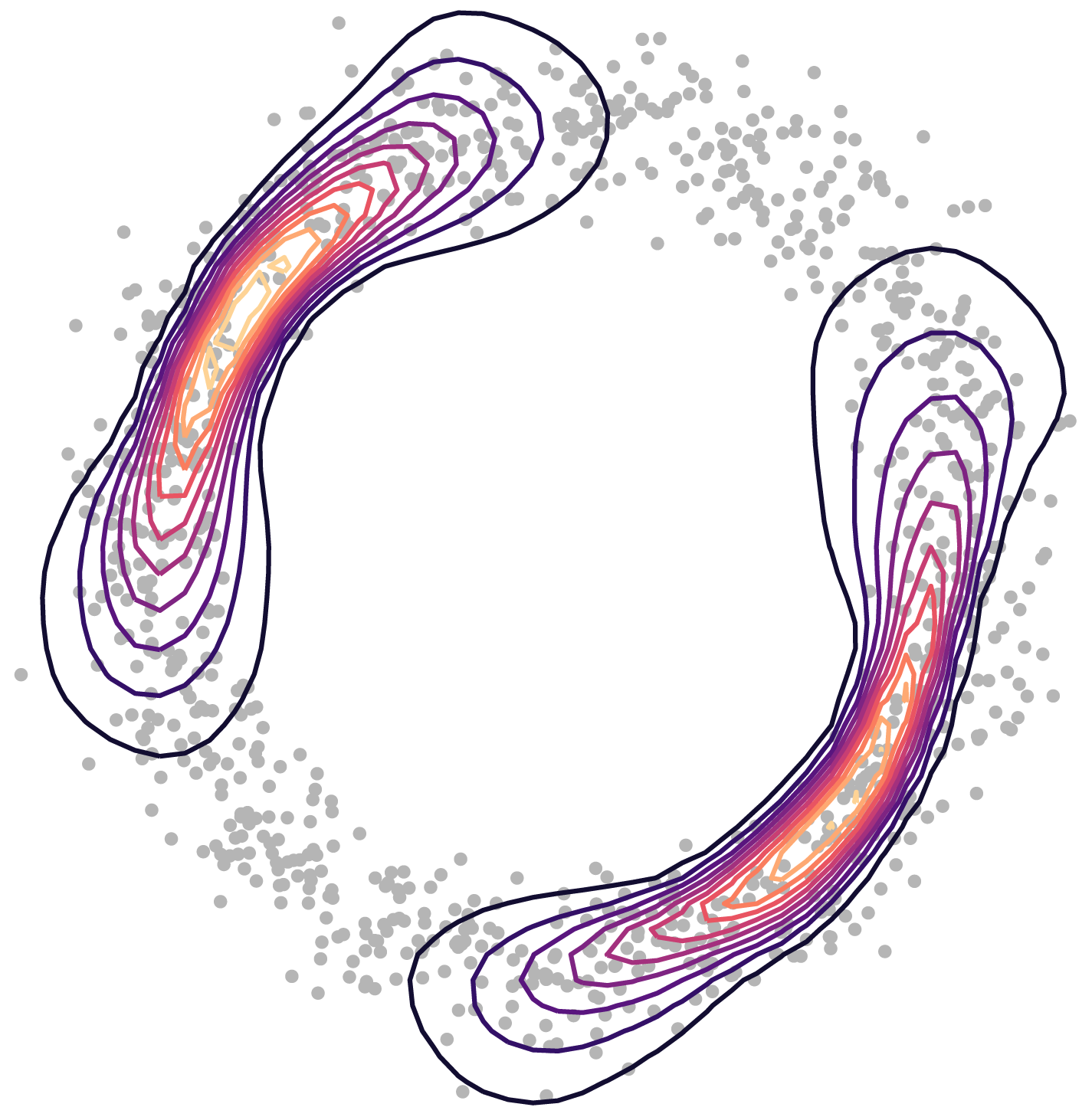}}
    \subfigure[\textsc{gmm}]{\label{fig:bcircle}\includegraphics[width=0.17\textwidth]{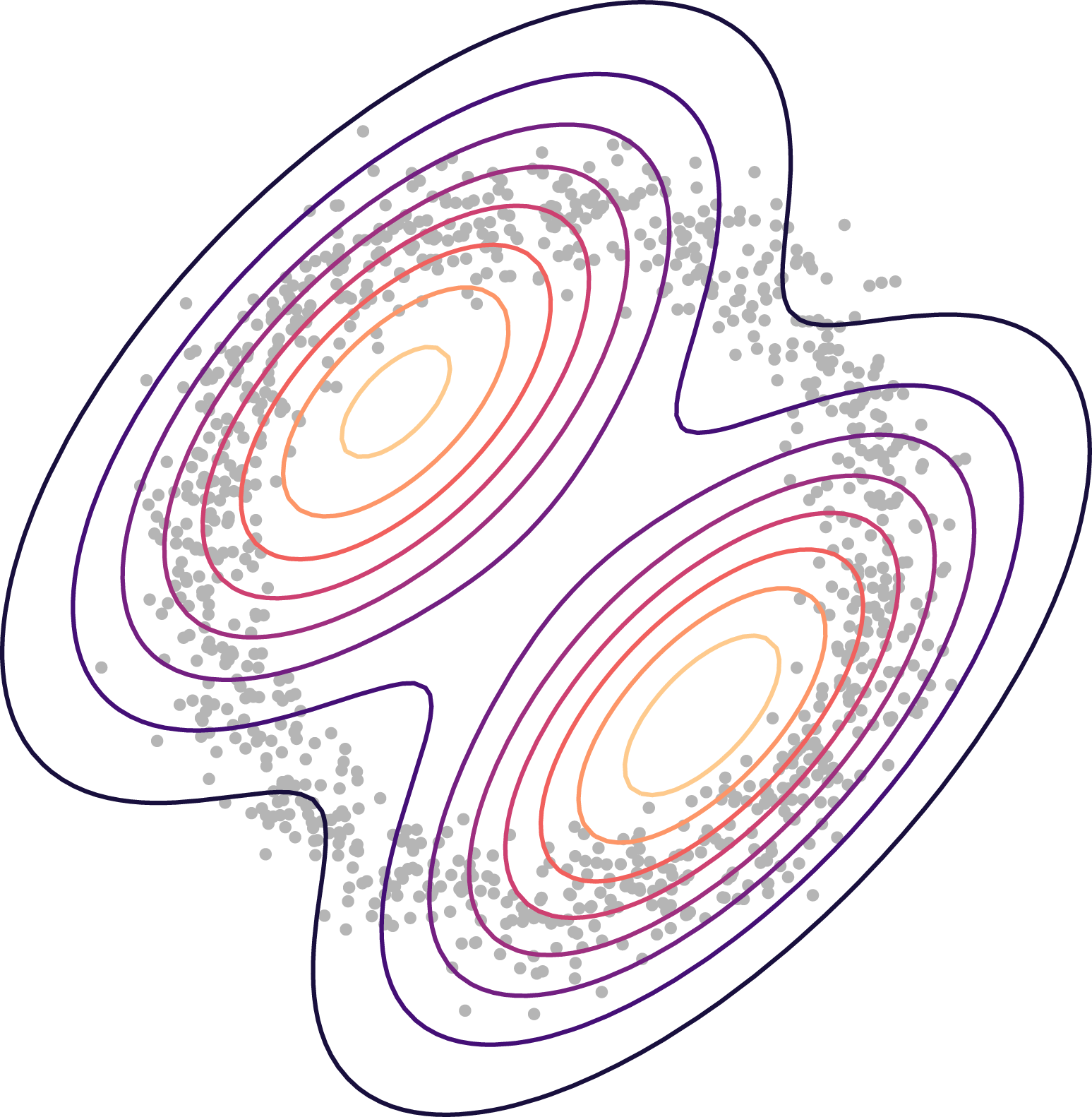}}
    \caption{Model comparison on \textsc{circle} toy data.}
    \label{fig:synth_land}
\end{figure}

\paragraph{Toy Data}
We generated three toy data sets and fitted the \textsc{land} model with pre-determined component numbers. Here we focus on a circle manifold with $1000$ data points, for which we compare the resulting \textsc{land} fit to the Euclidean Gaussian mixture model (\textsc{gmm}) in Fig.~\ref{fig:synth_land}, and report results for the other datasets in~\ref{app:experiments}.

\paragraph{Higher-dimensional Toy Data}
With increasing number of dimensions, new challenges for metric learning and geodesic solvers appear. With the simple kernel metric, almost all of the volume will be far from the data as the dimension increases, a phenomenon which we observe already in relatively low dimensions. Such metric behavior can lead to pathological integration problems, as the integrand may then become almost constant. In this experiment, we embed the circle toy data in higher dimensions by sampling random orthonormal matrices. After projecting the data, we add Gaussian noise $\epsilon_i \sim \N(0, 0.01)$ and standardize. Here we show the result for $d=5$ and report results for $d=\{3,4\}$ in~\ref{app:experiments}.

\subsection{Real-World Experiments}
\paragraph{\textsc{mnist}}
\begin{figure}
    \centering
    \subfigure[\textsc{land}]{\label{fig:mnista}\includegraphics[width=0.17\textwidth]{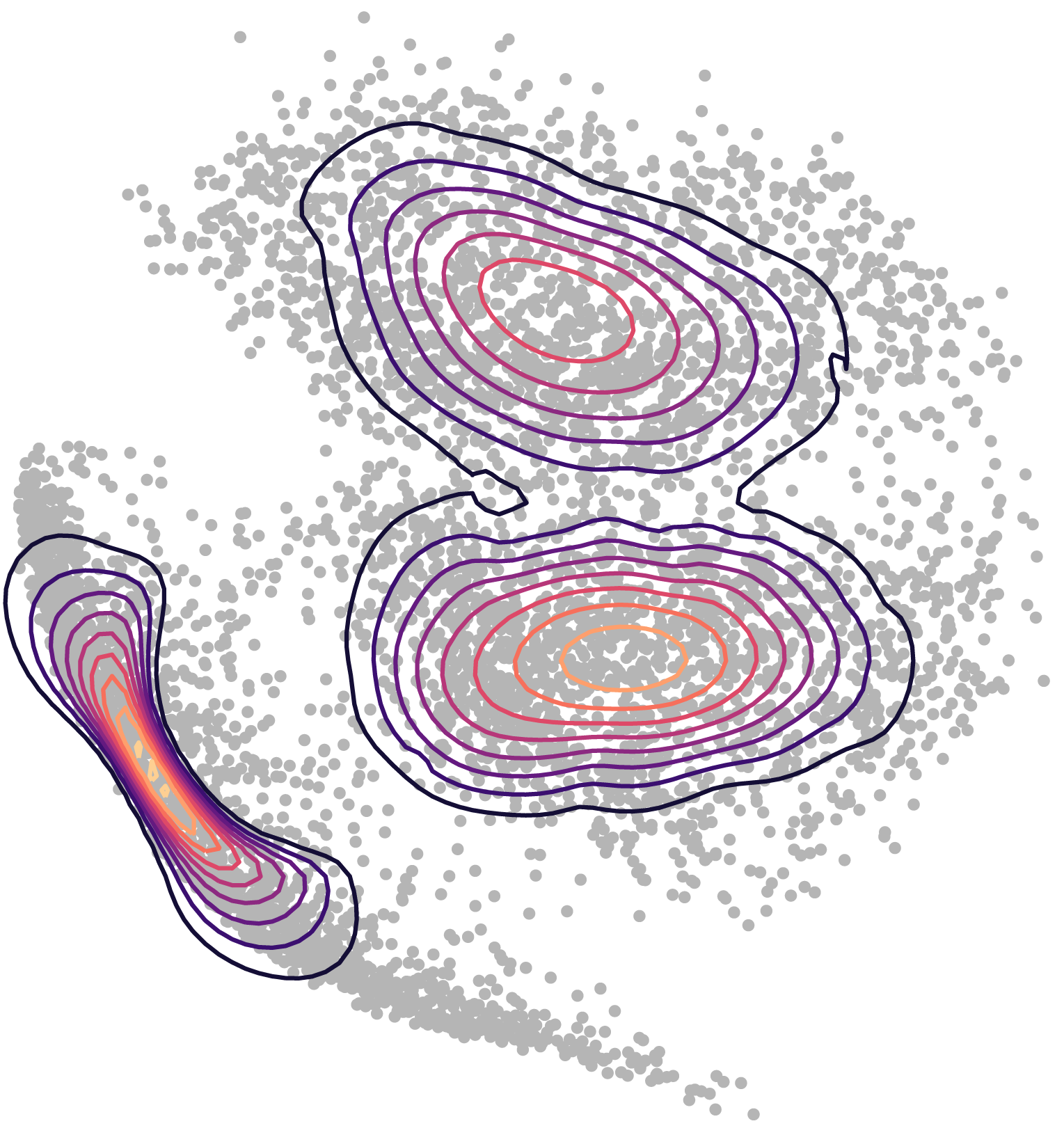}}
    \subfigure[\textsc{gmm}]{\label{fig:mnistb}\includegraphics[width=0.17\textwidth]{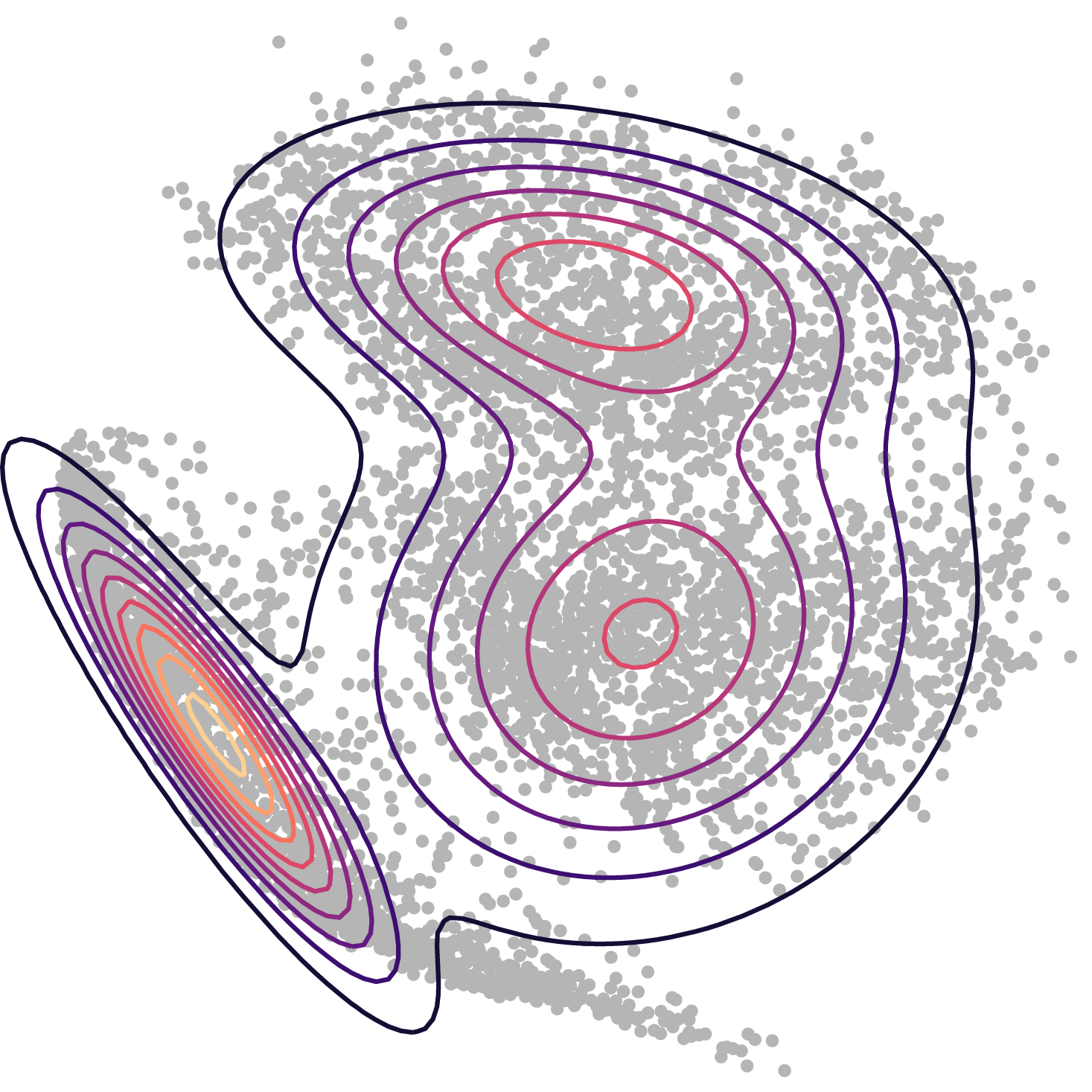}}
    \caption{Model comparison on three-digit \textsc{mnist}.}
    \label{fig:mnist_land}
\end{figure}
We trained a Variational Auto-Encoder on the first three digits of \textsc{mnist} using the aggregated posterior metric (Sec.~\ref{sec:metrics}) and found that the \textsc{land} is able to distinguish the three clusters more clearly than a Euclidean Gaussian mixture model, see Fig.~\ref{fig:mnist_land}. The \textsc{land} favors regions of higher density, where the \textsc{vae} has more training data. In this experiment, the gain in speed of \bq~is even more pronounced, since exponential maps are slow due to high curvature. \textsc{mc} with $1000$ samples achieves $2.78\%$ mean error on the whole \textsc{land} fit with a total runtime of $6$ hours and $56$ minutes, whereas \textsc{dcv} (18/2 exponential maps) achieves $2.84\%$ error within $21$ minutes; a speedup by a factor of $\approx 20$.

\paragraph{Molecular Dynamics}
In molecular dynamics, biophysical systems are simulated on the atomic level. This approach is useful to understand the conformational changes of a protein, i.e., the structural changes it undergoes. A Riemannian model is appropriate in this setting, because not all atom coordinates represent physically realistic conformations. For instance, a protein clearly does not self-intersect. Adapting locally to the data by space distortion is thus critical for modeling. More specifically, the \textsc{land} model is relevant because clustering conformations and finding representative states are of scientific interest (see e.g., \citealp{pca2009,wolf2013principal,spellmon2015molecular,dimredtraj}). The \textsc{land} can visualize the conformational landscape and generate realistic samples. Plausible interpolations (trajectories) between conformations may be conceived of as geodesics under the Riemannian metric.

We obtained multiple trajectories of the closed to open transition of the enzyme adenylate kinase (\textsc{adk}) \citep{adkdims}. Each observation consists of the Cartesian $(x,y,z)$ coordinates for each of the $3{,}341$ atoms, yielding a $10{,}023$ dimensional vector. As is common in the field, we used \textsc{pca} to extract the \textit{essential dynamics} \citep{amadei1993essential}, which clearly exhibit manifold structure (Fig.~\ref{fig:teaser}). The first two eigenvectors already explain $65\%$ of the total variance and suffice to capture the transition motion. We model the \textsc{adk} manifold with high curvature and large measure far from the data to account for the knowledge that realistic trajectories lie closely together. This makes for a challenging integration problem, since most mass is near the data boundary due to extreme metric values.

\begin{figure}
    \centering
    \scalebox{0.33}{\input{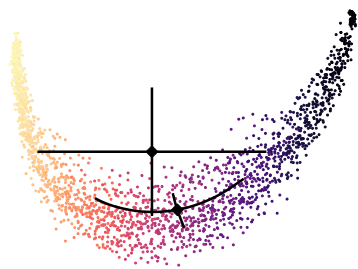}}
    \caption{Comparison of the Euclidean Gaussian vs. \textsc{land} mean and eigenvectors on \textsc{adk} data. Data is colored according to the \textit{radius of gyration}, a measure indicating how ``open'' the protein is, providing a visual argument for the manifold hypothesis.}
    \label{fig:adk_eigvecs}
\end{figure}

\begin{figure}
    \centering
    \includegraphics[width=0.39\textwidth]{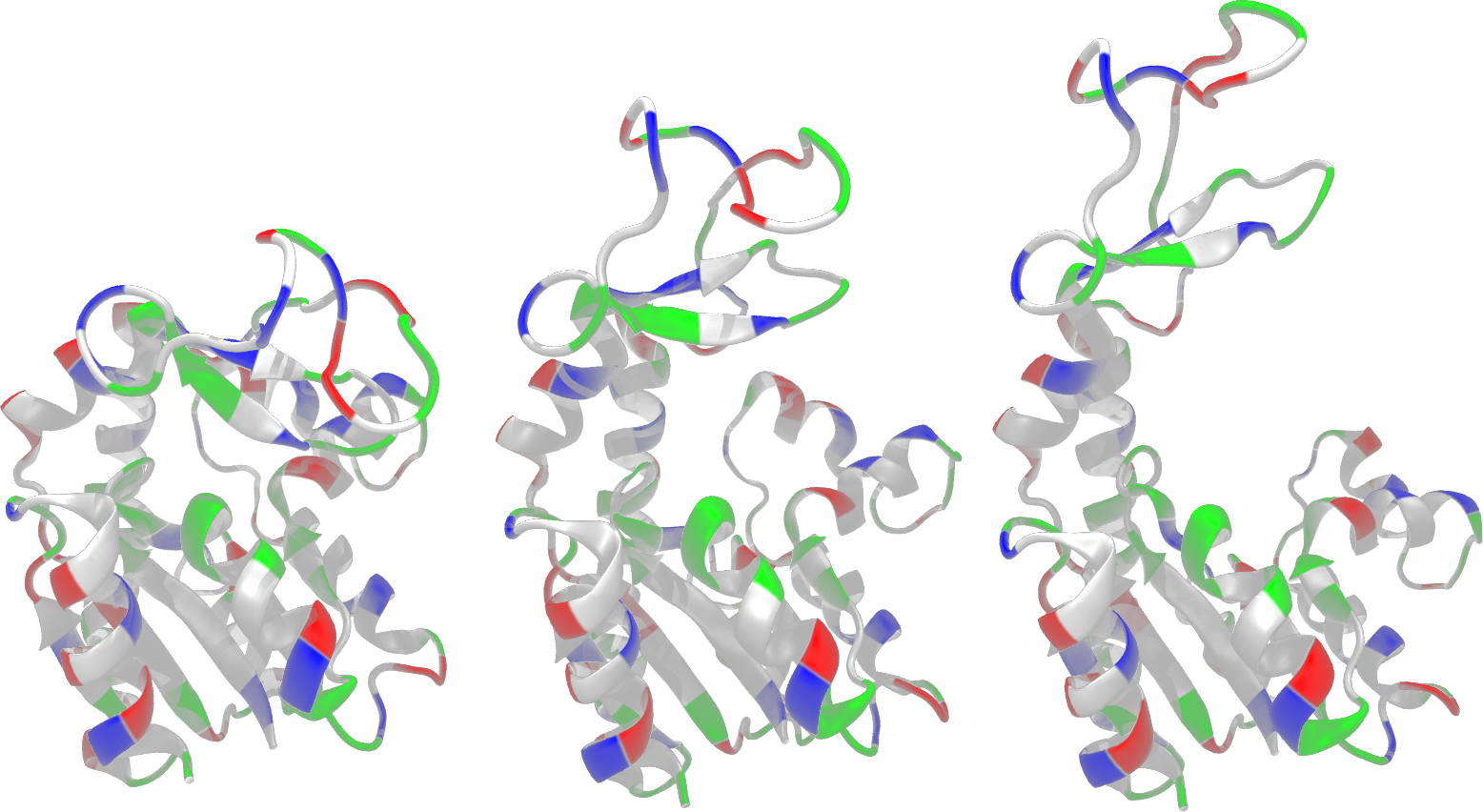}
    \caption{\textsc{adk} in closed, \textsc{land} mean and open state.}
    \label{fig:adks}
\end{figure}

A single-component \textsc{land} yields a representative state for the transition between the closed and open conformation. Whereas the Euclidean mean falls outside the data manifold, the \textsc{land} mean is reasonably situated. Plotting the eigenvectors of the covariance matrix makes it clear that the \textsc{land} captures the intrinsic dimensions of the data manifold (Fig.~\ref{fig:adk_eigvecs}) and that the mean interpolates between the closed and open state (Fig.~\ref{fig:adks}).
Our aim here is to demonstrate that molecular dynamics is an exciting application area for Riemannian statistics and sketch potential experiments, which are then for domain experts to design.

\subsection{Interpretation}
We find that \bq~consistently outperforms \textsc{mc} in terms of speed. Even on high-curvature manifolds with volume elements spanning multiple orders of magnitudes, such as \textsc{mnist} and \textsc{adk}, the \textsc{gp} succeeds to approximate the integrand well. Among the different \bq~candidates, we cannot discern a clear winner, since their performance depends on the specific problem geometry and exponential map runtimes. \textsc{dcv} performs especially well when geodesic computations are costly, such as for \textsc{mnist}. We note that geodesic solvers and metric learning are subject to new challenges in higher dimensions, which merit further research effort.

%% file: fig/boxplot_circle-1000_res.pgf
%% Creator: Matplotlib, PGF backend
%%
%% To include the figure in your LaTeX document, write
%%   \input{<filename>.pgf}
%%
%% Make sure the required packages are loaded in your preamble
%%   \usepackage{pgf}
%%
%% Figures using additional raster images can only be included by \input if
%% they are in the same directory as the main LaTeX file. For loading figures
%% from other directories you can use the `import` package
%% and then include the figures with
%%   \import{<path to file>}{<filename>.pgf}
%%
%% Matplotlib used the following preamble
%%
\begingroup%
\makeatletter%
\begin{pgfpicture}%
\pgfpathrectangle{\pgfpointorigin}{\pgfqpoint{5.920874in}{3.524405in}}%
\pgfusepath{use as bounding box, clip}%
\begin{pgfscope}%
\pgfsetbuttcap%
\pgfsetmiterjoin%
\definecolor{currentfill}{rgb}{1.000000,1.000000,1.000000}%
\pgfsetfillcolor{currentfill}%
\pgfsetlinewidth{0.000000pt}%
\definecolor{currentstroke}{rgb}{1.000000,1.000000,1.000000}%
\pgfsetstrokecolor{currentstroke}%
\pgfsetdash{}{0pt}%
\pgfpathmoveto{\pgfqpoint{0.000000in}{0.000000in}}%
\pgfpathlineto{\pgfqpoint{5.920874in}{0.000000in}}%
\pgfpathlineto{\pgfqpoint{5.920874in}{3.524405in}}%
\pgfpathlineto{\pgfqpoint{0.000000in}{3.524405in}}%
\pgfpathclose%
\pgfusepath{fill}%
\end{pgfscope}%
\begin{pgfscope}%
\pgfsetbuttcap%
\pgfsetmiterjoin%
\definecolor{currentfill}{rgb}{1.000000,1.000000,1.000000}%
\pgfsetfillcolor{currentfill}%
\pgfsetlinewidth{0.000000pt}%
\definecolor{currentstroke}{rgb}{0.000000,0.000000,0.000000}%
\pgfsetstrokecolor{currentstroke}%
\pgfsetstrokeopacity{0.000000}%
\pgfsetdash{}{0pt}%
\pgfpathmoveto{\pgfqpoint{1.220874in}{0.454405in}}%
\pgfpathlineto{\pgfqpoint{5.870874in}{0.454405in}}%
\pgfpathlineto{\pgfqpoint{5.870874in}{3.474405in}}%
\pgfpathlineto{\pgfqpoint{1.220874in}{3.474405in}}%
\pgfpathclose%
\pgfusepath{fill}%
\end{pgfscope}%
\begin{pgfscope}%
\definecolor{textcolor}{rgb}{0.000000,0.000000,0.000000}%
\pgfsetstrokecolor{textcolor}%
\pgfsetfillcolor{textcolor}%
\pgftext[x=1.802124in,y=0.357183in,,top]{\color{textcolor}\rmfamily\fontsize{24.200000}{29.040000}\selectfont \textsc{w-l}}%
\end{pgfscope}%
\begin{pgfscope}%
\definecolor{textcolor}{rgb}{0.000000,0.000000,0.000000}%
\pgfsetstrokecolor{textcolor}%
\pgfsetfillcolor{textcolor}%
\pgftext[x=2.964624in,y=0.357183in,,top]{\color{textcolor}\rmfamily\fontsize{24.200000}{29.040000}\selectfont \textsc{w-m}}%
\end{pgfscope}%
\begin{pgfscope}%
\definecolor{textcolor}{rgb}{0.000000,0.000000,0.000000}%
\pgfsetstrokecolor{textcolor}%
\pgfsetfillcolor{textcolor}%
\pgftext[x=4.127124in,y=0.357183in,,top]{\color{textcolor}\rmfamily\fontsize{24.200000}{29.040000}\selectfont \textsc{dcv}}%
\end{pgfscope}%
\begin{pgfscope}%
\definecolor{textcolor}{rgb}{0.000000,0.000000,0.000000}%
\pgfsetstrokecolor{textcolor}%
\pgfsetfillcolor{textcolor}%
\pgftext[x=5.289624in,y=0.357183in,,top]{\color{textcolor}\rmfamily\fontsize{24.200000}{29.040000}\selectfont \textsc{mc}}%
\end{pgfscope}%
\begin{pgfscope}%
\pgfpathrectangle{\pgfqpoint{1.220874in}{0.454405in}}{\pgfqpoint{4.650000in}{3.020000in}}%
\pgfusepath{clip}%
\pgfsetrectcap%
\pgfsetroundjoin%
\pgfsetlinewidth{2.509375pt}%
\definecolor{currentstroke}{rgb}{0.800000,0.800000,0.800000}%
\pgfsetstrokecolor{currentstroke}%
\pgfsetdash{}{0pt}%
\pgfpathmoveto{\pgfqpoint{1.220874in}{0.454405in}}%
\pgfpathlineto{\pgfqpoint{5.870874in}{0.454405in}}%
\pgfusepath{stroke}%
\end{pgfscope}%
\begin{pgfscope}%
\pgfsetbuttcap%
\pgfsetroundjoin%
\definecolor{currentfill}{rgb}{0.800000,0.800000,0.800000}%
\pgfsetfillcolor{currentfill}%
\pgfsetlinewidth{1.003750pt}%
\definecolor{currentstroke}{rgb}{0.800000,0.800000,0.800000}%
\pgfsetstrokecolor{currentstroke}%
\pgfsetdash{}{0pt}%
\pgfsys@defobject{currentmarker}{\pgfqpoint{-0.138889in}{0.000000in}}{\pgfqpoint{0.000000in}{0.000000in}}{%
\pgfpathmoveto{\pgfqpoint{0.000000in}{0.000000in}}%
\pgfpathlineto{\pgfqpoint{-0.138889in}{0.000000in}}%
\pgfusepath{stroke,fill}%
}%
\begin{pgfscope}%
\pgfsys@transformshift{1.220874in}{0.454405in}%
\pgfsys@useobject{currentmarker}{}%
\end{pgfscope}%
\end{pgfscope}%
\begin{pgfscope}%
\definecolor{textcolor}{rgb}{0.000000,0.000000,0.000000}%
\pgfsetstrokecolor{textcolor}%
\pgfsetfillcolor{textcolor}%
\pgftext[x=0.412738in,y=0.334420in,left,base]{\color{textcolor}\rmfamily\fontsize{24.200000}{29.040000}\selectfont \(\displaystyle 10^{-2}\)}%
\end{pgfscope}%
\begin{pgfscope}%
\pgfpathrectangle{\pgfqpoint{1.220874in}{0.454405in}}{\pgfqpoint{4.650000in}{3.020000in}}%
\pgfusepath{clip}%
\pgfsetrectcap%
\pgfsetroundjoin%
\pgfsetlinewidth{2.509375pt}%
\definecolor{currentstroke}{rgb}{0.611765,0.611765,0.611765}%
\pgfsetstrokecolor{currentstroke}%
\pgfsetdash{}{0pt}%
\pgfpathmoveto{\pgfqpoint{1.220874in}{2.394902in}}%
\pgfpathlineto{\pgfqpoint{5.870874in}{2.394902in}}%
\pgfusepath{stroke}%
\end{pgfscope}%
\begin{pgfscope}%
\pgfsetbuttcap%
\pgfsetroundjoin%
\definecolor{currentfill}{rgb}{0.800000,0.800000,0.800000}%
\pgfsetfillcolor{currentfill}%
\pgfsetlinewidth{1.003750pt}%
\definecolor{currentstroke}{rgb}{0.800000,0.800000,0.800000}%
\pgfsetstrokecolor{currentstroke}%
\pgfsetdash{}{0pt}%
\pgfsys@defobject{currentmarker}{\pgfqpoint{-0.138889in}{0.000000in}}{\pgfqpoint{0.000000in}{0.000000in}}{%
\pgfpathmoveto{\pgfqpoint{0.000000in}{0.000000in}}%
\pgfpathlineto{\pgfqpoint{-0.138889in}{0.000000in}}%
\pgfusepath{stroke,fill}%
}%
\begin{pgfscope}%
\pgfsys@transformshift{1.220874in}{2.394902in}%
\pgfsys@useobject{currentmarker}{}%
\end{pgfscope}%
\end{pgfscope}%
\begin{pgfscope}%
\definecolor{textcolor}{rgb}{0.000000,0.000000,0.000000}%
\pgfsetstrokecolor{textcolor}%
\pgfsetfillcolor{textcolor}%
\pgftext[x=0.412738in,y=2.274917in,left,base]{\color{textcolor}\rmfamily\fontsize{24.200000}{29.040000}\selectfont \(\displaystyle 10^{-1}\)}%
\end{pgfscope}%
\begin{pgfscope}%
\pgfpathrectangle{\pgfqpoint{1.220874in}{0.454405in}}{\pgfqpoint{4.650000in}{3.020000in}}%
\pgfusepath{clip}%
\pgfsetrectcap%
\pgfsetroundjoin%
\pgfsetlinewidth{0.401500pt}%
\definecolor{currentstroke}{rgb}{0.800000,0.800000,0.800000}%
\pgfsetstrokecolor{currentstroke}%
\pgfsetstrokeopacity{0.550000}%
\pgfsetdash{}{0pt}%
\pgfpathmoveto{\pgfqpoint{1.220874in}{1.038553in}}%
\pgfpathlineto{\pgfqpoint{5.870874in}{1.038553in}}%
\pgfusepath{stroke}%
\end{pgfscope}%
\begin{pgfscope}%
\pgfpathrectangle{\pgfqpoint{1.220874in}{0.454405in}}{\pgfqpoint{4.650000in}{3.020000in}}%
\pgfusepath{clip}%
\pgfsetrectcap%
\pgfsetroundjoin%
\pgfsetlinewidth{0.401500pt}%
\definecolor{currentstroke}{rgb}{0.800000,0.800000,0.800000}%
\pgfsetstrokecolor{currentstroke}%
\pgfsetstrokeopacity{0.550000}%
\pgfsetdash{}{0pt}%
\pgfpathmoveto{\pgfqpoint{1.220874in}{1.380257in}}%
\pgfpathlineto{\pgfqpoint{5.870874in}{1.380257in}}%
\pgfusepath{stroke}%
\end{pgfscope}%
\begin{pgfscope}%
\pgfpathrectangle{\pgfqpoint{1.220874in}{0.454405in}}{\pgfqpoint{4.650000in}{3.020000in}}%
\pgfusepath{clip}%
\pgfsetrectcap%
\pgfsetroundjoin%
\pgfsetlinewidth{0.401500pt}%
\definecolor{currentstroke}{rgb}{0.800000,0.800000,0.800000}%
\pgfsetstrokecolor{currentstroke}%
\pgfsetstrokeopacity{0.550000}%
\pgfsetdash{}{0pt}%
\pgfpathmoveto{\pgfqpoint{1.220874in}{1.622700in}}%
\pgfpathlineto{\pgfqpoint{5.870874in}{1.622700in}}%
\pgfusepath{stroke}%
\end{pgfscope}%
\begin{pgfscope}%
\pgfpathrectangle{\pgfqpoint{1.220874in}{0.454405in}}{\pgfqpoint{4.650000in}{3.020000in}}%
\pgfusepath{clip}%
\pgfsetrectcap%
\pgfsetroundjoin%
\pgfsetlinewidth{0.401500pt}%
\definecolor{currentstroke}{rgb}{0.800000,0.800000,0.800000}%
\pgfsetstrokecolor{currentstroke}%
\pgfsetstrokeopacity{0.550000}%
\pgfsetdash{}{0pt}%
\pgfpathmoveto{\pgfqpoint{1.220874in}{1.810754in}}%
\pgfpathlineto{\pgfqpoint{5.870874in}{1.810754in}}%
\pgfusepath{stroke}%
\end{pgfscope}%
\begin{pgfscope}%
\pgfpathrectangle{\pgfqpoint{1.220874in}{0.454405in}}{\pgfqpoint{4.650000in}{3.020000in}}%
\pgfusepath{clip}%
\pgfsetrectcap%
\pgfsetroundjoin%
\pgfsetlinewidth{0.401500pt}%
\definecolor{currentstroke}{rgb}{0.800000,0.800000,0.800000}%
\pgfsetstrokecolor{currentstroke}%
\pgfsetstrokeopacity{0.550000}%
\pgfsetdash{}{0pt}%
\pgfpathmoveto{\pgfqpoint{1.220874in}{1.964405in}}%
\pgfpathlineto{\pgfqpoint{5.870874in}{1.964405in}}%
\pgfusepath{stroke}%
\end{pgfscope}%
\begin{pgfscope}%
\pgfpathrectangle{\pgfqpoint{1.220874in}{0.454405in}}{\pgfqpoint{4.650000in}{3.020000in}}%
\pgfusepath{clip}%
\pgfsetrectcap%
\pgfsetroundjoin%
\pgfsetlinewidth{0.401500pt}%
\definecolor{currentstroke}{rgb}{0.800000,0.800000,0.800000}%
\pgfsetstrokecolor{currentstroke}%
\pgfsetstrokeopacity{0.550000}%
\pgfsetdash{}{0pt}%
\pgfpathmoveto{\pgfqpoint{1.220874in}{2.094315in}}%
\pgfpathlineto{\pgfqpoint{5.870874in}{2.094315in}}%
\pgfusepath{stroke}%
\end{pgfscope}%
\begin{pgfscope}%
\pgfpathrectangle{\pgfqpoint{1.220874in}{0.454405in}}{\pgfqpoint{4.650000in}{3.020000in}}%
\pgfusepath{clip}%
\pgfsetrectcap%
\pgfsetroundjoin%
\pgfsetlinewidth{0.401500pt}%
\definecolor{currentstroke}{rgb}{0.800000,0.800000,0.800000}%
\pgfsetstrokecolor{currentstroke}%
\pgfsetstrokeopacity{0.550000}%
\pgfsetdash{}{0pt}%
\pgfpathmoveto{\pgfqpoint{1.220874in}{2.206848in}}%
\pgfpathlineto{\pgfqpoint{5.870874in}{2.206848in}}%
\pgfusepath{stroke}%
\end{pgfscope}%
\begin{pgfscope}%
\pgfpathrectangle{\pgfqpoint{1.220874in}{0.454405in}}{\pgfqpoint{4.650000in}{3.020000in}}%
\pgfusepath{clip}%
\pgfsetrectcap%
\pgfsetroundjoin%
\pgfsetlinewidth{0.401500pt}%
\definecolor{currentstroke}{rgb}{0.800000,0.800000,0.800000}%
\pgfsetstrokecolor{currentstroke}%
\pgfsetstrokeopacity{0.550000}%
\pgfsetdash{}{0pt}%
\pgfpathmoveto{\pgfqpoint{1.220874in}{2.306110in}}%
\pgfpathlineto{\pgfqpoint{5.870874in}{2.306110in}}%
\pgfusepath{stroke}%
\end{pgfscope}%
\begin{pgfscope}%
\pgfpathrectangle{\pgfqpoint{1.220874in}{0.454405in}}{\pgfqpoint{4.650000in}{3.020000in}}%
\pgfusepath{clip}%
\pgfsetrectcap%
\pgfsetroundjoin%
\pgfsetlinewidth{0.401500pt}%
\definecolor{currentstroke}{rgb}{0.800000,0.800000,0.800000}%
\pgfsetstrokecolor{currentstroke}%
\pgfsetstrokeopacity{0.550000}%
\pgfsetdash{}{0pt}%
\pgfpathmoveto{\pgfqpoint{1.220874in}{2.979050in}}%
\pgfpathlineto{\pgfqpoint{5.870874in}{2.979050in}}%
\pgfusepath{stroke}%
\end{pgfscope}%
\begin{pgfscope}%
\pgfpathrectangle{\pgfqpoint{1.220874in}{0.454405in}}{\pgfqpoint{4.650000in}{3.020000in}}%
\pgfusepath{clip}%
\pgfsetrectcap%
\pgfsetroundjoin%
\pgfsetlinewidth{0.401500pt}%
\definecolor{currentstroke}{rgb}{0.800000,0.800000,0.800000}%
\pgfsetstrokecolor{currentstroke}%
\pgfsetstrokeopacity{0.550000}%
\pgfsetdash{}{0pt}%
\pgfpathmoveto{\pgfqpoint{1.220874in}{3.320754in}}%
\pgfpathlineto{\pgfqpoint{5.870874in}{3.320754in}}%
\pgfusepath{stroke}%
\end{pgfscope}%
\begin{pgfscope}%
\definecolor{textcolor}{rgb}{0.000000,0.000000,0.000000}%
\pgfsetstrokecolor{textcolor}%
\pgfsetfillcolor{textcolor}%
\pgftext[x=0.357183in,y=1.964405in,,bottom,rotate=90.000000]{\color{textcolor}\rmfamily\fontsize{26.400000}{31.680000}\selectfont relative error}%
\end{pgfscope}%
\begin{pgfscope}%
\pgfpathrectangle{\pgfqpoint{1.220874in}{0.454405in}}{\pgfqpoint{4.650000in}{3.020000in}}%
\pgfusepath{clip}%
\pgfsetbuttcap%
\pgfsetmiterjoin%
\definecolor{currentfill}{rgb}{0.347059,0.458824,0.641176}%
\pgfsetfillcolor{currentfill}%
\pgfsetlinewidth{0.803000pt}%
\definecolor{currentstroke}{rgb}{0.298039,0.298039,0.298039}%
\pgfsetstrokecolor{currentstroke}%
\pgfsetdash{}{0pt}%
\pgfpathmoveto{\pgfqpoint{1.337124in}{1.694483in}}%
\pgfpathlineto{\pgfqpoint{2.267124in}{1.694483in}}%
\pgfpathlineto{\pgfqpoint{2.267124in}{1.795545in}}%
\pgfpathlineto{\pgfqpoint{1.337124in}{1.795545in}}%
\pgfpathlineto{\pgfqpoint{1.337124in}{1.694483in}}%
\pgfpathclose%
\pgfusepath{stroke,fill}%
\end{pgfscope}%
\begin{pgfscope}%
\pgfpathrectangle{\pgfqpoint{1.220874in}{0.454405in}}{\pgfqpoint{4.650000in}{3.020000in}}%
\pgfusepath{clip}%
\pgfsetbuttcap%
\pgfsetmiterjoin%
\definecolor{currentfill}{rgb}{0.798529,0.536765,0.389706}%
\pgfsetfillcolor{currentfill}%
\pgfsetlinewidth{0.803000pt}%
\definecolor{currentstroke}{rgb}{0.298039,0.298039,0.298039}%
\pgfsetstrokecolor{currentstroke}%
\pgfsetdash{}{0pt}%
\pgfpathmoveto{\pgfqpoint{2.499624in}{1.512773in}}%
\pgfpathlineto{\pgfqpoint{3.429624in}{1.512773in}}%
\pgfpathlineto{\pgfqpoint{3.429624in}{1.596679in}}%
\pgfpathlineto{\pgfqpoint{2.499624in}{1.596679in}}%
\pgfpathlineto{\pgfqpoint{2.499624in}{1.512773in}}%
\pgfpathclose%
\pgfusepath{stroke,fill}%
\end{pgfscope}%
\begin{pgfscope}%
\pgfpathrectangle{\pgfqpoint{1.220874in}{0.454405in}}{\pgfqpoint{4.650000in}{3.020000in}}%
\pgfusepath{clip}%
\pgfsetbuttcap%
\pgfsetmiterjoin%
\definecolor{currentfill}{rgb}{0.374020,0.618137,0.429902}%
\pgfsetfillcolor{currentfill}%
\pgfsetlinewidth{0.803000pt}%
\definecolor{currentstroke}{rgb}{0.298039,0.298039,0.298039}%
\pgfsetstrokecolor{currentstroke}%
\pgfsetdash{}{0pt}%
\pgfpathmoveto{\pgfqpoint{3.662124in}{1.697232in}}%
\pgfpathlineto{\pgfqpoint{4.592124in}{1.697232in}}%
\pgfpathlineto{\pgfqpoint{4.592124in}{1.816955in}}%
\pgfpathlineto{\pgfqpoint{3.662124in}{1.816955in}}%
\pgfpathlineto{\pgfqpoint{3.662124in}{1.697232in}}%
\pgfpathclose%
\pgfusepath{stroke,fill}%
\end{pgfscope}%
\begin{pgfscope}%
\pgfpathrectangle{\pgfqpoint{1.220874in}{0.454405in}}{\pgfqpoint{4.650000in}{3.020000in}}%
\pgfusepath{clip}%
\pgfsetbuttcap%
\pgfsetmiterjoin%
\definecolor{currentfill}{rgb}{0.710784,0.363725,0.375490}%
\pgfsetfillcolor{currentfill}%
\pgfsetlinewidth{0.803000pt}%
\definecolor{currentstroke}{rgb}{0.298039,0.298039,0.298039}%
\pgfsetstrokecolor{currentstroke}%
\pgfsetdash{}{0pt}%
\pgfpathmoveto{\pgfqpoint{4.824624in}{2.563558in}}%
\pgfpathlineto{\pgfqpoint{5.754624in}{2.563558in}}%
\pgfpathlineto{\pgfqpoint{5.754624in}{2.631782in}}%
\pgfpathlineto{\pgfqpoint{4.824624in}{2.631782in}}%
\pgfpathlineto{\pgfqpoint{4.824624in}{2.563558in}}%
\pgfpathclose%
\pgfusepath{stroke,fill}%
\end{pgfscope}%
\begin{pgfscope}%
\pgfsetrectcap%
\pgfsetmiterjoin%
\pgfsetlinewidth{1.254687pt}%
\definecolor{currentstroke}{rgb}{0.800000,0.800000,0.800000}%
\pgfsetstrokecolor{currentstroke}%
\pgfsetdash{}{0pt}%
\pgfpathmoveto{\pgfqpoint{1.220874in}{0.454405in}}%
\pgfpathlineto{\pgfqpoint{1.220874in}{3.474405in}}%
\pgfusepath{stroke}%
\end{pgfscope}%
\begin{pgfscope}%
\pgfsetrectcap%
\pgfsetmiterjoin%
\pgfsetlinewidth{1.254687pt}%
\definecolor{currentstroke}{rgb}{0.800000,0.800000,0.800000}%
\pgfsetstrokecolor{currentstroke}%
\pgfsetdash{}{0pt}%
\pgfpathmoveto{\pgfqpoint{5.870874in}{0.454405in}}%
\pgfpathlineto{\pgfqpoint{5.870874in}{3.474405in}}%
\pgfusepath{stroke}%
\end{pgfscope}%
\begin{pgfscope}%
\pgfsetrectcap%
\pgfsetmiterjoin%
\pgfsetlinewidth{1.254687pt}%
\definecolor{currentstroke}{rgb}{0.800000,0.800000,0.800000}%
\pgfsetstrokecolor{currentstroke}%
\pgfsetdash{}{0pt}%
\pgfpathmoveto{\pgfqpoint{1.220874in}{0.454405in}}%
\pgfpathlineto{\pgfqpoint{5.870874in}{0.454405in}}%
\pgfusepath{stroke}%
\end{pgfscope}%
\begin{pgfscope}%
\pgfsetrectcap%
\pgfsetmiterjoin%
\pgfsetlinewidth{1.254687pt}%
\definecolor{currentstroke}{rgb}{0.800000,0.800000,0.800000}%
\pgfsetstrokecolor{currentstroke}%
\pgfsetdash{}{0pt}%
\pgfpathmoveto{\pgfqpoint{1.220874in}{3.474405in}}%
\pgfpathlineto{\pgfqpoint{5.870874in}{3.474405in}}%
\pgfusepath{stroke}%
\end{pgfscope}%
\begin{pgfscope}%
\pgfpathrectangle{\pgfqpoint{1.220874in}{0.454405in}}{\pgfqpoint{4.650000in}{3.020000in}}%
\pgfusepath{clip}%
\pgfsetrectcap%
\pgfsetroundjoin%
\pgfsetlinewidth{0.803000pt}%
\definecolor{currentstroke}{rgb}{0.298039,0.298039,0.298039}%
\pgfsetstrokecolor{currentstroke}%
\pgfsetdash{}{0pt}%
\pgfpathmoveto{\pgfqpoint{1.802124in}{1.694483in}}%
\pgfpathlineto{\pgfqpoint{1.802124in}{1.566892in}}%
\pgfusepath{stroke}%
\end{pgfscope}%
\begin{pgfscope}%
\pgfpathrectangle{\pgfqpoint{1.220874in}{0.454405in}}{\pgfqpoint{4.650000in}{3.020000in}}%
\pgfusepath{clip}%
\pgfsetrectcap%
\pgfsetroundjoin%
\pgfsetlinewidth{0.803000pt}%
\definecolor{currentstroke}{rgb}{0.298039,0.298039,0.298039}%
\pgfsetstrokecolor{currentstroke}%
\pgfsetdash{}{0pt}%
\pgfpathmoveto{\pgfqpoint{1.802124in}{1.795545in}}%
\pgfpathlineto{\pgfqpoint{1.802124in}{1.795545in}}%
\pgfusepath{stroke}%
\end{pgfscope}%
\begin{pgfscope}%
\pgfpathrectangle{\pgfqpoint{1.220874in}{0.454405in}}{\pgfqpoint{4.650000in}{3.020000in}}%
\pgfusepath{clip}%
\pgfsetrectcap%
\pgfsetroundjoin%
\pgfsetlinewidth{0.803000pt}%
\definecolor{currentstroke}{rgb}{0.298039,0.298039,0.298039}%
\pgfsetstrokecolor{currentstroke}%
\pgfsetdash{}{0pt}%
\pgfpathmoveto{\pgfqpoint{1.569624in}{1.566892in}}%
\pgfpathlineto{\pgfqpoint{2.034624in}{1.566892in}}%
\pgfusepath{stroke}%
\end{pgfscope}%
\begin{pgfscope}%
\pgfpathrectangle{\pgfqpoint{1.220874in}{0.454405in}}{\pgfqpoint{4.650000in}{3.020000in}}%
\pgfusepath{clip}%
\pgfsetrectcap%
\pgfsetroundjoin%
\pgfsetlinewidth{0.803000pt}%
\definecolor{currentstroke}{rgb}{0.298039,0.298039,0.298039}%
\pgfsetstrokecolor{currentstroke}%
\pgfsetdash{}{0pt}%
\pgfpathmoveto{\pgfqpoint{1.569624in}{1.795545in}}%
\pgfpathlineto{\pgfqpoint{2.034624in}{1.795545in}}%
\pgfusepath{stroke}%
\end{pgfscope}%
\begin{pgfscope}%
\pgfpathrectangle{\pgfqpoint{1.220874in}{0.454405in}}{\pgfqpoint{4.650000in}{3.020000in}}%
\pgfusepath{clip}%
\pgfsetrectcap%
\pgfsetroundjoin%
\pgfsetlinewidth{0.803000pt}%
\definecolor{currentstroke}{rgb}{0.298039,0.298039,0.298039}%
\pgfsetstrokecolor{currentstroke}%
\pgfsetdash{}{0pt}%
\pgfpathmoveto{\pgfqpoint{1.337124in}{1.791121in}}%
\pgfpathlineto{\pgfqpoint{2.267124in}{1.791121in}}%
\pgfusepath{stroke}%
\end{pgfscope}%
\begin{pgfscope}%
\pgfpathrectangle{\pgfqpoint{1.220874in}{0.454405in}}{\pgfqpoint{4.650000in}{3.020000in}}%
\pgfusepath{clip}%
\pgfsetrectcap%
\pgfsetroundjoin%
\pgfsetlinewidth{0.803000pt}%
\definecolor{currentstroke}{rgb}{0.298039,0.298039,0.298039}%
\pgfsetstrokecolor{currentstroke}%
\pgfsetdash{}{0pt}%
\pgfpathmoveto{\pgfqpoint{2.964624in}{1.512773in}}%
\pgfpathlineto{\pgfqpoint{2.964624in}{1.423750in}}%
\pgfusepath{stroke}%
\end{pgfscope}%
\begin{pgfscope}%
\pgfpathrectangle{\pgfqpoint{1.220874in}{0.454405in}}{\pgfqpoint{4.650000in}{3.020000in}}%
\pgfusepath{clip}%
\pgfsetrectcap%
\pgfsetroundjoin%
\pgfsetlinewidth{0.803000pt}%
\definecolor{currentstroke}{rgb}{0.298039,0.298039,0.298039}%
\pgfsetstrokecolor{currentstroke}%
\pgfsetdash{}{0pt}%
\pgfpathmoveto{\pgfqpoint{2.964624in}{1.596679in}}%
\pgfpathlineto{\pgfqpoint{2.964624in}{1.674223in}}%
\pgfusepath{stroke}%
\end{pgfscope}%
\begin{pgfscope}%
\pgfpathrectangle{\pgfqpoint{1.220874in}{0.454405in}}{\pgfqpoint{4.650000in}{3.020000in}}%
\pgfusepath{clip}%
\pgfsetrectcap%
\pgfsetroundjoin%
\pgfsetlinewidth{0.803000pt}%
\definecolor{currentstroke}{rgb}{0.298039,0.298039,0.298039}%
\pgfsetstrokecolor{currentstroke}%
\pgfsetdash{}{0pt}%
\pgfpathmoveto{\pgfqpoint{2.732124in}{1.423750in}}%
\pgfpathlineto{\pgfqpoint{3.197124in}{1.423750in}}%
\pgfusepath{stroke}%
\end{pgfscope}%
\begin{pgfscope}%
\pgfpathrectangle{\pgfqpoint{1.220874in}{0.454405in}}{\pgfqpoint{4.650000in}{3.020000in}}%
\pgfusepath{clip}%
\pgfsetrectcap%
\pgfsetroundjoin%
\pgfsetlinewidth{0.803000pt}%
\definecolor{currentstroke}{rgb}{0.298039,0.298039,0.298039}%
\pgfsetstrokecolor{currentstroke}%
\pgfsetdash{}{0pt}%
\pgfpathmoveto{\pgfqpoint{2.732124in}{1.674223in}}%
\pgfpathlineto{\pgfqpoint{3.197124in}{1.674223in}}%
\pgfusepath{stroke}%
\end{pgfscope}%
\begin{pgfscope}%
\pgfpathrectangle{\pgfqpoint{1.220874in}{0.454405in}}{\pgfqpoint{4.650000in}{3.020000in}}%
\pgfusepath{clip}%
\pgfsetrectcap%
\pgfsetroundjoin%
\pgfsetlinewidth{0.803000pt}%
\definecolor{currentstroke}{rgb}{0.298039,0.298039,0.298039}%
\pgfsetstrokecolor{currentstroke}%
\pgfsetdash{}{0pt}%
\pgfpathmoveto{\pgfqpoint{2.499624in}{1.535416in}}%
\pgfpathlineto{\pgfqpoint{3.429624in}{1.535416in}}%
\pgfusepath{stroke}%
\end{pgfscope}%
\begin{pgfscope}%
\pgfpathrectangle{\pgfqpoint{1.220874in}{0.454405in}}{\pgfqpoint{4.650000in}{3.020000in}}%
\pgfusepath{clip}%
\pgfsetbuttcap%
\pgfsetmiterjoin%
\definecolor{currentfill}{rgb}{0.298039,0.298039,0.298039}%
\pgfsetfillcolor{currentfill}%
\pgfsetlinewidth{1.003750pt}%
\definecolor{currentstroke}{rgb}{0.298039,0.298039,0.298039}%
\pgfsetstrokecolor{currentstroke}%
\pgfsetdash{}{0pt}%
\pgfsys@defobject{currentmarker}{\pgfqpoint{-0.029463in}{-0.049105in}}{\pgfqpoint{0.029463in}{0.049105in}}{%
\pgfpathmoveto{\pgfqpoint{0.000000in}{-0.049105in}}%
\pgfpathlineto{\pgfqpoint{0.029463in}{0.000000in}}%
\pgfpathlineto{\pgfqpoint{0.000000in}{0.049105in}}%
\pgfpathlineto{\pgfqpoint{-0.029463in}{0.000000in}}%
\pgfpathclose%
\pgfusepath{stroke,fill}%
}%
\begin{pgfscope}%
\pgfsys@transformshift{2.964624in}{1.339314in}%
\pgfsys@useobject{currentmarker}{}%
\end{pgfscope}%
\end{pgfscope}%
\begin{pgfscope}%
\pgfpathrectangle{\pgfqpoint{1.220874in}{0.454405in}}{\pgfqpoint{4.650000in}{3.020000in}}%
\pgfusepath{clip}%
\pgfsetrectcap%
\pgfsetroundjoin%
\pgfsetlinewidth{0.803000pt}%
\definecolor{currentstroke}{rgb}{0.298039,0.298039,0.298039}%
\pgfsetstrokecolor{currentstroke}%
\pgfsetdash{}{0pt}%
\pgfpathmoveto{\pgfqpoint{4.127124in}{1.697232in}}%
\pgfpathlineto{\pgfqpoint{4.127124in}{1.496408in}}%
\pgfusepath{stroke}%
\end{pgfscope}%
\begin{pgfscope}%
\pgfpathrectangle{\pgfqpoint{1.220874in}{0.454405in}}{\pgfqpoint{4.650000in}{3.020000in}}%
\pgfusepath{clip}%
\pgfsetrectcap%
\pgfsetroundjoin%
\pgfsetlinewidth{0.803000pt}%
\definecolor{currentstroke}{rgb}{0.298039,0.298039,0.298039}%
\pgfsetstrokecolor{currentstroke}%
\pgfsetdash{}{0pt}%
\pgfpathmoveto{\pgfqpoint{4.127124in}{1.816955in}}%
\pgfpathlineto{\pgfqpoint{4.127124in}{1.951222in}}%
\pgfusepath{stroke}%
\end{pgfscope}%
\begin{pgfscope}%
\pgfpathrectangle{\pgfqpoint{1.220874in}{0.454405in}}{\pgfqpoint{4.650000in}{3.020000in}}%
\pgfusepath{clip}%
\pgfsetrectcap%
\pgfsetroundjoin%
\pgfsetlinewidth{0.803000pt}%
\definecolor{currentstroke}{rgb}{0.298039,0.298039,0.298039}%
\pgfsetstrokecolor{currentstroke}%
\pgfsetdash{}{0pt}%
\pgfpathmoveto{\pgfqpoint{3.894624in}{1.496408in}}%
\pgfpathlineto{\pgfqpoint{4.359624in}{1.496408in}}%
\pgfusepath{stroke}%
\end{pgfscope}%
\begin{pgfscope}%
\pgfpathrectangle{\pgfqpoint{1.220874in}{0.454405in}}{\pgfqpoint{4.650000in}{3.020000in}}%
\pgfusepath{clip}%
\pgfsetrectcap%
\pgfsetroundjoin%
\pgfsetlinewidth{0.803000pt}%
\definecolor{currentstroke}{rgb}{0.298039,0.298039,0.298039}%
\pgfsetstrokecolor{currentstroke}%
\pgfsetdash{}{0pt}%
\pgfpathmoveto{\pgfqpoint{3.894624in}{1.951222in}}%
\pgfpathlineto{\pgfqpoint{4.359624in}{1.951222in}}%
\pgfusepath{stroke}%
\end{pgfscope}%
\begin{pgfscope}%
\pgfpathrectangle{\pgfqpoint{1.220874in}{0.454405in}}{\pgfqpoint{4.650000in}{3.020000in}}%
\pgfusepath{clip}%
\pgfsetrectcap%
\pgfsetroundjoin%
\pgfsetlinewidth{0.803000pt}%
\definecolor{currentstroke}{rgb}{0.298039,0.298039,0.298039}%
\pgfsetstrokecolor{currentstroke}%
\pgfsetdash{}{0pt}%
\pgfpathmoveto{\pgfqpoint{3.662124in}{1.792091in}}%
\pgfpathlineto{\pgfqpoint{4.592124in}{1.792091in}}%
\pgfusepath{stroke}%
\end{pgfscope}%
\begin{pgfscope}%
\pgfpathrectangle{\pgfqpoint{1.220874in}{0.454405in}}{\pgfqpoint{4.650000in}{3.020000in}}%
\pgfusepath{clip}%
\pgfsetrectcap%
\pgfsetroundjoin%
\pgfsetlinewidth{0.803000pt}%
\definecolor{currentstroke}{rgb}{0.298039,0.298039,0.298039}%
\pgfsetstrokecolor{currentstroke}%
\pgfsetdash{}{0pt}%
\pgfpathmoveto{\pgfqpoint{5.289624in}{2.563558in}}%
\pgfpathlineto{\pgfqpoint{5.289624in}{2.504909in}}%
\pgfusepath{stroke}%
\end{pgfscope}%
\begin{pgfscope}%
\pgfpathrectangle{\pgfqpoint{1.220874in}{0.454405in}}{\pgfqpoint{4.650000in}{3.020000in}}%
\pgfusepath{clip}%
\pgfsetrectcap%
\pgfsetroundjoin%
\pgfsetlinewidth{0.803000pt}%
\definecolor{currentstroke}{rgb}{0.298039,0.298039,0.298039}%
\pgfsetstrokecolor{currentstroke}%
\pgfsetdash{}{0pt}%
\pgfpathmoveto{\pgfqpoint{5.289624in}{2.631782in}}%
\pgfpathlineto{\pgfqpoint{5.289624in}{2.650913in}}%
\pgfusepath{stroke}%
\end{pgfscope}%
\begin{pgfscope}%
\pgfpathrectangle{\pgfqpoint{1.220874in}{0.454405in}}{\pgfqpoint{4.650000in}{3.020000in}}%
\pgfusepath{clip}%
\pgfsetrectcap%
\pgfsetroundjoin%
\pgfsetlinewidth{0.803000pt}%
\definecolor{currentstroke}{rgb}{0.298039,0.298039,0.298039}%
\pgfsetstrokecolor{currentstroke}%
\pgfsetdash{}{0pt}%
\pgfpathmoveto{\pgfqpoint{5.057124in}{2.504909in}}%
\pgfpathlineto{\pgfqpoint{5.522124in}{2.504909in}}%
\pgfusepath{stroke}%
\end{pgfscope}%
\begin{pgfscope}%
\pgfpathrectangle{\pgfqpoint{1.220874in}{0.454405in}}{\pgfqpoint{4.650000in}{3.020000in}}%
\pgfusepath{clip}%
\pgfsetrectcap%
\pgfsetroundjoin%
\pgfsetlinewidth{0.803000pt}%
\definecolor{currentstroke}{rgb}{0.298039,0.298039,0.298039}%
\pgfsetstrokecolor{currentstroke}%
\pgfsetdash{}{0pt}%
\pgfpathmoveto{\pgfqpoint{5.057124in}{2.650913in}}%
\pgfpathlineto{\pgfqpoint{5.522124in}{2.650913in}}%
\pgfusepath{stroke}%
\end{pgfscope}%
\begin{pgfscope}%
\pgfpathrectangle{\pgfqpoint{1.220874in}{0.454405in}}{\pgfqpoint{4.650000in}{3.020000in}}%
\pgfusepath{clip}%
\pgfsetrectcap%
\pgfsetroundjoin%
\pgfsetlinewidth{0.803000pt}%
\definecolor{currentstroke}{rgb}{0.298039,0.298039,0.298039}%
\pgfsetstrokecolor{currentstroke}%
\pgfsetdash{}{0pt}%
\pgfpathmoveto{\pgfqpoint{4.824624in}{2.602946in}}%
\pgfpathlineto{\pgfqpoint{5.754624in}{2.602946in}}%
\pgfusepath{stroke}%
\end{pgfscope}%
\begin{pgfscope}%
\pgfpathrectangle{\pgfqpoint{1.220874in}{0.454405in}}{\pgfqpoint{4.650000in}{3.020000in}}%
\pgfusepath{clip}%
\pgfsetbuttcap%
\pgfsetmiterjoin%
\definecolor{currentfill}{rgb}{0.298039,0.298039,0.298039}%
\pgfsetfillcolor{currentfill}%
\pgfsetlinewidth{1.003750pt}%
\definecolor{currentstroke}{rgb}{0.298039,0.298039,0.298039}%
\pgfsetstrokecolor{currentstroke}%
\pgfsetdash{}{0pt}%
\pgfsys@defobject{currentmarker}{\pgfqpoint{-0.029463in}{-0.049105in}}{\pgfqpoint{0.029463in}{0.049105in}}{%
\pgfpathmoveto{\pgfqpoint{0.000000in}{-0.049105in}}%
\pgfpathlineto{\pgfqpoint{0.029463in}{0.000000in}}%
\pgfpathlineto{\pgfqpoint{0.000000in}{0.049105in}}%
\pgfpathlineto{\pgfqpoint{-0.029463in}{0.000000in}}%
\pgfpathclose%
\pgfusepath{stroke,fill}%
}%
\begin{pgfscope}%
\pgfsys@transformshift{5.289624in}{2.414422in}%
\pgfsys@useobject{currentmarker}{}%
\end{pgfscope}%
\begin{pgfscope}%
\pgfsys@transformshift{5.289624in}{2.751222in}%
\pgfsys@useobject{currentmarker}{}%
\end{pgfscope}%
\end{pgfscope}%
\end{pgfpicture}%
\makeatother%
\endgroup%

%% file: fig/boxplot_circle-1000-5d_res.pgf
%% Creator: Matplotlib, PGF backend
%%
%% To include the figure in your LaTeX document, write
%%   \input{<filename>.pgf}
%%
%% Make sure the required packages are loaded in your preamble
%%   \usepackage{pgf}
%%
%% Figures using additional raster images can only be included by \input if
%% they are in the same directory as the main LaTeX file. For loading figures
%% from other directories you can use the `import` package
%% and then include the figures with
%%   \import{<path to file>}{<filename>.pgf}
%%
%% Matplotlib used the following preamble
%%
\begingroup%
\makeatletter%
\begin{pgfpicture}%
\pgfpathrectangle{\pgfpointorigin}{\pgfqpoint{4.888889in}{3.524405in}}%
\pgfusepath{use as bounding box, clip}%
\begin{pgfscope}%
\pgfsetbuttcap%
\pgfsetmiterjoin%
\definecolor{currentfill}{rgb}{1.000000,1.000000,1.000000}%
\pgfsetfillcolor{currentfill}%
\pgfsetlinewidth{0.000000pt}%
\definecolor{currentstroke}{rgb}{1.000000,1.000000,1.000000}%
\pgfsetstrokecolor{currentstroke}%
\pgfsetdash{}{0pt}%
\pgfpathmoveto{\pgfqpoint{0.000000in}{0.000000in}}%
\pgfpathlineto{\pgfqpoint{4.888889in}{0.000000in}}%
\pgfpathlineto{\pgfqpoint{4.888889in}{3.524405in}}%
\pgfpathlineto{\pgfqpoint{0.000000in}{3.524405in}}%
\pgfpathclose%
\pgfusepath{fill}%
\end{pgfscope}%
\begin{pgfscope}%
\pgfsetbuttcap%
\pgfsetmiterjoin%
\definecolor{currentfill}{rgb}{1.000000,1.000000,1.000000}%
\pgfsetfillcolor{currentfill}%
\pgfsetlinewidth{0.000000pt}%
\definecolor{currentstroke}{rgb}{0.000000,0.000000,0.000000}%
\pgfsetstrokecolor{currentstroke}%
\pgfsetstrokeopacity{0.000000}%
\pgfsetdash{}{0pt}%
\pgfpathmoveto{\pgfqpoint{0.188889in}{0.454405in}}%
\pgfpathlineto{\pgfqpoint{4.838889in}{0.454405in}}%
\pgfpathlineto{\pgfqpoint{4.838889in}{3.474405in}}%
\pgfpathlineto{\pgfqpoint{0.188889in}{3.474405in}}%
\pgfpathclose%
\pgfusepath{fill}%
\end{pgfscope}%
\begin{pgfscope}%
\definecolor{textcolor}{rgb}{0.000000,0.000000,0.000000}%
\pgfsetstrokecolor{textcolor}%
\pgfsetfillcolor{textcolor}%
\pgftext[x=0.770139in,y=0.357183in,,top]{\color{textcolor}\rmfamily\fontsize{24.200000}{29.040000}\selectfont \textsc{w-l}}%
\end{pgfscope}%
\begin{pgfscope}%
\definecolor{textcolor}{rgb}{0.000000,0.000000,0.000000}%
\pgfsetstrokecolor{textcolor}%
\pgfsetfillcolor{textcolor}%
\pgftext[x=1.932639in,y=0.357183in,,top]{\color{textcolor}\rmfamily\fontsize{24.200000}{29.040000}\selectfont \textsc{w-m}}%
\end{pgfscope}%
\begin{pgfscope}%
\definecolor{textcolor}{rgb}{0.000000,0.000000,0.000000}%
\pgfsetstrokecolor{textcolor}%
\pgfsetfillcolor{textcolor}%
\pgftext[x=3.095139in,y=0.357183in,,top]{\color{textcolor}\rmfamily\fontsize{24.200000}{29.040000}\selectfont \textsc{dcv}}%
\end{pgfscope}%
\begin{pgfscope}%
\definecolor{textcolor}{rgb}{0.000000,0.000000,0.000000}%
\pgfsetstrokecolor{textcolor}%
\pgfsetfillcolor{textcolor}%
\pgftext[x=4.257639in,y=0.357183in,,top]{\color{textcolor}\rmfamily\fontsize{24.200000}{29.040000}\selectfont \textsc{mc}}%
\end{pgfscope}%
\begin{pgfscope}%
\pgfpathrectangle{\pgfqpoint{0.188889in}{0.454405in}}{\pgfqpoint{4.650000in}{3.020000in}}%
\pgfusepath{clip}%
\pgfsetrectcap%
\pgfsetroundjoin%
\pgfsetlinewidth{2.509375pt}%
\definecolor{currentstroke}{rgb}{0.800000,0.800000,0.800000}%
\pgfsetstrokecolor{currentstroke}%
\pgfsetdash{}{0pt}%
\pgfpathmoveto{\pgfqpoint{0.188889in}{0.454405in}}%
\pgfpathlineto{\pgfqpoint{4.838889in}{0.454405in}}%
\pgfusepath{stroke}%
\end{pgfscope}%
\begin{pgfscope}%
\pgfsetbuttcap%
\pgfsetroundjoin%
\definecolor{currentfill}{rgb}{0.800000,0.800000,0.800000}%
\pgfsetfillcolor{currentfill}%
\pgfsetlinewidth{1.003750pt}%
\definecolor{currentstroke}{rgb}{0.800000,0.800000,0.800000}%
\pgfsetstrokecolor{currentstroke}%
\pgfsetdash{}{0pt}%
\pgfsys@defobject{currentmarker}{\pgfqpoint{-0.138889in}{0.000000in}}{\pgfqpoint{0.000000in}{0.000000in}}{%
\pgfpathmoveto{\pgfqpoint{0.000000in}{0.000000in}}%
\pgfpathlineto{\pgfqpoint{-0.138889in}{0.000000in}}%
\pgfusepath{stroke,fill}%
}%
\begin{pgfscope}%
\pgfsys@transformshift{0.188889in}{0.454405in}%
\pgfsys@useobject{currentmarker}{}%
\end{pgfscope}%
\end{pgfscope}%
\begin{pgfscope}%
\pgfpathrectangle{\pgfqpoint{0.188889in}{0.454405in}}{\pgfqpoint{4.650000in}{3.020000in}}%
\pgfusepath{clip}%
\pgfsetrectcap%
\pgfsetroundjoin%
\pgfsetlinewidth{2.509375pt}%
\definecolor{currentstroke}{rgb}{0.611765,0.611765,0.611765}%
\pgfsetstrokecolor{currentstroke}%
\pgfsetdash{}{0pt}%
\pgfpathmoveto{\pgfqpoint{0.188889in}{2.394902in}}%
\pgfpathlineto{\pgfqpoint{4.838889in}{2.394902in}}%
\pgfusepath{stroke}%
\end{pgfscope}%
\begin{pgfscope}%
\pgfsetbuttcap%
\pgfsetroundjoin%
\definecolor{currentfill}{rgb}{0.800000,0.800000,0.800000}%
\pgfsetfillcolor{currentfill}%
\pgfsetlinewidth{1.003750pt}%
\definecolor{currentstroke}{rgb}{0.800000,0.800000,0.800000}%
\pgfsetstrokecolor{currentstroke}%
\pgfsetdash{}{0pt}%
\pgfsys@defobject{currentmarker}{\pgfqpoint{-0.138889in}{0.000000in}}{\pgfqpoint{0.000000in}{0.000000in}}{%
\pgfpathmoveto{\pgfqpoint{0.000000in}{0.000000in}}%
\pgfpathlineto{\pgfqpoint{-0.138889in}{0.000000in}}%
\pgfusepath{stroke,fill}%
}%
\begin{pgfscope}%
\pgfsys@transformshift{0.188889in}{2.394902in}%
\pgfsys@useobject{currentmarker}{}%
\end{pgfscope}%
\end{pgfscope}%
\begin{pgfscope}%
\pgfpathrectangle{\pgfqpoint{0.188889in}{0.454405in}}{\pgfqpoint{4.650000in}{3.020000in}}%
\pgfusepath{clip}%
\pgfsetrectcap%
\pgfsetroundjoin%
\pgfsetlinewidth{0.401500pt}%
\definecolor{currentstroke}{rgb}{0.800000,0.800000,0.800000}%
\pgfsetstrokecolor{currentstroke}%
\pgfsetstrokeopacity{0.550000}%
\pgfsetdash{}{0pt}%
\pgfpathmoveto{\pgfqpoint{0.188889in}{1.038553in}}%
\pgfpathlineto{\pgfqpoint{4.838889in}{1.038553in}}%
\pgfusepath{stroke}%
\end{pgfscope}%
\begin{pgfscope}%
\pgfpathrectangle{\pgfqpoint{0.188889in}{0.454405in}}{\pgfqpoint{4.650000in}{3.020000in}}%
\pgfusepath{clip}%
\pgfsetrectcap%
\pgfsetroundjoin%
\pgfsetlinewidth{0.401500pt}%
\definecolor{currentstroke}{rgb}{0.800000,0.800000,0.800000}%
\pgfsetstrokecolor{currentstroke}%
\pgfsetstrokeopacity{0.550000}%
\pgfsetdash{}{0pt}%
\pgfpathmoveto{\pgfqpoint{0.188889in}{1.380257in}}%
\pgfpathlineto{\pgfqpoint{4.838889in}{1.380257in}}%
\pgfusepath{stroke}%
\end{pgfscope}%
\begin{pgfscope}%
\pgfpathrectangle{\pgfqpoint{0.188889in}{0.454405in}}{\pgfqpoint{4.650000in}{3.020000in}}%
\pgfusepath{clip}%
\pgfsetrectcap%
\pgfsetroundjoin%
\pgfsetlinewidth{0.401500pt}%
\definecolor{currentstroke}{rgb}{0.800000,0.800000,0.800000}%
\pgfsetstrokecolor{currentstroke}%
\pgfsetstrokeopacity{0.550000}%
\pgfsetdash{}{0pt}%
\pgfpathmoveto{\pgfqpoint{0.188889in}{1.622700in}}%
\pgfpathlineto{\pgfqpoint{4.838889in}{1.622700in}}%
\pgfusepath{stroke}%
\end{pgfscope}%
\begin{pgfscope}%
\pgfpathrectangle{\pgfqpoint{0.188889in}{0.454405in}}{\pgfqpoint{4.650000in}{3.020000in}}%
\pgfusepath{clip}%
\pgfsetrectcap%
\pgfsetroundjoin%
\pgfsetlinewidth{0.401500pt}%
\definecolor{currentstroke}{rgb}{0.800000,0.800000,0.800000}%
\pgfsetstrokecolor{currentstroke}%
\pgfsetstrokeopacity{0.550000}%
\pgfsetdash{}{0pt}%
\pgfpathmoveto{\pgfqpoint{0.188889in}{1.810754in}}%
\pgfpathlineto{\pgfqpoint{4.838889in}{1.810754in}}%
\pgfusepath{stroke}%
\end{pgfscope}%
\begin{pgfscope}%
\pgfpathrectangle{\pgfqpoint{0.188889in}{0.454405in}}{\pgfqpoint{4.650000in}{3.020000in}}%
\pgfusepath{clip}%
\pgfsetrectcap%
\pgfsetroundjoin%
\pgfsetlinewidth{0.401500pt}%
\definecolor{currentstroke}{rgb}{0.800000,0.800000,0.800000}%
\pgfsetstrokecolor{currentstroke}%
\pgfsetstrokeopacity{0.550000}%
\pgfsetdash{}{0pt}%
\pgfpathmoveto{\pgfqpoint{0.188889in}{1.964405in}}%
\pgfpathlineto{\pgfqpoint{4.838889in}{1.964405in}}%
\pgfusepath{stroke}%
\end{pgfscope}%
\begin{pgfscope}%
\pgfpathrectangle{\pgfqpoint{0.188889in}{0.454405in}}{\pgfqpoint{4.650000in}{3.020000in}}%
\pgfusepath{clip}%
\pgfsetrectcap%
\pgfsetroundjoin%
\pgfsetlinewidth{0.401500pt}%
\definecolor{currentstroke}{rgb}{0.800000,0.800000,0.800000}%
\pgfsetstrokecolor{currentstroke}%
\pgfsetstrokeopacity{0.550000}%
\pgfsetdash{}{0pt}%
\pgfpathmoveto{\pgfqpoint{0.188889in}{2.094315in}}%
\pgfpathlineto{\pgfqpoint{4.838889in}{2.094315in}}%
\pgfusepath{stroke}%
\end{pgfscope}%
\begin{pgfscope}%
\pgfpathrectangle{\pgfqpoint{0.188889in}{0.454405in}}{\pgfqpoint{4.650000in}{3.020000in}}%
\pgfusepath{clip}%
\pgfsetrectcap%
\pgfsetroundjoin%
\pgfsetlinewidth{0.401500pt}%
\definecolor{currentstroke}{rgb}{0.800000,0.800000,0.800000}%
\pgfsetstrokecolor{currentstroke}%
\pgfsetstrokeopacity{0.550000}%
\pgfsetdash{}{0pt}%
\pgfpathmoveto{\pgfqpoint{0.188889in}{2.206848in}}%
\pgfpathlineto{\pgfqpoint{4.838889in}{2.206848in}}%
\pgfusepath{stroke}%
\end{pgfscope}%
\begin{pgfscope}%
\pgfpathrectangle{\pgfqpoint{0.188889in}{0.454405in}}{\pgfqpoint{4.650000in}{3.020000in}}%
\pgfusepath{clip}%
\pgfsetrectcap%
\pgfsetroundjoin%
\pgfsetlinewidth{0.401500pt}%
\definecolor{currentstroke}{rgb}{0.800000,0.800000,0.800000}%
\pgfsetstrokecolor{currentstroke}%
\pgfsetstrokeopacity{0.550000}%
\pgfsetdash{}{0pt}%
\pgfpathmoveto{\pgfqpoint{0.188889in}{2.306110in}}%
\pgfpathlineto{\pgfqpoint{4.838889in}{2.306110in}}%
\pgfusepath{stroke}%
\end{pgfscope}%
\begin{pgfscope}%
\pgfpathrectangle{\pgfqpoint{0.188889in}{0.454405in}}{\pgfqpoint{4.650000in}{3.020000in}}%
\pgfusepath{clip}%
\pgfsetrectcap%
\pgfsetroundjoin%
\pgfsetlinewidth{0.401500pt}%
\definecolor{currentstroke}{rgb}{0.800000,0.800000,0.800000}%
\pgfsetstrokecolor{currentstroke}%
\pgfsetstrokeopacity{0.550000}%
\pgfsetdash{}{0pt}%
\pgfpathmoveto{\pgfqpoint{0.188889in}{2.979050in}}%
\pgfpathlineto{\pgfqpoint{4.838889in}{2.979050in}}%
\pgfusepath{stroke}%
\end{pgfscope}%
\begin{pgfscope}%
\pgfpathrectangle{\pgfqpoint{0.188889in}{0.454405in}}{\pgfqpoint{4.650000in}{3.020000in}}%
\pgfusepath{clip}%
\pgfsetrectcap%
\pgfsetroundjoin%
\pgfsetlinewidth{0.401500pt}%
\definecolor{currentstroke}{rgb}{0.800000,0.800000,0.800000}%
\pgfsetstrokecolor{currentstroke}%
\pgfsetstrokeopacity{0.550000}%
\pgfsetdash{}{0pt}%
\pgfpathmoveto{\pgfqpoint{0.188889in}{3.320754in}}%
\pgfpathlineto{\pgfqpoint{4.838889in}{3.320754in}}%
\pgfusepath{stroke}%
\end{pgfscope}%
\begin{pgfscope}%
\pgfpathrectangle{\pgfqpoint{0.188889in}{0.454405in}}{\pgfqpoint{4.650000in}{3.020000in}}%
\pgfusepath{clip}%
\pgfsetbuttcap%
\pgfsetmiterjoin%
\definecolor{currentfill}{rgb}{0.347059,0.458824,0.641176}%
\pgfsetfillcolor{currentfill}%
\pgfsetlinewidth{0.803000pt}%
\definecolor{currentstroke}{rgb}{0.298039,0.298039,0.298039}%
\pgfsetstrokecolor{currentstroke}%
\pgfsetdash{}{0pt}%
\pgfpathmoveto{\pgfqpoint{0.305139in}{3.162118in}}%
\pgfpathlineto{\pgfqpoint{1.235139in}{3.162118in}}%
\pgfpathlineto{\pgfqpoint{1.235139in}{3.189200in}}%
\pgfpathlineto{\pgfqpoint{0.305139in}{3.189200in}}%
\pgfpathlineto{\pgfqpoint{0.305139in}{3.162118in}}%
\pgfpathclose%
\pgfusepath{stroke,fill}%
\end{pgfscope}%
\begin{pgfscope}%
\pgfpathrectangle{\pgfqpoint{0.188889in}{0.454405in}}{\pgfqpoint{4.650000in}{3.020000in}}%
\pgfusepath{clip}%
\pgfsetbuttcap%
\pgfsetmiterjoin%
\definecolor{currentfill}{rgb}{0.798529,0.536765,0.389706}%
\pgfsetfillcolor{currentfill}%
\pgfsetlinewidth{0.803000pt}%
\definecolor{currentstroke}{rgb}{0.298039,0.298039,0.298039}%
\pgfsetstrokecolor{currentstroke}%
\pgfsetdash{}{0pt}%
\pgfpathmoveto{\pgfqpoint{1.467639in}{2.728856in}}%
\pgfpathlineto{\pgfqpoint{2.397639in}{2.728856in}}%
\pgfpathlineto{\pgfqpoint{2.397639in}{2.817950in}}%
\pgfpathlineto{\pgfqpoint{1.467639in}{2.817950in}}%
\pgfpathlineto{\pgfqpoint{1.467639in}{2.728856in}}%
\pgfpathclose%
\pgfusepath{stroke,fill}%
\end{pgfscope}%
\begin{pgfscope}%
\pgfpathrectangle{\pgfqpoint{0.188889in}{0.454405in}}{\pgfqpoint{4.650000in}{3.020000in}}%
\pgfusepath{clip}%
\pgfsetbuttcap%
\pgfsetmiterjoin%
\definecolor{currentfill}{rgb}{0.374020,0.618137,0.429902}%
\pgfsetfillcolor{currentfill}%
\pgfsetlinewidth{0.803000pt}%
\definecolor{currentstroke}{rgb}{0.298039,0.298039,0.298039}%
\pgfsetstrokecolor{currentstroke}%
\pgfsetdash{}{0pt}%
\pgfpathmoveto{\pgfqpoint{2.630139in}{2.535121in}}%
\pgfpathlineto{\pgfqpoint{3.560139in}{2.535121in}}%
\pgfpathlineto{\pgfqpoint{3.560139in}{2.869912in}}%
\pgfpathlineto{\pgfqpoint{2.630139in}{2.869912in}}%
\pgfpathlineto{\pgfqpoint{2.630139in}{2.535121in}}%
\pgfpathclose%
\pgfusepath{stroke,fill}%
\end{pgfscope}%
\begin{pgfscope}%
\pgfpathrectangle{\pgfqpoint{0.188889in}{0.454405in}}{\pgfqpoint{4.650000in}{3.020000in}}%
\pgfusepath{clip}%
\pgfsetbuttcap%
\pgfsetmiterjoin%
\definecolor{currentfill}{rgb}{0.710784,0.363725,0.375490}%
\pgfsetfillcolor{currentfill}%
\pgfsetlinewidth{0.803000pt}%
\definecolor{currentstroke}{rgb}{0.298039,0.298039,0.298039}%
\pgfsetstrokecolor{currentstroke}%
\pgfsetdash{}{0pt}%
\pgfpathmoveto{\pgfqpoint{3.792639in}{2.384818in}}%
\pgfpathlineto{\pgfqpoint{4.722639in}{2.384818in}}%
\pgfpathlineto{\pgfqpoint{4.722639in}{2.566201in}}%
\pgfpathlineto{\pgfqpoint{3.792639in}{2.566201in}}%
\pgfpathlineto{\pgfqpoint{3.792639in}{2.384818in}}%
\pgfpathclose%
\pgfusepath{stroke,fill}%
\end{pgfscope}%
\begin{pgfscope}%
\pgfsetrectcap%
\pgfsetmiterjoin%
\pgfsetlinewidth{1.254687pt}%
\definecolor{currentstroke}{rgb}{0.800000,0.800000,0.800000}%
\pgfsetstrokecolor{currentstroke}%
\pgfsetdash{}{0pt}%
\pgfpathmoveto{\pgfqpoint{0.188889in}{0.454405in}}%
\pgfpathlineto{\pgfqpoint{0.188889in}{3.474405in}}%
\pgfusepath{stroke}%
\end{pgfscope}%
\begin{pgfscope}%
\pgfsetrectcap%
\pgfsetmiterjoin%
\pgfsetlinewidth{1.254687pt}%
\definecolor{currentstroke}{rgb}{0.800000,0.800000,0.800000}%
\pgfsetstrokecolor{currentstroke}%
\pgfsetdash{}{0pt}%
\pgfpathmoveto{\pgfqpoint{4.838889in}{0.454405in}}%
\pgfpathlineto{\pgfqpoint{4.838889in}{3.474405in}}%
\pgfusepath{stroke}%
\end{pgfscope}%
\begin{pgfscope}%
\pgfsetrectcap%
\pgfsetmiterjoin%
\pgfsetlinewidth{1.254687pt}%
\definecolor{currentstroke}{rgb}{0.800000,0.800000,0.800000}%
\pgfsetstrokecolor{currentstroke}%
\pgfsetdash{}{0pt}%
\pgfpathmoveto{\pgfqpoint{0.188889in}{0.454405in}}%
\pgfpathlineto{\pgfqpoint{4.838889in}{0.454405in}}%
\pgfusepath{stroke}%
\end{pgfscope}%
\begin{pgfscope}%
\pgfsetrectcap%
\pgfsetmiterjoin%
\pgfsetlinewidth{1.254687pt}%
\definecolor{currentstroke}{rgb}{0.800000,0.800000,0.800000}%
\pgfsetstrokecolor{currentstroke}%
\pgfsetdash{}{0pt}%
\pgfpathmoveto{\pgfqpoint{0.188889in}{3.474405in}}%
\pgfpathlineto{\pgfqpoint{4.838889in}{3.474405in}}%
\pgfusepath{stroke}%
\end{pgfscope}%
\begin{pgfscope}%
\pgfpathrectangle{\pgfqpoint{0.188889in}{0.454405in}}{\pgfqpoint{4.650000in}{3.020000in}}%
\pgfusepath{clip}%
\pgfsetrectcap%
\pgfsetroundjoin%
\pgfsetlinewidth{0.803000pt}%
\definecolor{currentstroke}{rgb}{0.298039,0.298039,0.298039}%
\pgfsetstrokecolor{currentstroke}%
\pgfsetdash{}{0pt}%
\pgfpathmoveto{\pgfqpoint{0.770139in}{3.162118in}}%
\pgfpathlineto{\pgfqpoint{0.770139in}{3.162118in}}%
\pgfusepath{stroke}%
\end{pgfscope}%
\begin{pgfscope}%
\pgfpathrectangle{\pgfqpoint{0.188889in}{0.454405in}}{\pgfqpoint{4.650000in}{3.020000in}}%
\pgfusepath{clip}%
\pgfsetrectcap%
\pgfsetroundjoin%
\pgfsetlinewidth{0.803000pt}%
\definecolor{currentstroke}{rgb}{0.298039,0.298039,0.298039}%
\pgfsetstrokecolor{currentstroke}%
\pgfsetdash{}{0pt}%
\pgfpathmoveto{\pgfqpoint{0.770139in}{3.189200in}}%
\pgfpathlineto{\pgfqpoint{0.770139in}{3.195774in}}%
\pgfusepath{stroke}%
\end{pgfscope}%
\begin{pgfscope}%
\pgfpathrectangle{\pgfqpoint{0.188889in}{0.454405in}}{\pgfqpoint{4.650000in}{3.020000in}}%
\pgfusepath{clip}%
\pgfsetrectcap%
\pgfsetroundjoin%
\pgfsetlinewidth{0.803000pt}%
\definecolor{currentstroke}{rgb}{0.298039,0.298039,0.298039}%
\pgfsetstrokecolor{currentstroke}%
\pgfsetdash{}{0pt}%
\pgfpathmoveto{\pgfqpoint{0.537639in}{3.162118in}}%
\pgfpathlineto{\pgfqpoint{1.002639in}{3.162118in}}%
\pgfusepath{stroke}%
\end{pgfscope}%
\begin{pgfscope}%
\pgfpathrectangle{\pgfqpoint{0.188889in}{0.454405in}}{\pgfqpoint{4.650000in}{3.020000in}}%
\pgfusepath{clip}%
\pgfsetrectcap%
\pgfsetroundjoin%
\pgfsetlinewidth{0.803000pt}%
\definecolor{currentstroke}{rgb}{0.298039,0.298039,0.298039}%
\pgfsetstrokecolor{currentstroke}%
\pgfsetdash{}{0pt}%
\pgfpathmoveto{\pgfqpoint{0.537639in}{3.195774in}}%
\pgfpathlineto{\pgfqpoint{1.002639in}{3.195774in}}%
\pgfusepath{stroke}%
\end{pgfscope}%
\begin{pgfscope}%
\pgfpathrectangle{\pgfqpoint{0.188889in}{0.454405in}}{\pgfqpoint{4.650000in}{3.020000in}}%
\pgfusepath{clip}%
\pgfsetrectcap%
\pgfsetroundjoin%
\pgfsetlinewidth{0.803000pt}%
\definecolor{currentstroke}{rgb}{0.298039,0.298039,0.298039}%
\pgfsetstrokecolor{currentstroke}%
\pgfsetdash{}{0pt}%
\pgfpathmoveto{\pgfqpoint{0.305139in}{3.162118in}}%
\pgfpathlineto{\pgfqpoint{1.235139in}{3.162118in}}%
\pgfusepath{stroke}%
\end{pgfscope}%
\begin{pgfscope}%
\pgfpathrectangle{\pgfqpoint{0.188889in}{0.454405in}}{\pgfqpoint{4.650000in}{3.020000in}}%
\pgfusepath{clip}%
\pgfsetbuttcap%
\pgfsetmiterjoin%
\definecolor{currentfill}{rgb}{0.298039,0.298039,0.298039}%
\pgfsetfillcolor{currentfill}%
\pgfsetlinewidth{1.003750pt}%
\definecolor{currentstroke}{rgb}{0.298039,0.298039,0.298039}%
\pgfsetstrokecolor{currentstroke}%
\pgfsetdash{}{0pt}%
\pgfsys@defobject{currentmarker}{\pgfqpoint{-0.029463in}{-0.049105in}}{\pgfqpoint{0.029463in}{0.049105in}}{%
\pgfpathmoveto{\pgfqpoint{0.000000in}{-0.049105in}}%
\pgfpathlineto{\pgfqpoint{0.029463in}{0.000000in}}%
\pgfpathlineto{\pgfqpoint{0.000000in}{0.049105in}}%
\pgfpathlineto{\pgfqpoint{-0.029463in}{0.000000in}}%
\pgfpathclose%
\pgfusepath{stroke,fill}%
}%
\begin{pgfscope}%
\pgfsys@transformshift{0.770139in}{3.114632in}%
\pgfsys@useobject{currentmarker}{}%
\end{pgfscope}%
\begin{pgfscope}%
\pgfsys@transformshift{0.770139in}{3.114632in}%
\pgfsys@useobject{currentmarker}{}%
\end{pgfscope}%
\begin{pgfscope}%
\pgfsys@transformshift{0.770139in}{3.235736in}%
\pgfsys@useobject{currentmarker}{}%
\end{pgfscope}%
\begin{pgfscope}%
\pgfsys@transformshift{0.770139in}{3.242198in}%
\pgfsys@useobject{currentmarker}{}%
\end{pgfscope}%
\end{pgfscope}%
\begin{pgfscope}%
\pgfpathrectangle{\pgfqpoint{0.188889in}{0.454405in}}{\pgfqpoint{4.650000in}{3.020000in}}%
\pgfusepath{clip}%
\pgfsetrectcap%
\pgfsetroundjoin%
\pgfsetlinewidth{0.803000pt}%
\definecolor{currentstroke}{rgb}{0.298039,0.298039,0.298039}%
\pgfsetstrokecolor{currentstroke}%
\pgfsetdash{}{0pt}%
\pgfpathmoveto{\pgfqpoint{1.932639in}{2.728856in}}%
\pgfpathlineto{\pgfqpoint{1.932639in}{2.700387in}}%
\pgfusepath{stroke}%
\end{pgfscope}%
\begin{pgfscope}%
\pgfpathrectangle{\pgfqpoint{0.188889in}{0.454405in}}{\pgfqpoint{4.650000in}{3.020000in}}%
\pgfusepath{clip}%
\pgfsetrectcap%
\pgfsetroundjoin%
\pgfsetlinewidth{0.803000pt}%
\definecolor{currentstroke}{rgb}{0.298039,0.298039,0.298039}%
\pgfsetstrokecolor{currentstroke}%
\pgfsetdash{}{0pt}%
\pgfpathmoveto{\pgfqpoint{1.932639in}{2.817950in}}%
\pgfpathlineto{\pgfqpoint{1.932639in}{2.860152in}}%
\pgfusepath{stroke}%
\end{pgfscope}%
\begin{pgfscope}%
\pgfpathrectangle{\pgfqpoint{0.188889in}{0.454405in}}{\pgfqpoint{4.650000in}{3.020000in}}%
\pgfusepath{clip}%
\pgfsetrectcap%
\pgfsetroundjoin%
\pgfsetlinewidth{0.803000pt}%
\definecolor{currentstroke}{rgb}{0.298039,0.298039,0.298039}%
\pgfsetstrokecolor{currentstroke}%
\pgfsetdash{}{0pt}%
\pgfpathmoveto{\pgfqpoint{1.700139in}{2.700387in}}%
\pgfpathlineto{\pgfqpoint{2.165139in}{2.700387in}}%
\pgfusepath{stroke}%
\end{pgfscope}%
\begin{pgfscope}%
\pgfpathrectangle{\pgfqpoint{0.188889in}{0.454405in}}{\pgfqpoint{4.650000in}{3.020000in}}%
\pgfusepath{clip}%
\pgfsetrectcap%
\pgfsetroundjoin%
\pgfsetlinewidth{0.803000pt}%
\definecolor{currentstroke}{rgb}{0.298039,0.298039,0.298039}%
\pgfsetstrokecolor{currentstroke}%
\pgfsetdash{}{0pt}%
\pgfpathmoveto{\pgfqpoint{1.700139in}{2.860152in}}%
\pgfpathlineto{\pgfqpoint{2.165139in}{2.860152in}}%
\pgfusepath{stroke}%
\end{pgfscope}%
\begin{pgfscope}%
\pgfpathrectangle{\pgfqpoint{0.188889in}{0.454405in}}{\pgfqpoint{4.650000in}{3.020000in}}%
\pgfusepath{clip}%
\pgfsetrectcap%
\pgfsetroundjoin%
\pgfsetlinewidth{0.803000pt}%
\definecolor{currentstroke}{rgb}{0.298039,0.298039,0.298039}%
\pgfsetstrokecolor{currentstroke}%
\pgfsetdash{}{0pt}%
\pgfpathmoveto{\pgfqpoint{1.467639in}{2.747704in}}%
\pgfpathlineto{\pgfqpoint{2.397639in}{2.747704in}}%
\pgfusepath{stroke}%
\end{pgfscope}%
\begin{pgfscope}%
\pgfpathrectangle{\pgfqpoint{0.188889in}{0.454405in}}{\pgfqpoint{4.650000in}{3.020000in}}%
\pgfusepath{clip}%
\pgfsetbuttcap%
\pgfsetmiterjoin%
\definecolor{currentfill}{rgb}{0.298039,0.298039,0.298039}%
\pgfsetfillcolor{currentfill}%
\pgfsetlinewidth{1.003750pt}%
\definecolor{currentstroke}{rgb}{0.298039,0.298039,0.298039}%
\pgfsetstrokecolor{currentstroke}%
\pgfsetdash{}{0pt}%
\pgfsys@defobject{currentmarker}{\pgfqpoint{-0.029463in}{-0.049105in}}{\pgfqpoint{0.029463in}{0.049105in}}{%
\pgfpathmoveto{\pgfqpoint{0.000000in}{-0.049105in}}%
\pgfpathlineto{\pgfqpoint{0.029463in}{0.000000in}}%
\pgfpathlineto{\pgfqpoint{0.000000in}{0.049105in}}%
\pgfpathlineto{\pgfqpoint{-0.029463in}{0.000000in}}%
\pgfpathclose%
\pgfusepath{stroke,fill}%
}%
\begin{pgfscope}%
\pgfsys@transformshift{1.932639in}{2.562087in}%
\pgfsys@useobject{currentmarker}{}%
\end{pgfscope}%
\begin{pgfscope}%
\pgfsys@transformshift{1.932639in}{2.562087in}%
\pgfsys@useobject{currentmarker}{}%
\end{pgfscope}%
\end{pgfscope}%
\begin{pgfscope}%
\pgfpathrectangle{\pgfqpoint{0.188889in}{0.454405in}}{\pgfqpoint{4.650000in}{3.020000in}}%
\pgfusepath{clip}%
\pgfsetrectcap%
\pgfsetroundjoin%
\pgfsetlinewidth{0.803000pt}%
\definecolor{currentstroke}{rgb}{0.298039,0.298039,0.298039}%
\pgfsetstrokecolor{currentstroke}%
\pgfsetdash{}{0pt}%
\pgfpathmoveto{\pgfqpoint{3.095139in}{2.535121in}}%
\pgfpathlineto{\pgfqpoint{3.095139in}{2.447275in}}%
\pgfusepath{stroke}%
\end{pgfscope}%
\begin{pgfscope}%
\pgfpathrectangle{\pgfqpoint{0.188889in}{0.454405in}}{\pgfqpoint{4.650000in}{3.020000in}}%
\pgfusepath{clip}%
\pgfsetrectcap%
\pgfsetroundjoin%
\pgfsetlinewidth{0.803000pt}%
\definecolor{currentstroke}{rgb}{0.298039,0.298039,0.298039}%
\pgfsetstrokecolor{currentstroke}%
\pgfsetdash{}{0pt}%
\pgfpathmoveto{\pgfqpoint{3.095139in}{2.869912in}}%
\pgfpathlineto{\pgfqpoint{3.095139in}{2.971865in}}%
\pgfusepath{stroke}%
\end{pgfscope}%
\begin{pgfscope}%
\pgfpathrectangle{\pgfqpoint{0.188889in}{0.454405in}}{\pgfqpoint{4.650000in}{3.020000in}}%
\pgfusepath{clip}%
\pgfsetrectcap%
\pgfsetroundjoin%
\pgfsetlinewidth{0.803000pt}%
\definecolor{currentstroke}{rgb}{0.298039,0.298039,0.298039}%
\pgfsetstrokecolor{currentstroke}%
\pgfsetdash{}{0pt}%
\pgfpathmoveto{\pgfqpoint{2.862639in}{2.447275in}}%
\pgfpathlineto{\pgfqpoint{3.327639in}{2.447275in}}%
\pgfusepath{stroke}%
\end{pgfscope}%
\begin{pgfscope}%
\pgfpathrectangle{\pgfqpoint{0.188889in}{0.454405in}}{\pgfqpoint{4.650000in}{3.020000in}}%
\pgfusepath{clip}%
\pgfsetrectcap%
\pgfsetroundjoin%
\pgfsetlinewidth{0.803000pt}%
\definecolor{currentstroke}{rgb}{0.298039,0.298039,0.298039}%
\pgfsetstrokecolor{currentstroke}%
\pgfsetdash{}{0pt}%
\pgfpathmoveto{\pgfqpoint{2.862639in}{2.971865in}}%
\pgfpathlineto{\pgfqpoint{3.327639in}{2.971865in}}%
\pgfusepath{stroke}%
\end{pgfscope}%
\begin{pgfscope}%
\pgfpathrectangle{\pgfqpoint{0.188889in}{0.454405in}}{\pgfqpoint{4.650000in}{3.020000in}}%
\pgfusepath{clip}%
\pgfsetrectcap%
\pgfsetroundjoin%
\pgfsetlinewidth{0.803000pt}%
\definecolor{currentstroke}{rgb}{0.298039,0.298039,0.298039}%
\pgfsetstrokecolor{currentstroke}%
\pgfsetdash{}{0pt}%
\pgfpathmoveto{\pgfqpoint{2.630139in}{2.708372in}}%
\pgfpathlineto{\pgfqpoint{3.560139in}{2.708372in}}%
\pgfusepath{stroke}%
\end{pgfscope}%
\begin{pgfscope}%
\pgfpathrectangle{\pgfqpoint{0.188889in}{0.454405in}}{\pgfqpoint{4.650000in}{3.020000in}}%
\pgfusepath{clip}%
\pgfsetbuttcap%
\pgfsetmiterjoin%
\definecolor{currentfill}{rgb}{0.298039,0.298039,0.298039}%
\pgfsetfillcolor{currentfill}%
\pgfsetlinewidth{1.003750pt}%
\definecolor{currentstroke}{rgb}{0.298039,0.298039,0.298039}%
\pgfsetstrokecolor{currentstroke}%
\pgfsetdash{}{0pt}%
\pgfsys@defobject{currentmarker}{\pgfqpoint{-0.029463in}{-0.049105in}}{\pgfqpoint{0.029463in}{0.049105in}}{%
\pgfpathmoveto{\pgfqpoint{0.000000in}{-0.049105in}}%
\pgfpathlineto{\pgfqpoint{0.029463in}{0.000000in}}%
\pgfpathlineto{\pgfqpoint{0.000000in}{0.049105in}}%
\pgfpathlineto{\pgfqpoint{-0.029463in}{0.000000in}}%
\pgfpathclose%
\pgfusepath{stroke,fill}%
}%
\begin{pgfscope}%
\pgfsys@transformshift{3.095139in}{3.395500in}%
\pgfsys@useobject{currentmarker}{}%
\end{pgfscope}%
\end{pgfscope}%
\begin{pgfscope}%
\pgfpathrectangle{\pgfqpoint{0.188889in}{0.454405in}}{\pgfqpoint{4.650000in}{3.020000in}}%
\pgfusepath{clip}%
\pgfsetrectcap%
\pgfsetroundjoin%
\pgfsetlinewidth{0.803000pt}%
\definecolor{currentstroke}{rgb}{0.298039,0.298039,0.298039}%
\pgfsetstrokecolor{currentstroke}%
\pgfsetdash{}{0pt}%
\pgfpathmoveto{\pgfqpoint{4.257639in}{2.384818in}}%
\pgfpathlineto{\pgfqpoint{4.257639in}{2.247722in}}%
\pgfusepath{stroke}%
\end{pgfscope}%
\begin{pgfscope}%
\pgfpathrectangle{\pgfqpoint{0.188889in}{0.454405in}}{\pgfqpoint{4.650000in}{3.020000in}}%
\pgfusepath{clip}%
\pgfsetrectcap%
\pgfsetroundjoin%
\pgfsetlinewidth{0.803000pt}%
\definecolor{currentstroke}{rgb}{0.298039,0.298039,0.298039}%
\pgfsetstrokecolor{currentstroke}%
\pgfsetdash{}{0pt}%
\pgfpathmoveto{\pgfqpoint{4.257639in}{2.566201in}}%
\pgfpathlineto{\pgfqpoint{4.257639in}{2.718287in}}%
\pgfusepath{stroke}%
\end{pgfscope}%
\begin{pgfscope}%
\pgfpathrectangle{\pgfqpoint{0.188889in}{0.454405in}}{\pgfqpoint{4.650000in}{3.020000in}}%
\pgfusepath{clip}%
\pgfsetrectcap%
\pgfsetroundjoin%
\pgfsetlinewidth{0.803000pt}%
\definecolor{currentstroke}{rgb}{0.298039,0.298039,0.298039}%
\pgfsetstrokecolor{currentstroke}%
\pgfsetdash{}{0pt}%
\pgfpathmoveto{\pgfqpoint{4.025139in}{2.247722in}}%
\pgfpathlineto{\pgfqpoint{4.490139in}{2.247722in}}%
\pgfusepath{stroke}%
\end{pgfscope}%
\begin{pgfscope}%
\pgfpathrectangle{\pgfqpoint{0.188889in}{0.454405in}}{\pgfqpoint{4.650000in}{3.020000in}}%
\pgfusepath{clip}%
\pgfsetrectcap%
\pgfsetroundjoin%
\pgfsetlinewidth{0.803000pt}%
\definecolor{currentstroke}{rgb}{0.298039,0.298039,0.298039}%
\pgfsetstrokecolor{currentstroke}%
\pgfsetdash{}{0pt}%
\pgfpathmoveto{\pgfqpoint{4.025139in}{2.718287in}}%
\pgfpathlineto{\pgfqpoint{4.490139in}{2.718287in}}%
\pgfusepath{stroke}%
\end{pgfscope}%
\begin{pgfscope}%
\pgfpathrectangle{\pgfqpoint{0.188889in}{0.454405in}}{\pgfqpoint{4.650000in}{3.020000in}}%
\pgfusepath{clip}%
\pgfsetrectcap%
\pgfsetroundjoin%
\pgfsetlinewidth{0.803000pt}%
\definecolor{currentstroke}{rgb}{0.298039,0.298039,0.298039}%
\pgfsetstrokecolor{currentstroke}%
\pgfsetdash{}{0pt}%
\pgfpathmoveto{\pgfqpoint{3.792639in}{2.500978in}}%
\pgfpathlineto{\pgfqpoint{4.722639in}{2.500978in}}%
\pgfusepath{stroke}%
\end{pgfscope}%
\end{pgfpicture}%
\makeatother%
\endgroup%

%% file: fig/boxplot_MNIST_res.pgf
%% Creator: Matplotlib, PGF backend
%%
%% To include the figure in your LaTeX document, write
%%   \input{<filename>.pgf}
%%
%% Make sure the required packages are loaded in your preamble
%%   \usepackage{pgf}
%%
%% Figures using additional raster images can only be included by \input if
%% they are in the same directory as the main LaTeX file. For loading figures
%% from other directories you can use the `import` package
%% and then include the figures with
%%   \import{<path to file>}{<filename>.pgf}
%%
%% Matplotlib used the following preamble
%%
\begingroup%
\makeatletter%
\begin{pgfpicture}%
\pgfpathrectangle{\pgfpointorigin}{\pgfqpoint{4.888889in}{3.524405in}}%
\pgfusepath{use as bounding box, clip}%
\begin{pgfscope}%
\pgfsetbuttcap%
\pgfsetmiterjoin%
\definecolor{currentfill}{rgb}{1.000000,1.000000,1.000000}%
\pgfsetfillcolor{currentfill}%
\pgfsetlinewidth{0.000000pt}%
\definecolor{currentstroke}{rgb}{1.000000,1.000000,1.000000}%
\pgfsetstrokecolor{currentstroke}%
\pgfsetdash{}{0pt}%
\pgfpathmoveto{\pgfqpoint{0.000000in}{0.000000in}}%
\pgfpathlineto{\pgfqpoint{4.888889in}{0.000000in}}%
\pgfpathlineto{\pgfqpoint{4.888889in}{3.524405in}}%
\pgfpathlineto{\pgfqpoint{0.000000in}{3.524405in}}%
\pgfpathclose%
\pgfusepath{fill}%
\end{pgfscope}%
\begin{pgfscope}%
\pgfsetbuttcap%
\pgfsetmiterjoin%
\definecolor{currentfill}{rgb}{1.000000,1.000000,1.000000}%
\pgfsetfillcolor{currentfill}%
\pgfsetlinewidth{0.000000pt}%
\definecolor{currentstroke}{rgb}{0.000000,0.000000,0.000000}%
\pgfsetstrokecolor{currentstroke}%
\pgfsetstrokeopacity{0.000000}%
\pgfsetdash{}{0pt}%
\pgfpathmoveto{\pgfqpoint{0.188889in}{0.454405in}}%
\pgfpathlineto{\pgfqpoint{4.838889in}{0.454405in}}%
\pgfpathlineto{\pgfqpoint{4.838889in}{3.474405in}}%
\pgfpathlineto{\pgfqpoint{0.188889in}{3.474405in}}%
\pgfpathclose%
\pgfusepath{fill}%
\end{pgfscope}%
\begin{pgfscope}%
\definecolor{textcolor}{rgb}{0.000000,0.000000,0.000000}%
\pgfsetstrokecolor{textcolor}%
\pgfsetfillcolor{textcolor}%
\pgftext[x=0.770139in,y=0.357183in,,top]{\color{textcolor}\rmfamily\fontsize{24.200000}{29.040000}\selectfont \textsc{w-l}}%
\end{pgfscope}%
\begin{pgfscope}%
\definecolor{textcolor}{rgb}{0.000000,0.000000,0.000000}%
\pgfsetstrokecolor{textcolor}%
\pgfsetfillcolor{textcolor}%
\pgftext[x=1.932639in,y=0.357183in,,top]{\color{textcolor}\rmfamily\fontsize{24.200000}{29.040000}\selectfont \textsc{w-m}}%
\end{pgfscope}%
\begin{pgfscope}%
\definecolor{textcolor}{rgb}{0.000000,0.000000,0.000000}%
\pgfsetstrokecolor{textcolor}%
\pgfsetfillcolor{textcolor}%
\pgftext[x=3.095139in,y=0.357183in,,top]{\color{textcolor}\rmfamily\fontsize{24.200000}{29.040000}\selectfont \textsc{dcv}}%
\end{pgfscope}%
\begin{pgfscope}%
\definecolor{textcolor}{rgb}{0.000000,0.000000,0.000000}%
\pgfsetstrokecolor{textcolor}%
\pgfsetfillcolor{textcolor}%
\pgftext[x=4.257639in,y=0.357183in,,top]{\color{textcolor}\rmfamily\fontsize{24.200000}{29.040000}\selectfont \textsc{mc}}%
\end{pgfscope}%
\begin{pgfscope}%
\pgfpathrectangle{\pgfqpoint{0.188889in}{0.454405in}}{\pgfqpoint{4.650000in}{3.020000in}}%
\pgfusepath{clip}%
\pgfsetrectcap%
\pgfsetroundjoin%
\pgfsetlinewidth{2.509375pt}%
\definecolor{currentstroke}{rgb}{0.800000,0.800000,0.800000}%
\pgfsetstrokecolor{currentstroke}%
\pgfsetdash{}{0pt}%
\pgfpathmoveto{\pgfqpoint{0.188889in}{0.454405in}}%
\pgfpathlineto{\pgfqpoint{4.838889in}{0.454405in}}%
\pgfusepath{stroke}%
\end{pgfscope}%
\begin{pgfscope}%
\pgfsetbuttcap%
\pgfsetroundjoin%
\definecolor{currentfill}{rgb}{0.800000,0.800000,0.800000}%
\pgfsetfillcolor{currentfill}%
\pgfsetlinewidth{1.003750pt}%
\definecolor{currentstroke}{rgb}{0.800000,0.800000,0.800000}%
\pgfsetstrokecolor{currentstroke}%
\pgfsetdash{}{0pt}%
\pgfsys@defobject{currentmarker}{\pgfqpoint{-0.138889in}{0.000000in}}{\pgfqpoint{0.000000in}{0.000000in}}{%
\pgfpathmoveto{\pgfqpoint{0.000000in}{0.000000in}}%
\pgfpathlineto{\pgfqpoint{-0.138889in}{0.000000in}}%
\pgfusepath{stroke,fill}%
}%
\begin{pgfscope}%
\pgfsys@transformshift{0.188889in}{0.454405in}%
\pgfsys@useobject{currentmarker}{}%
\end{pgfscope}%
\end{pgfscope}%
\begin{pgfscope}%
\pgfpathrectangle{\pgfqpoint{0.188889in}{0.454405in}}{\pgfqpoint{4.650000in}{3.020000in}}%
\pgfusepath{clip}%
\pgfsetrectcap%
\pgfsetroundjoin%
\pgfsetlinewidth{2.509375pt}%
\definecolor{currentstroke}{rgb}{0.611765,0.611765,0.611765}%
\pgfsetstrokecolor{currentstroke}%
\pgfsetdash{}{0pt}%
\pgfpathmoveto{\pgfqpoint{0.188889in}{2.394902in}}%
\pgfpathlineto{\pgfqpoint{4.838889in}{2.394902in}}%
\pgfusepath{stroke}%
\end{pgfscope}%
\begin{pgfscope}%
\pgfsetbuttcap%
\pgfsetroundjoin%
\definecolor{currentfill}{rgb}{0.800000,0.800000,0.800000}%
\pgfsetfillcolor{currentfill}%
\pgfsetlinewidth{1.003750pt}%
\definecolor{currentstroke}{rgb}{0.800000,0.800000,0.800000}%
\pgfsetstrokecolor{currentstroke}%
\pgfsetdash{}{0pt}%
\pgfsys@defobject{currentmarker}{\pgfqpoint{-0.138889in}{0.000000in}}{\pgfqpoint{0.000000in}{0.000000in}}{%
\pgfpathmoveto{\pgfqpoint{0.000000in}{0.000000in}}%
\pgfpathlineto{\pgfqpoint{-0.138889in}{0.000000in}}%
\pgfusepath{stroke,fill}%
}%
\begin{pgfscope}%
\pgfsys@transformshift{0.188889in}{2.394902in}%
\pgfsys@useobject{currentmarker}{}%
\end{pgfscope}%
\end{pgfscope}%
\begin{pgfscope}%
\pgfpathrectangle{\pgfqpoint{0.188889in}{0.454405in}}{\pgfqpoint{4.650000in}{3.020000in}}%
\pgfusepath{clip}%
\pgfsetrectcap%
\pgfsetroundjoin%
\pgfsetlinewidth{0.401500pt}%
\definecolor{currentstroke}{rgb}{0.800000,0.800000,0.800000}%
\pgfsetstrokecolor{currentstroke}%
\pgfsetstrokeopacity{0.550000}%
\pgfsetdash{}{0pt}%
\pgfpathmoveto{\pgfqpoint{0.188889in}{1.038553in}}%
\pgfpathlineto{\pgfqpoint{4.838889in}{1.038553in}}%
\pgfusepath{stroke}%
\end{pgfscope}%
\begin{pgfscope}%
\pgfpathrectangle{\pgfqpoint{0.188889in}{0.454405in}}{\pgfqpoint{4.650000in}{3.020000in}}%
\pgfusepath{clip}%
\pgfsetrectcap%
\pgfsetroundjoin%
\pgfsetlinewidth{0.401500pt}%
\definecolor{currentstroke}{rgb}{0.800000,0.800000,0.800000}%
\pgfsetstrokecolor{currentstroke}%
\pgfsetstrokeopacity{0.550000}%
\pgfsetdash{}{0pt}%
\pgfpathmoveto{\pgfqpoint{0.188889in}{1.380257in}}%
\pgfpathlineto{\pgfqpoint{4.838889in}{1.380257in}}%
\pgfusepath{stroke}%
\end{pgfscope}%
\begin{pgfscope}%
\pgfpathrectangle{\pgfqpoint{0.188889in}{0.454405in}}{\pgfqpoint{4.650000in}{3.020000in}}%
\pgfusepath{clip}%
\pgfsetrectcap%
\pgfsetroundjoin%
\pgfsetlinewidth{0.401500pt}%
\definecolor{currentstroke}{rgb}{0.800000,0.800000,0.800000}%
\pgfsetstrokecolor{currentstroke}%
\pgfsetstrokeopacity{0.550000}%
\pgfsetdash{}{0pt}%
\pgfpathmoveto{\pgfqpoint{0.188889in}{1.622700in}}%
\pgfpathlineto{\pgfqpoint{4.838889in}{1.622700in}}%
\pgfusepath{stroke}%
\end{pgfscope}%
\begin{pgfscope}%
\pgfpathrectangle{\pgfqpoint{0.188889in}{0.454405in}}{\pgfqpoint{4.650000in}{3.020000in}}%
\pgfusepath{clip}%
\pgfsetrectcap%
\pgfsetroundjoin%
\pgfsetlinewidth{0.401500pt}%
\definecolor{currentstroke}{rgb}{0.800000,0.800000,0.800000}%
\pgfsetstrokecolor{currentstroke}%
\pgfsetstrokeopacity{0.550000}%
\pgfsetdash{}{0pt}%
\pgfpathmoveto{\pgfqpoint{0.188889in}{1.810754in}}%
\pgfpathlineto{\pgfqpoint{4.838889in}{1.810754in}}%
\pgfusepath{stroke}%
\end{pgfscope}%
\begin{pgfscope}%
\pgfpathrectangle{\pgfqpoint{0.188889in}{0.454405in}}{\pgfqpoint{4.650000in}{3.020000in}}%
\pgfusepath{clip}%
\pgfsetrectcap%
\pgfsetroundjoin%
\pgfsetlinewidth{0.401500pt}%
\definecolor{currentstroke}{rgb}{0.800000,0.800000,0.800000}%
\pgfsetstrokecolor{currentstroke}%
\pgfsetstrokeopacity{0.550000}%
\pgfsetdash{}{0pt}%
\pgfpathmoveto{\pgfqpoint{0.188889in}{1.964405in}}%
\pgfpathlineto{\pgfqpoint{4.838889in}{1.964405in}}%
\pgfusepath{stroke}%
\end{pgfscope}%
\begin{pgfscope}%
\pgfpathrectangle{\pgfqpoint{0.188889in}{0.454405in}}{\pgfqpoint{4.650000in}{3.020000in}}%
\pgfusepath{clip}%
\pgfsetrectcap%
\pgfsetroundjoin%
\pgfsetlinewidth{0.401500pt}%
\definecolor{currentstroke}{rgb}{0.800000,0.800000,0.800000}%
\pgfsetstrokecolor{currentstroke}%
\pgfsetstrokeopacity{0.550000}%
\pgfsetdash{}{0pt}%
\pgfpathmoveto{\pgfqpoint{0.188889in}{2.094315in}}%
\pgfpathlineto{\pgfqpoint{4.838889in}{2.094315in}}%
\pgfusepath{stroke}%
\end{pgfscope}%
\begin{pgfscope}%
\pgfpathrectangle{\pgfqpoint{0.188889in}{0.454405in}}{\pgfqpoint{4.650000in}{3.020000in}}%
\pgfusepath{clip}%
\pgfsetrectcap%
\pgfsetroundjoin%
\pgfsetlinewidth{0.401500pt}%
\definecolor{currentstroke}{rgb}{0.800000,0.800000,0.800000}%
\pgfsetstrokecolor{currentstroke}%
\pgfsetstrokeopacity{0.550000}%
\pgfsetdash{}{0pt}%
\pgfpathmoveto{\pgfqpoint{0.188889in}{2.206848in}}%
\pgfpathlineto{\pgfqpoint{4.838889in}{2.206848in}}%
\pgfusepath{stroke}%
\end{pgfscope}%
\begin{pgfscope}%
\pgfpathrectangle{\pgfqpoint{0.188889in}{0.454405in}}{\pgfqpoint{4.650000in}{3.020000in}}%
\pgfusepath{clip}%
\pgfsetrectcap%
\pgfsetroundjoin%
\pgfsetlinewidth{0.401500pt}%
\definecolor{currentstroke}{rgb}{0.800000,0.800000,0.800000}%
\pgfsetstrokecolor{currentstroke}%
\pgfsetstrokeopacity{0.550000}%
\pgfsetdash{}{0pt}%
\pgfpathmoveto{\pgfqpoint{0.188889in}{2.306110in}}%
\pgfpathlineto{\pgfqpoint{4.838889in}{2.306110in}}%
\pgfusepath{stroke}%
\end{pgfscope}%
\begin{pgfscope}%
\pgfpathrectangle{\pgfqpoint{0.188889in}{0.454405in}}{\pgfqpoint{4.650000in}{3.020000in}}%
\pgfusepath{clip}%
\pgfsetrectcap%
\pgfsetroundjoin%
\pgfsetlinewidth{0.401500pt}%
\definecolor{currentstroke}{rgb}{0.800000,0.800000,0.800000}%
\pgfsetstrokecolor{currentstroke}%
\pgfsetstrokeopacity{0.550000}%
\pgfsetdash{}{0pt}%
\pgfpathmoveto{\pgfqpoint{0.188889in}{2.979050in}}%
\pgfpathlineto{\pgfqpoint{4.838889in}{2.979050in}}%
\pgfusepath{stroke}%
\end{pgfscope}%
\begin{pgfscope}%
\pgfpathrectangle{\pgfqpoint{0.188889in}{0.454405in}}{\pgfqpoint{4.650000in}{3.020000in}}%
\pgfusepath{clip}%
\pgfsetrectcap%
\pgfsetroundjoin%
\pgfsetlinewidth{0.401500pt}%
\definecolor{currentstroke}{rgb}{0.800000,0.800000,0.800000}%
\pgfsetstrokecolor{currentstroke}%
\pgfsetstrokeopacity{0.550000}%
\pgfsetdash{}{0pt}%
\pgfpathmoveto{\pgfqpoint{0.188889in}{3.320754in}}%
\pgfpathlineto{\pgfqpoint{4.838889in}{3.320754in}}%
\pgfusepath{stroke}%
\end{pgfscope}%
\begin{pgfscope}%
\pgfpathrectangle{\pgfqpoint{0.188889in}{0.454405in}}{\pgfqpoint{4.650000in}{3.020000in}}%
\pgfusepath{clip}%
\pgfsetbuttcap%
\pgfsetmiterjoin%
\definecolor{currentfill}{rgb}{0.347059,0.458824,0.641176}%
\pgfsetfillcolor{currentfill}%
\pgfsetlinewidth{0.803000pt}%
\definecolor{currentstroke}{rgb}{0.298039,0.298039,0.298039}%
\pgfsetstrokecolor{currentstroke}%
\pgfsetdash{}{0pt}%
\pgfpathmoveto{\pgfqpoint{0.305139in}{1.401934in}}%
\pgfpathlineto{\pgfqpoint{1.235139in}{1.401934in}}%
\pgfpathlineto{\pgfqpoint{1.235139in}{1.406617in}}%
\pgfpathlineto{\pgfqpoint{0.305139in}{1.406617in}}%
\pgfpathlineto{\pgfqpoint{0.305139in}{1.401934in}}%
\pgfpathclose%
\pgfusepath{stroke,fill}%
\end{pgfscope}%
\begin{pgfscope}%
\pgfpathrectangle{\pgfqpoint{0.188889in}{0.454405in}}{\pgfqpoint{4.650000in}{3.020000in}}%
\pgfusepath{clip}%
\pgfsetbuttcap%
\pgfsetmiterjoin%
\definecolor{currentfill}{rgb}{0.798529,0.536765,0.389706}%
\pgfsetfillcolor{currentfill}%
\pgfsetlinewidth{0.803000pt}%
\definecolor{currentstroke}{rgb}{0.298039,0.298039,0.298039}%
\pgfsetstrokecolor{currentstroke}%
\pgfsetdash{}{0pt}%
\pgfpathmoveto{\pgfqpoint{1.467639in}{1.349093in}}%
\pgfpathlineto{\pgfqpoint{2.397639in}{1.349093in}}%
\pgfpathlineto{\pgfqpoint{2.397639in}{1.427445in}}%
\pgfpathlineto{\pgfqpoint{1.467639in}{1.427445in}}%
\pgfpathlineto{\pgfqpoint{1.467639in}{1.349093in}}%
\pgfpathclose%
\pgfusepath{stroke,fill}%
\end{pgfscope}%
\begin{pgfscope}%
\pgfpathrectangle{\pgfqpoint{0.188889in}{0.454405in}}{\pgfqpoint{4.650000in}{3.020000in}}%
\pgfusepath{clip}%
\pgfsetbuttcap%
\pgfsetmiterjoin%
\definecolor{currentfill}{rgb}{0.374020,0.618137,0.429902}%
\pgfsetfillcolor{currentfill}%
\pgfsetlinewidth{0.803000pt}%
\definecolor{currentstroke}{rgb}{0.298039,0.298039,0.298039}%
\pgfsetstrokecolor{currentstroke}%
\pgfsetdash{}{0pt}%
\pgfpathmoveto{\pgfqpoint{2.630139in}{1.270572in}}%
\pgfpathlineto{\pgfqpoint{3.560139in}{1.270572in}}%
\pgfpathlineto{\pgfqpoint{3.560139in}{1.371041in}}%
\pgfpathlineto{\pgfqpoint{2.630139in}{1.371041in}}%
\pgfpathlineto{\pgfqpoint{2.630139in}{1.270572in}}%
\pgfpathclose%
\pgfusepath{stroke,fill}%
\end{pgfscope}%
\begin{pgfscope}%
\pgfpathrectangle{\pgfqpoint{0.188889in}{0.454405in}}{\pgfqpoint{4.650000in}{3.020000in}}%
\pgfusepath{clip}%
\pgfsetbuttcap%
\pgfsetmiterjoin%
\definecolor{currentfill}{rgb}{0.710784,0.363725,0.375490}%
\pgfsetfillcolor{currentfill}%
\pgfsetlinewidth{0.803000pt}%
\definecolor{currentstroke}{rgb}{0.298039,0.298039,0.298039}%
\pgfsetstrokecolor{currentstroke}%
\pgfsetdash{}{0pt}%
\pgfpathmoveto{\pgfqpoint{3.792639in}{2.345311in}}%
\pgfpathlineto{\pgfqpoint{4.722639in}{2.345311in}}%
\pgfpathlineto{\pgfqpoint{4.722639in}{2.431531in}}%
\pgfpathlineto{\pgfqpoint{3.792639in}{2.431531in}}%
\pgfpathlineto{\pgfqpoint{3.792639in}{2.345311in}}%
\pgfpathclose%
\pgfusepath{stroke,fill}%
\end{pgfscope}%
\begin{pgfscope}%
\pgfsetrectcap%
\pgfsetmiterjoin%
\pgfsetlinewidth{1.254687pt}%
\definecolor{currentstroke}{rgb}{0.800000,0.800000,0.800000}%
\pgfsetstrokecolor{currentstroke}%
\pgfsetdash{}{0pt}%
\pgfpathmoveto{\pgfqpoint{0.188889in}{0.454405in}}%
\pgfpathlineto{\pgfqpoint{0.188889in}{3.474405in}}%
\pgfusepath{stroke}%
\end{pgfscope}%
\begin{pgfscope}%
\pgfsetrectcap%
\pgfsetmiterjoin%
\pgfsetlinewidth{1.254687pt}%
\definecolor{currentstroke}{rgb}{0.800000,0.800000,0.800000}%
\pgfsetstrokecolor{currentstroke}%
\pgfsetdash{}{0pt}%
\pgfpathmoveto{\pgfqpoint{4.838889in}{0.454405in}}%
\pgfpathlineto{\pgfqpoint{4.838889in}{3.474405in}}%
\pgfusepath{stroke}%
\end{pgfscope}%
\begin{pgfscope}%
\pgfsetrectcap%
\pgfsetmiterjoin%
\pgfsetlinewidth{1.254687pt}%
\definecolor{currentstroke}{rgb}{0.800000,0.800000,0.800000}%
\pgfsetstrokecolor{currentstroke}%
\pgfsetdash{}{0pt}%
\pgfpathmoveto{\pgfqpoint{0.188889in}{0.454405in}}%
\pgfpathlineto{\pgfqpoint{4.838889in}{0.454405in}}%
\pgfusepath{stroke}%
\end{pgfscope}%
\begin{pgfscope}%
\pgfsetrectcap%
\pgfsetmiterjoin%
\pgfsetlinewidth{1.254687pt}%
\definecolor{currentstroke}{rgb}{0.800000,0.800000,0.800000}%
\pgfsetstrokecolor{currentstroke}%
\pgfsetdash{}{0pt}%
\pgfpathmoveto{\pgfqpoint{0.188889in}{3.474405in}}%
\pgfpathlineto{\pgfqpoint{4.838889in}{3.474405in}}%
\pgfusepath{stroke}%
\end{pgfscope}%
\begin{pgfscope}%
\pgfpathrectangle{\pgfqpoint{0.188889in}{0.454405in}}{\pgfqpoint{4.650000in}{3.020000in}}%
\pgfusepath{clip}%
\pgfsetrectcap%
\pgfsetroundjoin%
\pgfsetlinewidth{0.803000pt}%
\definecolor{currentstroke}{rgb}{0.298039,0.298039,0.298039}%
\pgfsetstrokecolor{currentstroke}%
\pgfsetdash{}{0pt}%
\pgfpathmoveto{\pgfqpoint{0.770139in}{1.401934in}}%
\pgfpathlineto{\pgfqpoint{0.770139in}{1.399248in}}%
\pgfusepath{stroke}%
\end{pgfscope}%
\begin{pgfscope}%
\pgfpathrectangle{\pgfqpoint{0.188889in}{0.454405in}}{\pgfqpoint{4.650000in}{3.020000in}}%
\pgfusepath{clip}%
\pgfsetrectcap%
\pgfsetroundjoin%
\pgfsetlinewidth{0.803000pt}%
\definecolor{currentstroke}{rgb}{0.298039,0.298039,0.298039}%
\pgfsetstrokecolor{currentstroke}%
\pgfsetdash{}{0pt}%
\pgfpathmoveto{\pgfqpoint{0.770139in}{1.406617in}}%
\pgfpathlineto{\pgfqpoint{0.770139in}{1.406617in}}%
\pgfusepath{stroke}%
\end{pgfscope}%
\begin{pgfscope}%
\pgfpathrectangle{\pgfqpoint{0.188889in}{0.454405in}}{\pgfqpoint{4.650000in}{3.020000in}}%
\pgfusepath{clip}%
\pgfsetrectcap%
\pgfsetroundjoin%
\pgfsetlinewidth{0.803000pt}%
\definecolor{currentstroke}{rgb}{0.298039,0.298039,0.298039}%
\pgfsetstrokecolor{currentstroke}%
\pgfsetdash{}{0pt}%
\pgfpathmoveto{\pgfqpoint{0.537639in}{1.399248in}}%
\pgfpathlineto{\pgfqpoint{1.002639in}{1.399248in}}%
\pgfusepath{stroke}%
\end{pgfscope}%
\begin{pgfscope}%
\pgfpathrectangle{\pgfqpoint{0.188889in}{0.454405in}}{\pgfqpoint{4.650000in}{3.020000in}}%
\pgfusepath{clip}%
\pgfsetrectcap%
\pgfsetroundjoin%
\pgfsetlinewidth{0.803000pt}%
\definecolor{currentstroke}{rgb}{0.298039,0.298039,0.298039}%
\pgfsetstrokecolor{currentstroke}%
\pgfsetdash{}{0pt}%
\pgfpathmoveto{\pgfqpoint{0.537639in}{1.406617in}}%
\pgfpathlineto{\pgfqpoint{1.002639in}{1.406617in}}%
\pgfusepath{stroke}%
\end{pgfscope}%
\begin{pgfscope}%
\pgfpathrectangle{\pgfqpoint{0.188889in}{0.454405in}}{\pgfqpoint{4.650000in}{3.020000in}}%
\pgfusepath{clip}%
\pgfsetrectcap%
\pgfsetroundjoin%
\pgfsetlinewidth{0.803000pt}%
\definecolor{currentstroke}{rgb}{0.298039,0.298039,0.298039}%
\pgfsetstrokecolor{currentstroke}%
\pgfsetdash{}{0pt}%
\pgfpathmoveto{\pgfqpoint{0.305139in}{1.402828in}}%
\pgfpathlineto{\pgfqpoint{1.235139in}{1.402828in}}%
\pgfusepath{stroke}%
\end{pgfscope}%
\begin{pgfscope}%
\pgfpathrectangle{\pgfqpoint{0.188889in}{0.454405in}}{\pgfqpoint{4.650000in}{3.020000in}}%
\pgfusepath{clip}%
\pgfsetbuttcap%
\pgfsetmiterjoin%
\definecolor{currentfill}{rgb}{0.298039,0.298039,0.298039}%
\pgfsetfillcolor{currentfill}%
\pgfsetlinewidth{1.003750pt}%
\definecolor{currentstroke}{rgb}{0.298039,0.298039,0.298039}%
\pgfsetstrokecolor{currentstroke}%
\pgfsetdash{}{0pt}%
\pgfsys@defobject{currentmarker}{\pgfqpoint{-0.029463in}{-0.049105in}}{\pgfqpoint{0.029463in}{0.049105in}}{%
\pgfpathmoveto{\pgfqpoint{0.000000in}{-0.049105in}}%
\pgfpathlineto{\pgfqpoint{0.029463in}{0.000000in}}%
\pgfpathlineto{\pgfqpoint{0.000000in}{0.049105in}}%
\pgfpathlineto{\pgfqpoint{-0.029463in}{0.000000in}}%
\pgfpathclose%
\pgfusepath{stroke,fill}%
}%
\begin{pgfscope}%
\pgfsys@transformshift{0.770139in}{1.292873in}%
\pgfsys@useobject{currentmarker}{}%
\end{pgfscope}%
\begin{pgfscope}%
\pgfsys@transformshift{0.770139in}{1.346402in}%
\pgfsys@useobject{currentmarker}{}%
\end{pgfscope}%
\begin{pgfscope}%
\pgfsys@transformshift{0.770139in}{1.390391in}%
\pgfsys@useobject{currentmarker}{}%
\end{pgfscope}%
\begin{pgfscope}%
\pgfsys@transformshift{0.770139in}{1.417884in}%
\pgfsys@useobject{currentmarker}{}%
\end{pgfscope}%
\begin{pgfscope}%
\pgfsys@transformshift{0.770139in}{1.469327in}%
\pgfsys@useobject{currentmarker}{}%
\end{pgfscope}%
\begin{pgfscope}%
\pgfsys@transformshift{0.770139in}{1.469327in}%
\pgfsys@useobject{currentmarker}{}%
\end{pgfscope}%
\begin{pgfscope}%
\pgfsys@transformshift{0.770139in}{1.497500in}%
\pgfsys@useobject{currentmarker}{}%
\end{pgfscope}%
\end{pgfscope}%
\begin{pgfscope}%
\pgfpathrectangle{\pgfqpoint{0.188889in}{0.454405in}}{\pgfqpoint{4.650000in}{3.020000in}}%
\pgfusepath{clip}%
\pgfsetrectcap%
\pgfsetroundjoin%
\pgfsetlinewidth{0.803000pt}%
\definecolor{currentstroke}{rgb}{0.298039,0.298039,0.298039}%
\pgfsetstrokecolor{currentstroke}%
\pgfsetdash{}{0pt}%
\pgfpathmoveto{\pgfqpoint{1.932639in}{1.349093in}}%
\pgfpathlineto{\pgfqpoint{1.932639in}{1.311988in}}%
\pgfusepath{stroke}%
\end{pgfscope}%
\begin{pgfscope}%
\pgfpathrectangle{\pgfqpoint{0.188889in}{0.454405in}}{\pgfqpoint{4.650000in}{3.020000in}}%
\pgfusepath{clip}%
\pgfsetrectcap%
\pgfsetroundjoin%
\pgfsetlinewidth{0.803000pt}%
\definecolor{currentstroke}{rgb}{0.298039,0.298039,0.298039}%
\pgfsetstrokecolor{currentstroke}%
\pgfsetdash{}{0pt}%
\pgfpathmoveto{\pgfqpoint{1.932639in}{1.427445in}}%
\pgfpathlineto{\pgfqpoint{1.932639in}{1.445224in}}%
\pgfusepath{stroke}%
\end{pgfscope}%
\begin{pgfscope}%
\pgfpathrectangle{\pgfqpoint{0.188889in}{0.454405in}}{\pgfqpoint{4.650000in}{3.020000in}}%
\pgfusepath{clip}%
\pgfsetrectcap%
\pgfsetroundjoin%
\pgfsetlinewidth{0.803000pt}%
\definecolor{currentstroke}{rgb}{0.298039,0.298039,0.298039}%
\pgfsetstrokecolor{currentstroke}%
\pgfsetdash{}{0pt}%
\pgfpathmoveto{\pgfqpoint{1.700139in}{1.311988in}}%
\pgfpathlineto{\pgfqpoint{2.165139in}{1.311988in}}%
\pgfusepath{stroke}%
\end{pgfscope}%
\begin{pgfscope}%
\pgfpathrectangle{\pgfqpoint{0.188889in}{0.454405in}}{\pgfqpoint{4.650000in}{3.020000in}}%
\pgfusepath{clip}%
\pgfsetrectcap%
\pgfsetroundjoin%
\pgfsetlinewidth{0.803000pt}%
\definecolor{currentstroke}{rgb}{0.298039,0.298039,0.298039}%
\pgfsetstrokecolor{currentstroke}%
\pgfsetdash{}{0pt}%
\pgfpathmoveto{\pgfqpoint{1.700139in}{1.445224in}}%
\pgfpathlineto{\pgfqpoint{2.165139in}{1.445224in}}%
\pgfusepath{stroke}%
\end{pgfscope}%
\begin{pgfscope}%
\pgfpathrectangle{\pgfqpoint{0.188889in}{0.454405in}}{\pgfqpoint{4.650000in}{3.020000in}}%
\pgfusepath{clip}%
\pgfsetrectcap%
\pgfsetroundjoin%
\pgfsetlinewidth{0.803000pt}%
\definecolor{currentstroke}{rgb}{0.298039,0.298039,0.298039}%
\pgfsetstrokecolor{currentstroke}%
\pgfsetdash{}{0pt}%
\pgfpathmoveto{\pgfqpoint{1.467639in}{1.376403in}}%
\pgfpathlineto{\pgfqpoint{2.397639in}{1.376403in}}%
\pgfusepath{stroke}%
\end{pgfscope}%
\begin{pgfscope}%
\pgfpathrectangle{\pgfqpoint{0.188889in}{0.454405in}}{\pgfqpoint{4.650000in}{3.020000in}}%
\pgfusepath{clip}%
\pgfsetrectcap%
\pgfsetroundjoin%
\pgfsetlinewidth{0.803000pt}%
\definecolor{currentstroke}{rgb}{0.298039,0.298039,0.298039}%
\pgfsetstrokecolor{currentstroke}%
\pgfsetdash{}{0pt}%
\pgfpathmoveto{\pgfqpoint{3.095139in}{1.270572in}}%
\pgfpathlineto{\pgfqpoint{3.095139in}{1.189707in}}%
\pgfusepath{stroke}%
\end{pgfscope}%
\begin{pgfscope}%
\pgfpathrectangle{\pgfqpoint{0.188889in}{0.454405in}}{\pgfqpoint{4.650000in}{3.020000in}}%
\pgfusepath{clip}%
\pgfsetrectcap%
\pgfsetroundjoin%
\pgfsetlinewidth{0.803000pt}%
\definecolor{currentstroke}{rgb}{0.298039,0.298039,0.298039}%
\pgfsetstrokecolor{currentstroke}%
\pgfsetdash{}{0pt}%
\pgfpathmoveto{\pgfqpoint{3.095139in}{1.371041in}}%
\pgfpathlineto{\pgfqpoint{3.095139in}{1.430331in}}%
\pgfusepath{stroke}%
\end{pgfscope}%
\begin{pgfscope}%
\pgfpathrectangle{\pgfqpoint{0.188889in}{0.454405in}}{\pgfqpoint{4.650000in}{3.020000in}}%
\pgfusepath{clip}%
\pgfsetrectcap%
\pgfsetroundjoin%
\pgfsetlinewidth{0.803000pt}%
\definecolor{currentstroke}{rgb}{0.298039,0.298039,0.298039}%
\pgfsetstrokecolor{currentstroke}%
\pgfsetdash{}{0pt}%
\pgfpathmoveto{\pgfqpoint{2.862639in}{1.189707in}}%
\pgfpathlineto{\pgfqpoint{3.327639in}{1.189707in}}%
\pgfusepath{stroke}%
\end{pgfscope}%
\begin{pgfscope}%
\pgfpathrectangle{\pgfqpoint{0.188889in}{0.454405in}}{\pgfqpoint{4.650000in}{3.020000in}}%
\pgfusepath{clip}%
\pgfsetrectcap%
\pgfsetroundjoin%
\pgfsetlinewidth{0.803000pt}%
\definecolor{currentstroke}{rgb}{0.298039,0.298039,0.298039}%
\pgfsetstrokecolor{currentstroke}%
\pgfsetdash{}{0pt}%
\pgfpathmoveto{\pgfqpoint{2.862639in}{1.430331in}}%
\pgfpathlineto{\pgfqpoint{3.327639in}{1.430331in}}%
\pgfusepath{stroke}%
\end{pgfscope}%
\begin{pgfscope}%
\pgfpathrectangle{\pgfqpoint{0.188889in}{0.454405in}}{\pgfqpoint{4.650000in}{3.020000in}}%
\pgfusepath{clip}%
\pgfsetrectcap%
\pgfsetroundjoin%
\pgfsetlinewidth{0.803000pt}%
\definecolor{currentstroke}{rgb}{0.298039,0.298039,0.298039}%
\pgfsetstrokecolor{currentstroke}%
\pgfsetdash{}{0pt}%
\pgfpathmoveto{\pgfqpoint{2.630139in}{1.308954in}}%
\pgfpathlineto{\pgfqpoint{3.560139in}{1.308954in}}%
\pgfusepath{stroke}%
\end{pgfscope}%
\begin{pgfscope}%
\pgfpathrectangle{\pgfqpoint{0.188889in}{0.454405in}}{\pgfqpoint{4.650000in}{3.020000in}}%
\pgfusepath{clip}%
\pgfsetbuttcap%
\pgfsetmiterjoin%
\definecolor{currentfill}{rgb}{0.298039,0.298039,0.298039}%
\pgfsetfillcolor{currentfill}%
\pgfsetlinewidth{1.003750pt}%
\definecolor{currentstroke}{rgb}{0.298039,0.298039,0.298039}%
\pgfsetstrokecolor{currentstroke}%
\pgfsetdash{}{0pt}%
\pgfsys@defobject{currentmarker}{\pgfqpoint{-0.029463in}{-0.049105in}}{\pgfqpoint{0.029463in}{0.049105in}}{%
\pgfpathmoveto{\pgfqpoint{0.000000in}{-0.049105in}}%
\pgfpathlineto{\pgfqpoint{0.029463in}{0.000000in}}%
\pgfpathlineto{\pgfqpoint{0.000000in}{0.049105in}}%
\pgfpathlineto{\pgfqpoint{-0.029463in}{0.000000in}}%
\pgfpathclose%
\pgfusepath{stroke,fill}%
}%
\begin{pgfscope}%
\pgfsys@transformshift{3.095139in}{1.583861in}%
\pgfsys@useobject{currentmarker}{}%
\end{pgfscope}%
\end{pgfscope}%
\begin{pgfscope}%
\pgfpathrectangle{\pgfqpoint{0.188889in}{0.454405in}}{\pgfqpoint{4.650000in}{3.020000in}}%
\pgfusepath{clip}%
\pgfsetrectcap%
\pgfsetroundjoin%
\pgfsetlinewidth{0.803000pt}%
\definecolor{currentstroke}{rgb}{0.298039,0.298039,0.298039}%
\pgfsetstrokecolor{currentstroke}%
\pgfsetdash{}{0pt}%
\pgfpathmoveto{\pgfqpoint{4.257639in}{2.345311in}}%
\pgfpathlineto{\pgfqpoint{4.257639in}{2.245047in}}%
\pgfusepath{stroke}%
\end{pgfscope}%
\begin{pgfscope}%
\pgfpathrectangle{\pgfqpoint{0.188889in}{0.454405in}}{\pgfqpoint{4.650000in}{3.020000in}}%
\pgfusepath{clip}%
\pgfsetrectcap%
\pgfsetroundjoin%
\pgfsetlinewidth{0.803000pt}%
\definecolor{currentstroke}{rgb}{0.298039,0.298039,0.298039}%
\pgfsetstrokecolor{currentstroke}%
\pgfsetdash{}{0pt}%
\pgfpathmoveto{\pgfqpoint{4.257639in}{2.431531in}}%
\pgfpathlineto{\pgfqpoint{4.257639in}{2.541173in}}%
\pgfusepath{stroke}%
\end{pgfscope}%
\begin{pgfscope}%
\pgfpathrectangle{\pgfqpoint{0.188889in}{0.454405in}}{\pgfqpoint{4.650000in}{3.020000in}}%
\pgfusepath{clip}%
\pgfsetrectcap%
\pgfsetroundjoin%
\pgfsetlinewidth{0.803000pt}%
\definecolor{currentstroke}{rgb}{0.298039,0.298039,0.298039}%
\pgfsetstrokecolor{currentstroke}%
\pgfsetdash{}{0pt}%
\pgfpathmoveto{\pgfqpoint{4.025139in}{2.245047in}}%
\pgfpathlineto{\pgfqpoint{4.490139in}{2.245047in}}%
\pgfusepath{stroke}%
\end{pgfscope}%
\begin{pgfscope}%
\pgfpathrectangle{\pgfqpoint{0.188889in}{0.454405in}}{\pgfqpoint{4.650000in}{3.020000in}}%
\pgfusepath{clip}%
\pgfsetrectcap%
\pgfsetroundjoin%
\pgfsetlinewidth{0.803000pt}%
\definecolor{currentstroke}{rgb}{0.298039,0.298039,0.298039}%
\pgfsetstrokecolor{currentstroke}%
\pgfsetdash{}{0pt}%
\pgfpathmoveto{\pgfqpoint{4.025139in}{2.541173in}}%
\pgfpathlineto{\pgfqpoint{4.490139in}{2.541173in}}%
\pgfusepath{stroke}%
\end{pgfscope}%
\begin{pgfscope}%
\pgfpathrectangle{\pgfqpoint{0.188889in}{0.454405in}}{\pgfqpoint{4.650000in}{3.020000in}}%
\pgfusepath{clip}%
\pgfsetrectcap%
\pgfsetroundjoin%
\pgfsetlinewidth{0.803000pt}%
\definecolor{currentstroke}{rgb}{0.298039,0.298039,0.298039}%
\pgfsetstrokecolor{currentstroke}%
\pgfsetdash{}{0pt}%
\pgfpathmoveto{\pgfqpoint{3.792639in}{2.383495in}}%
\pgfpathlineto{\pgfqpoint{4.722639in}{2.383495in}}%
\pgfusepath{stroke}%
\end{pgfscope}%
\begin{pgfscope}%
\pgfpathrectangle{\pgfqpoint{0.188889in}{0.454405in}}{\pgfqpoint{4.650000in}{3.020000in}}%
\pgfusepath{clip}%
\pgfsetbuttcap%
\pgfsetmiterjoin%
\definecolor{currentfill}{rgb}{0.298039,0.298039,0.298039}%
\pgfsetfillcolor{currentfill}%
\pgfsetlinewidth{1.003750pt}%
\definecolor{currentstroke}{rgb}{0.298039,0.298039,0.298039}%
\pgfsetstrokecolor{currentstroke}%
\pgfsetdash{}{0pt}%
\pgfsys@defobject{currentmarker}{\pgfqpoint{-0.029463in}{-0.049105in}}{\pgfqpoint{0.029463in}{0.049105in}}{%
\pgfpathmoveto{\pgfqpoint{0.000000in}{-0.049105in}}%
\pgfpathlineto{\pgfqpoint{0.029463in}{0.000000in}}%
\pgfpathlineto{\pgfqpoint{0.000000in}{0.049105in}}%
\pgfpathlineto{\pgfqpoint{-0.029463in}{0.000000in}}%
\pgfpathclose%
\pgfusepath{stroke,fill}%
}%
\begin{pgfscope}%
\pgfsys@transformshift{4.257639in}{2.607720in}%
\pgfsys@useobject{currentmarker}{}%
\end{pgfscope}%
\end{pgfscope}%
\end{pgfpicture}%
\makeatother%
\endgroup%

%% file: fig/boxplot_ADK_res.pgf
%% Creator: Matplotlib, PGF backend
%%
%% To include the figure in your LaTeX document, write
%%   \input{<filename>.pgf}
%%
%% Make sure the required packages are loaded in your preamble
%%   \usepackage{pgf}
%%
%% Figures using additional raster images can only be included by \input if
%% they are in the same directory as the main LaTeX file. For loading figures
%% from other directories you can use the `import` package
%% and then include the figures with
%%   \import{<path to file>}{<filename>.pgf}
%%
%% Matplotlib used the following preamble
%%
\begingroup%
\makeatletter%
\begin{pgfpicture}%
\pgfpathrectangle{\pgfpointorigin}{\pgfqpoint{4.888889in}{3.524405in}}%
\pgfusepath{use as bounding box, clip}%
\begin{pgfscope}%
\pgfsetbuttcap%
\pgfsetmiterjoin%
\definecolor{currentfill}{rgb}{1.000000,1.000000,1.000000}%
\pgfsetfillcolor{currentfill}%
\pgfsetlinewidth{0.000000pt}%
\definecolor{currentstroke}{rgb}{1.000000,1.000000,1.000000}%
\pgfsetstrokecolor{currentstroke}%
\pgfsetdash{}{0pt}%
\pgfpathmoveto{\pgfqpoint{0.000000in}{0.000000in}}%
\pgfpathlineto{\pgfqpoint{4.888889in}{0.000000in}}%
\pgfpathlineto{\pgfqpoint{4.888889in}{3.524405in}}%
\pgfpathlineto{\pgfqpoint{0.000000in}{3.524405in}}%
\pgfpathclose%
\pgfusepath{fill}%
\end{pgfscope}%
\begin{pgfscope}%
\pgfsetbuttcap%
\pgfsetmiterjoin%
\definecolor{currentfill}{rgb}{1.000000,1.000000,1.000000}%
\pgfsetfillcolor{currentfill}%
\pgfsetlinewidth{0.000000pt}%
\definecolor{currentstroke}{rgb}{0.000000,0.000000,0.000000}%
\pgfsetstrokecolor{currentstroke}%
\pgfsetstrokeopacity{0.000000}%
\pgfsetdash{}{0pt}%
\pgfpathmoveto{\pgfqpoint{0.188889in}{0.454405in}}%
\pgfpathlineto{\pgfqpoint{4.838889in}{0.454405in}}%
\pgfpathlineto{\pgfqpoint{4.838889in}{3.474405in}}%
\pgfpathlineto{\pgfqpoint{0.188889in}{3.474405in}}%
\pgfpathclose%
\pgfusepath{fill}%
\end{pgfscope}%
\begin{pgfscope}%
\definecolor{textcolor}{rgb}{0.000000,0.000000,0.000000}%
\pgfsetstrokecolor{textcolor}%
\pgfsetfillcolor{textcolor}%
\pgftext[x=0.770139in,y=0.357183in,,top]{\color{textcolor}\rmfamily\fontsize{24.200000}{29.040000}\selectfont \textsc{w-l}}%
\end{pgfscope}%
\begin{pgfscope}%
\definecolor{textcolor}{rgb}{0.000000,0.000000,0.000000}%
\pgfsetstrokecolor{textcolor}%
\pgfsetfillcolor{textcolor}%
\pgftext[x=1.932639in,y=0.357183in,,top]{\color{textcolor}\rmfamily\fontsize{24.200000}{29.040000}\selectfont \textsc{w-m}}%
\end{pgfscope}%
\begin{pgfscope}%
\definecolor{textcolor}{rgb}{0.000000,0.000000,0.000000}%
\pgfsetstrokecolor{textcolor}%
\pgfsetfillcolor{textcolor}%
\pgftext[x=3.095139in,y=0.357183in,,top]{\color{textcolor}\rmfamily\fontsize{24.200000}{29.040000}\selectfont \textsc{dcv}}%
\end{pgfscope}%
\begin{pgfscope}%
\definecolor{textcolor}{rgb}{0.000000,0.000000,0.000000}%
\pgfsetstrokecolor{textcolor}%
\pgfsetfillcolor{textcolor}%
\pgftext[x=4.257639in,y=0.357183in,,top]{\color{textcolor}\rmfamily\fontsize{24.200000}{29.040000}\selectfont \textsc{mc}}%
\end{pgfscope}%
\begin{pgfscope}%
\pgfpathrectangle{\pgfqpoint{0.188889in}{0.454405in}}{\pgfqpoint{4.650000in}{3.020000in}}%
\pgfusepath{clip}%
\pgfsetrectcap%
\pgfsetroundjoin%
\pgfsetlinewidth{2.509375pt}%
\definecolor{currentstroke}{rgb}{0.800000,0.800000,0.800000}%
\pgfsetstrokecolor{currentstroke}%
\pgfsetdash{}{0pt}%
\pgfpathmoveto{\pgfqpoint{0.188889in}{0.454405in}}%
\pgfpathlineto{\pgfqpoint{4.838889in}{0.454405in}}%
\pgfusepath{stroke}%
\end{pgfscope}%
\begin{pgfscope}%
\pgfsetbuttcap%
\pgfsetroundjoin%
\definecolor{currentfill}{rgb}{0.800000,0.800000,0.800000}%
\pgfsetfillcolor{currentfill}%
\pgfsetlinewidth{1.003750pt}%
\definecolor{currentstroke}{rgb}{0.800000,0.800000,0.800000}%
\pgfsetstrokecolor{currentstroke}%
\pgfsetdash{}{0pt}%
\pgfsys@defobject{currentmarker}{\pgfqpoint{-0.138889in}{0.000000in}}{\pgfqpoint{0.000000in}{0.000000in}}{%
\pgfpathmoveto{\pgfqpoint{0.000000in}{0.000000in}}%
\pgfpathlineto{\pgfqpoint{-0.138889in}{0.000000in}}%
\pgfusepath{stroke,fill}%
}%
\begin{pgfscope}%
\pgfsys@transformshift{0.188889in}{0.454405in}%
\pgfsys@useobject{currentmarker}{}%
\end{pgfscope}%
\end{pgfscope}%
\begin{pgfscope}%
\pgfpathrectangle{\pgfqpoint{0.188889in}{0.454405in}}{\pgfqpoint{4.650000in}{3.020000in}}%
\pgfusepath{clip}%
\pgfsetrectcap%
\pgfsetroundjoin%
\pgfsetlinewidth{2.509375pt}%
\definecolor{currentstroke}{rgb}{0.611765,0.611765,0.611765}%
\pgfsetstrokecolor{currentstroke}%
\pgfsetdash{}{0pt}%
\pgfpathmoveto{\pgfqpoint{0.188889in}{2.394902in}}%
\pgfpathlineto{\pgfqpoint{4.838889in}{2.394902in}}%
\pgfusepath{stroke}%
\end{pgfscope}%
\begin{pgfscope}%
\pgfsetbuttcap%
\pgfsetroundjoin%
\definecolor{currentfill}{rgb}{0.800000,0.800000,0.800000}%
\pgfsetfillcolor{currentfill}%
\pgfsetlinewidth{1.003750pt}%
\definecolor{currentstroke}{rgb}{0.800000,0.800000,0.800000}%
\pgfsetstrokecolor{currentstroke}%
\pgfsetdash{}{0pt}%
\pgfsys@defobject{currentmarker}{\pgfqpoint{-0.138889in}{0.000000in}}{\pgfqpoint{0.000000in}{0.000000in}}{%
\pgfpathmoveto{\pgfqpoint{0.000000in}{0.000000in}}%
\pgfpathlineto{\pgfqpoint{-0.138889in}{0.000000in}}%
\pgfusepath{stroke,fill}%
}%
\begin{pgfscope}%
\pgfsys@transformshift{0.188889in}{2.394902in}%
\pgfsys@useobject{currentmarker}{}%
\end{pgfscope}%
\end{pgfscope}%
\begin{pgfscope}%
\pgfpathrectangle{\pgfqpoint{0.188889in}{0.454405in}}{\pgfqpoint{4.650000in}{3.020000in}}%
\pgfusepath{clip}%
\pgfsetrectcap%
\pgfsetroundjoin%
\pgfsetlinewidth{0.401500pt}%
\definecolor{currentstroke}{rgb}{0.800000,0.800000,0.800000}%
\pgfsetstrokecolor{currentstroke}%
\pgfsetstrokeopacity{0.550000}%
\pgfsetdash{}{0pt}%
\pgfpathmoveto{\pgfqpoint{0.188889in}{1.038553in}}%
\pgfpathlineto{\pgfqpoint{4.838889in}{1.038553in}}%
\pgfusepath{stroke}%
\end{pgfscope}%
\begin{pgfscope}%
\pgfpathrectangle{\pgfqpoint{0.188889in}{0.454405in}}{\pgfqpoint{4.650000in}{3.020000in}}%
\pgfusepath{clip}%
\pgfsetrectcap%
\pgfsetroundjoin%
\pgfsetlinewidth{0.401500pt}%
\definecolor{currentstroke}{rgb}{0.800000,0.800000,0.800000}%
\pgfsetstrokecolor{currentstroke}%
\pgfsetstrokeopacity{0.550000}%
\pgfsetdash{}{0pt}%
\pgfpathmoveto{\pgfqpoint{0.188889in}{1.380257in}}%
\pgfpathlineto{\pgfqpoint{4.838889in}{1.380257in}}%
\pgfusepath{stroke}%
\end{pgfscope}%
\begin{pgfscope}%
\pgfpathrectangle{\pgfqpoint{0.188889in}{0.454405in}}{\pgfqpoint{4.650000in}{3.020000in}}%
\pgfusepath{clip}%
\pgfsetrectcap%
\pgfsetroundjoin%
\pgfsetlinewidth{0.401500pt}%
\definecolor{currentstroke}{rgb}{0.800000,0.800000,0.800000}%
\pgfsetstrokecolor{currentstroke}%
\pgfsetstrokeopacity{0.550000}%
\pgfsetdash{}{0pt}%
\pgfpathmoveto{\pgfqpoint{0.188889in}{1.622700in}}%
\pgfpathlineto{\pgfqpoint{4.838889in}{1.622700in}}%
\pgfusepath{stroke}%
\end{pgfscope}%
\begin{pgfscope}%
\pgfpathrectangle{\pgfqpoint{0.188889in}{0.454405in}}{\pgfqpoint{4.650000in}{3.020000in}}%
\pgfusepath{clip}%
\pgfsetrectcap%
\pgfsetroundjoin%
\pgfsetlinewidth{0.401500pt}%
\definecolor{currentstroke}{rgb}{0.800000,0.800000,0.800000}%
\pgfsetstrokecolor{currentstroke}%
\pgfsetstrokeopacity{0.550000}%
\pgfsetdash{}{0pt}%
\pgfpathmoveto{\pgfqpoint{0.188889in}{1.810754in}}%
\pgfpathlineto{\pgfqpoint{4.838889in}{1.810754in}}%
\pgfusepath{stroke}%
\end{pgfscope}%
\begin{pgfscope}%
\pgfpathrectangle{\pgfqpoint{0.188889in}{0.454405in}}{\pgfqpoint{4.650000in}{3.020000in}}%
\pgfusepath{clip}%
\pgfsetrectcap%
\pgfsetroundjoin%
\pgfsetlinewidth{0.401500pt}%
\definecolor{currentstroke}{rgb}{0.800000,0.800000,0.800000}%
\pgfsetstrokecolor{currentstroke}%
\pgfsetstrokeopacity{0.550000}%
\pgfsetdash{}{0pt}%
\pgfpathmoveto{\pgfqpoint{0.188889in}{1.964405in}}%
\pgfpathlineto{\pgfqpoint{4.838889in}{1.964405in}}%
\pgfusepath{stroke}%
\end{pgfscope}%
\begin{pgfscope}%
\pgfpathrectangle{\pgfqpoint{0.188889in}{0.454405in}}{\pgfqpoint{4.650000in}{3.020000in}}%
\pgfusepath{clip}%
\pgfsetrectcap%
\pgfsetroundjoin%
\pgfsetlinewidth{0.401500pt}%
\definecolor{currentstroke}{rgb}{0.800000,0.800000,0.800000}%
\pgfsetstrokecolor{currentstroke}%
\pgfsetstrokeopacity{0.550000}%
\pgfsetdash{}{0pt}%
\pgfpathmoveto{\pgfqpoint{0.188889in}{2.094315in}}%
\pgfpathlineto{\pgfqpoint{4.838889in}{2.094315in}}%
\pgfusepath{stroke}%
\end{pgfscope}%
\begin{pgfscope}%
\pgfpathrectangle{\pgfqpoint{0.188889in}{0.454405in}}{\pgfqpoint{4.650000in}{3.020000in}}%
\pgfusepath{clip}%
\pgfsetrectcap%
\pgfsetroundjoin%
\pgfsetlinewidth{0.401500pt}%
\definecolor{currentstroke}{rgb}{0.800000,0.800000,0.800000}%
\pgfsetstrokecolor{currentstroke}%
\pgfsetstrokeopacity{0.550000}%
\pgfsetdash{}{0pt}%
\pgfpathmoveto{\pgfqpoint{0.188889in}{2.206848in}}%
\pgfpathlineto{\pgfqpoint{4.838889in}{2.206848in}}%
\pgfusepath{stroke}%
\end{pgfscope}%
\begin{pgfscope}%
\pgfpathrectangle{\pgfqpoint{0.188889in}{0.454405in}}{\pgfqpoint{4.650000in}{3.020000in}}%
\pgfusepath{clip}%
\pgfsetrectcap%
\pgfsetroundjoin%
\pgfsetlinewidth{0.401500pt}%
\definecolor{currentstroke}{rgb}{0.800000,0.800000,0.800000}%
\pgfsetstrokecolor{currentstroke}%
\pgfsetstrokeopacity{0.550000}%
\pgfsetdash{}{0pt}%
\pgfpathmoveto{\pgfqpoint{0.188889in}{2.306110in}}%
\pgfpathlineto{\pgfqpoint{4.838889in}{2.306110in}}%
\pgfusepath{stroke}%
\end{pgfscope}%
\begin{pgfscope}%
\pgfpathrectangle{\pgfqpoint{0.188889in}{0.454405in}}{\pgfqpoint{4.650000in}{3.020000in}}%
\pgfusepath{clip}%
\pgfsetrectcap%
\pgfsetroundjoin%
\pgfsetlinewidth{0.401500pt}%
\definecolor{currentstroke}{rgb}{0.800000,0.800000,0.800000}%
\pgfsetstrokecolor{currentstroke}%
\pgfsetstrokeopacity{0.550000}%
\pgfsetdash{}{0pt}%
\pgfpathmoveto{\pgfqpoint{0.188889in}{2.979050in}}%
\pgfpathlineto{\pgfqpoint{4.838889in}{2.979050in}}%
\pgfusepath{stroke}%
\end{pgfscope}%
\begin{pgfscope}%
\pgfpathrectangle{\pgfqpoint{0.188889in}{0.454405in}}{\pgfqpoint{4.650000in}{3.020000in}}%
\pgfusepath{clip}%
\pgfsetrectcap%
\pgfsetroundjoin%
\pgfsetlinewidth{0.401500pt}%
\definecolor{currentstroke}{rgb}{0.800000,0.800000,0.800000}%
\pgfsetstrokecolor{currentstroke}%
\pgfsetstrokeopacity{0.550000}%
\pgfsetdash{}{0pt}%
\pgfpathmoveto{\pgfqpoint{0.188889in}{3.320754in}}%
\pgfpathlineto{\pgfqpoint{4.838889in}{3.320754in}}%
\pgfusepath{stroke}%
\end{pgfscope}%
\begin{pgfscope}%
\pgfpathrectangle{\pgfqpoint{0.188889in}{0.454405in}}{\pgfqpoint{4.650000in}{3.020000in}}%
\pgfusepath{clip}%
\pgfsetbuttcap%
\pgfsetmiterjoin%
\definecolor{currentfill}{rgb}{0.347059,0.458824,0.641176}%
\pgfsetfillcolor{currentfill}%
\pgfsetlinewidth{0.803000pt}%
\definecolor{currentstroke}{rgb}{0.298039,0.298039,0.298039}%
\pgfsetstrokecolor{currentstroke}%
\pgfsetdash{}{0pt}%
\pgfpathmoveto{\pgfqpoint{0.305139in}{1.785007in}}%
\pgfpathlineto{\pgfqpoint{1.235139in}{1.785007in}}%
\pgfpathlineto{\pgfqpoint{1.235139in}{1.868402in}}%
\pgfpathlineto{\pgfqpoint{0.305139in}{1.868402in}}%
\pgfpathlineto{\pgfqpoint{0.305139in}{1.785007in}}%
\pgfpathclose%
\pgfusepath{stroke,fill}%
\end{pgfscope}%
\begin{pgfscope}%
\pgfpathrectangle{\pgfqpoint{0.188889in}{0.454405in}}{\pgfqpoint{4.650000in}{3.020000in}}%
\pgfusepath{clip}%
\pgfsetbuttcap%
\pgfsetmiterjoin%
\definecolor{currentfill}{rgb}{0.798529,0.536765,0.389706}%
\pgfsetfillcolor{currentfill}%
\pgfsetlinewidth{0.803000pt}%
\definecolor{currentstroke}{rgb}{0.298039,0.298039,0.298039}%
\pgfsetstrokecolor{currentstroke}%
\pgfsetdash{}{0pt}%
\pgfpathmoveto{\pgfqpoint{1.467639in}{1.773696in}}%
\pgfpathlineto{\pgfqpoint{2.397639in}{1.773696in}}%
\pgfpathlineto{\pgfqpoint{2.397639in}{1.895467in}}%
\pgfpathlineto{\pgfqpoint{1.467639in}{1.895467in}}%
\pgfpathlineto{\pgfqpoint{1.467639in}{1.773696in}}%
\pgfpathclose%
\pgfusepath{stroke,fill}%
\end{pgfscope}%
\begin{pgfscope}%
\pgfpathrectangle{\pgfqpoint{0.188889in}{0.454405in}}{\pgfqpoint{4.650000in}{3.020000in}}%
\pgfusepath{clip}%
\pgfsetbuttcap%
\pgfsetmiterjoin%
\definecolor{currentfill}{rgb}{0.374020,0.618137,0.429902}%
\pgfsetfillcolor{currentfill}%
\pgfsetlinewidth{0.803000pt}%
\definecolor{currentstroke}{rgb}{0.298039,0.298039,0.298039}%
\pgfsetstrokecolor{currentstroke}%
\pgfsetdash{}{0pt}%
\pgfpathmoveto{\pgfqpoint{2.630139in}{1.994272in}}%
\pgfpathlineto{\pgfqpoint{3.560139in}{1.994272in}}%
\pgfpathlineto{\pgfqpoint{3.560139in}{2.211896in}}%
\pgfpathlineto{\pgfqpoint{2.630139in}{2.211896in}}%
\pgfpathlineto{\pgfqpoint{2.630139in}{1.994272in}}%
\pgfpathclose%
\pgfusepath{stroke,fill}%
\end{pgfscope}%
\begin{pgfscope}%
\pgfpathrectangle{\pgfqpoint{0.188889in}{0.454405in}}{\pgfqpoint{4.650000in}{3.020000in}}%
\pgfusepath{clip}%
\pgfsetbuttcap%
\pgfsetmiterjoin%
\definecolor{currentfill}{rgb}{0.710784,0.363725,0.375490}%
\pgfsetfillcolor{currentfill}%
\pgfsetlinewidth{0.803000pt}%
\definecolor{currentstroke}{rgb}{0.298039,0.298039,0.298039}%
\pgfsetstrokecolor{currentstroke}%
\pgfsetdash{}{0pt}%
\pgfpathmoveto{\pgfqpoint{3.792639in}{2.958121in}}%
\pgfpathlineto{\pgfqpoint{4.722639in}{2.958121in}}%
\pgfpathlineto{\pgfqpoint{4.722639in}{3.119894in}}%
\pgfpathlineto{\pgfqpoint{3.792639in}{3.119894in}}%
\pgfpathlineto{\pgfqpoint{3.792639in}{2.958121in}}%
\pgfpathclose%
\pgfusepath{stroke,fill}%
\end{pgfscope}%
\begin{pgfscope}%
\pgfsetrectcap%
\pgfsetmiterjoin%
\pgfsetlinewidth{1.254687pt}%
\definecolor{currentstroke}{rgb}{0.800000,0.800000,0.800000}%
\pgfsetstrokecolor{currentstroke}%
\pgfsetdash{}{0pt}%
\pgfpathmoveto{\pgfqpoint{0.188889in}{0.454405in}}%
\pgfpathlineto{\pgfqpoint{0.188889in}{3.474405in}}%
\pgfusepath{stroke}%
\end{pgfscope}%
\begin{pgfscope}%
\pgfsetrectcap%
\pgfsetmiterjoin%
\pgfsetlinewidth{1.254687pt}%
\definecolor{currentstroke}{rgb}{0.800000,0.800000,0.800000}%
\pgfsetstrokecolor{currentstroke}%
\pgfsetdash{}{0pt}%
\pgfpathmoveto{\pgfqpoint{4.838889in}{0.454405in}}%
\pgfpathlineto{\pgfqpoint{4.838889in}{3.474405in}}%
\pgfusepath{stroke}%
\end{pgfscope}%
\begin{pgfscope}%
\pgfsetrectcap%
\pgfsetmiterjoin%
\pgfsetlinewidth{1.254687pt}%
\definecolor{currentstroke}{rgb}{0.800000,0.800000,0.800000}%
\pgfsetstrokecolor{currentstroke}%
\pgfsetdash{}{0pt}%
\pgfpathmoveto{\pgfqpoint{0.188889in}{0.454405in}}%
\pgfpathlineto{\pgfqpoint{4.838889in}{0.454405in}}%
\pgfusepath{stroke}%
\end{pgfscope}%
\begin{pgfscope}%
\pgfsetrectcap%
\pgfsetmiterjoin%
\pgfsetlinewidth{1.254687pt}%
\definecolor{currentstroke}{rgb}{0.800000,0.800000,0.800000}%
\pgfsetstrokecolor{currentstroke}%
\pgfsetdash{}{0pt}%
\pgfpathmoveto{\pgfqpoint{0.188889in}{3.474405in}}%
\pgfpathlineto{\pgfqpoint{4.838889in}{3.474405in}}%
\pgfusepath{stroke}%
\end{pgfscope}%
\begin{pgfscope}%
\pgfpathrectangle{\pgfqpoint{0.188889in}{0.454405in}}{\pgfqpoint{4.650000in}{3.020000in}}%
\pgfusepath{clip}%
\pgfsetrectcap%
\pgfsetroundjoin%
\pgfsetlinewidth{0.803000pt}%
\definecolor{currentstroke}{rgb}{0.298039,0.298039,0.298039}%
\pgfsetstrokecolor{currentstroke}%
\pgfsetdash{}{0pt}%
\pgfpathmoveto{\pgfqpoint{0.770139in}{1.785007in}}%
\pgfpathlineto{\pgfqpoint{0.770139in}{1.694641in}}%
\pgfusepath{stroke}%
\end{pgfscope}%
\begin{pgfscope}%
\pgfpathrectangle{\pgfqpoint{0.188889in}{0.454405in}}{\pgfqpoint{4.650000in}{3.020000in}}%
\pgfusepath{clip}%
\pgfsetrectcap%
\pgfsetroundjoin%
\pgfsetlinewidth{0.803000pt}%
\definecolor{currentstroke}{rgb}{0.298039,0.298039,0.298039}%
\pgfsetstrokecolor{currentstroke}%
\pgfsetdash{}{0pt}%
\pgfpathmoveto{\pgfqpoint{0.770139in}{1.868402in}}%
\pgfpathlineto{\pgfqpoint{0.770139in}{1.868402in}}%
\pgfusepath{stroke}%
\end{pgfscope}%
\begin{pgfscope}%
\pgfpathrectangle{\pgfqpoint{0.188889in}{0.454405in}}{\pgfqpoint{4.650000in}{3.020000in}}%
\pgfusepath{clip}%
\pgfsetrectcap%
\pgfsetroundjoin%
\pgfsetlinewidth{0.803000pt}%
\definecolor{currentstroke}{rgb}{0.298039,0.298039,0.298039}%
\pgfsetstrokecolor{currentstroke}%
\pgfsetdash{}{0pt}%
\pgfpathmoveto{\pgfqpoint{0.537639in}{1.694641in}}%
\pgfpathlineto{\pgfqpoint{1.002639in}{1.694641in}}%
\pgfusepath{stroke}%
\end{pgfscope}%
\begin{pgfscope}%
\pgfpathrectangle{\pgfqpoint{0.188889in}{0.454405in}}{\pgfqpoint{4.650000in}{3.020000in}}%
\pgfusepath{clip}%
\pgfsetrectcap%
\pgfsetroundjoin%
\pgfsetlinewidth{0.803000pt}%
\definecolor{currentstroke}{rgb}{0.298039,0.298039,0.298039}%
\pgfsetstrokecolor{currentstroke}%
\pgfsetdash{}{0pt}%
\pgfpathmoveto{\pgfqpoint{0.537639in}{1.868402in}}%
\pgfpathlineto{\pgfqpoint{1.002639in}{1.868402in}}%
\pgfusepath{stroke}%
\end{pgfscope}%
\begin{pgfscope}%
\pgfpathrectangle{\pgfqpoint{0.188889in}{0.454405in}}{\pgfqpoint{4.650000in}{3.020000in}}%
\pgfusepath{clip}%
\pgfsetrectcap%
\pgfsetroundjoin%
\pgfsetlinewidth{0.803000pt}%
\definecolor{currentstroke}{rgb}{0.298039,0.298039,0.298039}%
\pgfsetstrokecolor{currentstroke}%
\pgfsetdash{}{0pt}%
\pgfpathmoveto{\pgfqpoint{0.305139in}{1.785007in}}%
\pgfpathlineto{\pgfqpoint{1.235139in}{1.785007in}}%
\pgfusepath{stroke}%
\end{pgfscope}%
\begin{pgfscope}%
\pgfpathrectangle{\pgfqpoint{0.188889in}{0.454405in}}{\pgfqpoint{4.650000in}{3.020000in}}%
\pgfusepath{clip}%
\pgfsetbuttcap%
\pgfsetmiterjoin%
\definecolor{currentfill}{rgb}{0.298039,0.298039,0.298039}%
\pgfsetfillcolor{currentfill}%
\pgfsetlinewidth{1.003750pt}%
\definecolor{currentstroke}{rgb}{0.298039,0.298039,0.298039}%
\pgfsetstrokecolor{currentstroke}%
\pgfsetdash{}{0pt}%
\pgfsys@defobject{currentmarker}{\pgfqpoint{-0.029463in}{-0.049105in}}{\pgfqpoint{0.029463in}{0.049105in}}{%
\pgfpathmoveto{\pgfqpoint{0.000000in}{-0.049105in}}%
\pgfpathlineto{\pgfqpoint{0.029463in}{0.000000in}}%
\pgfpathlineto{\pgfqpoint{0.000000in}{0.049105in}}%
\pgfpathlineto{\pgfqpoint{-0.029463in}{0.000000in}}%
\pgfpathclose%
\pgfusepath{stroke,fill}%
}%
\begin{pgfscope}%
\pgfsys@transformshift{0.770139in}{1.598208in}%
\pgfsys@useobject{currentmarker}{}%
\end{pgfscope}%
\begin{pgfscope}%
\pgfsys@transformshift{0.770139in}{1.598208in}%
\pgfsys@useobject{currentmarker}{}%
\end{pgfscope}%
\begin{pgfscope}%
\pgfsys@transformshift{0.770139in}{2.086343in}%
\pgfsys@useobject{currentmarker}{}%
\end{pgfscope}%
\begin{pgfscope}%
\pgfsys@transformshift{0.770139in}{2.086343in}%
\pgfsys@useobject{currentmarker}{}%
\end{pgfscope}%
\begin{pgfscope}%
\pgfsys@transformshift{0.770139in}{2.086343in}%
\pgfsys@useobject{currentmarker}{}%
\end{pgfscope}%
\end{pgfscope}%
\begin{pgfscope}%
\pgfpathrectangle{\pgfqpoint{0.188889in}{0.454405in}}{\pgfqpoint{4.650000in}{3.020000in}}%
\pgfusepath{clip}%
\pgfsetrectcap%
\pgfsetroundjoin%
\pgfsetlinewidth{0.803000pt}%
\definecolor{currentstroke}{rgb}{0.298039,0.298039,0.298039}%
\pgfsetstrokecolor{currentstroke}%
\pgfsetdash{}{0pt}%
\pgfpathmoveto{\pgfqpoint{1.932639in}{1.773696in}}%
\pgfpathlineto{\pgfqpoint{1.932639in}{1.656527in}}%
\pgfusepath{stroke}%
\end{pgfscope}%
\begin{pgfscope}%
\pgfpathrectangle{\pgfqpoint{0.188889in}{0.454405in}}{\pgfqpoint{4.650000in}{3.020000in}}%
\pgfusepath{clip}%
\pgfsetrectcap%
\pgfsetroundjoin%
\pgfsetlinewidth{0.803000pt}%
\definecolor{currentstroke}{rgb}{0.298039,0.298039,0.298039}%
\pgfsetstrokecolor{currentstroke}%
\pgfsetdash{}{0pt}%
\pgfpathmoveto{\pgfqpoint{1.932639in}{1.895467in}}%
\pgfpathlineto{\pgfqpoint{1.932639in}{1.939965in}}%
\pgfusepath{stroke}%
\end{pgfscope}%
\begin{pgfscope}%
\pgfpathrectangle{\pgfqpoint{0.188889in}{0.454405in}}{\pgfqpoint{4.650000in}{3.020000in}}%
\pgfusepath{clip}%
\pgfsetrectcap%
\pgfsetroundjoin%
\pgfsetlinewidth{0.803000pt}%
\definecolor{currentstroke}{rgb}{0.298039,0.298039,0.298039}%
\pgfsetstrokecolor{currentstroke}%
\pgfsetdash{}{0pt}%
\pgfpathmoveto{\pgfqpoint{1.700139in}{1.656527in}}%
\pgfpathlineto{\pgfqpoint{2.165139in}{1.656527in}}%
\pgfusepath{stroke}%
\end{pgfscope}%
\begin{pgfscope}%
\pgfpathrectangle{\pgfqpoint{0.188889in}{0.454405in}}{\pgfqpoint{4.650000in}{3.020000in}}%
\pgfusepath{clip}%
\pgfsetrectcap%
\pgfsetroundjoin%
\pgfsetlinewidth{0.803000pt}%
\definecolor{currentstroke}{rgb}{0.298039,0.298039,0.298039}%
\pgfsetstrokecolor{currentstroke}%
\pgfsetdash{}{0pt}%
\pgfpathmoveto{\pgfqpoint{1.700139in}{1.939965in}}%
\pgfpathlineto{\pgfqpoint{2.165139in}{1.939965in}}%
\pgfusepath{stroke}%
\end{pgfscope}%
\begin{pgfscope}%
\pgfpathrectangle{\pgfqpoint{0.188889in}{0.454405in}}{\pgfqpoint{4.650000in}{3.020000in}}%
\pgfusepath{clip}%
\pgfsetrectcap%
\pgfsetroundjoin%
\pgfsetlinewidth{0.803000pt}%
\definecolor{currentstroke}{rgb}{0.298039,0.298039,0.298039}%
\pgfsetstrokecolor{currentstroke}%
\pgfsetdash{}{0pt}%
\pgfpathmoveto{\pgfqpoint{1.467639in}{1.889374in}}%
\pgfpathlineto{\pgfqpoint{2.397639in}{1.889374in}}%
\pgfusepath{stroke}%
\end{pgfscope}%
\begin{pgfscope}%
\pgfpathrectangle{\pgfqpoint{0.188889in}{0.454405in}}{\pgfqpoint{4.650000in}{3.020000in}}%
\pgfusepath{clip}%
\pgfsetbuttcap%
\pgfsetmiterjoin%
\definecolor{currentfill}{rgb}{0.298039,0.298039,0.298039}%
\pgfsetfillcolor{currentfill}%
\pgfsetlinewidth{1.003750pt}%
\definecolor{currentstroke}{rgb}{0.298039,0.298039,0.298039}%
\pgfsetstrokecolor{currentstroke}%
\pgfsetdash{}{0pt}%
\pgfsys@defobject{currentmarker}{\pgfqpoint{-0.029463in}{-0.049105in}}{\pgfqpoint{0.029463in}{0.049105in}}{%
\pgfpathmoveto{\pgfqpoint{0.000000in}{-0.049105in}}%
\pgfpathlineto{\pgfqpoint{0.029463in}{0.000000in}}%
\pgfpathlineto{\pgfqpoint{0.000000in}{0.049105in}}%
\pgfpathlineto{\pgfqpoint{-0.029463in}{0.000000in}}%
\pgfpathclose%
\pgfusepath{stroke,fill}%
}%
\begin{pgfscope}%
\pgfsys@transformshift{1.932639in}{2.279613in}%
\pgfsys@useobject{currentmarker}{}%
\end{pgfscope}%
\begin{pgfscope}%
\pgfsys@transformshift{1.932639in}{2.279613in}%
\pgfsys@useobject{currentmarker}{}%
\end{pgfscope}%
\end{pgfscope}%
\begin{pgfscope}%
\pgfpathrectangle{\pgfqpoint{0.188889in}{0.454405in}}{\pgfqpoint{4.650000in}{3.020000in}}%
\pgfusepath{clip}%
\pgfsetrectcap%
\pgfsetroundjoin%
\pgfsetlinewidth{0.803000pt}%
\definecolor{currentstroke}{rgb}{0.298039,0.298039,0.298039}%
\pgfsetstrokecolor{currentstroke}%
\pgfsetdash{}{0pt}%
\pgfpathmoveto{\pgfqpoint{3.095139in}{1.994272in}}%
\pgfpathlineto{\pgfqpoint{3.095139in}{1.935753in}}%
\pgfusepath{stroke}%
\end{pgfscope}%
\begin{pgfscope}%
\pgfpathrectangle{\pgfqpoint{0.188889in}{0.454405in}}{\pgfqpoint{4.650000in}{3.020000in}}%
\pgfusepath{clip}%
\pgfsetrectcap%
\pgfsetroundjoin%
\pgfsetlinewidth{0.803000pt}%
\definecolor{currentstroke}{rgb}{0.298039,0.298039,0.298039}%
\pgfsetstrokecolor{currentstroke}%
\pgfsetdash{}{0pt}%
\pgfpathmoveto{\pgfqpoint{3.095139in}{2.211896in}}%
\pgfpathlineto{\pgfqpoint{3.095139in}{2.211896in}}%
\pgfusepath{stroke}%
\end{pgfscope}%
\begin{pgfscope}%
\pgfpathrectangle{\pgfqpoint{0.188889in}{0.454405in}}{\pgfqpoint{4.650000in}{3.020000in}}%
\pgfusepath{clip}%
\pgfsetrectcap%
\pgfsetroundjoin%
\pgfsetlinewidth{0.803000pt}%
\definecolor{currentstroke}{rgb}{0.298039,0.298039,0.298039}%
\pgfsetstrokecolor{currentstroke}%
\pgfsetdash{}{0pt}%
\pgfpathmoveto{\pgfqpoint{2.862639in}{1.935753in}}%
\pgfpathlineto{\pgfqpoint{3.327639in}{1.935753in}}%
\pgfusepath{stroke}%
\end{pgfscope}%
\begin{pgfscope}%
\pgfpathrectangle{\pgfqpoint{0.188889in}{0.454405in}}{\pgfqpoint{4.650000in}{3.020000in}}%
\pgfusepath{clip}%
\pgfsetrectcap%
\pgfsetroundjoin%
\pgfsetlinewidth{0.803000pt}%
\definecolor{currentstroke}{rgb}{0.298039,0.298039,0.298039}%
\pgfsetstrokecolor{currentstroke}%
\pgfsetdash{}{0pt}%
\pgfpathmoveto{\pgfqpoint{2.862639in}{2.211896in}}%
\pgfpathlineto{\pgfqpoint{3.327639in}{2.211896in}}%
\pgfusepath{stroke}%
\end{pgfscope}%
\begin{pgfscope}%
\pgfpathrectangle{\pgfqpoint{0.188889in}{0.454405in}}{\pgfqpoint{4.650000in}{3.020000in}}%
\pgfusepath{clip}%
\pgfsetrectcap%
\pgfsetroundjoin%
\pgfsetlinewidth{0.803000pt}%
\definecolor{currentstroke}{rgb}{0.298039,0.298039,0.298039}%
\pgfsetstrokecolor{currentstroke}%
\pgfsetdash{}{0pt}%
\pgfpathmoveto{\pgfqpoint{2.630139in}{2.047369in}}%
\pgfpathlineto{\pgfqpoint{3.560139in}{2.047369in}}%
\pgfusepath{stroke}%
\end{pgfscope}%
\begin{pgfscope}%
\pgfpathrectangle{\pgfqpoint{0.188889in}{0.454405in}}{\pgfqpoint{4.650000in}{3.020000in}}%
\pgfusepath{clip}%
\pgfsetbuttcap%
\pgfsetmiterjoin%
\definecolor{currentfill}{rgb}{0.298039,0.298039,0.298039}%
\pgfsetfillcolor{currentfill}%
\pgfsetlinewidth{1.003750pt}%
\definecolor{currentstroke}{rgb}{0.298039,0.298039,0.298039}%
\pgfsetstrokecolor{currentstroke}%
\pgfsetdash{}{0pt}%
\pgfsys@defobject{currentmarker}{\pgfqpoint{-0.029463in}{-0.049105in}}{\pgfqpoint{0.029463in}{0.049105in}}{%
\pgfpathmoveto{\pgfqpoint{0.000000in}{-0.049105in}}%
\pgfpathlineto{\pgfqpoint{0.029463in}{0.000000in}}%
\pgfpathlineto{\pgfqpoint{0.000000in}{0.049105in}}%
\pgfpathlineto{\pgfqpoint{-0.029463in}{0.000000in}}%
\pgfpathclose%
\pgfusepath{stroke,fill}%
}%
\begin{pgfscope}%
\pgfsys@transformshift{3.095139in}{2.648201in}%
\pgfsys@useobject{currentmarker}{}%
\end{pgfscope}%
\end{pgfscope}%
\begin{pgfscope}%
\pgfpathrectangle{\pgfqpoint{0.188889in}{0.454405in}}{\pgfqpoint{4.650000in}{3.020000in}}%
\pgfusepath{clip}%
\pgfsetrectcap%
\pgfsetroundjoin%
\pgfsetlinewidth{0.803000pt}%
\definecolor{currentstroke}{rgb}{0.298039,0.298039,0.298039}%
\pgfsetstrokecolor{currentstroke}%
\pgfsetdash{}{0pt}%
\pgfpathmoveto{\pgfqpoint{4.257639in}{2.958121in}}%
\pgfpathlineto{\pgfqpoint{4.257639in}{2.690609in}}%
\pgfusepath{stroke}%
\end{pgfscope}%
\begin{pgfscope}%
\pgfpathrectangle{\pgfqpoint{0.188889in}{0.454405in}}{\pgfqpoint{4.650000in}{3.020000in}}%
\pgfusepath{clip}%
\pgfsetrectcap%
\pgfsetroundjoin%
\pgfsetlinewidth{0.803000pt}%
\definecolor{currentstroke}{rgb}{0.298039,0.298039,0.298039}%
\pgfsetstrokecolor{currentstroke}%
\pgfsetdash{}{0pt}%
\pgfpathmoveto{\pgfqpoint{4.257639in}{3.119894in}}%
\pgfpathlineto{\pgfqpoint{4.257639in}{3.223483in}}%
\pgfusepath{stroke}%
\end{pgfscope}%
\begin{pgfscope}%
\pgfpathrectangle{\pgfqpoint{0.188889in}{0.454405in}}{\pgfqpoint{4.650000in}{3.020000in}}%
\pgfusepath{clip}%
\pgfsetrectcap%
\pgfsetroundjoin%
\pgfsetlinewidth{0.803000pt}%
\definecolor{currentstroke}{rgb}{0.298039,0.298039,0.298039}%
\pgfsetstrokecolor{currentstroke}%
\pgfsetdash{}{0pt}%
\pgfpathmoveto{\pgfqpoint{4.025139in}{2.690609in}}%
\pgfpathlineto{\pgfqpoint{4.490139in}{2.690609in}}%
\pgfusepath{stroke}%
\end{pgfscope}%
\begin{pgfscope}%
\pgfpathrectangle{\pgfqpoint{0.188889in}{0.454405in}}{\pgfqpoint{4.650000in}{3.020000in}}%
\pgfusepath{clip}%
\pgfsetrectcap%
\pgfsetroundjoin%
\pgfsetlinewidth{0.803000pt}%
\definecolor{currentstroke}{rgb}{0.298039,0.298039,0.298039}%
\pgfsetstrokecolor{currentstroke}%
\pgfsetdash{}{0pt}%
\pgfpathmoveto{\pgfqpoint{4.025139in}{3.223483in}}%
\pgfpathlineto{\pgfqpoint{4.490139in}{3.223483in}}%
\pgfusepath{stroke}%
\end{pgfscope}%
\begin{pgfscope}%
\pgfpathrectangle{\pgfqpoint{0.188889in}{0.454405in}}{\pgfqpoint{4.650000in}{3.020000in}}%
\pgfusepath{clip}%
\pgfsetrectcap%
\pgfsetroundjoin%
\pgfsetlinewidth{0.803000pt}%
\definecolor{currentstroke}{rgb}{0.298039,0.298039,0.298039}%
\pgfsetstrokecolor{currentstroke}%
\pgfsetdash{}{0pt}%
\pgfpathmoveto{\pgfqpoint{3.792639in}{3.048590in}}%
\pgfpathlineto{\pgfqpoint{4.722639in}{3.048590in}}%
\pgfusepath{stroke}%
\end{pgfscope}%
\begin{pgfscope}%
\pgfpathrectangle{\pgfqpoint{0.188889in}{0.454405in}}{\pgfqpoint{4.650000in}{3.020000in}}%
\pgfusepath{clip}%
\pgfsetbuttcap%
\pgfsetmiterjoin%
\definecolor{currentfill}{rgb}{0.298039,0.298039,0.298039}%
\pgfsetfillcolor{currentfill}%
\pgfsetlinewidth{1.003750pt}%
\definecolor{currentstroke}{rgb}{0.298039,0.298039,0.298039}%
\pgfsetstrokecolor{currentstroke}%
\pgfsetdash{}{0pt}%
\pgfsys@defobject{currentmarker}{\pgfqpoint{-0.029463in}{-0.049105in}}{\pgfqpoint{0.029463in}{0.049105in}}{%
\pgfpathmoveto{\pgfqpoint{0.000000in}{-0.049105in}}%
\pgfpathlineto{\pgfqpoint{0.029463in}{0.000000in}}%
\pgfpathlineto{\pgfqpoint{0.000000in}{0.049105in}}%
\pgfpathlineto{\pgfqpoint{-0.029463in}{0.000000in}}%
\pgfpathclose%
\pgfusepath{stroke,fill}%
}%
\begin{pgfscope}%
\pgfsys@transformshift{4.257639in}{3.359588in}%
\pgfsys@useobject{currentmarker}{}%
\end{pgfscope}%
\end{pgfscope}%
\end{pgfpicture}%
\makeatother%
\endgroup%

%% file: fig/runtime_barplot_res.pgf
%% Creator: Matplotlib, PGF backend
%%
%% To include the figure in your LaTeX document, write
%%   \input{<filename>.pgf}
%%
%% Make sure the required packages are loaded in your preamble
%%   \usepackage{pgf}
%%
%% Figures using additional raster images can only be included by \input if
%% they are in the same directory as the main LaTeX file. For loading figures
%% from other directories you can use the `import` package
%% and then include the figures with
%%   \import{<path to file>}{<filename>.pgf}
%%
%% Matplotlib used the following preamble
%%
\begingroup%
\makeatletter%
\begin{pgfpicture}%
\pgfpathrectangle{\pgfpointorigin}{\pgfqpoint{5.512652in}{3.648626in}}%
\pgfusepath{use as bounding box, clip}%
\begin{pgfscope}%
\pgfsetbuttcap%
\pgfsetmiterjoin%
\definecolor{currentfill}{rgb}{1.000000,1.000000,1.000000}%
\pgfsetfillcolor{currentfill}%
\pgfsetlinewidth{0.000000pt}%
\definecolor{currentstroke}{rgb}{1.000000,1.000000,1.000000}%
\pgfsetstrokecolor{currentstroke}%
\pgfsetdash{}{0pt}%
\pgfpathmoveto{\pgfqpoint{0.000000in}{0.000000in}}%
\pgfpathlineto{\pgfqpoint{5.512652in}{0.000000in}}%
\pgfpathlineto{\pgfqpoint{5.512652in}{3.648626in}}%
\pgfpathlineto{\pgfqpoint{0.000000in}{3.648626in}}%
\pgfpathclose%
\pgfusepath{fill}%
\end{pgfscope}%
\begin{pgfscope}%
\pgfsetbuttcap%
\pgfsetmiterjoin%
\definecolor{currentfill}{rgb}{1.000000,1.000000,1.000000}%
\pgfsetfillcolor{currentfill}%
\pgfsetlinewidth{0.000000pt}%
\definecolor{currentstroke}{rgb}{0.000000,0.000000,0.000000}%
\pgfsetstrokecolor{currentstroke}%
\pgfsetstrokeopacity{0.000000}%
\pgfsetdash{}{0pt}%
\pgfpathmoveto{\pgfqpoint{0.762652in}{0.445293in}}%
\pgfpathlineto{\pgfqpoint{5.412652in}{0.445293in}}%
\pgfpathlineto{\pgfqpoint{5.412652in}{3.465293in}}%
\pgfpathlineto{\pgfqpoint{0.762652in}{3.465293in}}%
\pgfpathclose%
\pgfusepath{fill}%
\end{pgfscope}%
\begin{pgfscope}%
\definecolor{textcolor}{rgb}{0.150000,0.150000,0.150000}%
\pgfsetstrokecolor{textcolor}%
\pgfsetfillcolor{textcolor}%
\pgftext[x=1.343902in,y=0.313349in,,top]{\color{textcolor}\rmfamily\fontsize{16.500000}{19.800000}\selectfont \textsc{circle}}%
\end{pgfscope}%
\begin{pgfscope}%
\definecolor{textcolor}{rgb}{0.150000,0.150000,0.150000}%
\pgfsetstrokecolor{textcolor}%
\pgfsetfillcolor{textcolor}%
\pgftext[x=2.506402in,y=0.313349in,,top]{\color{textcolor}\rmfamily\fontsize{16.500000}{19.800000}\selectfont \textsc{circle} \oldstylenums{5}\textsc{d}}%
\end{pgfscope}%
\begin{pgfscope}%
\definecolor{textcolor}{rgb}{0.150000,0.150000,0.150000}%
\pgfsetstrokecolor{textcolor}%
\pgfsetfillcolor{textcolor}%
\pgftext[x=3.668902in,y=0.313349in,,top]{\color{textcolor}\rmfamily\fontsize{16.500000}{19.800000}\selectfont \textsc{mnist}}%
\end{pgfscope}%
\begin{pgfscope}%
\definecolor{textcolor}{rgb}{0.150000,0.150000,0.150000}%
\pgfsetstrokecolor{textcolor}%
\pgfsetfillcolor{textcolor}%
\pgftext[x=4.831402in,y=0.313349in,,top]{\color{textcolor}\rmfamily\fontsize{16.500000}{19.800000}\selectfont \textsc{adk}}%
\end{pgfscope}%
\begin{pgfscope}%
\pgfpathrectangle{\pgfqpoint{0.762652in}{0.445293in}}{\pgfqpoint{4.650000in}{3.020000in}}%
\pgfusepath{clip}%
\pgfsetroundcap%
\pgfsetroundjoin%
\pgfsetlinewidth{1.003750pt}%
\definecolor{currentstroke}{rgb}{0.800000,0.800000,0.800000}%
\pgfsetstrokecolor{currentstroke}%
\pgfsetdash{}{0pt}%
\pgfpathmoveto{\pgfqpoint{0.762652in}{0.677601in}}%
\pgfpathlineto{\pgfqpoint{5.412652in}{0.677601in}}%
\pgfusepath{stroke}%
\end{pgfscope}%
\begin{pgfscope}%
\definecolor{textcolor}{rgb}{0.150000,0.150000,0.150000}%
\pgfsetstrokecolor{textcolor}%
\pgfsetfillcolor{textcolor}%
\pgftext[x=0.520639in,y=0.594267in,left,base]{\color{textcolor}\rmfamily\fontsize{16.500000}{19.800000}\selectfont 6}%
\end{pgfscope}%
\begin{pgfscope}%
\pgfpathrectangle{\pgfqpoint{0.762652in}{0.445293in}}{\pgfqpoint{4.650000in}{3.020000in}}%
\pgfusepath{clip}%
\pgfsetroundcap%
\pgfsetroundjoin%
\pgfsetlinewidth{1.003750pt}%
\definecolor{currentstroke}{rgb}{0.800000,0.800000,0.800000}%
\pgfsetstrokecolor{currentstroke}%
\pgfsetdash{}{0pt}%
\pgfpathmoveto{\pgfqpoint{0.762652in}{1.142216in}}%
\pgfpathlineto{\pgfqpoint{5.412652in}{1.142216in}}%
\pgfusepath{stroke}%
\end{pgfscope}%
\begin{pgfscope}%
\definecolor{textcolor}{rgb}{0.150000,0.150000,0.150000}%
\pgfsetstrokecolor{textcolor}%
\pgfsetfillcolor{textcolor}%
\pgftext[x=0.520639in,y=1.058883in,left,base]{\color{textcolor}\rmfamily\fontsize{16.500000}{19.800000}\selectfont 8}%
\end{pgfscope}%
\begin{pgfscope}%
\pgfpathrectangle{\pgfqpoint{0.762652in}{0.445293in}}{\pgfqpoint{4.650000in}{3.020000in}}%
\pgfusepath{clip}%
\pgfsetroundcap%
\pgfsetroundjoin%
\pgfsetlinewidth{1.003750pt}%
\definecolor{currentstroke}{rgb}{0.800000,0.800000,0.800000}%
\pgfsetstrokecolor{currentstroke}%
\pgfsetdash{}{0pt}%
\pgfpathmoveto{\pgfqpoint{0.762652in}{1.606832in}}%
\pgfpathlineto{\pgfqpoint{5.412652in}{1.606832in}}%
\pgfusepath{stroke}%
\end{pgfscope}%
\begin{pgfscope}%
\definecolor{textcolor}{rgb}{0.150000,0.150000,0.150000}%
\pgfsetstrokecolor{textcolor}%
\pgfsetfillcolor{textcolor}%
\pgftext[x=0.410571in,y=1.523498in,left,base]{\color{textcolor}\rmfamily\fontsize{16.500000}{19.800000}\selectfont 10}%
\end{pgfscope}%
\begin{pgfscope}%
\pgfpathrectangle{\pgfqpoint{0.762652in}{0.445293in}}{\pgfqpoint{4.650000in}{3.020000in}}%
\pgfusepath{clip}%
\pgfsetroundcap%
\pgfsetroundjoin%
\pgfsetlinewidth{1.003750pt}%
\definecolor{currentstroke}{rgb}{0.800000,0.800000,0.800000}%
\pgfsetstrokecolor{currentstroke}%
\pgfsetdash{}{0pt}%
\pgfpathmoveto{\pgfqpoint{0.762652in}{2.071447in}}%
\pgfpathlineto{\pgfqpoint{5.412652in}{2.071447in}}%
\pgfusepath{stroke}%
\end{pgfscope}%
\begin{pgfscope}%
\definecolor{textcolor}{rgb}{0.150000,0.150000,0.150000}%
\pgfsetstrokecolor{textcolor}%
\pgfsetfillcolor{textcolor}%
\pgftext[x=0.410571in,y=1.988114in,left,base]{\color{textcolor}\rmfamily\fontsize{16.500000}{19.800000}\selectfont 12}%
\end{pgfscope}%
\begin{pgfscope}%
\pgfpathrectangle{\pgfqpoint{0.762652in}{0.445293in}}{\pgfqpoint{4.650000in}{3.020000in}}%
\pgfusepath{clip}%
\pgfsetroundcap%
\pgfsetroundjoin%
\pgfsetlinewidth{1.003750pt}%
\definecolor{currentstroke}{rgb}{0.800000,0.800000,0.800000}%
\pgfsetstrokecolor{currentstroke}%
\pgfsetdash{}{0pt}%
\pgfpathmoveto{\pgfqpoint{0.762652in}{2.536062in}}%
\pgfpathlineto{\pgfqpoint{5.412652in}{2.536062in}}%
\pgfusepath{stroke}%
\end{pgfscope}%
\begin{pgfscope}%
\definecolor{textcolor}{rgb}{0.150000,0.150000,0.150000}%
\pgfsetstrokecolor{textcolor}%
\pgfsetfillcolor{textcolor}%
\pgftext[x=0.410571in,y=2.452729in,left,base]{\color{textcolor}\rmfamily\fontsize{16.500000}{19.800000}\selectfont 14}%
\end{pgfscope}%
\begin{pgfscope}%
\pgfpathrectangle{\pgfqpoint{0.762652in}{0.445293in}}{\pgfqpoint{4.650000in}{3.020000in}}%
\pgfusepath{clip}%
\pgfsetroundcap%
\pgfsetroundjoin%
\pgfsetlinewidth{1.003750pt}%
\definecolor{currentstroke}{rgb}{0.800000,0.800000,0.800000}%
\pgfsetstrokecolor{currentstroke}%
\pgfsetdash{}{0pt}%
\pgfpathmoveto{\pgfqpoint{0.762652in}{3.000678in}}%
\pgfpathlineto{\pgfqpoint{5.412652in}{3.000678in}}%
\pgfusepath{stroke}%
\end{pgfscope}%
\begin{pgfscope}%
\definecolor{textcolor}{rgb}{0.150000,0.150000,0.150000}%
\pgfsetstrokecolor{textcolor}%
\pgfsetfillcolor{textcolor}%
\pgftext[x=0.410571in,y=2.917344in,left,base]{\color{textcolor}\rmfamily\fontsize{16.500000}{19.800000}\selectfont 16}%
\end{pgfscope}%
\begin{pgfscope}%
\pgfpathrectangle{\pgfqpoint{0.762652in}{0.445293in}}{\pgfqpoint{4.650000in}{3.020000in}}%
\pgfusepath{clip}%
\pgfsetroundcap%
\pgfsetroundjoin%
\pgfsetlinewidth{1.003750pt}%
\definecolor{currentstroke}{rgb}{0.800000,0.800000,0.800000}%
\pgfsetstrokecolor{currentstroke}%
\pgfsetdash{}{0pt}%
\pgfpathmoveto{\pgfqpoint{0.762652in}{3.465293in}}%
\pgfpathlineto{\pgfqpoint{5.412652in}{3.465293in}}%
\pgfusepath{stroke}%
\end{pgfscope}%
\begin{pgfscope}%
\definecolor{textcolor}{rgb}{0.150000,0.150000,0.150000}%
\pgfsetstrokecolor{textcolor}%
\pgfsetfillcolor{textcolor}%
\pgftext[x=0.410571in,y=3.381960in,left,base]{\color{textcolor}\rmfamily\fontsize{16.500000}{19.800000}\selectfont 18}%
\end{pgfscope}%
\begin{pgfscope}%
\definecolor{textcolor}{rgb}{0.150000,0.150000,0.150000}%
\pgfsetstrokecolor{textcolor}%
\pgfsetfillcolor{textcolor}%
\pgftext[x=0.313349in,y=1.955293in,,bottom,rotate=90.000000]{\color{textcolor}\rmfamily\fontsize{18.000000}{21.600000}\selectfont runtime in seconds}%
\end{pgfscope}%
\begin{pgfscope}%
\pgfpathrectangle{\pgfqpoint{0.762652in}{0.445293in}}{\pgfqpoint{4.650000in}{3.020000in}}%
\pgfusepath{clip}%
\pgfsetbuttcap%
\pgfsetmiterjoin%
\definecolor{currentfill}{rgb}{0.347059,0.458824,0.641176}%
\pgfsetfillcolor{currentfill}%
\pgfsetlinewidth{1.003750pt}%
\definecolor{currentstroke}{rgb}{1.000000,1.000000,1.000000}%
\pgfsetstrokecolor{currentstroke}%
\pgfsetdash{}{0pt}%
\pgfpathmoveto{\pgfqpoint{0.878902in}{-0.716245in}}%
\pgfpathlineto{\pgfqpoint{1.188902in}{-0.716245in}}%
\pgfpathlineto{\pgfqpoint{1.188902in}{1.542464in}}%
\pgfpathlineto{\pgfqpoint{0.878902in}{1.542464in}}%
\pgfpathclose%
\pgfusepath{stroke,fill}%
\end{pgfscope}%
\begin{pgfscope}%
\pgfpathrectangle{\pgfqpoint{0.762652in}{0.445293in}}{\pgfqpoint{4.650000in}{3.020000in}}%
\pgfusepath{clip}%
\pgfsetbuttcap%
\pgfsetmiterjoin%
\definecolor{currentfill}{rgb}{0.347059,0.458824,0.641176}%
\pgfsetfillcolor{currentfill}%
\pgfsetlinewidth{1.003750pt}%
\definecolor{currentstroke}{rgb}{1.000000,1.000000,1.000000}%
\pgfsetstrokecolor{currentstroke}%
\pgfsetdash{}{0pt}%
\pgfpathmoveto{\pgfqpoint{2.041402in}{-0.716245in}}%
\pgfpathlineto{\pgfqpoint{2.351402in}{-0.716245in}}%
\pgfpathlineto{\pgfqpoint{2.351402in}{2.752588in}}%
\pgfpathlineto{\pgfqpoint{2.041402in}{2.752588in}}%
\pgfpathclose%
\pgfusepath{stroke,fill}%
\end{pgfscope}%
\begin{pgfscope}%
\pgfpathrectangle{\pgfqpoint{0.762652in}{0.445293in}}{\pgfqpoint{4.650000in}{3.020000in}}%
\pgfusepath{clip}%
\pgfsetbuttcap%
\pgfsetmiterjoin%
\definecolor{currentfill}{rgb}{0.347059,0.458824,0.641176}%
\pgfsetfillcolor{currentfill}%
\pgfsetlinewidth{1.003750pt}%
\definecolor{currentstroke}{rgb}{1.000000,1.000000,1.000000}%
\pgfsetstrokecolor{currentstroke}%
\pgfsetdash{}{0pt}%
\pgfpathmoveto{\pgfqpoint{3.203902in}{-0.716245in}}%
\pgfpathlineto{\pgfqpoint{3.513902in}{-0.716245in}}%
\pgfpathlineto{\pgfqpoint{3.513902in}{2.744268in}}%
\pgfpathlineto{\pgfqpoint{3.203902in}{2.744268in}}%
\pgfpathclose%
\pgfusepath{stroke,fill}%
\end{pgfscope}%
\begin{pgfscope}%
\pgfpathrectangle{\pgfqpoint{0.762652in}{0.445293in}}{\pgfqpoint{4.650000in}{3.020000in}}%
\pgfusepath{clip}%
\pgfsetbuttcap%
\pgfsetmiterjoin%
\definecolor{currentfill}{rgb}{0.347059,0.458824,0.641176}%
\pgfsetfillcolor{currentfill}%
\pgfsetlinewidth{1.003750pt}%
\definecolor{currentstroke}{rgb}{1.000000,1.000000,1.000000}%
\pgfsetstrokecolor{currentstroke}%
\pgfsetdash{}{0pt}%
\pgfpathmoveto{\pgfqpoint{4.366402in}{-0.716245in}}%
\pgfpathlineto{\pgfqpoint{4.676402in}{-0.716245in}}%
\pgfpathlineto{\pgfqpoint{4.676402in}{1.307293in}}%
\pgfpathlineto{\pgfqpoint{4.366402in}{1.307293in}}%
\pgfpathclose%
\pgfusepath{stroke,fill}%
\end{pgfscope}%
\begin{pgfscope}%
\pgfpathrectangle{\pgfqpoint{0.762652in}{0.445293in}}{\pgfqpoint{4.650000in}{3.020000in}}%
\pgfusepath{clip}%
\pgfsetbuttcap%
\pgfsetmiterjoin%
\definecolor{currentfill}{rgb}{0.798529,0.536765,0.389706}%
\pgfsetfillcolor{currentfill}%
\pgfsetlinewidth{1.003750pt}%
\definecolor{currentstroke}{rgb}{1.000000,1.000000,1.000000}%
\pgfsetstrokecolor{currentstroke}%
\pgfsetdash{}{0pt}%
\pgfpathmoveto{\pgfqpoint{1.188902in}{-0.716245in}}%
\pgfpathlineto{\pgfqpoint{1.498902in}{-0.716245in}}%
\pgfpathlineto{\pgfqpoint{1.498902in}{1.553420in}}%
\pgfpathlineto{\pgfqpoint{1.188902in}{1.553420in}}%
\pgfpathclose%
\pgfusepath{stroke,fill}%
\end{pgfscope}%
\begin{pgfscope}%
\pgfpathrectangle{\pgfqpoint{0.762652in}{0.445293in}}{\pgfqpoint{4.650000in}{3.020000in}}%
\pgfusepath{clip}%
\pgfsetbuttcap%
\pgfsetmiterjoin%
\definecolor{currentfill}{rgb}{0.798529,0.536765,0.389706}%
\pgfsetfillcolor{currentfill}%
\pgfsetlinewidth{1.003750pt}%
\definecolor{currentstroke}{rgb}{1.000000,1.000000,1.000000}%
\pgfsetstrokecolor{currentstroke}%
\pgfsetdash{}{0pt}%
\pgfpathmoveto{\pgfqpoint{2.351402in}{-0.716245in}}%
\pgfpathlineto{\pgfqpoint{2.661402in}{-0.716245in}}%
\pgfpathlineto{\pgfqpoint{2.661402in}{2.744986in}}%
\pgfpathlineto{\pgfqpoint{2.351402in}{2.744986in}}%
\pgfpathclose%
\pgfusepath{stroke,fill}%
\end{pgfscope}%
\begin{pgfscope}%
\pgfpathrectangle{\pgfqpoint{0.762652in}{0.445293in}}{\pgfqpoint{4.650000in}{3.020000in}}%
\pgfusepath{clip}%
\pgfsetbuttcap%
\pgfsetmiterjoin%
\definecolor{currentfill}{rgb}{0.798529,0.536765,0.389706}%
\pgfsetfillcolor{currentfill}%
\pgfsetlinewidth{1.003750pt}%
\definecolor{currentstroke}{rgb}{1.000000,1.000000,1.000000}%
\pgfsetstrokecolor{currentstroke}%
\pgfsetdash{}{0pt}%
\pgfpathmoveto{\pgfqpoint{3.513902in}{-0.716245in}}%
\pgfpathlineto{\pgfqpoint{3.823902in}{-0.716245in}}%
\pgfpathlineto{\pgfqpoint{3.823902in}{2.748762in}}%
\pgfpathlineto{\pgfqpoint{3.513902in}{2.748762in}}%
\pgfpathclose%
\pgfusepath{stroke,fill}%
\end{pgfscope}%
\begin{pgfscope}%
\pgfpathrectangle{\pgfqpoint{0.762652in}{0.445293in}}{\pgfqpoint{4.650000in}{3.020000in}}%
\pgfusepath{clip}%
\pgfsetbuttcap%
\pgfsetmiterjoin%
\definecolor{currentfill}{rgb}{0.798529,0.536765,0.389706}%
\pgfsetfillcolor{currentfill}%
\pgfsetlinewidth{1.003750pt}%
\definecolor{currentstroke}{rgb}{1.000000,1.000000,1.000000}%
\pgfsetstrokecolor{currentstroke}%
\pgfsetdash{}{0pt}%
\pgfpathmoveto{\pgfqpoint{4.676402in}{-0.716245in}}%
\pgfpathlineto{\pgfqpoint{4.986402in}{-0.716245in}}%
\pgfpathlineto{\pgfqpoint{4.986402in}{1.310733in}}%
\pgfpathlineto{\pgfqpoint{4.676402in}{1.310733in}}%
\pgfpathclose%
\pgfusepath{stroke,fill}%
\end{pgfscope}%
\begin{pgfscope}%
\pgfpathrectangle{\pgfqpoint{0.762652in}{0.445293in}}{\pgfqpoint{4.650000in}{3.020000in}}%
\pgfusepath{clip}%
\pgfsetbuttcap%
\pgfsetmiterjoin%
\definecolor{currentfill}{rgb}{0.374020,0.618137,0.429902}%
\pgfsetfillcolor{currentfill}%
\pgfsetlinewidth{1.003750pt}%
\definecolor{currentstroke}{rgb}{1.000000,1.000000,1.000000}%
\pgfsetstrokecolor{currentstroke}%
\pgfsetdash{}{0pt}%
\pgfpathmoveto{\pgfqpoint{1.498902in}{-0.716245in}}%
\pgfpathlineto{\pgfqpoint{1.808902in}{-0.716245in}}%
\pgfpathlineto{\pgfqpoint{1.808902in}{1.600877in}}%
\pgfpathlineto{\pgfqpoint{1.498902in}{1.600877in}}%
\pgfpathclose%
\pgfusepath{stroke,fill}%
\end{pgfscope}%
\begin{pgfscope}%
\pgfpathrectangle{\pgfqpoint{0.762652in}{0.445293in}}{\pgfqpoint{4.650000in}{3.020000in}}%
\pgfusepath{clip}%
\pgfsetbuttcap%
\pgfsetmiterjoin%
\definecolor{currentfill}{rgb}{0.374020,0.618137,0.429902}%
\pgfsetfillcolor{currentfill}%
\pgfsetlinewidth{1.003750pt}%
\definecolor{currentstroke}{rgb}{1.000000,1.000000,1.000000}%
\pgfsetstrokecolor{currentstroke}%
\pgfsetdash{}{0pt}%
\pgfpathmoveto{\pgfqpoint{2.661402in}{-0.716245in}}%
\pgfpathlineto{\pgfqpoint{2.971402in}{-0.716245in}}%
\pgfpathlineto{\pgfqpoint{2.971402in}{0.937418in}}%
\pgfpathlineto{\pgfqpoint{2.661402in}{0.937418in}}%
\pgfpathclose%
\pgfusepath{stroke,fill}%
\end{pgfscope}%
\begin{pgfscope}%
\pgfpathrectangle{\pgfqpoint{0.762652in}{0.445293in}}{\pgfqpoint{4.650000in}{3.020000in}}%
\pgfusepath{clip}%
\pgfsetbuttcap%
\pgfsetmiterjoin%
\definecolor{currentfill}{rgb}{0.374020,0.618137,0.429902}%
\pgfsetfillcolor{currentfill}%
\pgfsetlinewidth{1.003750pt}%
\definecolor{currentstroke}{rgb}{1.000000,1.000000,1.000000}%
\pgfsetstrokecolor{currentstroke}%
\pgfsetdash{}{0pt}%
\pgfpathmoveto{\pgfqpoint{3.823902in}{-0.716245in}}%
\pgfpathlineto{\pgfqpoint{4.133902in}{-0.716245in}}%
\pgfpathlineto{\pgfqpoint{4.133902in}{2.096272in}}%
\pgfpathlineto{\pgfqpoint{3.823902in}{2.096272in}}%
\pgfpathclose%
\pgfusepath{stroke,fill}%
\end{pgfscope}%
\begin{pgfscope}%
\pgfpathrectangle{\pgfqpoint{0.762652in}{0.445293in}}{\pgfqpoint{4.650000in}{3.020000in}}%
\pgfusepath{clip}%
\pgfsetbuttcap%
\pgfsetmiterjoin%
\definecolor{currentfill}{rgb}{0.374020,0.618137,0.429902}%
\pgfsetfillcolor{currentfill}%
\pgfsetlinewidth{1.003750pt}%
\definecolor{currentstroke}{rgb}{1.000000,1.000000,1.000000}%
\pgfsetstrokecolor{currentstroke}%
\pgfsetdash{}{0pt}%
\pgfpathmoveto{\pgfqpoint{4.986402in}{-0.716245in}}%
\pgfpathlineto{\pgfqpoint{5.296402in}{-0.716245in}}%
\pgfpathlineto{\pgfqpoint{5.296402in}{1.205693in}}%
\pgfpathlineto{\pgfqpoint{4.986402in}{1.205693in}}%
\pgfpathclose%
\pgfusepath{stroke,fill}%
\end{pgfscope}%
\begin{pgfscope}%
\pgfpathrectangle{\pgfqpoint{0.762652in}{0.445293in}}{\pgfqpoint{4.650000in}{3.020000in}}%
\pgfusepath{clip}%
\pgfsetroundcap%
\pgfsetroundjoin%
\pgfsetlinewidth{2.710125pt}%
\definecolor{currentstroke}{rgb}{0.260000,0.260000,0.260000}%
\pgfsetstrokecolor{currentstroke}%
\pgfsetdash{}{0pt}%
\pgfpathmoveto{\pgfqpoint{1.033902in}{1.538251in}}%
\pgfpathlineto{\pgfqpoint{1.033902in}{1.546648in}}%
\pgfusepath{stroke}%
\end{pgfscope}%
\begin{pgfscope}%
\pgfpathrectangle{\pgfqpoint{0.762652in}{0.445293in}}{\pgfqpoint{4.650000in}{3.020000in}}%
\pgfusepath{clip}%
\pgfsetroundcap%
\pgfsetroundjoin%
\pgfsetlinewidth{2.710125pt}%
\definecolor{currentstroke}{rgb}{0.260000,0.260000,0.260000}%
\pgfsetstrokecolor{currentstroke}%
\pgfsetdash{}{0pt}%
\pgfpathmoveto{\pgfqpoint{2.196402in}{2.740132in}}%
\pgfpathlineto{\pgfqpoint{2.196402in}{2.764967in}}%
\pgfusepath{stroke}%
\end{pgfscope}%
\begin{pgfscope}%
\pgfpathrectangle{\pgfqpoint{0.762652in}{0.445293in}}{\pgfqpoint{4.650000in}{3.020000in}}%
\pgfusepath{clip}%
\pgfsetroundcap%
\pgfsetroundjoin%
\pgfsetlinewidth{2.710125pt}%
\definecolor{currentstroke}{rgb}{0.260000,0.260000,0.260000}%
\pgfsetstrokecolor{currentstroke}%
\pgfsetdash{}{0pt}%
\pgfpathmoveto{\pgfqpoint{3.358902in}{2.738316in}}%
\pgfpathlineto{\pgfqpoint{3.358902in}{2.750709in}}%
\pgfusepath{stroke}%
\end{pgfscope}%
\begin{pgfscope}%
\pgfpathrectangle{\pgfqpoint{0.762652in}{0.445293in}}{\pgfqpoint{4.650000in}{3.020000in}}%
\pgfusepath{clip}%
\pgfsetroundcap%
\pgfsetroundjoin%
\pgfsetlinewidth{2.710125pt}%
\definecolor{currentstroke}{rgb}{0.260000,0.260000,0.260000}%
\pgfsetstrokecolor{currentstroke}%
\pgfsetdash{}{0pt}%
\pgfpathmoveto{\pgfqpoint{4.521402in}{1.298380in}}%
\pgfpathlineto{\pgfqpoint{4.521402in}{1.316735in}}%
\pgfusepath{stroke}%
\end{pgfscope}%
\begin{pgfscope}%
\pgfpathrectangle{\pgfqpoint{0.762652in}{0.445293in}}{\pgfqpoint{4.650000in}{3.020000in}}%
\pgfusepath{clip}%
\pgfsetroundcap%
\pgfsetroundjoin%
\pgfsetlinewidth{2.710125pt}%
\definecolor{currentstroke}{rgb}{0.260000,0.260000,0.260000}%
\pgfsetstrokecolor{currentstroke}%
\pgfsetdash{}{0pt}%
\pgfpathmoveto{\pgfqpoint{1.343902in}{1.544108in}}%
\pgfpathlineto{\pgfqpoint{1.343902in}{1.563040in}}%
\pgfusepath{stroke}%
\end{pgfscope}%
\begin{pgfscope}%
\pgfpathrectangle{\pgfqpoint{0.762652in}{0.445293in}}{\pgfqpoint{4.650000in}{3.020000in}}%
\pgfusepath{clip}%
\pgfsetroundcap%
\pgfsetroundjoin%
\pgfsetlinewidth{2.710125pt}%
\definecolor{currentstroke}{rgb}{0.260000,0.260000,0.260000}%
\pgfsetstrokecolor{currentstroke}%
\pgfsetdash{}{0pt}%
\pgfpathmoveto{\pgfqpoint{2.506402in}{2.730514in}}%
\pgfpathlineto{\pgfqpoint{2.506402in}{2.761759in}}%
\pgfusepath{stroke}%
\end{pgfscope}%
\begin{pgfscope}%
\pgfpathrectangle{\pgfqpoint{0.762652in}{0.445293in}}{\pgfqpoint{4.650000in}{3.020000in}}%
\pgfusepath{clip}%
\pgfsetroundcap%
\pgfsetroundjoin%
\pgfsetlinewidth{2.710125pt}%
\definecolor{currentstroke}{rgb}{0.260000,0.260000,0.260000}%
\pgfsetstrokecolor{currentstroke}%
\pgfsetdash{}{0pt}%
\pgfpathmoveto{\pgfqpoint{3.668902in}{2.742927in}}%
\pgfpathlineto{\pgfqpoint{3.668902in}{2.754522in}}%
\pgfusepath{stroke}%
\end{pgfscope}%
\begin{pgfscope}%
\pgfpathrectangle{\pgfqpoint{0.762652in}{0.445293in}}{\pgfqpoint{4.650000in}{3.020000in}}%
\pgfusepath{clip}%
\pgfsetroundcap%
\pgfsetroundjoin%
\pgfsetlinewidth{2.710125pt}%
\definecolor{currentstroke}{rgb}{0.260000,0.260000,0.260000}%
\pgfsetstrokecolor{currentstroke}%
\pgfsetdash{}{0pt}%
\pgfpathmoveto{\pgfqpoint{4.831402in}{1.305293in}}%
\pgfpathlineto{\pgfqpoint{4.831402in}{1.316078in}}%
\pgfusepath{stroke}%
\end{pgfscope}%
\begin{pgfscope}%
\pgfpathrectangle{\pgfqpoint{0.762652in}{0.445293in}}{\pgfqpoint{4.650000in}{3.020000in}}%
\pgfusepath{clip}%
\pgfsetroundcap%
\pgfsetroundjoin%
\pgfsetlinewidth{2.710125pt}%
\definecolor{currentstroke}{rgb}{0.260000,0.260000,0.260000}%
\pgfsetstrokecolor{currentstroke}%
\pgfsetdash{}{0pt}%
\pgfpathmoveto{\pgfqpoint{1.653902in}{1.582411in}}%
\pgfpathlineto{\pgfqpoint{1.653902in}{1.620109in}}%
\pgfusepath{stroke}%
\end{pgfscope}%
\begin{pgfscope}%
\pgfpathrectangle{\pgfqpoint{0.762652in}{0.445293in}}{\pgfqpoint{4.650000in}{3.020000in}}%
\pgfusepath{clip}%
\pgfsetroundcap%
\pgfsetroundjoin%
\pgfsetlinewidth{2.710125pt}%
\definecolor{currentstroke}{rgb}{0.260000,0.260000,0.260000}%
\pgfsetstrokecolor{currentstroke}%
\pgfsetdash{}{0pt}%
\pgfpathmoveto{\pgfqpoint{2.816402in}{0.926312in}}%
\pgfpathlineto{\pgfqpoint{2.816402in}{0.948649in}}%
\pgfusepath{stroke}%
\end{pgfscope}%
\begin{pgfscope}%
\pgfpathrectangle{\pgfqpoint{0.762652in}{0.445293in}}{\pgfqpoint{4.650000in}{3.020000in}}%
\pgfusepath{clip}%
\pgfsetroundcap%
\pgfsetroundjoin%
\pgfsetlinewidth{2.710125pt}%
\definecolor{currentstroke}{rgb}{0.260000,0.260000,0.260000}%
\pgfsetstrokecolor{currentstroke}%
\pgfsetdash{}{0pt}%
\pgfpathmoveto{\pgfqpoint{3.978902in}{2.083376in}}%
\pgfpathlineto{\pgfqpoint{3.978902in}{2.109922in}}%
\pgfusepath{stroke}%
\end{pgfscope}%
\begin{pgfscope}%
\pgfpathrectangle{\pgfqpoint{0.762652in}{0.445293in}}{\pgfqpoint{4.650000in}{3.020000in}}%
\pgfusepath{clip}%
\pgfsetroundcap%
\pgfsetroundjoin%
\pgfsetlinewidth{2.710125pt}%
\definecolor{currentstroke}{rgb}{0.260000,0.260000,0.260000}%
\pgfsetstrokecolor{currentstroke}%
\pgfsetdash{}{0pt}%
\pgfpathmoveto{\pgfqpoint{5.141402in}{1.188945in}}%
\pgfpathlineto{\pgfqpoint{5.141402in}{1.221314in}}%
\pgfusepath{stroke}%
\end{pgfscope}%
\begin{pgfscope}%
\pgfsetrectcap%
\pgfsetmiterjoin%
\pgfsetlinewidth{1.254687pt}%
\definecolor{currentstroke}{rgb}{0.800000,0.800000,0.800000}%
\pgfsetstrokecolor{currentstroke}%
\pgfsetdash{}{0pt}%
\pgfpathmoveto{\pgfqpoint{0.762652in}{0.445293in}}%
\pgfpathlineto{\pgfqpoint{0.762652in}{3.465293in}}%
\pgfusepath{stroke}%
\end{pgfscope}%
\begin{pgfscope}%
\pgfsetrectcap%
\pgfsetmiterjoin%
\pgfsetlinewidth{1.254687pt}%
\definecolor{currentstroke}{rgb}{0.800000,0.800000,0.800000}%
\pgfsetstrokecolor{currentstroke}%
\pgfsetdash{}{0pt}%
\pgfpathmoveto{\pgfqpoint{5.412652in}{0.445293in}}%
\pgfpathlineto{\pgfqpoint{5.412652in}{3.465293in}}%
\pgfusepath{stroke}%
\end{pgfscope}%
\begin{pgfscope}%
\pgfsetrectcap%
\pgfsetmiterjoin%
\pgfsetlinewidth{1.254687pt}%
\definecolor{currentstroke}{rgb}{0.800000,0.800000,0.800000}%
\pgfsetstrokecolor{currentstroke}%
\pgfsetdash{}{0pt}%
\pgfpathmoveto{\pgfqpoint{0.762652in}{0.445293in}}%
\pgfpathlineto{\pgfqpoint{5.412652in}{0.445293in}}%
\pgfusepath{stroke}%
\end{pgfscope}%
\begin{pgfscope}%
\pgfsetrectcap%
\pgfsetmiterjoin%
\pgfsetlinewidth{1.254687pt}%
\definecolor{currentstroke}{rgb}{0.800000,0.800000,0.800000}%
\pgfsetstrokecolor{currentstroke}%
\pgfsetdash{}{0pt}%
\pgfpathmoveto{\pgfqpoint{0.762652in}{3.465293in}}%
\pgfpathlineto{\pgfqpoint{5.412652in}{3.465293in}}%
\pgfusepath{stroke}%
\end{pgfscope}%
\begin{pgfscope}%
\pgfsetbuttcap%
\pgfsetmiterjoin%
\definecolor{currentfill}{rgb}{1.000000,1.000000,1.000000}%
\pgfsetfillcolor{currentfill}%
\pgfsetfillopacity{0.800000}%
\pgfsetlinewidth{1.003750pt}%
\definecolor{currentstroke}{rgb}{0.800000,0.800000,0.800000}%
\pgfsetstrokecolor{currentstroke}%
\pgfsetstrokeopacity{0.800000}%
\pgfsetdash{}{0pt}%
\pgfpathmoveto{\pgfqpoint{1.473216in}{2.954028in}}%
\pgfpathlineto{\pgfqpoint{5.252235in}{2.954028in}}%
\pgfpathquadraticcurveto{\pgfqpoint{5.298068in}{2.954028in}}{\pgfqpoint{5.298068in}{2.999861in}}%
\pgfpathlineto{\pgfqpoint{5.298068in}{3.304876in}}%
\pgfpathquadraticcurveto{\pgfqpoint{5.298068in}{3.350710in}}{\pgfqpoint{5.252235in}{3.350710in}}%
\pgfpathlineto{\pgfqpoint{1.473216in}{3.350710in}}%
\pgfpathquadraticcurveto{\pgfqpoint{1.427383in}{3.350710in}}{\pgfqpoint{1.427383in}{3.304876in}}%
\pgfpathlineto{\pgfqpoint{1.427383in}{2.999861in}}%
\pgfpathquadraticcurveto{\pgfqpoint{1.427383in}{2.954028in}}{\pgfqpoint{1.473216in}{2.954028in}}%
\pgfpathclose%
\pgfusepath{stroke,fill}%
\end{pgfscope}%
\begin{pgfscope}%
\pgfsetbuttcap%
\pgfsetmiterjoin%
\definecolor{currentfill}{rgb}{0.347059,0.458824,0.641176}%
\pgfsetfillcolor{currentfill}%
\pgfsetlinewidth{1.003750pt}%
\definecolor{currentstroke}{rgb}{1.000000,1.000000,1.000000}%
\pgfsetstrokecolor{currentstroke}%
\pgfsetdash{}{0pt}%
\pgfpathmoveto{\pgfqpoint{1.519049in}{3.092376in}}%
\pgfpathlineto{\pgfqpoint{1.977383in}{3.092376in}}%
\pgfpathlineto{\pgfqpoint{1.977383in}{3.252793in}}%
\pgfpathlineto{\pgfqpoint{1.519049in}{3.252793in}}%
\pgfpathclose%
\pgfusepath{stroke,fill}%
\end{pgfscope}%
\begin{pgfscope}%
\definecolor{textcolor}{rgb}{0.150000,0.150000,0.150000}%
\pgfsetstrokecolor{textcolor}%
\pgfsetfillcolor{textcolor}%
\pgftext[x=2.160716in,y=3.092376in,left,base]{\color{textcolor}\rmfamily\fontsize{16.500000}{19.800000}\selectfont \textsc{w-l}}%
\end{pgfscope}%
\begin{pgfscope}%
\pgfsetbuttcap%
\pgfsetmiterjoin%
\definecolor{currentfill}{rgb}{0.798529,0.536765,0.389706}%
\pgfsetfillcolor{currentfill}%
\pgfsetlinewidth{1.003750pt}%
\definecolor{currentstroke}{rgb}{1.000000,1.000000,1.000000}%
\pgfsetstrokecolor{currentstroke}%
\pgfsetdash{}{0pt}%
\pgfpathmoveto{\pgfqpoint{2.760905in}{3.092376in}}%
\pgfpathlineto{\pgfqpoint{3.219238in}{3.092376in}}%
\pgfpathlineto{\pgfqpoint{3.219238in}{3.252793in}}%
\pgfpathlineto{\pgfqpoint{2.760905in}{3.252793in}}%
\pgfpathclose%
\pgfusepath{stroke,fill}%
\end{pgfscope}%
\begin{pgfscope}%
\definecolor{textcolor}{rgb}{0.150000,0.150000,0.150000}%
\pgfsetstrokecolor{textcolor}%
\pgfsetfillcolor{textcolor}%
\pgftext[x=3.402572in,y=3.092376in,left,base]{\color{textcolor}\rmfamily\fontsize{16.500000}{19.800000}\selectfont \textsc{w-m}}%
\end{pgfscope}%
\begin{pgfscope}%
\pgfsetbuttcap%
\pgfsetmiterjoin%
\definecolor{currentfill}{rgb}{0.374020,0.618137,0.429902}%
\pgfsetfillcolor{currentfill}%
\pgfsetlinewidth{1.003750pt}%
\definecolor{currentstroke}{rgb}{1.000000,1.000000,1.000000}%
\pgfsetstrokecolor{currentstroke}%
\pgfsetdash{}{0pt}%
\pgfpathmoveto{\pgfqpoint{4.068540in}{3.092376in}}%
\pgfpathlineto{\pgfqpoint{4.526873in}{3.092376in}}%
\pgfpathlineto{\pgfqpoint{4.526873in}{3.252793in}}%
\pgfpathlineto{\pgfqpoint{4.068540in}{3.252793in}}%
\pgfpathclose%
\pgfusepath{stroke,fill}%
\end{pgfscope}%
\begin{pgfscope}%
\definecolor{textcolor}{rgb}{0.150000,0.150000,0.150000}%
\pgfsetstrokecolor{textcolor}%
\pgfsetfillcolor{textcolor}%
\pgftext[x=4.710206in,y=3.092376in,left,base]{\color{textcolor}\rmfamily\fontsize{16.500000}{19.800000}\selectfont \textsc{dcv}}%
\end{pgfscope}%
\end{pgfpicture}%
\makeatother%
\endgroup%

%% file: fig/continuous_circle-1000_res.pgf
%% Creator: Matplotlib, PGF backend
%%
%% To include the figure in your LaTeX document, write
%%   \input{<filename>.pgf}
%%
%% Make sure the required packages are loaded in your preamble
%%   \usepackage{pgf}
%%
%% Figures using additional raster images can only be included by \input if
%% they are in the same directory as the main LaTeX file. For loading figures
%% from other directories you can use the `import` package
%% and then include the figures with
%%   \import{<path to file>}{<filename>.pgf}
%%
%% Matplotlib used the following preamble
%%
\begingroup%
\makeatletter%
\begin{pgfpicture}%
\pgfpathrectangle{\pgfpointorigin}{\pgfqpoint{5.979207in}{4.035754in}}%
\pgfusepath{use as bounding box, clip}%
\begin{pgfscope}%
\pgfsetbuttcap%
\pgfsetmiterjoin%
\definecolor{currentfill}{rgb}{1.000000,1.000000,1.000000}%
\pgfsetfillcolor{currentfill}%
\pgfsetlinewidth{0.000000pt}%
\definecolor{currentstroke}{rgb}{1.000000,1.000000,1.000000}%
\pgfsetstrokecolor{currentstroke}%
\pgfsetdash{}{0pt}%
\pgfpathmoveto{\pgfqpoint{0.000000in}{0.000000in}}%
\pgfpathlineto{\pgfqpoint{5.979207in}{0.000000in}}%
\pgfpathlineto{\pgfqpoint{5.979207in}{4.035754in}}%
\pgfpathlineto{\pgfqpoint{0.000000in}{4.035754in}}%
\pgfpathclose%
\pgfusepath{fill}%
\end{pgfscope}%
\begin{pgfscope}%
\pgfsetbuttcap%
\pgfsetmiterjoin%
\definecolor{currentfill}{rgb}{1.000000,1.000000,1.000000}%
\pgfsetfillcolor{currentfill}%
\pgfsetlinewidth{0.000000pt}%
\definecolor{currentstroke}{rgb}{0.000000,0.000000,0.000000}%
\pgfsetstrokecolor{currentstroke}%
\pgfsetstrokeopacity{0.000000}%
\pgfsetdash{}{0pt}%
\pgfpathmoveto{\pgfqpoint{1.229207in}{0.915754in}}%
\pgfpathlineto{\pgfqpoint{5.879208in}{0.915754in}}%
\pgfpathlineto{\pgfqpoint{5.879208in}{3.935754in}}%
\pgfpathlineto{\pgfqpoint{1.229207in}{3.935754in}}%
\pgfpathclose%
\pgfusepath{fill}%
\end{pgfscope}%
\begin{pgfscope}%
\pgfpathrectangle{\pgfqpoint{1.229207in}{0.915754in}}{\pgfqpoint{4.650000in}{3.020000in}}%
\pgfusepath{clip}%
\pgfsetroundcap%
\pgfsetroundjoin%
\pgfsetlinewidth{1.003750pt}%
\definecolor{currentstroke}{rgb}{0.800000,0.800000,0.800000}%
\pgfsetstrokecolor{currentstroke}%
\pgfsetdash{}{0pt}%
\pgfpathmoveto{\pgfqpoint{1.478315in}{0.915754in}}%
\pgfpathlineto{\pgfqpoint{1.478315in}{3.935754in}}%
\pgfusepath{stroke}%
\end{pgfscope}%
\begin{pgfscope}%
\definecolor{textcolor}{rgb}{0.150000,0.150000,0.150000}%
\pgfsetstrokecolor{textcolor}%
\pgfsetfillcolor{textcolor}%
\pgftext[x=1.478315in,y=0.783810in,,top]{\color{textcolor}\rmfamily\fontsize{23.100000}{27.720000}\selectfont 10}%
\end{pgfscope}%
\begin{pgfscope}%
\pgfpathrectangle{\pgfqpoint{1.229207in}{0.915754in}}{\pgfqpoint{4.650000in}{3.020000in}}%
\pgfusepath{clip}%
\pgfsetroundcap%
\pgfsetroundjoin%
\pgfsetlinewidth{1.003750pt}%
\definecolor{currentstroke}{rgb}{0.800000,0.800000,0.800000}%
\pgfsetstrokecolor{currentstroke}%
\pgfsetdash{}{0pt}%
\pgfpathmoveto{\pgfqpoint{2.308672in}{0.915754in}}%
\pgfpathlineto{\pgfqpoint{2.308672in}{3.935754in}}%
\pgfusepath{stroke}%
\end{pgfscope}%
\begin{pgfscope}%
\definecolor{textcolor}{rgb}{0.150000,0.150000,0.150000}%
\pgfsetstrokecolor{textcolor}%
\pgfsetfillcolor{textcolor}%
\pgftext[x=2.308672in,y=0.783810in,,top]{\color{textcolor}\rmfamily\fontsize{23.100000}{27.720000}\selectfont 20}%
\end{pgfscope}%
\begin{pgfscope}%
\pgfpathrectangle{\pgfqpoint{1.229207in}{0.915754in}}{\pgfqpoint{4.650000in}{3.020000in}}%
\pgfusepath{clip}%
\pgfsetroundcap%
\pgfsetroundjoin%
\pgfsetlinewidth{1.003750pt}%
\definecolor{currentstroke}{rgb}{0.800000,0.800000,0.800000}%
\pgfsetstrokecolor{currentstroke}%
\pgfsetdash{}{0pt}%
\pgfpathmoveto{\pgfqpoint{3.139029in}{0.915754in}}%
\pgfpathlineto{\pgfqpoint{3.139029in}{3.935754in}}%
\pgfusepath{stroke}%
\end{pgfscope}%
\begin{pgfscope}%
\definecolor{textcolor}{rgb}{0.150000,0.150000,0.150000}%
\pgfsetstrokecolor{textcolor}%
\pgfsetfillcolor{textcolor}%
\pgftext[x=3.139029in,y=0.783810in,,top]{\color{textcolor}\rmfamily\fontsize{23.100000}{27.720000}\selectfont 30}%
\end{pgfscope}%
\begin{pgfscope}%
\pgfpathrectangle{\pgfqpoint{1.229207in}{0.915754in}}{\pgfqpoint{4.650000in}{3.020000in}}%
\pgfusepath{clip}%
\pgfsetroundcap%
\pgfsetroundjoin%
\pgfsetlinewidth{1.003750pt}%
\definecolor{currentstroke}{rgb}{0.800000,0.800000,0.800000}%
\pgfsetstrokecolor{currentstroke}%
\pgfsetdash{}{0pt}%
\pgfpathmoveto{\pgfqpoint{3.969386in}{0.915754in}}%
\pgfpathlineto{\pgfqpoint{3.969386in}{3.935754in}}%
\pgfusepath{stroke}%
\end{pgfscope}%
\begin{pgfscope}%
\definecolor{textcolor}{rgb}{0.150000,0.150000,0.150000}%
\pgfsetstrokecolor{textcolor}%
\pgfsetfillcolor{textcolor}%
\pgftext[x=3.969386in,y=0.783810in,,top]{\color{textcolor}\rmfamily\fontsize{23.100000}{27.720000}\selectfont 40}%
\end{pgfscope}%
\begin{pgfscope}%
\pgfpathrectangle{\pgfqpoint{1.229207in}{0.915754in}}{\pgfqpoint{4.650000in}{3.020000in}}%
\pgfusepath{clip}%
\pgfsetroundcap%
\pgfsetroundjoin%
\pgfsetlinewidth{1.003750pt}%
\definecolor{currentstroke}{rgb}{0.800000,0.800000,0.800000}%
\pgfsetstrokecolor{currentstroke}%
\pgfsetdash{}{0pt}%
\pgfpathmoveto{\pgfqpoint{4.799743in}{0.915754in}}%
\pgfpathlineto{\pgfqpoint{4.799743in}{3.935754in}}%
\pgfusepath{stroke}%
\end{pgfscope}%
\begin{pgfscope}%
\definecolor{textcolor}{rgb}{0.150000,0.150000,0.150000}%
\pgfsetstrokecolor{textcolor}%
\pgfsetfillcolor{textcolor}%
\pgftext[x=4.799743in,y=0.783810in,,top]{\color{textcolor}\rmfamily\fontsize{23.100000}{27.720000}\selectfont 50}%
\end{pgfscope}%
\begin{pgfscope}%
\pgfpathrectangle{\pgfqpoint{1.229207in}{0.915754in}}{\pgfqpoint{4.650000in}{3.020000in}}%
\pgfusepath{clip}%
\pgfsetroundcap%
\pgfsetroundjoin%
\pgfsetlinewidth{1.003750pt}%
\definecolor{currentstroke}{rgb}{0.800000,0.800000,0.800000}%
\pgfsetstrokecolor{currentstroke}%
\pgfsetdash{}{0pt}%
\pgfpathmoveto{\pgfqpoint{5.630100in}{0.915754in}}%
\pgfpathlineto{\pgfqpoint{5.630100in}{3.935754in}}%
\pgfusepath{stroke}%
\end{pgfscope}%
\begin{pgfscope}%
\definecolor{textcolor}{rgb}{0.150000,0.150000,0.150000}%
\pgfsetstrokecolor{textcolor}%
\pgfsetfillcolor{textcolor}%
\pgftext[x=5.630100in,y=0.783810in,,top]{\color{textcolor}\rmfamily\fontsize{23.100000}{27.720000}\selectfont 60}%
\end{pgfscope}%
\begin{pgfscope}%
\definecolor{textcolor}{rgb}{0.150000,0.150000,0.150000}%
\pgfsetstrokecolor{textcolor}%
\pgfsetfillcolor{textcolor}%
\pgftext[x=3.554208in,y=0.407183in,,top]{\color{textcolor}\rmfamily\fontsize{25.200000}{30.240000}\selectfont runtime in seconds}%
\end{pgfscope}%
\begin{pgfscope}%
\pgfpathrectangle{\pgfqpoint{1.229207in}{0.915754in}}{\pgfqpoint{4.650000in}{3.020000in}}%
\pgfusepath{clip}%
\pgfsetroundcap%
\pgfsetroundjoin%
\pgfsetlinewidth{1.003750pt}%
\definecolor{currentstroke}{rgb}{0.800000,0.800000,0.800000}%
\pgfsetstrokecolor{currentstroke}%
\pgfsetdash{}{0pt}%
\pgfpathmoveto{\pgfqpoint{1.229207in}{1.328214in}}%
\pgfpathlineto{\pgfqpoint{5.879208in}{1.328214in}}%
\pgfusepath{stroke}%
\end{pgfscope}%
\begin{pgfscope}%
\definecolor{textcolor}{rgb}{0.150000,0.150000,0.150000}%
\pgfsetstrokecolor{textcolor}%
\pgfsetfillcolor{textcolor}%
\pgftext[x=0.476627in,y=1.208229in,left,base]{\color{textcolor}\rmfamily\fontsize{23.100000}{27.720000}\selectfont \(\displaystyle {10^{-2}}\)}%
\end{pgfscope}%
\begin{pgfscope}%
\pgfpathrectangle{\pgfqpoint{1.229207in}{0.915754in}}{\pgfqpoint{4.650000in}{3.020000in}}%
\pgfusepath{clip}%
\pgfsetroundcap%
\pgfsetroundjoin%
\pgfsetlinewidth{1.003750pt}%
\definecolor{currentstroke}{rgb}{0.800000,0.800000,0.800000}%
\pgfsetstrokecolor{currentstroke}%
\pgfsetdash{}{0pt}%
\pgfpathmoveto{\pgfqpoint{1.229207in}{2.698375in}}%
\pgfpathlineto{\pgfqpoint{5.879208in}{2.698375in}}%
\pgfusepath{stroke}%
\end{pgfscope}%
\begin{pgfscope}%
\definecolor{textcolor}{rgb}{0.150000,0.150000,0.150000}%
\pgfsetstrokecolor{textcolor}%
\pgfsetfillcolor{textcolor}%
\pgftext[x=0.476627in,y=2.578391in,left,base]{\color{textcolor}\rmfamily\fontsize{23.100000}{27.720000}\selectfont \(\displaystyle {10^{-1}}\)}%
\end{pgfscope}%
\begin{pgfscope}%
\definecolor{textcolor}{rgb}{0.150000,0.150000,0.150000}%
\pgfsetstrokecolor{textcolor}%
\pgfsetfillcolor{textcolor}%
\pgftext[x=0.407183in,y=2.425754in,,bottom,rotate=90.000000]{\color{textcolor}\rmfamily\fontsize{25.200000}{30.240000}\selectfont relative error}%
\end{pgfscope}%
\begin{pgfscope}%
\pgfpathrectangle{\pgfqpoint{1.229207in}{0.915754in}}{\pgfqpoint{4.650000in}{3.020000in}}%
\pgfusepath{clip}%
\pgfsetbuttcap%
\pgfsetroundjoin%
\definecolor{currentfill}{rgb}{0.298039,0.447059,0.690196}%
\pgfsetfillcolor{currentfill}%
\pgfsetfillopacity{0.200000}%
\pgfsetlinewidth{1.003750pt}%
\definecolor{currentstroke}{rgb}{0.298039,0.447059,0.690196}%
\pgfsetstrokecolor{currentstroke}%
\pgfsetstrokeopacity{0.200000}%
\pgfsetdash{}{0pt}%
\pgfpathmoveto{\pgfqpoint{1.103874in}{2.341424in}}%
\pgfpathlineto{\pgfqpoint{1.103874in}{2.012625in}}%
\pgfpathlineto{\pgfqpoint{1.278694in}{1.987244in}}%
\pgfpathlineto{\pgfqpoint{1.457176in}{1.904814in}}%
\pgfpathlineto{\pgfqpoint{1.635598in}{1.444835in}}%
\pgfpathlineto{\pgfqpoint{1.801986in}{1.746578in}}%
\pgfpathlineto{\pgfqpoint{1.982367in}{1.446591in}}%
\pgfpathlineto{\pgfqpoint{2.160208in}{1.601713in}}%
\pgfpathlineto{\pgfqpoint{2.335237in}{1.651246in}}%
\pgfpathlineto{\pgfqpoint{2.512035in}{1.737936in}}%
\pgfpathlineto{\pgfqpoint{2.686149in}{1.712185in}}%
\pgfpathlineto{\pgfqpoint{2.864490in}{1.508111in}}%
\pgfpathlineto{\pgfqpoint{3.040641in}{1.737972in}}%
\pgfpathlineto{\pgfqpoint{3.210816in}{1.726693in}}%
\pgfpathlineto{\pgfqpoint{3.388967in}{1.710489in}}%
\pgfpathlineto{\pgfqpoint{3.571066in}{1.830798in}}%
\pgfpathlineto{\pgfqpoint{3.743642in}{1.769902in}}%
\pgfpathlineto{\pgfqpoint{3.924748in}{1.901193in}}%
\pgfpathlineto{\pgfqpoint{4.094863in}{1.770869in}}%
\pgfpathlineto{\pgfqpoint{4.275599in}{1.670534in}}%
\pgfpathlineto{\pgfqpoint{4.452309in}{1.629495in}}%
\pgfpathlineto{\pgfqpoint{4.629138in}{1.616004in}}%
\pgfpathlineto{\pgfqpoint{4.806134in}{1.608538in}}%
\pgfpathlineto{\pgfqpoint{4.984390in}{1.669388in}}%
\pgfpathlineto{\pgfqpoint{5.179461in}{1.765055in}}%
\pgfpathlineto{\pgfqpoint{5.351963in}{1.618998in}}%
\pgfpathlineto{\pgfqpoint{5.532511in}{1.639554in}}%
\pgfpathlineto{\pgfqpoint{5.703570in}{1.617680in}}%
\pgfpathlineto{\pgfqpoint{5.884987in}{1.590347in}}%
\pgfpathlineto{\pgfqpoint{6.056972in}{1.465585in}}%
\pgfpathlineto{\pgfqpoint{6.235442in}{1.624065in}}%
\pgfpathlineto{\pgfqpoint{6.235442in}{1.883470in}}%
\pgfpathlineto{\pgfqpoint{6.235442in}{1.883470in}}%
\pgfpathlineto{\pgfqpoint{6.056972in}{1.756076in}}%
\pgfpathlineto{\pgfqpoint{5.884987in}{1.905635in}}%
\pgfpathlineto{\pgfqpoint{5.703570in}{1.940392in}}%
\pgfpathlineto{\pgfqpoint{5.532511in}{1.942409in}}%
\pgfpathlineto{\pgfqpoint{5.351963in}{1.902109in}}%
\pgfpathlineto{\pgfqpoint{5.179461in}{2.064265in}}%
\pgfpathlineto{\pgfqpoint{4.984390in}{1.930683in}}%
\pgfpathlineto{\pgfqpoint{4.806134in}{1.957988in}}%
\pgfpathlineto{\pgfqpoint{4.629138in}{1.922670in}}%
\pgfpathlineto{\pgfqpoint{4.452309in}{1.885953in}}%
\pgfpathlineto{\pgfqpoint{4.275599in}{1.923276in}}%
\pgfpathlineto{\pgfqpoint{4.094863in}{2.079783in}}%
\pgfpathlineto{\pgfqpoint{3.924748in}{2.180050in}}%
\pgfpathlineto{\pgfqpoint{3.743642in}{2.120040in}}%
\pgfpathlineto{\pgfqpoint{3.571066in}{2.108445in}}%
\pgfpathlineto{\pgfqpoint{3.388967in}{2.057586in}}%
\pgfpathlineto{\pgfqpoint{3.210816in}{2.036672in}}%
\pgfpathlineto{\pgfqpoint{3.040641in}{2.070349in}}%
\pgfpathlineto{\pgfqpoint{2.864490in}{1.869933in}}%
\pgfpathlineto{\pgfqpoint{2.686149in}{2.041961in}}%
\pgfpathlineto{\pgfqpoint{2.512035in}{1.998399in}}%
\pgfpathlineto{\pgfqpoint{2.335237in}{2.004549in}}%
\pgfpathlineto{\pgfqpoint{2.160208in}{1.985137in}}%
\pgfpathlineto{\pgfqpoint{1.982367in}{1.838453in}}%
\pgfpathlineto{\pgfqpoint{1.801986in}{2.031684in}}%
\pgfpathlineto{\pgfqpoint{1.635598in}{1.794343in}}%
\pgfpathlineto{\pgfqpoint{1.457176in}{2.174142in}}%
\pgfpathlineto{\pgfqpoint{1.278694in}{2.205097in}}%
\pgfpathlineto{\pgfqpoint{1.103874in}{2.341424in}}%
\pgfpathclose%
\pgfusepath{stroke,fill}%
\end{pgfscope}%
\begin{pgfscope}%
\pgfpathrectangle{\pgfqpoint{1.229207in}{0.915754in}}{\pgfqpoint{4.650000in}{3.020000in}}%
\pgfusepath{clip}%
\pgfsetbuttcap%
\pgfsetroundjoin%
\definecolor{currentfill}{rgb}{0.866667,0.517647,0.321569}%
\pgfsetfillcolor{currentfill}%
\pgfsetfillopacity{0.200000}%
\pgfsetlinewidth{1.003750pt}%
\definecolor{currentstroke}{rgb}{0.866667,0.517647,0.321569}%
\pgfsetstrokecolor{currentstroke}%
\pgfsetstrokeopacity{0.200000}%
\pgfsetdash{}{0pt}%
\pgfpathmoveto{\pgfqpoint{1.103439in}{2.877638in}}%
\pgfpathlineto{\pgfqpoint{1.103439in}{2.241524in}}%
\pgfpathlineto{\pgfqpoint{1.280202in}{1.768093in}}%
\pgfpathlineto{\pgfqpoint{1.458894in}{1.552788in}}%
\pgfpathlineto{\pgfqpoint{1.627647in}{2.008920in}}%
\pgfpathlineto{\pgfqpoint{1.804594in}{2.013214in}}%
\pgfpathlineto{\pgfqpoint{1.979562in}{1.926472in}}%
\pgfpathlineto{\pgfqpoint{2.152541in}{2.121339in}}%
\pgfpathlineto{\pgfqpoint{2.336959in}{1.935992in}}%
\pgfpathlineto{\pgfqpoint{2.511175in}{1.987096in}}%
\pgfpathlineto{\pgfqpoint{2.685677in}{1.778690in}}%
\pgfpathlineto{\pgfqpoint{2.864498in}{2.037094in}}%
\pgfpathlineto{\pgfqpoint{3.039720in}{1.889242in}}%
\pgfpathlineto{\pgfqpoint{3.215436in}{1.844753in}}%
\pgfpathlineto{\pgfqpoint{3.390580in}{1.916743in}}%
\pgfpathlineto{\pgfqpoint{3.574503in}{1.687953in}}%
\pgfpathlineto{\pgfqpoint{3.747011in}{1.893345in}}%
\pgfpathlineto{\pgfqpoint{3.924376in}{1.717109in}}%
\pgfpathlineto{\pgfqpoint{4.097909in}{1.809800in}}%
\pgfpathlineto{\pgfqpoint{4.277942in}{1.789101in}}%
\pgfpathlineto{\pgfqpoint{4.457364in}{1.951908in}}%
\pgfpathlineto{\pgfqpoint{4.629898in}{1.733993in}}%
\pgfpathlineto{\pgfqpoint{4.810888in}{1.882709in}}%
\pgfpathlineto{\pgfqpoint{4.990903in}{1.985456in}}%
\pgfpathlineto{\pgfqpoint{5.173143in}{2.020349in}}%
\pgfpathlineto{\pgfqpoint{5.348261in}{1.779142in}}%
\pgfpathlineto{\pgfqpoint{5.528637in}{1.922000in}}%
\pgfpathlineto{\pgfqpoint{5.705495in}{1.877237in}}%
\pgfpathlineto{\pgfqpoint{5.884432in}{1.897731in}}%
\pgfpathlineto{\pgfqpoint{6.061464in}{1.781290in}}%
\pgfpathlineto{\pgfqpoint{6.238675in}{1.738953in}}%
\pgfpathlineto{\pgfqpoint{6.238675in}{2.100018in}}%
\pgfpathlineto{\pgfqpoint{6.238675in}{2.100018in}}%
\pgfpathlineto{\pgfqpoint{6.061464in}{2.138327in}}%
\pgfpathlineto{\pgfqpoint{5.884432in}{2.149288in}}%
\pgfpathlineto{\pgfqpoint{5.705495in}{2.184376in}}%
\pgfpathlineto{\pgfqpoint{5.528637in}{2.307360in}}%
\pgfpathlineto{\pgfqpoint{5.348261in}{2.118060in}}%
\pgfpathlineto{\pgfqpoint{5.173143in}{2.324586in}}%
\pgfpathlineto{\pgfqpoint{4.990903in}{2.321094in}}%
\pgfpathlineto{\pgfqpoint{4.810888in}{2.205218in}}%
\pgfpathlineto{\pgfqpoint{4.629898in}{2.106114in}}%
\pgfpathlineto{\pgfqpoint{4.457364in}{2.241748in}}%
\pgfpathlineto{\pgfqpoint{4.277942in}{2.289754in}}%
\pgfpathlineto{\pgfqpoint{4.097909in}{2.230874in}}%
\pgfpathlineto{\pgfqpoint{3.924376in}{2.030019in}}%
\pgfpathlineto{\pgfqpoint{3.747011in}{2.303224in}}%
\pgfpathlineto{\pgfqpoint{3.574503in}{2.044387in}}%
\pgfpathlineto{\pgfqpoint{3.390580in}{2.238199in}}%
\pgfpathlineto{\pgfqpoint{3.215436in}{2.215558in}}%
\pgfpathlineto{\pgfqpoint{3.039720in}{2.257052in}}%
\pgfpathlineto{\pgfqpoint{2.864498in}{2.316482in}}%
\pgfpathlineto{\pgfqpoint{2.685677in}{2.181353in}}%
\pgfpathlineto{\pgfqpoint{2.511175in}{2.236813in}}%
\pgfpathlineto{\pgfqpoint{2.336959in}{2.315762in}}%
\pgfpathlineto{\pgfqpoint{2.152541in}{2.394600in}}%
\pgfpathlineto{\pgfqpoint{1.979562in}{2.235333in}}%
\pgfpathlineto{\pgfqpoint{1.804594in}{2.350004in}}%
\pgfpathlineto{\pgfqpoint{1.627647in}{2.377530in}}%
\pgfpathlineto{\pgfqpoint{1.458894in}{1.948447in}}%
\pgfpathlineto{\pgfqpoint{1.280202in}{2.180949in}}%
\pgfpathlineto{\pgfqpoint{1.103439in}{2.877638in}}%
\pgfpathclose%
\pgfusepath{stroke,fill}%
\end{pgfscope}%
\begin{pgfscope}%
\pgfpathrectangle{\pgfqpoint{1.229207in}{0.915754in}}{\pgfqpoint{4.650000in}{3.020000in}}%
\pgfusepath{clip}%
\pgfsetbuttcap%
\pgfsetroundjoin%
\definecolor{currentfill}{rgb}{0.333333,0.658824,0.407843}%
\pgfsetfillcolor{currentfill}%
\pgfsetfillopacity{0.200000}%
\pgfsetlinewidth{1.003750pt}%
\definecolor{currentstroke}{rgb}{0.333333,0.658824,0.407843}%
\pgfsetstrokecolor{currentstroke}%
\pgfsetstrokeopacity{0.200000}%
\pgfsetdash{}{0pt}%
\pgfpathmoveto{\pgfqpoint{1.159646in}{3.676940in}}%
\pgfpathlineto{\pgfqpoint{1.159646in}{2.983100in}}%
\pgfpathlineto{\pgfqpoint{1.342404in}{2.527125in}}%
\pgfpathlineto{\pgfqpoint{1.520378in}{2.411863in}}%
\pgfpathlineto{\pgfqpoint{1.712457in}{2.419209in}}%
\pgfpathlineto{\pgfqpoint{1.896077in}{2.243647in}}%
\pgfpathlineto{\pgfqpoint{2.059563in}{2.175275in}}%
\pgfpathlineto{\pgfqpoint{2.253724in}{2.071074in}}%
\pgfpathlineto{\pgfqpoint{2.435330in}{2.009757in}}%
\pgfpathlineto{\pgfqpoint{2.605206in}{2.162901in}}%
\pgfpathlineto{\pgfqpoint{2.802842in}{2.084444in}}%
\pgfpathlineto{\pgfqpoint{2.982234in}{2.062571in}}%
\pgfpathlineto{\pgfqpoint{3.141515in}{1.994850in}}%
\pgfpathlineto{\pgfqpoint{3.337853in}{2.134925in}}%
\pgfpathlineto{\pgfqpoint{3.490970in}{1.776935in}}%
\pgfpathlineto{\pgfqpoint{3.711765in}{2.000862in}}%
\pgfpathlineto{\pgfqpoint{3.909236in}{1.978351in}}%
\pgfpathlineto{\pgfqpoint{4.062994in}{1.904339in}}%
\pgfpathlineto{\pgfqpoint{4.258508in}{1.992409in}}%
\pgfpathlineto{\pgfqpoint{4.438295in}{1.976410in}}%
\pgfpathlineto{\pgfqpoint{4.618481in}{2.083964in}}%
\pgfpathlineto{\pgfqpoint{4.780693in}{1.809223in}}%
\pgfpathlineto{\pgfqpoint{4.985800in}{1.951264in}}%
\pgfpathlineto{\pgfqpoint{5.151776in}{1.902958in}}%
\pgfpathlineto{\pgfqpoint{5.364386in}{1.868051in}}%
\pgfpathlineto{\pgfqpoint{5.500672in}{1.733871in}}%
\pgfpathlineto{\pgfqpoint{5.675271in}{1.761704in}}%
\pgfpathlineto{\pgfqpoint{5.867141in}{1.758726in}}%
\pgfpathlineto{\pgfqpoint{6.050413in}{1.756356in}}%
\pgfpathlineto{\pgfqpoint{6.237552in}{1.700237in}}%
\pgfpathlineto{\pgfqpoint{6.435487in}{1.738140in}}%
\pgfpathlineto{\pgfqpoint{6.435487in}{1.977746in}}%
\pgfpathlineto{\pgfqpoint{6.435487in}{1.977746in}}%
\pgfpathlineto{\pgfqpoint{6.237552in}{2.011375in}}%
\pgfpathlineto{\pgfqpoint{6.050413in}{2.045998in}}%
\pgfpathlineto{\pgfqpoint{5.867141in}{2.039269in}}%
\pgfpathlineto{\pgfqpoint{5.675271in}{2.060684in}}%
\pgfpathlineto{\pgfqpoint{5.500672in}{2.067733in}}%
\pgfpathlineto{\pgfqpoint{5.364386in}{2.188174in}}%
\pgfpathlineto{\pgfqpoint{5.151776in}{2.198187in}}%
\pgfpathlineto{\pgfqpoint{4.985800in}{2.213396in}}%
\pgfpathlineto{\pgfqpoint{4.780693in}{2.187534in}}%
\pgfpathlineto{\pgfqpoint{4.618481in}{2.317463in}}%
\pgfpathlineto{\pgfqpoint{4.438295in}{2.275154in}}%
\pgfpathlineto{\pgfqpoint{4.258508in}{2.273099in}}%
\pgfpathlineto{\pgfqpoint{4.062994in}{2.225709in}}%
\pgfpathlineto{\pgfqpoint{3.909236in}{2.215580in}}%
\pgfpathlineto{\pgfqpoint{3.711765in}{2.331814in}}%
\pgfpathlineto{\pgfqpoint{3.490970in}{2.185074in}}%
\pgfpathlineto{\pgfqpoint{3.337853in}{2.447275in}}%
\pgfpathlineto{\pgfqpoint{3.141515in}{2.364627in}}%
\pgfpathlineto{\pgfqpoint{2.982234in}{2.337594in}}%
\pgfpathlineto{\pgfqpoint{2.802842in}{2.399798in}}%
\pgfpathlineto{\pgfqpoint{2.605206in}{2.397757in}}%
\pgfpathlineto{\pgfqpoint{2.435330in}{2.333172in}}%
\pgfpathlineto{\pgfqpoint{2.253724in}{2.381289in}}%
\pgfpathlineto{\pgfqpoint{2.059563in}{2.494831in}}%
\pgfpathlineto{\pgfqpoint{1.896077in}{2.638328in}}%
\pgfpathlineto{\pgfqpoint{1.712457in}{2.740025in}}%
\pgfpathlineto{\pgfqpoint{1.520378in}{3.027633in}}%
\pgfpathlineto{\pgfqpoint{1.342404in}{3.712066in}}%
\pgfpathlineto{\pgfqpoint{1.159646in}{3.676940in}}%
\pgfpathclose%
\pgfusepath{stroke,fill}%
\end{pgfscope}%
\begin{pgfscope}%
\pgfpathrectangle{\pgfqpoint{1.229207in}{0.915754in}}{\pgfqpoint{4.650000in}{3.020000in}}%
\pgfusepath{clip}%
\pgfsetbuttcap%
\pgfsetroundjoin%
\definecolor{currentfill}{rgb}{0.768627,0.305882,0.321569}%
\pgfsetfillcolor{currentfill}%
\pgfsetfillopacity{0.200000}%
\pgfsetlinewidth{1.003750pt}%
\definecolor{currentstroke}{rgb}{0.768627,0.305882,0.321569}%
\pgfsetstrokecolor{currentstroke}%
\pgfsetstrokeopacity{0.200000}%
\pgfsetdash{}{0pt}%
\pgfpathmoveto{\pgfqpoint{1.063136in}{3.431425in}}%
\pgfpathlineto{\pgfqpoint{1.063136in}{3.086228in}}%
\pgfpathlineto{\pgfqpoint{1.234934in}{3.085120in}}%
\pgfpathlineto{\pgfqpoint{1.406732in}{2.990748in}}%
\pgfpathlineto{\pgfqpoint{1.578530in}{2.977720in}}%
\pgfpathlineto{\pgfqpoint{1.750328in}{2.900667in}}%
\pgfpathlineto{\pgfqpoint{1.922126in}{2.797798in}}%
\pgfpathlineto{\pgfqpoint{2.093924in}{2.759694in}}%
\pgfpathlineto{\pgfqpoint{2.265722in}{2.633265in}}%
\pgfpathlineto{\pgfqpoint{2.437520in}{2.722640in}}%
\pgfpathlineto{\pgfqpoint{2.609318in}{2.817359in}}%
\pgfpathlineto{\pgfqpoint{2.781116in}{2.704093in}}%
\pgfpathlineto{\pgfqpoint{2.952914in}{2.795426in}}%
\pgfpathlineto{\pgfqpoint{3.124712in}{2.595257in}}%
\pgfpathlineto{\pgfqpoint{3.296510in}{2.461223in}}%
\pgfpathlineto{\pgfqpoint{3.468308in}{2.752416in}}%
\pgfpathlineto{\pgfqpoint{3.640107in}{2.598368in}}%
\pgfpathlineto{\pgfqpoint{3.811905in}{2.565697in}}%
\pgfpathlineto{\pgfqpoint{3.983703in}{2.600312in}}%
\pgfpathlineto{\pgfqpoint{4.155501in}{2.655681in}}%
\pgfpathlineto{\pgfqpoint{4.327299in}{2.455872in}}%
\pgfpathlineto{\pgfqpoint{4.499097in}{2.525346in}}%
\pgfpathlineto{\pgfqpoint{4.670895in}{2.469543in}}%
\pgfpathlineto{\pgfqpoint{4.842693in}{2.571746in}}%
\pgfpathlineto{\pgfqpoint{5.014491in}{2.616350in}}%
\pgfpathlineto{\pgfqpoint{5.186289in}{2.351721in}}%
\pgfpathlineto{\pgfqpoint{5.358087in}{2.528401in}}%
\pgfpathlineto{\pgfqpoint{5.529885in}{2.358393in}}%
\pgfpathlineto{\pgfqpoint{5.701683in}{2.319044in}}%
\pgfpathlineto{\pgfqpoint{5.873481in}{2.370037in}}%
\pgfpathlineto{\pgfqpoint{6.045279in}{2.511084in}}%
\pgfpathlineto{\pgfqpoint{6.045279in}{2.814248in}}%
\pgfpathlineto{\pgfqpoint{6.045279in}{2.814248in}}%
\pgfpathlineto{\pgfqpoint{5.873481in}{2.748003in}}%
\pgfpathlineto{\pgfqpoint{5.701683in}{2.683085in}}%
\pgfpathlineto{\pgfqpoint{5.529885in}{2.607375in}}%
\pgfpathlineto{\pgfqpoint{5.358087in}{2.804090in}}%
\pgfpathlineto{\pgfqpoint{5.186289in}{2.689907in}}%
\pgfpathlineto{\pgfqpoint{5.014491in}{2.941974in}}%
\pgfpathlineto{\pgfqpoint{4.842693in}{2.869599in}}%
\pgfpathlineto{\pgfqpoint{4.670895in}{2.825962in}}%
\pgfpathlineto{\pgfqpoint{4.499097in}{2.838562in}}%
\pgfpathlineto{\pgfqpoint{4.327299in}{2.772691in}}%
\pgfpathlineto{\pgfqpoint{4.155501in}{2.970707in}}%
\pgfpathlineto{\pgfqpoint{3.983703in}{2.955907in}}%
\pgfpathlineto{\pgfqpoint{3.811905in}{2.900376in}}%
\pgfpathlineto{\pgfqpoint{3.640107in}{2.889982in}}%
\pgfpathlineto{\pgfqpoint{3.468308in}{3.043108in}}%
\pgfpathlineto{\pgfqpoint{3.296510in}{2.838758in}}%
\pgfpathlineto{\pgfqpoint{3.124712in}{2.987396in}}%
\pgfpathlineto{\pgfqpoint{2.952914in}{3.085676in}}%
\pgfpathlineto{\pgfqpoint{2.781116in}{2.997870in}}%
\pgfpathlineto{\pgfqpoint{2.609318in}{3.097176in}}%
\pgfpathlineto{\pgfqpoint{2.437520in}{3.025779in}}%
\pgfpathlineto{\pgfqpoint{2.265722in}{2.984173in}}%
\pgfpathlineto{\pgfqpoint{2.093924in}{3.067356in}}%
\pgfpathlineto{\pgfqpoint{1.922126in}{3.098596in}}%
\pgfpathlineto{\pgfqpoint{1.750328in}{3.227237in}}%
\pgfpathlineto{\pgfqpoint{1.578530in}{3.283764in}}%
\pgfpathlineto{\pgfqpoint{1.406732in}{3.311449in}}%
\pgfpathlineto{\pgfqpoint{1.234934in}{3.380678in}}%
\pgfpathlineto{\pgfqpoint{1.063136in}{3.431425in}}%
\pgfpathclose%
\pgfusepath{stroke,fill}%
\end{pgfscope}%
\begin{pgfscope}%
\pgfpathrectangle{\pgfqpoint{1.229207in}{0.915754in}}{\pgfqpoint{4.650000in}{3.020000in}}%
\pgfusepath{clip}%
\pgfsetroundcap%
\pgfsetroundjoin%
\pgfsetlinewidth{1.505625pt}%
\definecolor{currentstroke}{rgb}{0.298039,0.447059,0.690196}%
\pgfsetstrokecolor{currentstroke}%
\pgfsetdash{}{0pt}%
\pgfpathmoveto{\pgfqpoint{1.215319in}{2.141312in}}%
\pgfpathlineto{\pgfqpoint{1.278694in}{2.109687in}}%
\pgfpathlineto{\pgfqpoint{1.457176in}{2.042654in}}%
\pgfpathlineto{\pgfqpoint{1.635598in}{1.630027in}}%
\pgfpathlineto{\pgfqpoint{1.801986in}{1.901140in}}%
\pgfpathlineto{\pgfqpoint{1.982367in}{1.682711in}}%
\pgfpathlineto{\pgfqpoint{2.160208in}{1.817121in}}%
\pgfpathlineto{\pgfqpoint{2.335237in}{1.849499in}}%
\pgfpathlineto{\pgfqpoint{2.512035in}{1.885375in}}%
\pgfpathlineto{\pgfqpoint{2.686149in}{1.897455in}}%
\pgfpathlineto{\pgfqpoint{2.864490in}{1.706189in}}%
\pgfpathlineto{\pgfqpoint{3.040641in}{1.926703in}}%
\pgfpathlineto{\pgfqpoint{3.210816in}{1.907297in}}%
\pgfpathlineto{\pgfqpoint{3.388967in}{1.908221in}}%
\pgfpathlineto{\pgfqpoint{3.571066in}{1.986851in}}%
\pgfpathlineto{\pgfqpoint{3.743642in}{1.962378in}}%
\pgfpathlineto{\pgfqpoint{3.924748in}{2.055999in}}%
\pgfpathlineto{\pgfqpoint{4.094863in}{1.938415in}}%
\pgfpathlineto{\pgfqpoint{4.275599in}{1.816976in}}%
\pgfpathlineto{\pgfqpoint{4.452309in}{1.772071in}}%
\pgfpathlineto{\pgfqpoint{4.629138in}{1.786797in}}%
\pgfpathlineto{\pgfqpoint{4.806134in}{1.802846in}}%
\pgfpathlineto{\pgfqpoint{4.984390in}{1.814988in}}%
\pgfpathlineto{\pgfqpoint{5.179461in}{1.923206in}}%
\pgfpathlineto{\pgfqpoint{5.351963in}{1.774221in}}%
\pgfpathlineto{\pgfqpoint{5.532511in}{1.815779in}}%
\pgfpathlineto{\pgfqpoint{5.703570in}{1.798902in}}%
\pgfpathlineto{\pgfqpoint{5.884987in}{1.760506in}}%
\pgfpathlineto{\pgfqpoint{5.893096in}{1.754380in}}%
\pgfusepath{stroke}%
\end{pgfscope}%
\begin{pgfscope}%
\pgfpathrectangle{\pgfqpoint{1.229207in}{0.915754in}}{\pgfqpoint{4.650000in}{3.020000in}}%
\pgfusepath{clip}%
\pgfsetroundcap%
\pgfsetroundjoin%
\pgfsetlinewidth{1.505625pt}%
\definecolor{currentstroke}{rgb}{0.866667,0.517647,0.321569}%
\pgfsetstrokecolor{currentstroke}%
\pgfsetdash{}{0pt}%
\pgfpathmoveto{\pgfqpoint{1.215319in}{2.209119in}}%
\pgfpathlineto{\pgfqpoint{1.280202in}{2.004057in}}%
\pgfpathlineto{\pgfqpoint{1.458894in}{1.765821in}}%
\pgfpathlineto{\pgfqpoint{1.627647in}{2.201040in}}%
\pgfpathlineto{\pgfqpoint{1.804594in}{2.206496in}}%
\pgfpathlineto{\pgfqpoint{1.979562in}{2.101844in}}%
\pgfpathlineto{\pgfqpoint{2.152541in}{2.268898in}}%
\pgfpathlineto{\pgfqpoint{2.336959in}{2.146606in}}%
\pgfpathlineto{\pgfqpoint{2.511175in}{2.124340in}}%
\pgfpathlineto{\pgfqpoint{2.685677in}{1.988497in}}%
\pgfpathlineto{\pgfqpoint{2.864498in}{2.184607in}}%
\pgfpathlineto{\pgfqpoint{3.039720in}{2.095728in}}%
\pgfpathlineto{\pgfqpoint{3.215436in}{2.054539in}}%
\pgfpathlineto{\pgfqpoint{3.390580in}{2.091330in}}%
\pgfpathlineto{\pgfqpoint{3.574503in}{1.883253in}}%
\pgfpathlineto{\pgfqpoint{3.747011in}{2.117556in}}%
\pgfpathlineto{\pgfqpoint{3.924376in}{1.893709in}}%
\pgfpathlineto{\pgfqpoint{4.097909in}{2.047286in}}%
\pgfpathlineto{\pgfqpoint{4.277942in}{2.059381in}}%
\pgfpathlineto{\pgfqpoint{4.457364in}{2.108904in}}%
\pgfpathlineto{\pgfqpoint{4.629898in}{1.939274in}}%
\pgfpathlineto{\pgfqpoint{4.810888in}{2.054183in}}%
\pgfpathlineto{\pgfqpoint{4.990903in}{2.171603in}}%
\pgfpathlineto{\pgfqpoint{5.173143in}{2.181683in}}%
\pgfpathlineto{\pgfqpoint{5.348261in}{1.972648in}}%
\pgfpathlineto{\pgfqpoint{5.528637in}{2.120135in}}%
\pgfpathlineto{\pgfqpoint{5.705495in}{2.028289in}}%
\pgfpathlineto{\pgfqpoint{5.884432in}{2.036342in}}%
\pgfpathlineto{\pgfqpoint{5.893096in}{2.033761in}}%
\pgfusepath{stroke}%
\end{pgfscope}%
\begin{pgfscope}%
\pgfpathrectangle{\pgfqpoint{1.229207in}{0.915754in}}{\pgfqpoint{4.650000in}{3.020000in}}%
\pgfusepath{clip}%
\pgfsetroundcap%
\pgfsetroundjoin%
\pgfsetlinewidth{1.505625pt}%
\definecolor{currentstroke}{rgb}{0.333333,0.658824,0.407843}%
\pgfsetstrokecolor{currentstroke}%
\pgfsetdash{}{0pt}%
\pgfpathmoveto{\pgfqpoint{1.215319in}{3.326790in}}%
\pgfpathlineto{\pgfqpoint{1.342404in}{3.232296in}}%
\pgfpathlineto{\pgfqpoint{1.520378in}{2.757713in}}%
\pgfpathlineto{\pgfqpoint{1.712457in}{2.588341in}}%
\pgfpathlineto{\pgfqpoint{1.896077in}{2.460274in}}%
\pgfpathlineto{\pgfqpoint{2.059563in}{2.346326in}}%
\pgfpathlineto{\pgfqpoint{2.253724in}{2.241835in}}%
\pgfpathlineto{\pgfqpoint{2.435330in}{2.185484in}}%
\pgfpathlineto{\pgfqpoint{2.605206in}{2.284317in}}%
\pgfpathlineto{\pgfqpoint{2.802842in}{2.265463in}}%
\pgfpathlineto{\pgfqpoint{2.982234in}{2.221356in}}%
\pgfpathlineto{\pgfqpoint{3.141515in}{2.195038in}}%
\pgfpathlineto{\pgfqpoint{3.337853in}{2.305657in}}%
\pgfpathlineto{\pgfqpoint{3.490970in}{1.990369in}}%
\pgfpathlineto{\pgfqpoint{3.711765in}{2.188310in}}%
\pgfpathlineto{\pgfqpoint{3.909236in}{2.114326in}}%
\pgfpathlineto{\pgfqpoint{4.062994in}{2.081739in}}%
\pgfpathlineto{\pgfqpoint{4.258508in}{2.145019in}}%
\pgfpathlineto{\pgfqpoint{4.438295in}{2.137991in}}%
\pgfpathlineto{\pgfqpoint{4.618481in}{2.212637in}}%
\pgfpathlineto{\pgfqpoint{4.780693in}{2.022481in}}%
\pgfpathlineto{\pgfqpoint{4.985800in}{2.101176in}}%
\pgfpathlineto{\pgfqpoint{5.151776in}{2.063554in}}%
\pgfpathlineto{\pgfqpoint{5.364386in}{2.037837in}}%
\pgfpathlineto{\pgfqpoint{5.500672in}{1.913490in}}%
\pgfpathlineto{\pgfqpoint{5.675271in}{1.926483in}}%
\pgfpathlineto{\pgfqpoint{5.867141in}{1.916677in}}%
\pgfpathlineto{\pgfqpoint{5.893096in}{1.916530in}}%
\pgfusepath{stroke}%
\end{pgfscope}%
\begin{pgfscope}%
\pgfpathrectangle{\pgfqpoint{1.229207in}{0.915754in}}{\pgfqpoint{4.650000in}{3.020000in}}%
\pgfusepath{clip}%
\pgfsetroundcap%
\pgfsetroundjoin%
\pgfsetlinewidth{1.505625pt}%
\definecolor{currentstroke}{rgb}{0.768627,0.305882,0.321569}%
\pgfsetstrokecolor{currentstroke}%
\pgfsetdash{}{0pt}%
\pgfpathmoveto{\pgfqpoint{1.215319in}{3.247485in}}%
\pgfpathlineto{\pgfqpoint{1.234934in}{3.243000in}}%
\pgfpathlineto{\pgfqpoint{1.406732in}{3.163308in}}%
\pgfpathlineto{\pgfqpoint{1.578530in}{3.144804in}}%
\pgfpathlineto{\pgfqpoint{1.750328in}{3.081353in}}%
\pgfpathlineto{\pgfqpoint{1.922126in}{2.964699in}}%
\pgfpathlineto{\pgfqpoint{2.093924in}{2.931305in}}%
\pgfpathlineto{\pgfqpoint{2.265722in}{2.833932in}}%
\pgfpathlineto{\pgfqpoint{2.437520in}{2.880178in}}%
\pgfpathlineto{\pgfqpoint{2.609318in}{2.960037in}}%
\pgfpathlineto{\pgfqpoint{2.781116in}{2.868165in}}%
\pgfpathlineto{\pgfqpoint{2.952914in}{2.953276in}}%
\pgfpathlineto{\pgfqpoint{3.124712in}{2.807760in}}%
\pgfpathlineto{\pgfqpoint{3.296510in}{2.667868in}}%
\pgfpathlineto{\pgfqpoint{3.468308in}{2.908685in}}%
\pgfpathlineto{\pgfqpoint{3.640107in}{2.762823in}}%
\pgfpathlineto{\pgfqpoint{3.811905in}{2.752937in}}%
\pgfpathlineto{\pgfqpoint{3.983703in}{2.793376in}}%
\pgfpathlineto{\pgfqpoint{4.155501in}{2.824560in}}%
\pgfpathlineto{\pgfqpoint{4.327299in}{2.628994in}}%
\pgfpathlineto{\pgfqpoint{4.499097in}{2.697642in}}%
\pgfpathlineto{\pgfqpoint{4.670895in}{2.663162in}}%
\pgfpathlineto{\pgfqpoint{4.842693in}{2.733272in}}%
\pgfpathlineto{\pgfqpoint{5.014491in}{2.793298in}}%
\pgfpathlineto{\pgfqpoint{5.186289in}{2.542410in}}%
\pgfpathlineto{\pgfqpoint{5.358087in}{2.680647in}}%
\pgfpathlineto{\pgfqpoint{5.529885in}{2.495524in}}%
\pgfpathlineto{\pgfqpoint{5.701683in}{2.526875in}}%
\pgfpathlineto{\pgfqpoint{5.873481in}{2.572731in}}%
\pgfpathlineto{\pgfqpoint{5.893096in}{2.585111in}}%
\pgfusepath{stroke}%
\end{pgfscope}%
\begin{pgfscope}%
\pgfsetrectcap%
\pgfsetmiterjoin%
\pgfsetlinewidth{1.254687pt}%
\definecolor{currentstroke}{rgb}{0.800000,0.800000,0.800000}%
\pgfsetstrokecolor{currentstroke}%
\pgfsetdash{}{0pt}%
\pgfpathmoveto{\pgfqpoint{1.229207in}{0.915754in}}%
\pgfpathlineto{\pgfqpoint{1.229207in}{3.935754in}}%
\pgfusepath{stroke}%
\end{pgfscope}%
\begin{pgfscope}%
\pgfsetrectcap%
\pgfsetmiterjoin%
\pgfsetlinewidth{1.254687pt}%
\definecolor{currentstroke}{rgb}{0.800000,0.800000,0.800000}%
\pgfsetstrokecolor{currentstroke}%
\pgfsetdash{}{0pt}%
\pgfpathmoveto{\pgfqpoint{5.879208in}{0.915754in}}%
\pgfpathlineto{\pgfqpoint{5.879208in}{3.935754in}}%
\pgfusepath{stroke}%
\end{pgfscope}%
\begin{pgfscope}%
\pgfsetrectcap%
\pgfsetmiterjoin%
\pgfsetlinewidth{1.254687pt}%
\definecolor{currentstroke}{rgb}{0.800000,0.800000,0.800000}%
\pgfsetstrokecolor{currentstroke}%
\pgfsetdash{}{0pt}%
\pgfpathmoveto{\pgfqpoint{1.229207in}{0.915754in}}%
\pgfpathlineto{\pgfqpoint{5.879208in}{0.915754in}}%
\pgfusepath{stroke}%
\end{pgfscope}%
\begin{pgfscope}%
\pgfsetrectcap%
\pgfsetmiterjoin%
\pgfsetlinewidth{1.254687pt}%
\definecolor{currentstroke}{rgb}{0.800000,0.800000,0.800000}%
\pgfsetstrokecolor{currentstroke}%
\pgfsetdash{}{0pt}%
\pgfpathmoveto{\pgfqpoint{1.229207in}{3.935754in}}%
\pgfpathlineto{\pgfqpoint{5.879208in}{3.935754in}}%
\pgfusepath{stroke}%
\end{pgfscope}%
\begin{pgfscope}%
\pgfsetbuttcap%
\pgfsetmiterjoin%
\definecolor{currentfill}{rgb}{1.000000,1.000000,1.000000}%
\pgfsetfillcolor{currentfill}%
\pgfsetfillopacity{0.800000}%
\pgfsetlinewidth{1.003750pt}%
\definecolor{currentstroke}{rgb}{0.800000,0.800000,0.800000}%
\pgfsetstrokecolor{currentstroke}%
\pgfsetstrokeopacity{0.800000}%
\pgfsetdash{}{0pt}%
\pgfpathmoveto{\pgfqpoint{1.966460in}{2.840139in}}%
\pgfpathlineto{\pgfqpoint{5.654624in}{2.840139in}}%
\pgfpathquadraticcurveto{\pgfqpoint{5.718791in}{2.840139in}}{\pgfqpoint{5.718791in}{2.904306in}}%
\pgfpathlineto{\pgfqpoint{5.718791in}{3.711171in}}%
\pgfpathquadraticcurveto{\pgfqpoint{5.718791in}{3.775338in}}{\pgfqpoint{5.654624in}{3.775338in}}%
\pgfpathlineto{\pgfqpoint{1.966460in}{3.775338in}}%
\pgfpathquadraticcurveto{\pgfqpoint{1.902294in}{3.775338in}}{\pgfqpoint{1.902294in}{3.711171in}}%
\pgfpathlineto{\pgfqpoint{1.902294in}{2.904306in}}%
\pgfpathquadraticcurveto{\pgfqpoint{1.902294in}{2.840139in}}{\pgfqpoint{1.966460in}{2.840139in}}%
\pgfpathclose%
\pgfusepath{stroke,fill}%
\end{pgfscope}%
\begin{pgfscope}%
\pgfsetroundcap%
\pgfsetroundjoin%
\pgfsetlinewidth{5.018750pt}%
\definecolor{currentstroke}{rgb}{0.298039,0.447059,0.690196}%
\pgfsetstrokecolor{currentstroke}%
\pgfsetdash{}{0pt}%
\pgfpathmoveto{\pgfqpoint{2.030627in}{3.519327in}}%
\pgfpathlineto{\pgfqpoint{2.672294in}{3.519327in}}%
\pgfusepath{stroke}%
\end{pgfscope}%
\begin{pgfscope}%
\definecolor{textcolor}{rgb}{0.150000,0.150000,0.150000}%
\pgfsetstrokecolor{textcolor}%
\pgfsetfillcolor{textcolor}%
\pgftext[x=2.928960in,y=3.407035in,left,base]{\color{textcolor}\rmfamily\fontsize{23.100000}{27.720000}\selectfont W-L}%
\end{pgfscope}%
\begin{pgfscope}%
\pgfsetroundcap%
\pgfsetroundjoin%
\pgfsetlinewidth{5.018750pt}%
\definecolor{currentstroke}{rgb}{0.866667,0.517647,0.321569}%
\pgfsetstrokecolor{currentstroke}%
\pgfsetdash{}{0pt}%
\pgfpathmoveto{\pgfqpoint{2.030627in}{3.147977in}}%
\pgfpathlineto{\pgfqpoint{2.672294in}{3.147977in}}%
\pgfusepath{stroke}%
\end{pgfscope}%
\begin{pgfscope}%
\definecolor{textcolor}{rgb}{0.150000,0.150000,0.150000}%
\pgfsetstrokecolor{textcolor}%
\pgfsetfillcolor{textcolor}%
\pgftext[x=2.928960in,y=3.035686in,left,base]{\color{textcolor}\rmfamily\fontsize{23.100000}{27.720000}\selectfont W-M}%
\end{pgfscope}%
\begin{pgfscope}%
\pgfsetroundcap%
\pgfsetroundjoin%
\pgfsetlinewidth{5.018750pt}%
\definecolor{currentstroke}{rgb}{0.333333,0.658824,0.407843}%
\pgfsetstrokecolor{currentstroke}%
\pgfsetdash{}{0pt}%
\pgfpathmoveto{\pgfqpoint{3.977694in}{3.519327in}}%
\pgfpathlineto{\pgfqpoint{4.619361in}{3.519327in}}%
\pgfusepath{stroke}%
\end{pgfscope}%
\begin{pgfscope}%
\definecolor{textcolor}{rgb}{0.150000,0.150000,0.150000}%
\pgfsetstrokecolor{textcolor}%
\pgfsetfillcolor{textcolor}%
\pgftext[x=4.876028in,y=3.407035in,left,base]{\color{textcolor}\rmfamily\fontsize{23.100000}{27.720000}\selectfont DCV}%
\end{pgfscope}%
\begin{pgfscope}%
\pgfsetroundcap%
\pgfsetroundjoin%
\pgfsetlinewidth{5.018750pt}%
\definecolor{currentstroke}{rgb}{0.768627,0.305882,0.321569}%
\pgfsetstrokecolor{currentstroke}%
\pgfsetdash{}{0pt}%
\pgfpathmoveto{\pgfqpoint{3.977694in}{3.147977in}}%
\pgfpathlineto{\pgfqpoint{4.619361in}{3.147977in}}%
\pgfusepath{stroke}%
\end{pgfscope}%
\begin{pgfscope}%
\definecolor{textcolor}{rgb}{0.150000,0.150000,0.150000}%
\pgfsetstrokecolor{textcolor}%
\pgfsetfillcolor{textcolor}%
\pgftext[x=4.876028in,y=3.035686in,left,base]{\color{textcolor}\rmfamily\fontsize{23.100000}{27.720000}\selectfont MC}%
\end{pgfscope}%
\end{pgfpicture}%
\makeatother%
\endgroup%

%% file: fig/continuous_circle-1000-5d_res.pgf
%% Creator: Matplotlib, PGF backend
%%
%% To include the figure in your LaTeX document, write
%%   \input{<filename>.pgf}
%%
%% Make sure the required packages are loaded in your preamble
%%   \usepackage{pgf}
%%
%% Figures using additional raster images can only be included by \input if
%% they are in the same directory as the main LaTeX file. For loading figures
%% from other directories you can use the `import` package
%% and then include the figures with
%%   \import{<path to file>}{<filename>.pgf}
%%
%% Matplotlib used the following preamble
%%
\begingroup%
\makeatletter%
\begin{pgfpicture}%
\pgfpathrectangle{\pgfpointorigin}{\pgfqpoint{4.850000in}{4.035754in}}%
\pgfusepath{use as bounding box, clip}%
\begin{pgfscope}%
\pgfsetbuttcap%
\pgfsetmiterjoin%
\definecolor{currentfill}{rgb}{1.000000,1.000000,1.000000}%
\pgfsetfillcolor{currentfill}%
\pgfsetlinewidth{0.000000pt}%
\definecolor{currentstroke}{rgb}{1.000000,1.000000,1.000000}%
\pgfsetstrokecolor{currentstroke}%
\pgfsetdash{}{0pt}%
\pgfpathmoveto{\pgfqpoint{0.000000in}{0.000000in}}%
\pgfpathlineto{\pgfqpoint{4.850000in}{0.000000in}}%
\pgfpathlineto{\pgfqpoint{4.850000in}{4.035754in}}%
\pgfpathlineto{\pgfqpoint{0.000000in}{4.035754in}}%
\pgfpathclose%
\pgfusepath{fill}%
\end{pgfscope}%
\begin{pgfscope}%
\pgfsetbuttcap%
\pgfsetmiterjoin%
\definecolor{currentfill}{rgb}{1.000000,1.000000,1.000000}%
\pgfsetfillcolor{currentfill}%
\pgfsetlinewidth{0.000000pt}%
\definecolor{currentstroke}{rgb}{0.000000,0.000000,0.000000}%
\pgfsetstrokecolor{currentstroke}%
\pgfsetstrokeopacity{0.000000}%
\pgfsetdash{}{0pt}%
\pgfpathmoveto{\pgfqpoint{0.100000in}{0.915754in}}%
\pgfpathlineto{\pgfqpoint{4.750000in}{0.915754in}}%
\pgfpathlineto{\pgfqpoint{4.750000in}{3.935754in}}%
\pgfpathlineto{\pgfqpoint{0.100000in}{3.935754in}}%
\pgfpathclose%
\pgfusepath{fill}%
\end{pgfscope}%
\begin{pgfscope}%
\pgfpathrectangle{\pgfqpoint{0.100000in}{0.915754in}}{\pgfqpoint{4.650000in}{3.020000in}}%
\pgfusepath{clip}%
\pgfsetroundcap%
\pgfsetroundjoin%
\pgfsetlinewidth{1.003750pt}%
\definecolor{currentstroke}{rgb}{0.800000,0.800000,0.800000}%
\pgfsetstrokecolor{currentstroke}%
\pgfsetdash{}{0pt}%
\pgfpathmoveto{\pgfqpoint{0.349107in}{0.915754in}}%
\pgfpathlineto{\pgfqpoint{0.349107in}{3.935754in}}%
\pgfusepath{stroke}%
\end{pgfscope}%
\begin{pgfscope}%
\definecolor{textcolor}{rgb}{0.150000,0.150000,0.150000}%
\pgfsetstrokecolor{textcolor}%
\pgfsetfillcolor{textcolor}%
\pgftext[x=0.349107in,y=0.783810in,,top]{\color{textcolor}\rmfamily\fontsize{23.100000}{27.720000}\selectfont 10}%
\end{pgfscope}%
\begin{pgfscope}%
\pgfpathrectangle{\pgfqpoint{0.100000in}{0.915754in}}{\pgfqpoint{4.650000in}{3.020000in}}%
\pgfusepath{clip}%
\pgfsetroundcap%
\pgfsetroundjoin%
\pgfsetlinewidth{1.003750pt}%
\definecolor{currentstroke}{rgb}{0.800000,0.800000,0.800000}%
\pgfsetstrokecolor{currentstroke}%
\pgfsetdash{}{0pt}%
\pgfpathmoveto{\pgfqpoint{1.179464in}{0.915754in}}%
\pgfpathlineto{\pgfqpoint{1.179464in}{3.935754in}}%
\pgfusepath{stroke}%
\end{pgfscope}%
\begin{pgfscope}%
\definecolor{textcolor}{rgb}{0.150000,0.150000,0.150000}%
\pgfsetstrokecolor{textcolor}%
\pgfsetfillcolor{textcolor}%
\pgftext[x=1.179464in,y=0.783810in,,top]{\color{textcolor}\rmfamily\fontsize{23.100000}{27.720000}\selectfont 20}%
\end{pgfscope}%
\begin{pgfscope}%
\pgfpathrectangle{\pgfqpoint{0.100000in}{0.915754in}}{\pgfqpoint{4.650000in}{3.020000in}}%
\pgfusepath{clip}%
\pgfsetroundcap%
\pgfsetroundjoin%
\pgfsetlinewidth{1.003750pt}%
\definecolor{currentstroke}{rgb}{0.800000,0.800000,0.800000}%
\pgfsetstrokecolor{currentstroke}%
\pgfsetdash{}{0pt}%
\pgfpathmoveto{\pgfqpoint{2.009821in}{0.915754in}}%
\pgfpathlineto{\pgfqpoint{2.009821in}{3.935754in}}%
\pgfusepath{stroke}%
\end{pgfscope}%
\begin{pgfscope}%
\definecolor{textcolor}{rgb}{0.150000,0.150000,0.150000}%
\pgfsetstrokecolor{textcolor}%
\pgfsetfillcolor{textcolor}%
\pgftext[x=2.009821in,y=0.783810in,,top]{\color{textcolor}\rmfamily\fontsize{23.100000}{27.720000}\selectfont 30}%
\end{pgfscope}%
\begin{pgfscope}%
\pgfpathrectangle{\pgfqpoint{0.100000in}{0.915754in}}{\pgfqpoint{4.650000in}{3.020000in}}%
\pgfusepath{clip}%
\pgfsetroundcap%
\pgfsetroundjoin%
\pgfsetlinewidth{1.003750pt}%
\definecolor{currentstroke}{rgb}{0.800000,0.800000,0.800000}%
\pgfsetstrokecolor{currentstroke}%
\pgfsetdash{}{0pt}%
\pgfpathmoveto{\pgfqpoint{2.840179in}{0.915754in}}%
\pgfpathlineto{\pgfqpoint{2.840179in}{3.935754in}}%
\pgfusepath{stroke}%
\end{pgfscope}%
\begin{pgfscope}%
\definecolor{textcolor}{rgb}{0.150000,0.150000,0.150000}%
\pgfsetstrokecolor{textcolor}%
\pgfsetfillcolor{textcolor}%
\pgftext[x=2.840179in,y=0.783810in,,top]{\color{textcolor}\rmfamily\fontsize{23.100000}{27.720000}\selectfont 40}%
\end{pgfscope}%
\begin{pgfscope}%
\pgfpathrectangle{\pgfqpoint{0.100000in}{0.915754in}}{\pgfqpoint{4.650000in}{3.020000in}}%
\pgfusepath{clip}%
\pgfsetroundcap%
\pgfsetroundjoin%
\pgfsetlinewidth{1.003750pt}%
\definecolor{currentstroke}{rgb}{0.800000,0.800000,0.800000}%
\pgfsetstrokecolor{currentstroke}%
\pgfsetdash{}{0pt}%
\pgfpathmoveto{\pgfqpoint{3.670536in}{0.915754in}}%
\pgfpathlineto{\pgfqpoint{3.670536in}{3.935754in}}%
\pgfusepath{stroke}%
\end{pgfscope}%
\begin{pgfscope}%
\definecolor{textcolor}{rgb}{0.150000,0.150000,0.150000}%
\pgfsetstrokecolor{textcolor}%
\pgfsetfillcolor{textcolor}%
\pgftext[x=3.670536in,y=0.783810in,,top]{\color{textcolor}\rmfamily\fontsize{23.100000}{27.720000}\selectfont 50}%
\end{pgfscope}%
\begin{pgfscope}%
\pgfpathrectangle{\pgfqpoint{0.100000in}{0.915754in}}{\pgfqpoint{4.650000in}{3.020000in}}%
\pgfusepath{clip}%
\pgfsetroundcap%
\pgfsetroundjoin%
\pgfsetlinewidth{1.003750pt}%
\definecolor{currentstroke}{rgb}{0.800000,0.800000,0.800000}%
\pgfsetstrokecolor{currentstroke}%
\pgfsetdash{}{0pt}%
\pgfpathmoveto{\pgfqpoint{4.500893in}{0.915754in}}%
\pgfpathlineto{\pgfqpoint{4.500893in}{3.935754in}}%
\pgfusepath{stroke}%
\end{pgfscope}%
\begin{pgfscope}%
\definecolor{textcolor}{rgb}{0.150000,0.150000,0.150000}%
\pgfsetstrokecolor{textcolor}%
\pgfsetfillcolor{textcolor}%
\pgftext[x=4.500893in,y=0.783810in,,top]{\color{textcolor}\rmfamily\fontsize{23.100000}{27.720000}\selectfont 60}%
\end{pgfscope}%
\begin{pgfscope}%
\definecolor{textcolor}{rgb}{0.150000,0.150000,0.150000}%
\pgfsetstrokecolor{textcolor}%
\pgfsetfillcolor{textcolor}%
\pgftext[x=2.425000in,y=0.407183in,,top]{\color{textcolor}\rmfamily\fontsize{25.200000}{30.240000}\selectfont runtime in seconds}%
\end{pgfscope}%
\begin{pgfscope}%
\pgfpathrectangle{\pgfqpoint{0.100000in}{0.915754in}}{\pgfqpoint{4.650000in}{3.020000in}}%
\pgfusepath{clip}%
\pgfsetroundcap%
\pgfsetroundjoin%
\pgfsetlinewidth{1.003750pt}%
\definecolor{currentstroke}{rgb}{0.800000,0.800000,0.800000}%
\pgfsetstrokecolor{currentstroke}%
\pgfsetdash{}{0pt}%
\pgfpathmoveto{\pgfqpoint{0.100000in}{1.328214in}}%
\pgfpathlineto{\pgfqpoint{4.750000in}{1.328214in}}%
\pgfusepath{stroke}%
\end{pgfscope}%
\begin{pgfscope}%
\pgfpathrectangle{\pgfqpoint{0.100000in}{0.915754in}}{\pgfqpoint{4.650000in}{3.020000in}}%
\pgfusepath{clip}%
\pgfsetroundcap%
\pgfsetroundjoin%
\pgfsetlinewidth{1.003750pt}%
\definecolor{currentstroke}{rgb}{0.800000,0.800000,0.800000}%
\pgfsetstrokecolor{currentstroke}%
\pgfsetdash{}{0pt}%
\pgfpathmoveto{\pgfqpoint{0.100000in}{2.698375in}}%
\pgfpathlineto{\pgfqpoint{4.750000in}{2.698375in}}%
\pgfusepath{stroke}%
\end{pgfscope}%
\begin{pgfscope}%
\pgfpathrectangle{\pgfqpoint{0.100000in}{0.915754in}}{\pgfqpoint{4.650000in}{3.020000in}}%
\pgfusepath{clip}%
\pgfsetbuttcap%
\pgfsetroundjoin%
\definecolor{currentfill}{rgb}{0.298039,0.447059,0.690196}%
\pgfsetfillcolor{currentfill}%
\pgfsetfillopacity{0.200000}%
\pgfsetlinewidth{1.003750pt}%
\definecolor{currentstroke}{rgb}{0.298039,0.447059,0.690196}%
\pgfsetstrokecolor{currentstroke}%
\pgfsetstrokeopacity{0.200000}%
\pgfsetdash{}{0pt}%
\pgfpathmoveto{\pgfqpoint{-0.024487in}{2.007009in}}%
\pgfpathlineto{\pgfqpoint{-0.024487in}{1.757429in}}%
\pgfpathlineto{\pgfqpoint{0.153801in}{1.845293in}}%
\pgfpathlineto{\pgfqpoint{0.326333in}{2.075465in}}%
\pgfpathlineto{\pgfqpoint{0.502402in}{2.167266in}}%
\pgfpathlineto{\pgfqpoint{0.676104in}{2.209370in}}%
\pgfpathlineto{\pgfqpoint{0.847730in}{2.267045in}}%
\pgfpathlineto{\pgfqpoint{1.026665in}{1.989678in}}%
\pgfpathlineto{\pgfqpoint{1.197882in}{2.085058in}}%
\pgfpathlineto{\pgfqpoint{1.373116in}{1.906621in}}%
\pgfpathlineto{\pgfqpoint{1.544081in}{2.067309in}}%
\pgfpathlineto{\pgfqpoint{1.720038in}{2.073791in}}%
\pgfpathlineto{\pgfqpoint{1.898061in}{2.010780in}}%
\pgfpathlineto{\pgfqpoint{2.068535in}{1.986843in}}%
\pgfpathlineto{\pgfqpoint{2.250102in}{2.053717in}}%
\pgfpathlineto{\pgfqpoint{2.419418in}{1.891750in}}%
\pgfpathlineto{\pgfqpoint{2.597532in}{2.054949in}}%
\pgfpathlineto{\pgfqpoint{2.771691in}{1.989388in}}%
\pgfpathlineto{\pgfqpoint{2.942440in}{2.050856in}}%
\pgfpathlineto{\pgfqpoint{3.118824in}{2.058549in}}%
\pgfpathlineto{\pgfqpoint{3.289559in}{2.040944in}}%
\pgfpathlineto{\pgfqpoint{3.466392in}{1.896572in}}%
\pgfpathlineto{\pgfqpoint{3.642509in}{2.020126in}}%
\pgfpathlineto{\pgfqpoint{3.823342in}{1.981886in}}%
\pgfpathlineto{\pgfqpoint{3.996368in}{2.027984in}}%
\pgfpathlineto{\pgfqpoint{4.170776in}{1.984018in}}%
\pgfpathlineto{\pgfqpoint{4.343945in}{1.998764in}}%
\pgfpathlineto{\pgfqpoint{4.519249in}{1.892350in}}%
\pgfpathlineto{\pgfqpoint{4.697520in}{2.083726in}}%
\pgfpathlineto{\pgfqpoint{4.871872in}{1.993549in}}%
\pgfpathlineto{\pgfqpoint{5.048739in}{2.005883in}}%
\pgfpathlineto{\pgfqpoint{5.048739in}{2.162096in}}%
\pgfpathlineto{\pgfqpoint{5.048739in}{2.162096in}}%
\pgfpathlineto{\pgfqpoint{4.871872in}{2.131127in}}%
\pgfpathlineto{\pgfqpoint{4.697520in}{2.215591in}}%
\pgfpathlineto{\pgfqpoint{4.519249in}{2.075041in}}%
\pgfpathlineto{\pgfqpoint{4.343945in}{2.116295in}}%
\pgfpathlineto{\pgfqpoint{4.170776in}{2.139489in}}%
\pgfpathlineto{\pgfqpoint{3.996368in}{2.139912in}}%
\pgfpathlineto{\pgfqpoint{3.823342in}{2.173696in}}%
\pgfpathlineto{\pgfqpoint{3.642509in}{2.169682in}}%
\pgfpathlineto{\pgfqpoint{3.466392in}{2.100680in}}%
\pgfpathlineto{\pgfqpoint{3.289559in}{2.181205in}}%
\pgfpathlineto{\pgfqpoint{3.118824in}{2.213784in}}%
\pgfpathlineto{\pgfqpoint{2.942440in}{2.199374in}}%
\pgfpathlineto{\pgfqpoint{2.771691in}{2.169360in}}%
\pgfpathlineto{\pgfqpoint{2.597532in}{2.201084in}}%
\pgfpathlineto{\pgfqpoint{2.419418in}{2.055298in}}%
\pgfpathlineto{\pgfqpoint{2.250102in}{2.208470in}}%
\pgfpathlineto{\pgfqpoint{2.068535in}{2.141165in}}%
\pgfpathlineto{\pgfqpoint{1.898061in}{2.167519in}}%
\pgfpathlineto{\pgfqpoint{1.720038in}{2.252662in}}%
\pgfpathlineto{\pgfqpoint{1.544081in}{2.252994in}}%
\pgfpathlineto{\pgfqpoint{1.373116in}{2.163979in}}%
\pgfpathlineto{\pgfqpoint{1.197882in}{2.270113in}}%
\pgfpathlineto{\pgfqpoint{1.026665in}{2.206826in}}%
\pgfpathlineto{\pgfqpoint{0.847730in}{2.377281in}}%
\pgfpathlineto{\pgfqpoint{0.676104in}{2.354812in}}%
\pgfpathlineto{\pgfqpoint{0.502402in}{2.321791in}}%
\pgfpathlineto{\pgfqpoint{0.326333in}{2.259191in}}%
\pgfpathlineto{\pgfqpoint{0.153801in}{2.136052in}}%
\pgfpathlineto{\pgfqpoint{-0.024487in}{2.007009in}}%
\pgfpathclose%
\pgfusepath{stroke,fill}%
\end{pgfscope}%
\begin{pgfscope}%
\pgfpathrectangle{\pgfqpoint{0.100000in}{0.915754in}}{\pgfqpoint{4.650000in}{3.020000in}}%
\pgfusepath{clip}%
\pgfsetbuttcap%
\pgfsetroundjoin%
\definecolor{currentfill}{rgb}{0.866667,0.517647,0.321569}%
\pgfsetfillcolor{currentfill}%
\pgfsetfillopacity{0.200000}%
\pgfsetlinewidth{1.003750pt}%
\definecolor{currentstroke}{rgb}{0.866667,0.517647,0.321569}%
\pgfsetstrokecolor{currentstroke}%
\pgfsetstrokeopacity{0.200000}%
\pgfsetdash{}{0pt}%
\pgfpathmoveto{\pgfqpoint{-0.020824in}{2.162119in}}%
\pgfpathlineto{\pgfqpoint{-0.020824in}{1.940689in}}%
\pgfpathlineto{\pgfqpoint{0.152413in}{1.371042in}}%
\pgfpathlineto{\pgfqpoint{0.330375in}{1.727704in}}%
\pgfpathlineto{\pgfqpoint{0.499912in}{1.662874in}}%
\pgfpathlineto{\pgfqpoint{0.676644in}{1.743416in}}%
\pgfpathlineto{\pgfqpoint{0.849328in}{1.773398in}}%
\pgfpathlineto{\pgfqpoint{1.026253in}{1.559110in}}%
\pgfpathlineto{\pgfqpoint{1.198311in}{1.398947in}}%
\pgfpathlineto{\pgfqpoint{1.368900in}{1.431885in}}%
\pgfpathlineto{\pgfqpoint{1.545534in}{1.387245in}}%
\pgfpathlineto{\pgfqpoint{1.722310in}{1.415780in}}%
\pgfpathlineto{\pgfqpoint{1.895741in}{1.396626in}}%
\pgfpathlineto{\pgfqpoint{2.070309in}{1.309820in}}%
\pgfpathlineto{\pgfqpoint{2.249437in}{1.537920in}}%
\pgfpathlineto{\pgfqpoint{2.420811in}{1.423596in}}%
\pgfpathlineto{\pgfqpoint{2.592900in}{1.450756in}}%
\pgfpathlineto{\pgfqpoint{2.771634in}{1.505502in}}%
\pgfpathlineto{\pgfqpoint{2.948049in}{1.468509in}}%
\pgfpathlineto{\pgfqpoint{3.122146in}{1.389730in}}%
\pgfpathlineto{\pgfqpoint{3.293776in}{1.295359in}}%
\pgfpathlineto{\pgfqpoint{3.473490in}{1.395630in}}%
\pgfpathlineto{\pgfqpoint{3.647130in}{1.419708in}}%
\pgfpathlineto{\pgfqpoint{3.818284in}{1.512590in}}%
\pgfpathlineto{\pgfqpoint{3.993824in}{1.287781in}}%
\pgfpathlineto{\pgfqpoint{4.170370in}{1.571208in}}%
\pgfpathlineto{\pgfqpoint{4.350245in}{1.410906in}}%
\pgfpathlineto{\pgfqpoint{4.519724in}{1.401279in}}%
\pgfpathlineto{\pgfqpoint{4.698489in}{1.475715in}}%
\pgfpathlineto{\pgfqpoint{4.870727in}{1.632662in}}%
\pgfpathlineto{\pgfqpoint{5.047994in}{1.631352in}}%
\pgfpathlineto{\pgfqpoint{5.047994in}{1.848130in}}%
\pgfpathlineto{\pgfqpoint{5.047994in}{1.848130in}}%
\pgfpathlineto{\pgfqpoint{4.870727in}{1.908984in}}%
\pgfpathlineto{\pgfqpoint{4.698489in}{1.778517in}}%
\pgfpathlineto{\pgfqpoint{4.519724in}{1.674990in}}%
\pgfpathlineto{\pgfqpoint{4.350245in}{1.747676in}}%
\pgfpathlineto{\pgfqpoint{4.170370in}{1.842305in}}%
\pgfpathlineto{\pgfqpoint{3.993824in}{1.658137in}}%
\pgfpathlineto{\pgfqpoint{3.818284in}{1.738055in}}%
\pgfpathlineto{\pgfqpoint{3.647130in}{1.708734in}}%
\pgfpathlineto{\pgfqpoint{3.473490in}{1.683870in}}%
\pgfpathlineto{\pgfqpoint{3.293776in}{1.653820in}}%
\pgfpathlineto{\pgfqpoint{3.122146in}{1.665311in}}%
\pgfpathlineto{\pgfqpoint{2.948049in}{1.767166in}}%
\pgfpathlineto{\pgfqpoint{2.771634in}{1.792276in}}%
\pgfpathlineto{\pgfqpoint{2.592900in}{1.752420in}}%
\pgfpathlineto{\pgfqpoint{2.420811in}{1.782769in}}%
\pgfpathlineto{\pgfqpoint{2.249437in}{1.800327in}}%
\pgfpathlineto{\pgfqpoint{2.070309in}{1.682348in}}%
\pgfpathlineto{\pgfqpoint{1.895741in}{1.732839in}}%
\pgfpathlineto{\pgfqpoint{1.722310in}{1.727355in}}%
\pgfpathlineto{\pgfqpoint{1.545534in}{1.748402in}}%
\pgfpathlineto{\pgfqpoint{1.368900in}{1.734227in}}%
\pgfpathlineto{\pgfqpoint{1.198311in}{1.712985in}}%
\pgfpathlineto{\pgfqpoint{1.026253in}{1.804623in}}%
\pgfpathlineto{\pgfqpoint{0.849328in}{1.984356in}}%
\pgfpathlineto{\pgfqpoint{0.676644in}{1.995041in}}%
\pgfpathlineto{\pgfqpoint{0.499912in}{1.901012in}}%
\pgfpathlineto{\pgfqpoint{0.330375in}{1.917092in}}%
\pgfpathlineto{\pgfqpoint{0.152413in}{1.753745in}}%
\pgfpathlineto{\pgfqpoint{-0.020824in}{2.162119in}}%
\pgfpathclose%
\pgfusepath{stroke,fill}%
\end{pgfscope}%
\begin{pgfscope}%
\pgfpathrectangle{\pgfqpoint{0.100000in}{0.915754in}}{\pgfqpoint{4.650000in}{3.020000in}}%
\pgfusepath{clip}%
\pgfsetbuttcap%
\pgfsetroundjoin%
\definecolor{currentfill}{rgb}{0.333333,0.658824,0.407843}%
\pgfsetfillcolor{currentfill}%
\pgfsetfillopacity{0.200000}%
\pgfsetlinewidth{1.003750pt}%
\definecolor{currentstroke}{rgb}{0.333333,0.658824,0.407843}%
\pgfsetstrokecolor{currentstroke}%
\pgfsetstrokeopacity{0.200000}%
\pgfsetdash{}{0pt}%
\pgfpathmoveto{\pgfqpoint{0.012317in}{3.148138in}}%
\pgfpathlineto{\pgfqpoint{0.012317in}{2.682960in}}%
\pgfpathlineto{\pgfqpoint{0.194795in}{2.704588in}}%
\pgfpathlineto{\pgfqpoint{0.376496in}{2.385280in}}%
\pgfpathlineto{\pgfqpoint{0.566349in}{2.406082in}}%
\pgfpathlineto{\pgfqpoint{0.742415in}{2.320591in}}%
\pgfpathlineto{\pgfqpoint{0.926648in}{2.222282in}}%
\pgfpathlineto{\pgfqpoint{1.104568in}{2.203319in}}%
\pgfpathlineto{\pgfqpoint{1.291435in}{2.098533in}}%
\pgfpathlineto{\pgfqpoint{1.499325in}{2.096216in}}%
\pgfpathlineto{\pgfqpoint{1.662708in}{2.017501in}}%
\pgfpathlineto{\pgfqpoint{1.853700in}{2.071458in}}%
\pgfpathlineto{\pgfqpoint{2.051143in}{2.058797in}}%
\pgfpathlineto{\pgfqpoint{2.232877in}{1.915287in}}%
\pgfpathlineto{\pgfqpoint{2.410870in}{1.962256in}}%
\pgfpathlineto{\pgfqpoint{2.611535in}{1.914758in}}%
\pgfpathlineto{\pgfqpoint{2.802309in}{2.031242in}}%
\pgfpathlineto{\pgfqpoint{2.976814in}{1.896204in}}%
\pgfpathlineto{\pgfqpoint{3.162703in}{1.872231in}}%
\pgfpathlineto{\pgfqpoint{3.364005in}{2.041948in}}%
\pgfpathlineto{\pgfqpoint{3.522904in}{1.930839in}}%
\pgfpathlineto{\pgfqpoint{3.717669in}{1.876319in}}%
\pgfpathlineto{\pgfqpoint{3.906390in}{2.046444in}}%
\pgfpathlineto{\pgfqpoint{4.100743in}{1.966677in}}%
\pgfpathlineto{\pgfqpoint{4.261438in}{1.830475in}}%
\pgfpathlineto{\pgfqpoint{4.456709in}{1.814069in}}%
\pgfpathlineto{\pgfqpoint{4.645937in}{1.960716in}}%
\pgfpathlineto{\pgfqpoint{4.824816in}{1.723272in}}%
\pgfpathlineto{\pgfqpoint{5.015718in}{1.747172in}}%
\pgfpathlineto{\pgfqpoint{5.200794in}{1.712843in}}%
\pgfpathlineto{\pgfqpoint{5.403632in}{1.798981in}}%
\pgfpathlineto{\pgfqpoint{5.403632in}{2.140920in}}%
\pgfpathlineto{\pgfqpoint{5.403632in}{2.140920in}}%
\pgfpathlineto{\pgfqpoint{5.200794in}{2.229823in}}%
\pgfpathlineto{\pgfqpoint{5.015718in}{2.113187in}}%
\pgfpathlineto{\pgfqpoint{4.824816in}{2.071037in}}%
\pgfpathlineto{\pgfqpoint{4.645937in}{2.211355in}}%
\pgfpathlineto{\pgfqpoint{4.456709in}{2.123935in}}%
\pgfpathlineto{\pgfqpoint{4.261438in}{2.213077in}}%
\pgfpathlineto{\pgfqpoint{4.100743in}{2.211242in}}%
\pgfpathlineto{\pgfqpoint{3.906390in}{2.382695in}}%
\pgfpathlineto{\pgfqpoint{3.717669in}{2.193854in}}%
\pgfpathlineto{\pgfqpoint{3.522904in}{2.222148in}}%
\pgfpathlineto{\pgfqpoint{3.364005in}{2.356804in}}%
\pgfpathlineto{\pgfqpoint{3.162703in}{2.232266in}}%
\pgfpathlineto{\pgfqpoint{2.976814in}{2.270094in}}%
\pgfpathlineto{\pgfqpoint{2.802309in}{2.449955in}}%
\pgfpathlineto{\pgfqpoint{2.611535in}{2.243129in}}%
\pgfpathlineto{\pgfqpoint{2.410870in}{2.260048in}}%
\pgfpathlineto{\pgfqpoint{2.232877in}{2.291214in}}%
\pgfpathlineto{\pgfqpoint{2.051143in}{2.462799in}}%
\pgfpathlineto{\pgfqpoint{1.853700in}{2.421632in}}%
\pgfpathlineto{\pgfqpoint{1.662708in}{2.699568in}}%
\pgfpathlineto{\pgfqpoint{1.499325in}{2.518735in}}%
\pgfpathlineto{\pgfqpoint{1.291435in}{2.593579in}}%
\pgfpathlineto{\pgfqpoint{1.104568in}{2.584448in}}%
\pgfpathlineto{\pgfqpoint{0.926648in}{2.624754in}}%
\pgfpathlineto{\pgfqpoint{0.742415in}{2.691750in}}%
\pgfpathlineto{\pgfqpoint{0.566349in}{2.769558in}}%
\pgfpathlineto{\pgfqpoint{0.376496in}{2.731670in}}%
\pgfpathlineto{\pgfqpoint{0.194795in}{3.068850in}}%
\pgfpathlineto{\pgfqpoint{0.012317in}{3.148138in}}%
\pgfpathclose%
\pgfusepath{stroke,fill}%
\end{pgfscope}%
\begin{pgfscope}%
\pgfpathrectangle{\pgfqpoint{0.100000in}{0.915754in}}{\pgfqpoint{4.650000in}{3.020000in}}%
\pgfusepath{clip}%
\pgfsetbuttcap%
\pgfsetroundjoin%
\definecolor{currentfill}{rgb}{0.768627,0.305882,0.321569}%
\pgfsetfillcolor{currentfill}%
\pgfsetfillopacity{0.200000}%
\pgfsetlinewidth{1.003750pt}%
\definecolor{currentstroke}{rgb}{0.768627,0.305882,0.321569}%
\pgfsetstrokecolor{currentstroke}%
\pgfsetstrokeopacity{0.200000}%
\pgfsetdash{}{0pt}%
\pgfpathmoveto{\pgfqpoint{-0.066071in}{3.107934in}}%
\pgfpathlineto{\pgfqpoint{-0.066071in}{2.693965in}}%
\pgfpathlineto{\pgfqpoint{0.105727in}{2.692150in}}%
\pgfpathlineto{\pgfqpoint{0.277525in}{2.776507in}}%
\pgfpathlineto{\pgfqpoint{0.449323in}{2.621148in}}%
\pgfpathlineto{\pgfqpoint{0.621121in}{2.257307in}}%
\pgfpathlineto{\pgfqpoint{0.792919in}{2.490539in}}%
\pgfpathlineto{\pgfqpoint{0.964717in}{2.626299in}}%
\pgfpathlineto{\pgfqpoint{1.136515in}{2.347380in}}%
\pgfpathlineto{\pgfqpoint{1.308313in}{2.381347in}}%
\pgfpathlineto{\pgfqpoint{1.480111in}{2.538599in}}%
\pgfpathlineto{\pgfqpoint{1.651909in}{2.499865in}}%
\pgfpathlineto{\pgfqpoint{1.823707in}{2.265362in}}%
\pgfpathlineto{\pgfqpoint{1.995505in}{2.196399in}}%
\pgfpathlineto{\pgfqpoint{2.167303in}{2.383328in}}%
\pgfpathlineto{\pgfqpoint{2.339101in}{2.199172in}}%
\pgfpathlineto{\pgfqpoint{2.510899in}{2.267880in}}%
\pgfpathlineto{\pgfqpoint{2.682697in}{2.232971in}}%
\pgfpathlineto{\pgfqpoint{2.854495in}{2.222802in}}%
\pgfpathlineto{\pgfqpoint{3.026293in}{2.230232in}}%
\pgfpathlineto{\pgfqpoint{3.198091in}{2.235149in}}%
\pgfpathlineto{\pgfqpoint{3.369889in}{2.251771in}}%
\pgfpathlineto{\pgfqpoint{3.541687in}{2.143535in}}%
\pgfpathlineto{\pgfqpoint{3.713485in}{2.157166in}}%
\pgfpathlineto{\pgfqpoint{3.885283in}{2.104518in}}%
\pgfpathlineto{\pgfqpoint{4.057081in}{2.116447in}}%
\pgfpathlineto{\pgfqpoint{4.228879in}{2.122842in}}%
\pgfpathlineto{\pgfqpoint{4.400677in}{2.064917in}}%
\pgfpathlineto{\pgfqpoint{4.572475in}{2.100335in}}%
\pgfpathlineto{\pgfqpoint{4.744273in}{2.095616in}}%
\pgfpathlineto{\pgfqpoint{4.916071in}{2.020124in}}%
\pgfpathlineto{\pgfqpoint{4.916071in}{2.356509in}}%
\pgfpathlineto{\pgfqpoint{4.916071in}{2.356509in}}%
\pgfpathlineto{\pgfqpoint{4.744273in}{2.381927in}}%
\pgfpathlineto{\pgfqpoint{4.572475in}{2.510350in}}%
\pgfpathlineto{\pgfqpoint{4.400677in}{2.439643in}}%
\pgfpathlineto{\pgfqpoint{4.228879in}{2.413167in}}%
\pgfpathlineto{\pgfqpoint{4.057081in}{2.384287in}}%
\pgfpathlineto{\pgfqpoint{3.885283in}{2.402281in}}%
\pgfpathlineto{\pgfqpoint{3.713485in}{2.490402in}}%
\pgfpathlineto{\pgfqpoint{3.541687in}{2.420332in}}%
\pgfpathlineto{\pgfqpoint{3.369889in}{2.573009in}}%
\pgfpathlineto{\pgfqpoint{3.198091in}{2.503748in}}%
\pgfpathlineto{\pgfqpoint{3.026293in}{2.539081in}}%
\pgfpathlineto{\pgfqpoint{2.854495in}{2.517602in}}%
\pgfpathlineto{\pgfqpoint{2.682697in}{2.509839in}}%
\pgfpathlineto{\pgfqpoint{2.510899in}{2.612704in}}%
\pgfpathlineto{\pgfqpoint{2.339101in}{2.522894in}}%
\pgfpathlineto{\pgfqpoint{2.167303in}{2.640179in}}%
\pgfpathlineto{\pgfqpoint{1.995505in}{2.493732in}}%
\pgfpathlineto{\pgfqpoint{1.823707in}{2.607882in}}%
\pgfpathlineto{\pgfqpoint{1.651909in}{2.777239in}}%
\pgfpathlineto{\pgfqpoint{1.480111in}{2.782985in}}%
\pgfpathlineto{\pgfqpoint{1.308313in}{2.648609in}}%
\pgfpathlineto{\pgfqpoint{1.136515in}{2.688200in}}%
\pgfpathlineto{\pgfqpoint{0.964717in}{2.937024in}}%
\pgfpathlineto{\pgfqpoint{0.792919in}{2.809590in}}%
\pgfpathlineto{\pgfqpoint{0.621121in}{2.668708in}}%
\pgfpathlineto{\pgfqpoint{0.449323in}{2.931655in}}%
\pgfpathlineto{\pgfqpoint{0.277525in}{3.052989in}}%
\pgfpathlineto{\pgfqpoint{0.105727in}{2.996174in}}%
\pgfpathlineto{\pgfqpoint{-0.066071in}{3.107934in}}%
\pgfpathclose%
\pgfusepath{stroke,fill}%
\end{pgfscope}%
\begin{pgfscope}%
\pgfpathrectangle{\pgfqpoint{0.100000in}{0.915754in}}{\pgfqpoint{4.650000in}{3.020000in}}%
\pgfusepath{clip}%
\pgfsetroundcap%
\pgfsetroundjoin%
\pgfsetlinewidth{1.505625pt}%
\definecolor{currentstroke}{rgb}{0.298039,0.447059,0.690196}%
\pgfsetstrokecolor{currentstroke}%
\pgfsetdash{}{0pt}%
\pgfpathmoveto{\pgfqpoint{0.086111in}{1.957930in}}%
\pgfpathlineto{\pgfqpoint{0.153801in}{2.003553in}}%
\pgfpathlineto{\pgfqpoint{0.326333in}{2.169395in}}%
\pgfpathlineto{\pgfqpoint{0.502402in}{2.252217in}}%
\pgfpathlineto{\pgfqpoint{0.676104in}{2.287033in}}%
\pgfpathlineto{\pgfqpoint{0.847730in}{2.321930in}}%
\pgfpathlineto{\pgfqpoint{1.026665in}{2.108566in}}%
\pgfpathlineto{\pgfqpoint{1.197882in}{2.185584in}}%
\pgfpathlineto{\pgfqpoint{1.373116in}{2.047767in}}%
\pgfpathlineto{\pgfqpoint{1.544081in}{2.168876in}}%
\pgfpathlineto{\pgfqpoint{1.720038in}{2.167645in}}%
\pgfpathlineto{\pgfqpoint{1.898061in}{2.099970in}}%
\pgfpathlineto{\pgfqpoint{2.068535in}{2.073879in}}%
\pgfpathlineto{\pgfqpoint{2.250102in}{2.133762in}}%
\pgfpathlineto{\pgfqpoint{2.419418in}{1.981034in}}%
\pgfpathlineto{\pgfqpoint{2.597532in}{2.135891in}}%
\pgfpathlineto{\pgfqpoint{2.771691in}{2.082292in}}%
\pgfpathlineto{\pgfqpoint{2.942440in}{2.127580in}}%
\pgfpathlineto{\pgfqpoint{3.118824in}{2.143275in}}%
\pgfpathlineto{\pgfqpoint{3.289559in}{2.114790in}}%
\pgfpathlineto{\pgfqpoint{3.466392in}{2.008291in}}%
\pgfpathlineto{\pgfqpoint{3.642509in}{2.099031in}}%
\pgfpathlineto{\pgfqpoint{3.823342in}{2.084189in}}%
\pgfpathlineto{\pgfqpoint{3.996368in}{2.085695in}}%
\pgfpathlineto{\pgfqpoint{4.170776in}{2.066564in}}%
\pgfpathlineto{\pgfqpoint{4.343945in}{2.059806in}}%
\pgfpathlineto{\pgfqpoint{4.519249in}{1.991812in}}%
\pgfpathlineto{\pgfqpoint{4.697520in}{2.156006in}}%
\pgfpathlineto{\pgfqpoint{4.763889in}{2.121007in}}%
\pgfusepath{stroke}%
\end{pgfscope}%
\begin{pgfscope}%
\pgfpathrectangle{\pgfqpoint{0.100000in}{0.915754in}}{\pgfqpoint{4.650000in}{3.020000in}}%
\pgfusepath{clip}%
\pgfsetroundcap%
\pgfsetroundjoin%
\pgfsetlinewidth{1.505625pt}%
\definecolor{currentstroke}{rgb}{0.866667,0.517647,0.321569}%
\pgfsetstrokecolor{currentstroke}%
\pgfsetdash{}{0pt}%
\pgfpathmoveto{\pgfqpoint{0.086111in}{1.768498in}}%
\pgfpathlineto{\pgfqpoint{0.152413in}{1.591294in}}%
\pgfpathlineto{\pgfqpoint{0.330375in}{1.834734in}}%
\pgfpathlineto{\pgfqpoint{0.499912in}{1.785037in}}%
\pgfpathlineto{\pgfqpoint{0.676644in}{1.875432in}}%
\pgfpathlineto{\pgfqpoint{0.849328in}{1.881944in}}%
\pgfpathlineto{\pgfqpoint{1.026253in}{1.696637in}}%
\pgfpathlineto{\pgfqpoint{1.198311in}{1.572472in}}%
\pgfpathlineto{\pgfqpoint{1.368900in}{1.597625in}}%
\pgfpathlineto{\pgfqpoint{1.545534in}{1.591633in}}%
\pgfpathlineto{\pgfqpoint{1.722310in}{1.588084in}}%
\pgfpathlineto{\pgfqpoint{1.895741in}{1.582771in}}%
\pgfpathlineto{\pgfqpoint{2.070309in}{1.509881in}}%
\pgfpathlineto{\pgfqpoint{2.249437in}{1.688238in}}%
\pgfpathlineto{\pgfqpoint{2.420811in}{1.625186in}}%
\pgfpathlineto{\pgfqpoint{2.592900in}{1.618541in}}%
\pgfpathlineto{\pgfqpoint{2.771634in}{1.663327in}}%
\pgfpathlineto{\pgfqpoint{2.948049in}{1.632914in}}%
\pgfpathlineto{\pgfqpoint{3.122146in}{1.543380in}}%
\pgfpathlineto{\pgfqpoint{3.293776in}{1.501169in}}%
\pgfpathlineto{\pgfqpoint{3.473490in}{1.554934in}}%
\pgfpathlineto{\pgfqpoint{3.647130in}{1.578652in}}%
\pgfpathlineto{\pgfqpoint{3.818284in}{1.626210in}}%
\pgfpathlineto{\pgfqpoint{3.993824in}{1.486238in}}%
\pgfpathlineto{\pgfqpoint{4.170370in}{1.719911in}}%
\pgfpathlineto{\pgfqpoint{4.350245in}{1.598235in}}%
\pgfpathlineto{\pgfqpoint{4.519724in}{1.562027in}}%
\pgfpathlineto{\pgfqpoint{4.698489in}{1.639690in}}%
\pgfpathlineto{\pgfqpoint{4.763889in}{1.693082in}}%
\pgfusepath{stroke}%
\end{pgfscope}%
\begin{pgfscope}%
\pgfpathrectangle{\pgfqpoint{0.100000in}{0.915754in}}{\pgfqpoint{4.650000in}{3.020000in}}%
\pgfusepath{clip}%
\pgfsetroundcap%
\pgfsetroundjoin%
\pgfsetlinewidth{1.505625pt}%
\definecolor{currentstroke}{rgb}{0.333333,0.658824,0.407843}%
\pgfsetstrokecolor{currentstroke}%
\pgfsetdash{}{0pt}%
\pgfpathmoveto{\pgfqpoint{0.086111in}{2.927625in}}%
\pgfpathlineto{\pgfqpoint{0.194795in}{2.906855in}}%
\pgfpathlineto{\pgfqpoint{0.376496in}{2.572583in}}%
\pgfpathlineto{\pgfqpoint{0.566349in}{2.589402in}}%
\pgfpathlineto{\pgfqpoint{0.742415in}{2.510620in}}%
\pgfpathlineto{\pgfqpoint{0.926648in}{2.447782in}}%
\pgfpathlineto{\pgfqpoint{1.104568in}{2.420620in}}%
\pgfpathlineto{\pgfqpoint{1.291435in}{2.380860in}}%
\pgfpathlineto{\pgfqpoint{1.499325in}{2.309151in}}%
\pgfpathlineto{\pgfqpoint{1.662708in}{2.395983in}}%
\pgfpathlineto{\pgfqpoint{1.853700in}{2.258245in}}%
\pgfpathlineto{\pgfqpoint{2.051143in}{2.274712in}}%
\pgfpathlineto{\pgfqpoint{2.232877in}{2.124880in}}%
\pgfpathlineto{\pgfqpoint{2.410870in}{2.119415in}}%
\pgfpathlineto{\pgfqpoint{2.611535in}{2.095473in}}%
\pgfpathlineto{\pgfqpoint{2.802309in}{2.264165in}}%
\pgfpathlineto{\pgfqpoint{2.976814in}{2.099558in}}%
\pgfpathlineto{\pgfqpoint{3.162703in}{2.058609in}}%
\pgfpathlineto{\pgfqpoint{3.364005in}{2.210199in}}%
\pgfpathlineto{\pgfqpoint{3.522904in}{2.086566in}}%
\pgfpathlineto{\pgfqpoint{3.717669in}{2.047568in}}%
\pgfpathlineto{\pgfqpoint{3.906390in}{2.227741in}}%
\pgfpathlineto{\pgfqpoint{4.100743in}{2.100185in}}%
\pgfpathlineto{\pgfqpoint{4.261438in}{2.041391in}}%
\pgfpathlineto{\pgfqpoint{4.456709in}{1.985889in}}%
\pgfpathlineto{\pgfqpoint{4.645937in}{2.102248in}}%
\pgfpathlineto{\pgfqpoint{4.763889in}{1.981099in}}%
\pgfusepath{stroke}%
\end{pgfscope}%
\begin{pgfscope}%
\pgfpathrectangle{\pgfqpoint{0.100000in}{0.915754in}}{\pgfqpoint{4.650000in}{3.020000in}}%
\pgfusepath{clip}%
\pgfsetroundcap%
\pgfsetroundjoin%
\pgfsetlinewidth{1.505625pt}%
\definecolor{currentstroke}{rgb}{0.768627,0.305882,0.321569}%
\pgfsetstrokecolor{currentstroke}%
\pgfsetdash{}{0pt}%
\pgfpathmoveto{\pgfqpoint{0.086111in}{2.868527in}}%
\pgfpathlineto{\pgfqpoint{0.105727in}{2.861796in}}%
\pgfpathlineto{\pgfqpoint{0.277525in}{2.927751in}}%
\pgfpathlineto{\pgfqpoint{0.449323in}{2.789066in}}%
\pgfpathlineto{\pgfqpoint{0.621121in}{2.482234in}}%
\pgfpathlineto{\pgfqpoint{0.792919in}{2.668657in}}%
\pgfpathlineto{\pgfqpoint{0.964717in}{2.794715in}}%
\pgfpathlineto{\pgfqpoint{1.136515in}{2.543371in}}%
\pgfpathlineto{\pgfqpoint{1.308313in}{2.533231in}}%
\pgfpathlineto{\pgfqpoint{1.480111in}{2.667856in}}%
\pgfpathlineto{\pgfqpoint{1.651909in}{2.643738in}}%
\pgfpathlineto{\pgfqpoint{1.823707in}{2.460374in}}%
\pgfpathlineto{\pgfqpoint{1.995505in}{2.361583in}}%
\pgfpathlineto{\pgfqpoint{2.167303in}{2.519720in}}%
\pgfpathlineto{\pgfqpoint{2.339101in}{2.377672in}}%
\pgfpathlineto{\pgfqpoint{2.510899in}{2.448622in}}%
\pgfpathlineto{\pgfqpoint{2.682697in}{2.391989in}}%
\pgfpathlineto{\pgfqpoint{2.854495in}{2.386288in}}%
\pgfpathlineto{\pgfqpoint{3.026293in}{2.401790in}}%
\pgfpathlineto{\pgfqpoint{3.198091in}{2.375306in}}%
\pgfpathlineto{\pgfqpoint{3.369889in}{2.426505in}}%
\pgfpathlineto{\pgfqpoint{3.541687in}{2.296032in}}%
\pgfpathlineto{\pgfqpoint{3.713485in}{2.340283in}}%
\pgfpathlineto{\pgfqpoint{3.885283in}{2.277563in}}%
\pgfpathlineto{\pgfqpoint{4.057081in}{2.257005in}}%
\pgfpathlineto{\pgfqpoint{4.228879in}{2.276957in}}%
\pgfpathlineto{\pgfqpoint{4.400677in}{2.270619in}}%
\pgfpathlineto{\pgfqpoint{4.572475in}{2.329202in}}%
\pgfpathlineto{\pgfqpoint{4.744273in}{2.244511in}}%
\pgfpathlineto{\pgfqpoint{4.763889in}{2.240300in}}%
\pgfusepath{stroke}%
\end{pgfscope}%
\begin{pgfscope}%
\pgfsetrectcap%
\pgfsetmiterjoin%
\pgfsetlinewidth{1.254687pt}%
\definecolor{currentstroke}{rgb}{0.800000,0.800000,0.800000}%
\pgfsetstrokecolor{currentstroke}%
\pgfsetdash{}{0pt}%
\pgfpathmoveto{\pgfqpoint{0.100000in}{0.915754in}}%
\pgfpathlineto{\pgfqpoint{0.100000in}{3.935754in}}%
\pgfusepath{stroke}%
\end{pgfscope}%
\begin{pgfscope}%
\pgfsetrectcap%
\pgfsetmiterjoin%
\pgfsetlinewidth{1.254687pt}%
\definecolor{currentstroke}{rgb}{0.800000,0.800000,0.800000}%
\pgfsetstrokecolor{currentstroke}%
\pgfsetdash{}{0pt}%
\pgfpathmoveto{\pgfqpoint{4.750000in}{0.915754in}}%
\pgfpathlineto{\pgfqpoint{4.750000in}{3.935754in}}%
\pgfusepath{stroke}%
\end{pgfscope}%
\begin{pgfscope}%
\pgfsetrectcap%
\pgfsetmiterjoin%
\pgfsetlinewidth{1.254687pt}%
\definecolor{currentstroke}{rgb}{0.800000,0.800000,0.800000}%
\pgfsetstrokecolor{currentstroke}%
\pgfsetdash{}{0pt}%
\pgfpathmoveto{\pgfqpoint{0.100000in}{0.915754in}}%
\pgfpathlineto{\pgfqpoint{4.750000in}{0.915754in}}%
\pgfusepath{stroke}%
\end{pgfscope}%
\begin{pgfscope}%
\pgfsetrectcap%
\pgfsetmiterjoin%
\pgfsetlinewidth{1.254687pt}%
\definecolor{currentstroke}{rgb}{0.800000,0.800000,0.800000}%
\pgfsetstrokecolor{currentstroke}%
\pgfsetdash{}{0pt}%
\pgfpathmoveto{\pgfqpoint{0.100000in}{3.935754in}}%
\pgfpathlineto{\pgfqpoint{4.750000in}{3.935754in}}%
\pgfusepath{stroke}%
\end{pgfscope}%
\end{pgfpicture}%
\makeatother%
\endgroup%

%% file: fig/continuous_MNIST_res.pgf
%% Creator: Matplotlib, PGF backend
%%
%% To include the figure in your LaTeX document, write
%%   \input{<filename>.pgf}
%%
%% Make sure the required packages are loaded in your preamble
%%   \usepackage{pgf}
%%
%% Figures using additional raster images can only be included by \input if
%% they are in the same directory as the main LaTeX file. For loading figures
%% from other directories you can use the `import` package
%% and then include the figures with
%%   \import{<path to file>}{<filename>.pgf}
%%
%% Matplotlib used the following preamble
%%
\begingroup%
\makeatletter%
\begin{pgfpicture}%
\pgfpathrectangle{\pgfpointorigin}{\pgfqpoint{4.850000in}{4.035754in}}%
\pgfusepath{use as bounding box, clip}%
\begin{pgfscope}%
\pgfsetbuttcap%
\pgfsetmiterjoin%
\definecolor{currentfill}{rgb}{1.000000,1.000000,1.000000}%
\pgfsetfillcolor{currentfill}%
\pgfsetlinewidth{0.000000pt}%
\definecolor{currentstroke}{rgb}{1.000000,1.000000,1.000000}%
\pgfsetstrokecolor{currentstroke}%
\pgfsetdash{}{0pt}%
\pgfpathmoveto{\pgfqpoint{0.000000in}{0.000000in}}%
\pgfpathlineto{\pgfqpoint{4.850000in}{0.000000in}}%
\pgfpathlineto{\pgfqpoint{4.850000in}{4.035754in}}%
\pgfpathlineto{\pgfqpoint{0.000000in}{4.035754in}}%
\pgfpathclose%
\pgfusepath{fill}%
\end{pgfscope}%
\begin{pgfscope}%
\pgfsetbuttcap%
\pgfsetmiterjoin%
\definecolor{currentfill}{rgb}{1.000000,1.000000,1.000000}%
\pgfsetfillcolor{currentfill}%
\pgfsetlinewidth{0.000000pt}%
\definecolor{currentstroke}{rgb}{0.000000,0.000000,0.000000}%
\pgfsetstrokecolor{currentstroke}%
\pgfsetstrokeopacity{0.000000}%
\pgfsetdash{}{0pt}%
\pgfpathmoveto{\pgfqpoint{0.100000in}{0.915754in}}%
\pgfpathlineto{\pgfqpoint{4.750000in}{0.915754in}}%
\pgfpathlineto{\pgfqpoint{4.750000in}{3.935754in}}%
\pgfpathlineto{\pgfqpoint{0.100000in}{3.935754in}}%
\pgfpathclose%
\pgfusepath{fill}%
\end{pgfscope}%
\begin{pgfscope}%
\pgfpathrectangle{\pgfqpoint{0.100000in}{0.915754in}}{\pgfqpoint{4.650000in}{3.020000in}}%
\pgfusepath{clip}%
\pgfsetroundcap%
\pgfsetroundjoin%
\pgfsetlinewidth{1.003750pt}%
\definecolor{currentstroke}{rgb}{0.800000,0.800000,0.800000}%
\pgfsetstrokecolor{currentstroke}%
\pgfsetdash{}{0pt}%
\pgfpathmoveto{\pgfqpoint{0.349107in}{0.915754in}}%
\pgfpathlineto{\pgfqpoint{0.349107in}{3.935754in}}%
\pgfusepath{stroke}%
\end{pgfscope}%
\begin{pgfscope}%
\definecolor{textcolor}{rgb}{0.150000,0.150000,0.150000}%
\pgfsetstrokecolor{textcolor}%
\pgfsetfillcolor{textcolor}%
\pgftext[x=0.349107in,y=0.783810in,,top]{\color{textcolor}\rmfamily\fontsize{23.100000}{27.720000}\selectfont 10}%
\end{pgfscope}%
\begin{pgfscope}%
\pgfpathrectangle{\pgfqpoint{0.100000in}{0.915754in}}{\pgfqpoint{4.650000in}{3.020000in}}%
\pgfusepath{clip}%
\pgfsetroundcap%
\pgfsetroundjoin%
\pgfsetlinewidth{1.003750pt}%
\definecolor{currentstroke}{rgb}{0.800000,0.800000,0.800000}%
\pgfsetstrokecolor{currentstroke}%
\pgfsetdash{}{0pt}%
\pgfpathmoveto{\pgfqpoint{1.179464in}{0.915754in}}%
\pgfpathlineto{\pgfqpoint{1.179464in}{3.935754in}}%
\pgfusepath{stroke}%
\end{pgfscope}%
\begin{pgfscope}%
\definecolor{textcolor}{rgb}{0.150000,0.150000,0.150000}%
\pgfsetstrokecolor{textcolor}%
\pgfsetfillcolor{textcolor}%
\pgftext[x=1.179464in,y=0.783810in,,top]{\color{textcolor}\rmfamily\fontsize{23.100000}{27.720000}\selectfont 20}%
\end{pgfscope}%
\begin{pgfscope}%
\pgfpathrectangle{\pgfqpoint{0.100000in}{0.915754in}}{\pgfqpoint{4.650000in}{3.020000in}}%
\pgfusepath{clip}%
\pgfsetroundcap%
\pgfsetroundjoin%
\pgfsetlinewidth{1.003750pt}%
\definecolor{currentstroke}{rgb}{0.800000,0.800000,0.800000}%
\pgfsetstrokecolor{currentstroke}%
\pgfsetdash{}{0pt}%
\pgfpathmoveto{\pgfqpoint{2.009821in}{0.915754in}}%
\pgfpathlineto{\pgfqpoint{2.009821in}{3.935754in}}%
\pgfusepath{stroke}%
\end{pgfscope}%
\begin{pgfscope}%
\definecolor{textcolor}{rgb}{0.150000,0.150000,0.150000}%
\pgfsetstrokecolor{textcolor}%
\pgfsetfillcolor{textcolor}%
\pgftext[x=2.009821in,y=0.783810in,,top]{\color{textcolor}\rmfamily\fontsize{23.100000}{27.720000}\selectfont 30}%
\end{pgfscope}%
\begin{pgfscope}%
\pgfpathrectangle{\pgfqpoint{0.100000in}{0.915754in}}{\pgfqpoint{4.650000in}{3.020000in}}%
\pgfusepath{clip}%
\pgfsetroundcap%
\pgfsetroundjoin%
\pgfsetlinewidth{1.003750pt}%
\definecolor{currentstroke}{rgb}{0.800000,0.800000,0.800000}%
\pgfsetstrokecolor{currentstroke}%
\pgfsetdash{}{0pt}%
\pgfpathmoveto{\pgfqpoint{2.840179in}{0.915754in}}%
\pgfpathlineto{\pgfqpoint{2.840179in}{3.935754in}}%
\pgfusepath{stroke}%
\end{pgfscope}%
\begin{pgfscope}%
\definecolor{textcolor}{rgb}{0.150000,0.150000,0.150000}%
\pgfsetstrokecolor{textcolor}%
\pgfsetfillcolor{textcolor}%
\pgftext[x=2.840179in,y=0.783810in,,top]{\color{textcolor}\rmfamily\fontsize{23.100000}{27.720000}\selectfont 40}%
\end{pgfscope}%
\begin{pgfscope}%
\pgfpathrectangle{\pgfqpoint{0.100000in}{0.915754in}}{\pgfqpoint{4.650000in}{3.020000in}}%
\pgfusepath{clip}%
\pgfsetroundcap%
\pgfsetroundjoin%
\pgfsetlinewidth{1.003750pt}%
\definecolor{currentstroke}{rgb}{0.800000,0.800000,0.800000}%
\pgfsetstrokecolor{currentstroke}%
\pgfsetdash{}{0pt}%
\pgfpathmoveto{\pgfqpoint{3.670536in}{0.915754in}}%
\pgfpathlineto{\pgfqpoint{3.670536in}{3.935754in}}%
\pgfusepath{stroke}%
\end{pgfscope}%
\begin{pgfscope}%
\definecolor{textcolor}{rgb}{0.150000,0.150000,0.150000}%
\pgfsetstrokecolor{textcolor}%
\pgfsetfillcolor{textcolor}%
\pgftext[x=3.670536in,y=0.783810in,,top]{\color{textcolor}\rmfamily\fontsize{23.100000}{27.720000}\selectfont 50}%
\end{pgfscope}%
\begin{pgfscope}%
\pgfpathrectangle{\pgfqpoint{0.100000in}{0.915754in}}{\pgfqpoint{4.650000in}{3.020000in}}%
\pgfusepath{clip}%
\pgfsetroundcap%
\pgfsetroundjoin%
\pgfsetlinewidth{1.003750pt}%
\definecolor{currentstroke}{rgb}{0.800000,0.800000,0.800000}%
\pgfsetstrokecolor{currentstroke}%
\pgfsetdash{}{0pt}%
\pgfpathmoveto{\pgfqpoint{4.500893in}{0.915754in}}%
\pgfpathlineto{\pgfqpoint{4.500893in}{3.935754in}}%
\pgfusepath{stroke}%
\end{pgfscope}%
\begin{pgfscope}%
\definecolor{textcolor}{rgb}{0.150000,0.150000,0.150000}%
\pgfsetstrokecolor{textcolor}%
\pgfsetfillcolor{textcolor}%
\pgftext[x=4.500893in,y=0.783810in,,top]{\color{textcolor}\rmfamily\fontsize{23.100000}{27.720000}\selectfont 60}%
\end{pgfscope}%
\begin{pgfscope}%
\definecolor{textcolor}{rgb}{0.150000,0.150000,0.150000}%
\pgfsetstrokecolor{textcolor}%
\pgfsetfillcolor{textcolor}%
\pgftext[x=2.425000in,y=0.407183in,,top]{\color{textcolor}\rmfamily\fontsize{25.200000}{30.240000}\selectfont runtime in seconds}%
\end{pgfscope}%
\begin{pgfscope}%
\pgfpathrectangle{\pgfqpoint{0.100000in}{0.915754in}}{\pgfqpoint{4.650000in}{3.020000in}}%
\pgfusepath{clip}%
\pgfsetroundcap%
\pgfsetroundjoin%
\pgfsetlinewidth{1.003750pt}%
\definecolor{currentstroke}{rgb}{0.800000,0.800000,0.800000}%
\pgfsetstrokecolor{currentstroke}%
\pgfsetdash{}{0pt}%
\pgfpathmoveto{\pgfqpoint{0.100000in}{1.328214in}}%
\pgfpathlineto{\pgfqpoint{4.750000in}{1.328214in}}%
\pgfusepath{stroke}%
\end{pgfscope}%
\begin{pgfscope}%
\pgfpathrectangle{\pgfqpoint{0.100000in}{0.915754in}}{\pgfqpoint{4.650000in}{3.020000in}}%
\pgfusepath{clip}%
\pgfsetroundcap%
\pgfsetroundjoin%
\pgfsetlinewidth{1.003750pt}%
\definecolor{currentstroke}{rgb}{0.800000,0.800000,0.800000}%
\pgfsetstrokecolor{currentstroke}%
\pgfsetdash{}{0pt}%
\pgfpathmoveto{\pgfqpoint{0.100000in}{2.698375in}}%
\pgfpathlineto{\pgfqpoint{4.750000in}{2.698375in}}%
\pgfusepath{stroke}%
\end{pgfscope}%
\begin{pgfscope}%
\pgfpathrectangle{\pgfqpoint{0.100000in}{0.915754in}}{\pgfqpoint{4.650000in}{3.020000in}}%
\pgfusepath{clip}%
\pgfsetbuttcap%
\pgfsetroundjoin%
\definecolor{currentfill}{rgb}{0.298039,0.447059,0.690196}%
\pgfsetfillcolor{currentfill}%
\pgfsetfillopacity{0.200000}%
\pgfsetlinewidth{1.003750pt}%
\definecolor{currentstroke}{rgb}{0.298039,0.447059,0.690196}%
\pgfsetstrokecolor{currentstroke}%
\pgfsetstrokeopacity{0.200000}%
\pgfsetdash{}{0pt}%
\pgfpathmoveto{\pgfqpoint{-0.019391in}{2.709450in}}%
\pgfpathlineto{\pgfqpoint{-0.019391in}{2.430322in}}%
\pgfpathlineto{\pgfqpoint{0.152086in}{2.116456in}}%
\pgfpathlineto{\pgfqpoint{0.325373in}{1.756176in}}%
\pgfpathlineto{\pgfqpoint{0.502890in}{1.711829in}}%
\pgfpathlineto{\pgfqpoint{0.673302in}{1.825552in}}%
\pgfpathlineto{\pgfqpoint{0.856156in}{1.750133in}}%
\pgfpathlineto{\pgfqpoint{1.024360in}{1.706699in}}%
\pgfpathlineto{\pgfqpoint{1.200312in}{1.584510in}}%
\pgfpathlineto{\pgfqpoint{1.377466in}{1.609232in}}%
\pgfpathlineto{\pgfqpoint{1.545153in}{1.643653in}}%
\pgfpathlineto{\pgfqpoint{1.724380in}{1.647601in}}%
\pgfpathlineto{\pgfqpoint{1.899998in}{1.616110in}}%
\pgfpathlineto{\pgfqpoint{2.079568in}{1.554133in}}%
\pgfpathlineto{\pgfqpoint{2.247244in}{1.456076in}}%
\pgfpathlineto{\pgfqpoint{2.423306in}{1.398276in}}%
\pgfpathlineto{\pgfqpoint{2.598607in}{1.290515in}}%
\pgfpathlineto{\pgfqpoint{2.777201in}{1.243452in}}%
\pgfpathlineto{\pgfqpoint{2.946057in}{1.331600in}}%
\pgfpathlineto{\pgfqpoint{3.124932in}{1.238474in}}%
\pgfpathlineto{\pgfqpoint{3.297946in}{1.200729in}}%
\pgfpathlineto{\pgfqpoint{3.473696in}{1.364778in}}%
\pgfpathlineto{\pgfqpoint{3.647530in}{1.316537in}}%
\pgfpathlineto{\pgfqpoint{3.827452in}{1.308253in}}%
\pgfpathlineto{\pgfqpoint{4.005065in}{1.269256in}}%
\pgfpathlineto{\pgfqpoint{4.177060in}{1.244601in}}%
\pgfpathlineto{\pgfqpoint{4.348488in}{1.212473in}}%
\pgfpathlineto{\pgfqpoint{4.521492in}{1.303870in}}%
\pgfpathlineto{\pgfqpoint{4.698420in}{1.197972in}}%
\pgfpathlineto{\pgfqpoint{4.881078in}{1.194379in}}%
\pgfpathlineto{\pgfqpoint{5.048128in}{1.018337in}}%
\pgfpathlineto{\pgfqpoint{5.048128in}{1.355194in}}%
\pgfpathlineto{\pgfqpoint{5.048128in}{1.355194in}}%
\pgfpathlineto{\pgfqpoint{4.881078in}{1.423531in}}%
\pgfpathlineto{\pgfqpoint{4.698420in}{1.472303in}}%
\pgfpathlineto{\pgfqpoint{4.521492in}{1.568100in}}%
\pgfpathlineto{\pgfqpoint{4.348488in}{1.494292in}}%
\pgfpathlineto{\pgfqpoint{4.177060in}{1.553254in}}%
\pgfpathlineto{\pgfqpoint{4.005065in}{1.523017in}}%
\pgfpathlineto{\pgfqpoint{3.827452in}{1.547196in}}%
\pgfpathlineto{\pgfqpoint{3.647530in}{1.564457in}}%
\pgfpathlineto{\pgfqpoint{3.473696in}{1.683962in}}%
\pgfpathlineto{\pgfqpoint{3.297946in}{1.525264in}}%
\pgfpathlineto{\pgfqpoint{3.124932in}{1.537248in}}%
\pgfpathlineto{\pgfqpoint{2.946057in}{1.616379in}}%
\pgfpathlineto{\pgfqpoint{2.777201in}{1.500398in}}%
\pgfpathlineto{\pgfqpoint{2.598607in}{1.628967in}}%
\pgfpathlineto{\pgfqpoint{2.423306in}{1.698534in}}%
\pgfpathlineto{\pgfqpoint{2.247244in}{1.693842in}}%
\pgfpathlineto{\pgfqpoint{2.079568in}{1.791951in}}%
\pgfpathlineto{\pgfqpoint{1.899998in}{1.881240in}}%
\pgfpathlineto{\pgfqpoint{1.724380in}{1.888015in}}%
\pgfpathlineto{\pgfqpoint{1.545153in}{1.888437in}}%
\pgfpathlineto{\pgfqpoint{1.377466in}{1.922818in}}%
\pgfpathlineto{\pgfqpoint{1.200312in}{1.875801in}}%
\pgfpathlineto{\pgfqpoint{1.024360in}{2.024088in}}%
\pgfpathlineto{\pgfqpoint{0.856156in}{2.004009in}}%
\pgfpathlineto{\pgfqpoint{0.673302in}{2.061508in}}%
\pgfpathlineto{\pgfqpoint{0.502890in}{2.047359in}}%
\pgfpathlineto{\pgfqpoint{0.325373in}{2.064632in}}%
\pgfpathlineto{\pgfqpoint{0.152086in}{2.463629in}}%
\pgfpathlineto{\pgfqpoint{-0.019391in}{2.709450in}}%
\pgfpathclose%
\pgfusepath{stroke,fill}%
\end{pgfscope}%
\begin{pgfscope}%
\pgfpathrectangle{\pgfqpoint{0.100000in}{0.915754in}}{\pgfqpoint{4.650000in}{3.020000in}}%
\pgfusepath{clip}%
\pgfsetbuttcap%
\pgfsetroundjoin%
\definecolor{currentfill}{rgb}{0.866667,0.517647,0.321569}%
\pgfsetfillcolor{currentfill}%
\pgfsetfillopacity{0.200000}%
\pgfsetlinewidth{1.003750pt}%
\definecolor{currentstroke}{rgb}{0.866667,0.517647,0.321569}%
\pgfsetstrokecolor{currentstroke}%
\pgfsetstrokeopacity{0.200000}%
\pgfsetdash{}{0pt}%
\pgfpathmoveto{\pgfqpoint{-0.014923in}{2.645009in}}%
\pgfpathlineto{\pgfqpoint{-0.014923in}{2.312979in}}%
\pgfpathlineto{\pgfqpoint{0.155326in}{2.445935in}}%
\pgfpathlineto{\pgfqpoint{0.322346in}{2.295626in}}%
\pgfpathlineto{\pgfqpoint{0.502199in}{1.883076in}}%
\pgfpathlineto{\pgfqpoint{0.681051in}{1.730142in}}%
\pgfpathlineto{\pgfqpoint{0.849652in}{1.564877in}}%
\pgfpathlineto{\pgfqpoint{1.025750in}{1.586793in}}%
\pgfpathlineto{\pgfqpoint{1.201022in}{1.537967in}}%
\pgfpathlineto{\pgfqpoint{1.374235in}{1.500072in}}%
\pgfpathlineto{\pgfqpoint{1.553200in}{1.561230in}}%
\pgfpathlineto{\pgfqpoint{1.726769in}{1.440416in}}%
\pgfpathlineto{\pgfqpoint{1.901770in}{1.435462in}}%
\pgfpathlineto{\pgfqpoint{2.071803in}{1.218074in}}%
\pgfpathlineto{\pgfqpoint{2.249597in}{1.356233in}}%
\pgfpathlineto{\pgfqpoint{2.421511in}{1.211167in}}%
\pgfpathlineto{\pgfqpoint{2.598394in}{1.348990in}}%
\pgfpathlineto{\pgfqpoint{2.769806in}{0.997924in}}%
\pgfpathlineto{\pgfqpoint{2.945845in}{1.189020in}}%
\pgfpathlineto{\pgfqpoint{3.120944in}{1.244065in}}%
\pgfpathlineto{\pgfqpoint{3.295882in}{1.306308in}}%
\pgfpathlineto{\pgfqpoint{3.472242in}{1.268643in}}%
\pgfpathlineto{\pgfqpoint{3.647479in}{1.162395in}}%
\pgfpathlineto{\pgfqpoint{3.833581in}{1.267862in}}%
\pgfpathlineto{\pgfqpoint{3.997646in}{1.288642in}}%
\pgfpathlineto{\pgfqpoint{4.171573in}{1.046718in}}%
\pgfpathlineto{\pgfqpoint{4.348425in}{1.155954in}}%
\pgfpathlineto{\pgfqpoint{4.519423in}{1.086188in}}%
\pgfpathlineto{\pgfqpoint{4.704724in}{1.065339in}}%
\pgfpathlineto{\pgfqpoint{4.878032in}{1.202619in}}%
\pgfpathlineto{\pgfqpoint{5.058110in}{0.889922in}}%
\pgfpathlineto{\pgfqpoint{5.058110in}{1.198010in}}%
\pgfpathlineto{\pgfqpoint{5.058110in}{1.198010in}}%
\pgfpathlineto{\pgfqpoint{4.878032in}{1.482398in}}%
\pgfpathlineto{\pgfqpoint{4.704724in}{1.383641in}}%
\pgfpathlineto{\pgfqpoint{4.519423in}{1.413909in}}%
\pgfpathlineto{\pgfqpoint{4.348425in}{1.481007in}}%
\pgfpathlineto{\pgfqpoint{4.171573in}{1.348946in}}%
\pgfpathlineto{\pgfqpoint{3.997646in}{1.542076in}}%
\pgfpathlineto{\pgfqpoint{3.833581in}{1.572283in}}%
\pgfpathlineto{\pgfqpoint{3.647479in}{1.443461in}}%
\pgfpathlineto{\pgfqpoint{3.472242in}{1.571287in}}%
\pgfpathlineto{\pgfqpoint{3.295882in}{1.566691in}}%
\pgfpathlineto{\pgfqpoint{3.120944in}{1.545523in}}%
\pgfpathlineto{\pgfqpoint{2.945845in}{1.512849in}}%
\pgfpathlineto{\pgfqpoint{2.769806in}{1.401067in}}%
\pgfpathlineto{\pgfqpoint{2.598394in}{1.648023in}}%
\pgfpathlineto{\pgfqpoint{2.421511in}{1.521207in}}%
\pgfpathlineto{\pgfqpoint{2.249597in}{1.729682in}}%
\pgfpathlineto{\pgfqpoint{2.071803in}{1.536468in}}%
\pgfpathlineto{\pgfqpoint{1.901770in}{1.732593in}}%
\pgfpathlineto{\pgfqpoint{1.726769in}{1.717882in}}%
\pgfpathlineto{\pgfqpoint{1.553200in}{1.873025in}}%
\pgfpathlineto{\pgfqpoint{1.374235in}{1.821351in}}%
\pgfpathlineto{\pgfqpoint{1.201022in}{1.811229in}}%
\pgfpathlineto{\pgfqpoint{1.025750in}{1.863523in}}%
\pgfpathlineto{\pgfqpoint{0.849652in}{1.935043in}}%
\pgfpathlineto{\pgfqpoint{0.681051in}{2.097465in}}%
\pgfpathlineto{\pgfqpoint{0.502199in}{2.178353in}}%
\pgfpathlineto{\pgfqpoint{0.322346in}{2.507143in}}%
\pgfpathlineto{\pgfqpoint{0.155326in}{2.720449in}}%
\pgfpathlineto{\pgfqpoint{-0.014923in}{2.645009in}}%
\pgfpathclose%
\pgfusepath{stroke,fill}%
\end{pgfscope}%
\begin{pgfscope}%
\pgfpathrectangle{\pgfqpoint{0.100000in}{0.915754in}}{\pgfqpoint{4.650000in}{3.020000in}}%
\pgfusepath{clip}%
\pgfsetbuttcap%
\pgfsetroundjoin%
\definecolor{currentfill}{rgb}{0.333333,0.658824,0.407843}%
\pgfsetfillcolor{currentfill}%
\pgfsetfillopacity{0.200000}%
\pgfsetlinewidth{1.003750pt}%
\definecolor{currentstroke}{rgb}{0.333333,0.658824,0.407843}%
\pgfsetstrokecolor{currentstroke}%
\pgfsetstrokeopacity{0.200000}%
\pgfsetdash{}{0pt}%
\pgfpathmoveto{\pgfqpoint{0.024321in}{3.687919in}}%
\pgfpathlineto{\pgfqpoint{0.024321in}{3.135989in}}%
\pgfpathlineto{\pgfqpoint{0.203458in}{2.732420in}}%
\pgfpathlineto{\pgfqpoint{0.404762in}{2.756974in}}%
\pgfpathlineto{\pgfqpoint{0.563835in}{2.240211in}}%
\pgfpathlineto{\pgfqpoint{0.744066in}{2.181196in}}%
\pgfpathlineto{\pgfqpoint{0.931310in}{2.077176in}}%
\pgfpathlineto{\pgfqpoint{1.113175in}{1.791729in}}%
\pgfpathlineto{\pgfqpoint{1.271257in}{1.765415in}}%
\pgfpathlineto{\pgfqpoint{1.453934in}{1.740913in}}%
\pgfpathlineto{\pgfqpoint{1.634601in}{1.458511in}}%
\pgfpathlineto{\pgfqpoint{1.821217in}{1.450051in}}%
\pgfpathlineto{\pgfqpoint{1.999614in}{1.447140in}}%
\pgfpathlineto{\pgfqpoint{2.187991in}{1.457285in}}%
\pgfpathlineto{\pgfqpoint{2.347791in}{1.254405in}}%
\pgfpathlineto{\pgfqpoint{2.528958in}{1.136449in}}%
\pgfpathlineto{\pgfqpoint{2.725574in}{1.202355in}}%
\pgfpathlineto{\pgfqpoint{2.871469in}{0.997916in}}%
\pgfpathlineto{\pgfqpoint{3.067534in}{1.272374in}}%
\pgfpathlineto{\pgfqpoint{3.248891in}{1.174220in}}%
\pgfpathlineto{\pgfqpoint{3.436789in}{1.297874in}}%
\pgfpathlineto{\pgfqpoint{3.609385in}{1.118614in}}%
\pgfpathlineto{\pgfqpoint{3.789587in}{1.109496in}}%
\pgfpathlineto{\pgfqpoint{3.968320in}{1.139813in}}%
\pgfpathlineto{\pgfqpoint{4.169346in}{1.232957in}}%
\pgfpathlineto{\pgfqpoint{4.328482in}{0.829355in}}%
\pgfpathlineto{\pgfqpoint{4.495397in}{1.168261in}}%
\pgfpathlineto{\pgfqpoint{4.690334in}{0.975795in}}%
\pgfpathlineto{\pgfqpoint{4.882766in}{0.967527in}}%
\pgfpathlineto{\pgfqpoint{5.055855in}{0.955933in}}%
\pgfpathlineto{\pgfqpoint{5.226014in}{1.038818in}}%
\pgfpathlineto{\pgfqpoint{5.226014in}{1.362070in}}%
\pgfpathlineto{\pgfqpoint{5.226014in}{1.362070in}}%
\pgfpathlineto{\pgfqpoint{5.055855in}{1.285489in}}%
\pgfpathlineto{\pgfqpoint{4.882766in}{1.287693in}}%
\pgfpathlineto{\pgfqpoint{4.690334in}{1.292119in}}%
\pgfpathlineto{\pgfqpoint{4.495397in}{1.458056in}}%
\pgfpathlineto{\pgfqpoint{4.328482in}{1.237523in}}%
\pgfpathlineto{\pgfqpoint{4.169346in}{1.556642in}}%
\pgfpathlineto{\pgfqpoint{3.968320in}{1.437796in}}%
\pgfpathlineto{\pgfqpoint{3.789587in}{1.442853in}}%
\pgfpathlineto{\pgfqpoint{3.609385in}{1.457641in}}%
\pgfpathlineto{\pgfqpoint{3.436789in}{1.576339in}}%
\pgfpathlineto{\pgfqpoint{3.248891in}{1.465501in}}%
\pgfpathlineto{\pgfqpoint{3.067534in}{1.625531in}}%
\pgfpathlineto{\pgfqpoint{2.871469in}{1.354800in}}%
\pgfpathlineto{\pgfqpoint{2.725574in}{1.521338in}}%
\pgfpathlineto{\pgfqpoint{2.528958in}{1.474522in}}%
\pgfpathlineto{\pgfqpoint{2.347791in}{1.598577in}}%
\pgfpathlineto{\pgfqpoint{2.187991in}{1.743296in}}%
\pgfpathlineto{\pgfqpoint{1.999614in}{1.756371in}}%
\pgfpathlineto{\pgfqpoint{1.821217in}{1.778125in}}%
\pgfpathlineto{\pgfqpoint{1.634601in}{1.867822in}}%
\pgfpathlineto{\pgfqpoint{1.453934in}{2.037611in}}%
\pgfpathlineto{\pgfqpoint{1.271257in}{2.046973in}}%
\pgfpathlineto{\pgfqpoint{1.113175in}{2.112095in}}%
\pgfpathlineto{\pgfqpoint{0.931310in}{2.551078in}}%
\pgfpathlineto{\pgfqpoint{0.744066in}{2.524684in}}%
\pgfpathlineto{\pgfqpoint{0.563835in}{2.787046in}}%
\pgfpathlineto{\pgfqpoint{0.404762in}{3.174311in}}%
\pgfpathlineto{\pgfqpoint{0.203458in}{3.126442in}}%
\pgfpathlineto{\pgfqpoint{0.024321in}{3.687919in}}%
\pgfpathclose%
\pgfusepath{stroke,fill}%
\end{pgfscope}%
\begin{pgfscope}%
\pgfpathrectangle{\pgfqpoint{0.100000in}{0.915754in}}{\pgfqpoint{4.650000in}{3.020000in}}%
\pgfusepath{clip}%
\pgfsetbuttcap%
\pgfsetroundjoin%
\definecolor{currentfill}{rgb}{0.768627,0.305882,0.321569}%
\pgfsetfillcolor{currentfill}%
\pgfsetfillopacity{0.200000}%
\pgfsetlinewidth{1.003750pt}%
\definecolor{currentstroke}{rgb}{0.768627,0.305882,0.321569}%
\pgfsetstrokecolor{currentstroke}%
\pgfsetstrokeopacity{0.200000}%
\pgfsetdash{}{0pt}%
\pgfpathmoveto{\pgfqpoint{-0.066071in}{3.346897in}}%
\pgfpathlineto{\pgfqpoint{-0.066071in}{2.997886in}}%
\pgfpathlineto{\pgfqpoint{0.105727in}{2.960492in}}%
\pgfpathlineto{\pgfqpoint{0.277525in}{2.702699in}}%
\pgfpathlineto{\pgfqpoint{0.449323in}{2.652345in}}%
\pgfpathlineto{\pgfqpoint{0.621121in}{2.743735in}}%
\pgfpathlineto{\pgfqpoint{0.792919in}{2.531917in}}%
\pgfpathlineto{\pgfqpoint{0.964717in}{2.593899in}}%
\pgfpathlineto{\pgfqpoint{1.136515in}{2.571242in}}%
\pgfpathlineto{\pgfqpoint{1.308313in}{2.383365in}}%
\pgfpathlineto{\pgfqpoint{1.480111in}{2.330132in}}%
\pgfpathlineto{\pgfqpoint{1.651909in}{2.340196in}}%
\pgfpathlineto{\pgfqpoint{1.823707in}{2.359016in}}%
\pgfpathlineto{\pgfqpoint{1.995505in}{2.466734in}}%
\pgfpathlineto{\pgfqpoint{2.167303in}{2.370972in}}%
\pgfpathlineto{\pgfqpoint{2.339101in}{2.200388in}}%
\pgfpathlineto{\pgfqpoint{2.510899in}{2.490631in}}%
\pgfpathlineto{\pgfqpoint{2.682697in}{2.366108in}}%
\pgfpathlineto{\pgfqpoint{2.854495in}{2.244324in}}%
\pgfpathlineto{\pgfqpoint{3.026293in}{2.239368in}}%
\pgfpathlineto{\pgfqpoint{3.198091in}{2.283087in}}%
\pgfpathlineto{\pgfqpoint{3.369889in}{2.191097in}}%
\pgfpathlineto{\pgfqpoint{3.541687in}{2.252690in}}%
\pgfpathlineto{\pgfqpoint{3.713485in}{2.162737in}}%
\pgfpathlineto{\pgfqpoint{3.885283in}{2.114858in}}%
\pgfpathlineto{\pgfqpoint{4.057081in}{2.090171in}}%
\pgfpathlineto{\pgfqpoint{4.228879in}{2.083829in}}%
\pgfpathlineto{\pgfqpoint{4.400677in}{2.100912in}}%
\pgfpathlineto{\pgfqpoint{4.572475in}{2.242428in}}%
\pgfpathlineto{\pgfqpoint{4.744273in}{1.881308in}}%
\pgfpathlineto{\pgfqpoint{4.916071in}{2.287720in}}%
\pgfpathlineto{\pgfqpoint{4.916071in}{2.529313in}}%
\pgfpathlineto{\pgfqpoint{4.916071in}{2.529313in}}%
\pgfpathlineto{\pgfqpoint{4.744273in}{2.254916in}}%
\pgfpathlineto{\pgfqpoint{4.572475in}{2.563831in}}%
\pgfpathlineto{\pgfqpoint{4.400677in}{2.432619in}}%
\pgfpathlineto{\pgfqpoint{4.228879in}{2.360795in}}%
\pgfpathlineto{\pgfqpoint{4.057081in}{2.420410in}}%
\pgfpathlineto{\pgfqpoint{3.885283in}{2.470568in}}%
\pgfpathlineto{\pgfqpoint{3.713485in}{2.490399in}}%
\pgfpathlineto{\pgfqpoint{3.541687in}{2.586464in}}%
\pgfpathlineto{\pgfqpoint{3.369889in}{2.584375in}}%
\pgfpathlineto{\pgfqpoint{3.198091in}{2.585374in}}%
\pgfpathlineto{\pgfqpoint{3.026293in}{2.537698in}}%
\pgfpathlineto{\pgfqpoint{2.854495in}{2.572241in}}%
\pgfpathlineto{\pgfqpoint{2.682697in}{2.669842in}}%
\pgfpathlineto{\pgfqpoint{2.510899in}{2.827320in}}%
\pgfpathlineto{\pgfqpoint{2.339101in}{2.580268in}}%
\pgfpathlineto{\pgfqpoint{2.167303in}{2.660563in}}%
\pgfpathlineto{\pgfqpoint{1.995505in}{2.729718in}}%
\pgfpathlineto{\pgfqpoint{1.823707in}{2.677459in}}%
\pgfpathlineto{\pgfqpoint{1.651909in}{2.699161in}}%
\pgfpathlineto{\pgfqpoint{1.480111in}{2.665780in}}%
\pgfpathlineto{\pgfqpoint{1.308313in}{2.728595in}}%
\pgfpathlineto{\pgfqpoint{1.136515in}{2.891867in}}%
\pgfpathlineto{\pgfqpoint{0.964717in}{2.892495in}}%
\pgfpathlineto{\pgfqpoint{0.792919in}{2.841870in}}%
\pgfpathlineto{\pgfqpoint{0.621121in}{3.035854in}}%
\pgfpathlineto{\pgfqpoint{0.449323in}{2.917902in}}%
\pgfpathlineto{\pgfqpoint{0.277525in}{2.987969in}}%
\pgfpathlineto{\pgfqpoint{0.105727in}{3.221034in}}%
\pgfpathlineto{\pgfqpoint{-0.066071in}{3.346897in}}%
\pgfpathclose%
\pgfusepath{stroke,fill}%
\end{pgfscope}%
\begin{pgfscope}%
\pgfpathrectangle{\pgfqpoint{0.100000in}{0.915754in}}{\pgfqpoint{4.650000in}{3.020000in}}%
\pgfusepath{clip}%
\pgfsetroundcap%
\pgfsetroundjoin%
\pgfsetlinewidth{1.505625pt}%
\definecolor{currentstroke}{rgb}{0.298039,0.447059,0.690196}%
\pgfsetstrokecolor{currentstroke}%
\pgfsetdash{}{0pt}%
\pgfpathmoveto{\pgfqpoint{0.086111in}{2.414328in}}%
\pgfpathlineto{\pgfqpoint{0.152086in}{2.307455in}}%
\pgfpathlineto{\pgfqpoint{0.325373in}{1.921146in}}%
\pgfpathlineto{\pgfqpoint{0.502890in}{1.886353in}}%
\pgfpathlineto{\pgfqpoint{0.673302in}{1.956816in}}%
\pgfpathlineto{\pgfqpoint{0.856156in}{1.892392in}}%
\pgfpathlineto{\pgfqpoint{1.024360in}{1.886222in}}%
\pgfpathlineto{\pgfqpoint{1.200312in}{1.743773in}}%
\pgfpathlineto{\pgfqpoint{1.377466in}{1.776210in}}%
\pgfpathlineto{\pgfqpoint{1.545153in}{1.772432in}}%
\pgfpathlineto{\pgfqpoint{1.724380in}{1.779829in}}%
\pgfpathlineto{\pgfqpoint{1.899998in}{1.773319in}}%
\pgfpathlineto{\pgfqpoint{2.079568in}{1.678583in}}%
\pgfpathlineto{\pgfqpoint{2.247244in}{1.590792in}}%
\pgfpathlineto{\pgfqpoint{2.423306in}{1.556013in}}%
\pgfpathlineto{\pgfqpoint{2.598607in}{1.483927in}}%
\pgfpathlineto{\pgfqpoint{2.777201in}{1.387715in}}%
\pgfpathlineto{\pgfqpoint{2.946057in}{1.478956in}}%
\pgfpathlineto{\pgfqpoint{3.124932in}{1.402468in}}%
\pgfpathlineto{\pgfqpoint{3.297946in}{1.375640in}}%
\pgfpathlineto{\pgfqpoint{3.473696in}{1.541477in}}%
\pgfpathlineto{\pgfqpoint{3.647530in}{1.448939in}}%
\pgfpathlineto{\pgfqpoint{3.827452in}{1.432081in}}%
\pgfpathlineto{\pgfqpoint{4.005065in}{1.405702in}}%
\pgfpathlineto{\pgfqpoint{4.177060in}{1.406011in}}%
\pgfpathlineto{\pgfqpoint{4.348488in}{1.368916in}}%
\pgfpathlineto{\pgfqpoint{4.521492in}{1.453194in}}%
\pgfpathlineto{\pgfqpoint{4.698420in}{1.353376in}}%
\pgfpathlineto{\pgfqpoint{4.763889in}{1.338598in}}%
\pgfusepath{stroke}%
\end{pgfscope}%
\begin{pgfscope}%
\pgfpathrectangle{\pgfqpoint{0.100000in}{0.915754in}}{\pgfqpoint{4.650000in}{3.020000in}}%
\pgfusepath{clip}%
\pgfsetroundcap%
\pgfsetroundjoin%
\pgfsetlinewidth{1.505625pt}%
\definecolor{currentstroke}{rgb}{0.866667,0.517647,0.321569}%
\pgfsetstrokecolor{currentstroke}%
\pgfsetdash{}{0pt}%
\pgfpathmoveto{\pgfqpoint{0.086111in}{2.556579in}}%
\pgfpathlineto{\pgfqpoint{0.155326in}{2.596928in}}%
\pgfpathlineto{\pgfqpoint{0.322346in}{2.404604in}}%
\pgfpathlineto{\pgfqpoint{0.502199in}{2.041569in}}%
\pgfpathlineto{\pgfqpoint{0.681051in}{1.941659in}}%
\pgfpathlineto{\pgfqpoint{0.849652in}{1.769981in}}%
\pgfpathlineto{\pgfqpoint{1.025750in}{1.731834in}}%
\pgfpathlineto{\pgfqpoint{1.201022in}{1.696357in}}%
\pgfpathlineto{\pgfqpoint{1.374235in}{1.677427in}}%
\pgfpathlineto{\pgfqpoint{1.553200in}{1.734567in}}%
\pgfpathlineto{\pgfqpoint{1.726769in}{1.587953in}}%
\pgfpathlineto{\pgfqpoint{1.901770in}{1.603785in}}%
\pgfpathlineto{\pgfqpoint{2.071803in}{1.400809in}}%
\pgfpathlineto{\pgfqpoint{2.249597in}{1.557606in}}%
\pgfpathlineto{\pgfqpoint{2.421511in}{1.381481in}}%
\pgfpathlineto{\pgfqpoint{2.598394in}{1.516272in}}%
\pgfpathlineto{\pgfqpoint{2.769806in}{1.222346in}}%
\pgfpathlineto{\pgfqpoint{2.945845in}{1.375622in}}%
\pgfpathlineto{\pgfqpoint{3.120944in}{1.411274in}}%
\pgfpathlineto{\pgfqpoint{3.295882in}{1.439606in}}%
\pgfpathlineto{\pgfqpoint{3.472242in}{1.428544in}}%
\pgfpathlineto{\pgfqpoint{3.647479in}{1.317341in}}%
\pgfpathlineto{\pgfqpoint{3.833581in}{1.432597in}}%
\pgfpathlineto{\pgfqpoint{3.997646in}{1.423125in}}%
\pgfpathlineto{\pgfqpoint{4.171573in}{1.205981in}}%
\pgfpathlineto{\pgfqpoint{4.348425in}{1.338429in}}%
\pgfpathlineto{\pgfqpoint{4.519423in}{1.268892in}}%
\pgfpathlineto{\pgfqpoint{4.704724in}{1.239284in}}%
\pgfpathlineto{\pgfqpoint{4.763889in}{1.278352in}}%
\pgfusepath{stroke}%
\end{pgfscope}%
\begin{pgfscope}%
\pgfpathrectangle{\pgfqpoint{0.100000in}{0.915754in}}{\pgfqpoint{4.650000in}{3.020000in}}%
\pgfusepath{clip}%
\pgfsetroundcap%
\pgfsetroundjoin%
\pgfsetlinewidth{1.505625pt}%
\definecolor{currentstroke}{rgb}{0.333333,0.658824,0.407843}%
\pgfsetstrokecolor{currentstroke}%
\pgfsetdash{}{0pt}%
\pgfpathmoveto{\pgfqpoint{0.086111in}{3.282634in}}%
\pgfpathlineto{\pgfqpoint{0.203458in}{2.963778in}}%
\pgfpathlineto{\pgfqpoint{0.404762in}{2.978195in}}%
\pgfpathlineto{\pgfqpoint{0.563835in}{2.544607in}}%
\pgfpathlineto{\pgfqpoint{0.744066in}{2.359557in}}%
\pgfpathlineto{\pgfqpoint{0.931310in}{2.344296in}}%
\pgfpathlineto{\pgfqpoint{1.113175in}{1.956469in}}%
\pgfpathlineto{\pgfqpoint{1.271257in}{1.916369in}}%
\pgfpathlineto{\pgfqpoint{1.453934in}{1.899497in}}%
\pgfpathlineto{\pgfqpoint{1.634601in}{1.688674in}}%
\pgfpathlineto{\pgfqpoint{1.821217in}{1.631133in}}%
\pgfpathlineto{\pgfqpoint{1.999614in}{1.614363in}}%
\pgfpathlineto{\pgfqpoint{2.187991in}{1.612670in}}%
\pgfpathlineto{\pgfqpoint{2.347791in}{1.434289in}}%
\pgfpathlineto{\pgfqpoint{2.528958in}{1.319289in}}%
\pgfpathlineto{\pgfqpoint{2.725574in}{1.377232in}}%
\pgfpathlineto{\pgfqpoint{2.871469in}{1.190121in}}%
\pgfpathlineto{\pgfqpoint{3.067534in}{1.456493in}}%
\pgfpathlineto{\pgfqpoint{3.248891in}{1.327952in}}%
\pgfpathlineto{\pgfqpoint{3.436789in}{1.450075in}}%
\pgfpathlineto{\pgfqpoint{3.609385in}{1.307865in}}%
\pgfpathlineto{\pgfqpoint{3.789587in}{1.291669in}}%
\pgfpathlineto{\pgfqpoint{3.968320in}{1.299754in}}%
\pgfpathlineto{\pgfqpoint{4.169346in}{1.415053in}}%
\pgfpathlineto{\pgfqpoint{4.328482in}{1.043633in}}%
\pgfpathlineto{\pgfqpoint{4.495397in}{1.320962in}}%
\pgfpathlineto{\pgfqpoint{4.690334in}{1.153439in}}%
\pgfpathlineto{\pgfqpoint{4.763889in}{1.148080in}}%
\pgfusepath{stroke}%
\end{pgfscope}%
\begin{pgfscope}%
\pgfpathrectangle{\pgfqpoint{0.100000in}{0.915754in}}{\pgfqpoint{4.650000in}{3.020000in}}%
\pgfusepath{clip}%
\pgfsetroundcap%
\pgfsetroundjoin%
\pgfsetlinewidth{1.505625pt}%
\definecolor{currentstroke}{rgb}{0.768627,0.305882,0.321569}%
\pgfsetstrokecolor{currentstroke}%
\pgfsetdash{}{0pt}%
\pgfpathmoveto{\pgfqpoint{0.086111in}{3.114346in}}%
\pgfpathlineto{\pgfqpoint{0.105727in}{3.104520in}}%
\pgfpathlineto{\pgfqpoint{0.277525in}{2.848031in}}%
\pgfpathlineto{\pgfqpoint{0.449323in}{2.801227in}}%
\pgfpathlineto{\pgfqpoint{0.621121in}{2.905833in}}%
\pgfpathlineto{\pgfqpoint{0.792919in}{2.706638in}}%
\pgfpathlineto{\pgfqpoint{0.964717in}{2.757483in}}%
\pgfpathlineto{\pgfqpoint{1.136515in}{2.747068in}}%
\pgfpathlineto{\pgfqpoint{1.308313in}{2.571621in}}%
\pgfpathlineto{\pgfqpoint{1.480111in}{2.512017in}}%
\pgfpathlineto{\pgfqpoint{1.651909in}{2.538869in}}%
\pgfpathlineto{\pgfqpoint{1.823707in}{2.543082in}}%
\pgfpathlineto{\pgfqpoint{1.995505in}{2.611442in}}%
\pgfpathlineto{\pgfqpoint{2.167303in}{2.533010in}}%
\pgfpathlineto{\pgfqpoint{2.339101in}{2.412648in}}%
\pgfpathlineto{\pgfqpoint{2.510899in}{2.677565in}}%
\pgfpathlineto{\pgfqpoint{2.682697in}{2.535378in}}%
\pgfpathlineto{\pgfqpoint{2.854495in}{2.421159in}}%
\pgfpathlineto{\pgfqpoint{3.026293in}{2.400215in}}%
\pgfpathlineto{\pgfqpoint{3.198091in}{2.446647in}}%
\pgfpathlineto{\pgfqpoint{3.369889in}{2.404024in}}%
\pgfpathlineto{\pgfqpoint{3.541687in}{2.436320in}}%
\pgfpathlineto{\pgfqpoint{3.713485in}{2.335386in}}%
\pgfpathlineto{\pgfqpoint{3.885283in}{2.317459in}}%
\pgfpathlineto{\pgfqpoint{4.057081in}{2.269780in}}%
\pgfpathlineto{\pgfqpoint{4.228879in}{2.237300in}}%
\pgfpathlineto{\pgfqpoint{4.400677in}{2.286307in}}%
\pgfpathlineto{\pgfqpoint{4.572475in}{2.418242in}}%
\pgfpathlineto{\pgfqpoint{4.744273in}{2.077074in}}%
\pgfpathlineto{\pgfqpoint{4.763889in}{2.115905in}}%
\pgfusepath{stroke}%
\end{pgfscope}%
\begin{pgfscope}%
\pgfsetrectcap%
\pgfsetmiterjoin%
\pgfsetlinewidth{1.254687pt}%
\definecolor{currentstroke}{rgb}{0.800000,0.800000,0.800000}%
\pgfsetstrokecolor{currentstroke}%
\pgfsetdash{}{0pt}%
\pgfpathmoveto{\pgfqpoint{0.100000in}{0.915754in}}%
\pgfpathlineto{\pgfqpoint{0.100000in}{3.935754in}}%
\pgfusepath{stroke}%
\end{pgfscope}%
\begin{pgfscope}%
\pgfsetrectcap%
\pgfsetmiterjoin%
\pgfsetlinewidth{1.254687pt}%
\definecolor{currentstroke}{rgb}{0.800000,0.800000,0.800000}%
\pgfsetstrokecolor{currentstroke}%
\pgfsetdash{}{0pt}%
\pgfpathmoveto{\pgfqpoint{4.750000in}{0.915754in}}%
\pgfpathlineto{\pgfqpoint{4.750000in}{3.935754in}}%
\pgfusepath{stroke}%
\end{pgfscope}%
\begin{pgfscope}%
\pgfsetrectcap%
\pgfsetmiterjoin%
\pgfsetlinewidth{1.254687pt}%
\definecolor{currentstroke}{rgb}{0.800000,0.800000,0.800000}%
\pgfsetstrokecolor{currentstroke}%
\pgfsetdash{}{0pt}%
\pgfpathmoveto{\pgfqpoint{0.100000in}{0.915754in}}%
\pgfpathlineto{\pgfqpoint{4.750000in}{0.915754in}}%
\pgfusepath{stroke}%
\end{pgfscope}%
\begin{pgfscope}%
\pgfsetrectcap%
\pgfsetmiterjoin%
\pgfsetlinewidth{1.254687pt}%
\definecolor{currentstroke}{rgb}{0.800000,0.800000,0.800000}%
\pgfsetstrokecolor{currentstroke}%
\pgfsetdash{}{0pt}%
\pgfpathmoveto{\pgfqpoint{0.100000in}{3.935754in}}%
\pgfpathlineto{\pgfqpoint{4.750000in}{3.935754in}}%
\pgfusepath{stroke}%
\end{pgfscope}%
\end{pgfpicture}%
\makeatother%
\endgroup%

%% file: fig/continuous_ADK_res.pgf
%% Creator: Matplotlib, PGF backend
%%
%% To include the figure in your LaTeX document, write
%%   \input{<filename>.pgf}
%%
%% Make sure the required packages are loaded in your preamble
%%   \usepackage{pgf}
%%
%% Figures using additional raster images can only be included by \input if
%% they are in the same directory as the main LaTeX file. For loading figures
%% from other directories you can use the `import` package
%% and then include the figures with
%%   \import{<path to file>}{<filename>.pgf}
%%
%% Matplotlib used the following preamble
%%
\begingroup%
\makeatletter%
\begin{pgfpicture}%
\pgfpathrectangle{\pgfpointorigin}{\pgfqpoint{4.850000in}{4.035754in}}%
\pgfusepath{use as bounding box, clip}%
\begin{pgfscope}%
\pgfsetbuttcap%
\pgfsetmiterjoin%
\definecolor{currentfill}{rgb}{1.000000,1.000000,1.000000}%
\pgfsetfillcolor{currentfill}%
\pgfsetlinewidth{0.000000pt}%
\definecolor{currentstroke}{rgb}{1.000000,1.000000,1.000000}%
\pgfsetstrokecolor{currentstroke}%
\pgfsetdash{}{0pt}%
\pgfpathmoveto{\pgfqpoint{0.000000in}{0.000000in}}%
\pgfpathlineto{\pgfqpoint{4.850000in}{0.000000in}}%
\pgfpathlineto{\pgfqpoint{4.850000in}{4.035754in}}%
\pgfpathlineto{\pgfqpoint{0.000000in}{4.035754in}}%
\pgfpathclose%
\pgfusepath{fill}%
\end{pgfscope}%
\begin{pgfscope}%
\pgfsetbuttcap%
\pgfsetmiterjoin%
\definecolor{currentfill}{rgb}{1.000000,1.000000,1.000000}%
\pgfsetfillcolor{currentfill}%
\pgfsetlinewidth{0.000000pt}%
\definecolor{currentstroke}{rgb}{0.000000,0.000000,0.000000}%
\pgfsetstrokecolor{currentstroke}%
\pgfsetstrokeopacity{0.000000}%
\pgfsetdash{}{0pt}%
\pgfpathmoveto{\pgfqpoint{0.100000in}{0.915754in}}%
\pgfpathlineto{\pgfqpoint{4.750000in}{0.915754in}}%
\pgfpathlineto{\pgfqpoint{4.750000in}{3.935754in}}%
\pgfpathlineto{\pgfqpoint{0.100000in}{3.935754in}}%
\pgfpathclose%
\pgfusepath{fill}%
\end{pgfscope}%
\begin{pgfscope}%
\pgfpathrectangle{\pgfqpoint{0.100000in}{0.915754in}}{\pgfqpoint{4.650000in}{3.020000in}}%
\pgfusepath{clip}%
\pgfsetroundcap%
\pgfsetroundjoin%
\pgfsetlinewidth{1.003750pt}%
\definecolor{currentstroke}{rgb}{0.800000,0.800000,0.800000}%
\pgfsetstrokecolor{currentstroke}%
\pgfsetdash{}{0pt}%
\pgfpathmoveto{\pgfqpoint{0.349107in}{0.915754in}}%
\pgfpathlineto{\pgfqpoint{0.349107in}{3.935754in}}%
\pgfusepath{stroke}%
\end{pgfscope}%
\begin{pgfscope}%
\definecolor{textcolor}{rgb}{0.150000,0.150000,0.150000}%
\pgfsetstrokecolor{textcolor}%
\pgfsetfillcolor{textcolor}%
\pgftext[x=0.349107in,y=0.783810in,,top]{\color{textcolor}\rmfamily\fontsize{23.100000}{27.720000}\selectfont 10}%
\end{pgfscope}%
\begin{pgfscope}%
\pgfpathrectangle{\pgfqpoint{0.100000in}{0.915754in}}{\pgfqpoint{4.650000in}{3.020000in}}%
\pgfusepath{clip}%
\pgfsetroundcap%
\pgfsetroundjoin%
\pgfsetlinewidth{1.003750pt}%
\definecolor{currentstroke}{rgb}{0.800000,0.800000,0.800000}%
\pgfsetstrokecolor{currentstroke}%
\pgfsetdash{}{0pt}%
\pgfpathmoveto{\pgfqpoint{1.179464in}{0.915754in}}%
\pgfpathlineto{\pgfqpoint{1.179464in}{3.935754in}}%
\pgfusepath{stroke}%
\end{pgfscope}%
\begin{pgfscope}%
\definecolor{textcolor}{rgb}{0.150000,0.150000,0.150000}%
\pgfsetstrokecolor{textcolor}%
\pgfsetfillcolor{textcolor}%
\pgftext[x=1.179464in,y=0.783810in,,top]{\color{textcolor}\rmfamily\fontsize{23.100000}{27.720000}\selectfont 20}%
\end{pgfscope}%
\begin{pgfscope}%
\pgfpathrectangle{\pgfqpoint{0.100000in}{0.915754in}}{\pgfqpoint{4.650000in}{3.020000in}}%
\pgfusepath{clip}%
\pgfsetroundcap%
\pgfsetroundjoin%
\pgfsetlinewidth{1.003750pt}%
\definecolor{currentstroke}{rgb}{0.800000,0.800000,0.800000}%
\pgfsetstrokecolor{currentstroke}%
\pgfsetdash{}{0pt}%
\pgfpathmoveto{\pgfqpoint{2.009821in}{0.915754in}}%
\pgfpathlineto{\pgfqpoint{2.009821in}{3.935754in}}%
\pgfusepath{stroke}%
\end{pgfscope}%
\begin{pgfscope}%
\definecolor{textcolor}{rgb}{0.150000,0.150000,0.150000}%
\pgfsetstrokecolor{textcolor}%
\pgfsetfillcolor{textcolor}%
\pgftext[x=2.009821in,y=0.783810in,,top]{\color{textcolor}\rmfamily\fontsize{23.100000}{27.720000}\selectfont 30}%
\end{pgfscope}%
\begin{pgfscope}%
\pgfpathrectangle{\pgfqpoint{0.100000in}{0.915754in}}{\pgfqpoint{4.650000in}{3.020000in}}%
\pgfusepath{clip}%
\pgfsetroundcap%
\pgfsetroundjoin%
\pgfsetlinewidth{1.003750pt}%
\definecolor{currentstroke}{rgb}{0.800000,0.800000,0.800000}%
\pgfsetstrokecolor{currentstroke}%
\pgfsetdash{}{0pt}%
\pgfpathmoveto{\pgfqpoint{2.840179in}{0.915754in}}%
\pgfpathlineto{\pgfqpoint{2.840179in}{3.935754in}}%
\pgfusepath{stroke}%
\end{pgfscope}%
\begin{pgfscope}%
\definecolor{textcolor}{rgb}{0.150000,0.150000,0.150000}%
\pgfsetstrokecolor{textcolor}%
\pgfsetfillcolor{textcolor}%
\pgftext[x=2.840179in,y=0.783810in,,top]{\color{textcolor}\rmfamily\fontsize{23.100000}{27.720000}\selectfont 40}%
\end{pgfscope}%
\begin{pgfscope}%
\pgfpathrectangle{\pgfqpoint{0.100000in}{0.915754in}}{\pgfqpoint{4.650000in}{3.020000in}}%
\pgfusepath{clip}%
\pgfsetroundcap%
\pgfsetroundjoin%
\pgfsetlinewidth{1.003750pt}%
\definecolor{currentstroke}{rgb}{0.800000,0.800000,0.800000}%
\pgfsetstrokecolor{currentstroke}%
\pgfsetdash{}{0pt}%
\pgfpathmoveto{\pgfqpoint{3.670536in}{0.915754in}}%
\pgfpathlineto{\pgfqpoint{3.670536in}{3.935754in}}%
\pgfusepath{stroke}%
\end{pgfscope}%
\begin{pgfscope}%
\definecolor{textcolor}{rgb}{0.150000,0.150000,0.150000}%
\pgfsetstrokecolor{textcolor}%
\pgfsetfillcolor{textcolor}%
\pgftext[x=3.670536in,y=0.783810in,,top]{\color{textcolor}\rmfamily\fontsize{23.100000}{27.720000}\selectfont 50}%
\end{pgfscope}%
\begin{pgfscope}%
\pgfpathrectangle{\pgfqpoint{0.100000in}{0.915754in}}{\pgfqpoint{4.650000in}{3.020000in}}%
\pgfusepath{clip}%
\pgfsetroundcap%
\pgfsetroundjoin%
\pgfsetlinewidth{1.003750pt}%
\definecolor{currentstroke}{rgb}{0.800000,0.800000,0.800000}%
\pgfsetstrokecolor{currentstroke}%
\pgfsetdash{}{0pt}%
\pgfpathmoveto{\pgfqpoint{4.500893in}{0.915754in}}%
\pgfpathlineto{\pgfqpoint{4.500893in}{3.935754in}}%
\pgfusepath{stroke}%
\end{pgfscope}%
\begin{pgfscope}%
\definecolor{textcolor}{rgb}{0.150000,0.150000,0.150000}%
\pgfsetstrokecolor{textcolor}%
\pgfsetfillcolor{textcolor}%
\pgftext[x=4.500893in,y=0.783810in,,top]{\color{textcolor}\rmfamily\fontsize{23.100000}{27.720000}\selectfont 60}%
\end{pgfscope}%
\begin{pgfscope}%
\definecolor{textcolor}{rgb}{0.150000,0.150000,0.150000}%
\pgfsetstrokecolor{textcolor}%
\pgfsetfillcolor{textcolor}%
\pgftext[x=2.425000in,y=0.407183in,,top]{\color{textcolor}\rmfamily\fontsize{25.200000}{30.240000}\selectfont runtime in seconds}%
\end{pgfscope}%
\begin{pgfscope}%
\pgfpathrectangle{\pgfqpoint{0.100000in}{0.915754in}}{\pgfqpoint{4.650000in}{3.020000in}}%
\pgfusepath{clip}%
\pgfsetroundcap%
\pgfsetroundjoin%
\pgfsetlinewidth{1.003750pt}%
\definecolor{currentstroke}{rgb}{0.800000,0.800000,0.800000}%
\pgfsetstrokecolor{currentstroke}%
\pgfsetdash{}{0pt}%
\pgfpathmoveto{\pgfqpoint{0.100000in}{1.328214in}}%
\pgfpathlineto{\pgfqpoint{4.750000in}{1.328214in}}%
\pgfusepath{stroke}%
\end{pgfscope}%
\begin{pgfscope}%
\pgfpathrectangle{\pgfqpoint{0.100000in}{0.915754in}}{\pgfqpoint{4.650000in}{3.020000in}}%
\pgfusepath{clip}%
\pgfsetroundcap%
\pgfsetroundjoin%
\pgfsetlinewidth{1.003750pt}%
\definecolor{currentstroke}{rgb}{0.800000,0.800000,0.800000}%
\pgfsetstrokecolor{currentstroke}%
\pgfsetdash{}{0pt}%
\pgfpathmoveto{\pgfqpoint{0.100000in}{2.698375in}}%
\pgfpathlineto{\pgfqpoint{4.750000in}{2.698375in}}%
\pgfusepath{stroke}%
\end{pgfscope}%
\begin{pgfscope}%
\pgfpathrectangle{\pgfqpoint{0.100000in}{0.915754in}}{\pgfqpoint{4.650000in}{3.020000in}}%
\pgfusepath{clip}%
\pgfsetbuttcap%
\pgfsetroundjoin%
\definecolor{currentfill}{rgb}{0.298039,0.447059,0.690196}%
\pgfsetfillcolor{currentfill}%
\pgfsetfillopacity{0.200000}%
\pgfsetlinewidth{1.003750pt}%
\definecolor{currentstroke}{rgb}{0.298039,0.447059,0.690196}%
\pgfsetstrokecolor{currentstroke}%
\pgfsetstrokeopacity{0.200000}%
\pgfsetdash{}{0pt}%
\pgfpathmoveto{\pgfqpoint{-0.026151in}{2.295013in}}%
\pgfpathlineto{\pgfqpoint{-0.026151in}{2.033061in}}%
\pgfpathlineto{\pgfqpoint{0.154098in}{2.008403in}}%
\pgfpathlineto{\pgfqpoint{0.330412in}{2.342184in}}%
\pgfpathlineto{\pgfqpoint{0.503375in}{2.319662in}}%
\pgfpathlineto{\pgfqpoint{0.677756in}{2.212711in}}%
\pgfpathlineto{\pgfqpoint{0.854951in}{2.146327in}}%
\pgfpathlineto{\pgfqpoint{1.032765in}{2.293581in}}%
\pgfpathlineto{\pgfqpoint{1.205706in}{2.158638in}}%
\pgfpathlineto{\pgfqpoint{1.383173in}{2.161539in}}%
\pgfpathlineto{\pgfqpoint{1.557256in}{2.093074in}}%
\pgfpathlineto{\pgfqpoint{1.737294in}{2.144372in}}%
\pgfpathlineto{\pgfqpoint{1.911281in}{2.118476in}}%
\pgfpathlineto{\pgfqpoint{2.089026in}{2.142453in}}%
\pgfpathlineto{\pgfqpoint{2.266038in}{2.046516in}}%
\pgfpathlineto{\pgfqpoint{2.442835in}{2.087954in}}%
\pgfpathlineto{\pgfqpoint{2.619209in}{2.075939in}}%
\pgfpathlineto{\pgfqpoint{2.795858in}{2.149884in}}%
\pgfpathlineto{\pgfqpoint{2.967521in}{2.106620in}}%
\pgfpathlineto{\pgfqpoint{3.147918in}{1.996990in}}%
\pgfpathlineto{\pgfqpoint{3.324794in}{2.120406in}}%
\pgfpathlineto{\pgfqpoint{3.505504in}{2.157225in}}%
\pgfpathlineto{\pgfqpoint{3.682021in}{2.066670in}}%
\pgfpathlineto{\pgfqpoint{3.863226in}{2.109635in}}%
\pgfpathlineto{\pgfqpoint{4.033288in}{2.191067in}}%
\pgfpathlineto{\pgfqpoint{4.215585in}{2.056664in}}%
\pgfpathlineto{\pgfqpoint{4.392704in}{2.053136in}}%
\pgfpathlineto{\pgfqpoint{4.569194in}{1.997600in}}%
\pgfpathlineto{\pgfqpoint{4.750665in}{2.061387in}}%
\pgfpathlineto{\pgfqpoint{4.924830in}{1.913290in}}%
\pgfpathlineto{\pgfqpoint{5.104652in}{2.088700in}}%
\pgfpathlineto{\pgfqpoint{5.104652in}{2.254278in}}%
\pgfpathlineto{\pgfqpoint{5.104652in}{2.254278in}}%
\pgfpathlineto{\pgfqpoint{4.924830in}{2.088195in}}%
\pgfpathlineto{\pgfqpoint{4.750665in}{2.233885in}}%
\pgfpathlineto{\pgfqpoint{4.569194in}{2.153293in}}%
\pgfpathlineto{\pgfqpoint{4.392704in}{2.216003in}}%
\pgfpathlineto{\pgfqpoint{4.215585in}{2.256838in}}%
\pgfpathlineto{\pgfqpoint{4.033288in}{2.309847in}}%
\pgfpathlineto{\pgfqpoint{3.863226in}{2.256264in}}%
\pgfpathlineto{\pgfqpoint{3.682021in}{2.219826in}}%
\pgfpathlineto{\pgfqpoint{3.505504in}{2.306586in}}%
\pgfpathlineto{\pgfqpoint{3.324794in}{2.266414in}}%
\pgfpathlineto{\pgfqpoint{3.147918in}{2.213887in}}%
\pgfpathlineto{\pgfqpoint{2.967521in}{2.299914in}}%
\pgfpathlineto{\pgfqpoint{2.795858in}{2.338190in}}%
\pgfpathlineto{\pgfqpoint{2.619209in}{2.257374in}}%
\pgfpathlineto{\pgfqpoint{2.442835in}{2.303211in}}%
\pgfpathlineto{\pgfqpoint{2.266038in}{2.234407in}}%
\pgfpathlineto{\pgfqpoint{2.089026in}{2.324895in}}%
\pgfpathlineto{\pgfqpoint{1.911281in}{2.305183in}}%
\pgfpathlineto{\pgfqpoint{1.737294in}{2.362406in}}%
\pgfpathlineto{\pgfqpoint{1.557256in}{2.273059in}}%
\pgfpathlineto{\pgfqpoint{1.383173in}{2.354016in}}%
\pgfpathlineto{\pgfqpoint{1.205706in}{2.377084in}}%
\pgfpathlineto{\pgfqpoint{1.032765in}{2.447991in}}%
\pgfpathlineto{\pgfqpoint{0.854951in}{2.389800in}}%
\pgfpathlineto{\pgfqpoint{0.677756in}{2.434140in}}%
\pgfpathlineto{\pgfqpoint{0.503375in}{2.524561in}}%
\pgfpathlineto{\pgfqpoint{0.330412in}{2.479975in}}%
\pgfpathlineto{\pgfqpoint{0.154098in}{2.272733in}}%
\pgfpathlineto{\pgfqpoint{-0.026151in}{2.295013in}}%
\pgfpathclose%
\pgfusepath{stroke,fill}%
\end{pgfscope}%
\begin{pgfscope}%
\pgfpathrectangle{\pgfqpoint{0.100000in}{0.915754in}}{\pgfqpoint{4.650000in}{3.020000in}}%
\pgfusepath{clip}%
\pgfsetbuttcap%
\pgfsetroundjoin%
\definecolor{currentfill}{rgb}{0.866667,0.517647,0.321569}%
\pgfsetfillcolor{currentfill}%
\pgfsetfillopacity{0.200000}%
\pgfsetlinewidth{1.003750pt}%
\definecolor{currentstroke}{rgb}{0.866667,0.517647,0.321569}%
\pgfsetstrokecolor{currentstroke}%
\pgfsetstrokeopacity{0.200000}%
\pgfsetdash{}{0pt}%
\pgfpathmoveto{\pgfqpoint{-0.022496in}{2.137220in}}%
\pgfpathlineto{\pgfqpoint{-0.022496in}{1.846361in}}%
\pgfpathlineto{\pgfqpoint{0.148268in}{1.729667in}}%
\pgfpathlineto{\pgfqpoint{0.327168in}{2.087144in}}%
\pgfpathlineto{\pgfqpoint{0.503417in}{1.952616in}}%
\pgfpathlineto{\pgfqpoint{0.680317in}{1.878257in}}%
\pgfpathlineto{\pgfqpoint{0.854169in}{1.818655in}}%
\pgfpathlineto{\pgfqpoint{1.028178in}{1.881233in}}%
\pgfpathlineto{\pgfqpoint{1.207774in}{1.841213in}}%
\pgfpathlineto{\pgfqpoint{1.382505in}{1.856864in}}%
\pgfpathlineto{\pgfqpoint{1.559014in}{1.915351in}}%
\pgfpathlineto{\pgfqpoint{1.734406in}{1.696530in}}%
\pgfpathlineto{\pgfqpoint{1.913397in}{1.885822in}}%
\pgfpathlineto{\pgfqpoint{2.088138in}{1.905462in}}%
\pgfpathlineto{\pgfqpoint{2.265881in}{1.872287in}}%
\pgfpathlineto{\pgfqpoint{2.444035in}{1.803659in}}%
\pgfpathlineto{\pgfqpoint{2.617618in}{1.910143in}}%
\pgfpathlineto{\pgfqpoint{2.793993in}{1.808210in}}%
\pgfpathlineto{\pgfqpoint{2.974909in}{1.963228in}}%
\pgfpathlineto{\pgfqpoint{3.152710in}{1.843468in}}%
\pgfpathlineto{\pgfqpoint{3.330953in}{1.890371in}}%
\pgfpathlineto{\pgfqpoint{3.505301in}{1.895283in}}%
\pgfpathlineto{\pgfqpoint{3.681956in}{1.650903in}}%
\pgfpathlineto{\pgfqpoint{3.860657in}{1.729351in}}%
\pgfpathlineto{\pgfqpoint{4.033430in}{1.829825in}}%
\pgfpathlineto{\pgfqpoint{4.209323in}{1.879997in}}%
\pgfpathlineto{\pgfqpoint{4.391473in}{1.811641in}}%
\pgfpathlineto{\pgfqpoint{4.566080in}{1.697445in}}%
\pgfpathlineto{\pgfqpoint{4.747491in}{1.864622in}}%
\pgfpathlineto{\pgfqpoint{4.926529in}{1.863732in}}%
\pgfpathlineto{\pgfqpoint{5.104838in}{1.795664in}}%
\pgfpathlineto{\pgfqpoint{5.104838in}{2.032034in}}%
\pgfpathlineto{\pgfqpoint{5.104838in}{2.032034in}}%
\pgfpathlineto{\pgfqpoint{4.926529in}{2.078682in}}%
\pgfpathlineto{\pgfqpoint{4.747491in}{2.104798in}}%
\pgfpathlineto{\pgfqpoint{4.566080in}{1.996030in}}%
\pgfpathlineto{\pgfqpoint{4.391473in}{2.080541in}}%
\pgfpathlineto{\pgfqpoint{4.209323in}{2.092560in}}%
\pgfpathlineto{\pgfqpoint{4.033430in}{2.071769in}}%
\pgfpathlineto{\pgfqpoint{3.860657in}{1.998244in}}%
\pgfpathlineto{\pgfqpoint{3.681956in}{1.902433in}}%
\pgfpathlineto{\pgfqpoint{3.505301in}{2.100648in}}%
\pgfpathlineto{\pgfqpoint{3.330953in}{2.116665in}}%
\pgfpathlineto{\pgfqpoint{3.152710in}{2.053008in}}%
\pgfpathlineto{\pgfqpoint{2.974909in}{2.189574in}}%
\pgfpathlineto{\pgfqpoint{2.793993in}{2.099598in}}%
\pgfpathlineto{\pgfqpoint{2.617618in}{2.139309in}}%
\pgfpathlineto{\pgfqpoint{2.444035in}{2.081417in}}%
\pgfpathlineto{\pgfqpoint{2.265881in}{2.129578in}}%
\pgfpathlineto{\pgfqpoint{2.088138in}{2.135205in}}%
\pgfpathlineto{\pgfqpoint{1.913397in}{2.125556in}}%
\pgfpathlineto{\pgfqpoint{1.734406in}{1.962958in}}%
\pgfpathlineto{\pgfqpoint{1.559014in}{2.099648in}}%
\pgfpathlineto{\pgfqpoint{1.382505in}{2.184892in}}%
\pgfpathlineto{\pgfqpoint{1.207774in}{2.068341in}}%
\pgfpathlineto{\pgfqpoint{1.028178in}{2.142612in}}%
\pgfpathlineto{\pgfqpoint{0.854169in}{2.087384in}}%
\pgfpathlineto{\pgfqpoint{0.680317in}{2.179749in}}%
\pgfpathlineto{\pgfqpoint{0.503417in}{2.230399in}}%
\pgfpathlineto{\pgfqpoint{0.327168in}{2.325898in}}%
\pgfpathlineto{\pgfqpoint{0.148268in}{2.051130in}}%
\pgfpathlineto{\pgfqpoint{-0.022496in}{2.137220in}}%
\pgfpathclose%
\pgfusepath{stroke,fill}%
\end{pgfscope}%
\begin{pgfscope}%
\pgfpathrectangle{\pgfqpoint{0.100000in}{0.915754in}}{\pgfqpoint{4.650000in}{3.020000in}}%
\pgfusepath{clip}%
\pgfsetbuttcap%
\pgfsetroundjoin%
\definecolor{currentfill}{rgb}{0.333333,0.658824,0.407843}%
\pgfsetfillcolor{currentfill}%
\pgfsetfillopacity{0.200000}%
\pgfsetlinewidth{1.003750pt}%
\definecolor{currentstroke}{rgb}{0.333333,0.658824,0.407843}%
\pgfsetstrokecolor{currentstroke}%
\pgfsetstrokeopacity{0.200000}%
\pgfsetdash{}{0pt}%
\pgfpathmoveto{\pgfqpoint{0.010696in}{2.964811in}}%
\pgfpathlineto{\pgfqpoint{0.010696in}{2.561095in}}%
\pgfpathlineto{\pgfqpoint{0.204636in}{2.511972in}}%
\pgfpathlineto{\pgfqpoint{0.373456in}{2.275481in}}%
\pgfpathlineto{\pgfqpoint{0.563217in}{2.156305in}}%
\pgfpathlineto{\pgfqpoint{0.754639in}{2.140593in}}%
\pgfpathlineto{\pgfqpoint{0.930587in}{1.961972in}}%
\pgfpathlineto{\pgfqpoint{1.126404in}{2.146384in}}%
\pgfpathlineto{\pgfqpoint{1.294782in}{2.076421in}}%
\pgfpathlineto{\pgfqpoint{1.473885in}{1.939497in}}%
\pgfpathlineto{\pgfqpoint{1.657099in}{2.063672in}}%
\pgfpathlineto{\pgfqpoint{1.837444in}{2.113831in}}%
\pgfpathlineto{\pgfqpoint{2.044262in}{1.947825in}}%
\pgfpathlineto{\pgfqpoint{2.204047in}{1.959077in}}%
\pgfpathlineto{\pgfqpoint{2.400004in}{1.760737in}}%
\pgfpathlineto{\pgfqpoint{2.596749in}{1.908514in}}%
\pgfpathlineto{\pgfqpoint{2.747124in}{1.856556in}}%
\pgfpathlineto{\pgfqpoint{2.937611in}{1.769607in}}%
\pgfpathlineto{\pgfqpoint{3.115926in}{1.912367in}}%
\pgfpathlineto{\pgfqpoint{3.317385in}{1.960563in}}%
\pgfpathlineto{\pgfqpoint{3.477786in}{1.924816in}}%
\pgfpathlineto{\pgfqpoint{3.656894in}{1.829478in}}%
\pgfpathlineto{\pgfqpoint{3.826815in}{1.867625in}}%
\pgfpathlineto{\pgfqpoint{4.017268in}{1.935308in}}%
\pgfpathlineto{\pgfqpoint{4.236592in}{1.853671in}}%
\pgfpathlineto{\pgfqpoint{4.382197in}{2.081531in}}%
\pgfpathlineto{\pgfqpoint{4.567584in}{1.975543in}}%
\pgfpathlineto{\pgfqpoint{4.759765in}{1.802910in}}%
\pgfpathlineto{\pgfqpoint{4.960713in}{1.969412in}}%
\pgfpathlineto{\pgfqpoint{5.166756in}{1.751392in}}%
\pgfpathlineto{\pgfqpoint{5.302516in}{1.846529in}}%
\pgfpathlineto{\pgfqpoint{5.302516in}{2.037933in}}%
\pgfpathlineto{\pgfqpoint{5.302516in}{2.037933in}}%
\pgfpathlineto{\pgfqpoint{5.166756in}{2.000064in}}%
\pgfpathlineto{\pgfqpoint{4.960713in}{2.144279in}}%
\pgfpathlineto{\pgfqpoint{4.759765in}{2.082607in}}%
\pgfpathlineto{\pgfqpoint{4.567584in}{2.183518in}}%
\pgfpathlineto{\pgfqpoint{4.382197in}{2.210778in}}%
\pgfpathlineto{\pgfqpoint{4.236592in}{2.051294in}}%
\pgfpathlineto{\pgfqpoint{4.017268in}{2.174691in}}%
\pgfpathlineto{\pgfqpoint{3.826815in}{2.113977in}}%
\pgfpathlineto{\pgfqpoint{3.656894in}{2.137722in}}%
\pgfpathlineto{\pgfqpoint{3.477786in}{2.157399in}}%
\pgfpathlineto{\pgfqpoint{3.317385in}{2.240662in}}%
\pgfpathlineto{\pgfqpoint{3.115926in}{2.183649in}}%
\pgfpathlineto{\pgfqpoint{2.937611in}{2.058214in}}%
\pgfpathlineto{\pgfqpoint{2.747124in}{2.194961in}}%
\pgfpathlineto{\pgfqpoint{2.596749in}{2.140174in}}%
\pgfpathlineto{\pgfqpoint{2.400004in}{2.100128in}}%
\pgfpathlineto{\pgfqpoint{2.204047in}{2.269843in}}%
\pgfpathlineto{\pgfqpoint{2.044262in}{2.225897in}}%
\pgfpathlineto{\pgfqpoint{1.837444in}{2.419407in}}%
\pgfpathlineto{\pgfqpoint{1.657099in}{2.356002in}}%
\pgfpathlineto{\pgfqpoint{1.473885in}{2.318063in}}%
\pgfpathlineto{\pgfqpoint{1.294782in}{2.798970in}}%
\pgfpathlineto{\pgfqpoint{1.126404in}{2.410227in}}%
\pgfpathlineto{\pgfqpoint{0.930587in}{2.294201in}}%
\pgfpathlineto{\pgfqpoint{0.754639in}{2.499294in}}%
\pgfpathlineto{\pgfqpoint{0.563217in}{2.475719in}}%
\pgfpathlineto{\pgfqpoint{0.373456in}{2.543381in}}%
\pgfpathlineto{\pgfqpoint{0.204636in}{2.955626in}}%
\pgfpathlineto{\pgfqpoint{0.010696in}{2.964811in}}%
\pgfpathclose%
\pgfusepath{stroke,fill}%
\end{pgfscope}%
\begin{pgfscope}%
\pgfpathrectangle{\pgfqpoint{0.100000in}{0.915754in}}{\pgfqpoint{4.650000in}{3.020000in}}%
\pgfusepath{clip}%
\pgfsetbuttcap%
\pgfsetroundjoin%
\definecolor{currentfill}{rgb}{0.768627,0.305882,0.321569}%
\pgfsetfillcolor{currentfill}%
\pgfsetfillopacity{0.200000}%
\pgfsetlinewidth{1.003750pt}%
\definecolor{currentstroke}{rgb}{0.768627,0.305882,0.321569}%
\pgfsetstrokecolor{currentstroke}%
\pgfsetstrokeopacity{0.200000}%
\pgfsetdash{}{0pt}%
\pgfpathmoveto{\pgfqpoint{-0.066071in}{3.487670in}}%
\pgfpathlineto{\pgfqpoint{-0.066071in}{3.101931in}}%
\pgfpathlineto{\pgfqpoint{0.105727in}{2.982939in}}%
\pgfpathlineto{\pgfqpoint{0.277525in}{2.976678in}}%
\pgfpathlineto{\pgfqpoint{0.449323in}{2.888983in}}%
\pgfpathlineto{\pgfqpoint{0.621121in}{2.846559in}}%
\pgfpathlineto{\pgfqpoint{0.792919in}{2.816569in}}%
\pgfpathlineto{\pgfqpoint{0.964717in}{2.668350in}}%
\pgfpathlineto{\pgfqpoint{1.136515in}{2.713805in}}%
\pgfpathlineto{\pgfqpoint{1.308313in}{2.635048in}}%
\pgfpathlineto{\pgfqpoint{1.480111in}{2.629715in}}%
\pgfpathlineto{\pgfqpoint{1.651909in}{2.604018in}}%
\pgfpathlineto{\pgfqpoint{1.823707in}{2.609343in}}%
\pgfpathlineto{\pgfqpoint{1.995505in}{2.555536in}}%
\pgfpathlineto{\pgfqpoint{2.167303in}{2.710736in}}%
\pgfpathlineto{\pgfqpoint{2.339101in}{2.646239in}}%
\pgfpathlineto{\pgfqpoint{2.510899in}{2.473121in}}%
\pgfpathlineto{\pgfqpoint{2.682697in}{2.460894in}}%
\pgfpathlineto{\pgfqpoint{2.854495in}{2.514806in}}%
\pgfpathlineto{\pgfqpoint{3.026293in}{2.483957in}}%
\pgfpathlineto{\pgfqpoint{3.198091in}{2.478905in}}%
\pgfpathlineto{\pgfqpoint{3.369889in}{2.468011in}}%
\pgfpathlineto{\pgfqpoint{3.541687in}{2.415595in}}%
\pgfpathlineto{\pgfqpoint{3.713485in}{2.482538in}}%
\pgfpathlineto{\pgfqpoint{3.885283in}{2.510446in}}%
\pgfpathlineto{\pgfqpoint{4.057081in}{2.437015in}}%
\pgfpathlineto{\pgfqpoint{4.228879in}{2.132575in}}%
\pgfpathlineto{\pgfqpoint{4.400677in}{2.425693in}}%
\pgfpathlineto{\pgfqpoint{4.572475in}{2.457495in}}%
\pgfpathlineto{\pgfqpoint{4.744273in}{2.415875in}}%
\pgfpathlineto{\pgfqpoint{4.916071in}{2.453161in}}%
\pgfpathlineto{\pgfqpoint{4.916071in}{2.735602in}}%
\pgfpathlineto{\pgfqpoint{4.916071in}{2.735602in}}%
\pgfpathlineto{\pgfqpoint{4.744273in}{2.682221in}}%
\pgfpathlineto{\pgfqpoint{4.572475in}{2.794736in}}%
\pgfpathlineto{\pgfqpoint{4.400677in}{2.749179in}}%
\pgfpathlineto{\pgfqpoint{4.228879in}{2.500087in}}%
\pgfpathlineto{\pgfqpoint{4.057081in}{2.795274in}}%
\pgfpathlineto{\pgfqpoint{3.885283in}{2.822963in}}%
\pgfpathlineto{\pgfqpoint{3.713485in}{2.765828in}}%
\pgfpathlineto{\pgfqpoint{3.541687in}{2.799374in}}%
\pgfpathlineto{\pgfqpoint{3.369889in}{2.810385in}}%
\pgfpathlineto{\pgfqpoint{3.198091in}{2.847777in}}%
\pgfpathlineto{\pgfqpoint{3.026293in}{2.827971in}}%
\pgfpathlineto{\pgfqpoint{2.854495in}{2.809486in}}%
\pgfpathlineto{\pgfqpoint{2.682697in}{2.912949in}}%
\pgfpathlineto{\pgfqpoint{2.510899in}{2.747989in}}%
\pgfpathlineto{\pgfqpoint{2.339101in}{2.923488in}}%
\pgfpathlineto{\pgfqpoint{2.167303in}{3.045467in}}%
\pgfpathlineto{\pgfqpoint{1.995505in}{2.847424in}}%
\pgfpathlineto{\pgfqpoint{1.823707in}{2.926034in}}%
\pgfpathlineto{\pgfqpoint{1.651909in}{2.933309in}}%
\pgfpathlineto{\pgfqpoint{1.480111in}{2.917393in}}%
\pgfpathlineto{\pgfqpoint{1.308313in}{2.950629in}}%
\pgfpathlineto{\pgfqpoint{1.136515in}{3.009294in}}%
\pgfpathlineto{\pgfqpoint{0.964717in}{2.916985in}}%
\pgfpathlineto{\pgfqpoint{0.792919in}{3.118797in}}%
\pgfpathlineto{\pgfqpoint{0.621121in}{3.115442in}}%
\pgfpathlineto{\pgfqpoint{0.449323in}{3.226137in}}%
\pgfpathlineto{\pgfqpoint{0.277525in}{3.309069in}}%
\pgfpathlineto{\pgfqpoint{0.105727in}{3.299186in}}%
\pgfpathlineto{\pgfqpoint{-0.066071in}{3.487670in}}%
\pgfpathclose%
\pgfusepath{stroke,fill}%
\end{pgfscope}%
\begin{pgfscope}%
\pgfpathrectangle{\pgfqpoint{0.100000in}{0.915754in}}{\pgfqpoint{4.650000in}{3.020000in}}%
\pgfusepath{clip}%
\pgfsetroundcap%
\pgfsetroundjoin%
\pgfsetlinewidth{1.505625pt}%
\definecolor{currentstroke}{rgb}{0.298039,0.447059,0.690196}%
\pgfsetstrokecolor{currentstroke}%
\pgfsetdash{}{0pt}%
\pgfpathmoveto{\pgfqpoint{0.086111in}{2.160896in}}%
\pgfpathlineto{\pgfqpoint{0.154098in}{2.152237in}}%
\pgfpathlineto{\pgfqpoint{0.330412in}{2.417022in}}%
\pgfpathlineto{\pgfqpoint{0.503375in}{2.429253in}}%
\pgfpathlineto{\pgfqpoint{0.677756in}{2.331077in}}%
\pgfpathlineto{\pgfqpoint{0.854951in}{2.277385in}}%
\pgfpathlineto{\pgfqpoint{1.032765in}{2.377399in}}%
\pgfpathlineto{\pgfqpoint{1.205706in}{2.282741in}}%
\pgfpathlineto{\pgfqpoint{1.383173in}{2.264529in}}%
\pgfpathlineto{\pgfqpoint{1.557256in}{2.186918in}}%
\pgfpathlineto{\pgfqpoint{1.737294in}{2.256758in}}%
\pgfpathlineto{\pgfqpoint{1.911281in}{2.222628in}}%
\pgfpathlineto{\pgfqpoint{2.089026in}{2.234721in}}%
\pgfpathlineto{\pgfqpoint{2.266038in}{2.146371in}}%
\pgfpathlineto{\pgfqpoint{2.442835in}{2.208796in}}%
\pgfpathlineto{\pgfqpoint{2.619209in}{2.176526in}}%
\pgfpathlineto{\pgfqpoint{2.795858in}{2.250760in}}%
\pgfpathlineto{\pgfqpoint{2.967521in}{2.206507in}}%
\pgfpathlineto{\pgfqpoint{3.147918in}{2.112197in}}%
\pgfpathlineto{\pgfqpoint{3.324794in}{2.197077in}}%
\pgfpathlineto{\pgfqpoint{3.505504in}{2.236921in}}%
\pgfpathlineto{\pgfqpoint{3.682021in}{2.149759in}}%
\pgfpathlineto{\pgfqpoint{3.863226in}{2.187122in}}%
\pgfpathlineto{\pgfqpoint{4.033288in}{2.252323in}}%
\pgfpathlineto{\pgfqpoint{4.215585in}{2.170313in}}%
\pgfpathlineto{\pgfqpoint{4.392704in}{2.143416in}}%
\pgfpathlineto{\pgfqpoint{4.569194in}{2.080825in}}%
\pgfpathlineto{\pgfqpoint{4.750665in}{2.153124in}}%
\pgfpathlineto{\pgfqpoint{4.763889in}{2.141657in}}%
\pgfusepath{stroke}%
\end{pgfscope}%
\begin{pgfscope}%
\pgfpathrectangle{\pgfqpoint{0.100000in}{0.915754in}}{\pgfqpoint{4.650000in}{3.020000in}}%
\pgfusepath{clip}%
\pgfsetroundcap%
\pgfsetroundjoin%
\pgfsetlinewidth{1.505625pt}%
\definecolor{currentstroke}{rgb}{0.866667,0.517647,0.321569}%
\pgfsetstrokecolor{currentstroke}%
\pgfsetdash{}{0pt}%
\pgfpathmoveto{\pgfqpoint{0.086111in}{1.941542in}}%
\pgfpathlineto{\pgfqpoint{0.148268in}{1.905980in}}%
\pgfpathlineto{\pgfqpoint{0.327168in}{2.207387in}}%
\pgfpathlineto{\pgfqpoint{0.503417in}{2.109071in}}%
\pgfpathlineto{\pgfqpoint{0.680317in}{2.043222in}}%
\pgfpathlineto{\pgfqpoint{0.854169in}{1.971563in}}%
\pgfpathlineto{\pgfqpoint{1.028178in}{2.022190in}}%
\pgfpathlineto{\pgfqpoint{1.207774in}{1.965208in}}%
\pgfpathlineto{\pgfqpoint{1.382505in}{2.031995in}}%
\pgfpathlineto{\pgfqpoint{1.559014in}{2.013322in}}%
\pgfpathlineto{\pgfqpoint{1.734406in}{1.840329in}}%
\pgfpathlineto{\pgfqpoint{1.913397in}{2.021235in}}%
\pgfpathlineto{\pgfqpoint{2.088138in}{2.032369in}}%
\pgfpathlineto{\pgfqpoint{2.265881in}{2.016847in}}%
\pgfpathlineto{\pgfqpoint{2.444035in}{1.959138in}}%
\pgfpathlineto{\pgfqpoint{2.617618in}{2.035623in}}%
\pgfpathlineto{\pgfqpoint{2.793993in}{1.950570in}}%
\pgfpathlineto{\pgfqpoint{2.974909in}{2.083762in}}%
\pgfpathlineto{\pgfqpoint{3.152710in}{1.958883in}}%
\pgfpathlineto{\pgfqpoint{3.330953in}{2.012165in}}%
\pgfpathlineto{\pgfqpoint{3.505301in}{2.008193in}}%
\pgfpathlineto{\pgfqpoint{3.681956in}{1.790425in}}%
\pgfpathlineto{\pgfqpoint{3.860657in}{1.881655in}}%
\pgfpathlineto{\pgfqpoint{4.033430in}{1.954071in}}%
\pgfpathlineto{\pgfqpoint{4.209323in}{1.995397in}}%
\pgfpathlineto{\pgfqpoint{4.391473in}{1.959293in}}%
\pgfpathlineto{\pgfqpoint{4.566080in}{1.857054in}}%
\pgfpathlineto{\pgfqpoint{4.747491in}{1.989322in}}%
\pgfpathlineto{\pgfqpoint{4.763889in}{1.988592in}}%
\pgfusepath{stroke}%
\end{pgfscope}%
\begin{pgfscope}%
\pgfpathrectangle{\pgfqpoint{0.100000in}{0.915754in}}{\pgfqpoint{4.650000in}{3.020000in}}%
\pgfusepath{clip}%
\pgfsetroundcap%
\pgfsetroundjoin%
\pgfsetlinewidth{1.505625pt}%
\definecolor{currentstroke}{rgb}{0.333333,0.658824,0.407843}%
\pgfsetstrokecolor{currentstroke}%
\pgfsetdash{}{0pt}%
\pgfpathmoveto{\pgfqpoint{0.086111in}{2.771585in}}%
\pgfpathlineto{\pgfqpoint{0.204636in}{2.760285in}}%
\pgfpathlineto{\pgfqpoint{0.373456in}{2.421201in}}%
\pgfpathlineto{\pgfqpoint{0.563217in}{2.327300in}}%
\pgfpathlineto{\pgfqpoint{0.754639in}{2.340274in}}%
\pgfpathlineto{\pgfqpoint{0.930587in}{2.145082in}}%
\pgfpathlineto{\pgfqpoint{1.126404in}{2.281109in}}%
\pgfpathlineto{\pgfqpoint{1.294782in}{2.460808in}}%
\pgfpathlineto{\pgfqpoint{1.473885in}{2.155557in}}%
\pgfpathlineto{\pgfqpoint{1.657099in}{2.214940in}}%
\pgfpathlineto{\pgfqpoint{1.837444in}{2.288626in}}%
\pgfpathlineto{\pgfqpoint{2.044262in}{2.101676in}}%
\pgfpathlineto{\pgfqpoint{2.204047in}{2.130171in}}%
\pgfpathlineto{\pgfqpoint{2.400004in}{1.948683in}}%
\pgfpathlineto{\pgfqpoint{2.596749in}{2.038426in}}%
\pgfpathlineto{\pgfqpoint{2.747124in}{2.027724in}}%
\pgfpathlineto{\pgfqpoint{2.937611in}{1.922766in}}%
\pgfpathlineto{\pgfqpoint{3.115926in}{2.058469in}}%
\pgfpathlineto{\pgfqpoint{3.317385in}{2.106494in}}%
\pgfpathlineto{\pgfqpoint{3.477786in}{2.050014in}}%
\pgfpathlineto{\pgfqpoint{3.656894in}{1.992730in}}%
\pgfpathlineto{\pgfqpoint{3.826815in}{2.000034in}}%
\pgfpathlineto{\pgfqpoint{4.017268in}{2.064700in}}%
\pgfpathlineto{\pgfqpoint{4.236592in}{1.956132in}}%
\pgfpathlineto{\pgfqpoint{4.382197in}{2.150339in}}%
\pgfpathlineto{\pgfqpoint{4.567584in}{2.088531in}}%
\pgfpathlineto{\pgfqpoint{4.759765in}{1.951597in}}%
\pgfpathlineto{\pgfqpoint{4.763889in}{1.953902in}}%
\pgfusepath{stroke}%
\end{pgfscope}%
\begin{pgfscope}%
\pgfpathrectangle{\pgfqpoint{0.100000in}{0.915754in}}{\pgfqpoint{4.650000in}{3.020000in}}%
\pgfusepath{clip}%
\pgfsetroundcap%
\pgfsetroundjoin%
\pgfsetlinewidth{1.505625pt}%
\definecolor{currentstroke}{rgb}{0.768627,0.305882,0.321569}%
\pgfsetstrokecolor{currentstroke}%
\pgfsetdash{}{0pt}%
\pgfpathmoveto{\pgfqpoint{0.086111in}{3.173931in}}%
\pgfpathlineto{\pgfqpoint{0.105727in}{3.155540in}}%
\pgfpathlineto{\pgfqpoint{0.277525in}{3.162299in}}%
\pgfpathlineto{\pgfqpoint{0.449323in}{3.066176in}}%
\pgfpathlineto{\pgfqpoint{0.621121in}{2.991887in}}%
\pgfpathlineto{\pgfqpoint{0.792919in}{2.987292in}}%
\pgfpathlineto{\pgfqpoint{0.964717in}{2.799690in}}%
\pgfpathlineto{\pgfqpoint{1.136515in}{2.870504in}}%
\pgfpathlineto{\pgfqpoint{1.308313in}{2.800693in}}%
\pgfpathlineto{\pgfqpoint{1.480111in}{2.786611in}}%
\pgfpathlineto{\pgfqpoint{1.651909in}{2.789060in}}%
\pgfpathlineto{\pgfqpoint{1.823707in}{2.769066in}}%
\pgfpathlineto{\pgfqpoint{1.995505in}{2.715929in}}%
\pgfpathlineto{\pgfqpoint{2.167303in}{2.894470in}}%
\pgfpathlineto{\pgfqpoint{2.339101in}{2.796099in}}%
\pgfpathlineto{\pgfqpoint{2.510899in}{2.624984in}}%
\pgfpathlineto{\pgfqpoint{2.682697in}{2.711026in}}%
\pgfpathlineto{\pgfqpoint{2.854495in}{2.679344in}}%
\pgfpathlineto{\pgfqpoint{3.026293in}{2.679985in}}%
\pgfpathlineto{\pgfqpoint{3.198091in}{2.672638in}}%
\pgfpathlineto{\pgfqpoint{3.369889in}{2.652246in}}%
\pgfpathlineto{\pgfqpoint{3.541687in}{2.626626in}}%
\pgfpathlineto{\pgfqpoint{3.713485in}{2.638216in}}%
\pgfpathlineto{\pgfqpoint{3.885283in}{2.677361in}}%
\pgfpathlineto{\pgfqpoint{4.057081in}{2.638104in}}%
\pgfpathlineto{\pgfqpoint{4.228879in}{2.328911in}}%
\pgfpathlineto{\pgfqpoint{4.400677in}{2.604814in}}%
\pgfpathlineto{\pgfqpoint{4.572475in}{2.640611in}}%
\pgfpathlineto{\pgfqpoint{4.744273in}{2.567935in}}%
\pgfpathlineto{\pgfqpoint{4.763889in}{2.572758in}}%
\pgfusepath{stroke}%
\end{pgfscope}%
\begin{pgfscope}%
\pgfsetrectcap%
\pgfsetmiterjoin%
\pgfsetlinewidth{1.254687pt}%
\definecolor{currentstroke}{rgb}{0.800000,0.800000,0.800000}%
\pgfsetstrokecolor{currentstroke}%
\pgfsetdash{}{0pt}%
\pgfpathmoveto{\pgfqpoint{0.100000in}{0.915754in}}%
\pgfpathlineto{\pgfqpoint{0.100000in}{3.935754in}}%
\pgfusepath{stroke}%
\end{pgfscope}%
\begin{pgfscope}%
\pgfsetrectcap%
\pgfsetmiterjoin%
\pgfsetlinewidth{1.254687pt}%
\definecolor{currentstroke}{rgb}{0.800000,0.800000,0.800000}%
\pgfsetstrokecolor{currentstroke}%
\pgfsetdash{}{0pt}%
\pgfpathmoveto{\pgfqpoint{4.750000in}{0.915754in}}%
\pgfpathlineto{\pgfqpoint{4.750000in}{3.935754in}}%
\pgfusepath{stroke}%
\end{pgfscope}%
\begin{pgfscope}%
\pgfsetrectcap%
\pgfsetmiterjoin%
\pgfsetlinewidth{1.254687pt}%
\definecolor{currentstroke}{rgb}{0.800000,0.800000,0.800000}%
\pgfsetstrokecolor{currentstroke}%
\pgfsetdash{}{0pt}%
\pgfpathmoveto{\pgfqpoint{0.100000in}{0.915754in}}%
\pgfpathlineto{\pgfqpoint{4.750000in}{0.915754in}}%
\pgfusepath{stroke}%
\end{pgfscope}%
\begin{pgfscope}%
\pgfsetrectcap%
\pgfsetmiterjoin%
\pgfsetlinewidth{1.254687pt}%
\definecolor{currentstroke}{rgb}{0.800000,0.800000,0.800000}%
\pgfsetstrokecolor{currentstroke}%
\pgfsetdash{}{0pt}%
\pgfpathmoveto{\pgfqpoint{0.100000in}{3.935754in}}%
\pgfpathlineto{\pgfqpoint{4.750000in}{3.935754in}}%
\pgfusepath{stroke}%
\end{pgfscope}%
\end{pgfpicture}%
\makeatother%
\endgroup%

%% file: sections/06_discussion.tex
\section{Conclusion and Discussion}
\label{sec:discussion}
Riemannian statistics is the appropriate framework to model real data with nonlinear geometry.
Yet, its wide adoption is hampered by the prohibitive cost of numerical computations required to learn geometry from data and operate on manifolds.
In this work, we have demonstrated on the example of numerical integration the great potential of probabilistic numerical methods (\textsc{pnm}) to reduce this computational burden.
\textsc{pnm} adaptively select actions in a decision-theoretic manner and thus handle information with greater care than classic methods, e.g.,~Monte Carlo. 
Consequently, the deliberate choice of informative computations saves unnecessary operations on the manifold.
% \alex{All of the above sounds a bit like it could be the abstract. Maybe we should build up the paragraph the other way round, starting from what we did (think in a journalist manner: Information should be ordered by decreasing importance so the reader can stop anytime and still get the maximum of content given the amount of text read.)}
% \chris{I actually quite like it. perhaps we can get some feedback on it today. I think this first paragraph is a good summary of our mesage.}

We have extended Bayesian quadrature to Riemannian manifolds, where it outperforms Monte Carlo over a large number of integration problems owing to its increased sample efficiency.
Beyond known active learning schemes for \bq, we have introduced a version of uncertainty sampling adapted to the manifold setting that allows to further reduce the number of expensive geodesic evaluations needed to estimate the integral. 

Numerical integration is just one of multiple numerical tasks in the context of statistics on Riemannian manifolds where \textsc{pnm} suggest promising improvements.
The key operations on data manifolds are geodesic computations, i.e.,~solutions of ordinary differential equations. Geodesics have been viewed through the \textsc{pn} lens, e.g.,~by \citet{hennig:aistats:2014}, but still offer a margin for increasing the performance of statistical models such as the considered \textsc{land}.

% The combined effort of different \textsc{pnm} serves to illustrate \alex{it sounds as if we showed anything along these lines} the paradigm of \textit{uncertainty propagation} within the use-case of Riemannian statistics.
Once multiple \textsc{pnm}~are established for Riemannian statistics, the future avenue directs towards having them operate in a concerted fashion. As data-driven Riemannian models rely on complex computation pipelines with multiple sources of epistemic and aleatory uncertainty, their robustness and efficiency can benefit from modeling and propagating uncertainty through the computations. 
% As a concrete example, uncertainty arising in the integration process could be used to inform the optimizer, as the gradients for the mean and covariance depend on the normalization constant.

% Several other probabilistic numerical methods can be used to tackle these additional issues. For instance, uncertainty arising in the integration process could be used to inform the optimization. We currently optimize the covariance matrix over a symmetric positive definite manifold~(\ref{app:land}). The robustness of the optimizer might be improved from uncertainty propagation to its linesearch subroutine. To this end, we plan to adapt \textit{probabilistic line search} \citep{mahsereci2017} to the manifold setting.
% Robustness of this algorithm could be improved by explicitly modeling the uncertainty arising from quadrature in the line search. 
%  Since covariance matrices are constrained to be symmetric positive definite (SPD), we employ a manifold optimization scheme on the manifold of SPD matrices~(\ref{app:land}). This uses a manifold line search as a subroutine, which makes a deterministic decision about whether to shrink or expand the search space.

All in all, we believe the coalition of \textit{geometry-} and \textit{uncertainty}-aware methods to be a fruitful endeavor, as these approaches are united by their common intention to respect structure in data and computation that is otherwise often neglected.

%% file: sections/07_supplementary.tex
\onecolumn

\renewcommand{\thesection}{A.\arabic{section}}
\renewcommand{\thesubsection}{\thesection.\arabic{subsection}}

\icmltitle{Supplementary Material\\ \large{Bayesian Quadrature on Riemannian Data Manifolds}}

\thispagestyle{empty}

%%%%%%%%%%%%%%%%%%%%%%%%%%%%%%%%%%%%%%%%%%
\section{Riemannian Geometry}
%%%%%%%%%%%%%%%%%%%%%%%%%%%%%%%%%%%%%%%%%%
\label{app:riemann}
A \textit{manifold} $\mathcal{M}$ of dimension $D$ is a topological space which does not carry a global vector space structure. As opposed to the familiar $\mathbb{R}^D$, a manifold lacks the possibility of adding or scaling vectors globally. Instead, an \textit{atlas} it used to cover the manifold in \textit{charts}, which only locally give a Euclidean view of the manifold. If transition maps between overlapping charts are smooth, we call $\mathcal{M}$ a \textit{smooth manifold}, which provides the means for doing calculus. Our analysis is heavily simplified by viewing $\R^D$ as a manifold and employing the identity map as a global chart map, which covers the whole $\R^D$, thereby endowing the manifold automatically with the smoothness property. We use global Euclidean coordinates, which implies that we can solve the geodesic equations directly in this global chart.
 
\subsection{Geodesic Equations}
The energy or action functional of a curve $\bmgamma$ with time derivative $\dot{\bmgamma}(t)$ is defined as 
\begin{equation}
    E(\bmgamma) = \frac{1}{2} \int_{0}^1 \underbrace{\langle \dot{\bmgamma}(t), \bm{M}(\bmgamma(t)) \dot{\bmgamma}(t)  \rangle}_{\eqqcolon \mathscr{L}} \d t.
\end{equation}

In physics, the argument of the integral is known as \emph{Lagrangian} and we therefore abbreviate the inner product as $\mathscr{L}~\coloneqq~\langle \dot{\bmgamma}(t), \bm{M}(\bmgamma(t)) \dot{\bmgamma}(t)  \rangle$. Geodesics are the stationary curves of this functional. We are interested in the minimizers, i.e., shortest paths. Minimizing curve energy instead of length avoids the issue of arbitrary reparameterization. Let $\gamma^i$ denote the $i$-th coordinate of the curve $\bmgamma$ at time $t$ and $M_{ik}$ the metric component at row $i$ and column $k$, if it is represented as a matrix. We leave sums over repeated indices implicit (Einstein summation convention). Applying the Euler-Lagrange equations to the functional $E$ results in a system of equations involving $\mathscr{L}$
\begin{equation}
    \frac{\partial \mathscr{L}}{\partial \gamma^k} = \frac{\partial}{\partial t} \frac{\partial \mathscr{L}}{\partial \dot{\gamma}^k}, \quad \text{for } k \in 1,\dots, D.
\end{equation}
which is a system of $2^{nd}$ order differential equations. We first consider the left-hand side 
\begin{equation}
    \RN{1} \coloneqq \frac{\partial \mathscr{L}}{\partial \gamma^k} = \frac{1}{2} \frac{\partial M_{ij}}{\partial \gamma^k} \dot{\gamma}^i \dot{\gamma}^j,
\end{equation}
which holds due to independence of the coordinates. The right-hand side is
\begin{equation}
    \RN{2} \coloneqq \frac{\partial}{\partial t} \left[M_{ik} \dot{\gamma}^i \right] = \frac{\partial M_{ik}}{\partial \gamma^j} \dot{\gamma}^i \dot{\gamma}^j + M_{ik} \ddot{\gamma}^i.
\end{equation}
We expand this using a small index rearrangement trick
\begin{equation}
    \RN{2} = \frac{1}{2} \frac{\partial M_{ik}}{\partial \gamma^j} \dot{\gamma}^i \dot{\gamma}^j + \frac{1}{2} \frac{\partial M_{jk}}{\partial \gamma^i} \dot{\gamma}^i \dot{\gamma}^j + M_{ik} \ddot{\gamma}^i.
\end{equation}
This allows us to write $\RN{1} = \RN{2} \Leftrightarrow \RN{2} - \RN{1} = 0$ as
\begin{equation}
    M_{ik} \ddot{\gamma}^i + \frac{1}{2} \left(\frac{\partial M_{ik}}{\partial \gamma^j} + \frac{\partial M_{jk}}{\partial \gamma^i} - \frac{\partial M_{ij}}{\partial \gamma^k}  \right) \dot{\gamma}^j  = 0.
\end{equation}
the next step is to left multiply with the inverse metric tensor and plug in the \textit{Christoffel symbols} defined as follows
\begin{equation}
    \Gamma_{ij}^k = \frac{1}{2} M^{-1}_{kh} \left(\frac{\partial M_{ih}}{\partial \gamma^j} + \frac{\partial M_{jh}}{\partial \gamma^i} - \frac{\partial M_{ij}}{\partial \gamma^h}  \right),
\end{equation}
so we finally obtain the geodesic equations in the canonical form
\begin{equation}
    \ddot{\gamma}^k + \Gamma_{ij}^k  \dot{\gamma}^j \dot{\gamma}^j = 0, \quad \text{for } k \in 1,\dots, D.
\end{equation}
% \paragraph{Riemannian Normal Coordinates}
We assume our manifold to be \textit{geodesically complete} \citep{pennec2006}, which means that geodesics can be infinitely extended, i.e., their domain is $\R$. As a consequence, the exponential map is then defined on the whole tangent space. In theory, the exponential map $\expmap_{\bm{\mu}}(\cdot)$ is a \textit{diffeomorphism} only in some open neighborhood around $\bm{\mu}$ and thus it only admits a smooth inverse, i.e., $\logmap_{\bm{\mu}}(\cdot)$, in said neighborhood. However, we assume this to be true on the whole manifold in practice to keep the analysis tractable. For long geodesics on high-curvature data manifolds, often $\logmap_{\bm{\mu}}(\expmap_{\bm{\mu}}(\bm{v})) \neq \bm{v}$. This is rather unproblematic since if $\|\logmap_{\bm{\mu}_l}(\bmx_n)\|$ is high, the responsibility $r_{nl}$ will be low (see~\ref{app:land}), so this logarithmic map will play a minimal role in the Mahalanobis distance of the \textsc{land} density. Thus, the optimization process on its own favors mean and covariances such that the density is concentrated in sufficiently small neighborhoods where the exponential map approximately admits an inverse.

%\paragraph{Covariance and Precision Matrices}
\subsection{Covariance and Precision Matrices}
We here elaborate on Footnote 1 of the paper. The Riemannian normal distribution \citep{pennec2006} is defined using the precision matrix $\bm{\Gamma}$. This matrix lives on the tangent space $\mathcal{T}_{\bm{\mu}}\mathcal{M}$, i.e., it may be represented as a matrix in $\R^{D \times D}$, where $D$ is the dimension of the tangent space, which is equal to the topological dimension of the manifold. In our applied setting, $D$ matches the dimension of the data space, as we view the whole $\R^D$ as the manifold. We can use the tangent space ``covariance'' matrix $\bm{\Sigma} = \bm{\Gamma}^{-1}$ for our reasoning and the optimization process. However, to obtain the true covariance on the manifold $\mathcal{M}$, a subtle correction is necessary \citep{pennec2006}
\begin{equation}
    \bm{\Sigma}_{\mathcal{M}} = \mathbb{E}\left[\logmap_{\bm{\mu}}(\bm{x}) \logmap_{\bm{\mu}}(\bm{x})^\intercal\right] = \frac{1}{\mathcal{C}} \int_{\mathcal{M}} \logmap_{\bm{\mu}}(\bm{x}) \logmap_{\bm{\mu}}(\bm{x})^\intercal \exp\left( - \frac{1}{2} \left\langle \logmap_{\bm{\mu}}(\bm{x}), \bm{\Gamma} \logmap_{\bm{\mu}}(\bm{x})\right\rangle \right) \d \mathcal{M}(\bm{x}),
\end{equation}
with respect to the density on the manifold. For conceptual ease, we focus on the tangent space view in the paper. To plot the eigenvectors of the \textsc{adk} \textsc{land} covariance (Fig.~\ref{fig:adk_eigvecs}), we used the exponential map on the tangent space covariance matrix, i.e., we evaluate and plot $\expmap_{\bm{\mu}}(\bm{v}_{1:2})$, where $\bm{v}_{1:2}$ are the eigenvectors of $\bm{\Sigma}$.

%\paragraph{Geodesic Solvers}
\subsection{Geodesic Solvers}
To solve the geodesic equations, we combine two solvers, which have different strengths and weaknesses. By chaining them together, we obtain a more robust computational pipeline.

First, we make use of the fast and robust fixed-point solver (\textsc{fp}) introduced by \citet{arvanitidis2019fast}. This solver pursues a \textsc{gp}-based approach that avoids the often ill-behaved Jacobians of the geodesic \textsc{ode} system. However, the resulting logarithmic maps are subject to significant approximation error, depending on the curvature of the manifold. The parameters of this solver are as follows:
\begin{center}
\begin{tabular}{lll}
	\toprule
	Parameter & Value & Description \\ 
	\midrule
	$\mathtt{iter_{max}}$ & $1000$ & maximum number of iterations \\ 
	$\mathtt{N}$ & $10$ & number of mesh nodes. \\ 
	$\mathtt{tol}$ & $0.1$ & tolerance used to evaluate solution correctness. \\ 
	$\mathtt{\sigma}$ & $10^{-4}$ & noise of the \textsc{gp}. \\ 
	\bottomrule
\end{tabular} 
\end{center}
For \textsc{mnist}, we set $\mathtt{iter_{max}}=500$, and $\mathtt{tol}=0.2$, since this high-curvature manifold easily leads to failing geodesics.

The second solver we employ is a precise, albeit less robust one. This is the \textsc{bvp} solver available in the module \texttt{scipy.integrate.solve\_bvp}. On high-curvature manifolds, this solver often fails (especially for long curves) and takes a significant amount of time to run. When it succeeds, however, the logarithmic maps are reliable. For this solver, we set the maximum number of mesh nodes to $100$ and the tolerance to $0.1$. We empirically found that choosing a high maximum number of mesh nodes (e.g., $500$) can lead to high runtimes for failing geodesic computations.

To obtain fast and robust geodesics, these solvers may be chained together, i.e., we initialize the \textsc{bvp} solver with the \textsc{fp} solution, which is often worth the extra effort for speedup and improved robustness. For initialization, we use $20$ mesh nodes, evenly spaced on the \textsc{fp} solution. If the \textsc{fp} solver already failed, it is very unlikely for the \textsc{bvp} solver to succeed, so we abort the computation. 

Furthermore, we exploit previously computed \textsc{bvp} solutions: assume we want to compute $\logmap_{\bm{\mu}_t}(\bm{x})$. We search for past results $\logmap_{\bm{\mu}_t^*}(\bm{x})$, with $t^*<t$, $t^* = \argmin \|\bm{\mu}_t - \bm{\mu}_{t*}\|$ and $\|\bm{\mu}_t - \bm{\mu}_{t*}\| < \epsilon_d$, where we choose $\epsilon_d = 0.5$. Since we compute logarithmic maps for data points $\bm{x}_{1:N}$, which do not change during \textsc{land} optimization, we can use them as hash keys in a dictionary, where we store the solutions. Looking up the solution is then linear in the number of previous \textsc{land} iterations. If such a solution is found, the \textsc{fp} is skipped and the solution is used to directly initialize the \textsc{bvp} solver.

For the exponential maps, we use \texttt{scipy.integrate.solve\_ivp} with a tolerance of $10^{-3}$.

%%%%%%%%%%%%%%%%%%%%%%%%%%%%%%%%%%%%%%%%%%
\section{The \textsc{land} Objective and Gradients}
\label{app:land}
%%%%%%%%%%%%%%%%%%%%%%%%%%%%%%%%%%%%%%%%%%
Given a dataset $\bm{x}_{1:N}$ assumed to be i.i.d., the negative log-likelihood of the \textit{Locally Adaptive Normal Distribution} (\textsc{LAND}) mixture can be stated as \citep{arvanitidis2016land}

\begin{equation}
    \mathcal{L}\left(\{\bm{\mu}_k,\bm{\Sigma}_k\}_{1:K}\right) = \sum_{k=1}^{K} \sum_{n=1}^{N} r_{nk} \left[
    \frac{1}{2} \langle \logmap_{\bm{\mu}_k}(\bm{x}_n), \bm{\Sigma}_k^{-1} \logmap_{\bm{\mu_k}}(\bm{x}_n) \rangle + \log\left(\mathcal{C}(\bm{\mu}_k,\bm{\Sigma}_k)\right) - \log(\pi_k)
    \right]
\end{equation}

where $\pi_k$ is the weight of the $k^{\text{th}}$ component, $\sum_{k=1}^{K} \pi_k = 1$ and $r_{nk} = \frac{\pi_k p(x_n \mid \bm{\mu}_k, \bm{\Sigma}_k)}{\sum_{l=1}^{K} \pi_l p(x_n \mid \bm{\mu}_l, \bm{\Sigma}_l)}$ is the responsibility of the $k^{\text{th}}$ component for the $n^{\text{th}}$ datum. The maximum likelihood solution can be obtained by non-convex optimization, alternating between gradient descent updates of $\bm{\mu}$ and $\bm{\Sigma}$ and cycling through the components $k$, as described in Alg.~\ref{alg:land}. 

% LAND MIXTURE PSEUDOCODE
\begin{algorithm}[t]
\caption{\textsc{land} mixture main loop}
\label{alg:land}
\begin{algorithmic}
   \STATE {\bfseries Input:} data $\bm{x}_{1:N}$, manifold $\mathcal{M}$ with $\expmap$ and $\logmap$ operators, max. number of iterations $t_{max}$, \\ initial stepsize $\alpha_{\bm{\mu}}^1 \in \R$, gradient tolerance $\epsilon_{\nabla_{\bm{\mu}}}$, likelihood tolerance $\epsilon_{\mathcal{L}}$ 
   \STATE {\bfseries Output:}
   estimates $\left(\bm{\mu}_k,\bm{\Sigma}_k,\mathcal{C}_k,\pi_k \right)_{1:K}$
   \STATE Initialize \textsc{land} parameters$\left(\bm{\mu}_k^1,\bm{\Sigma}_k^1,\mathcal{C}_k^1,\pi_k^1 \right)_{1:K}$, $t \leftarrow 1$.
   \REPEAT
   \STATE $\textbf{Expectation step}$:
   $r_{nk} = \frac{\pi_k p(x_n \mid \bm{\mu}_k, \bm{\Sigma}_k)}{\sum_{l=1}^{K} \pi_l p(x_n \mid \bm{\mu}_l, \bm{\Sigma}_l)}$ 
   \STATE $\textbf{Maximization step}$:
 
   \FOR{$k=1$ {\bfseries to} $K$}
   \STATE Compute $\mathcal{C}_k^t(\bm{\mu}_k^t,\bm{\Sigma}_k^t)$ 
   \STATE Compute $d_{\bm{\mu}_k} \mathcal{L}(\bm{\mu}_k^t,\bm{\Sigma}_k^t)$ using Eq.~\eqref{eq:steepestd_mu}
   \IF{$||d_{\bm{\mu}_k} \mathcal{L}|| < \epsilon_{\nabla_{\bm{\mu}}}$}
   \STATE $\textbf{Continue}$
   \ENDIF
   \STATE $\bm{\mu}_k^{t+1} \leftarrow \expmap_{\bm{\mu}_k^t}(\alpha_{\bm{\mu}}^t d_{\bm{\mu}_k}
   \mathcal{L})$
   \STATE Compute $\logmap_{\bm{\mu}_k^{t+1}}(\bm{x}_{1:N})$
   \STATE Compute $\mathcal{C}_k^{t+1}(\bm{\mu}_k^{t+1},\bm{\Sigma}_k^t)$ 
   \STATE $\bm{\Sigma}_k^{t+1} \leftarrow \text{update}_{\bm{\Sigma}_k^t}$ using Alg.~\ref{alg:covupdate}
   \STATE $\pi_k^t = \frac{1}{N} \sum_{n=1}^{N} r_{nk}$
   \ENDFOR
   \IF{$\mathcal{L}^{t+1} < \mathcal{L}^t$}
   \STATE $\alpha_{\bm{\mu}}^{t+1} \leftarrow 1.1 \cdot \alpha_{\bm{\mu}}^{t}$ \COMMENT{optimism}
   \ELSE
   \STATE $\alpha_{\bm{\mu}}^{t+1} \leftarrow 0.75 \cdot \alpha_{\bm{\mu}}^{t}$ \COMMENT{pessimism}
   \ENDIF
   \STATE $t \leftarrow t+1$
   \UNTIL{$||\mathcal{L}^{t+1}-\mathcal{L}^t|| \leq \epsilon_{\mathcal{L}}$ \OR $t=t_{max}$}
\end{algorithmic}
\end{algorithm}

% LAND COVARIANCE PSEUDOCODE
\begin{algorithm}[t]
\caption{\textsc{land} $\text{update}_{\bm{\Sigma}_k}$ on the symmetric positive definite manifold $\mathcal{S}_{+}$}
\label{alg:covupdate}
\begin{algorithmic}
\STATE {\bfseries Input:} Covariance $\bm{\Sigma}_k^t$, mean $\bm{\mu}_k$, max. linesearch iterations $t_{max,\bm{\Sigma}}$, last stepsize $\alpha_k$, initial stepsize $\alpha_1=1.0$, \\sufficient decrease factor $c_0=0.5$, contraction factor $c_1=0.5$
\STATE {\bfseries Output:} $\bm{\Sigma}_k^{t+1}$, $\alpha_k$ (for reuse) 
\STATE $\text{ }$
\STATE \COMMENT{define the exp. map on the $\mathcal{S}_+$ manifold, where $\bm{X}$ is an SPD matrix and $\bm{\Xi}$ is a tangent vector, i.e., a symmetric matrix}
\STATE \textbf{Function}  $\expmap_{\bm{X}}^+(\bm{\Xi})$: 
\STATE \quad \textbf{return} $\bm{X}^{\frac{1}{2}} \exp\left(\bm{X}^{-\frac{1}{2}} \bm{\Xi} \bm{X}^{-\frac{1}{2}}\right) \bm{X}^{\frac{1}{2}}$, \quad where $\exp$ denotes the matrix exponential.
\STATE \textbf{EndFunction}
% SPD norm
\STATE \COMMENT{define the norm of a vector $\bm{\Xi}$ in the tangent space of $\bm{X} \in \mathcal{S}_{+}$}
\STATE \textbf{Function}  $\left(\|\cdot\|_{\bm{X}}^+\right)(\bm{\Xi})$:
\STATE \quad $\bm{X} \leftarrow \bm{L} \bm{L}^\intercal$ \hfill\COMMENT{cholesky decomposition}
\STATE \quad \textbf{return} $\|\bm{L}^{-1} \bm{\Xi} \bm{L}^{-\intercal}\|_2$
% \STATE \quad \textbf{return} $\sqrt{\tr((\bm{x} \backslash \bm{\xi})^2)}$
\STATE \textbf{EndFunction}
\STATE $\text{ }$
% gradient descent loop
\FOR{$i=1$ {\bfseries to} $2$ \COMMENT{outer gradient descent loop}}
\STATE Compute (or retrieve from cache) $\mathcal{L}(\bm{\Sigma}_k^t)$ 
\STATE Compute (or retrieve from cache) Euclidean gradient $\nabla_{\bm{\Sigma}_k^t} \mathcal{L}(\Sigma_k^t)$ using Eq.~\eqref{eq:sigmagrad}
\STATE Obtain manifold gradient: $g \coloneqq \nabla_{\bm{\Sigma}_k^t;\mathcal{S}_{+}} = \frac{1}{2} \bm{\Sigma}_k^{t} \left(\nabla_{\bm{\Sigma}_k^t} + \nabla_{\bm{\Sigma}_k^t}^\intercal \right)\bm{\Sigma}_k^{t}$
\IF{$\alpha_k$ \textbf{is None or } $\alpha_k = 0$}
\STATE $\alpha_k \leftarrow \frac{\alpha_0}{\|g\|}$
\ENDIF
% take the step with expmap
\STATE $\bm{\Sigma}_k^{t+1} \leftarrow \expmap_{\bm{\Sigma}_k^t}^+\left(- \alpha_k \cdot g\right)$
% compute new constant
\STATE Compute $\mathcal{C}_k(\bm{\mu}_k,\bm{\Sigma}_k^{t+1})$
\STATE Evaluate \textsc{land} objective $\mathcal{L}(\bm{\Sigma}_k^{t+1})$
% LINESEARCH
\STATE \COMMENT{Linesearch subroutine}
\STATE $j \leftarrow 1$
%linesearch condition
\WHILE{$\mathcal{L}(\bm{\Sigma}_k^{t+1}) > \mathcal{L}(\bm{\Sigma}_k^{t}) - c_0 \cdot \alpha_k \cdot \|g\|^2$ and $j \leq t_{max,\bm{\Sigma}}$}
\STATE \COMMENT {while no sufficient decrease, contract}
\STATE $\alpha_k \leftarrow \alpha_k \cdot c_1$
\STATE $\bm{\Sigma}_k^{t+1} \leftarrow \expmap_{\bm{\Sigma}_k^t}^+\left(- \alpha_k \cdot g\right)$
% evaluate constant in the linesearch loop
\STATE Compute $\mathcal{C}_k(\bm{\mu}_k,\bm{\Sigma}_k^{t+1})$
\STATE Evaluate \textsc{land} objective $\mathcal{L}(\bm{\Sigma}_k^{t+1})$
\STATE $j \leftarrow j + 1$
\ENDWHILE
\IF{$\mathcal{L}(\bm{\Sigma}_k^{t+1}) > \mathcal{L}(\bm{\Sigma}_k^{t})$}
\STATE $\alpha_k \leftarrow 0$
\ENDIF 
\IF{$j \neq 2$}
\STATE $\alpha_k = 1.3 \cdot \alpha_k$ \COMMENT{optimism}
\ENDIF
\STATE $t \leftarrow t+1$
\ENDFOR
\end{algorithmic}
\end{algorithm}

For $\bm{\mu}$, we use the steepest descent direction as in \citet{arvanitidis2016land}
\begin{equation}
\label{eq:steepestd_mu}
    d_{\bm{\mu}_k} \mathcal{L} = \sum_{n=1}^{N} r_{nk} \logmap_{\bm{\mu}_k}(\bm{x}_n) - \frac{\mathcal{Z}_k \cdot R_k}{\mathcal{C}_k(\bm{\mu}_k,\bm{\Sigma}_k)} \int_{\mathcal{T}_{\bm{\mu}_k}\mathcal{M}} \bm{v} g_{\bm{\mu}_k} (\bm{v})  \N (\bm{v}; \bm{0}, \bm{\Sigma}_k) \d\bm{v},  
\end{equation}
where the vector-valued integral stems from \bq~and $R_k=\sum_{n=1}^{N} r_{nk}$, $\mathcal{Z}_k=\sqrt{(2\pi)^d |\bm{\Sigma}_k|}$. 

\citet{arvanitidis2016land} decomposed the precision $\bm{\Sigma}_k^{-1}=\bm{A}^\intercal \bm{A}$ for unconstrained optimization using gradient descent. We opt for a more principled approach by exploiting geometric structure of the symmetric positive definite (SPD) manifold, to which the covariance is confined. More specifically, we use the \textit{bi-invariant} metric \citep{bhatia2009positive}. Under this metric, geodesics from $\bm{A}$ to $\bm{B}$ may be parameterized as $
    \gamma(t) ~= \bm{A}^{\frac{1}{2}} \left(\bm{A}^{-\frac{1}{2}} \bm{B}^{\frac{1}{2}} \bm{A}^{-\frac{1}{2}} \right)^t \bm{A}^{\frac{1}{2}}, \quad 0 \leq t < 1
$, and the distance from $\bm{A}$ to $\bm{B}$ is $
    d(\bm{A},\bm{B}) ~= \left\Vert \log{\bm{A}^{-\frac{1}{2}} \bm{B}^{\frac{1}{2}} \bm{A}^{-\frac{1}{2}}} \right\Vert_2
$.
The name stems from the fact that this distance is invariant under multiplication with any invertible square matrix $\bm{\Xi}$, i.e., $d(\bm{A},\bm{B}) ~= d(\bm{\Xi} \cdot \bm{A},\bm{\Xi} \cdot \bm{B})$. For manifold gradient descent, we calculate the Euclidean gradient and then project it onto the manifold. We begin with the first term
\begin{equation}
    \nabla_{\bm{\Sigma}_k} \left(\sum_{n=1}^{N} r_{nk} \left[
    \frac{1}{2} \langle \logmap_{\bm{\mu}_k}(\bm{x}_n), \bm{\Sigma}_k^{-1} \logmap_{\bm{\mu_k}}(\bm{x}_n) \rangle \right] \right) = 
    - \frac{1}{2} \sum_{n=1}^{N} r_{nk} 
      \bm{\Sigma}_k^{-\intercal} \logmap_{\bm{\mu_k}}(\bm{x}_n)
      \logmap_{\bm{\mu_k}}(\bm{x}_n)^\intercal \bm{\Sigma}_k^{-\intercal}. 
\end{equation}
For the gradient of the normalization constant we get
\begin{equation}
\begin{aligned}
\nabla_{\bm{\Sigma}_k}  \log(\mathcal{C}(\bm{\mu}_k,\bm{\Sigma}_k)) &= \frac{1}{\mathcal{C}(\bm{\mu}_k,\bm{\Sigma}_k)} \int_{\mathcal{M}} \nabla_{\bm{\Sigma}_k} \exp\left(  \frac{1}{2} \langle \logmap_{\bm{\mu}_k}(\bm{x}), \bm{\Sigma}^{-1} \logmap_{\bm{\mu}_k}(\bm{x})\rangle\right) \d \mathcal{M}_{\bm{x}}\\
&= \frac{1}{2\cdot\mathcal{C}(\bm{\mu}_k,\bm{\Sigma}_k)} \int_{\mathcal{M}} \bm{\Sigma}_k^{-\intercal} \logmap_{\bm{\mu}_k}(\bm{x}) \logmap_{\bm{\mu}_k}(\bm{x})^\intercal \bm{\Sigma}_k^{-\intercal} \exp\left(-\frac{1}{2} \langle \logmap_{\bm{\mu}_k}(\bm{x}), \bm{\Sigma}^{-1} \logmap_{\bm{\mu}_k}(\bm{x})\rangle\right) \d \mathcal{M}_{\bm{x}} \\
&=  \frac{1}{2\cdot\mathcal{C}(\bm{\mu}_k,\bm{\Sigma}_k)} \int_{\mathcal{T}_{\bm{\mu}_k}\mathcal{M}} \bm{\Sigma}_k^{-\intercal} \bm{v} \bm{v}^\intercal g_{\bm{\mu}_k}(\bm{v}) \bm{\Sigma}_k^{-\intercal} \exp\left(-\frac{1}{2} \langle \bm{v}, \bm{\Sigma}^{-1} \bm{v}\rangle\right) \d \bm{v}. \\
\end{aligned}
\end{equation}
Taking this together, we obtain the gradient
\begin{equation}
\label{eq:sigmagrad}
\begin{aligned}
\nabla_{\bm{\Sigma}_k} \mathcal{L} = 
    &- \frac{1}{2} \sum_{n=1}^{N} r_{nk} 
      \bm{\Sigma}_k^{-\intercal} \logmap_{\bm{\mu_k}}(\bm{x}_n)
      \logmap_{\bm{\mu_k}}(\bm{x}_n)^\intercal \bm{\Sigma}_k^{-\intercal}\\
      &+ \frac{R_k}{2\cdot\mathcal{C}(\bm{\mu}_k,\bm{\Sigma}_k)} \int_{\mathcal{T}_{\bm{\mu}_k}\mathcal{M}} \bm{\Sigma}_k^{-\intercal} \bm{v} \bm{v}^\intercal g_{\bm{\mu}_k}(\bm{v}) \bm{\Sigma}_k^{-\intercal} \exp\left(-\frac{1}{2} \langle \bm{v}, \bm{\Sigma}^{-1} \bm{v}\rangle\right) \d \bm{v},
\end{aligned}
\end{equation}
where the matrix-valued integral again stems from \bq. To project the Euclidean gradient $\nabla_{\bm{\Sigma}_k}$ onto the tangent space of a SPD matrix $\bm{\Sigma}_k$, we simply calculate $\frac{1}{2} \bm{\Sigma}_k \left(\nabla_{\bm{\Sigma}_k} + \nabla_{\bm{\Sigma}_k}^\intercal \right)\bm{\Sigma}_k$. We optimize with gradient descent and a deterministic manifold linesearch as a subroutine, which adaptively chooses its step lengths. This procedure as well as the SPD manifold are conveniently available in the \texttt{Pymanopt} \citep{townsend2016pymanopt} library. 

In sum, the optimization process is as follows: we cycle through the components $K$. After taking a single steepest-direction step for $\bm{\mu}_k$, we perform two gradient descent steps for $\bm{\Sigma}_k$, each of which may use up to $4$ steps in the linesearch subroutine to satisfy a sufficient decrease criterion. We provide pseudocode for the covariance update in Alg.~\ref{alg:covupdate}.
The optimizer has the following hyperparameters:
\begin{center}
\begin{tabular}{lll}
	\toprule
	Parameter & Value & Description \\ 
	\midrule
	$t_{max}$ & - & update each component $\mathtt{t_{max}}$ times. \\ 
	$\alpha_{\bm{\mu}}^1$ & - & initial stepsize for mean updates.\\
	$\epsilon_{\nabla_{\bm{\mu}}}$ & - &  tolerance for mean gradients \\
	$\epsilon_{\mathcal{L}}$ & $2$ & likelihood tolerance \\
	$t_{max,\bm{\Sigma}}$ & $4$ & max. $\bm{\Sigma}$ linesearch steps. \\
	$\alpha_1$ & $1.0$ & initial step size ($\bm{\Sigma}$ linesearch).\\
	$c_0$ & $0.5$ & sufficient decrease factor ($\bm{\Sigma}$ linesearch).\\
	$c_1$ & $0.5$ & contraction factor ($\bm{\Sigma}$ linesearch)\\

	\bottomrule
\end{tabular} 
\end{center}
Cells with unspecified values (-) imply that the value of the respective parameter is not equal across all experiments and problems. Experiment-specific parameter details are in \ref{app:experiments}.

%%%%%%%%%%%%%%%%%%%%%%%%%%%%%%%%%%%%%%%%%%
\section{More Details on \textsc{bq}}
\label{app:dcv}
%%%%%%%%%%%%%%%%%%%%%%%%%%%%%%%%%%%%%%%%%%

\subsection{General \textsc{bq}}
Since we use the Mat\'ern-5/2 kernel and we require further integrals for the \textsc{land} objective gradients, we use the \textsc{gp} as an emulator of the function we wish to model; that is, we do not calculate integrals analytically, but use extensive Monte Carlo (\textsc{mc}) sampling on top of the \textsc{gp}, which implies evaluating the posterior mean at the locations randomly drawn from the integration measure. To compute the integral without loss of precision, we use $S=30{,}000$ samples to estimate the integrals. The time overhead and approximation error of this procedure are negligible in practice.

We optimize the marginal likelihood of the \textsc{gp} with respect to the hyperparameters and use their final values to initialize the next iteration, since during the optimization the function changes smoothly from each step to the next. This information is not shared across the $K$ components, but kept separately.

Our implementation of \textsc{bq} builds upon the \texttt{bayesquad} python library \citep{wagstaff2018batch}, which is available at \url{https://github.com/OxfordML/bayesquad}.

\subsection{\textsc{dcv} - Derivations and Technical Details}
The \textsc{dcv} acquisition function is
\begin{equation}
    \bar{u} (\bm{r}) = \int_0^\infty  u (\beta \bm{r}) \d\beta
    =  \int_0^\infty \tilde{k}_{\mathcal{D}}(\beta \bm{r},\beta \bm{r}) \pi(\beta \bm{r})^2 \d\beta,
\end{equation}
with derivative
\begin{equation}
    \frac{\partial}{\partial \bm{r}} \bar{u}(\bm{r}) = \int_0^\infty \beta \pi(\beta \bm{r}) \left[2 \tilde{k}_{\mathcal{D}}(\beta \bm{r},\beta \bm{r}) \frac{\partial}{\partial \beta\bm{r}} \pi(\beta \bm{r}) + \pi(\beta \bm{r}) \frac{\partial}{\partial \beta\bm{r}} \tilde{k}_{\mathcal{D}}(\beta \bm{r},\beta \bm{r})  \right] \d\beta.
\end{equation}
Since the integration measure is Gaussian, i.e., $\pi(\beta \bm{r}) = \mathcal{N}(\beta \bm{r}; 0, \bm{\Sigma})$, its derivative is
\begin{equation} 
    \frac{\partial}{\partial \beta\bm{r}} \pi(\beta \bm{r}) = - \pi(\beta \bm{r}) \bm{\Sigma}^{-1} \beta \bm{r}.
\end{equation}
For simplicity, we always use \wsabil~in combination with \textsc{dcv}, so the derivative of the variance of the warped GP is
\begin{equation}
    \frac{\partial}{\partial \beta\bm{r}} \tilde{k}_{\mathcal{D}}(\beta \bm{r},\beta \bm{r}) = \frac{\partial}{\partial \beta\bm{r}} \left[ m_{\mathcal{D}} (\beta\bm{r})^2 k_{\mathcal{D}} (\beta\bm{r}, \beta\bm{r}) \right] = 2 m_{\mathcal{D}} (\beta\bm{r}) k_{\mathcal{D}} (\beta\bm{r}, \beta\bm{r}) \frac{\partial}{\partial \beta\bm{r}} m_{\mathcal{D}} (\beta\bm{r}) + \frac{\partial}{\partial \beta\bm{r}}  k_{\mathcal{D}} (\beta\bm{r}, \beta\bm{r}) m_{\mathcal{D}} (\beta\bm{r})^2.
\end{equation}
The derivative of the \textsc{dcv} acquisition function is significantly more costly to evaluate than the objective, because it requires predictive gradients of the underlying GP. Instead of using a quadrature routine like \texttt{scipy.quad}, which would evaluate the integral for every dimension sequentially, we use Simpson's rule on 50 evenly spaced points between $0$ and $\alpha_{\text{max}}$ (defined below). Since these are multiple univariate integrals of a smooth function, the errors are practically negligible.

The scalar $\alpha_{\text{max}}$ simultaneously constitutes an upper bound for the integration and the length of the exponential map. A bound is reasonable since longer exponential maps are slower to compute and the integration measure concentrates the mass near the center, so very far-away locations become irrelevant. For a sensible bound, we use the chi-square distribution:
\begin{equation}
    \langle \alpha \cdot \bm{r}, \bm{\Sigma}^{-1} \alpha \cdot \bm{r} \rangle = \chi_{p}^{2}
\end{equation}
by choosing a high value $p=99.5\%$, we make sure that there is no significant amount of mass outside of this isoprobability contour. Note that this limit applies only to the computation of exponential maps and the collection of observations, not to the main quadrature itself.

Since $\bm{r}$ is constrained to lie on the unit hypersphere, we employ manifold gradient descent with a linesearch subroutine. 
Conveniently, the linesearch only evaluates the objective and not its gradient, which saves a significant amount of time. Overall, optimizing this acquisition function is costly, however.

For completeness, we briefly describe the geometry of the unit (hyper)sphere. If the tangent space of our data manifold is $\mathcal{T}_{\bm{\mu}}\mathcal{M} = \R^D$, then a direction in this tangent space is a point on $\mathbb{S}^{D-1}$, which we represent as a unit norm vector in $\R^D$. For a point $\bm{x}$ on the sphere and a tangent vector $\bm{\xi}$, which lies in the plane touching the sphere tangentially, the exponential map is $  \expmap_{\bm{x}}(\bm{\xi}) = \cos(\|\bm{\xi}\|_2)\bm{x} + \sin(\|\bm{\xi}\|_2) \frac{\bm{\xi}}{\|\bm{\xi}\|_2}
$. However, the optimizer uses a \textit{retraction map} $\operatorname{Retr}_{\bm{x}}(\bm{\xi}) = \frac{\bm{x}+\bm{\xi}}{\|\bm{\xi}\|_2}$ instead of the exponential map to take a descent step. To obtain the gradient on the manifold, the Euclidean gradient is orthogonally projected onto the tangent plane.

The gradient descent is allowed a maximum of $15$ steps in the ``error vs. runtime experiment'', whereas in the boxplot experiment we decrease this number to $5$, as this experiment focuses more on speed given a fixed number of samples. 
The linesearch may use up to $5$ steps. We set the optimism of the linesearch to $2.0$ and the initial stepsize to $1.0$. If a descent step has norm less than $10^{-10}$, the optimization is aborted.

After an exponential map is computed according to \textsc{dcv}, we discretize the resulting straight line in the tangent space into $30$ evenly spaced points and sequentially select $6$ points using the standard \textsc{wsabi} objective, updating the \textsc{gp} after each observation.

%%%%%%%%%%%%%%%%%%%%%%%%%%%%%%%%%%%%%%%%%%
\section{More Details on the Experiments}
\begin{figure*}
\centering     %%% not \center
\hspace{-1.7mm}
\subfigure[\textsc{circle} \oldstylenums{3}\textsc{d}]{\label{fig:a1}\scalebox{0.3}{\input{fig/continuous_circle-1000-3d_res.pgf}}}
\hspace{1.2mm}
\subfigure[\textsc{circle} \oldstylenums{4}\textsc{d}]{\label{fig:b1}\scalebox{0.3}{\input{fig/continuous_circle-1000-4d_res.pgf}}} 
\hspace{1.2mm}
\subfigure[\textsc{curly}]{\label{fig:c1}\scalebox{0.3}{\input{fig/continuous_sparse_moon_res.pgf}}} 
\hspace{1.2mm}
\subfigure[\oldstylenums{2}-\textsc{circles}]{\label{fig:d1}\scalebox{0.3}{\input{fig/continuous_two_circles_res.pgf}}}
\hfill
\caption{Comparison of \bq~and \textsc{mc}
errors (vertical log scale, shared legend and axes) for different manifolds, on the first integration problem of the respective \textsc{land} fit. Shaded regions indicate $95\%$ confidence intervals w.r.t. the $30$ independent runs.}
\label{fig:app_continuous_graph}
\end{figure*}
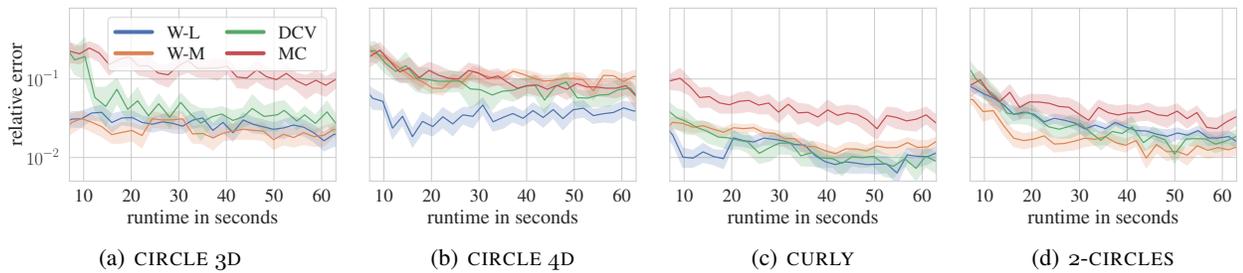

%%%%%%%%%%%%%%%%%%%%%%%%%%%%%%%%%%%%%%%%%%
\label{app:experiments}
In this supplementary section, we give details about the conducted experiments and report further results, not included in the main paper due to space limitations. Fig.~\ref{fig:app_boxplots} and Fig.~\ref{fig:app_continuous_graph} follow the methodology as sketched in the main paper. The runtimes belonging to Fig.~\ref{fig:app_boxplots} are displayed in Fig.~\ref{fig:app:runtime_barplot}. We here also report mean runtimes of exponential maps (Tab.~\ref{app:tab:expmaps_runtimes}).
\begin{table}
    \centering
    \caption{Mean exponential map runtime in milliseconds, obtained by averaging over \textsc{mc} runtimes on the entire \textsc{land} fit.}
    \vskip 0.15in
    \begin{center}
    \begin{small}
    \begin{tabular}{cccccccc}
        \toprule
         \textsc{circle} & \textsc{circle} \oldstylenums{5}\textsc{d} & \textsc{mnist} & \textsc{adk} & \textsc{circle} \oldstylenums{3}\textsc{d} & \textsc{circle} \oldstylenums{4}\textsc{d} & \textsc{curly} & \oldstylenums{2}-\textsc{circles} \\
         \midrule
         $60$ & $50$ & $238$ & $68$ & $32$ & $45$ & $62$ & $36$\\
         \bottomrule
        \end{tabular}
        \end{small}
        \end{center}
        \vskip -0.15in
    \label{app:tab:expmaps_runtimes}
\end{table}

\subsection{Synthetic Experiments}
We created two further synthetic datasets \textsc{curly} and  \oldstylenums{2}-\textsc{circles} that are not shown in the main paper (Fig.~\ref{app:fig:more_lands}).
\begin{figure}
    \centering
    \subfigure[\textsc{curly}]{\includegraphics[width=0.2\textwidth]{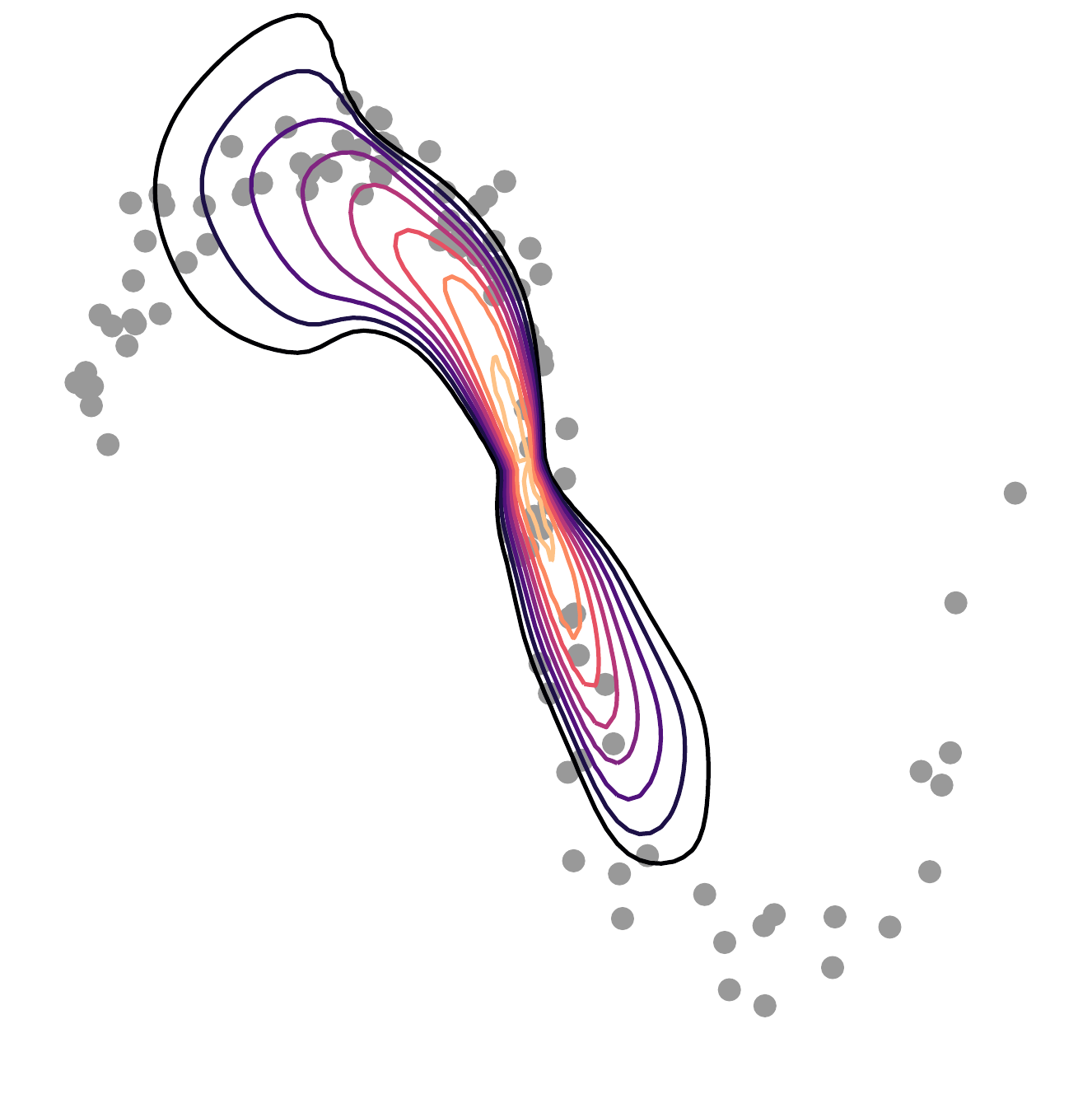}}
    \subfigure[\oldstylenums{2}-\textsc{circles}]{\includegraphics[width=0.2\textwidth]{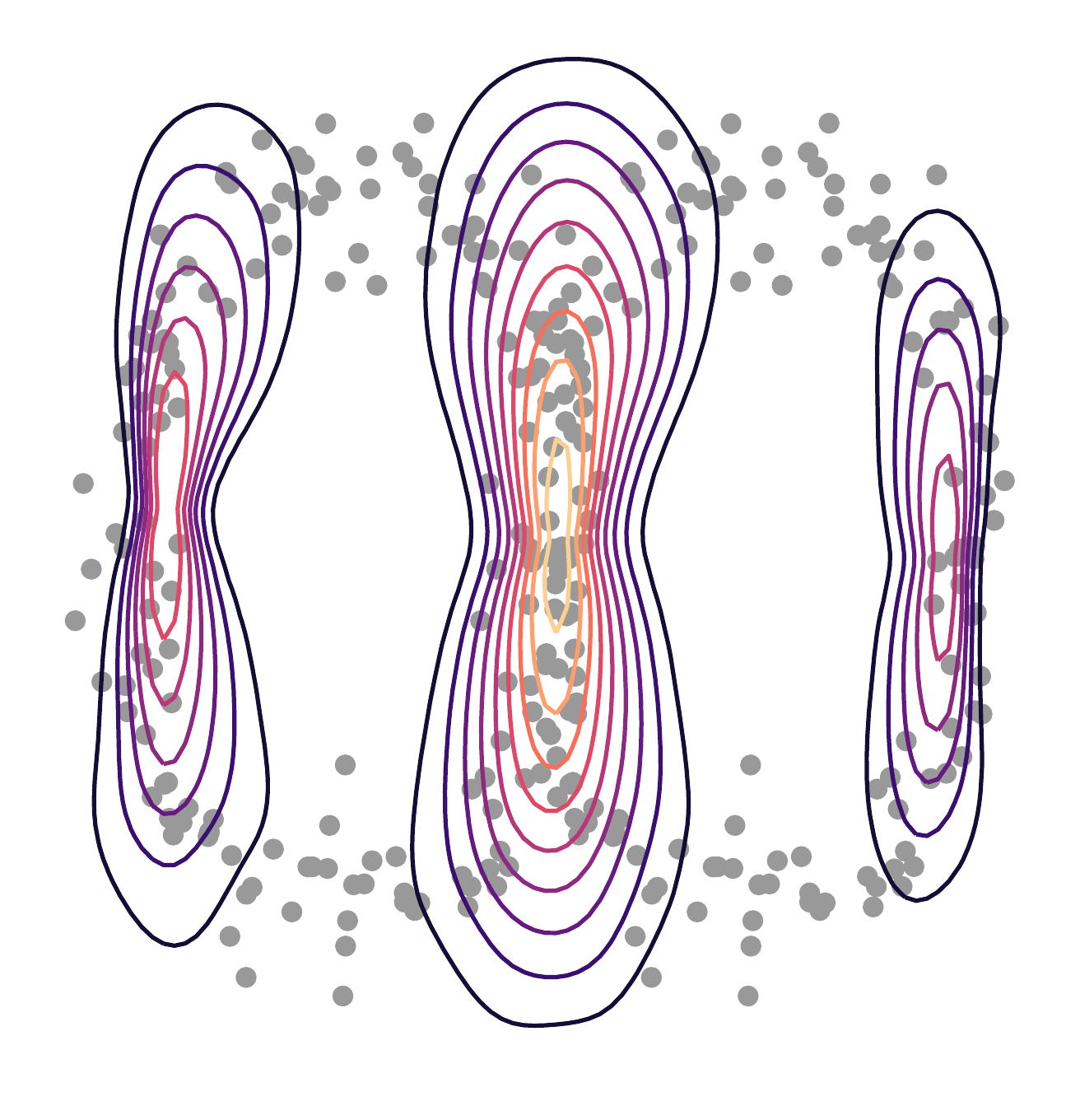}}
    \caption{Toy data \textsc{land} fits}
    \label{app:fig:more_lands}
\end{figure}
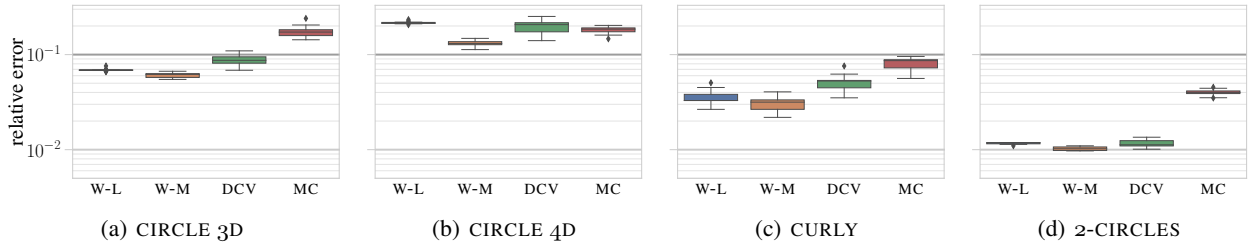
\begin{figure*}
\centering     %%% not \center
\hspace{-1.2mm}
\subfigure[\textsc{circle} \oldstylenums{3}\textsc{d}]{\scalebox{0.3}{\input{fig/boxplot_circle-1000-3d_res.pgf}}}
\hspace{1.2mm}
\subfigure[\textsc{circle} \oldstylenums{4}\textsc{d}]{\scalebox{0.3}{\input{fig/boxplot_circle-1000-4d_res.pgf}}}
\hspace{1.2mm}
\subfigure[\textsc{curly}]{\scalebox{0.3}{\input{fig/boxplot_sparse_moon_res.pgf}}}
\hspace{1.2mm}
\subfigure[\oldstylenums{2}-\textsc{circles}]{\scalebox{0.3}{\input{fig/boxplot_two_circles_res.pgf}}}
\hfill
\caption{Boxplot error comparison (log scale, shared y-axis) of \bq~and \textsc{mc} on whole \textsc{land} fit for different manifolds. For \textsc{mc}, we allocate the runtime of the mean slowest \bq~method. Each box contains 16 independent runs.}
\label{fig:app_boxplots}
\end{figure*}  
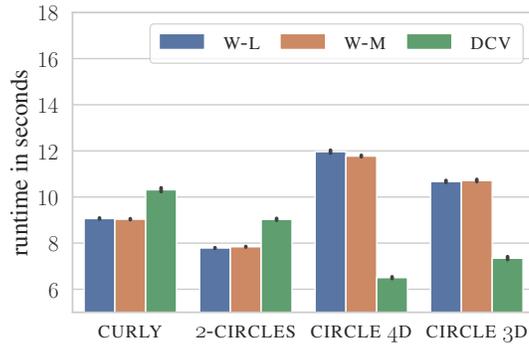
\begin{figure}
    \centering
    \scalebox{0.52}{\input{fig/runtime_barplot_res_app.pgf}}
    \caption{Mean runtime comparison (for a single integration) of the \bq~methods. Errorbars indicate 95\% confidence intervals w.r.t the 16 runs on each \textsc{land} fit. 
    The reported runtimes belong to the boxplots in Fig.~\ref{fig:app_boxplots}.
    }
    \label{fig:app:runtime_barplot}
\end{figure}

\subsection{\textsc{mnist}}
We sampled $5{,}504$ random data points from the first three digits of \textsc{mnist} \citep{mnist}, which were preprocessed by normalizing them feature-wise to $[-1,+1]$ using \texttt{sklearn.preprocessing.MinMaxScaler}.
We trained a simple Variational-Autoencoder (VAE) to embed the $784$ dimensional input in a latent space of dimension $2$. The architecture uses separate encoders $\bm{\mu}_\phi$, $\bm{\sigma}_\phi$ and decoders $\bm{\mu}_\theta$, $\bm{\sigma}_\theta$. In summary:
\begin{center}
\begin{tabular}{crrr}
	\toprule 
	Encoder/Decoder & Layer 1 & Layer 2 & Layer 3 \\ 
	\midrule
	$\bm{\mu}_\phi$ & 128 (tanh) & 64~~(tanh) & 2~~~~(linear) \\ 
	$\bm{\sigma}_\phi$ & 128 (tanh) & 64~~(tanh) & 2 (softplus) \\ 
	$\bm{\mu}_\theta$ & 64 (linear) & 128 (linear) & 784~~~~(linear) \\ 
	$\bm{\sigma}_\theta$ & 64 (linear) & 128 (linear) & ~784 (softplus) \\ 
	\bottomrule 
\end{tabular} 
\end{center}
We trained the network for $200$ epochs using \textsc{adam} with a learning rate of $10^{-3}$.
The resulting latent codes were used to construct the \textit{Aggregated Posterior Metric}, with $\rho=0.001$, such that the measure far from the data is $1000$. The small variances cause high curvature, which makes the integration tasks challenging and geodesic computations slow. To fit the \textsc{land}, we used $250$ subsampled points to lower the amount of time spent on \textsc{bvps}. In contrast, the \textsc{gmm} was fitted on the whole $5{,}504$ points. Note that Fig.~\ref{fig:mnist_land} shows this training data.

\subsection{\textsc{adk}}
We obtained protein trajectory data of adenylate kinase from
\begin{center}
    \url{https://www.mdanalysis.org/MDAnalysisData/adk_transitions.html#adk-dims-transitions-ensemble-dataset}
\end{center}
\citep{adkdims}. We use the \textsc{dims} variant, a dataset which comprises 200 trajectories and select a subset consisting of the trajectories $160-200$, which contain in total $2{,}038$ data points. To model the assumed high curvature of the trajectory space, we choose the kernel metric with $\sigma=0.035$ and $\rho=10^{-5}$. To visualize spatial protein structure, we used the software \textsc{vmd} \citep{vmd} with the ``new cartoon'' representation, colored according to ``residue type''.

According to \citet{adkdims}, ``AdK’s closed/open transition [..] is a standard test case that captures general, essential
features of conformational changes in proteins''. This well-studies transition involves the movement of the \textit{LID} and \textit{NMP} domains against the rather stable core domain. As a consequence, it can be described by two angles $\theta_{LID}$ and $\theta_{NMP}$. In Fig.~\ref{fig:adks}, it is visible how the \textit{LID} opens to the top, whereas the \textit{NMP} domain moves towards the bottom right (from this particular perspective).

\subsection{General Methodology}
For the aforementioned manifolds we fitted the \textsc{land} mixture model with a pre-determined component number $K$. 

As the ground truth, we obtained $S=40{,}000$ \textsc{mc} samples on each integration problem. Since obtaining a large number of exponential maps is computationally extremely expensive, we subsampled from this pool of ground truth samples when \textsc{mc} samples were required in the experiments, instead of running \textsc{mc} again. For example, in the ``error vs. runtime'' experiment, we calculated the mean \textsc{mc} runtime per sample from the ground truth pool of this particular problem and then subsampled as many samples as the given runtime limit affords. For the boxplot experiments, we averaged the \textsc{mc} runtimes over the whole \textsc{land} fit and always obtained the same number of samples per integration. Note that the \textsc{mc} runtime practically corresponds to the runtime of the exponential maps, since the overhead is minimal.

All experiments were run in a cloud setting on $8$ virtual CPUs. We restricted the core usage of \textsc{blas} linear algebra subroutines to a single core, so as not to create interference between multiple processes.

\subsection{Manifold and Optimization Hyperparameters}
In Table~\ref{tab:landparams}, we report the relevant hyperparameters for the metrics ($\sigma$, $\rho$), which were used to construct the manifolds, and those optimization parameters which are not equal across all problems.
\begin{table}[H]
\centering
\begin{tabular}{lrrrrrrrr}
	\toprule
	Parameter & \textsc{circle} & \textsc{circle} \oldstylenums{3}\textsc{d} & \textsc{circle} \oldstylenums{4}\textsc{d} & \textsc{circle} \oldstylenums{5}\textsc{d} & \textsc{mnist} & \textsc{adk} & \textsc{curly} & \oldstylenums{2}-\textsc{circles} \\ 
	\midrule
	$\sigma$ & 0.1 & 0.25 & 0.25 & 0.25 & - & 0.035 & 0.2 & 0.15 \\ 
	$\rho$ & 0.001 & 0.01 & 0.0316 & 0.063 & 0.001 & 0.00001 & 0.01 & 0.01 \\ 
	$K$ & 2 & 2 & 2 & 2 & 3 & 1 & 1 & 3 \\ 
	$t_{max}$ & 7 & 4 & 4 & 4 & 7 & 7 & 7 & 7 \\ 
	$\alpha_{\bm{\mu}}^1$ & $0.3$ & $0.3$ & $0.3$  & $0.3$ & $0.3$ & $0.2$ & $0.3$ & $0.3$ \\
	$\epsilon_{\nabla_{\bm{\mu}}}$ & $0.01$ & $0.01$ & $0.01$ & $0.01$ & $0.015$ & $0.01$ & $0.01$ & $0.01$\\
	integrations & 67 & 39 & 40 & 34 & 105 & 36 & 33 & 111 \\ 
	\bottomrule
\end{tabular} 
\caption{Manifold and \textsc{land} optimization hyperparameters and resulting number of integrations.}
\label{tab:landparams}
\end{table}

\subsection{Boxplot Experiments (Fig.~\ref{fig:boxplots}, Fig.~\ref{fig:app_boxplots})}
These experiment were conducted on whole \textsc{land} fits, with $16$ independent runs for each of the $3$ \textsc{bq} methods. From Table~\ref{tab:landparams}, we can easily calculate the total number of runs as $48 \cdot (67+39+40+34+105+36+33+111) = 22{,}320$.

\subsection{Error vs. Runtime Experiments (Fig.~\ref{fig:continuous_graph}, Fig.~\ref{fig:app_continuous_graph})}
We evenly space $30$ runtime limits between $5$ and $65$ seconds using \texttt{np.linspace(5., 65., 30)}. For each of these runtime limits, we let each \textsc{bq} method run $30$ times. \textsc{bq} will stop collecting more samples as soon as the runtime limit is reached. After this, however, it will take some more time to finalize, as an ongoing computation is not interrupted. We then record the actually resulting runtimes and average over the 30 runs. These averages are then used for the x-axes of the plots, whereas the mean relative error is on the y-axes. In total, each \textsc{bq} method thus has $900$ runs on each problem. The $8$ plots, $4$ in the main paper and $4$ in the supplementary, contain $3 \cdot 900 \cdot 8 = 21{,}600$ runs. Together with the boxplot experiments, we obtain $21{,}600 + 22{,} = 43{,}920$ \textsc{bq} runs, that is, $14{,}640$ for each of the $3$ methods.

In Fig.~\ref{fig:contc}, we removed $4$ extreme \textsc{dcv} outliers, where seemingly the \textsc{gp} ``broke''. This amounts to $\frac{4}{21{,}600} = 0.01852\%$ of the \textsc{bq} runs in the $8$ plots.

%% file: fig/boxplot_circle-1000-3d_res.pgf
%% Creator: Matplotlib, PGF backend
%%
%% To include the figure in your LaTeX document, write
%%   \input{<filename>.pgf}
%%
%% Make sure the required packages are loaded in your preamble
%%   \usepackage{pgf}
%%
%% Figures using additional raster images can only be included by \input if
%% they are in the same directory as the main LaTeX file. For loading figures
%% from other directories you can use the `import` package
%% and then include the figures with
%%   \import{<path to file>}{<filename>.pgf}
%%
%% Matplotlib used the following preamble
%%
\begingroup%
\makeatletter%
\begin{pgfpicture}%
\pgfpathrectangle{\pgfpointorigin}{\pgfqpoint{5.920874in}{3.524405in}}%
\pgfusepath{use as bounding box, clip}%
\begin{pgfscope}%
\pgfsetbuttcap%
\pgfsetmiterjoin%
\definecolor{currentfill}{rgb}{1.000000,1.000000,1.000000}%
\pgfsetfillcolor{currentfill}%
\pgfsetlinewidth{0.000000pt}%
\definecolor{currentstroke}{rgb}{1.000000,1.000000,1.000000}%
\pgfsetstrokecolor{currentstroke}%
\pgfsetdash{}{0pt}%
\pgfpathmoveto{\pgfqpoint{0.000000in}{0.000000in}}%
\pgfpathlineto{\pgfqpoint{5.920874in}{0.000000in}}%
\pgfpathlineto{\pgfqpoint{5.920874in}{3.524405in}}%
\pgfpathlineto{\pgfqpoint{0.000000in}{3.524405in}}%
\pgfpathclose%
\pgfusepath{fill}%
\end{pgfscope}%
\begin{pgfscope}%
\pgfsetbuttcap%
\pgfsetmiterjoin%
\definecolor{currentfill}{rgb}{1.000000,1.000000,1.000000}%
\pgfsetfillcolor{currentfill}%
\pgfsetlinewidth{0.000000pt}%
\definecolor{currentstroke}{rgb}{0.000000,0.000000,0.000000}%
\pgfsetstrokecolor{currentstroke}%
\pgfsetstrokeopacity{0.000000}%
\pgfsetdash{}{0pt}%
\pgfpathmoveto{\pgfqpoint{1.220874in}{0.454405in}}%
\pgfpathlineto{\pgfqpoint{5.870874in}{0.454405in}}%
\pgfpathlineto{\pgfqpoint{5.870874in}{3.474405in}}%
\pgfpathlineto{\pgfqpoint{1.220874in}{3.474405in}}%
\pgfpathclose%
\pgfusepath{fill}%
\end{pgfscope}%
\begin{pgfscope}%
\definecolor{textcolor}{rgb}{0.000000,0.000000,0.000000}%
\pgfsetstrokecolor{textcolor}%
\pgfsetfillcolor{textcolor}%
\pgftext[x=1.802124in,y=0.357183in,,top]{\color{textcolor}\rmfamily\fontsize{24.200000}{29.040000}\selectfont \textsc{w-l}}%
\end{pgfscope}%
\begin{pgfscope}%
\definecolor{textcolor}{rgb}{0.000000,0.000000,0.000000}%
\pgfsetstrokecolor{textcolor}%
\pgfsetfillcolor{textcolor}%
\pgftext[x=2.964624in,y=0.357183in,,top]{\color{textcolor}\rmfamily\fontsize{24.200000}{29.040000}\selectfont \textsc{w-m}}%
\end{pgfscope}%
\begin{pgfscope}%
\definecolor{textcolor}{rgb}{0.000000,0.000000,0.000000}%
\pgfsetstrokecolor{textcolor}%
\pgfsetfillcolor{textcolor}%
\pgftext[x=4.127124in,y=0.357183in,,top]{\color{textcolor}\rmfamily\fontsize{24.200000}{29.040000}\selectfont \textsc{dcv}}%
\end{pgfscope}%
\begin{pgfscope}%
\definecolor{textcolor}{rgb}{0.000000,0.000000,0.000000}%
\pgfsetstrokecolor{textcolor}%
\pgfsetfillcolor{textcolor}%
\pgftext[x=5.289624in,y=0.357183in,,top]{\color{textcolor}\rmfamily\fontsize{24.200000}{29.040000}\selectfont \textsc{mc}}%
\end{pgfscope}%
\begin{pgfscope}%
\pgfpathrectangle{\pgfqpoint{1.220874in}{0.454405in}}{\pgfqpoint{4.650000in}{3.020000in}}%
\pgfusepath{clip}%
\pgfsetrectcap%
\pgfsetroundjoin%
\pgfsetlinewidth{2.509375pt}%
\definecolor{currentstroke}{rgb}{0.800000,0.800000,0.800000}%
\pgfsetstrokecolor{currentstroke}%
\pgfsetdash{}{0pt}%
\pgfpathmoveto{\pgfqpoint{1.220874in}{0.954042in}}%
\pgfpathlineto{\pgfqpoint{5.870874in}{0.954042in}}%
\pgfusepath{stroke}%
\end{pgfscope}%
\begin{pgfscope}%
\pgfsetbuttcap%
\pgfsetroundjoin%
\definecolor{currentfill}{rgb}{0.800000,0.800000,0.800000}%
\pgfsetfillcolor{currentfill}%
\pgfsetlinewidth{1.003750pt}%
\definecolor{currentstroke}{rgb}{0.800000,0.800000,0.800000}%
\pgfsetstrokecolor{currentstroke}%
\pgfsetdash{}{0pt}%
\pgfsys@defobject{currentmarker}{\pgfqpoint{-0.138889in}{0.000000in}}{\pgfqpoint{0.000000in}{0.000000in}}{%
\pgfpathmoveto{\pgfqpoint{0.000000in}{0.000000in}}%
\pgfpathlineto{\pgfqpoint{-0.138889in}{0.000000in}}%
\pgfusepath{stroke,fill}%
}%
\begin{pgfscope}%
\pgfsys@transformshift{1.220874in}{0.954042in}%
\pgfsys@useobject{currentmarker}{}%
\end{pgfscope}%
\end{pgfscope}%
\begin{pgfscope}%
\definecolor{textcolor}{rgb}{0.000000,0.000000,0.000000}%
\pgfsetstrokecolor{textcolor}%
\pgfsetfillcolor{textcolor}%
\pgftext[x=0.412738in,y=0.834057in,left,base]{\color{textcolor}\rmfamily\fontsize{24.200000}{29.040000}\selectfont \(\displaystyle 10^{-2}\)}%
\end{pgfscope}%
\begin{pgfscope}%
\pgfpathrectangle{\pgfqpoint{1.220874in}{0.454405in}}{\pgfqpoint{4.650000in}{3.020000in}}%
\pgfusepath{clip}%
\pgfsetrectcap%
\pgfsetroundjoin%
\pgfsetlinewidth{2.509375pt}%
\definecolor{currentstroke}{rgb}{0.611765,0.611765,0.611765}%
\pgfsetstrokecolor{currentstroke}%
\pgfsetdash{}{0pt}%
\pgfpathmoveto{\pgfqpoint{1.220874in}{2.613798in}}%
\pgfpathlineto{\pgfqpoint{5.870874in}{2.613798in}}%
\pgfusepath{stroke}%
\end{pgfscope}%
\begin{pgfscope}%
\pgfsetbuttcap%
\pgfsetroundjoin%
\definecolor{currentfill}{rgb}{0.800000,0.800000,0.800000}%
\pgfsetfillcolor{currentfill}%
\pgfsetlinewidth{1.003750pt}%
\definecolor{currentstroke}{rgb}{0.800000,0.800000,0.800000}%
\pgfsetstrokecolor{currentstroke}%
\pgfsetdash{}{0pt}%
\pgfsys@defobject{currentmarker}{\pgfqpoint{-0.138889in}{0.000000in}}{\pgfqpoint{0.000000in}{0.000000in}}{%
\pgfpathmoveto{\pgfqpoint{0.000000in}{0.000000in}}%
\pgfpathlineto{\pgfqpoint{-0.138889in}{0.000000in}}%
\pgfusepath{stroke,fill}%
}%
\begin{pgfscope}%
\pgfsys@transformshift{1.220874in}{2.613798in}%
\pgfsys@useobject{currentmarker}{}%
\end{pgfscope}%
\end{pgfscope}%
\begin{pgfscope}%
\definecolor{textcolor}{rgb}{0.000000,0.000000,0.000000}%
\pgfsetstrokecolor{textcolor}%
\pgfsetfillcolor{textcolor}%
\pgftext[x=0.412738in,y=2.493813in,left,base]{\color{textcolor}\rmfamily\fontsize{24.200000}{29.040000}\selectfont \(\displaystyle 10^{-1}\)}%
\end{pgfscope}%
\begin{pgfscope}%
\pgfpathrectangle{\pgfqpoint{1.220874in}{0.454405in}}{\pgfqpoint{4.650000in}{3.020000in}}%
\pgfusepath{clip}%
\pgfsetrectcap%
\pgfsetroundjoin%
\pgfsetlinewidth{0.401500pt}%
\definecolor{currentstroke}{rgb}{0.800000,0.800000,0.800000}%
\pgfsetstrokecolor{currentstroke}%
\pgfsetstrokeopacity{0.550000}%
\pgfsetdash{}{0pt}%
\pgfpathmoveto{\pgfqpoint{1.220874in}{0.454405in}}%
\pgfpathlineto{\pgfqpoint{5.870874in}{0.454405in}}%
\pgfusepath{stroke}%
\end{pgfscope}%
\begin{pgfscope}%
\pgfpathrectangle{\pgfqpoint{1.220874in}{0.454405in}}{\pgfqpoint{4.650000in}{3.020000in}}%
\pgfusepath{clip}%
\pgfsetrectcap%
\pgfsetroundjoin%
\pgfsetlinewidth{0.401500pt}%
\definecolor{currentstroke}{rgb}{0.800000,0.800000,0.800000}%
\pgfsetstrokecolor{currentstroke}%
\pgfsetstrokeopacity{0.550000}%
\pgfsetdash{}{0pt}%
\pgfpathmoveto{\pgfqpoint{1.220874in}{0.585827in}}%
\pgfpathlineto{\pgfqpoint{5.870874in}{0.585827in}}%
\pgfusepath{stroke}%
\end{pgfscope}%
\begin{pgfscope}%
\pgfpathrectangle{\pgfqpoint{1.220874in}{0.454405in}}{\pgfqpoint{4.650000in}{3.020000in}}%
\pgfusepath{clip}%
\pgfsetrectcap%
\pgfsetroundjoin%
\pgfsetlinewidth{0.401500pt}%
\definecolor{currentstroke}{rgb}{0.800000,0.800000,0.800000}%
\pgfsetstrokecolor{currentstroke}%
\pgfsetstrokeopacity{0.550000}%
\pgfsetdash{}{0pt}%
\pgfpathmoveto{\pgfqpoint{1.220874in}{0.696942in}}%
\pgfpathlineto{\pgfqpoint{5.870874in}{0.696942in}}%
\pgfusepath{stroke}%
\end{pgfscope}%
\begin{pgfscope}%
\pgfpathrectangle{\pgfqpoint{1.220874in}{0.454405in}}{\pgfqpoint{4.650000in}{3.020000in}}%
\pgfusepath{clip}%
\pgfsetrectcap%
\pgfsetroundjoin%
\pgfsetlinewidth{0.401500pt}%
\definecolor{currentstroke}{rgb}{0.800000,0.800000,0.800000}%
\pgfsetstrokecolor{currentstroke}%
\pgfsetstrokeopacity{0.550000}%
\pgfsetdash{}{0pt}%
\pgfpathmoveto{\pgfqpoint{1.220874in}{0.793194in}}%
\pgfpathlineto{\pgfqpoint{5.870874in}{0.793194in}}%
\pgfusepath{stroke}%
\end{pgfscope}%
\begin{pgfscope}%
\pgfpathrectangle{\pgfqpoint{1.220874in}{0.454405in}}{\pgfqpoint{4.650000in}{3.020000in}}%
\pgfusepath{clip}%
\pgfsetrectcap%
\pgfsetroundjoin%
\pgfsetlinewidth{0.401500pt}%
\definecolor{currentstroke}{rgb}{0.800000,0.800000,0.800000}%
\pgfsetstrokecolor{currentstroke}%
\pgfsetstrokeopacity{0.550000}%
\pgfsetdash{}{0pt}%
\pgfpathmoveto{\pgfqpoint{1.220874in}{0.878095in}}%
\pgfpathlineto{\pgfqpoint{5.870874in}{0.878095in}}%
\pgfusepath{stroke}%
\end{pgfscope}%
\begin{pgfscope}%
\pgfpathrectangle{\pgfqpoint{1.220874in}{0.454405in}}{\pgfqpoint{4.650000in}{3.020000in}}%
\pgfusepath{clip}%
\pgfsetrectcap%
\pgfsetroundjoin%
\pgfsetlinewidth{0.401500pt}%
\definecolor{currentstroke}{rgb}{0.800000,0.800000,0.800000}%
\pgfsetstrokecolor{currentstroke}%
\pgfsetstrokeopacity{0.550000}%
\pgfsetdash{}{0pt}%
\pgfpathmoveto{\pgfqpoint{1.220874in}{1.453678in}}%
\pgfpathlineto{\pgfqpoint{5.870874in}{1.453678in}}%
\pgfusepath{stroke}%
\end{pgfscope}%
\begin{pgfscope}%
\pgfpathrectangle{\pgfqpoint{1.220874in}{0.454405in}}{\pgfqpoint{4.650000in}{3.020000in}}%
\pgfusepath{clip}%
\pgfsetrectcap%
\pgfsetroundjoin%
\pgfsetlinewidth{0.401500pt}%
\definecolor{currentstroke}{rgb}{0.800000,0.800000,0.800000}%
\pgfsetstrokecolor{currentstroke}%
\pgfsetstrokeopacity{0.550000}%
\pgfsetdash{}{0pt}%
\pgfpathmoveto{\pgfqpoint{1.220874in}{1.745947in}}%
\pgfpathlineto{\pgfqpoint{5.870874in}{1.745947in}}%
\pgfusepath{stroke}%
\end{pgfscope}%
\begin{pgfscope}%
\pgfpathrectangle{\pgfqpoint{1.220874in}{0.454405in}}{\pgfqpoint{4.650000in}{3.020000in}}%
\pgfusepath{clip}%
\pgfsetrectcap%
\pgfsetroundjoin%
\pgfsetlinewidth{0.401500pt}%
\definecolor{currentstroke}{rgb}{0.800000,0.800000,0.800000}%
\pgfsetstrokecolor{currentstroke}%
\pgfsetstrokeopacity{0.550000}%
\pgfsetdash{}{0pt}%
\pgfpathmoveto{\pgfqpoint{1.220874in}{1.953315in}}%
\pgfpathlineto{\pgfqpoint{5.870874in}{1.953315in}}%
\pgfusepath{stroke}%
\end{pgfscope}%
\begin{pgfscope}%
\pgfpathrectangle{\pgfqpoint{1.220874in}{0.454405in}}{\pgfqpoint{4.650000in}{3.020000in}}%
\pgfusepath{clip}%
\pgfsetrectcap%
\pgfsetroundjoin%
\pgfsetlinewidth{0.401500pt}%
\definecolor{currentstroke}{rgb}{0.800000,0.800000,0.800000}%
\pgfsetstrokecolor{currentstroke}%
\pgfsetstrokeopacity{0.550000}%
\pgfsetdash{}{0pt}%
\pgfpathmoveto{\pgfqpoint{1.220874in}{2.114162in}}%
\pgfpathlineto{\pgfqpoint{5.870874in}{2.114162in}}%
\pgfusepath{stroke}%
\end{pgfscope}%
\begin{pgfscope}%
\pgfpathrectangle{\pgfqpoint{1.220874in}{0.454405in}}{\pgfqpoint{4.650000in}{3.020000in}}%
\pgfusepath{clip}%
\pgfsetrectcap%
\pgfsetroundjoin%
\pgfsetlinewidth{0.401500pt}%
\definecolor{currentstroke}{rgb}{0.800000,0.800000,0.800000}%
\pgfsetstrokecolor{currentstroke}%
\pgfsetstrokeopacity{0.550000}%
\pgfsetdash{}{0pt}%
\pgfpathmoveto{\pgfqpoint{1.220874in}{2.245583in}}%
\pgfpathlineto{\pgfqpoint{5.870874in}{2.245583in}}%
\pgfusepath{stroke}%
\end{pgfscope}%
\begin{pgfscope}%
\pgfpathrectangle{\pgfqpoint{1.220874in}{0.454405in}}{\pgfqpoint{4.650000in}{3.020000in}}%
\pgfusepath{clip}%
\pgfsetrectcap%
\pgfsetroundjoin%
\pgfsetlinewidth{0.401500pt}%
\definecolor{currentstroke}{rgb}{0.800000,0.800000,0.800000}%
\pgfsetstrokecolor{currentstroke}%
\pgfsetstrokeopacity{0.550000}%
\pgfsetdash{}{0pt}%
\pgfpathmoveto{\pgfqpoint{1.220874in}{2.356699in}}%
\pgfpathlineto{\pgfqpoint{5.870874in}{2.356699in}}%
\pgfusepath{stroke}%
\end{pgfscope}%
\begin{pgfscope}%
\pgfpathrectangle{\pgfqpoint{1.220874in}{0.454405in}}{\pgfqpoint{4.650000in}{3.020000in}}%
\pgfusepath{clip}%
\pgfsetrectcap%
\pgfsetroundjoin%
\pgfsetlinewidth{0.401500pt}%
\definecolor{currentstroke}{rgb}{0.800000,0.800000,0.800000}%
\pgfsetstrokecolor{currentstroke}%
\pgfsetstrokeopacity{0.550000}%
\pgfsetdash{}{0pt}%
\pgfpathmoveto{\pgfqpoint{1.220874in}{2.452951in}}%
\pgfpathlineto{\pgfqpoint{5.870874in}{2.452951in}}%
\pgfusepath{stroke}%
\end{pgfscope}%
\begin{pgfscope}%
\pgfpathrectangle{\pgfqpoint{1.220874in}{0.454405in}}{\pgfqpoint{4.650000in}{3.020000in}}%
\pgfusepath{clip}%
\pgfsetrectcap%
\pgfsetroundjoin%
\pgfsetlinewidth{0.401500pt}%
\definecolor{currentstroke}{rgb}{0.800000,0.800000,0.800000}%
\pgfsetstrokecolor{currentstroke}%
\pgfsetstrokeopacity{0.550000}%
\pgfsetdash{}{0pt}%
\pgfpathmoveto{\pgfqpoint{1.220874in}{2.537852in}}%
\pgfpathlineto{\pgfqpoint{5.870874in}{2.537852in}}%
\pgfusepath{stroke}%
\end{pgfscope}%
\begin{pgfscope}%
\pgfpathrectangle{\pgfqpoint{1.220874in}{0.454405in}}{\pgfqpoint{4.650000in}{3.020000in}}%
\pgfusepath{clip}%
\pgfsetrectcap%
\pgfsetroundjoin%
\pgfsetlinewidth{0.401500pt}%
\definecolor{currentstroke}{rgb}{0.800000,0.800000,0.800000}%
\pgfsetstrokecolor{currentstroke}%
\pgfsetstrokeopacity{0.550000}%
\pgfsetdash{}{0pt}%
\pgfpathmoveto{\pgfqpoint{1.220874in}{3.113435in}}%
\pgfpathlineto{\pgfqpoint{5.870874in}{3.113435in}}%
\pgfusepath{stroke}%
\end{pgfscope}%
\begin{pgfscope}%
\pgfpathrectangle{\pgfqpoint{1.220874in}{0.454405in}}{\pgfqpoint{4.650000in}{3.020000in}}%
\pgfusepath{clip}%
\pgfsetrectcap%
\pgfsetroundjoin%
\pgfsetlinewidth{0.401500pt}%
\definecolor{currentstroke}{rgb}{0.800000,0.800000,0.800000}%
\pgfsetstrokecolor{currentstroke}%
\pgfsetstrokeopacity{0.550000}%
\pgfsetdash{}{0pt}%
\pgfpathmoveto{\pgfqpoint{1.220874in}{3.405703in}}%
\pgfpathlineto{\pgfqpoint{5.870874in}{3.405703in}}%
\pgfusepath{stroke}%
\end{pgfscope}%
\begin{pgfscope}%
\definecolor{textcolor}{rgb}{0.000000,0.000000,0.000000}%
\pgfsetstrokecolor{textcolor}%
\pgfsetfillcolor{textcolor}%
\pgftext[x=0.357183in,y=1.964405in,,bottom,rotate=90.000000]{\color{textcolor}\rmfamily\fontsize{26.400000}{31.680000}\selectfont relative error}%
\end{pgfscope}%
\begin{pgfscope}%
\pgfpathrectangle{\pgfqpoint{1.220874in}{0.454405in}}{\pgfqpoint{4.650000in}{3.020000in}}%
\pgfusepath{clip}%
\pgfsetbuttcap%
\pgfsetmiterjoin%
\definecolor{currentfill}{rgb}{0.347059,0.458824,0.641176}%
\pgfsetfillcolor{currentfill}%
\pgfsetlinewidth{0.803000pt}%
\definecolor{currentstroke}{rgb}{0.298039,0.298039,0.298039}%
\pgfsetstrokecolor{currentstroke}%
\pgfsetdash{}{0pt}%
\pgfpathmoveto{\pgfqpoint{1.337124in}{2.340771in}}%
\pgfpathlineto{\pgfqpoint{2.267124in}{2.340771in}}%
\pgfpathlineto{\pgfqpoint{2.267124in}{2.345798in}}%
\pgfpathlineto{\pgfqpoint{1.337124in}{2.345798in}}%
\pgfpathlineto{\pgfqpoint{1.337124in}{2.340771in}}%
\pgfpathclose%
\pgfusepath{stroke,fill}%
\end{pgfscope}%
\begin{pgfscope}%
\pgfpathrectangle{\pgfqpoint{1.220874in}{0.454405in}}{\pgfqpoint{4.650000in}{3.020000in}}%
\pgfusepath{clip}%
\pgfsetbuttcap%
\pgfsetmiterjoin%
\definecolor{currentfill}{rgb}{0.798529,0.536765,0.389706}%
\pgfsetfillcolor{currentfill}%
\pgfsetlinewidth{0.803000pt}%
\definecolor{currentstroke}{rgb}{0.298039,0.298039,0.298039}%
\pgfsetstrokecolor{currentstroke}%
\pgfsetdash{}{0pt}%
\pgfpathmoveto{\pgfqpoint{2.499624in}{2.215192in}}%
\pgfpathlineto{\pgfqpoint{3.429624in}{2.215192in}}%
\pgfpathlineto{\pgfqpoint{3.429624in}{2.281678in}}%
\pgfpathlineto{\pgfqpoint{2.499624in}{2.281678in}}%
\pgfpathlineto{\pgfqpoint{2.499624in}{2.215192in}}%
\pgfpathclose%
\pgfusepath{stroke,fill}%
\end{pgfscope}%
\begin{pgfscope}%
\pgfpathrectangle{\pgfqpoint{1.220874in}{0.454405in}}{\pgfqpoint{4.650000in}{3.020000in}}%
\pgfusepath{clip}%
\pgfsetbuttcap%
\pgfsetmiterjoin%
\definecolor{currentfill}{rgb}{0.374020,0.618137,0.429902}%
\pgfsetfillcolor{currentfill}%
\pgfsetlinewidth{0.803000pt}%
\definecolor{currentstroke}{rgb}{0.298039,0.298039,0.298039}%
\pgfsetstrokecolor{currentstroke}%
\pgfsetdash{}{0pt}%
\pgfpathmoveto{\pgfqpoint{3.662124in}{2.463481in}}%
\pgfpathlineto{\pgfqpoint{4.592124in}{2.463481in}}%
\pgfpathlineto{\pgfqpoint{4.592124in}{2.575324in}}%
\pgfpathlineto{\pgfqpoint{3.662124in}{2.575324in}}%
\pgfpathlineto{\pgfqpoint{3.662124in}{2.463481in}}%
\pgfpathclose%
\pgfusepath{stroke,fill}%
\end{pgfscope}%
\begin{pgfscope}%
\pgfpathrectangle{\pgfqpoint{1.220874in}{0.454405in}}{\pgfqpoint{4.650000in}{3.020000in}}%
\pgfusepath{clip}%
\pgfsetbuttcap%
\pgfsetmiterjoin%
\definecolor{currentfill}{rgb}{0.710784,0.363725,0.375490}%
\pgfsetfillcolor{currentfill}%
\pgfsetlinewidth{0.803000pt}%
\definecolor{currentstroke}{rgb}{0.298039,0.298039,0.298039}%
\pgfsetstrokecolor{currentstroke}%
\pgfsetdash{}{0pt}%
\pgfpathmoveto{\pgfqpoint{4.824624in}{2.945817in}}%
\pgfpathlineto{\pgfqpoint{5.754624in}{2.945817in}}%
\pgfpathlineto{\pgfqpoint{5.754624in}{3.048085in}}%
\pgfpathlineto{\pgfqpoint{4.824624in}{3.048085in}}%
\pgfpathlineto{\pgfqpoint{4.824624in}{2.945817in}}%
\pgfpathclose%
\pgfusepath{stroke,fill}%
\end{pgfscope}%
\begin{pgfscope}%
\pgfsetrectcap%
\pgfsetmiterjoin%
\pgfsetlinewidth{1.254687pt}%
\definecolor{currentstroke}{rgb}{0.800000,0.800000,0.800000}%
\pgfsetstrokecolor{currentstroke}%
\pgfsetdash{}{0pt}%
\pgfpathmoveto{\pgfqpoint{1.220874in}{0.454405in}}%
\pgfpathlineto{\pgfqpoint{1.220874in}{3.474405in}}%
\pgfusepath{stroke}%
\end{pgfscope}%
\begin{pgfscope}%
\pgfsetrectcap%
\pgfsetmiterjoin%
\pgfsetlinewidth{1.254687pt}%
\definecolor{currentstroke}{rgb}{0.800000,0.800000,0.800000}%
\pgfsetstrokecolor{currentstroke}%
\pgfsetdash{}{0pt}%
\pgfpathmoveto{\pgfqpoint{5.870874in}{0.454405in}}%
\pgfpathlineto{\pgfqpoint{5.870874in}{3.474405in}}%
\pgfusepath{stroke}%
\end{pgfscope}%
\begin{pgfscope}%
\pgfsetrectcap%
\pgfsetmiterjoin%
\pgfsetlinewidth{1.254687pt}%
\definecolor{currentstroke}{rgb}{0.800000,0.800000,0.800000}%
\pgfsetstrokecolor{currentstroke}%
\pgfsetdash{}{0pt}%
\pgfpathmoveto{\pgfqpoint{1.220874in}{0.454405in}}%
\pgfpathlineto{\pgfqpoint{5.870874in}{0.454405in}}%
\pgfusepath{stroke}%
\end{pgfscope}%
\begin{pgfscope}%
\pgfsetrectcap%
\pgfsetmiterjoin%
\pgfsetlinewidth{1.254687pt}%
\definecolor{currentstroke}{rgb}{0.800000,0.800000,0.800000}%
\pgfsetstrokecolor{currentstroke}%
\pgfsetdash{}{0pt}%
\pgfpathmoveto{\pgfqpoint{1.220874in}{3.474405in}}%
\pgfpathlineto{\pgfqpoint{5.870874in}{3.474405in}}%
\pgfusepath{stroke}%
\end{pgfscope}%
\begin{pgfscope}%
\pgfpathrectangle{\pgfqpoint{1.220874in}{0.454405in}}{\pgfqpoint{4.650000in}{3.020000in}}%
\pgfusepath{clip}%
\pgfsetrectcap%
\pgfsetroundjoin%
\pgfsetlinewidth{0.803000pt}%
\definecolor{currentstroke}{rgb}{0.298039,0.298039,0.298039}%
\pgfsetstrokecolor{currentstroke}%
\pgfsetdash{}{0pt}%
\pgfpathmoveto{\pgfqpoint{1.802124in}{2.340771in}}%
\pgfpathlineto{\pgfqpoint{1.802124in}{2.336172in}}%
\pgfusepath{stroke}%
\end{pgfscope}%
\begin{pgfscope}%
\pgfpathrectangle{\pgfqpoint{1.220874in}{0.454405in}}{\pgfqpoint{4.650000in}{3.020000in}}%
\pgfusepath{clip}%
\pgfsetrectcap%
\pgfsetroundjoin%
\pgfsetlinewidth{0.803000pt}%
\definecolor{currentstroke}{rgb}{0.298039,0.298039,0.298039}%
\pgfsetstrokecolor{currentstroke}%
\pgfsetdash{}{0pt}%
\pgfpathmoveto{\pgfqpoint{1.802124in}{2.345798in}}%
\pgfpathlineto{\pgfqpoint{1.802124in}{2.345798in}}%
\pgfusepath{stroke}%
\end{pgfscope}%
\begin{pgfscope}%
\pgfpathrectangle{\pgfqpoint{1.220874in}{0.454405in}}{\pgfqpoint{4.650000in}{3.020000in}}%
\pgfusepath{clip}%
\pgfsetrectcap%
\pgfsetroundjoin%
\pgfsetlinewidth{0.803000pt}%
\definecolor{currentstroke}{rgb}{0.298039,0.298039,0.298039}%
\pgfsetstrokecolor{currentstroke}%
\pgfsetdash{}{0pt}%
\pgfpathmoveto{\pgfqpoint{1.569624in}{2.336172in}}%
\pgfpathlineto{\pgfqpoint{2.034624in}{2.336172in}}%
\pgfusepath{stroke}%
\end{pgfscope}%
\begin{pgfscope}%
\pgfpathrectangle{\pgfqpoint{1.220874in}{0.454405in}}{\pgfqpoint{4.650000in}{3.020000in}}%
\pgfusepath{clip}%
\pgfsetrectcap%
\pgfsetroundjoin%
\pgfsetlinewidth{0.803000pt}%
\definecolor{currentstroke}{rgb}{0.298039,0.298039,0.298039}%
\pgfsetstrokecolor{currentstroke}%
\pgfsetdash{}{0pt}%
\pgfpathmoveto{\pgfqpoint{1.569624in}{2.345798in}}%
\pgfpathlineto{\pgfqpoint{2.034624in}{2.345798in}}%
\pgfusepath{stroke}%
\end{pgfscope}%
\begin{pgfscope}%
\pgfpathrectangle{\pgfqpoint{1.220874in}{0.454405in}}{\pgfqpoint{4.650000in}{3.020000in}}%
\pgfusepath{clip}%
\pgfsetrectcap%
\pgfsetroundjoin%
\pgfsetlinewidth{0.803000pt}%
\definecolor{currentstroke}{rgb}{0.298039,0.298039,0.298039}%
\pgfsetstrokecolor{currentstroke}%
\pgfsetdash{}{0pt}%
\pgfpathmoveto{\pgfqpoint{1.337124in}{2.342298in}}%
\pgfpathlineto{\pgfqpoint{2.267124in}{2.342298in}}%
\pgfusepath{stroke}%
\end{pgfscope}%
\begin{pgfscope}%
\pgfpathrectangle{\pgfqpoint{1.220874in}{0.454405in}}{\pgfqpoint{4.650000in}{3.020000in}}%
\pgfusepath{clip}%
\pgfsetbuttcap%
\pgfsetmiterjoin%
\definecolor{currentfill}{rgb}{0.298039,0.298039,0.298039}%
\pgfsetfillcolor{currentfill}%
\pgfsetlinewidth{1.003750pt}%
\definecolor{currentstroke}{rgb}{0.298039,0.298039,0.298039}%
\pgfsetstrokecolor{currentstroke}%
\pgfsetdash{}{0pt}%
\pgfsys@defobject{currentmarker}{\pgfqpoint{-0.029463in}{-0.049105in}}{\pgfqpoint{0.029463in}{0.049105in}}{%
\pgfpathmoveto{\pgfqpoint{0.000000in}{-0.049105in}}%
\pgfpathlineto{\pgfqpoint{0.029463in}{0.000000in}}%
\pgfpathlineto{\pgfqpoint{0.000000in}{0.049105in}}%
\pgfpathlineto{\pgfqpoint{-0.029463in}{0.000000in}}%
\pgfpathclose%
\pgfusepath{stroke,fill}%
}%
\begin{pgfscope}%
\pgfsys@transformshift{1.802124in}{2.317626in}%
\pgfsys@useobject{currentmarker}{}%
\end{pgfscope}%
\begin{pgfscope}%
\pgfsys@transformshift{1.802124in}{2.318536in}%
\pgfsys@useobject{currentmarker}{}%
\end{pgfscope}%
\begin{pgfscope}%
\pgfsys@transformshift{1.802124in}{2.318536in}%
\pgfsys@useobject{currentmarker}{}%
\end{pgfscope}%
\begin{pgfscope}%
\pgfsys@transformshift{1.802124in}{2.356195in}%
\pgfsys@useobject{currentmarker}{}%
\end{pgfscope}%
\begin{pgfscope}%
\pgfsys@transformshift{1.802124in}{2.356195in}%
\pgfsys@useobject{currentmarker}{}%
\end{pgfscope}%
\begin{pgfscope}%
\pgfsys@transformshift{1.802124in}{2.358194in}%
\pgfsys@useobject{currentmarker}{}%
\end{pgfscope}%
\begin{pgfscope}%
\pgfsys@transformshift{1.802124in}{2.415939in}%
\pgfsys@useobject{currentmarker}{}%
\end{pgfscope}%
\end{pgfscope}%
\begin{pgfscope}%
\pgfpathrectangle{\pgfqpoint{1.220874in}{0.454405in}}{\pgfqpoint{4.650000in}{3.020000in}}%
\pgfusepath{clip}%
\pgfsetrectcap%
\pgfsetroundjoin%
\pgfsetlinewidth{0.803000pt}%
\definecolor{currentstroke}{rgb}{0.298039,0.298039,0.298039}%
\pgfsetstrokecolor{currentstroke}%
\pgfsetdash{}{0pt}%
\pgfpathmoveto{\pgfqpoint{2.964624in}{2.215192in}}%
\pgfpathlineto{\pgfqpoint{2.964624in}{2.180698in}}%
\pgfusepath{stroke}%
\end{pgfscope}%
\begin{pgfscope}%
\pgfpathrectangle{\pgfqpoint{1.220874in}{0.454405in}}{\pgfqpoint{4.650000in}{3.020000in}}%
\pgfusepath{clip}%
\pgfsetrectcap%
\pgfsetroundjoin%
\pgfsetlinewidth{0.803000pt}%
\definecolor{currentstroke}{rgb}{0.298039,0.298039,0.298039}%
\pgfsetstrokecolor{currentstroke}%
\pgfsetdash{}{0pt}%
\pgfpathmoveto{\pgfqpoint{2.964624in}{2.281678in}}%
\pgfpathlineto{\pgfqpoint{2.964624in}{2.323203in}}%
\pgfusepath{stroke}%
\end{pgfscope}%
\begin{pgfscope}%
\pgfpathrectangle{\pgfqpoint{1.220874in}{0.454405in}}{\pgfqpoint{4.650000in}{3.020000in}}%
\pgfusepath{clip}%
\pgfsetrectcap%
\pgfsetroundjoin%
\pgfsetlinewidth{0.803000pt}%
\definecolor{currentstroke}{rgb}{0.298039,0.298039,0.298039}%
\pgfsetstrokecolor{currentstroke}%
\pgfsetdash{}{0pt}%
\pgfpathmoveto{\pgfqpoint{2.732124in}{2.180698in}}%
\pgfpathlineto{\pgfqpoint{3.197124in}{2.180698in}}%
\pgfusepath{stroke}%
\end{pgfscope}%
\begin{pgfscope}%
\pgfpathrectangle{\pgfqpoint{1.220874in}{0.454405in}}{\pgfqpoint{4.650000in}{3.020000in}}%
\pgfusepath{clip}%
\pgfsetrectcap%
\pgfsetroundjoin%
\pgfsetlinewidth{0.803000pt}%
\definecolor{currentstroke}{rgb}{0.298039,0.298039,0.298039}%
\pgfsetstrokecolor{currentstroke}%
\pgfsetdash{}{0pt}%
\pgfpathmoveto{\pgfqpoint{2.732124in}{2.323203in}}%
\pgfpathlineto{\pgfqpoint{3.197124in}{2.323203in}}%
\pgfusepath{stroke}%
\end{pgfscope}%
\begin{pgfscope}%
\pgfpathrectangle{\pgfqpoint{1.220874in}{0.454405in}}{\pgfqpoint{4.650000in}{3.020000in}}%
\pgfusepath{clip}%
\pgfsetrectcap%
\pgfsetroundjoin%
\pgfsetlinewidth{0.803000pt}%
\definecolor{currentstroke}{rgb}{0.298039,0.298039,0.298039}%
\pgfsetstrokecolor{currentstroke}%
\pgfsetdash{}{0pt}%
\pgfpathmoveto{\pgfqpoint{2.499624in}{2.268264in}}%
\pgfpathlineto{\pgfqpoint{3.429624in}{2.268264in}}%
\pgfusepath{stroke}%
\end{pgfscope}%
\begin{pgfscope}%
\pgfpathrectangle{\pgfqpoint{1.220874in}{0.454405in}}{\pgfqpoint{4.650000in}{3.020000in}}%
\pgfusepath{clip}%
\pgfsetrectcap%
\pgfsetroundjoin%
\pgfsetlinewidth{0.803000pt}%
\definecolor{currentstroke}{rgb}{0.298039,0.298039,0.298039}%
\pgfsetstrokecolor{currentstroke}%
\pgfsetdash{}{0pt}%
\pgfpathmoveto{\pgfqpoint{4.127124in}{2.463481in}}%
\pgfpathlineto{\pgfqpoint{4.127124in}{2.340423in}}%
\pgfusepath{stroke}%
\end{pgfscope}%
\begin{pgfscope}%
\pgfpathrectangle{\pgfqpoint{1.220874in}{0.454405in}}{\pgfqpoint{4.650000in}{3.020000in}}%
\pgfusepath{clip}%
\pgfsetrectcap%
\pgfsetroundjoin%
\pgfsetlinewidth{0.803000pt}%
\definecolor{currentstroke}{rgb}{0.298039,0.298039,0.298039}%
\pgfsetstrokecolor{currentstroke}%
\pgfsetdash{}{0pt}%
\pgfpathmoveto{\pgfqpoint{4.127124in}{2.575324in}}%
\pgfpathlineto{\pgfqpoint{4.127124in}{2.679188in}}%
\pgfusepath{stroke}%
\end{pgfscope}%
\begin{pgfscope}%
\pgfpathrectangle{\pgfqpoint{1.220874in}{0.454405in}}{\pgfqpoint{4.650000in}{3.020000in}}%
\pgfusepath{clip}%
\pgfsetrectcap%
\pgfsetroundjoin%
\pgfsetlinewidth{0.803000pt}%
\definecolor{currentstroke}{rgb}{0.298039,0.298039,0.298039}%
\pgfsetstrokecolor{currentstroke}%
\pgfsetdash{}{0pt}%
\pgfpathmoveto{\pgfqpoint{3.894624in}{2.340423in}}%
\pgfpathlineto{\pgfqpoint{4.359624in}{2.340423in}}%
\pgfusepath{stroke}%
\end{pgfscope}%
\begin{pgfscope}%
\pgfpathrectangle{\pgfqpoint{1.220874in}{0.454405in}}{\pgfqpoint{4.650000in}{3.020000in}}%
\pgfusepath{clip}%
\pgfsetrectcap%
\pgfsetroundjoin%
\pgfsetlinewidth{0.803000pt}%
\definecolor{currentstroke}{rgb}{0.298039,0.298039,0.298039}%
\pgfsetstrokecolor{currentstroke}%
\pgfsetdash{}{0pt}%
\pgfpathmoveto{\pgfqpoint{3.894624in}{2.679188in}}%
\pgfpathlineto{\pgfqpoint{4.359624in}{2.679188in}}%
\pgfusepath{stroke}%
\end{pgfscope}%
\begin{pgfscope}%
\pgfpathrectangle{\pgfqpoint{1.220874in}{0.454405in}}{\pgfqpoint{4.650000in}{3.020000in}}%
\pgfusepath{clip}%
\pgfsetrectcap%
\pgfsetroundjoin%
\pgfsetlinewidth{0.803000pt}%
\definecolor{currentstroke}{rgb}{0.298039,0.298039,0.298039}%
\pgfsetstrokecolor{currentstroke}%
\pgfsetdash{}{0pt}%
\pgfpathmoveto{\pgfqpoint{3.662124in}{2.513899in}}%
\pgfpathlineto{\pgfqpoint{4.592124in}{2.513899in}}%
\pgfusepath{stroke}%
\end{pgfscope}%
\begin{pgfscope}%
\pgfpathrectangle{\pgfqpoint{1.220874in}{0.454405in}}{\pgfqpoint{4.650000in}{3.020000in}}%
\pgfusepath{clip}%
\pgfsetrectcap%
\pgfsetroundjoin%
\pgfsetlinewidth{0.803000pt}%
\definecolor{currentstroke}{rgb}{0.298039,0.298039,0.298039}%
\pgfsetstrokecolor{currentstroke}%
\pgfsetdash{}{0pt}%
\pgfpathmoveto{\pgfqpoint{5.289624in}{2.945817in}}%
\pgfpathlineto{\pgfqpoint{5.289624in}{2.872864in}}%
\pgfusepath{stroke}%
\end{pgfscope}%
\begin{pgfscope}%
\pgfpathrectangle{\pgfqpoint{1.220874in}{0.454405in}}{\pgfqpoint{4.650000in}{3.020000in}}%
\pgfusepath{clip}%
\pgfsetrectcap%
\pgfsetroundjoin%
\pgfsetlinewidth{0.803000pt}%
\definecolor{currentstroke}{rgb}{0.298039,0.298039,0.298039}%
\pgfsetstrokecolor{currentstroke}%
\pgfsetdash{}{0pt}%
\pgfpathmoveto{\pgfqpoint{5.289624in}{3.048085in}}%
\pgfpathlineto{\pgfqpoint{5.289624in}{3.130755in}}%
\pgfusepath{stroke}%
\end{pgfscope}%
\begin{pgfscope}%
\pgfpathrectangle{\pgfqpoint{1.220874in}{0.454405in}}{\pgfqpoint{4.650000in}{3.020000in}}%
\pgfusepath{clip}%
\pgfsetrectcap%
\pgfsetroundjoin%
\pgfsetlinewidth{0.803000pt}%
\definecolor{currentstroke}{rgb}{0.298039,0.298039,0.298039}%
\pgfsetstrokecolor{currentstroke}%
\pgfsetdash{}{0pt}%
\pgfpathmoveto{\pgfqpoint{5.057124in}{2.872864in}}%
\pgfpathlineto{\pgfqpoint{5.522124in}{2.872864in}}%
\pgfusepath{stroke}%
\end{pgfscope}%
\begin{pgfscope}%
\pgfpathrectangle{\pgfqpoint{1.220874in}{0.454405in}}{\pgfqpoint{4.650000in}{3.020000in}}%
\pgfusepath{clip}%
\pgfsetrectcap%
\pgfsetroundjoin%
\pgfsetlinewidth{0.803000pt}%
\definecolor{currentstroke}{rgb}{0.298039,0.298039,0.298039}%
\pgfsetstrokecolor{currentstroke}%
\pgfsetdash{}{0pt}%
\pgfpathmoveto{\pgfqpoint{5.057124in}{3.130755in}}%
\pgfpathlineto{\pgfqpoint{5.522124in}{3.130755in}}%
\pgfusepath{stroke}%
\end{pgfscope}%
\begin{pgfscope}%
\pgfpathrectangle{\pgfqpoint{1.220874in}{0.454405in}}{\pgfqpoint{4.650000in}{3.020000in}}%
\pgfusepath{clip}%
\pgfsetrectcap%
\pgfsetroundjoin%
\pgfsetlinewidth{0.803000pt}%
\definecolor{currentstroke}{rgb}{0.298039,0.298039,0.298039}%
\pgfsetstrokecolor{currentstroke}%
\pgfsetdash{}{0pt}%
\pgfpathmoveto{\pgfqpoint{4.824624in}{3.007072in}}%
\pgfpathlineto{\pgfqpoint{5.754624in}{3.007072in}}%
\pgfusepath{stroke}%
\end{pgfscope}%
\begin{pgfscope}%
\pgfpathrectangle{\pgfqpoint{1.220874in}{0.454405in}}{\pgfqpoint{4.650000in}{3.020000in}}%
\pgfusepath{clip}%
\pgfsetbuttcap%
\pgfsetmiterjoin%
\definecolor{currentfill}{rgb}{0.298039,0.298039,0.298039}%
\pgfsetfillcolor{currentfill}%
\pgfsetlinewidth{1.003750pt}%
\definecolor{currentstroke}{rgb}{0.298039,0.298039,0.298039}%
\pgfsetstrokecolor{currentstroke}%
\pgfsetdash{}{0pt}%
\pgfsys@defobject{currentmarker}{\pgfqpoint{-0.029463in}{-0.049105in}}{\pgfqpoint{0.029463in}{0.049105in}}{%
\pgfpathmoveto{\pgfqpoint{0.000000in}{-0.049105in}}%
\pgfpathlineto{\pgfqpoint{0.029463in}{0.000000in}}%
\pgfpathlineto{\pgfqpoint{0.000000in}{0.049105in}}%
\pgfpathlineto{\pgfqpoint{-0.029463in}{0.000000in}}%
\pgfpathclose%
\pgfusepath{stroke,fill}%
}%
\begin{pgfscope}%
\pgfsys@transformshift{5.289624in}{3.245642in}%
\pgfsys@useobject{currentmarker}{}%
\end{pgfscope}%
\end{pgfscope}%
\end{pgfpicture}%
\makeatother%
\endgroup%

%% file: fig/boxplot_circle-1000-4d_res.pgf
%% Creator: Matplotlib, PGF backend
%%
%% To include the figure in your LaTeX document, write
%%   \input{<filename>.pgf}
%%
%% Make sure the required packages are loaded in your preamble
%%   \usepackage{pgf}
%%
%% Figures using additional raster images can only be included by \input if
%% they are in the same directory as the main LaTeX file. For loading figures
%% from other directories you can use the `import` package
%% and then include the figures with
%%   \import{<path to file>}{<filename>.pgf}
%%
%% Matplotlib used the following preamble
%%
\begingroup%
\makeatletter%
\begin{pgfpicture}%
\pgfpathrectangle{\pgfpointorigin}{\pgfqpoint{4.888889in}{3.524405in}}%
\pgfusepath{use as bounding box, clip}%
\begin{pgfscope}%
\pgfsetbuttcap%
\pgfsetmiterjoin%
\definecolor{currentfill}{rgb}{1.000000,1.000000,1.000000}%
\pgfsetfillcolor{currentfill}%
\pgfsetlinewidth{0.000000pt}%
\definecolor{currentstroke}{rgb}{1.000000,1.000000,1.000000}%
\pgfsetstrokecolor{currentstroke}%
\pgfsetdash{}{0pt}%
\pgfpathmoveto{\pgfqpoint{0.000000in}{0.000000in}}%
\pgfpathlineto{\pgfqpoint{4.888889in}{0.000000in}}%
\pgfpathlineto{\pgfqpoint{4.888889in}{3.524405in}}%
\pgfpathlineto{\pgfqpoint{0.000000in}{3.524405in}}%
\pgfpathclose%
\pgfusepath{fill}%
\end{pgfscope}%
\begin{pgfscope}%
\pgfsetbuttcap%
\pgfsetmiterjoin%
\definecolor{currentfill}{rgb}{1.000000,1.000000,1.000000}%
\pgfsetfillcolor{currentfill}%
\pgfsetlinewidth{0.000000pt}%
\definecolor{currentstroke}{rgb}{0.000000,0.000000,0.000000}%
\pgfsetstrokecolor{currentstroke}%
\pgfsetstrokeopacity{0.000000}%
\pgfsetdash{}{0pt}%
\pgfpathmoveto{\pgfqpoint{0.188889in}{0.454405in}}%
\pgfpathlineto{\pgfqpoint{4.838889in}{0.454405in}}%
\pgfpathlineto{\pgfqpoint{4.838889in}{3.474405in}}%
\pgfpathlineto{\pgfqpoint{0.188889in}{3.474405in}}%
\pgfpathclose%
\pgfusepath{fill}%
\end{pgfscope}%
\begin{pgfscope}%
\definecolor{textcolor}{rgb}{0.000000,0.000000,0.000000}%
\pgfsetstrokecolor{textcolor}%
\pgfsetfillcolor{textcolor}%
\pgftext[x=0.770139in,y=0.357183in,,top]{\color{textcolor}\rmfamily\fontsize{24.200000}{29.040000}\selectfont \textsc{w-l}}%
\end{pgfscope}%
\begin{pgfscope}%
\definecolor{textcolor}{rgb}{0.000000,0.000000,0.000000}%
\pgfsetstrokecolor{textcolor}%
\pgfsetfillcolor{textcolor}%
\pgftext[x=1.932639in,y=0.357183in,,top]{\color{textcolor}\rmfamily\fontsize{24.200000}{29.040000}\selectfont \textsc{w-m}}%
\end{pgfscope}%
\begin{pgfscope}%
\definecolor{textcolor}{rgb}{0.000000,0.000000,0.000000}%
\pgfsetstrokecolor{textcolor}%
\pgfsetfillcolor{textcolor}%
\pgftext[x=3.095139in,y=0.357183in,,top]{\color{textcolor}\rmfamily\fontsize{24.200000}{29.040000}\selectfont \textsc{dcv}}%
\end{pgfscope}%
\begin{pgfscope}%
\definecolor{textcolor}{rgb}{0.000000,0.000000,0.000000}%
\pgfsetstrokecolor{textcolor}%
\pgfsetfillcolor{textcolor}%
\pgftext[x=4.257639in,y=0.357183in,,top]{\color{textcolor}\rmfamily\fontsize{24.200000}{29.040000}\selectfont \textsc{mc}}%
\end{pgfscope}%
\begin{pgfscope}%
\pgfpathrectangle{\pgfqpoint{0.188889in}{0.454405in}}{\pgfqpoint{4.650000in}{3.020000in}}%
\pgfusepath{clip}%
\pgfsetrectcap%
\pgfsetroundjoin%
\pgfsetlinewidth{2.509375pt}%
\definecolor{currentstroke}{rgb}{0.800000,0.800000,0.800000}%
\pgfsetstrokecolor{currentstroke}%
\pgfsetdash{}{0pt}%
\pgfpathmoveto{\pgfqpoint{0.188889in}{0.954042in}}%
\pgfpathlineto{\pgfqpoint{4.838889in}{0.954042in}}%
\pgfusepath{stroke}%
\end{pgfscope}%
\begin{pgfscope}%
\pgfsetbuttcap%
\pgfsetroundjoin%
\definecolor{currentfill}{rgb}{0.800000,0.800000,0.800000}%
\pgfsetfillcolor{currentfill}%
\pgfsetlinewidth{1.003750pt}%
\definecolor{currentstroke}{rgb}{0.800000,0.800000,0.800000}%
\pgfsetstrokecolor{currentstroke}%
\pgfsetdash{}{0pt}%
\pgfsys@defobject{currentmarker}{\pgfqpoint{-0.138889in}{0.000000in}}{\pgfqpoint{0.000000in}{0.000000in}}{%
\pgfpathmoveto{\pgfqpoint{0.000000in}{0.000000in}}%
\pgfpathlineto{\pgfqpoint{-0.138889in}{0.000000in}}%
\pgfusepath{stroke,fill}%
}%
\begin{pgfscope}%
\pgfsys@transformshift{0.188889in}{0.954042in}%
\pgfsys@useobject{currentmarker}{}%
\end{pgfscope}%
\end{pgfscope}%
\begin{pgfscope}%
\pgfpathrectangle{\pgfqpoint{0.188889in}{0.454405in}}{\pgfqpoint{4.650000in}{3.020000in}}%
\pgfusepath{clip}%
\pgfsetrectcap%
\pgfsetroundjoin%
\pgfsetlinewidth{2.509375pt}%
\definecolor{currentstroke}{rgb}{0.611765,0.611765,0.611765}%
\pgfsetstrokecolor{currentstroke}%
\pgfsetdash{}{0pt}%
\pgfpathmoveto{\pgfqpoint{0.188889in}{2.613798in}}%
\pgfpathlineto{\pgfqpoint{4.838889in}{2.613798in}}%
\pgfusepath{stroke}%
\end{pgfscope}%
\begin{pgfscope}%
\pgfsetbuttcap%
\pgfsetroundjoin%
\definecolor{currentfill}{rgb}{0.800000,0.800000,0.800000}%
\pgfsetfillcolor{currentfill}%
\pgfsetlinewidth{1.003750pt}%
\definecolor{currentstroke}{rgb}{0.800000,0.800000,0.800000}%
\pgfsetstrokecolor{currentstroke}%
\pgfsetdash{}{0pt}%
\pgfsys@defobject{currentmarker}{\pgfqpoint{-0.138889in}{0.000000in}}{\pgfqpoint{0.000000in}{0.000000in}}{%
\pgfpathmoveto{\pgfqpoint{0.000000in}{0.000000in}}%
\pgfpathlineto{\pgfqpoint{-0.138889in}{0.000000in}}%
\pgfusepath{stroke,fill}%
}%
\begin{pgfscope}%
\pgfsys@transformshift{0.188889in}{2.613798in}%
\pgfsys@useobject{currentmarker}{}%
\end{pgfscope}%
\end{pgfscope}%
\begin{pgfscope}%
\pgfpathrectangle{\pgfqpoint{0.188889in}{0.454405in}}{\pgfqpoint{4.650000in}{3.020000in}}%
\pgfusepath{clip}%
\pgfsetrectcap%
\pgfsetroundjoin%
\pgfsetlinewidth{0.401500pt}%
\definecolor{currentstroke}{rgb}{0.800000,0.800000,0.800000}%
\pgfsetstrokecolor{currentstroke}%
\pgfsetstrokeopacity{0.550000}%
\pgfsetdash{}{0pt}%
\pgfpathmoveto{\pgfqpoint{0.188889in}{0.454405in}}%
\pgfpathlineto{\pgfqpoint{4.838889in}{0.454405in}}%
\pgfusepath{stroke}%
\end{pgfscope}%
\begin{pgfscope}%
\pgfpathrectangle{\pgfqpoint{0.188889in}{0.454405in}}{\pgfqpoint{4.650000in}{3.020000in}}%
\pgfusepath{clip}%
\pgfsetrectcap%
\pgfsetroundjoin%
\pgfsetlinewidth{0.401500pt}%
\definecolor{currentstroke}{rgb}{0.800000,0.800000,0.800000}%
\pgfsetstrokecolor{currentstroke}%
\pgfsetstrokeopacity{0.550000}%
\pgfsetdash{}{0pt}%
\pgfpathmoveto{\pgfqpoint{0.188889in}{0.585827in}}%
\pgfpathlineto{\pgfqpoint{4.838889in}{0.585827in}}%
\pgfusepath{stroke}%
\end{pgfscope}%
\begin{pgfscope}%
\pgfpathrectangle{\pgfqpoint{0.188889in}{0.454405in}}{\pgfqpoint{4.650000in}{3.020000in}}%
\pgfusepath{clip}%
\pgfsetrectcap%
\pgfsetroundjoin%
\pgfsetlinewidth{0.401500pt}%
\definecolor{currentstroke}{rgb}{0.800000,0.800000,0.800000}%
\pgfsetstrokecolor{currentstroke}%
\pgfsetstrokeopacity{0.550000}%
\pgfsetdash{}{0pt}%
\pgfpathmoveto{\pgfqpoint{0.188889in}{0.696942in}}%
\pgfpathlineto{\pgfqpoint{4.838889in}{0.696942in}}%
\pgfusepath{stroke}%
\end{pgfscope}%
\begin{pgfscope}%
\pgfpathrectangle{\pgfqpoint{0.188889in}{0.454405in}}{\pgfqpoint{4.650000in}{3.020000in}}%
\pgfusepath{clip}%
\pgfsetrectcap%
\pgfsetroundjoin%
\pgfsetlinewidth{0.401500pt}%
\definecolor{currentstroke}{rgb}{0.800000,0.800000,0.800000}%
\pgfsetstrokecolor{currentstroke}%
\pgfsetstrokeopacity{0.550000}%
\pgfsetdash{}{0pt}%
\pgfpathmoveto{\pgfqpoint{0.188889in}{0.793194in}}%
\pgfpathlineto{\pgfqpoint{4.838889in}{0.793194in}}%
\pgfusepath{stroke}%
\end{pgfscope}%
\begin{pgfscope}%
\pgfpathrectangle{\pgfqpoint{0.188889in}{0.454405in}}{\pgfqpoint{4.650000in}{3.020000in}}%
\pgfusepath{clip}%
\pgfsetrectcap%
\pgfsetroundjoin%
\pgfsetlinewidth{0.401500pt}%
\definecolor{currentstroke}{rgb}{0.800000,0.800000,0.800000}%
\pgfsetstrokecolor{currentstroke}%
\pgfsetstrokeopacity{0.550000}%
\pgfsetdash{}{0pt}%
\pgfpathmoveto{\pgfqpoint{0.188889in}{0.878095in}}%
\pgfpathlineto{\pgfqpoint{4.838889in}{0.878095in}}%
\pgfusepath{stroke}%
\end{pgfscope}%
\begin{pgfscope}%
\pgfpathrectangle{\pgfqpoint{0.188889in}{0.454405in}}{\pgfqpoint{4.650000in}{3.020000in}}%
\pgfusepath{clip}%
\pgfsetrectcap%
\pgfsetroundjoin%
\pgfsetlinewidth{0.401500pt}%
\definecolor{currentstroke}{rgb}{0.800000,0.800000,0.800000}%
\pgfsetstrokecolor{currentstroke}%
\pgfsetstrokeopacity{0.550000}%
\pgfsetdash{}{0pt}%
\pgfpathmoveto{\pgfqpoint{0.188889in}{1.453678in}}%
\pgfpathlineto{\pgfqpoint{4.838889in}{1.453678in}}%
\pgfusepath{stroke}%
\end{pgfscope}%
\begin{pgfscope}%
\pgfpathrectangle{\pgfqpoint{0.188889in}{0.454405in}}{\pgfqpoint{4.650000in}{3.020000in}}%
\pgfusepath{clip}%
\pgfsetrectcap%
\pgfsetroundjoin%
\pgfsetlinewidth{0.401500pt}%
\definecolor{currentstroke}{rgb}{0.800000,0.800000,0.800000}%
\pgfsetstrokecolor{currentstroke}%
\pgfsetstrokeopacity{0.550000}%
\pgfsetdash{}{0pt}%
\pgfpathmoveto{\pgfqpoint{0.188889in}{1.745947in}}%
\pgfpathlineto{\pgfqpoint{4.838889in}{1.745947in}}%
\pgfusepath{stroke}%
\end{pgfscope}%
\begin{pgfscope}%
\pgfpathrectangle{\pgfqpoint{0.188889in}{0.454405in}}{\pgfqpoint{4.650000in}{3.020000in}}%
\pgfusepath{clip}%
\pgfsetrectcap%
\pgfsetroundjoin%
\pgfsetlinewidth{0.401500pt}%
\definecolor{currentstroke}{rgb}{0.800000,0.800000,0.800000}%
\pgfsetstrokecolor{currentstroke}%
\pgfsetstrokeopacity{0.550000}%
\pgfsetdash{}{0pt}%
\pgfpathmoveto{\pgfqpoint{0.188889in}{1.953315in}}%
\pgfpathlineto{\pgfqpoint{4.838889in}{1.953315in}}%
\pgfusepath{stroke}%
\end{pgfscope}%
\begin{pgfscope}%
\pgfpathrectangle{\pgfqpoint{0.188889in}{0.454405in}}{\pgfqpoint{4.650000in}{3.020000in}}%
\pgfusepath{clip}%
\pgfsetrectcap%
\pgfsetroundjoin%
\pgfsetlinewidth{0.401500pt}%
\definecolor{currentstroke}{rgb}{0.800000,0.800000,0.800000}%
\pgfsetstrokecolor{currentstroke}%
\pgfsetstrokeopacity{0.550000}%
\pgfsetdash{}{0pt}%
\pgfpathmoveto{\pgfqpoint{0.188889in}{2.114162in}}%
\pgfpathlineto{\pgfqpoint{4.838889in}{2.114162in}}%
\pgfusepath{stroke}%
\end{pgfscope}%
\begin{pgfscope}%
\pgfpathrectangle{\pgfqpoint{0.188889in}{0.454405in}}{\pgfqpoint{4.650000in}{3.020000in}}%
\pgfusepath{clip}%
\pgfsetrectcap%
\pgfsetroundjoin%
\pgfsetlinewidth{0.401500pt}%
\definecolor{currentstroke}{rgb}{0.800000,0.800000,0.800000}%
\pgfsetstrokecolor{currentstroke}%
\pgfsetstrokeopacity{0.550000}%
\pgfsetdash{}{0pt}%
\pgfpathmoveto{\pgfqpoint{0.188889in}{2.245583in}}%
\pgfpathlineto{\pgfqpoint{4.838889in}{2.245583in}}%
\pgfusepath{stroke}%
\end{pgfscope}%
\begin{pgfscope}%
\pgfpathrectangle{\pgfqpoint{0.188889in}{0.454405in}}{\pgfqpoint{4.650000in}{3.020000in}}%
\pgfusepath{clip}%
\pgfsetrectcap%
\pgfsetroundjoin%
\pgfsetlinewidth{0.401500pt}%
\definecolor{currentstroke}{rgb}{0.800000,0.800000,0.800000}%
\pgfsetstrokecolor{currentstroke}%
\pgfsetstrokeopacity{0.550000}%
\pgfsetdash{}{0pt}%
\pgfpathmoveto{\pgfqpoint{0.188889in}{2.356699in}}%
\pgfpathlineto{\pgfqpoint{4.838889in}{2.356699in}}%
\pgfusepath{stroke}%
\end{pgfscope}%
\begin{pgfscope}%
\pgfpathrectangle{\pgfqpoint{0.188889in}{0.454405in}}{\pgfqpoint{4.650000in}{3.020000in}}%
\pgfusepath{clip}%
\pgfsetrectcap%
\pgfsetroundjoin%
\pgfsetlinewidth{0.401500pt}%
\definecolor{currentstroke}{rgb}{0.800000,0.800000,0.800000}%
\pgfsetstrokecolor{currentstroke}%
\pgfsetstrokeopacity{0.550000}%
\pgfsetdash{}{0pt}%
\pgfpathmoveto{\pgfqpoint{0.188889in}{2.452951in}}%
\pgfpathlineto{\pgfqpoint{4.838889in}{2.452951in}}%
\pgfusepath{stroke}%
\end{pgfscope}%
\begin{pgfscope}%
\pgfpathrectangle{\pgfqpoint{0.188889in}{0.454405in}}{\pgfqpoint{4.650000in}{3.020000in}}%
\pgfusepath{clip}%
\pgfsetrectcap%
\pgfsetroundjoin%
\pgfsetlinewidth{0.401500pt}%
\definecolor{currentstroke}{rgb}{0.800000,0.800000,0.800000}%
\pgfsetstrokecolor{currentstroke}%
\pgfsetstrokeopacity{0.550000}%
\pgfsetdash{}{0pt}%
\pgfpathmoveto{\pgfqpoint{0.188889in}{2.537852in}}%
\pgfpathlineto{\pgfqpoint{4.838889in}{2.537852in}}%
\pgfusepath{stroke}%
\end{pgfscope}%
\begin{pgfscope}%
\pgfpathrectangle{\pgfqpoint{0.188889in}{0.454405in}}{\pgfqpoint{4.650000in}{3.020000in}}%
\pgfusepath{clip}%
\pgfsetrectcap%
\pgfsetroundjoin%
\pgfsetlinewidth{0.401500pt}%
\definecolor{currentstroke}{rgb}{0.800000,0.800000,0.800000}%
\pgfsetstrokecolor{currentstroke}%
\pgfsetstrokeopacity{0.550000}%
\pgfsetdash{}{0pt}%
\pgfpathmoveto{\pgfqpoint{0.188889in}{3.113435in}}%
\pgfpathlineto{\pgfqpoint{4.838889in}{3.113435in}}%
\pgfusepath{stroke}%
\end{pgfscope}%
\begin{pgfscope}%
\pgfpathrectangle{\pgfqpoint{0.188889in}{0.454405in}}{\pgfqpoint{4.650000in}{3.020000in}}%
\pgfusepath{clip}%
\pgfsetrectcap%
\pgfsetroundjoin%
\pgfsetlinewidth{0.401500pt}%
\definecolor{currentstroke}{rgb}{0.800000,0.800000,0.800000}%
\pgfsetstrokecolor{currentstroke}%
\pgfsetstrokeopacity{0.550000}%
\pgfsetdash{}{0pt}%
\pgfpathmoveto{\pgfqpoint{0.188889in}{3.405703in}}%
\pgfpathlineto{\pgfqpoint{4.838889in}{3.405703in}}%
\pgfusepath{stroke}%
\end{pgfscope}%
\begin{pgfscope}%
\pgfpathrectangle{\pgfqpoint{0.188889in}{0.454405in}}{\pgfqpoint{4.650000in}{3.020000in}}%
\pgfusepath{clip}%
\pgfsetbuttcap%
\pgfsetmiterjoin%
\definecolor{currentfill}{rgb}{0.347059,0.458824,0.641176}%
\pgfsetfillcolor{currentfill}%
\pgfsetlinewidth{0.803000pt}%
\definecolor{currentstroke}{rgb}{0.298039,0.298039,0.298039}%
\pgfsetstrokecolor{currentstroke}%
\pgfsetdash{}{0pt}%
\pgfpathmoveto{\pgfqpoint{0.305139in}{3.161143in}}%
\pgfpathlineto{\pgfqpoint{1.235139in}{3.161143in}}%
\pgfpathlineto{\pgfqpoint{1.235139in}{3.175793in}}%
\pgfpathlineto{\pgfqpoint{0.305139in}{3.175793in}}%
\pgfpathlineto{\pgfqpoint{0.305139in}{3.161143in}}%
\pgfpathclose%
\pgfusepath{stroke,fill}%
\end{pgfscope}%
\begin{pgfscope}%
\pgfpathrectangle{\pgfqpoint{0.188889in}{0.454405in}}{\pgfqpoint{4.650000in}{3.020000in}}%
\pgfusepath{clip}%
\pgfsetbuttcap%
\pgfsetmiterjoin%
\definecolor{currentfill}{rgb}{0.798529,0.536765,0.389706}%
\pgfsetfillcolor{currentfill}%
\pgfsetlinewidth{0.803000pt}%
\definecolor{currentstroke}{rgb}{0.298039,0.298039,0.298039}%
\pgfsetstrokecolor{currentstroke}%
\pgfsetdash{}{0pt}%
\pgfpathmoveto{\pgfqpoint{1.467639in}{2.786900in}}%
\pgfpathlineto{\pgfqpoint{2.397639in}{2.786900in}}%
\pgfpathlineto{\pgfqpoint{2.397639in}{2.839837in}}%
\pgfpathlineto{\pgfqpoint{1.467639in}{2.839837in}}%
\pgfpathlineto{\pgfqpoint{1.467639in}{2.786900in}}%
\pgfpathclose%
\pgfusepath{stroke,fill}%
\end{pgfscope}%
\begin{pgfscope}%
\pgfpathrectangle{\pgfqpoint{0.188889in}{0.454405in}}{\pgfqpoint{4.650000in}{3.020000in}}%
\pgfusepath{clip}%
\pgfsetbuttcap%
\pgfsetmiterjoin%
\definecolor{currentfill}{rgb}{0.374020,0.618137,0.429902}%
\pgfsetfillcolor{currentfill}%
\pgfsetlinewidth{0.803000pt}%
\definecolor{currentstroke}{rgb}{0.298039,0.298039,0.298039}%
\pgfsetstrokecolor{currentstroke}%
\pgfsetdash{}{0pt}%
\pgfpathmoveto{\pgfqpoint{2.630139in}{3.011790in}}%
\pgfpathlineto{\pgfqpoint{3.560139in}{3.011790in}}%
\pgfpathlineto{\pgfqpoint{3.560139in}{3.170907in}}%
\pgfpathlineto{\pgfqpoint{2.630139in}{3.170907in}}%
\pgfpathlineto{\pgfqpoint{2.630139in}{3.011790in}}%
\pgfpathclose%
\pgfusepath{stroke,fill}%
\end{pgfscope}%
\begin{pgfscope}%
\pgfpathrectangle{\pgfqpoint{0.188889in}{0.454405in}}{\pgfqpoint{4.650000in}{3.020000in}}%
\pgfusepath{clip}%
\pgfsetbuttcap%
\pgfsetmiterjoin%
\definecolor{currentfill}{rgb}{0.710784,0.363725,0.375490}%
\pgfsetfillcolor{currentfill}%
\pgfsetlinewidth{0.803000pt}%
\definecolor{currentstroke}{rgb}{0.298039,0.298039,0.298039}%
\pgfsetstrokecolor{currentstroke}%
\pgfsetdash{}{0pt}%
\pgfpathmoveto{\pgfqpoint{3.792639in}{3.012573in}}%
\pgfpathlineto{\pgfqpoint{4.722639in}{3.012573in}}%
\pgfpathlineto{\pgfqpoint{4.722639in}{3.080057in}}%
\pgfpathlineto{\pgfqpoint{3.792639in}{3.080057in}}%
\pgfpathlineto{\pgfqpoint{3.792639in}{3.012573in}}%
\pgfpathclose%
\pgfusepath{stroke,fill}%
\end{pgfscope}%
\begin{pgfscope}%
\pgfsetrectcap%
\pgfsetmiterjoin%
\pgfsetlinewidth{1.254687pt}%
\definecolor{currentstroke}{rgb}{0.800000,0.800000,0.800000}%
\pgfsetstrokecolor{currentstroke}%
\pgfsetdash{}{0pt}%
\pgfpathmoveto{\pgfqpoint{0.188889in}{0.454405in}}%
\pgfpathlineto{\pgfqpoint{0.188889in}{3.474405in}}%
\pgfusepath{stroke}%
\end{pgfscope}%
\begin{pgfscope}%
\pgfsetrectcap%
\pgfsetmiterjoin%
\pgfsetlinewidth{1.254687pt}%
\definecolor{currentstroke}{rgb}{0.800000,0.800000,0.800000}%
\pgfsetstrokecolor{currentstroke}%
\pgfsetdash{}{0pt}%
\pgfpathmoveto{\pgfqpoint{4.838889in}{0.454405in}}%
\pgfpathlineto{\pgfqpoint{4.838889in}{3.474405in}}%
\pgfusepath{stroke}%
\end{pgfscope}%
\begin{pgfscope}%
\pgfsetrectcap%
\pgfsetmiterjoin%
\pgfsetlinewidth{1.254687pt}%
\definecolor{currentstroke}{rgb}{0.800000,0.800000,0.800000}%
\pgfsetstrokecolor{currentstroke}%
\pgfsetdash{}{0pt}%
\pgfpathmoveto{\pgfqpoint{0.188889in}{0.454405in}}%
\pgfpathlineto{\pgfqpoint{4.838889in}{0.454405in}}%
\pgfusepath{stroke}%
\end{pgfscope}%
\begin{pgfscope}%
\pgfsetrectcap%
\pgfsetmiterjoin%
\pgfsetlinewidth{1.254687pt}%
\definecolor{currentstroke}{rgb}{0.800000,0.800000,0.800000}%
\pgfsetstrokecolor{currentstroke}%
\pgfsetdash{}{0pt}%
\pgfpathmoveto{\pgfqpoint{0.188889in}{3.474405in}}%
\pgfpathlineto{\pgfqpoint{4.838889in}{3.474405in}}%
\pgfusepath{stroke}%
\end{pgfscope}%
\begin{pgfscope}%
\pgfpathrectangle{\pgfqpoint{0.188889in}{0.454405in}}{\pgfqpoint{4.650000in}{3.020000in}}%
\pgfusepath{clip}%
\pgfsetrectcap%
\pgfsetroundjoin%
\pgfsetlinewidth{0.803000pt}%
\definecolor{currentstroke}{rgb}{0.298039,0.298039,0.298039}%
\pgfsetstrokecolor{currentstroke}%
\pgfsetdash{}{0pt}%
\pgfpathmoveto{\pgfqpoint{0.770139in}{3.161143in}}%
\pgfpathlineto{\pgfqpoint{0.770139in}{3.151876in}}%
\pgfusepath{stroke}%
\end{pgfscope}%
\begin{pgfscope}%
\pgfpathrectangle{\pgfqpoint{0.188889in}{0.454405in}}{\pgfqpoint{4.650000in}{3.020000in}}%
\pgfusepath{clip}%
\pgfsetrectcap%
\pgfsetroundjoin%
\pgfsetlinewidth{0.803000pt}%
\definecolor{currentstroke}{rgb}{0.298039,0.298039,0.298039}%
\pgfsetstrokecolor{currentstroke}%
\pgfsetdash{}{0pt}%
\pgfpathmoveto{\pgfqpoint{0.770139in}{3.175793in}}%
\pgfpathlineto{\pgfqpoint{0.770139in}{3.180584in}}%
\pgfusepath{stroke}%
\end{pgfscope}%
\begin{pgfscope}%
\pgfpathrectangle{\pgfqpoint{0.188889in}{0.454405in}}{\pgfqpoint{4.650000in}{3.020000in}}%
\pgfusepath{clip}%
\pgfsetrectcap%
\pgfsetroundjoin%
\pgfsetlinewidth{0.803000pt}%
\definecolor{currentstroke}{rgb}{0.298039,0.298039,0.298039}%
\pgfsetstrokecolor{currentstroke}%
\pgfsetdash{}{0pt}%
\pgfpathmoveto{\pgfqpoint{0.537639in}{3.151876in}}%
\pgfpathlineto{\pgfqpoint{1.002639in}{3.151876in}}%
\pgfusepath{stroke}%
\end{pgfscope}%
\begin{pgfscope}%
\pgfpathrectangle{\pgfqpoint{0.188889in}{0.454405in}}{\pgfqpoint{4.650000in}{3.020000in}}%
\pgfusepath{clip}%
\pgfsetrectcap%
\pgfsetroundjoin%
\pgfsetlinewidth{0.803000pt}%
\definecolor{currentstroke}{rgb}{0.298039,0.298039,0.298039}%
\pgfsetstrokecolor{currentstroke}%
\pgfsetdash{}{0pt}%
\pgfpathmoveto{\pgfqpoint{0.537639in}{3.180584in}}%
\pgfpathlineto{\pgfqpoint{1.002639in}{3.180584in}}%
\pgfusepath{stroke}%
\end{pgfscope}%
\begin{pgfscope}%
\pgfpathrectangle{\pgfqpoint{0.188889in}{0.454405in}}{\pgfqpoint{4.650000in}{3.020000in}}%
\pgfusepath{clip}%
\pgfsetrectcap%
\pgfsetroundjoin%
\pgfsetlinewidth{0.803000pt}%
\definecolor{currentstroke}{rgb}{0.298039,0.298039,0.298039}%
\pgfsetstrokecolor{currentstroke}%
\pgfsetdash{}{0pt}%
\pgfpathmoveto{\pgfqpoint{0.305139in}{3.161143in}}%
\pgfpathlineto{\pgfqpoint{1.235139in}{3.161143in}}%
\pgfusepath{stroke}%
\end{pgfscope}%
\begin{pgfscope}%
\pgfpathrectangle{\pgfqpoint{0.188889in}{0.454405in}}{\pgfqpoint{4.650000in}{3.020000in}}%
\pgfusepath{clip}%
\pgfsetbuttcap%
\pgfsetmiterjoin%
\definecolor{currentfill}{rgb}{0.298039,0.298039,0.298039}%
\pgfsetfillcolor{currentfill}%
\pgfsetlinewidth{1.003750pt}%
\definecolor{currentstroke}{rgb}{0.298039,0.298039,0.298039}%
\pgfsetstrokecolor{currentstroke}%
\pgfsetdash{}{0pt}%
\pgfsys@defobject{currentmarker}{\pgfqpoint{-0.029463in}{-0.049105in}}{\pgfqpoint{0.029463in}{0.049105in}}{%
\pgfpathmoveto{\pgfqpoint{0.000000in}{-0.049105in}}%
\pgfpathlineto{\pgfqpoint{0.029463in}{0.000000in}}%
\pgfpathlineto{\pgfqpoint{0.000000in}{0.049105in}}%
\pgfpathlineto{\pgfqpoint{-0.029463in}{0.000000in}}%
\pgfpathclose%
\pgfusepath{stroke,fill}%
}%
\begin{pgfscope}%
\pgfsys@transformshift{0.770139in}{3.136474in}%
\pgfsys@useobject{currentmarker}{}%
\end{pgfscope}%
\begin{pgfscope}%
\pgfsys@transformshift{0.770139in}{3.199127in}%
\pgfsys@useobject{currentmarker}{}%
\end{pgfscope}%
\begin{pgfscope}%
\pgfsys@transformshift{0.770139in}{3.199127in}%
\pgfsys@useobject{currentmarker}{}%
\end{pgfscope}%
\begin{pgfscope}%
\pgfsys@transformshift{0.770139in}{3.225545in}%
\pgfsys@useobject{currentmarker}{}%
\end{pgfscope}%
\end{pgfscope}%
\begin{pgfscope}%
\pgfpathrectangle{\pgfqpoint{0.188889in}{0.454405in}}{\pgfqpoint{4.650000in}{3.020000in}}%
\pgfusepath{clip}%
\pgfsetrectcap%
\pgfsetroundjoin%
\pgfsetlinewidth{0.803000pt}%
\definecolor{currentstroke}{rgb}{0.298039,0.298039,0.298039}%
\pgfsetstrokecolor{currentstroke}%
\pgfsetdash{}{0pt}%
\pgfpathmoveto{\pgfqpoint{1.932639in}{2.786900in}}%
\pgfpathlineto{\pgfqpoint{1.932639in}{2.701176in}}%
\pgfusepath{stroke}%
\end{pgfscope}%
\begin{pgfscope}%
\pgfpathrectangle{\pgfqpoint{0.188889in}{0.454405in}}{\pgfqpoint{4.650000in}{3.020000in}}%
\pgfusepath{clip}%
\pgfsetrectcap%
\pgfsetroundjoin%
\pgfsetlinewidth{0.803000pt}%
\definecolor{currentstroke}{rgb}{0.298039,0.298039,0.298039}%
\pgfsetstrokecolor{currentstroke}%
\pgfsetdash{}{0pt}%
\pgfpathmoveto{\pgfqpoint{1.932639in}{2.839837in}}%
\pgfpathlineto{\pgfqpoint{1.932639in}{2.897162in}}%
\pgfusepath{stroke}%
\end{pgfscope}%
\begin{pgfscope}%
\pgfpathrectangle{\pgfqpoint{0.188889in}{0.454405in}}{\pgfqpoint{4.650000in}{3.020000in}}%
\pgfusepath{clip}%
\pgfsetrectcap%
\pgfsetroundjoin%
\pgfsetlinewidth{0.803000pt}%
\definecolor{currentstroke}{rgb}{0.298039,0.298039,0.298039}%
\pgfsetstrokecolor{currentstroke}%
\pgfsetdash{}{0pt}%
\pgfpathmoveto{\pgfqpoint{1.700139in}{2.701176in}}%
\pgfpathlineto{\pgfqpoint{2.165139in}{2.701176in}}%
\pgfusepath{stroke}%
\end{pgfscope}%
\begin{pgfscope}%
\pgfpathrectangle{\pgfqpoint{0.188889in}{0.454405in}}{\pgfqpoint{4.650000in}{3.020000in}}%
\pgfusepath{clip}%
\pgfsetrectcap%
\pgfsetroundjoin%
\pgfsetlinewidth{0.803000pt}%
\definecolor{currentstroke}{rgb}{0.298039,0.298039,0.298039}%
\pgfsetstrokecolor{currentstroke}%
\pgfsetdash{}{0pt}%
\pgfpathmoveto{\pgfqpoint{1.700139in}{2.897162in}}%
\pgfpathlineto{\pgfqpoint{2.165139in}{2.897162in}}%
\pgfusepath{stroke}%
\end{pgfscope}%
\begin{pgfscope}%
\pgfpathrectangle{\pgfqpoint{0.188889in}{0.454405in}}{\pgfqpoint{4.650000in}{3.020000in}}%
\pgfusepath{clip}%
\pgfsetrectcap%
\pgfsetroundjoin%
\pgfsetlinewidth{0.803000pt}%
\definecolor{currentstroke}{rgb}{0.298039,0.298039,0.298039}%
\pgfsetstrokecolor{currentstroke}%
\pgfsetdash{}{0pt}%
\pgfpathmoveto{\pgfqpoint{1.467639in}{2.806739in}}%
\pgfpathlineto{\pgfqpoint{2.397639in}{2.806739in}}%
\pgfusepath{stroke}%
\end{pgfscope}%
\begin{pgfscope}%
\pgfpathrectangle{\pgfqpoint{0.188889in}{0.454405in}}{\pgfqpoint{4.650000in}{3.020000in}}%
\pgfusepath{clip}%
\pgfsetrectcap%
\pgfsetroundjoin%
\pgfsetlinewidth{0.803000pt}%
\definecolor{currentstroke}{rgb}{0.298039,0.298039,0.298039}%
\pgfsetstrokecolor{currentstroke}%
\pgfsetdash{}{0pt}%
\pgfpathmoveto{\pgfqpoint{3.095139in}{3.011790in}}%
\pgfpathlineto{\pgfqpoint{3.095139in}{2.857663in}}%
\pgfusepath{stroke}%
\end{pgfscope}%
\begin{pgfscope}%
\pgfpathrectangle{\pgfqpoint{0.188889in}{0.454405in}}{\pgfqpoint{4.650000in}{3.020000in}}%
\pgfusepath{clip}%
\pgfsetrectcap%
\pgfsetroundjoin%
\pgfsetlinewidth{0.803000pt}%
\definecolor{currentstroke}{rgb}{0.298039,0.298039,0.298039}%
\pgfsetstrokecolor{currentstroke}%
\pgfsetdash{}{0pt}%
\pgfpathmoveto{\pgfqpoint{3.095139in}{3.170907in}}%
\pgfpathlineto{\pgfqpoint{3.095139in}{3.280252in}}%
\pgfusepath{stroke}%
\end{pgfscope}%
\begin{pgfscope}%
\pgfpathrectangle{\pgfqpoint{0.188889in}{0.454405in}}{\pgfqpoint{4.650000in}{3.020000in}}%
\pgfusepath{clip}%
\pgfsetrectcap%
\pgfsetroundjoin%
\pgfsetlinewidth{0.803000pt}%
\definecolor{currentstroke}{rgb}{0.298039,0.298039,0.298039}%
\pgfsetstrokecolor{currentstroke}%
\pgfsetdash{}{0pt}%
\pgfpathmoveto{\pgfqpoint{2.862639in}{2.857663in}}%
\pgfpathlineto{\pgfqpoint{3.327639in}{2.857663in}}%
\pgfusepath{stroke}%
\end{pgfscope}%
\begin{pgfscope}%
\pgfpathrectangle{\pgfqpoint{0.188889in}{0.454405in}}{\pgfqpoint{4.650000in}{3.020000in}}%
\pgfusepath{clip}%
\pgfsetrectcap%
\pgfsetroundjoin%
\pgfsetlinewidth{0.803000pt}%
\definecolor{currentstroke}{rgb}{0.298039,0.298039,0.298039}%
\pgfsetstrokecolor{currentstroke}%
\pgfsetdash{}{0pt}%
\pgfpathmoveto{\pgfqpoint{2.862639in}{3.280252in}}%
\pgfpathlineto{\pgfqpoint{3.327639in}{3.280252in}}%
\pgfusepath{stroke}%
\end{pgfscope}%
\begin{pgfscope}%
\pgfpathrectangle{\pgfqpoint{0.188889in}{0.454405in}}{\pgfqpoint{4.650000in}{3.020000in}}%
\pgfusepath{clip}%
\pgfsetrectcap%
\pgfsetroundjoin%
\pgfsetlinewidth{0.803000pt}%
\definecolor{currentstroke}{rgb}{0.298039,0.298039,0.298039}%
\pgfsetstrokecolor{currentstroke}%
\pgfsetdash{}{0pt}%
\pgfpathmoveto{\pgfqpoint{2.630139in}{3.137109in}}%
\pgfpathlineto{\pgfqpoint{3.560139in}{3.137109in}}%
\pgfusepath{stroke}%
\end{pgfscope}%
\begin{pgfscope}%
\pgfpathrectangle{\pgfqpoint{0.188889in}{0.454405in}}{\pgfqpoint{4.650000in}{3.020000in}}%
\pgfusepath{clip}%
\pgfsetrectcap%
\pgfsetroundjoin%
\pgfsetlinewidth{0.803000pt}%
\definecolor{currentstroke}{rgb}{0.298039,0.298039,0.298039}%
\pgfsetstrokecolor{currentstroke}%
\pgfsetdash{}{0pt}%
\pgfpathmoveto{\pgfqpoint{4.257639in}{3.012573in}}%
\pgfpathlineto{\pgfqpoint{4.257639in}{2.954325in}}%
\pgfusepath{stroke}%
\end{pgfscope}%
\begin{pgfscope}%
\pgfpathrectangle{\pgfqpoint{0.188889in}{0.454405in}}{\pgfqpoint{4.650000in}{3.020000in}}%
\pgfusepath{clip}%
\pgfsetrectcap%
\pgfsetroundjoin%
\pgfsetlinewidth{0.803000pt}%
\definecolor{currentstroke}{rgb}{0.298039,0.298039,0.298039}%
\pgfsetstrokecolor{currentstroke}%
\pgfsetdash{}{0pt}%
\pgfpathmoveto{\pgfqpoint{4.257639in}{3.080057in}}%
\pgfpathlineto{\pgfqpoint{4.257639in}{3.123851in}}%
\pgfusepath{stroke}%
\end{pgfscope}%
\begin{pgfscope}%
\pgfpathrectangle{\pgfqpoint{0.188889in}{0.454405in}}{\pgfqpoint{4.650000in}{3.020000in}}%
\pgfusepath{clip}%
\pgfsetrectcap%
\pgfsetroundjoin%
\pgfsetlinewidth{0.803000pt}%
\definecolor{currentstroke}{rgb}{0.298039,0.298039,0.298039}%
\pgfsetstrokecolor{currentstroke}%
\pgfsetdash{}{0pt}%
\pgfpathmoveto{\pgfqpoint{4.025139in}{2.954325in}}%
\pgfpathlineto{\pgfqpoint{4.490139in}{2.954325in}}%
\pgfusepath{stroke}%
\end{pgfscope}%
\begin{pgfscope}%
\pgfpathrectangle{\pgfqpoint{0.188889in}{0.454405in}}{\pgfqpoint{4.650000in}{3.020000in}}%
\pgfusepath{clip}%
\pgfsetrectcap%
\pgfsetroundjoin%
\pgfsetlinewidth{0.803000pt}%
\definecolor{currentstroke}{rgb}{0.298039,0.298039,0.298039}%
\pgfsetstrokecolor{currentstroke}%
\pgfsetdash{}{0pt}%
\pgfpathmoveto{\pgfqpoint{4.025139in}{3.123851in}}%
\pgfpathlineto{\pgfqpoint{4.490139in}{3.123851in}}%
\pgfusepath{stroke}%
\end{pgfscope}%
\begin{pgfscope}%
\pgfpathrectangle{\pgfqpoint{0.188889in}{0.454405in}}{\pgfqpoint{4.650000in}{3.020000in}}%
\pgfusepath{clip}%
\pgfsetrectcap%
\pgfsetroundjoin%
\pgfsetlinewidth{0.803000pt}%
\definecolor{currentstroke}{rgb}{0.298039,0.298039,0.298039}%
\pgfsetstrokecolor{currentstroke}%
\pgfsetdash{}{0pt}%
\pgfpathmoveto{\pgfqpoint{3.792639in}{3.058193in}}%
\pgfpathlineto{\pgfqpoint{4.722639in}{3.058193in}}%
\pgfusepath{stroke}%
\end{pgfscope}%
\begin{pgfscope}%
\pgfpathrectangle{\pgfqpoint{0.188889in}{0.454405in}}{\pgfqpoint{4.650000in}{3.020000in}}%
\pgfusepath{clip}%
\pgfsetbuttcap%
\pgfsetmiterjoin%
\definecolor{currentfill}{rgb}{0.298039,0.298039,0.298039}%
\pgfsetfillcolor{currentfill}%
\pgfsetlinewidth{1.003750pt}%
\definecolor{currentstroke}{rgb}{0.298039,0.298039,0.298039}%
\pgfsetstrokecolor{currentstroke}%
\pgfsetdash{}{0pt}%
\pgfsys@defobject{currentmarker}{\pgfqpoint{-0.029463in}{-0.049105in}}{\pgfqpoint{0.029463in}{0.049105in}}{%
\pgfpathmoveto{\pgfqpoint{0.000000in}{-0.049105in}}%
\pgfpathlineto{\pgfqpoint{0.029463in}{0.000000in}}%
\pgfpathlineto{\pgfqpoint{0.000000in}{0.049105in}}%
\pgfpathlineto{\pgfqpoint{-0.029463in}{0.000000in}}%
\pgfpathclose%
\pgfusepath{stroke,fill}%
}%
\begin{pgfscope}%
\pgfsys@transformshift{4.257639in}{2.889621in}%
\pgfsys@useobject{currentmarker}{}%
\end{pgfscope}%
\end{pgfscope}%
\end{pgfpicture}%
\makeatother%
\endgroup%

%% file: fig/boxplot_sparse_moon_res.pgf
%% Creator: Matplotlib, PGF backend
%%
%% To include the figure in your LaTeX document, write
%%   \input{<filename>.pgf}
%%
%% Make sure the required packages are loaded in your preamble
%%   \usepackage{pgf}
%%
%% Figures using additional raster images can only be included by \input if
%% they are in the same directory as the main LaTeX file. For loading figures
%% from other directories you can use the `import` package
%% and then include the figures with
%%   \import{<path to file>}{<filename>.pgf}
%%
%% Matplotlib used the following preamble
%%
\begingroup%
\makeatletter%
\begin{pgfpicture}%
\pgfpathrectangle{\pgfpointorigin}{\pgfqpoint{4.888889in}{3.524405in}}%
\pgfusepath{use as bounding box, clip}%
\begin{pgfscope}%
\pgfsetbuttcap%
\pgfsetmiterjoin%
\definecolor{currentfill}{rgb}{1.000000,1.000000,1.000000}%
\pgfsetfillcolor{currentfill}%
\pgfsetlinewidth{0.000000pt}%
\definecolor{currentstroke}{rgb}{1.000000,1.000000,1.000000}%
\pgfsetstrokecolor{currentstroke}%
\pgfsetdash{}{0pt}%
\pgfpathmoveto{\pgfqpoint{0.000000in}{0.000000in}}%
\pgfpathlineto{\pgfqpoint{4.888889in}{0.000000in}}%
\pgfpathlineto{\pgfqpoint{4.888889in}{3.524405in}}%
\pgfpathlineto{\pgfqpoint{0.000000in}{3.524405in}}%
\pgfpathclose%
\pgfusepath{fill}%
\end{pgfscope}%
\begin{pgfscope}%
\pgfsetbuttcap%
\pgfsetmiterjoin%
\definecolor{currentfill}{rgb}{1.000000,1.000000,1.000000}%
\pgfsetfillcolor{currentfill}%
\pgfsetlinewidth{0.000000pt}%
\definecolor{currentstroke}{rgb}{0.000000,0.000000,0.000000}%
\pgfsetstrokecolor{currentstroke}%
\pgfsetstrokeopacity{0.000000}%
\pgfsetdash{}{0pt}%
\pgfpathmoveto{\pgfqpoint{0.188889in}{0.454405in}}%
\pgfpathlineto{\pgfqpoint{4.838889in}{0.454405in}}%
\pgfpathlineto{\pgfqpoint{4.838889in}{3.474405in}}%
\pgfpathlineto{\pgfqpoint{0.188889in}{3.474405in}}%
\pgfpathclose%
\pgfusepath{fill}%
\end{pgfscope}%
\begin{pgfscope}%
\definecolor{textcolor}{rgb}{0.000000,0.000000,0.000000}%
\pgfsetstrokecolor{textcolor}%
\pgfsetfillcolor{textcolor}%
\pgftext[x=0.770139in,y=0.357183in,,top]{\color{textcolor}\rmfamily\fontsize{24.200000}{29.040000}\selectfont \textsc{w-l}}%
\end{pgfscope}%
\begin{pgfscope}%
\definecolor{textcolor}{rgb}{0.000000,0.000000,0.000000}%
\pgfsetstrokecolor{textcolor}%
\pgfsetfillcolor{textcolor}%
\pgftext[x=1.932639in,y=0.357183in,,top]{\color{textcolor}\rmfamily\fontsize{24.200000}{29.040000}\selectfont \textsc{w-m}}%
\end{pgfscope}%
\begin{pgfscope}%
\definecolor{textcolor}{rgb}{0.000000,0.000000,0.000000}%
\pgfsetstrokecolor{textcolor}%
\pgfsetfillcolor{textcolor}%
\pgftext[x=3.095139in,y=0.357183in,,top]{\color{textcolor}\rmfamily\fontsize{24.200000}{29.040000}\selectfont \textsc{dcv}}%
\end{pgfscope}%
\begin{pgfscope}%
\definecolor{textcolor}{rgb}{0.000000,0.000000,0.000000}%
\pgfsetstrokecolor{textcolor}%
\pgfsetfillcolor{textcolor}%
\pgftext[x=4.257639in,y=0.357183in,,top]{\color{textcolor}\rmfamily\fontsize{24.200000}{29.040000}\selectfont \textsc{mc}}%
\end{pgfscope}%
\begin{pgfscope}%
\pgfpathrectangle{\pgfqpoint{0.188889in}{0.454405in}}{\pgfqpoint{4.650000in}{3.020000in}}%
\pgfusepath{clip}%
\pgfsetrectcap%
\pgfsetroundjoin%
\pgfsetlinewidth{2.509375pt}%
\definecolor{currentstroke}{rgb}{0.800000,0.800000,0.800000}%
\pgfsetstrokecolor{currentstroke}%
\pgfsetdash{}{0pt}%
\pgfpathmoveto{\pgfqpoint{0.188889in}{0.954042in}}%
\pgfpathlineto{\pgfqpoint{4.838889in}{0.954042in}}%
\pgfusepath{stroke}%
\end{pgfscope}%
\begin{pgfscope}%
\pgfsetbuttcap%
\pgfsetroundjoin%
\definecolor{currentfill}{rgb}{0.800000,0.800000,0.800000}%
\pgfsetfillcolor{currentfill}%
\pgfsetlinewidth{1.003750pt}%
\definecolor{currentstroke}{rgb}{0.800000,0.800000,0.800000}%
\pgfsetstrokecolor{currentstroke}%
\pgfsetdash{}{0pt}%
\pgfsys@defobject{currentmarker}{\pgfqpoint{-0.138889in}{0.000000in}}{\pgfqpoint{0.000000in}{0.000000in}}{%
\pgfpathmoveto{\pgfqpoint{0.000000in}{0.000000in}}%
\pgfpathlineto{\pgfqpoint{-0.138889in}{0.000000in}}%
\pgfusepath{stroke,fill}%
}%
\begin{pgfscope}%
\pgfsys@transformshift{0.188889in}{0.954042in}%
\pgfsys@useobject{currentmarker}{}%
\end{pgfscope}%
\end{pgfscope}%
\begin{pgfscope}%
\pgfpathrectangle{\pgfqpoint{0.188889in}{0.454405in}}{\pgfqpoint{4.650000in}{3.020000in}}%
\pgfusepath{clip}%
\pgfsetrectcap%
\pgfsetroundjoin%
\pgfsetlinewidth{2.509375pt}%
\definecolor{currentstroke}{rgb}{0.611765,0.611765,0.611765}%
\pgfsetstrokecolor{currentstroke}%
\pgfsetdash{}{0pt}%
\pgfpathmoveto{\pgfqpoint{0.188889in}{2.613798in}}%
\pgfpathlineto{\pgfqpoint{4.838889in}{2.613798in}}%
\pgfusepath{stroke}%
\end{pgfscope}%
\begin{pgfscope}%
\pgfsetbuttcap%
\pgfsetroundjoin%
\definecolor{currentfill}{rgb}{0.800000,0.800000,0.800000}%
\pgfsetfillcolor{currentfill}%
\pgfsetlinewidth{1.003750pt}%
\definecolor{currentstroke}{rgb}{0.800000,0.800000,0.800000}%
\pgfsetstrokecolor{currentstroke}%
\pgfsetdash{}{0pt}%
\pgfsys@defobject{currentmarker}{\pgfqpoint{-0.138889in}{0.000000in}}{\pgfqpoint{0.000000in}{0.000000in}}{%
\pgfpathmoveto{\pgfqpoint{0.000000in}{0.000000in}}%
\pgfpathlineto{\pgfqpoint{-0.138889in}{0.000000in}}%
\pgfusepath{stroke,fill}%
}%
\begin{pgfscope}%
\pgfsys@transformshift{0.188889in}{2.613798in}%
\pgfsys@useobject{currentmarker}{}%
\end{pgfscope}%
\end{pgfscope}%
\begin{pgfscope}%
\pgfpathrectangle{\pgfqpoint{0.188889in}{0.454405in}}{\pgfqpoint{4.650000in}{3.020000in}}%
\pgfusepath{clip}%
\pgfsetrectcap%
\pgfsetroundjoin%
\pgfsetlinewidth{0.401500pt}%
\definecolor{currentstroke}{rgb}{0.800000,0.800000,0.800000}%
\pgfsetstrokecolor{currentstroke}%
\pgfsetstrokeopacity{0.550000}%
\pgfsetdash{}{0pt}%
\pgfpathmoveto{\pgfqpoint{0.188889in}{0.454405in}}%
\pgfpathlineto{\pgfqpoint{4.838889in}{0.454405in}}%
\pgfusepath{stroke}%
\end{pgfscope}%
\begin{pgfscope}%
\pgfpathrectangle{\pgfqpoint{0.188889in}{0.454405in}}{\pgfqpoint{4.650000in}{3.020000in}}%
\pgfusepath{clip}%
\pgfsetrectcap%
\pgfsetroundjoin%
\pgfsetlinewidth{0.401500pt}%
\definecolor{currentstroke}{rgb}{0.800000,0.800000,0.800000}%
\pgfsetstrokecolor{currentstroke}%
\pgfsetstrokeopacity{0.550000}%
\pgfsetdash{}{0pt}%
\pgfpathmoveto{\pgfqpoint{0.188889in}{0.585827in}}%
\pgfpathlineto{\pgfqpoint{4.838889in}{0.585827in}}%
\pgfusepath{stroke}%
\end{pgfscope}%
\begin{pgfscope}%
\pgfpathrectangle{\pgfqpoint{0.188889in}{0.454405in}}{\pgfqpoint{4.650000in}{3.020000in}}%
\pgfusepath{clip}%
\pgfsetrectcap%
\pgfsetroundjoin%
\pgfsetlinewidth{0.401500pt}%
\definecolor{currentstroke}{rgb}{0.800000,0.800000,0.800000}%
\pgfsetstrokecolor{currentstroke}%
\pgfsetstrokeopacity{0.550000}%
\pgfsetdash{}{0pt}%
\pgfpathmoveto{\pgfqpoint{0.188889in}{0.696942in}}%
\pgfpathlineto{\pgfqpoint{4.838889in}{0.696942in}}%
\pgfusepath{stroke}%
\end{pgfscope}%
\begin{pgfscope}%
\pgfpathrectangle{\pgfqpoint{0.188889in}{0.454405in}}{\pgfqpoint{4.650000in}{3.020000in}}%
\pgfusepath{clip}%
\pgfsetrectcap%
\pgfsetroundjoin%
\pgfsetlinewidth{0.401500pt}%
\definecolor{currentstroke}{rgb}{0.800000,0.800000,0.800000}%
\pgfsetstrokecolor{currentstroke}%
\pgfsetstrokeopacity{0.550000}%
\pgfsetdash{}{0pt}%
\pgfpathmoveto{\pgfqpoint{0.188889in}{0.793194in}}%
\pgfpathlineto{\pgfqpoint{4.838889in}{0.793194in}}%
\pgfusepath{stroke}%
\end{pgfscope}%
\begin{pgfscope}%
\pgfpathrectangle{\pgfqpoint{0.188889in}{0.454405in}}{\pgfqpoint{4.650000in}{3.020000in}}%
\pgfusepath{clip}%
\pgfsetrectcap%
\pgfsetroundjoin%
\pgfsetlinewidth{0.401500pt}%
\definecolor{currentstroke}{rgb}{0.800000,0.800000,0.800000}%
\pgfsetstrokecolor{currentstroke}%
\pgfsetstrokeopacity{0.550000}%
\pgfsetdash{}{0pt}%
\pgfpathmoveto{\pgfqpoint{0.188889in}{0.878095in}}%
\pgfpathlineto{\pgfqpoint{4.838889in}{0.878095in}}%
\pgfusepath{stroke}%
\end{pgfscope}%
\begin{pgfscope}%
\pgfpathrectangle{\pgfqpoint{0.188889in}{0.454405in}}{\pgfqpoint{4.650000in}{3.020000in}}%
\pgfusepath{clip}%
\pgfsetrectcap%
\pgfsetroundjoin%
\pgfsetlinewidth{0.401500pt}%
\definecolor{currentstroke}{rgb}{0.800000,0.800000,0.800000}%
\pgfsetstrokecolor{currentstroke}%
\pgfsetstrokeopacity{0.550000}%
\pgfsetdash{}{0pt}%
\pgfpathmoveto{\pgfqpoint{0.188889in}{1.453678in}}%
\pgfpathlineto{\pgfqpoint{4.838889in}{1.453678in}}%
\pgfusepath{stroke}%
\end{pgfscope}%
\begin{pgfscope}%
\pgfpathrectangle{\pgfqpoint{0.188889in}{0.454405in}}{\pgfqpoint{4.650000in}{3.020000in}}%
\pgfusepath{clip}%
\pgfsetrectcap%
\pgfsetroundjoin%
\pgfsetlinewidth{0.401500pt}%
\definecolor{currentstroke}{rgb}{0.800000,0.800000,0.800000}%
\pgfsetstrokecolor{currentstroke}%
\pgfsetstrokeopacity{0.550000}%
\pgfsetdash{}{0pt}%
\pgfpathmoveto{\pgfqpoint{0.188889in}{1.745947in}}%
\pgfpathlineto{\pgfqpoint{4.838889in}{1.745947in}}%
\pgfusepath{stroke}%
\end{pgfscope}%
\begin{pgfscope}%
\pgfpathrectangle{\pgfqpoint{0.188889in}{0.454405in}}{\pgfqpoint{4.650000in}{3.020000in}}%
\pgfusepath{clip}%
\pgfsetrectcap%
\pgfsetroundjoin%
\pgfsetlinewidth{0.401500pt}%
\definecolor{currentstroke}{rgb}{0.800000,0.800000,0.800000}%
\pgfsetstrokecolor{currentstroke}%
\pgfsetstrokeopacity{0.550000}%
\pgfsetdash{}{0pt}%
\pgfpathmoveto{\pgfqpoint{0.188889in}{1.953315in}}%
\pgfpathlineto{\pgfqpoint{4.838889in}{1.953315in}}%
\pgfusepath{stroke}%
\end{pgfscope}%
\begin{pgfscope}%
\pgfpathrectangle{\pgfqpoint{0.188889in}{0.454405in}}{\pgfqpoint{4.650000in}{3.020000in}}%
\pgfusepath{clip}%
\pgfsetrectcap%
\pgfsetroundjoin%
\pgfsetlinewidth{0.401500pt}%
\definecolor{currentstroke}{rgb}{0.800000,0.800000,0.800000}%
\pgfsetstrokecolor{currentstroke}%
\pgfsetstrokeopacity{0.550000}%
\pgfsetdash{}{0pt}%
\pgfpathmoveto{\pgfqpoint{0.188889in}{2.114162in}}%
\pgfpathlineto{\pgfqpoint{4.838889in}{2.114162in}}%
\pgfusepath{stroke}%
\end{pgfscope}%
\begin{pgfscope}%
\pgfpathrectangle{\pgfqpoint{0.188889in}{0.454405in}}{\pgfqpoint{4.650000in}{3.020000in}}%
\pgfusepath{clip}%
\pgfsetrectcap%
\pgfsetroundjoin%
\pgfsetlinewidth{0.401500pt}%
\definecolor{currentstroke}{rgb}{0.800000,0.800000,0.800000}%
\pgfsetstrokecolor{currentstroke}%
\pgfsetstrokeopacity{0.550000}%
\pgfsetdash{}{0pt}%
\pgfpathmoveto{\pgfqpoint{0.188889in}{2.245583in}}%
\pgfpathlineto{\pgfqpoint{4.838889in}{2.245583in}}%
\pgfusepath{stroke}%
\end{pgfscope}%
\begin{pgfscope}%
\pgfpathrectangle{\pgfqpoint{0.188889in}{0.454405in}}{\pgfqpoint{4.650000in}{3.020000in}}%
\pgfusepath{clip}%
\pgfsetrectcap%
\pgfsetroundjoin%
\pgfsetlinewidth{0.401500pt}%
\definecolor{currentstroke}{rgb}{0.800000,0.800000,0.800000}%
\pgfsetstrokecolor{currentstroke}%
\pgfsetstrokeopacity{0.550000}%
\pgfsetdash{}{0pt}%
\pgfpathmoveto{\pgfqpoint{0.188889in}{2.356699in}}%
\pgfpathlineto{\pgfqpoint{4.838889in}{2.356699in}}%
\pgfusepath{stroke}%
\end{pgfscope}%
\begin{pgfscope}%
\pgfpathrectangle{\pgfqpoint{0.188889in}{0.454405in}}{\pgfqpoint{4.650000in}{3.020000in}}%
\pgfusepath{clip}%
\pgfsetrectcap%
\pgfsetroundjoin%
\pgfsetlinewidth{0.401500pt}%
\definecolor{currentstroke}{rgb}{0.800000,0.800000,0.800000}%
\pgfsetstrokecolor{currentstroke}%
\pgfsetstrokeopacity{0.550000}%
\pgfsetdash{}{0pt}%
\pgfpathmoveto{\pgfqpoint{0.188889in}{2.452951in}}%
\pgfpathlineto{\pgfqpoint{4.838889in}{2.452951in}}%
\pgfusepath{stroke}%
\end{pgfscope}%
\begin{pgfscope}%
\pgfpathrectangle{\pgfqpoint{0.188889in}{0.454405in}}{\pgfqpoint{4.650000in}{3.020000in}}%
\pgfusepath{clip}%
\pgfsetrectcap%
\pgfsetroundjoin%
\pgfsetlinewidth{0.401500pt}%
\definecolor{currentstroke}{rgb}{0.800000,0.800000,0.800000}%
\pgfsetstrokecolor{currentstroke}%
\pgfsetstrokeopacity{0.550000}%
\pgfsetdash{}{0pt}%
\pgfpathmoveto{\pgfqpoint{0.188889in}{2.537852in}}%
\pgfpathlineto{\pgfqpoint{4.838889in}{2.537852in}}%
\pgfusepath{stroke}%
\end{pgfscope}%
\begin{pgfscope}%
\pgfpathrectangle{\pgfqpoint{0.188889in}{0.454405in}}{\pgfqpoint{4.650000in}{3.020000in}}%
\pgfusepath{clip}%
\pgfsetrectcap%
\pgfsetroundjoin%
\pgfsetlinewidth{0.401500pt}%
\definecolor{currentstroke}{rgb}{0.800000,0.800000,0.800000}%
\pgfsetstrokecolor{currentstroke}%
\pgfsetstrokeopacity{0.550000}%
\pgfsetdash{}{0pt}%
\pgfpathmoveto{\pgfqpoint{0.188889in}{3.113435in}}%
\pgfpathlineto{\pgfqpoint{4.838889in}{3.113435in}}%
\pgfusepath{stroke}%
\end{pgfscope}%
\begin{pgfscope}%
\pgfpathrectangle{\pgfqpoint{0.188889in}{0.454405in}}{\pgfqpoint{4.650000in}{3.020000in}}%
\pgfusepath{clip}%
\pgfsetrectcap%
\pgfsetroundjoin%
\pgfsetlinewidth{0.401500pt}%
\definecolor{currentstroke}{rgb}{0.800000,0.800000,0.800000}%
\pgfsetstrokecolor{currentstroke}%
\pgfsetstrokeopacity{0.550000}%
\pgfsetdash{}{0pt}%
\pgfpathmoveto{\pgfqpoint{0.188889in}{3.405703in}}%
\pgfpathlineto{\pgfqpoint{4.838889in}{3.405703in}}%
\pgfusepath{stroke}%
\end{pgfscope}%
\begin{pgfscope}%
\pgfpathrectangle{\pgfqpoint{0.188889in}{0.454405in}}{\pgfqpoint{4.650000in}{3.020000in}}%
\pgfusepath{clip}%
\pgfsetbuttcap%
\pgfsetmiterjoin%
\definecolor{currentfill}{rgb}{0.347059,0.458824,0.641176}%
\pgfsetfillcolor{currentfill}%
\pgfsetlinewidth{0.803000pt}%
\definecolor{currentstroke}{rgb}{0.298039,0.298039,0.298039}%
\pgfsetstrokecolor{currentstroke}%
\pgfsetdash{}{0pt}%
\pgfpathmoveto{\pgfqpoint{0.305139in}{1.812093in}}%
\pgfpathlineto{\pgfqpoint{1.235139in}{1.812093in}}%
\pgfpathlineto{\pgfqpoint{1.235139in}{1.919858in}}%
\pgfpathlineto{\pgfqpoint{0.305139in}{1.919858in}}%
\pgfpathlineto{\pgfqpoint{0.305139in}{1.812093in}}%
\pgfpathclose%
\pgfusepath{stroke,fill}%
\end{pgfscope}%
\begin{pgfscope}%
\pgfpathrectangle{\pgfqpoint{0.188889in}{0.454405in}}{\pgfqpoint{4.650000in}{3.020000in}}%
\pgfusepath{clip}%
\pgfsetbuttcap%
\pgfsetmiterjoin%
\definecolor{currentfill}{rgb}{0.798529,0.536765,0.389706}%
\pgfsetfillcolor{currentfill}%
\pgfsetlinewidth{0.803000pt}%
\definecolor{currentstroke}{rgb}{0.298039,0.298039,0.298039}%
\pgfsetstrokecolor{currentstroke}%
\pgfsetdash{}{0pt}%
\pgfpathmoveto{\pgfqpoint{1.467639in}{1.658177in}}%
\pgfpathlineto{\pgfqpoint{2.397639in}{1.658177in}}%
\pgfpathlineto{\pgfqpoint{2.397639in}{1.823137in}}%
\pgfpathlineto{\pgfqpoint{1.467639in}{1.823137in}}%
\pgfpathlineto{\pgfqpoint{1.467639in}{1.658177in}}%
\pgfpathclose%
\pgfusepath{stroke,fill}%
\end{pgfscope}%
\begin{pgfscope}%
\pgfpathrectangle{\pgfqpoint{0.188889in}{0.454405in}}{\pgfqpoint{4.650000in}{3.020000in}}%
\pgfusepath{clip}%
\pgfsetbuttcap%
\pgfsetmiterjoin%
\definecolor{currentfill}{rgb}{0.374020,0.618137,0.429902}%
\pgfsetfillcolor{currentfill}%
\pgfsetlinewidth{0.803000pt}%
\definecolor{currentstroke}{rgb}{0.298039,0.298039,0.298039}%
\pgfsetstrokecolor{currentstroke}%
\pgfsetdash{}{0pt}%
\pgfpathmoveto{\pgfqpoint{2.630139in}{2.034003in}}%
\pgfpathlineto{\pgfqpoint{3.560139in}{2.034003in}}%
\pgfpathlineto{\pgfqpoint{3.560139in}{2.163029in}}%
\pgfpathlineto{\pgfqpoint{2.630139in}{2.163029in}}%
\pgfpathlineto{\pgfqpoint{2.630139in}{2.034003in}}%
\pgfpathclose%
\pgfusepath{stroke,fill}%
\end{pgfscope}%
\begin{pgfscope}%
\pgfpathrectangle{\pgfqpoint{0.188889in}{0.454405in}}{\pgfqpoint{4.650000in}{3.020000in}}%
\pgfusepath{clip}%
\pgfsetbuttcap%
\pgfsetmiterjoin%
\definecolor{currentfill}{rgb}{0.710784,0.363725,0.375490}%
\pgfsetfillcolor{currentfill}%
\pgfsetlinewidth{0.803000pt}%
\definecolor{currentstroke}{rgb}{0.298039,0.298039,0.298039}%
\pgfsetstrokecolor{currentstroke}%
\pgfsetdash{}{0pt}%
\pgfpathmoveto{\pgfqpoint{3.792639in}{2.381828in}}%
\pgfpathlineto{\pgfqpoint{4.722639in}{2.381828in}}%
\pgfpathlineto{\pgfqpoint{4.722639in}{2.524601in}}%
\pgfpathlineto{\pgfqpoint{3.792639in}{2.524601in}}%
\pgfpathlineto{\pgfqpoint{3.792639in}{2.381828in}}%
\pgfpathclose%
\pgfusepath{stroke,fill}%
\end{pgfscope}%
\begin{pgfscope}%
\pgfsetrectcap%
\pgfsetmiterjoin%
\pgfsetlinewidth{1.254687pt}%
\definecolor{currentstroke}{rgb}{0.800000,0.800000,0.800000}%
\pgfsetstrokecolor{currentstroke}%
\pgfsetdash{}{0pt}%
\pgfpathmoveto{\pgfqpoint{0.188889in}{0.454405in}}%
\pgfpathlineto{\pgfqpoint{0.188889in}{3.474405in}}%
\pgfusepath{stroke}%
\end{pgfscope}%
\begin{pgfscope}%
\pgfsetrectcap%
\pgfsetmiterjoin%
\pgfsetlinewidth{1.254687pt}%
\definecolor{currentstroke}{rgb}{0.800000,0.800000,0.800000}%
\pgfsetstrokecolor{currentstroke}%
\pgfsetdash{}{0pt}%
\pgfpathmoveto{\pgfqpoint{4.838889in}{0.454405in}}%
\pgfpathlineto{\pgfqpoint{4.838889in}{3.474405in}}%
\pgfusepath{stroke}%
\end{pgfscope}%
\begin{pgfscope}%
\pgfsetrectcap%
\pgfsetmiterjoin%
\pgfsetlinewidth{1.254687pt}%
\definecolor{currentstroke}{rgb}{0.800000,0.800000,0.800000}%
\pgfsetstrokecolor{currentstroke}%
\pgfsetdash{}{0pt}%
\pgfpathmoveto{\pgfqpoint{0.188889in}{0.454405in}}%
\pgfpathlineto{\pgfqpoint{4.838889in}{0.454405in}}%
\pgfusepath{stroke}%
\end{pgfscope}%
\begin{pgfscope}%
\pgfsetrectcap%
\pgfsetmiterjoin%
\pgfsetlinewidth{1.254687pt}%
\definecolor{currentstroke}{rgb}{0.800000,0.800000,0.800000}%
\pgfsetstrokecolor{currentstroke}%
\pgfsetdash{}{0pt}%
\pgfpathmoveto{\pgfqpoint{0.188889in}{3.474405in}}%
\pgfpathlineto{\pgfqpoint{4.838889in}{3.474405in}}%
\pgfusepath{stroke}%
\end{pgfscope}%
\begin{pgfscope}%
\pgfpathrectangle{\pgfqpoint{0.188889in}{0.454405in}}{\pgfqpoint{4.650000in}{3.020000in}}%
\pgfusepath{clip}%
\pgfsetrectcap%
\pgfsetroundjoin%
\pgfsetlinewidth{0.803000pt}%
\definecolor{currentstroke}{rgb}{0.298039,0.298039,0.298039}%
\pgfsetstrokecolor{currentstroke}%
\pgfsetdash{}{0pt}%
\pgfpathmoveto{\pgfqpoint{0.770139in}{1.812093in}}%
\pgfpathlineto{\pgfqpoint{0.770139in}{1.657900in}}%
\pgfusepath{stroke}%
\end{pgfscope}%
\begin{pgfscope}%
\pgfpathrectangle{\pgfqpoint{0.188889in}{0.454405in}}{\pgfqpoint{4.650000in}{3.020000in}}%
\pgfusepath{clip}%
\pgfsetrectcap%
\pgfsetroundjoin%
\pgfsetlinewidth{0.803000pt}%
\definecolor{currentstroke}{rgb}{0.298039,0.298039,0.298039}%
\pgfsetstrokecolor{currentstroke}%
\pgfsetdash{}{0pt}%
\pgfpathmoveto{\pgfqpoint{0.770139in}{1.919858in}}%
\pgfpathlineto{\pgfqpoint{0.770139in}{2.039663in}}%
\pgfusepath{stroke}%
\end{pgfscope}%
\begin{pgfscope}%
\pgfpathrectangle{\pgfqpoint{0.188889in}{0.454405in}}{\pgfqpoint{4.650000in}{3.020000in}}%
\pgfusepath{clip}%
\pgfsetrectcap%
\pgfsetroundjoin%
\pgfsetlinewidth{0.803000pt}%
\definecolor{currentstroke}{rgb}{0.298039,0.298039,0.298039}%
\pgfsetstrokecolor{currentstroke}%
\pgfsetdash{}{0pt}%
\pgfpathmoveto{\pgfqpoint{0.537639in}{1.657900in}}%
\pgfpathlineto{\pgfqpoint{1.002639in}{1.657900in}}%
\pgfusepath{stroke}%
\end{pgfscope}%
\begin{pgfscope}%
\pgfpathrectangle{\pgfqpoint{0.188889in}{0.454405in}}{\pgfqpoint{4.650000in}{3.020000in}}%
\pgfusepath{clip}%
\pgfsetrectcap%
\pgfsetroundjoin%
\pgfsetlinewidth{0.803000pt}%
\definecolor{currentstroke}{rgb}{0.298039,0.298039,0.298039}%
\pgfsetstrokecolor{currentstroke}%
\pgfsetdash{}{0pt}%
\pgfpathmoveto{\pgfqpoint{0.537639in}{2.039663in}}%
\pgfpathlineto{\pgfqpoint{1.002639in}{2.039663in}}%
\pgfusepath{stroke}%
\end{pgfscope}%
\begin{pgfscope}%
\pgfpathrectangle{\pgfqpoint{0.188889in}{0.454405in}}{\pgfqpoint{4.650000in}{3.020000in}}%
\pgfusepath{clip}%
\pgfsetrectcap%
\pgfsetroundjoin%
\pgfsetlinewidth{0.803000pt}%
\definecolor{currentstroke}{rgb}{0.298039,0.298039,0.298039}%
\pgfsetstrokecolor{currentstroke}%
\pgfsetdash{}{0pt}%
\pgfpathmoveto{\pgfqpoint{0.305139in}{1.812093in}}%
\pgfpathlineto{\pgfqpoint{1.235139in}{1.812093in}}%
\pgfusepath{stroke}%
\end{pgfscope}%
\begin{pgfscope}%
\pgfpathrectangle{\pgfqpoint{0.188889in}{0.454405in}}{\pgfqpoint{4.650000in}{3.020000in}}%
\pgfusepath{clip}%
\pgfsetbuttcap%
\pgfsetmiterjoin%
\definecolor{currentfill}{rgb}{0.298039,0.298039,0.298039}%
\pgfsetfillcolor{currentfill}%
\pgfsetlinewidth{1.003750pt}%
\definecolor{currentstroke}{rgb}{0.298039,0.298039,0.298039}%
\pgfsetstrokecolor{currentstroke}%
\pgfsetdash{}{0pt}%
\pgfsys@defobject{currentmarker}{\pgfqpoint{-0.029463in}{-0.049105in}}{\pgfqpoint{0.029463in}{0.049105in}}{%
\pgfpathmoveto{\pgfqpoint{0.000000in}{-0.049105in}}%
\pgfpathlineto{\pgfqpoint{0.029463in}{0.000000in}}%
\pgfpathlineto{\pgfqpoint{0.000000in}{0.049105in}}%
\pgfpathlineto{\pgfqpoint{-0.029463in}{0.000000in}}%
\pgfpathclose%
\pgfusepath{stroke,fill}%
}%
\begin{pgfscope}%
\pgfsys@transformshift{0.770139in}{2.121423in}%
\pgfsys@useobject{currentmarker}{}%
\end{pgfscope}%
\end{pgfscope}%
\begin{pgfscope}%
\pgfpathrectangle{\pgfqpoint{0.188889in}{0.454405in}}{\pgfqpoint{4.650000in}{3.020000in}}%
\pgfusepath{clip}%
\pgfsetrectcap%
\pgfsetroundjoin%
\pgfsetlinewidth{0.803000pt}%
\definecolor{currentstroke}{rgb}{0.298039,0.298039,0.298039}%
\pgfsetstrokecolor{currentstroke}%
\pgfsetdash{}{0pt}%
\pgfpathmoveto{\pgfqpoint{1.932639in}{1.658177in}}%
\pgfpathlineto{\pgfqpoint{1.932639in}{1.518007in}}%
\pgfusepath{stroke}%
\end{pgfscope}%
\begin{pgfscope}%
\pgfpathrectangle{\pgfqpoint{0.188889in}{0.454405in}}{\pgfqpoint{4.650000in}{3.020000in}}%
\pgfusepath{clip}%
\pgfsetrectcap%
\pgfsetroundjoin%
\pgfsetlinewidth{0.803000pt}%
\definecolor{currentstroke}{rgb}{0.298039,0.298039,0.298039}%
\pgfsetstrokecolor{currentstroke}%
\pgfsetdash{}{0pt}%
\pgfpathmoveto{\pgfqpoint{1.932639in}{1.823137in}}%
\pgfpathlineto{\pgfqpoint{1.932639in}{1.963473in}}%
\pgfusepath{stroke}%
\end{pgfscope}%
\begin{pgfscope}%
\pgfpathrectangle{\pgfqpoint{0.188889in}{0.454405in}}{\pgfqpoint{4.650000in}{3.020000in}}%
\pgfusepath{clip}%
\pgfsetrectcap%
\pgfsetroundjoin%
\pgfsetlinewidth{0.803000pt}%
\definecolor{currentstroke}{rgb}{0.298039,0.298039,0.298039}%
\pgfsetstrokecolor{currentstroke}%
\pgfsetdash{}{0pt}%
\pgfpathmoveto{\pgfqpoint{1.700139in}{1.518007in}}%
\pgfpathlineto{\pgfqpoint{2.165139in}{1.518007in}}%
\pgfusepath{stroke}%
\end{pgfscope}%
\begin{pgfscope}%
\pgfpathrectangle{\pgfqpoint{0.188889in}{0.454405in}}{\pgfqpoint{4.650000in}{3.020000in}}%
\pgfusepath{clip}%
\pgfsetrectcap%
\pgfsetroundjoin%
\pgfsetlinewidth{0.803000pt}%
\definecolor{currentstroke}{rgb}{0.298039,0.298039,0.298039}%
\pgfsetstrokecolor{currentstroke}%
\pgfsetdash{}{0pt}%
\pgfpathmoveto{\pgfqpoint{1.700139in}{1.963473in}}%
\pgfpathlineto{\pgfqpoint{2.165139in}{1.963473in}}%
\pgfusepath{stroke}%
\end{pgfscope}%
\begin{pgfscope}%
\pgfpathrectangle{\pgfqpoint{0.188889in}{0.454405in}}{\pgfqpoint{4.650000in}{3.020000in}}%
\pgfusepath{clip}%
\pgfsetrectcap%
\pgfsetroundjoin%
\pgfsetlinewidth{0.803000pt}%
\definecolor{currentstroke}{rgb}{0.298039,0.298039,0.298039}%
\pgfsetstrokecolor{currentstroke}%
\pgfsetdash{}{0pt}%
\pgfpathmoveto{\pgfqpoint{1.467639in}{1.783990in}}%
\pgfpathlineto{\pgfqpoint{2.397639in}{1.783990in}}%
\pgfusepath{stroke}%
\end{pgfscope}%
\begin{pgfscope}%
\pgfpathrectangle{\pgfqpoint{0.188889in}{0.454405in}}{\pgfqpoint{4.650000in}{3.020000in}}%
\pgfusepath{clip}%
\pgfsetrectcap%
\pgfsetroundjoin%
\pgfsetlinewidth{0.803000pt}%
\definecolor{currentstroke}{rgb}{0.298039,0.298039,0.298039}%
\pgfsetstrokecolor{currentstroke}%
\pgfsetdash{}{0pt}%
\pgfpathmoveto{\pgfqpoint{3.095139in}{2.034003in}}%
\pgfpathlineto{\pgfqpoint{3.095139in}{1.859257in}}%
\pgfusepath{stroke}%
\end{pgfscope}%
\begin{pgfscope}%
\pgfpathrectangle{\pgfqpoint{0.188889in}{0.454405in}}{\pgfqpoint{4.650000in}{3.020000in}}%
\pgfusepath{clip}%
\pgfsetrectcap%
\pgfsetroundjoin%
\pgfsetlinewidth{0.803000pt}%
\definecolor{currentstroke}{rgb}{0.298039,0.298039,0.298039}%
\pgfsetstrokecolor{currentstroke}%
\pgfsetdash{}{0pt}%
\pgfpathmoveto{\pgfqpoint{3.095139in}{2.163029in}}%
\pgfpathlineto{\pgfqpoint{3.095139in}{2.273047in}}%
\pgfusepath{stroke}%
\end{pgfscope}%
\begin{pgfscope}%
\pgfpathrectangle{\pgfqpoint{0.188889in}{0.454405in}}{\pgfqpoint{4.650000in}{3.020000in}}%
\pgfusepath{clip}%
\pgfsetrectcap%
\pgfsetroundjoin%
\pgfsetlinewidth{0.803000pt}%
\definecolor{currentstroke}{rgb}{0.298039,0.298039,0.298039}%
\pgfsetstrokecolor{currentstroke}%
\pgfsetdash{}{0pt}%
\pgfpathmoveto{\pgfqpoint{2.862639in}{1.859257in}}%
\pgfpathlineto{\pgfqpoint{3.327639in}{1.859257in}}%
\pgfusepath{stroke}%
\end{pgfscope}%
\begin{pgfscope}%
\pgfpathrectangle{\pgfqpoint{0.188889in}{0.454405in}}{\pgfqpoint{4.650000in}{3.020000in}}%
\pgfusepath{clip}%
\pgfsetrectcap%
\pgfsetroundjoin%
\pgfsetlinewidth{0.803000pt}%
\definecolor{currentstroke}{rgb}{0.298039,0.298039,0.298039}%
\pgfsetstrokecolor{currentstroke}%
\pgfsetdash{}{0pt}%
\pgfpathmoveto{\pgfqpoint{2.862639in}{2.273047in}}%
\pgfpathlineto{\pgfqpoint{3.327639in}{2.273047in}}%
\pgfusepath{stroke}%
\end{pgfscope}%
\begin{pgfscope}%
\pgfpathrectangle{\pgfqpoint{0.188889in}{0.454405in}}{\pgfqpoint{4.650000in}{3.020000in}}%
\pgfusepath{clip}%
\pgfsetrectcap%
\pgfsetroundjoin%
\pgfsetlinewidth{0.803000pt}%
\definecolor{currentstroke}{rgb}{0.298039,0.298039,0.298039}%
\pgfsetstrokecolor{currentstroke}%
\pgfsetdash{}{0pt}%
\pgfpathmoveto{\pgfqpoint{2.630139in}{2.152573in}}%
\pgfpathlineto{\pgfqpoint{3.560139in}{2.152573in}}%
\pgfusepath{stroke}%
\end{pgfscope}%
\begin{pgfscope}%
\pgfpathrectangle{\pgfqpoint{0.188889in}{0.454405in}}{\pgfqpoint{4.650000in}{3.020000in}}%
\pgfusepath{clip}%
\pgfsetbuttcap%
\pgfsetmiterjoin%
\definecolor{currentfill}{rgb}{0.298039,0.298039,0.298039}%
\pgfsetfillcolor{currentfill}%
\pgfsetlinewidth{1.003750pt}%
\definecolor{currentstroke}{rgb}{0.298039,0.298039,0.298039}%
\pgfsetstrokecolor{currentstroke}%
\pgfsetdash{}{0pt}%
\pgfsys@defobject{currentmarker}{\pgfqpoint{-0.029463in}{-0.049105in}}{\pgfqpoint{0.029463in}{0.049105in}}{%
\pgfpathmoveto{\pgfqpoint{0.000000in}{-0.049105in}}%
\pgfpathlineto{\pgfqpoint{0.029463in}{0.000000in}}%
\pgfpathlineto{\pgfqpoint{0.000000in}{0.049105in}}%
\pgfpathlineto{\pgfqpoint{-0.029463in}{0.000000in}}%
\pgfpathclose%
\pgfusepath{stroke,fill}%
}%
\begin{pgfscope}%
\pgfsys@transformshift{3.095139in}{2.414844in}%
\pgfsys@useobject{currentmarker}{}%
\end{pgfscope}%
\end{pgfscope}%
\begin{pgfscope}%
\pgfpathrectangle{\pgfqpoint{0.188889in}{0.454405in}}{\pgfqpoint{4.650000in}{3.020000in}}%
\pgfusepath{clip}%
\pgfsetrectcap%
\pgfsetroundjoin%
\pgfsetlinewidth{0.803000pt}%
\definecolor{currentstroke}{rgb}{0.298039,0.298039,0.298039}%
\pgfsetstrokecolor{currentstroke}%
\pgfsetdash{}{0pt}%
\pgfpathmoveto{\pgfqpoint{4.257639in}{2.381828in}}%
\pgfpathlineto{\pgfqpoint{4.257639in}{2.197660in}}%
\pgfusepath{stroke}%
\end{pgfscope}%
\begin{pgfscope}%
\pgfpathrectangle{\pgfqpoint{0.188889in}{0.454405in}}{\pgfqpoint{4.650000in}{3.020000in}}%
\pgfusepath{clip}%
\pgfsetrectcap%
\pgfsetroundjoin%
\pgfsetlinewidth{0.803000pt}%
\definecolor{currentstroke}{rgb}{0.298039,0.298039,0.298039}%
\pgfsetstrokecolor{currentstroke}%
\pgfsetdash{}{0pt}%
\pgfpathmoveto{\pgfqpoint{4.257639in}{2.524601in}}%
\pgfpathlineto{\pgfqpoint{4.257639in}{2.581862in}}%
\pgfusepath{stroke}%
\end{pgfscope}%
\begin{pgfscope}%
\pgfpathrectangle{\pgfqpoint{0.188889in}{0.454405in}}{\pgfqpoint{4.650000in}{3.020000in}}%
\pgfusepath{clip}%
\pgfsetrectcap%
\pgfsetroundjoin%
\pgfsetlinewidth{0.803000pt}%
\definecolor{currentstroke}{rgb}{0.298039,0.298039,0.298039}%
\pgfsetstrokecolor{currentstroke}%
\pgfsetdash{}{0pt}%
\pgfpathmoveto{\pgfqpoint{4.025139in}{2.197660in}}%
\pgfpathlineto{\pgfqpoint{4.490139in}{2.197660in}}%
\pgfusepath{stroke}%
\end{pgfscope}%
\begin{pgfscope}%
\pgfpathrectangle{\pgfqpoint{0.188889in}{0.454405in}}{\pgfqpoint{4.650000in}{3.020000in}}%
\pgfusepath{clip}%
\pgfsetrectcap%
\pgfsetroundjoin%
\pgfsetlinewidth{0.803000pt}%
\definecolor{currentstroke}{rgb}{0.298039,0.298039,0.298039}%
\pgfsetstrokecolor{currentstroke}%
\pgfsetdash{}{0pt}%
\pgfpathmoveto{\pgfqpoint{4.025139in}{2.581862in}}%
\pgfpathlineto{\pgfqpoint{4.490139in}{2.581862in}}%
\pgfusepath{stroke}%
\end{pgfscope}%
\begin{pgfscope}%
\pgfpathrectangle{\pgfqpoint{0.188889in}{0.454405in}}{\pgfqpoint{4.650000in}{3.020000in}}%
\pgfusepath{clip}%
\pgfsetrectcap%
\pgfsetroundjoin%
\pgfsetlinewidth{0.803000pt}%
\definecolor{currentstroke}{rgb}{0.298039,0.298039,0.298039}%
\pgfsetstrokecolor{currentstroke}%
\pgfsetdash{}{0pt}%
\pgfpathmoveto{\pgfqpoint{3.792639in}{2.508410in}}%
\pgfpathlineto{\pgfqpoint{4.722639in}{2.508410in}}%
\pgfusepath{stroke}%
\end{pgfscope}%
\end{pgfpicture}%
\makeatother%
\endgroup%

%% file: fig/boxplot_two_circles_res.pgf
%% Creator: Matplotlib, PGF backend
%%
%% To include the figure in your LaTeX document, write
%%   \input{<filename>.pgf}
%%
%% Make sure the required packages are loaded in your preamble
%%   \usepackage{pgf}
%%
%% Figures using additional raster images can only be included by \input if
%% they are in the same directory as the main LaTeX file. For loading figures
%% from other directories you can use the `import` package
%% and then include the figures with
%%   \import{<path to file>}{<filename>.pgf}
%%
%% Matplotlib used the following preamble
%%
\begingroup%
\makeatletter%
\begin{pgfpicture}%
\pgfpathrectangle{\pgfpointorigin}{\pgfqpoint{4.888889in}{3.524405in}}%
\pgfusepath{use as bounding box, clip}%
\begin{pgfscope}%
\pgfsetbuttcap%
\pgfsetmiterjoin%
\definecolor{currentfill}{rgb}{1.000000,1.000000,1.000000}%
\pgfsetfillcolor{currentfill}%
\pgfsetlinewidth{0.000000pt}%
\definecolor{currentstroke}{rgb}{1.000000,1.000000,1.000000}%
\pgfsetstrokecolor{currentstroke}%
\pgfsetdash{}{0pt}%
\pgfpathmoveto{\pgfqpoint{0.000000in}{0.000000in}}%
\pgfpathlineto{\pgfqpoint{4.888889in}{0.000000in}}%
\pgfpathlineto{\pgfqpoint{4.888889in}{3.524405in}}%
\pgfpathlineto{\pgfqpoint{0.000000in}{3.524405in}}%
\pgfpathclose%
\pgfusepath{fill}%
\end{pgfscope}%
\begin{pgfscope}%
\pgfsetbuttcap%
\pgfsetmiterjoin%
\definecolor{currentfill}{rgb}{1.000000,1.000000,1.000000}%
\pgfsetfillcolor{currentfill}%
\pgfsetlinewidth{0.000000pt}%
\definecolor{currentstroke}{rgb}{0.000000,0.000000,0.000000}%
\pgfsetstrokecolor{currentstroke}%
\pgfsetstrokeopacity{0.000000}%
\pgfsetdash{}{0pt}%
\pgfpathmoveto{\pgfqpoint{0.188889in}{0.454405in}}%
\pgfpathlineto{\pgfqpoint{4.838889in}{0.454405in}}%
\pgfpathlineto{\pgfqpoint{4.838889in}{3.474405in}}%
\pgfpathlineto{\pgfqpoint{0.188889in}{3.474405in}}%
\pgfpathclose%
\pgfusepath{fill}%
\end{pgfscope}%
\begin{pgfscope}%
\definecolor{textcolor}{rgb}{0.000000,0.000000,0.000000}%
\pgfsetstrokecolor{textcolor}%
\pgfsetfillcolor{textcolor}%
\pgftext[x=0.770139in,y=0.357183in,,top]{\color{textcolor}\rmfamily\fontsize{24.200000}{29.040000}\selectfont \textsc{w-l}}%
\end{pgfscope}%
\begin{pgfscope}%
\definecolor{textcolor}{rgb}{0.000000,0.000000,0.000000}%
\pgfsetstrokecolor{textcolor}%
\pgfsetfillcolor{textcolor}%
\pgftext[x=1.932639in,y=0.357183in,,top]{\color{textcolor}\rmfamily\fontsize{24.200000}{29.040000}\selectfont \textsc{w-m}}%
\end{pgfscope}%
\begin{pgfscope}%
\definecolor{textcolor}{rgb}{0.000000,0.000000,0.000000}%
\pgfsetstrokecolor{textcolor}%
\pgfsetfillcolor{textcolor}%
\pgftext[x=3.095139in,y=0.357183in,,top]{\color{textcolor}\rmfamily\fontsize{24.200000}{29.040000}\selectfont \textsc{dcv}}%
\end{pgfscope}%
\begin{pgfscope}%
\definecolor{textcolor}{rgb}{0.000000,0.000000,0.000000}%
\pgfsetstrokecolor{textcolor}%
\pgfsetfillcolor{textcolor}%
\pgftext[x=4.257639in,y=0.357183in,,top]{\color{textcolor}\rmfamily\fontsize{24.200000}{29.040000}\selectfont \textsc{mc}}%
\end{pgfscope}%
\begin{pgfscope}%
\pgfpathrectangle{\pgfqpoint{0.188889in}{0.454405in}}{\pgfqpoint{4.650000in}{3.020000in}}%
\pgfusepath{clip}%
\pgfsetrectcap%
\pgfsetroundjoin%
\pgfsetlinewidth{2.509375pt}%
\definecolor{currentstroke}{rgb}{0.800000,0.800000,0.800000}%
\pgfsetstrokecolor{currentstroke}%
\pgfsetdash{}{0pt}%
\pgfpathmoveto{\pgfqpoint{0.188889in}{0.954042in}}%
\pgfpathlineto{\pgfqpoint{4.838889in}{0.954042in}}%
\pgfusepath{stroke}%
\end{pgfscope}%
\begin{pgfscope}%
\pgfsetbuttcap%
\pgfsetroundjoin%
\definecolor{currentfill}{rgb}{0.800000,0.800000,0.800000}%
\pgfsetfillcolor{currentfill}%
\pgfsetlinewidth{1.003750pt}%
\definecolor{currentstroke}{rgb}{0.800000,0.800000,0.800000}%
\pgfsetstrokecolor{currentstroke}%
\pgfsetdash{}{0pt}%
\pgfsys@defobject{currentmarker}{\pgfqpoint{-0.138889in}{0.000000in}}{\pgfqpoint{0.000000in}{0.000000in}}{%
\pgfpathmoveto{\pgfqpoint{0.000000in}{0.000000in}}%
\pgfpathlineto{\pgfqpoint{-0.138889in}{0.000000in}}%
\pgfusepath{stroke,fill}%
}%
\begin{pgfscope}%
\pgfsys@transformshift{0.188889in}{0.954042in}%
\pgfsys@useobject{currentmarker}{}%
\end{pgfscope}%
\end{pgfscope}%
\begin{pgfscope}%
\pgfpathrectangle{\pgfqpoint{0.188889in}{0.454405in}}{\pgfqpoint{4.650000in}{3.020000in}}%
\pgfusepath{clip}%
\pgfsetrectcap%
\pgfsetroundjoin%
\pgfsetlinewidth{2.509375pt}%
\definecolor{currentstroke}{rgb}{0.611765,0.611765,0.611765}%
\pgfsetstrokecolor{currentstroke}%
\pgfsetdash{}{0pt}%
\pgfpathmoveto{\pgfqpoint{0.188889in}{2.613798in}}%
\pgfpathlineto{\pgfqpoint{4.838889in}{2.613798in}}%
\pgfusepath{stroke}%
\end{pgfscope}%
\begin{pgfscope}%
\pgfsetbuttcap%
\pgfsetroundjoin%
\definecolor{currentfill}{rgb}{0.800000,0.800000,0.800000}%
\pgfsetfillcolor{currentfill}%
\pgfsetlinewidth{1.003750pt}%
\definecolor{currentstroke}{rgb}{0.800000,0.800000,0.800000}%
\pgfsetstrokecolor{currentstroke}%
\pgfsetdash{}{0pt}%
\pgfsys@defobject{currentmarker}{\pgfqpoint{-0.138889in}{0.000000in}}{\pgfqpoint{0.000000in}{0.000000in}}{%
\pgfpathmoveto{\pgfqpoint{0.000000in}{0.000000in}}%
\pgfpathlineto{\pgfqpoint{-0.138889in}{0.000000in}}%
\pgfusepath{stroke,fill}%
}%
\begin{pgfscope}%
\pgfsys@transformshift{0.188889in}{2.613798in}%
\pgfsys@useobject{currentmarker}{}%
\end{pgfscope}%
\end{pgfscope}%
\begin{pgfscope}%
\pgfpathrectangle{\pgfqpoint{0.188889in}{0.454405in}}{\pgfqpoint{4.650000in}{3.020000in}}%
\pgfusepath{clip}%
\pgfsetrectcap%
\pgfsetroundjoin%
\pgfsetlinewidth{0.401500pt}%
\definecolor{currentstroke}{rgb}{0.800000,0.800000,0.800000}%
\pgfsetstrokecolor{currentstroke}%
\pgfsetstrokeopacity{0.550000}%
\pgfsetdash{}{0pt}%
\pgfpathmoveto{\pgfqpoint{0.188889in}{0.454405in}}%
\pgfpathlineto{\pgfqpoint{4.838889in}{0.454405in}}%
\pgfusepath{stroke}%
\end{pgfscope}%
\begin{pgfscope}%
\pgfpathrectangle{\pgfqpoint{0.188889in}{0.454405in}}{\pgfqpoint{4.650000in}{3.020000in}}%
\pgfusepath{clip}%
\pgfsetrectcap%
\pgfsetroundjoin%
\pgfsetlinewidth{0.401500pt}%
\definecolor{currentstroke}{rgb}{0.800000,0.800000,0.800000}%
\pgfsetstrokecolor{currentstroke}%
\pgfsetstrokeopacity{0.550000}%
\pgfsetdash{}{0pt}%
\pgfpathmoveto{\pgfqpoint{0.188889in}{0.585827in}}%
\pgfpathlineto{\pgfqpoint{4.838889in}{0.585827in}}%
\pgfusepath{stroke}%
\end{pgfscope}%
\begin{pgfscope}%
\pgfpathrectangle{\pgfqpoint{0.188889in}{0.454405in}}{\pgfqpoint{4.650000in}{3.020000in}}%
\pgfusepath{clip}%
\pgfsetrectcap%
\pgfsetroundjoin%
\pgfsetlinewidth{0.401500pt}%
\definecolor{currentstroke}{rgb}{0.800000,0.800000,0.800000}%
\pgfsetstrokecolor{currentstroke}%
\pgfsetstrokeopacity{0.550000}%
\pgfsetdash{}{0pt}%
\pgfpathmoveto{\pgfqpoint{0.188889in}{0.696942in}}%
\pgfpathlineto{\pgfqpoint{4.838889in}{0.696942in}}%
\pgfusepath{stroke}%
\end{pgfscope}%
\begin{pgfscope}%
\pgfpathrectangle{\pgfqpoint{0.188889in}{0.454405in}}{\pgfqpoint{4.650000in}{3.020000in}}%
\pgfusepath{clip}%
\pgfsetrectcap%
\pgfsetroundjoin%
\pgfsetlinewidth{0.401500pt}%
\definecolor{currentstroke}{rgb}{0.800000,0.800000,0.800000}%
\pgfsetstrokecolor{currentstroke}%
\pgfsetstrokeopacity{0.550000}%
\pgfsetdash{}{0pt}%
\pgfpathmoveto{\pgfqpoint{0.188889in}{0.793194in}}%
\pgfpathlineto{\pgfqpoint{4.838889in}{0.793194in}}%
\pgfusepath{stroke}%
\end{pgfscope}%
\begin{pgfscope}%
\pgfpathrectangle{\pgfqpoint{0.188889in}{0.454405in}}{\pgfqpoint{4.650000in}{3.020000in}}%
\pgfusepath{clip}%
\pgfsetrectcap%
\pgfsetroundjoin%
\pgfsetlinewidth{0.401500pt}%
\definecolor{currentstroke}{rgb}{0.800000,0.800000,0.800000}%
\pgfsetstrokecolor{currentstroke}%
\pgfsetstrokeopacity{0.550000}%
\pgfsetdash{}{0pt}%
\pgfpathmoveto{\pgfqpoint{0.188889in}{0.878095in}}%
\pgfpathlineto{\pgfqpoint{4.838889in}{0.878095in}}%
\pgfusepath{stroke}%
\end{pgfscope}%
\begin{pgfscope}%
\pgfpathrectangle{\pgfqpoint{0.188889in}{0.454405in}}{\pgfqpoint{4.650000in}{3.020000in}}%
\pgfusepath{clip}%
\pgfsetrectcap%
\pgfsetroundjoin%
\pgfsetlinewidth{0.401500pt}%
\definecolor{currentstroke}{rgb}{0.800000,0.800000,0.800000}%
\pgfsetstrokecolor{currentstroke}%
\pgfsetstrokeopacity{0.550000}%
\pgfsetdash{}{0pt}%
\pgfpathmoveto{\pgfqpoint{0.188889in}{1.453678in}}%
\pgfpathlineto{\pgfqpoint{4.838889in}{1.453678in}}%
\pgfusepath{stroke}%
\end{pgfscope}%
\begin{pgfscope}%
\pgfpathrectangle{\pgfqpoint{0.188889in}{0.454405in}}{\pgfqpoint{4.650000in}{3.020000in}}%
\pgfusepath{clip}%
\pgfsetrectcap%
\pgfsetroundjoin%
\pgfsetlinewidth{0.401500pt}%
\definecolor{currentstroke}{rgb}{0.800000,0.800000,0.800000}%
\pgfsetstrokecolor{currentstroke}%
\pgfsetstrokeopacity{0.550000}%
\pgfsetdash{}{0pt}%
\pgfpathmoveto{\pgfqpoint{0.188889in}{1.745947in}}%
\pgfpathlineto{\pgfqpoint{4.838889in}{1.745947in}}%
\pgfusepath{stroke}%
\end{pgfscope}%
\begin{pgfscope}%
\pgfpathrectangle{\pgfqpoint{0.188889in}{0.454405in}}{\pgfqpoint{4.650000in}{3.020000in}}%
\pgfusepath{clip}%
\pgfsetrectcap%
\pgfsetroundjoin%
\pgfsetlinewidth{0.401500pt}%
\definecolor{currentstroke}{rgb}{0.800000,0.800000,0.800000}%
\pgfsetstrokecolor{currentstroke}%
\pgfsetstrokeopacity{0.550000}%
\pgfsetdash{}{0pt}%
\pgfpathmoveto{\pgfqpoint{0.188889in}{1.953315in}}%
\pgfpathlineto{\pgfqpoint{4.838889in}{1.953315in}}%
\pgfusepath{stroke}%
\end{pgfscope}%
\begin{pgfscope}%
\pgfpathrectangle{\pgfqpoint{0.188889in}{0.454405in}}{\pgfqpoint{4.650000in}{3.020000in}}%
\pgfusepath{clip}%
\pgfsetrectcap%
\pgfsetroundjoin%
\pgfsetlinewidth{0.401500pt}%
\definecolor{currentstroke}{rgb}{0.800000,0.800000,0.800000}%
\pgfsetstrokecolor{currentstroke}%
\pgfsetstrokeopacity{0.550000}%
\pgfsetdash{}{0pt}%
\pgfpathmoveto{\pgfqpoint{0.188889in}{2.114162in}}%
\pgfpathlineto{\pgfqpoint{4.838889in}{2.114162in}}%
\pgfusepath{stroke}%
\end{pgfscope}%
\begin{pgfscope}%
\pgfpathrectangle{\pgfqpoint{0.188889in}{0.454405in}}{\pgfqpoint{4.650000in}{3.020000in}}%
\pgfusepath{clip}%
\pgfsetrectcap%
\pgfsetroundjoin%
\pgfsetlinewidth{0.401500pt}%
\definecolor{currentstroke}{rgb}{0.800000,0.800000,0.800000}%
\pgfsetstrokecolor{currentstroke}%
\pgfsetstrokeopacity{0.550000}%
\pgfsetdash{}{0pt}%
\pgfpathmoveto{\pgfqpoint{0.188889in}{2.245583in}}%
\pgfpathlineto{\pgfqpoint{4.838889in}{2.245583in}}%
\pgfusepath{stroke}%
\end{pgfscope}%
\begin{pgfscope}%
\pgfpathrectangle{\pgfqpoint{0.188889in}{0.454405in}}{\pgfqpoint{4.650000in}{3.020000in}}%
\pgfusepath{clip}%
\pgfsetrectcap%
\pgfsetroundjoin%
\pgfsetlinewidth{0.401500pt}%
\definecolor{currentstroke}{rgb}{0.800000,0.800000,0.800000}%
\pgfsetstrokecolor{currentstroke}%
\pgfsetstrokeopacity{0.550000}%
\pgfsetdash{}{0pt}%
\pgfpathmoveto{\pgfqpoint{0.188889in}{2.356699in}}%
\pgfpathlineto{\pgfqpoint{4.838889in}{2.356699in}}%
\pgfusepath{stroke}%
\end{pgfscope}%
\begin{pgfscope}%
\pgfpathrectangle{\pgfqpoint{0.188889in}{0.454405in}}{\pgfqpoint{4.650000in}{3.020000in}}%
\pgfusepath{clip}%
\pgfsetrectcap%
\pgfsetroundjoin%
\pgfsetlinewidth{0.401500pt}%
\definecolor{currentstroke}{rgb}{0.800000,0.800000,0.800000}%
\pgfsetstrokecolor{currentstroke}%
\pgfsetstrokeopacity{0.550000}%
\pgfsetdash{}{0pt}%
\pgfpathmoveto{\pgfqpoint{0.188889in}{2.452951in}}%
\pgfpathlineto{\pgfqpoint{4.838889in}{2.452951in}}%
\pgfusepath{stroke}%
\end{pgfscope}%
\begin{pgfscope}%
\pgfpathrectangle{\pgfqpoint{0.188889in}{0.454405in}}{\pgfqpoint{4.650000in}{3.020000in}}%
\pgfusepath{clip}%
\pgfsetrectcap%
\pgfsetroundjoin%
\pgfsetlinewidth{0.401500pt}%
\definecolor{currentstroke}{rgb}{0.800000,0.800000,0.800000}%
\pgfsetstrokecolor{currentstroke}%
\pgfsetstrokeopacity{0.550000}%
\pgfsetdash{}{0pt}%
\pgfpathmoveto{\pgfqpoint{0.188889in}{2.537852in}}%
\pgfpathlineto{\pgfqpoint{4.838889in}{2.537852in}}%
\pgfusepath{stroke}%
\end{pgfscope}%
\begin{pgfscope}%
\pgfpathrectangle{\pgfqpoint{0.188889in}{0.454405in}}{\pgfqpoint{4.650000in}{3.020000in}}%
\pgfusepath{clip}%
\pgfsetrectcap%
\pgfsetroundjoin%
\pgfsetlinewidth{0.401500pt}%
\definecolor{currentstroke}{rgb}{0.800000,0.800000,0.800000}%
\pgfsetstrokecolor{currentstroke}%
\pgfsetstrokeopacity{0.550000}%
\pgfsetdash{}{0pt}%
\pgfpathmoveto{\pgfqpoint{0.188889in}{3.113435in}}%
\pgfpathlineto{\pgfqpoint{4.838889in}{3.113435in}}%
\pgfusepath{stroke}%
\end{pgfscope}%
\begin{pgfscope}%
\pgfpathrectangle{\pgfqpoint{0.188889in}{0.454405in}}{\pgfqpoint{4.650000in}{3.020000in}}%
\pgfusepath{clip}%
\pgfsetrectcap%
\pgfsetroundjoin%
\pgfsetlinewidth{0.401500pt}%
\definecolor{currentstroke}{rgb}{0.800000,0.800000,0.800000}%
\pgfsetstrokecolor{currentstroke}%
\pgfsetstrokeopacity{0.550000}%
\pgfsetdash{}{0pt}%
\pgfpathmoveto{\pgfqpoint{0.188889in}{3.405703in}}%
\pgfpathlineto{\pgfqpoint{4.838889in}{3.405703in}}%
\pgfusepath{stroke}%
\end{pgfscope}%
\begin{pgfscope}%
\pgfpathrectangle{\pgfqpoint{0.188889in}{0.454405in}}{\pgfqpoint{4.650000in}{3.020000in}}%
\pgfusepath{clip}%
\pgfsetbuttcap%
\pgfsetmiterjoin%
\definecolor{currentfill}{rgb}{0.347059,0.458824,0.641176}%
\pgfsetfillcolor{currentfill}%
\pgfsetlinewidth{0.803000pt}%
\definecolor{currentstroke}{rgb}{0.298039,0.298039,0.298039}%
\pgfsetstrokecolor{currentstroke}%
\pgfsetdash{}{0pt}%
\pgfpathmoveto{\pgfqpoint{0.305139in}{1.056293in}}%
\pgfpathlineto{\pgfqpoint{1.235139in}{1.056293in}}%
\pgfpathlineto{\pgfqpoint{1.235139in}{1.076465in}}%
\pgfpathlineto{\pgfqpoint{0.305139in}{1.076465in}}%
\pgfpathlineto{\pgfqpoint{0.305139in}{1.056293in}}%
\pgfpathclose%
\pgfusepath{stroke,fill}%
\end{pgfscope}%
\begin{pgfscope}%
\pgfpathrectangle{\pgfqpoint{0.188889in}{0.454405in}}{\pgfqpoint{4.650000in}{3.020000in}}%
\pgfusepath{clip}%
\pgfsetbuttcap%
\pgfsetmiterjoin%
\definecolor{currentfill}{rgb}{0.798529,0.536765,0.389706}%
\pgfsetfillcolor{currentfill}%
\pgfsetlinewidth{0.803000pt}%
\definecolor{currentstroke}{rgb}{0.298039,0.298039,0.298039}%
\pgfsetstrokecolor{currentstroke}%
\pgfsetdash{}{0pt}%
\pgfpathmoveto{\pgfqpoint{1.467639in}{0.939168in}}%
\pgfpathlineto{\pgfqpoint{2.397639in}{0.939168in}}%
\pgfpathlineto{\pgfqpoint{2.397639in}{1.001542in}}%
\pgfpathlineto{\pgfqpoint{1.467639in}{1.001542in}}%
\pgfpathlineto{\pgfqpoint{1.467639in}{0.939168in}}%
\pgfpathclose%
\pgfusepath{stroke,fill}%
\end{pgfscope}%
\begin{pgfscope}%
\pgfpathrectangle{\pgfqpoint{0.188889in}{0.454405in}}{\pgfqpoint{4.650000in}{3.020000in}}%
\pgfusepath{clip}%
\pgfsetbuttcap%
\pgfsetmiterjoin%
\definecolor{currentfill}{rgb}{0.374020,0.618137,0.429902}%
\pgfsetfillcolor{currentfill}%
\pgfsetlinewidth{0.803000pt}%
\definecolor{currentstroke}{rgb}{0.298039,0.298039,0.298039}%
\pgfsetstrokecolor{currentstroke}%
\pgfsetdash{}{0pt}%
\pgfpathmoveto{\pgfqpoint{2.630139in}{1.019059in}}%
\pgfpathlineto{\pgfqpoint{3.560139in}{1.019059in}}%
\pgfpathlineto{\pgfqpoint{3.560139in}{1.111514in}}%
\pgfpathlineto{\pgfqpoint{2.630139in}{1.111514in}}%
\pgfpathlineto{\pgfqpoint{2.630139in}{1.019059in}}%
\pgfpathclose%
\pgfusepath{stroke,fill}%
\end{pgfscope}%
\begin{pgfscope}%
\pgfpathrectangle{\pgfqpoint{0.188889in}{0.454405in}}{\pgfqpoint{4.650000in}{3.020000in}}%
\pgfusepath{clip}%
\pgfsetbuttcap%
\pgfsetmiterjoin%
\definecolor{currentfill}{rgb}{0.710784,0.363725,0.375490}%
\pgfsetfillcolor{currentfill}%
\pgfsetlinewidth{0.803000pt}%
\definecolor{currentstroke}{rgb}{0.298039,0.298039,0.298039}%
\pgfsetstrokecolor{currentstroke}%
\pgfsetdash{}{0pt}%
\pgfpathmoveto{\pgfqpoint{3.792639in}{1.933553in}}%
\pgfpathlineto{\pgfqpoint{4.722639in}{1.933553in}}%
\pgfpathlineto{\pgfqpoint{4.722639in}{1.979361in}}%
\pgfpathlineto{\pgfqpoint{3.792639in}{1.979361in}}%
\pgfpathlineto{\pgfqpoint{3.792639in}{1.933553in}}%
\pgfpathclose%
\pgfusepath{stroke,fill}%
\end{pgfscope}%
\begin{pgfscope}%
\pgfsetrectcap%
\pgfsetmiterjoin%
\pgfsetlinewidth{1.254687pt}%
\definecolor{currentstroke}{rgb}{0.800000,0.800000,0.800000}%
\pgfsetstrokecolor{currentstroke}%
\pgfsetdash{}{0pt}%
\pgfpathmoveto{\pgfqpoint{0.188889in}{0.454405in}}%
\pgfpathlineto{\pgfqpoint{0.188889in}{3.474405in}}%
\pgfusepath{stroke}%
\end{pgfscope}%
\begin{pgfscope}%
\pgfsetrectcap%
\pgfsetmiterjoin%
\pgfsetlinewidth{1.254687pt}%
\definecolor{currentstroke}{rgb}{0.800000,0.800000,0.800000}%
\pgfsetstrokecolor{currentstroke}%
\pgfsetdash{}{0pt}%
\pgfpathmoveto{\pgfqpoint{4.838889in}{0.454405in}}%
\pgfpathlineto{\pgfqpoint{4.838889in}{3.474405in}}%
\pgfusepath{stroke}%
\end{pgfscope}%
\begin{pgfscope}%
\pgfsetrectcap%
\pgfsetmiterjoin%
\pgfsetlinewidth{1.254687pt}%
\definecolor{currentstroke}{rgb}{0.800000,0.800000,0.800000}%
\pgfsetstrokecolor{currentstroke}%
\pgfsetdash{}{0pt}%
\pgfpathmoveto{\pgfqpoint{0.188889in}{0.454405in}}%
\pgfpathlineto{\pgfqpoint{4.838889in}{0.454405in}}%
\pgfusepath{stroke}%
\end{pgfscope}%
\begin{pgfscope}%
\pgfsetrectcap%
\pgfsetmiterjoin%
\pgfsetlinewidth{1.254687pt}%
\definecolor{currentstroke}{rgb}{0.800000,0.800000,0.800000}%
\pgfsetstrokecolor{currentstroke}%
\pgfsetdash{}{0pt}%
\pgfpathmoveto{\pgfqpoint{0.188889in}{3.474405in}}%
\pgfpathlineto{\pgfqpoint{4.838889in}{3.474405in}}%
\pgfusepath{stroke}%
\end{pgfscope}%
\begin{pgfscope}%
\pgfpathrectangle{\pgfqpoint{0.188889in}{0.454405in}}{\pgfqpoint{4.650000in}{3.020000in}}%
\pgfusepath{clip}%
\pgfsetrectcap%
\pgfsetroundjoin%
\pgfsetlinewidth{0.803000pt}%
\definecolor{currentstroke}{rgb}{0.298039,0.298039,0.298039}%
\pgfsetstrokecolor{currentstroke}%
\pgfsetdash{}{0pt}%
\pgfpathmoveto{\pgfqpoint{0.770139in}{1.056293in}}%
\pgfpathlineto{\pgfqpoint{0.770139in}{1.045905in}}%
\pgfusepath{stroke}%
\end{pgfscope}%
\begin{pgfscope}%
\pgfpathrectangle{\pgfqpoint{0.188889in}{0.454405in}}{\pgfqpoint{4.650000in}{3.020000in}}%
\pgfusepath{clip}%
\pgfsetrectcap%
\pgfsetroundjoin%
\pgfsetlinewidth{0.803000pt}%
\definecolor{currentstroke}{rgb}{0.298039,0.298039,0.298039}%
\pgfsetstrokecolor{currentstroke}%
\pgfsetdash{}{0pt}%
\pgfpathmoveto{\pgfqpoint{0.770139in}{1.076465in}}%
\pgfpathlineto{\pgfqpoint{0.770139in}{1.078472in}}%
\pgfusepath{stroke}%
\end{pgfscope}%
\begin{pgfscope}%
\pgfpathrectangle{\pgfqpoint{0.188889in}{0.454405in}}{\pgfqpoint{4.650000in}{3.020000in}}%
\pgfusepath{clip}%
\pgfsetrectcap%
\pgfsetroundjoin%
\pgfsetlinewidth{0.803000pt}%
\definecolor{currentstroke}{rgb}{0.298039,0.298039,0.298039}%
\pgfsetstrokecolor{currentstroke}%
\pgfsetdash{}{0pt}%
\pgfpathmoveto{\pgfqpoint{0.537639in}{1.045905in}}%
\pgfpathlineto{\pgfqpoint{1.002639in}{1.045905in}}%
\pgfusepath{stroke}%
\end{pgfscope}%
\begin{pgfscope}%
\pgfpathrectangle{\pgfqpoint{0.188889in}{0.454405in}}{\pgfqpoint{4.650000in}{3.020000in}}%
\pgfusepath{clip}%
\pgfsetrectcap%
\pgfsetroundjoin%
\pgfsetlinewidth{0.803000pt}%
\definecolor{currentstroke}{rgb}{0.298039,0.298039,0.298039}%
\pgfsetstrokecolor{currentstroke}%
\pgfsetdash{}{0pt}%
\pgfpathmoveto{\pgfqpoint{0.537639in}{1.078472in}}%
\pgfpathlineto{\pgfqpoint{1.002639in}{1.078472in}}%
\pgfusepath{stroke}%
\end{pgfscope}%
\begin{pgfscope}%
\pgfpathrectangle{\pgfqpoint{0.188889in}{0.454405in}}{\pgfqpoint{4.650000in}{3.020000in}}%
\pgfusepath{clip}%
\pgfsetrectcap%
\pgfsetroundjoin%
\pgfsetlinewidth{0.803000pt}%
\definecolor{currentstroke}{rgb}{0.298039,0.298039,0.298039}%
\pgfsetstrokecolor{currentstroke}%
\pgfsetdash{}{0pt}%
\pgfpathmoveto{\pgfqpoint{0.305139in}{1.076465in}}%
\pgfpathlineto{\pgfqpoint{1.235139in}{1.076465in}}%
\pgfusepath{stroke}%
\end{pgfscope}%
\begin{pgfscope}%
\pgfpathrectangle{\pgfqpoint{0.188889in}{0.454405in}}{\pgfqpoint{4.650000in}{3.020000in}}%
\pgfusepath{clip}%
\pgfsetbuttcap%
\pgfsetmiterjoin%
\definecolor{currentfill}{rgb}{0.298039,0.298039,0.298039}%
\pgfsetfillcolor{currentfill}%
\pgfsetlinewidth{1.003750pt}%
\definecolor{currentstroke}{rgb}{0.298039,0.298039,0.298039}%
\pgfsetstrokecolor{currentstroke}%
\pgfsetdash{}{0pt}%
\pgfsys@defobject{currentmarker}{\pgfqpoint{-0.029463in}{-0.049105in}}{\pgfqpoint{0.029463in}{0.049105in}}{%
\pgfpathmoveto{\pgfqpoint{0.000000in}{-0.049105in}}%
\pgfpathlineto{\pgfqpoint{0.029463in}{0.000000in}}%
\pgfpathlineto{\pgfqpoint{0.000000in}{0.049105in}}%
\pgfpathlineto{\pgfqpoint{-0.029463in}{0.000000in}}%
\pgfpathclose%
\pgfusepath{stroke,fill}%
}%
\begin{pgfscope}%
\pgfsys@transformshift{0.770139in}{1.022052in}%
\pgfsys@useobject{currentmarker}{}%
\end{pgfscope}%
\end{pgfscope}%
\begin{pgfscope}%
\pgfpathrectangle{\pgfqpoint{0.188889in}{0.454405in}}{\pgfqpoint{4.650000in}{3.020000in}}%
\pgfusepath{clip}%
\pgfsetrectcap%
\pgfsetroundjoin%
\pgfsetlinewidth{0.803000pt}%
\definecolor{currentstroke}{rgb}{0.298039,0.298039,0.298039}%
\pgfsetstrokecolor{currentstroke}%
\pgfsetdash{}{0pt}%
\pgfpathmoveto{\pgfqpoint{1.932639in}{0.939168in}}%
\pgfpathlineto{\pgfqpoint{1.932639in}{0.932613in}}%
\pgfusepath{stroke}%
\end{pgfscope}%
\begin{pgfscope}%
\pgfpathrectangle{\pgfqpoint{0.188889in}{0.454405in}}{\pgfqpoint{4.650000in}{3.020000in}}%
\pgfusepath{clip}%
\pgfsetrectcap%
\pgfsetroundjoin%
\pgfsetlinewidth{0.803000pt}%
\definecolor{currentstroke}{rgb}{0.298039,0.298039,0.298039}%
\pgfsetstrokecolor{currentstroke}%
\pgfsetdash{}{0pt}%
\pgfpathmoveto{\pgfqpoint{1.932639in}{1.001542in}}%
\pgfpathlineto{\pgfqpoint{1.932639in}{1.021521in}}%
\pgfusepath{stroke}%
\end{pgfscope}%
\begin{pgfscope}%
\pgfpathrectangle{\pgfqpoint{0.188889in}{0.454405in}}{\pgfqpoint{4.650000in}{3.020000in}}%
\pgfusepath{clip}%
\pgfsetrectcap%
\pgfsetroundjoin%
\pgfsetlinewidth{0.803000pt}%
\definecolor{currentstroke}{rgb}{0.298039,0.298039,0.298039}%
\pgfsetstrokecolor{currentstroke}%
\pgfsetdash{}{0pt}%
\pgfpathmoveto{\pgfqpoint{1.700139in}{0.932613in}}%
\pgfpathlineto{\pgfqpoint{2.165139in}{0.932613in}}%
\pgfusepath{stroke}%
\end{pgfscope}%
\begin{pgfscope}%
\pgfpathrectangle{\pgfqpoint{0.188889in}{0.454405in}}{\pgfqpoint{4.650000in}{3.020000in}}%
\pgfusepath{clip}%
\pgfsetrectcap%
\pgfsetroundjoin%
\pgfsetlinewidth{0.803000pt}%
\definecolor{currentstroke}{rgb}{0.298039,0.298039,0.298039}%
\pgfsetstrokecolor{currentstroke}%
\pgfsetdash{}{0pt}%
\pgfpathmoveto{\pgfqpoint{1.700139in}{1.021521in}}%
\pgfpathlineto{\pgfqpoint{2.165139in}{1.021521in}}%
\pgfusepath{stroke}%
\end{pgfscope}%
\begin{pgfscope}%
\pgfpathrectangle{\pgfqpoint{0.188889in}{0.454405in}}{\pgfqpoint{4.650000in}{3.020000in}}%
\pgfusepath{clip}%
\pgfsetrectcap%
\pgfsetroundjoin%
\pgfsetlinewidth{0.803000pt}%
\definecolor{currentstroke}{rgb}{0.298039,0.298039,0.298039}%
\pgfsetstrokecolor{currentstroke}%
\pgfsetdash{}{0pt}%
\pgfpathmoveto{\pgfqpoint{1.467639in}{0.969609in}}%
\pgfpathlineto{\pgfqpoint{2.397639in}{0.969609in}}%
\pgfusepath{stroke}%
\end{pgfscope}%
\begin{pgfscope}%
\pgfpathrectangle{\pgfqpoint{0.188889in}{0.454405in}}{\pgfqpoint{4.650000in}{3.020000in}}%
\pgfusepath{clip}%
\pgfsetrectcap%
\pgfsetroundjoin%
\pgfsetlinewidth{0.803000pt}%
\definecolor{currentstroke}{rgb}{0.298039,0.298039,0.298039}%
\pgfsetstrokecolor{currentstroke}%
\pgfsetdash{}{0pt}%
\pgfpathmoveto{\pgfqpoint{3.095139in}{1.019059in}}%
\pgfpathlineto{\pgfqpoint{3.095139in}{0.962147in}}%
\pgfusepath{stroke}%
\end{pgfscope}%
\begin{pgfscope}%
\pgfpathrectangle{\pgfqpoint{0.188889in}{0.454405in}}{\pgfqpoint{4.650000in}{3.020000in}}%
\pgfusepath{clip}%
\pgfsetrectcap%
\pgfsetroundjoin%
\pgfsetlinewidth{0.803000pt}%
\definecolor{currentstroke}{rgb}{0.298039,0.298039,0.298039}%
\pgfsetstrokecolor{currentstroke}%
\pgfsetdash{}{0pt}%
\pgfpathmoveto{\pgfqpoint{3.095139in}{1.111514in}}%
\pgfpathlineto{\pgfqpoint{3.095139in}{1.171359in}}%
\pgfusepath{stroke}%
\end{pgfscope}%
\begin{pgfscope}%
\pgfpathrectangle{\pgfqpoint{0.188889in}{0.454405in}}{\pgfqpoint{4.650000in}{3.020000in}}%
\pgfusepath{clip}%
\pgfsetrectcap%
\pgfsetroundjoin%
\pgfsetlinewidth{0.803000pt}%
\definecolor{currentstroke}{rgb}{0.298039,0.298039,0.298039}%
\pgfsetstrokecolor{currentstroke}%
\pgfsetdash{}{0pt}%
\pgfpathmoveto{\pgfqpoint{2.862639in}{0.962147in}}%
\pgfpathlineto{\pgfqpoint{3.327639in}{0.962147in}}%
\pgfusepath{stroke}%
\end{pgfscope}%
\begin{pgfscope}%
\pgfpathrectangle{\pgfqpoint{0.188889in}{0.454405in}}{\pgfqpoint{4.650000in}{3.020000in}}%
\pgfusepath{clip}%
\pgfsetrectcap%
\pgfsetroundjoin%
\pgfsetlinewidth{0.803000pt}%
\definecolor{currentstroke}{rgb}{0.298039,0.298039,0.298039}%
\pgfsetstrokecolor{currentstroke}%
\pgfsetdash{}{0pt}%
\pgfpathmoveto{\pgfqpoint{2.862639in}{1.171359in}}%
\pgfpathlineto{\pgfqpoint{3.327639in}{1.171359in}}%
\pgfusepath{stroke}%
\end{pgfscope}%
\begin{pgfscope}%
\pgfpathrectangle{\pgfqpoint{0.188889in}{0.454405in}}{\pgfqpoint{4.650000in}{3.020000in}}%
\pgfusepath{clip}%
\pgfsetrectcap%
\pgfsetroundjoin%
\pgfsetlinewidth{0.803000pt}%
\definecolor{currentstroke}{rgb}{0.298039,0.298039,0.298039}%
\pgfsetstrokecolor{currentstroke}%
\pgfsetdash{}{0pt}%
\pgfpathmoveto{\pgfqpoint{2.630139in}{1.039441in}}%
\pgfpathlineto{\pgfqpoint{3.560139in}{1.039441in}}%
\pgfusepath{stroke}%
\end{pgfscope}%
\begin{pgfscope}%
\pgfpathrectangle{\pgfqpoint{0.188889in}{0.454405in}}{\pgfqpoint{4.650000in}{3.020000in}}%
\pgfusepath{clip}%
\pgfsetrectcap%
\pgfsetroundjoin%
\pgfsetlinewidth{0.803000pt}%
\definecolor{currentstroke}{rgb}{0.298039,0.298039,0.298039}%
\pgfsetstrokecolor{currentstroke}%
\pgfsetdash{}{0pt}%
\pgfpathmoveto{\pgfqpoint{4.257639in}{1.933553in}}%
\pgfpathlineto{\pgfqpoint{4.257639in}{1.860787in}}%
\pgfusepath{stroke}%
\end{pgfscope}%
\begin{pgfscope}%
\pgfpathrectangle{\pgfqpoint{0.188889in}{0.454405in}}{\pgfqpoint{4.650000in}{3.020000in}}%
\pgfusepath{clip}%
\pgfsetrectcap%
\pgfsetroundjoin%
\pgfsetlinewidth{0.803000pt}%
\definecolor{currentstroke}{rgb}{0.298039,0.298039,0.298039}%
\pgfsetstrokecolor{currentstroke}%
\pgfsetdash{}{0pt}%
\pgfpathmoveto{\pgfqpoint{4.257639in}{1.979361in}}%
\pgfpathlineto{\pgfqpoint{4.257639in}{2.026110in}}%
\pgfusepath{stroke}%
\end{pgfscope}%
\begin{pgfscope}%
\pgfpathrectangle{\pgfqpoint{0.188889in}{0.454405in}}{\pgfqpoint{4.650000in}{3.020000in}}%
\pgfusepath{clip}%
\pgfsetrectcap%
\pgfsetroundjoin%
\pgfsetlinewidth{0.803000pt}%
\definecolor{currentstroke}{rgb}{0.298039,0.298039,0.298039}%
\pgfsetstrokecolor{currentstroke}%
\pgfsetdash{}{0pt}%
\pgfpathmoveto{\pgfqpoint{4.025139in}{1.860787in}}%
\pgfpathlineto{\pgfqpoint{4.490139in}{1.860787in}}%
\pgfusepath{stroke}%
\end{pgfscope}%
\begin{pgfscope}%
\pgfpathrectangle{\pgfqpoint{0.188889in}{0.454405in}}{\pgfqpoint{4.650000in}{3.020000in}}%
\pgfusepath{clip}%
\pgfsetrectcap%
\pgfsetroundjoin%
\pgfsetlinewidth{0.803000pt}%
\definecolor{currentstroke}{rgb}{0.298039,0.298039,0.298039}%
\pgfsetstrokecolor{currentstroke}%
\pgfsetdash{}{0pt}%
\pgfpathmoveto{\pgfqpoint{4.025139in}{2.026110in}}%
\pgfpathlineto{\pgfqpoint{4.490139in}{2.026110in}}%
\pgfusepath{stroke}%
\end{pgfscope}%
\begin{pgfscope}%
\pgfpathrectangle{\pgfqpoint{0.188889in}{0.454405in}}{\pgfqpoint{4.650000in}{3.020000in}}%
\pgfusepath{clip}%
\pgfsetrectcap%
\pgfsetroundjoin%
\pgfsetlinewidth{0.803000pt}%
\definecolor{currentstroke}{rgb}{0.298039,0.298039,0.298039}%
\pgfsetstrokecolor{currentstroke}%
\pgfsetdash{}{0pt}%
\pgfpathmoveto{\pgfqpoint{3.792639in}{1.954010in}}%
\pgfpathlineto{\pgfqpoint{4.722639in}{1.954010in}}%
\pgfusepath{stroke}%
\end{pgfscope}%
\begin{pgfscope}%
\pgfpathrectangle{\pgfqpoint{0.188889in}{0.454405in}}{\pgfqpoint{4.650000in}{3.020000in}}%
\pgfusepath{clip}%
\pgfsetbuttcap%
\pgfsetmiterjoin%
\definecolor{currentfill}{rgb}{0.298039,0.298039,0.298039}%
\pgfsetfillcolor{currentfill}%
\pgfsetlinewidth{1.003750pt}%
\definecolor{currentstroke}{rgb}{0.298039,0.298039,0.298039}%
\pgfsetstrokecolor{currentstroke}%
\pgfsetdash{}{0pt}%
\pgfsys@defobject{currentmarker}{\pgfqpoint{-0.029463in}{-0.049105in}}{\pgfqpoint{0.029463in}{0.049105in}}{%
\pgfpathmoveto{\pgfqpoint{0.000000in}{-0.049105in}}%
\pgfpathlineto{\pgfqpoint{0.029463in}{0.000000in}}%
\pgfpathlineto{\pgfqpoint{0.000000in}{0.049105in}}%
\pgfpathlineto{\pgfqpoint{-0.029463in}{0.000000in}}%
\pgfpathclose%
\pgfusepath{stroke,fill}%
}%
\begin{pgfscope}%
\pgfsys@transformshift{4.257639in}{1.855884in}%
\pgfsys@useobject{currentmarker}{}%
\end{pgfscope}%
\begin{pgfscope}%
\pgfsys@transformshift{4.257639in}{2.043894in}%
\pgfsys@useobject{currentmarker}{}%
\end{pgfscope}%
\end{pgfscope}%
\end{pgfpicture}%
\makeatother%
\endgroup%

%% file: fig/runtime_barplot_res_app.pgf
%% Creator: Matplotlib, PGF backend
%%
%% To include the figure in your LaTeX document, write
%%   \input{<filename>.pgf}
%%
%% Make sure the required packages are loaded in your preamble
%%   \usepackage{pgf}
%%
%% Figures using additional raster images can only be included by \input if
%% they are in the same directory as the main LaTeX file. For loading figures
%% from other directories you can use the `import` package
%% and then include the figures with
%%   \import{<path to file>}{<filename>.pgf}
%%
%% Matplotlib used the following preamble
%%
\begingroup%
\makeatletter%
\begin{pgfpicture}%
\pgfpathrectangle{\pgfpointorigin}{\pgfqpoint{5.512652in}{3.648626in}}%
\pgfusepath{use as bounding box, clip}%
\begin{pgfscope}%
\pgfsetbuttcap%
\pgfsetmiterjoin%
\definecolor{currentfill}{rgb}{1.000000,1.000000,1.000000}%
\pgfsetfillcolor{currentfill}%
\pgfsetlinewidth{0.000000pt}%
\definecolor{currentstroke}{rgb}{1.000000,1.000000,1.000000}%
\pgfsetstrokecolor{currentstroke}%
\pgfsetdash{}{0pt}%
\pgfpathmoveto{\pgfqpoint{0.000000in}{0.000000in}}%
\pgfpathlineto{\pgfqpoint{5.512652in}{0.000000in}}%
\pgfpathlineto{\pgfqpoint{5.512652in}{3.648626in}}%
\pgfpathlineto{\pgfqpoint{0.000000in}{3.648626in}}%
\pgfpathclose%
\pgfusepath{fill}%
\end{pgfscope}%
\begin{pgfscope}%
\pgfsetbuttcap%
\pgfsetmiterjoin%
\definecolor{currentfill}{rgb}{1.000000,1.000000,1.000000}%
\pgfsetfillcolor{currentfill}%
\pgfsetlinewidth{0.000000pt}%
\definecolor{currentstroke}{rgb}{0.000000,0.000000,0.000000}%
\pgfsetstrokecolor{currentstroke}%
\pgfsetstrokeopacity{0.000000}%
\pgfsetdash{}{0pt}%
\pgfpathmoveto{\pgfqpoint{0.762652in}{0.445293in}}%
\pgfpathlineto{\pgfqpoint{5.412652in}{0.445293in}}%
\pgfpathlineto{\pgfqpoint{5.412652in}{3.465293in}}%
\pgfpathlineto{\pgfqpoint{0.762652in}{3.465293in}}%
\pgfpathclose%
\pgfusepath{fill}%
\end{pgfscope}%
\begin{pgfscope}%
\definecolor{textcolor}{rgb}{0.150000,0.150000,0.150000}%
\pgfsetstrokecolor{textcolor}%
\pgfsetfillcolor{textcolor}%
\pgftext[x=1.343902in,y=0.313349in,,top]{\color{textcolor}\rmfamily\fontsize{16.500000}{19.800000}\selectfont \textsc{curly}}%
\end{pgfscope}%
\begin{pgfscope}%
\definecolor{textcolor}{rgb}{0.150000,0.150000,0.150000}%
\pgfsetstrokecolor{textcolor}%
\pgfsetfillcolor{textcolor}%
\pgftext[x=2.506402in,y=0.313349in,,top]{\color{textcolor}\rmfamily\fontsize{16.500000}{19.800000}\selectfont \oldstylenums{2}-\textsc{circles}}%
\end{pgfscope}%
\begin{pgfscope}%
\definecolor{textcolor}{rgb}{0.150000,0.150000,0.150000}%
\pgfsetstrokecolor{textcolor}%
\pgfsetfillcolor{textcolor}%
\pgftext[x=3.668902in,y=0.313349in,,top]{\color{textcolor}\rmfamily\fontsize{16.500000}{19.800000}\selectfont \textsc{circle}~\oldstylenums{4}\textsc{d}}%
\end{pgfscope}%
\begin{pgfscope}%
\definecolor{textcolor}{rgb}{0.150000,0.150000,0.150000}%
\pgfsetstrokecolor{textcolor}%
\pgfsetfillcolor{textcolor}%
\pgftext[x=4.831402in,y=0.313349in,,top]{\color{textcolor}\rmfamily\fontsize{16.500000}{19.800000}\selectfont \textsc{circle}~\oldstylenums{3}\textsc{d}}%
\end{pgfscope}%
\begin{pgfscope}%
\pgfpathrectangle{\pgfqpoint{0.762652in}{0.445293in}}{\pgfqpoint{4.650000in}{3.020000in}}%
\pgfusepath{clip}%
\pgfsetroundcap%
\pgfsetroundjoin%
\pgfsetlinewidth{1.003750pt}%
\definecolor{currentstroke}{rgb}{0.800000,0.800000,0.800000}%
\pgfsetstrokecolor{currentstroke}%
\pgfsetdash{}{0pt}%
\pgfpathmoveto{\pgfqpoint{0.762652in}{0.677601in}}%
\pgfpathlineto{\pgfqpoint{5.412652in}{0.677601in}}%
\pgfusepath{stroke}%
\end{pgfscope}%
\begin{pgfscope}%
\definecolor{textcolor}{rgb}{0.150000,0.150000,0.150000}%
\pgfsetstrokecolor{textcolor}%
\pgfsetfillcolor{textcolor}%
\pgftext[x=0.520639in,y=0.594267in,left,base]{\color{textcolor}\rmfamily\fontsize{16.500000}{19.800000}\selectfont \(\displaystyle 6\)}%
\end{pgfscope}%
\begin{pgfscope}%
\pgfpathrectangle{\pgfqpoint{0.762652in}{0.445293in}}{\pgfqpoint{4.650000in}{3.020000in}}%
\pgfusepath{clip}%
\pgfsetroundcap%
\pgfsetroundjoin%
\pgfsetlinewidth{1.003750pt}%
\definecolor{currentstroke}{rgb}{0.800000,0.800000,0.800000}%
\pgfsetstrokecolor{currentstroke}%
\pgfsetdash{}{0pt}%
\pgfpathmoveto{\pgfqpoint{0.762652in}{1.142216in}}%
\pgfpathlineto{\pgfqpoint{5.412652in}{1.142216in}}%
\pgfusepath{stroke}%
\end{pgfscope}%
\begin{pgfscope}%
\definecolor{textcolor}{rgb}{0.150000,0.150000,0.150000}%
\pgfsetstrokecolor{textcolor}%
\pgfsetfillcolor{textcolor}%
\pgftext[x=0.520639in,y=1.058883in,left,base]{\color{textcolor}\rmfamily\fontsize{16.500000}{19.800000}\selectfont \(\displaystyle 8\)}%
\end{pgfscope}%
\begin{pgfscope}%
\pgfpathrectangle{\pgfqpoint{0.762652in}{0.445293in}}{\pgfqpoint{4.650000in}{3.020000in}}%
\pgfusepath{clip}%
\pgfsetroundcap%
\pgfsetroundjoin%
\pgfsetlinewidth{1.003750pt}%
\definecolor{currentstroke}{rgb}{0.800000,0.800000,0.800000}%
\pgfsetstrokecolor{currentstroke}%
\pgfsetdash{}{0pt}%
\pgfpathmoveto{\pgfqpoint{0.762652in}{1.606832in}}%
\pgfpathlineto{\pgfqpoint{5.412652in}{1.606832in}}%
\pgfusepath{stroke}%
\end{pgfscope}%
\begin{pgfscope}%
\definecolor{textcolor}{rgb}{0.150000,0.150000,0.150000}%
\pgfsetstrokecolor{textcolor}%
\pgfsetfillcolor{textcolor}%
\pgftext[x=0.410571in,y=1.523498in,left,base]{\color{textcolor}\rmfamily\fontsize{16.500000}{19.800000}\selectfont \(\displaystyle 10\)}%
\end{pgfscope}%
\begin{pgfscope}%
\pgfpathrectangle{\pgfqpoint{0.762652in}{0.445293in}}{\pgfqpoint{4.650000in}{3.020000in}}%
\pgfusepath{clip}%
\pgfsetroundcap%
\pgfsetroundjoin%
\pgfsetlinewidth{1.003750pt}%
\definecolor{currentstroke}{rgb}{0.800000,0.800000,0.800000}%
\pgfsetstrokecolor{currentstroke}%
\pgfsetdash{}{0pt}%
\pgfpathmoveto{\pgfqpoint{0.762652in}{2.071447in}}%
\pgfpathlineto{\pgfqpoint{5.412652in}{2.071447in}}%
\pgfusepath{stroke}%
\end{pgfscope}%
\begin{pgfscope}%
\definecolor{textcolor}{rgb}{0.150000,0.150000,0.150000}%
\pgfsetstrokecolor{textcolor}%
\pgfsetfillcolor{textcolor}%
\pgftext[x=0.410571in,y=1.988114in,left,base]{\color{textcolor}\rmfamily\fontsize{16.500000}{19.800000}\selectfont \(\displaystyle 12\)}%
\end{pgfscope}%
\begin{pgfscope}%
\pgfpathrectangle{\pgfqpoint{0.762652in}{0.445293in}}{\pgfqpoint{4.650000in}{3.020000in}}%
\pgfusepath{clip}%
\pgfsetroundcap%
\pgfsetroundjoin%
\pgfsetlinewidth{1.003750pt}%
\definecolor{currentstroke}{rgb}{0.800000,0.800000,0.800000}%
\pgfsetstrokecolor{currentstroke}%
\pgfsetdash{}{0pt}%
\pgfpathmoveto{\pgfqpoint{0.762652in}{2.536062in}}%
\pgfpathlineto{\pgfqpoint{5.412652in}{2.536062in}}%
\pgfusepath{stroke}%
\end{pgfscope}%
\begin{pgfscope}%
\definecolor{textcolor}{rgb}{0.150000,0.150000,0.150000}%
\pgfsetstrokecolor{textcolor}%
\pgfsetfillcolor{textcolor}%
\pgftext[x=0.410571in,y=2.452729in,left,base]{\color{textcolor}\rmfamily\fontsize{16.500000}{19.800000}\selectfont \(\displaystyle 14\)}%
\end{pgfscope}%
\begin{pgfscope}%
\pgfpathrectangle{\pgfqpoint{0.762652in}{0.445293in}}{\pgfqpoint{4.650000in}{3.020000in}}%
\pgfusepath{clip}%
\pgfsetroundcap%
\pgfsetroundjoin%
\pgfsetlinewidth{1.003750pt}%
\definecolor{currentstroke}{rgb}{0.800000,0.800000,0.800000}%
\pgfsetstrokecolor{currentstroke}%
\pgfsetdash{}{0pt}%
\pgfpathmoveto{\pgfqpoint{0.762652in}{3.000678in}}%
\pgfpathlineto{\pgfqpoint{5.412652in}{3.000678in}}%
\pgfusepath{stroke}%
\end{pgfscope}%
\begin{pgfscope}%
\definecolor{textcolor}{rgb}{0.150000,0.150000,0.150000}%
\pgfsetstrokecolor{textcolor}%
\pgfsetfillcolor{textcolor}%
\pgftext[x=0.410571in,y=2.917344in,left,base]{\color{textcolor}\rmfamily\fontsize{16.500000}{19.800000}\selectfont \(\displaystyle 16\)}%
\end{pgfscope}%
\begin{pgfscope}%
\pgfpathrectangle{\pgfqpoint{0.762652in}{0.445293in}}{\pgfqpoint{4.650000in}{3.020000in}}%
\pgfusepath{clip}%
\pgfsetroundcap%
\pgfsetroundjoin%
\pgfsetlinewidth{1.003750pt}%
\definecolor{currentstroke}{rgb}{0.800000,0.800000,0.800000}%
\pgfsetstrokecolor{currentstroke}%
\pgfsetdash{}{0pt}%
\pgfpathmoveto{\pgfqpoint{0.762652in}{3.465293in}}%
\pgfpathlineto{\pgfqpoint{5.412652in}{3.465293in}}%
\pgfusepath{stroke}%
\end{pgfscope}%
\begin{pgfscope}%
\definecolor{textcolor}{rgb}{0.150000,0.150000,0.150000}%
\pgfsetstrokecolor{textcolor}%
\pgfsetfillcolor{textcolor}%
\pgftext[x=0.410571in,y=3.381960in,left,base]{\color{textcolor}\rmfamily\fontsize{16.500000}{19.800000}\selectfont \(\displaystyle 18\)}%
\end{pgfscope}%
\begin{pgfscope}%
\definecolor{textcolor}{rgb}{0.150000,0.150000,0.150000}%
\pgfsetstrokecolor{textcolor}%
\pgfsetfillcolor{textcolor}%
\pgftext[x=0.313349in,y=1.955293in,,bottom,rotate=90.000000]{\color{textcolor}\rmfamily\fontsize{18.000000}{21.600000}\selectfont runtime in seconds}%
\end{pgfscope}%
\begin{pgfscope}%
\pgfpathrectangle{\pgfqpoint{0.762652in}{0.445293in}}{\pgfqpoint{4.650000in}{3.020000in}}%
\pgfusepath{clip}%
\pgfsetbuttcap%
\pgfsetmiterjoin%
\definecolor{currentfill}{rgb}{0.347059,0.458824,0.641176}%
\pgfsetfillcolor{currentfill}%
\pgfsetlinewidth{1.003750pt}%
\definecolor{currentstroke}{rgb}{1.000000,1.000000,1.000000}%
\pgfsetstrokecolor{currentstroke}%
\pgfsetdash{}{0pt}%
\pgfpathmoveto{\pgfqpoint{0.878902in}{-0.716245in}}%
\pgfpathlineto{\pgfqpoint{1.188902in}{-0.716245in}}%
\pgfpathlineto{\pgfqpoint{1.188902in}{1.389660in}}%
\pgfpathlineto{\pgfqpoint{0.878902in}{1.389660in}}%
\pgfpathclose%
\pgfusepath{stroke,fill}%
\end{pgfscope}%
\begin{pgfscope}%
\pgfpathrectangle{\pgfqpoint{0.762652in}{0.445293in}}{\pgfqpoint{4.650000in}{3.020000in}}%
\pgfusepath{clip}%
\pgfsetbuttcap%
\pgfsetmiterjoin%
\definecolor{currentfill}{rgb}{0.347059,0.458824,0.641176}%
\pgfsetfillcolor{currentfill}%
\pgfsetlinewidth{1.003750pt}%
\definecolor{currentstroke}{rgb}{1.000000,1.000000,1.000000}%
\pgfsetstrokecolor{currentstroke}%
\pgfsetdash{}{0pt}%
\pgfpathmoveto{\pgfqpoint{2.041402in}{-0.716245in}}%
\pgfpathlineto{\pgfqpoint{2.351402in}{-0.716245in}}%
\pgfpathlineto{\pgfqpoint{2.351402in}{1.093042in}}%
\pgfpathlineto{\pgfqpoint{2.041402in}{1.093042in}}%
\pgfpathclose%
\pgfusepath{stroke,fill}%
\end{pgfscope}%
\begin{pgfscope}%
\pgfpathrectangle{\pgfqpoint{0.762652in}{0.445293in}}{\pgfqpoint{4.650000in}{3.020000in}}%
\pgfusepath{clip}%
\pgfsetbuttcap%
\pgfsetmiterjoin%
\definecolor{currentfill}{rgb}{0.347059,0.458824,0.641176}%
\pgfsetfillcolor{currentfill}%
\pgfsetlinewidth{1.003750pt}%
\definecolor{currentstroke}{rgb}{1.000000,1.000000,1.000000}%
\pgfsetstrokecolor{currentstroke}%
\pgfsetdash{}{0pt}%
\pgfpathmoveto{\pgfqpoint{3.203902in}{-0.716245in}}%
\pgfpathlineto{\pgfqpoint{3.513902in}{-0.716245in}}%
\pgfpathlineto{\pgfqpoint{3.513902in}{2.063482in}}%
\pgfpathlineto{\pgfqpoint{3.203902in}{2.063482in}}%
\pgfpathclose%
\pgfusepath{stroke,fill}%
\end{pgfscope}%
\begin{pgfscope}%
\pgfpathrectangle{\pgfqpoint{0.762652in}{0.445293in}}{\pgfqpoint{4.650000in}{3.020000in}}%
\pgfusepath{clip}%
\pgfsetbuttcap%
\pgfsetmiterjoin%
\definecolor{currentfill}{rgb}{0.347059,0.458824,0.641176}%
\pgfsetfillcolor{currentfill}%
\pgfsetlinewidth{1.003750pt}%
\definecolor{currentstroke}{rgb}{1.000000,1.000000,1.000000}%
\pgfsetstrokecolor{currentstroke}%
\pgfsetdash{}{0pt}%
\pgfpathmoveto{\pgfqpoint{4.366402in}{-0.716245in}}%
\pgfpathlineto{\pgfqpoint{4.676402in}{-0.716245in}}%
\pgfpathlineto{\pgfqpoint{4.676402in}{1.762738in}}%
\pgfpathlineto{\pgfqpoint{4.366402in}{1.762738in}}%
\pgfpathclose%
\pgfusepath{stroke,fill}%
\end{pgfscope}%
\begin{pgfscope}%
\pgfpathrectangle{\pgfqpoint{0.762652in}{0.445293in}}{\pgfqpoint{4.650000in}{3.020000in}}%
\pgfusepath{clip}%
\pgfsetbuttcap%
\pgfsetmiterjoin%
\definecolor{currentfill}{rgb}{0.798529,0.536765,0.389706}%
\pgfsetfillcolor{currentfill}%
\pgfsetlinewidth{1.003750pt}%
\definecolor{currentstroke}{rgb}{1.000000,1.000000,1.000000}%
\pgfsetstrokecolor{currentstroke}%
\pgfsetdash{}{0pt}%
\pgfpathmoveto{\pgfqpoint{1.188902in}{-0.716245in}}%
\pgfpathlineto{\pgfqpoint{1.498902in}{-0.716245in}}%
\pgfpathlineto{\pgfqpoint{1.498902in}{1.383077in}}%
\pgfpathlineto{\pgfqpoint{1.188902in}{1.383077in}}%
\pgfpathclose%
\pgfusepath{stroke,fill}%
\end{pgfscope}%
\begin{pgfscope}%
\pgfpathrectangle{\pgfqpoint{0.762652in}{0.445293in}}{\pgfqpoint{4.650000in}{3.020000in}}%
\pgfusepath{clip}%
\pgfsetbuttcap%
\pgfsetmiterjoin%
\definecolor{currentfill}{rgb}{0.798529,0.536765,0.389706}%
\pgfsetfillcolor{currentfill}%
\pgfsetlinewidth{1.003750pt}%
\definecolor{currentstroke}{rgb}{1.000000,1.000000,1.000000}%
\pgfsetstrokecolor{currentstroke}%
\pgfsetdash{}{0pt}%
\pgfpathmoveto{\pgfqpoint{2.351402in}{-0.716245in}}%
\pgfpathlineto{\pgfqpoint{2.661402in}{-0.716245in}}%
\pgfpathlineto{\pgfqpoint{2.661402in}{1.105407in}}%
\pgfpathlineto{\pgfqpoint{2.351402in}{1.105407in}}%
\pgfpathclose%
\pgfusepath{stroke,fill}%
\end{pgfscope}%
\begin{pgfscope}%
\pgfpathrectangle{\pgfqpoint{0.762652in}{0.445293in}}{\pgfqpoint{4.650000in}{3.020000in}}%
\pgfusepath{clip}%
\pgfsetbuttcap%
\pgfsetmiterjoin%
\definecolor{currentfill}{rgb}{0.798529,0.536765,0.389706}%
\pgfsetfillcolor{currentfill}%
\pgfsetlinewidth{1.003750pt}%
\definecolor{currentstroke}{rgb}{1.000000,1.000000,1.000000}%
\pgfsetstrokecolor{currentstroke}%
\pgfsetdash{}{0pt}%
\pgfpathmoveto{\pgfqpoint{3.513902in}{-0.716245in}}%
\pgfpathlineto{\pgfqpoint{3.823902in}{-0.716245in}}%
\pgfpathlineto{\pgfqpoint{3.823902in}{2.018887in}}%
\pgfpathlineto{\pgfqpoint{3.513902in}{2.018887in}}%
\pgfpathclose%
\pgfusepath{stroke,fill}%
\end{pgfscope}%
\begin{pgfscope}%
\pgfpathrectangle{\pgfqpoint{0.762652in}{0.445293in}}{\pgfqpoint{4.650000in}{3.020000in}}%
\pgfusepath{clip}%
\pgfsetbuttcap%
\pgfsetmiterjoin%
\definecolor{currentfill}{rgb}{0.798529,0.536765,0.389706}%
\pgfsetfillcolor{currentfill}%
\pgfsetlinewidth{1.003750pt}%
\definecolor{currentstroke}{rgb}{1.000000,1.000000,1.000000}%
\pgfsetstrokecolor{currentstroke}%
\pgfsetdash{}{0pt}%
\pgfpathmoveto{\pgfqpoint{4.676402in}{-0.716245in}}%
\pgfpathlineto{\pgfqpoint{4.986402in}{-0.716245in}}%
\pgfpathlineto{\pgfqpoint{4.986402in}{1.772926in}}%
\pgfpathlineto{\pgfqpoint{4.676402in}{1.772926in}}%
\pgfpathclose%
\pgfusepath{stroke,fill}%
\end{pgfscope}%
\begin{pgfscope}%
\pgfpathrectangle{\pgfqpoint{0.762652in}{0.445293in}}{\pgfqpoint{4.650000in}{3.020000in}}%
\pgfusepath{clip}%
\pgfsetbuttcap%
\pgfsetmiterjoin%
\definecolor{currentfill}{rgb}{0.374020,0.618137,0.429902}%
\pgfsetfillcolor{currentfill}%
\pgfsetlinewidth{1.003750pt}%
\definecolor{currentstroke}{rgb}{1.000000,1.000000,1.000000}%
\pgfsetstrokecolor{currentstroke}%
\pgfsetdash{}{0pt}%
\pgfpathmoveto{\pgfqpoint{1.498902in}{-0.716245in}}%
\pgfpathlineto{\pgfqpoint{1.808902in}{-0.716245in}}%
\pgfpathlineto{\pgfqpoint{1.808902in}{1.679495in}}%
\pgfpathlineto{\pgfqpoint{1.498902in}{1.679495in}}%
\pgfpathclose%
\pgfusepath{stroke,fill}%
\end{pgfscope}%
\begin{pgfscope}%
\pgfpathrectangle{\pgfqpoint{0.762652in}{0.445293in}}{\pgfqpoint{4.650000in}{3.020000in}}%
\pgfusepath{clip}%
\pgfsetbuttcap%
\pgfsetmiterjoin%
\definecolor{currentfill}{rgb}{0.374020,0.618137,0.429902}%
\pgfsetfillcolor{currentfill}%
\pgfsetlinewidth{1.003750pt}%
\definecolor{currentstroke}{rgb}{1.000000,1.000000,1.000000}%
\pgfsetstrokecolor{currentstroke}%
\pgfsetdash{}{0pt}%
\pgfpathmoveto{\pgfqpoint{2.661402in}{-0.716245in}}%
\pgfpathlineto{\pgfqpoint{2.971402in}{-0.716245in}}%
\pgfpathlineto{\pgfqpoint{2.971402in}{1.381650in}}%
\pgfpathlineto{\pgfqpoint{2.661402in}{1.381650in}}%
\pgfpathclose%
\pgfusepath{stroke,fill}%
\end{pgfscope}%
\begin{pgfscope}%
\pgfpathrectangle{\pgfqpoint{0.762652in}{0.445293in}}{\pgfqpoint{4.650000in}{3.020000in}}%
\pgfusepath{clip}%
\pgfsetbuttcap%
\pgfsetmiterjoin%
\definecolor{currentfill}{rgb}{0.374020,0.618137,0.429902}%
\pgfsetfillcolor{currentfill}%
\pgfsetlinewidth{1.003750pt}%
\definecolor{currentstroke}{rgb}{1.000000,1.000000,1.000000}%
\pgfsetstrokecolor{currentstroke}%
\pgfsetdash{}{0pt}%
\pgfpathmoveto{\pgfqpoint{3.823902in}{-0.716245in}}%
\pgfpathlineto{\pgfqpoint{4.133902in}{-0.716245in}}%
\pgfpathlineto{\pgfqpoint{4.133902in}{0.794768in}}%
\pgfpathlineto{\pgfqpoint{3.823902in}{0.794768in}}%
\pgfpathclose%
\pgfusepath{stroke,fill}%
\end{pgfscope}%
\begin{pgfscope}%
\pgfpathrectangle{\pgfqpoint{0.762652in}{0.445293in}}{\pgfqpoint{4.650000in}{3.020000in}}%
\pgfusepath{clip}%
\pgfsetbuttcap%
\pgfsetmiterjoin%
\definecolor{currentfill}{rgb}{0.374020,0.618137,0.429902}%
\pgfsetfillcolor{currentfill}%
\pgfsetlinewidth{1.003750pt}%
\definecolor{currentstroke}{rgb}{1.000000,1.000000,1.000000}%
\pgfsetstrokecolor{currentstroke}%
\pgfsetdash{}{0pt}%
\pgfpathmoveto{\pgfqpoint{4.986402in}{-0.716245in}}%
\pgfpathlineto{\pgfqpoint{5.296402in}{-0.716245in}}%
\pgfpathlineto{\pgfqpoint{5.296402in}{0.990090in}}%
\pgfpathlineto{\pgfqpoint{4.986402in}{0.990090in}}%
\pgfpathclose%
\pgfusepath{stroke,fill}%
\end{pgfscope}%
\begin{pgfscope}%
\pgfpathrectangle{\pgfqpoint{0.762652in}{0.445293in}}{\pgfqpoint{4.650000in}{3.020000in}}%
\pgfusepath{clip}%
\pgfsetroundcap%
\pgfsetroundjoin%
\pgfsetlinewidth{2.710125pt}%
\definecolor{currentstroke}{rgb}{0.260000,0.260000,0.260000}%
\pgfsetstrokecolor{currentstroke}%
\pgfsetdash{}{0pt}%
\pgfpathmoveto{\pgfqpoint{1.033902in}{1.384447in}}%
\pgfpathlineto{\pgfqpoint{1.033902in}{1.395478in}}%
\pgfusepath{stroke}%
\end{pgfscope}%
\begin{pgfscope}%
\pgfpathrectangle{\pgfqpoint{0.762652in}{0.445293in}}{\pgfqpoint{4.650000in}{3.020000in}}%
\pgfusepath{clip}%
\pgfsetroundcap%
\pgfsetroundjoin%
\pgfsetlinewidth{2.710125pt}%
\definecolor{currentstroke}{rgb}{0.260000,0.260000,0.260000}%
\pgfsetstrokecolor{currentstroke}%
\pgfsetdash{}{0pt}%
\pgfpathmoveto{\pgfqpoint{2.196402in}{1.089947in}}%
\pgfpathlineto{\pgfqpoint{2.196402in}{1.096444in}}%
\pgfusepath{stroke}%
\end{pgfscope}%
\begin{pgfscope}%
\pgfpathrectangle{\pgfqpoint{0.762652in}{0.445293in}}{\pgfqpoint{4.650000in}{3.020000in}}%
\pgfusepath{clip}%
\pgfsetroundcap%
\pgfsetroundjoin%
\pgfsetlinewidth{2.710125pt}%
\definecolor{currentstroke}{rgb}{0.260000,0.260000,0.260000}%
\pgfsetstrokecolor{currentstroke}%
\pgfsetdash{}{0pt}%
\pgfpathmoveto{\pgfqpoint{3.358902in}{2.048365in}}%
\pgfpathlineto{\pgfqpoint{3.358902in}{2.078348in}}%
\pgfusepath{stroke}%
\end{pgfscope}%
\begin{pgfscope}%
\pgfpathrectangle{\pgfqpoint{0.762652in}{0.445293in}}{\pgfqpoint{4.650000in}{3.020000in}}%
\pgfusepath{clip}%
\pgfsetroundcap%
\pgfsetroundjoin%
\pgfsetlinewidth{2.710125pt}%
\definecolor{currentstroke}{rgb}{0.260000,0.260000,0.260000}%
\pgfsetstrokecolor{currentstroke}%
\pgfsetdash{}{0pt}%
\pgfpathmoveto{\pgfqpoint{4.521402in}{1.753033in}}%
\pgfpathlineto{\pgfqpoint{4.521402in}{1.773840in}}%
\pgfusepath{stroke}%
\end{pgfscope}%
\begin{pgfscope}%
\pgfpathrectangle{\pgfqpoint{0.762652in}{0.445293in}}{\pgfqpoint{4.650000in}{3.020000in}}%
\pgfusepath{clip}%
\pgfsetroundcap%
\pgfsetroundjoin%
\pgfsetlinewidth{2.710125pt}%
\definecolor{currentstroke}{rgb}{0.260000,0.260000,0.260000}%
\pgfsetstrokecolor{currentstroke}%
\pgfsetdash{}{0pt}%
\pgfpathmoveto{\pgfqpoint{1.343902in}{1.378300in}}%
\pgfpathlineto{\pgfqpoint{1.343902in}{1.388159in}}%
\pgfusepath{stroke}%
\end{pgfscope}%
\begin{pgfscope}%
\pgfpathrectangle{\pgfqpoint{0.762652in}{0.445293in}}{\pgfqpoint{4.650000in}{3.020000in}}%
\pgfusepath{clip}%
\pgfsetroundcap%
\pgfsetroundjoin%
\pgfsetlinewidth{2.710125pt}%
\definecolor{currentstroke}{rgb}{0.260000,0.260000,0.260000}%
\pgfsetstrokecolor{currentstroke}%
\pgfsetdash{}{0pt}%
\pgfpathmoveto{\pgfqpoint{2.506402in}{1.102658in}}%
\pgfpathlineto{\pgfqpoint{2.506402in}{1.108285in}}%
\pgfusepath{stroke}%
\end{pgfscope}%
\begin{pgfscope}%
\pgfpathrectangle{\pgfqpoint{0.762652in}{0.445293in}}{\pgfqpoint{4.650000in}{3.020000in}}%
\pgfusepath{clip}%
\pgfsetroundcap%
\pgfsetroundjoin%
\pgfsetlinewidth{2.710125pt}%
\definecolor{currentstroke}{rgb}{0.260000,0.260000,0.260000}%
\pgfsetstrokecolor{currentstroke}%
\pgfsetdash{}{0pt}%
\pgfpathmoveto{\pgfqpoint{3.668902in}{2.012055in}}%
\pgfpathlineto{\pgfqpoint{3.668902in}{2.026414in}}%
\pgfusepath{stroke}%
\end{pgfscope}%
\begin{pgfscope}%
\pgfpathrectangle{\pgfqpoint{0.762652in}{0.445293in}}{\pgfqpoint{4.650000in}{3.020000in}}%
\pgfusepath{clip}%
\pgfsetroundcap%
\pgfsetroundjoin%
\pgfsetlinewidth{2.710125pt}%
\definecolor{currentstroke}{rgb}{0.260000,0.260000,0.260000}%
\pgfsetstrokecolor{currentstroke}%
\pgfsetdash{}{0pt}%
\pgfpathmoveto{\pgfqpoint{4.831402in}{1.762280in}}%
\pgfpathlineto{\pgfqpoint{4.831402in}{1.784146in}}%
\pgfusepath{stroke}%
\end{pgfscope}%
\begin{pgfscope}%
\pgfpathrectangle{\pgfqpoint{0.762652in}{0.445293in}}{\pgfqpoint{4.650000in}{3.020000in}}%
\pgfusepath{clip}%
\pgfsetroundcap%
\pgfsetroundjoin%
\pgfsetlinewidth{2.710125pt}%
\definecolor{currentstroke}{rgb}{0.260000,0.260000,0.260000}%
\pgfsetstrokecolor{currentstroke}%
\pgfsetdash{}{0pt}%
\pgfpathmoveto{\pgfqpoint{1.653902in}{1.661125in}}%
\pgfpathlineto{\pgfqpoint{1.653902in}{1.699550in}}%
\pgfusepath{stroke}%
\end{pgfscope}%
\begin{pgfscope}%
\pgfpathrectangle{\pgfqpoint{0.762652in}{0.445293in}}{\pgfqpoint{4.650000in}{3.020000in}}%
\pgfusepath{clip}%
\pgfsetroundcap%
\pgfsetroundjoin%
\pgfsetlinewidth{2.710125pt}%
\definecolor{currentstroke}{rgb}{0.260000,0.260000,0.260000}%
\pgfsetstrokecolor{currentstroke}%
\pgfsetdash{}{0pt}%
\pgfpathmoveto{\pgfqpoint{2.816402in}{1.372691in}}%
\pgfpathlineto{\pgfqpoint{2.816402in}{1.391282in}}%
\pgfusepath{stroke}%
\end{pgfscope}%
\begin{pgfscope}%
\pgfpathrectangle{\pgfqpoint{0.762652in}{0.445293in}}{\pgfqpoint{4.650000in}{3.020000in}}%
\pgfusepath{clip}%
\pgfsetroundcap%
\pgfsetroundjoin%
\pgfsetlinewidth{2.710125pt}%
\definecolor{currentstroke}{rgb}{0.260000,0.260000,0.260000}%
\pgfsetstrokecolor{currentstroke}%
\pgfsetdash{}{0pt}%
\pgfpathmoveto{\pgfqpoint{3.978902in}{0.785351in}}%
\pgfpathlineto{\pgfqpoint{3.978902in}{0.803299in}}%
\pgfusepath{stroke}%
\end{pgfscope}%
\begin{pgfscope}%
\pgfpathrectangle{\pgfqpoint{0.762652in}{0.445293in}}{\pgfqpoint{4.650000in}{3.020000in}}%
\pgfusepath{clip}%
\pgfsetroundcap%
\pgfsetroundjoin%
\pgfsetlinewidth{2.710125pt}%
\definecolor{currentstroke}{rgb}{0.260000,0.260000,0.260000}%
\pgfsetstrokecolor{currentstroke}%
\pgfsetdash{}{0pt}%
\pgfpathmoveto{\pgfqpoint{5.141402in}{0.976458in}}%
\pgfpathlineto{\pgfqpoint{5.141402in}{1.006286in}}%
\pgfusepath{stroke}%
\end{pgfscope}%
\begin{pgfscope}%
\pgfsetrectcap%
\pgfsetmiterjoin%
\pgfsetlinewidth{1.254687pt}%
\definecolor{currentstroke}{rgb}{0.800000,0.800000,0.800000}%
\pgfsetstrokecolor{currentstroke}%
\pgfsetdash{}{0pt}%
\pgfpathmoveto{\pgfqpoint{0.762652in}{0.445293in}}%
\pgfpathlineto{\pgfqpoint{0.762652in}{3.465293in}}%
\pgfusepath{stroke}%
\end{pgfscope}%
\begin{pgfscope}%
\pgfsetrectcap%
\pgfsetmiterjoin%
\pgfsetlinewidth{1.254687pt}%
\definecolor{currentstroke}{rgb}{0.800000,0.800000,0.800000}%
\pgfsetstrokecolor{currentstroke}%
\pgfsetdash{}{0pt}%
\pgfpathmoveto{\pgfqpoint{5.412652in}{0.445293in}}%
\pgfpathlineto{\pgfqpoint{5.412652in}{3.465293in}}%
\pgfusepath{stroke}%
\end{pgfscope}%
\begin{pgfscope}%
\pgfsetrectcap%
\pgfsetmiterjoin%
\pgfsetlinewidth{1.254687pt}%
\definecolor{currentstroke}{rgb}{0.800000,0.800000,0.800000}%
\pgfsetstrokecolor{currentstroke}%
\pgfsetdash{}{0pt}%
\pgfpathmoveto{\pgfqpoint{0.762652in}{0.445293in}}%
\pgfpathlineto{\pgfqpoint{5.412652in}{0.445293in}}%
\pgfusepath{stroke}%
\end{pgfscope}%
\begin{pgfscope}%
\pgfsetrectcap%
\pgfsetmiterjoin%
\pgfsetlinewidth{1.254687pt}%
\definecolor{currentstroke}{rgb}{0.800000,0.800000,0.800000}%
\pgfsetstrokecolor{currentstroke}%
\pgfsetdash{}{0pt}%
\pgfpathmoveto{\pgfqpoint{0.762652in}{3.465293in}}%
\pgfpathlineto{\pgfqpoint{5.412652in}{3.465293in}}%
\pgfusepath{stroke}%
\end{pgfscope}%
\begin{pgfscope}%
\pgfsetbuttcap%
\pgfsetmiterjoin%
\definecolor{currentfill}{rgb}{1.000000,1.000000,1.000000}%
\pgfsetfillcolor{currentfill}%
\pgfsetfillopacity{0.800000}%
\pgfsetlinewidth{1.003750pt}%
\definecolor{currentstroke}{rgb}{0.800000,0.800000,0.800000}%
\pgfsetstrokecolor{currentstroke}%
\pgfsetstrokeopacity{0.800000}%
\pgfsetdash{}{0pt}%
\pgfpathmoveto{\pgfqpoint{1.588342in}{2.954028in}}%
\pgfpathlineto{\pgfqpoint{5.252235in}{2.954028in}}%
\pgfpathquadraticcurveto{\pgfqpoint{5.298068in}{2.954028in}}{\pgfqpoint{5.298068in}{2.999861in}}%
\pgfpathlineto{\pgfqpoint{5.298068in}{3.304876in}}%
\pgfpathquadraticcurveto{\pgfqpoint{5.298068in}{3.350710in}}{\pgfqpoint{5.252235in}{3.350710in}}%
\pgfpathlineto{\pgfqpoint{1.588342in}{3.350710in}}%
\pgfpathquadraticcurveto{\pgfqpoint{1.542508in}{3.350710in}}{\pgfqpoint{1.542508in}{3.304876in}}%
\pgfpathlineto{\pgfqpoint{1.542508in}{2.999861in}}%
\pgfpathquadraticcurveto{\pgfqpoint{1.542508in}{2.954028in}}{\pgfqpoint{1.588342in}{2.954028in}}%
\pgfpathclose%
\pgfusepath{stroke,fill}%
\end{pgfscope}%
\begin{pgfscope}%
\pgfsetbuttcap%
\pgfsetmiterjoin%
\definecolor{currentfill}{rgb}{0.347059,0.458824,0.641176}%
\pgfsetfillcolor{currentfill}%
\pgfsetlinewidth{1.003750pt}%
\definecolor{currentstroke}{rgb}{1.000000,1.000000,1.000000}%
\pgfsetstrokecolor{currentstroke}%
\pgfsetdash{}{0pt}%
\pgfpathmoveto{\pgfqpoint{1.634175in}{3.092376in}}%
\pgfpathlineto{\pgfqpoint{2.092508in}{3.092376in}}%
\pgfpathlineto{\pgfqpoint{2.092508in}{3.252793in}}%
\pgfpathlineto{\pgfqpoint{1.634175in}{3.252793in}}%
\pgfpathclose%
\pgfusepath{stroke,fill}%
\end{pgfscope}%
\begin{pgfscope}%
\definecolor{textcolor}{rgb}{0.150000,0.150000,0.150000}%
\pgfsetstrokecolor{textcolor}%
\pgfsetfillcolor{textcolor}%
\pgftext[x=2.275842in,y=3.092376in,left,base]{\color{textcolor}\rmfamily\fontsize{16.500000}{19.800000}\selectfont \textsc{w-l}}%
\end{pgfscope}%
\begin{pgfscope}%
\pgfsetbuttcap%
\pgfsetmiterjoin%
\definecolor{currentfill}{rgb}{0.798529,0.536765,0.389706}%
\pgfsetfillcolor{currentfill}%
\pgfsetlinewidth{1.003750pt}%
\definecolor{currentstroke}{rgb}{1.000000,1.000000,1.000000}%
\pgfsetstrokecolor{currentstroke}%
\pgfsetdash{}{0pt}%
\pgfpathmoveto{\pgfqpoint{2.850656in}{3.092376in}}%
\pgfpathlineto{\pgfqpoint{3.308989in}{3.092376in}}%
\pgfpathlineto{\pgfqpoint{3.308989in}{3.252793in}}%
\pgfpathlineto{\pgfqpoint{2.850656in}{3.252793in}}%
\pgfpathclose%
\pgfusepath{stroke,fill}%
\end{pgfscope}%
\begin{pgfscope}%
\definecolor{textcolor}{rgb}{0.150000,0.150000,0.150000}%
\pgfsetstrokecolor{textcolor}%
\pgfsetfillcolor{textcolor}%
\pgftext[x=3.492323in,y=3.092376in,left,base]{\color{textcolor}\rmfamily\fontsize{16.500000}{19.800000}\selectfont \textsc{w-m}}%
\end{pgfscope}%
\begin{pgfscope}%
\pgfsetbuttcap%
\pgfsetmiterjoin%
\definecolor{currentfill}{rgb}{0.374020,0.618137,0.429902}%
\pgfsetfillcolor{currentfill}%
\pgfsetlinewidth{1.003750pt}%
\definecolor{currentstroke}{rgb}{1.000000,1.000000,1.000000}%
\pgfsetstrokecolor{currentstroke}%
\pgfsetdash{}{0pt}%
\pgfpathmoveto{\pgfqpoint{4.123137in}{3.092376in}}%
\pgfpathlineto{\pgfqpoint{4.581470in}{3.092376in}}%
\pgfpathlineto{\pgfqpoint{4.581470in}{3.252793in}}%
\pgfpathlineto{\pgfqpoint{4.123137in}{3.252793in}}%
\pgfpathclose%
\pgfusepath{stroke,fill}%
\end{pgfscope}%
\begin{pgfscope}%
\definecolor{textcolor}{rgb}{0.150000,0.150000,0.150000}%
\pgfsetstrokecolor{textcolor}%
\pgfsetfillcolor{textcolor}%
\pgftext[x=4.764803in,y=3.092376in,left,base]{\color{textcolor}\rmfamily\fontsize{16.500000}{19.800000}\selectfont \textsc{dcv}}%
\end{pgfscope}%
\end{pgfpicture}%
\makeatother%
\endgroup%

%% file: fig/continuous_circle-1000-3d_res.pgf
%% Creator: Matplotlib, PGF backend
%%
%% To include the figure in your LaTeX document, write
%%   \input{<filename>.pgf}
%%
%% Make sure the required packages are loaded in your preamble
%%   \usepackage{pgf}
%%
%% Figures using additional raster images can only be included by \input if
%% they are in the same directory as the main LaTeX file. For loading figures
%% from other directories you can use the `import` package
%% and then include the figures with
%%   \import{<path to file>}{<filename>.pgf}
%%
%% Matplotlib used the following preamble
%%
\begingroup%
\makeatletter%
\begin{pgfpicture}%
\pgfpathrectangle{\pgfpointorigin}{\pgfqpoint{5.979207in}{4.035754in}}%
\pgfusepath{use as bounding box, clip}%
\begin{pgfscope}%
\pgfsetbuttcap%
\pgfsetmiterjoin%
\definecolor{currentfill}{rgb}{1.000000,1.000000,1.000000}%
\pgfsetfillcolor{currentfill}%
\pgfsetlinewidth{0.000000pt}%
\definecolor{currentstroke}{rgb}{1.000000,1.000000,1.000000}%
\pgfsetstrokecolor{currentstroke}%
\pgfsetdash{}{0pt}%
\pgfpathmoveto{\pgfqpoint{0.000000in}{0.000000in}}%
\pgfpathlineto{\pgfqpoint{5.979207in}{0.000000in}}%
\pgfpathlineto{\pgfqpoint{5.979207in}{4.035754in}}%
\pgfpathlineto{\pgfqpoint{0.000000in}{4.035754in}}%
\pgfpathclose%
\pgfusepath{fill}%
\end{pgfscope}%
\begin{pgfscope}%
\pgfsetbuttcap%
\pgfsetmiterjoin%
\definecolor{currentfill}{rgb}{1.000000,1.000000,1.000000}%
\pgfsetfillcolor{currentfill}%
\pgfsetlinewidth{0.000000pt}%
\definecolor{currentstroke}{rgb}{0.000000,0.000000,0.000000}%
\pgfsetstrokecolor{currentstroke}%
\pgfsetstrokeopacity{0.000000}%
\pgfsetdash{}{0pt}%
\pgfpathmoveto{\pgfqpoint{1.229207in}{0.915754in}}%
\pgfpathlineto{\pgfqpoint{5.879208in}{0.915754in}}%
\pgfpathlineto{\pgfqpoint{5.879208in}{3.935754in}}%
\pgfpathlineto{\pgfqpoint{1.229207in}{3.935754in}}%
\pgfpathclose%
\pgfusepath{fill}%
\end{pgfscope}%
\begin{pgfscope}%
\pgfpathrectangle{\pgfqpoint{1.229207in}{0.915754in}}{\pgfqpoint{4.650000in}{3.020000in}}%
\pgfusepath{clip}%
\pgfsetroundcap%
\pgfsetroundjoin%
\pgfsetlinewidth{1.003750pt}%
\definecolor{currentstroke}{rgb}{0.800000,0.800000,0.800000}%
\pgfsetstrokecolor{currentstroke}%
\pgfsetdash{}{0pt}%
\pgfpathmoveto{\pgfqpoint{1.478315in}{0.915754in}}%
\pgfpathlineto{\pgfqpoint{1.478315in}{3.935754in}}%
\pgfusepath{stroke}%
\end{pgfscope}%
\begin{pgfscope}%
\definecolor{textcolor}{rgb}{0.150000,0.150000,0.150000}%
\pgfsetstrokecolor{textcolor}%
\pgfsetfillcolor{textcolor}%
\pgftext[x=1.478315in,y=0.783810in,,top]{\color{textcolor}\rmfamily\fontsize{23.100000}{27.720000}\selectfont 10}%
\end{pgfscope}%
\begin{pgfscope}%
\pgfpathrectangle{\pgfqpoint{1.229207in}{0.915754in}}{\pgfqpoint{4.650000in}{3.020000in}}%
\pgfusepath{clip}%
\pgfsetroundcap%
\pgfsetroundjoin%
\pgfsetlinewidth{1.003750pt}%
\definecolor{currentstroke}{rgb}{0.800000,0.800000,0.800000}%
\pgfsetstrokecolor{currentstroke}%
\pgfsetdash{}{0pt}%
\pgfpathmoveto{\pgfqpoint{2.308672in}{0.915754in}}%
\pgfpathlineto{\pgfqpoint{2.308672in}{3.935754in}}%
\pgfusepath{stroke}%
\end{pgfscope}%
\begin{pgfscope}%
\definecolor{textcolor}{rgb}{0.150000,0.150000,0.150000}%
\pgfsetstrokecolor{textcolor}%
\pgfsetfillcolor{textcolor}%
\pgftext[x=2.308672in,y=0.783810in,,top]{\color{textcolor}\rmfamily\fontsize{23.100000}{27.720000}\selectfont 20}%
\end{pgfscope}%
\begin{pgfscope}%
\pgfpathrectangle{\pgfqpoint{1.229207in}{0.915754in}}{\pgfqpoint{4.650000in}{3.020000in}}%
\pgfusepath{clip}%
\pgfsetroundcap%
\pgfsetroundjoin%
\pgfsetlinewidth{1.003750pt}%
\definecolor{currentstroke}{rgb}{0.800000,0.800000,0.800000}%
\pgfsetstrokecolor{currentstroke}%
\pgfsetdash{}{0pt}%
\pgfpathmoveto{\pgfqpoint{3.139029in}{0.915754in}}%
\pgfpathlineto{\pgfqpoint{3.139029in}{3.935754in}}%
\pgfusepath{stroke}%
\end{pgfscope}%
\begin{pgfscope}%
\definecolor{textcolor}{rgb}{0.150000,0.150000,0.150000}%
\pgfsetstrokecolor{textcolor}%
\pgfsetfillcolor{textcolor}%
\pgftext[x=3.139029in,y=0.783810in,,top]{\color{textcolor}\rmfamily\fontsize{23.100000}{27.720000}\selectfont 30}%
\end{pgfscope}%
\begin{pgfscope}%
\pgfpathrectangle{\pgfqpoint{1.229207in}{0.915754in}}{\pgfqpoint{4.650000in}{3.020000in}}%
\pgfusepath{clip}%
\pgfsetroundcap%
\pgfsetroundjoin%
\pgfsetlinewidth{1.003750pt}%
\definecolor{currentstroke}{rgb}{0.800000,0.800000,0.800000}%
\pgfsetstrokecolor{currentstroke}%
\pgfsetdash{}{0pt}%
\pgfpathmoveto{\pgfqpoint{3.969386in}{0.915754in}}%
\pgfpathlineto{\pgfqpoint{3.969386in}{3.935754in}}%
\pgfusepath{stroke}%
\end{pgfscope}%
\begin{pgfscope}%
\definecolor{textcolor}{rgb}{0.150000,0.150000,0.150000}%
\pgfsetstrokecolor{textcolor}%
\pgfsetfillcolor{textcolor}%
\pgftext[x=3.969386in,y=0.783810in,,top]{\color{textcolor}\rmfamily\fontsize{23.100000}{27.720000}\selectfont 40}%
\end{pgfscope}%
\begin{pgfscope}%
\pgfpathrectangle{\pgfqpoint{1.229207in}{0.915754in}}{\pgfqpoint{4.650000in}{3.020000in}}%
\pgfusepath{clip}%
\pgfsetroundcap%
\pgfsetroundjoin%
\pgfsetlinewidth{1.003750pt}%
\definecolor{currentstroke}{rgb}{0.800000,0.800000,0.800000}%
\pgfsetstrokecolor{currentstroke}%
\pgfsetdash{}{0pt}%
\pgfpathmoveto{\pgfqpoint{4.799743in}{0.915754in}}%
\pgfpathlineto{\pgfqpoint{4.799743in}{3.935754in}}%
\pgfusepath{stroke}%
\end{pgfscope}%
\begin{pgfscope}%
\definecolor{textcolor}{rgb}{0.150000,0.150000,0.150000}%
\pgfsetstrokecolor{textcolor}%
\pgfsetfillcolor{textcolor}%
\pgftext[x=4.799743in,y=0.783810in,,top]{\color{textcolor}\rmfamily\fontsize{23.100000}{27.720000}\selectfont 50}%
\end{pgfscope}%
\begin{pgfscope}%
\pgfpathrectangle{\pgfqpoint{1.229207in}{0.915754in}}{\pgfqpoint{4.650000in}{3.020000in}}%
\pgfusepath{clip}%
\pgfsetroundcap%
\pgfsetroundjoin%
\pgfsetlinewidth{1.003750pt}%
\definecolor{currentstroke}{rgb}{0.800000,0.800000,0.800000}%
\pgfsetstrokecolor{currentstroke}%
\pgfsetdash{}{0pt}%
\pgfpathmoveto{\pgfqpoint{5.630100in}{0.915754in}}%
\pgfpathlineto{\pgfqpoint{5.630100in}{3.935754in}}%
\pgfusepath{stroke}%
\end{pgfscope}%
\begin{pgfscope}%
\definecolor{textcolor}{rgb}{0.150000,0.150000,0.150000}%
\pgfsetstrokecolor{textcolor}%
\pgfsetfillcolor{textcolor}%
\pgftext[x=5.630100in,y=0.783810in,,top]{\color{textcolor}\rmfamily\fontsize{23.100000}{27.720000}\selectfont 60}%
\end{pgfscope}%
\begin{pgfscope}%
\definecolor{textcolor}{rgb}{0.150000,0.150000,0.150000}%
\pgfsetstrokecolor{textcolor}%
\pgfsetfillcolor{textcolor}%
\pgftext[x=3.554208in,y=0.407183in,,top]{\color{textcolor}\rmfamily\fontsize{25.200000}{30.240000}\selectfont runtime in seconds}%
\end{pgfscope}%
\begin{pgfscope}%
\pgfpathrectangle{\pgfqpoint{1.229207in}{0.915754in}}{\pgfqpoint{4.650000in}{3.020000in}}%
\pgfusepath{clip}%
\pgfsetroundcap%
\pgfsetroundjoin%
\pgfsetlinewidth{1.003750pt}%
\definecolor{currentstroke}{rgb}{0.800000,0.800000,0.800000}%
\pgfsetstrokecolor{currentstroke}%
\pgfsetdash{}{0pt}%
\pgfpathmoveto{\pgfqpoint{1.229207in}{1.328214in}}%
\pgfpathlineto{\pgfqpoint{5.879208in}{1.328214in}}%
\pgfusepath{stroke}%
\end{pgfscope}%
\begin{pgfscope}%
\definecolor{textcolor}{rgb}{0.150000,0.150000,0.150000}%
\pgfsetstrokecolor{textcolor}%
\pgfsetfillcolor{textcolor}%
\pgftext[x=0.476627in,y=1.208229in,left,base]{\color{textcolor}\rmfamily\fontsize{23.100000}{27.720000}\selectfont \(\displaystyle {10^{-2}}\)}%
\end{pgfscope}%
\begin{pgfscope}%
\pgfpathrectangle{\pgfqpoint{1.229207in}{0.915754in}}{\pgfqpoint{4.650000in}{3.020000in}}%
\pgfusepath{clip}%
\pgfsetroundcap%
\pgfsetroundjoin%
\pgfsetlinewidth{1.003750pt}%
\definecolor{currentstroke}{rgb}{0.800000,0.800000,0.800000}%
\pgfsetstrokecolor{currentstroke}%
\pgfsetdash{}{0pt}%
\pgfpathmoveto{\pgfqpoint{1.229207in}{2.698375in}}%
\pgfpathlineto{\pgfqpoint{5.879208in}{2.698375in}}%
\pgfusepath{stroke}%
\end{pgfscope}%
\begin{pgfscope}%
\definecolor{textcolor}{rgb}{0.150000,0.150000,0.150000}%
\pgfsetstrokecolor{textcolor}%
\pgfsetfillcolor{textcolor}%
\pgftext[x=0.476627in,y=2.578391in,left,base]{\color{textcolor}\rmfamily\fontsize{23.100000}{27.720000}\selectfont \(\displaystyle {10^{-1}}\)}%
\end{pgfscope}%
\begin{pgfscope}%
\definecolor{textcolor}{rgb}{0.150000,0.150000,0.150000}%
\pgfsetstrokecolor{textcolor}%
\pgfsetfillcolor{textcolor}%
\pgftext[x=0.407183in,y=2.425754in,,bottom,rotate=90.000000]{\color{textcolor}\rmfamily\fontsize{25.200000}{30.240000}\selectfont relative error}%
\end{pgfscope}%
\begin{pgfscope}%
\pgfpathrectangle{\pgfqpoint{1.229207in}{0.915754in}}{\pgfqpoint{4.650000in}{3.020000in}}%
\pgfusepath{clip}%
\pgfsetbuttcap%
\pgfsetroundjoin%
\definecolor{currentfill}{rgb}{0.298039,0.447059,0.690196}%
\pgfsetfillcolor{currentfill}%
\pgfsetfillopacity{0.200000}%
\pgfsetlinewidth{1.003750pt}%
\definecolor{currentstroke}{rgb}{0.298039,0.447059,0.690196}%
\pgfsetstrokecolor{currentstroke}%
\pgfsetstrokeopacity{0.200000}%
\pgfsetdash{}{0pt}%
\pgfpathmoveto{\pgfqpoint{1.089028in}{2.022898in}}%
\pgfpathlineto{\pgfqpoint{1.089028in}{1.687899in}}%
\pgfpathlineto{\pgfqpoint{1.264273in}{1.859165in}}%
\pgfpathlineto{\pgfqpoint{1.441156in}{1.855479in}}%
\pgfpathlineto{\pgfqpoint{1.616568in}{1.986802in}}%
\pgfpathlineto{\pgfqpoint{1.791150in}{1.961087in}}%
\pgfpathlineto{\pgfqpoint{1.968891in}{1.768738in}}%
\pgfpathlineto{\pgfqpoint{2.143118in}{1.899654in}}%
\pgfpathlineto{\pgfqpoint{2.322225in}{1.870983in}}%
\pgfpathlineto{\pgfqpoint{2.498742in}{1.818853in}}%
\pgfpathlineto{\pgfqpoint{2.674590in}{1.771707in}}%
\pgfpathlineto{\pgfqpoint{2.851517in}{1.805965in}}%
\pgfpathlineto{\pgfqpoint{3.026279in}{1.755741in}}%
\pgfpathlineto{\pgfqpoint{3.204726in}{1.745488in}}%
\pgfpathlineto{\pgfqpoint{3.381811in}{1.841878in}}%
\pgfpathlineto{\pgfqpoint{3.554796in}{1.868118in}}%
\pgfpathlineto{\pgfqpoint{3.731924in}{1.657087in}}%
\pgfpathlineto{\pgfqpoint{3.909831in}{1.766729in}}%
\pgfpathlineto{\pgfqpoint{4.083198in}{1.609190in}}%
\pgfpathlineto{\pgfqpoint{4.262040in}{1.842271in}}%
\pgfpathlineto{\pgfqpoint{4.437908in}{1.755138in}}%
\pgfpathlineto{\pgfqpoint{4.609306in}{1.741166in}}%
\pgfpathlineto{\pgfqpoint{4.789162in}{1.677735in}}%
\pgfpathlineto{\pgfqpoint{4.967437in}{1.661787in}}%
\pgfpathlineto{\pgfqpoint{5.140365in}{1.731979in}}%
\pgfpathlineto{\pgfqpoint{5.321320in}{1.701898in}}%
\pgfpathlineto{\pgfqpoint{5.493001in}{1.604550in}}%
\pgfpathlineto{\pgfqpoint{5.670989in}{1.444064in}}%
\pgfpathlineto{\pgfqpoint{5.847329in}{1.525572in}}%
\pgfpathlineto{\pgfqpoint{6.022461in}{1.482452in}}%
\pgfpathlineto{\pgfqpoint{6.201261in}{1.733249in}}%
\pgfpathlineto{\pgfqpoint{6.201261in}{1.978950in}}%
\pgfpathlineto{\pgfqpoint{6.201261in}{1.978950in}}%
\pgfpathlineto{\pgfqpoint{6.022461in}{1.876332in}}%
\pgfpathlineto{\pgfqpoint{5.847329in}{1.881356in}}%
\pgfpathlineto{\pgfqpoint{5.670989in}{1.767007in}}%
\pgfpathlineto{\pgfqpoint{5.493001in}{1.886663in}}%
\pgfpathlineto{\pgfqpoint{5.321320in}{1.976378in}}%
\pgfpathlineto{\pgfqpoint{5.140365in}{2.040956in}}%
\pgfpathlineto{\pgfqpoint{4.967437in}{1.987822in}}%
\pgfpathlineto{\pgfqpoint{4.789162in}{1.933759in}}%
\pgfpathlineto{\pgfqpoint{4.609306in}{1.987314in}}%
\pgfpathlineto{\pgfqpoint{4.437908in}{2.023785in}}%
\pgfpathlineto{\pgfqpoint{4.262040in}{2.105858in}}%
\pgfpathlineto{\pgfqpoint{4.083198in}{1.895328in}}%
\pgfpathlineto{\pgfqpoint{3.909831in}{2.076879in}}%
\pgfpathlineto{\pgfqpoint{3.731924in}{1.913907in}}%
\pgfpathlineto{\pgfqpoint{3.554796in}{2.145174in}}%
\pgfpathlineto{\pgfqpoint{3.381811in}{2.104793in}}%
\pgfpathlineto{\pgfqpoint{3.204726in}{1.994683in}}%
\pgfpathlineto{\pgfqpoint{3.026279in}{2.072163in}}%
\pgfpathlineto{\pgfqpoint{2.851517in}{2.078718in}}%
\pgfpathlineto{\pgfqpoint{2.674590in}{2.018692in}}%
\pgfpathlineto{\pgfqpoint{2.498742in}{2.053787in}}%
\pgfpathlineto{\pgfqpoint{2.322225in}{2.142433in}}%
\pgfpathlineto{\pgfqpoint{2.143118in}{2.131022in}}%
\pgfpathlineto{\pgfqpoint{1.968891in}{2.085336in}}%
\pgfpathlineto{\pgfqpoint{1.791150in}{2.240807in}}%
\pgfpathlineto{\pgfqpoint{1.616568in}{2.200216in}}%
\pgfpathlineto{\pgfqpoint{1.441156in}{2.136480in}}%
\pgfpathlineto{\pgfqpoint{1.264273in}{2.124344in}}%
\pgfpathlineto{\pgfqpoint{1.089028in}{2.022898in}}%
\pgfpathclose%
\pgfusepath{stroke,fill}%
\end{pgfscope}%
\begin{pgfscope}%
\pgfpathrectangle{\pgfqpoint{1.229207in}{0.915754in}}{\pgfqpoint{4.650000in}{3.020000in}}%
\pgfusepath{clip}%
\pgfsetbuttcap%
\pgfsetroundjoin%
\definecolor{currentfill}{rgb}{0.866667,0.517647,0.321569}%
\pgfsetfillcolor{currentfill}%
\pgfsetfillopacity{0.200000}%
\pgfsetlinewidth{1.003750pt}%
\definecolor{currentstroke}{rgb}{0.866667,0.517647,0.321569}%
\pgfsetstrokecolor{currentstroke}%
\pgfsetstrokeopacity{0.200000}%
\pgfsetdash{}{0pt}%
\pgfpathmoveto{\pgfqpoint{1.089666in}{1.903121in}}%
\pgfpathlineto{\pgfqpoint{1.089666in}{1.540363in}}%
\pgfpathlineto{\pgfqpoint{1.265689in}{1.747022in}}%
\pgfpathlineto{\pgfqpoint{1.441110in}{1.855239in}}%
\pgfpathlineto{\pgfqpoint{1.615478in}{1.776592in}}%
\pgfpathlineto{\pgfqpoint{1.794538in}{1.689402in}}%
\pgfpathlineto{\pgfqpoint{1.968346in}{1.574750in}}%
\pgfpathlineto{\pgfqpoint{2.144822in}{1.579309in}}%
\pgfpathlineto{\pgfqpoint{2.319359in}{1.620114in}}%
\pgfpathlineto{\pgfqpoint{2.495445in}{1.494950in}}%
\pgfpathlineto{\pgfqpoint{2.673755in}{1.852895in}}%
\pgfpathlineto{\pgfqpoint{2.849552in}{1.840019in}}%
\pgfpathlineto{\pgfqpoint{3.026042in}{1.825512in}}%
\pgfpathlineto{\pgfqpoint{3.205178in}{1.839591in}}%
\pgfpathlineto{\pgfqpoint{3.377738in}{1.580344in}}%
\pgfpathlineto{\pgfqpoint{3.556219in}{1.568113in}}%
\pgfpathlineto{\pgfqpoint{3.732015in}{1.455922in}}%
\pgfpathlineto{\pgfqpoint{3.911362in}{1.619566in}}%
\pgfpathlineto{\pgfqpoint{4.088755in}{1.644104in}}%
\pgfpathlineto{\pgfqpoint{4.260223in}{1.646174in}}%
\pgfpathlineto{\pgfqpoint{4.439105in}{1.645599in}}%
\pgfpathlineto{\pgfqpoint{4.614493in}{1.624156in}}%
\pgfpathlineto{\pgfqpoint{4.790769in}{1.468913in}}%
\pgfpathlineto{\pgfqpoint{4.964881in}{1.547510in}}%
\pgfpathlineto{\pgfqpoint{5.142310in}{1.508758in}}%
\pgfpathlineto{\pgfqpoint{5.316986in}{1.555442in}}%
\pgfpathlineto{\pgfqpoint{5.494555in}{1.522814in}}%
\pgfpathlineto{\pgfqpoint{5.671909in}{1.525830in}}%
\pgfpathlineto{\pgfqpoint{5.845348in}{1.694069in}}%
\pgfpathlineto{\pgfqpoint{6.023142in}{1.554503in}}%
\pgfpathlineto{\pgfqpoint{6.200859in}{1.443819in}}%
\pgfpathlineto{\pgfqpoint{6.200859in}{1.804246in}}%
\pgfpathlineto{\pgfqpoint{6.200859in}{1.804246in}}%
\pgfpathlineto{\pgfqpoint{6.023142in}{1.887545in}}%
\pgfpathlineto{\pgfqpoint{5.845348in}{1.933267in}}%
\pgfpathlineto{\pgfqpoint{5.671909in}{1.871633in}}%
\pgfpathlineto{\pgfqpoint{5.494555in}{1.812997in}}%
\pgfpathlineto{\pgfqpoint{5.316986in}{1.915756in}}%
\pgfpathlineto{\pgfqpoint{5.142310in}{1.823495in}}%
\pgfpathlineto{\pgfqpoint{4.964881in}{1.889916in}}%
\pgfpathlineto{\pgfqpoint{4.790769in}{1.775921in}}%
\pgfpathlineto{\pgfqpoint{4.614493in}{1.935057in}}%
\pgfpathlineto{\pgfqpoint{4.439105in}{1.922096in}}%
\pgfpathlineto{\pgfqpoint{4.260223in}{1.972337in}}%
\pgfpathlineto{\pgfqpoint{4.088755in}{1.915127in}}%
\pgfpathlineto{\pgfqpoint{3.911362in}{1.869015in}}%
\pgfpathlineto{\pgfqpoint{3.732015in}{1.833381in}}%
\pgfpathlineto{\pgfqpoint{3.556219in}{1.896121in}}%
\pgfpathlineto{\pgfqpoint{3.377738in}{1.867131in}}%
\pgfpathlineto{\pgfqpoint{3.205178in}{2.123637in}}%
\pgfpathlineto{\pgfqpoint{3.026042in}{2.128227in}}%
\pgfpathlineto{\pgfqpoint{2.849552in}{2.091639in}}%
\pgfpathlineto{\pgfqpoint{2.673755in}{2.128179in}}%
\pgfpathlineto{\pgfqpoint{2.495445in}{1.868304in}}%
\pgfpathlineto{\pgfqpoint{2.319359in}{1.937151in}}%
\pgfpathlineto{\pgfqpoint{2.144822in}{1.939175in}}%
\pgfpathlineto{\pgfqpoint{1.968346in}{1.855840in}}%
\pgfpathlineto{\pgfqpoint{1.794538in}{2.000303in}}%
\pgfpathlineto{\pgfqpoint{1.615478in}{2.072287in}}%
\pgfpathlineto{\pgfqpoint{1.441110in}{2.134887in}}%
\pgfpathlineto{\pgfqpoint{1.265689in}{2.086547in}}%
\pgfpathlineto{\pgfqpoint{1.089666in}{1.903121in}}%
\pgfpathclose%
\pgfusepath{stroke,fill}%
\end{pgfscope}%
\begin{pgfscope}%
\pgfpathrectangle{\pgfqpoint{1.229207in}{0.915754in}}{\pgfqpoint{4.650000in}{3.020000in}}%
\pgfusepath{clip}%
\pgfsetbuttcap%
\pgfsetroundjoin%
\definecolor{currentfill}{rgb}{0.333333,0.658824,0.407843}%
\pgfsetfillcolor{currentfill}%
\pgfsetfillopacity{0.200000}%
\pgfsetlinewidth{1.003750pt}%
\definecolor{currentstroke}{rgb}{0.333333,0.658824,0.407843}%
\pgfsetstrokecolor{currentstroke}%
\pgfsetstrokeopacity{0.200000}%
\pgfsetdash{}{0pt}%
\pgfpathmoveto{\pgfqpoint{1.144757in}{3.439721in}}%
\pgfpathlineto{\pgfqpoint{1.144757in}{3.085166in}}%
\pgfpathlineto{\pgfqpoint{1.322897in}{2.776396in}}%
\pgfpathlineto{\pgfqpoint{1.500065in}{2.691306in}}%
\pgfpathlineto{\pgfqpoint{1.673292in}{2.150543in}}%
\pgfpathlineto{\pgfqpoint{1.863605in}{1.877495in}}%
\pgfpathlineto{\pgfqpoint{2.058614in}{2.273375in}}%
\pgfpathlineto{\pgfqpoint{2.246967in}{1.958046in}}%
\pgfpathlineto{\pgfqpoint{2.419962in}{2.051913in}}%
\pgfpathlineto{\pgfqpoint{2.608677in}{1.907972in}}%
\pgfpathlineto{\pgfqpoint{2.788308in}{2.078264in}}%
\pgfpathlineto{\pgfqpoint{2.977959in}{1.834466in}}%
\pgfpathlineto{\pgfqpoint{3.171638in}{2.121173in}}%
\pgfpathlineto{\pgfqpoint{3.340789in}{1.896916in}}%
\pgfpathlineto{\pgfqpoint{3.519887in}{1.621338in}}%
\pgfpathlineto{\pgfqpoint{3.711443in}{1.873358in}}%
\pgfpathlineto{\pgfqpoint{3.885866in}{1.920129in}}%
\pgfpathlineto{\pgfqpoint{4.081932in}{1.751846in}}%
\pgfpathlineto{\pgfqpoint{4.264802in}{1.867477in}}%
\pgfpathlineto{\pgfqpoint{4.418574in}{1.917512in}}%
\pgfpathlineto{\pgfqpoint{4.627086in}{1.796470in}}%
\pgfpathlineto{\pgfqpoint{4.858007in}{1.788329in}}%
\pgfpathlineto{\pgfqpoint{4.977441in}{1.972475in}}%
\pgfpathlineto{\pgfqpoint{5.197900in}{1.812629in}}%
\pgfpathlineto{\pgfqpoint{5.333556in}{1.907280in}}%
\pgfpathlineto{\pgfqpoint{5.521432in}{1.756113in}}%
\pgfpathlineto{\pgfqpoint{5.723730in}{1.882844in}}%
\pgfpathlineto{\pgfqpoint{5.918497in}{1.682397in}}%
\pgfpathlineto{\pgfqpoint{6.090552in}{1.773172in}}%
\pgfpathlineto{\pgfqpoint{6.257240in}{1.820412in}}%
\pgfpathlineto{\pgfqpoint{6.459097in}{1.895319in}}%
\pgfpathlineto{\pgfqpoint{6.459097in}{2.244733in}}%
\pgfpathlineto{\pgfqpoint{6.459097in}{2.244733in}}%
\pgfpathlineto{\pgfqpoint{6.257240in}{2.117592in}}%
\pgfpathlineto{\pgfqpoint{6.090552in}{2.131008in}}%
\pgfpathlineto{\pgfqpoint{5.918497in}{2.019176in}}%
\pgfpathlineto{\pgfqpoint{5.723730in}{2.314263in}}%
\pgfpathlineto{\pgfqpoint{5.521432in}{2.062920in}}%
\pgfpathlineto{\pgfqpoint{5.333556in}{2.216698in}}%
\pgfpathlineto{\pgfqpoint{5.197900in}{2.090691in}}%
\pgfpathlineto{\pgfqpoint{4.977441in}{2.319046in}}%
\pgfpathlineto{\pgfqpoint{4.858007in}{2.302525in}}%
\pgfpathlineto{\pgfqpoint{4.627086in}{2.227042in}}%
\pgfpathlineto{\pgfqpoint{4.418574in}{2.423258in}}%
\pgfpathlineto{\pgfqpoint{4.264802in}{2.192214in}}%
\pgfpathlineto{\pgfqpoint{4.081932in}{2.148457in}}%
\pgfpathlineto{\pgfqpoint{3.885866in}{2.215892in}}%
\pgfpathlineto{\pgfqpoint{3.711443in}{2.177303in}}%
\pgfpathlineto{\pgfqpoint{3.519887in}{2.184738in}}%
\pgfpathlineto{\pgfqpoint{3.340789in}{2.268619in}}%
\pgfpathlineto{\pgfqpoint{3.171638in}{2.451257in}}%
\pgfpathlineto{\pgfqpoint{2.977959in}{2.174263in}}%
\pgfpathlineto{\pgfqpoint{2.788308in}{2.436290in}}%
\pgfpathlineto{\pgfqpoint{2.608677in}{2.186899in}}%
\pgfpathlineto{\pgfqpoint{2.419962in}{2.547966in}}%
\pgfpathlineto{\pgfqpoint{2.246967in}{2.336602in}}%
\pgfpathlineto{\pgfqpoint{2.058614in}{2.712697in}}%
\pgfpathlineto{\pgfqpoint{1.863605in}{2.555752in}}%
\pgfpathlineto{\pgfqpoint{1.673292in}{2.551083in}}%
\pgfpathlineto{\pgfqpoint{1.500065in}{3.416327in}}%
\pgfpathlineto{\pgfqpoint{1.322897in}{3.246314in}}%
\pgfpathlineto{\pgfqpoint{1.144757in}{3.439721in}}%
\pgfpathclose%
\pgfusepath{stroke,fill}%
\end{pgfscope}%
\begin{pgfscope}%
\pgfpathrectangle{\pgfqpoint{1.229207in}{0.915754in}}{\pgfqpoint{4.650000in}{3.020000in}}%
\pgfusepath{clip}%
\pgfsetbuttcap%
\pgfsetroundjoin%
\definecolor{currentfill}{rgb}{0.768627,0.305882,0.321569}%
\pgfsetfillcolor{currentfill}%
\pgfsetfillopacity{0.200000}%
\pgfsetlinewidth{1.003750pt}%
\definecolor{currentstroke}{rgb}{0.768627,0.305882,0.321569}%
\pgfsetstrokecolor{currentstroke}%
\pgfsetstrokeopacity{0.200000}%
\pgfsetdash{}{0pt}%
\pgfpathmoveto{\pgfqpoint{1.063136in}{3.728957in}}%
\pgfpathlineto{\pgfqpoint{1.063136in}{3.350756in}}%
\pgfpathlineto{\pgfqpoint{1.234934in}{2.986452in}}%
\pgfpathlineto{\pgfqpoint{1.406732in}{2.961694in}}%
\pgfpathlineto{\pgfqpoint{1.578530in}{3.089917in}}%
\pgfpathlineto{\pgfqpoint{1.750328in}{2.995780in}}%
\pgfpathlineto{\pgfqpoint{1.922126in}{2.764660in}}%
\pgfpathlineto{\pgfqpoint{2.093924in}{2.904856in}}%
\pgfpathlineto{\pgfqpoint{2.265722in}{2.932507in}}%
\pgfpathlineto{\pgfqpoint{2.437520in}{2.767224in}}%
\pgfpathlineto{\pgfqpoint{2.609318in}{2.730214in}}%
\pgfpathlineto{\pgfqpoint{2.781116in}{2.621520in}}%
\pgfpathlineto{\pgfqpoint{2.952914in}{2.555572in}}%
\pgfpathlineto{\pgfqpoint{3.124712in}{2.749184in}}%
\pgfpathlineto{\pgfqpoint{3.296510in}{2.814766in}}%
\pgfpathlineto{\pgfqpoint{3.468308in}{2.727867in}}%
\pgfpathlineto{\pgfqpoint{3.640107in}{2.725380in}}%
\pgfpathlineto{\pgfqpoint{3.811905in}{2.677517in}}%
\pgfpathlineto{\pgfqpoint{3.983703in}{2.793944in}}%
\pgfpathlineto{\pgfqpoint{4.155501in}{2.545660in}}%
\pgfpathlineto{\pgfqpoint{4.327299in}{2.533222in}}%
\pgfpathlineto{\pgfqpoint{4.499097in}{2.652939in}}%
\pgfpathlineto{\pgfqpoint{4.670895in}{2.546675in}}%
\pgfpathlineto{\pgfqpoint{4.842693in}{2.498867in}}%
\pgfpathlineto{\pgfqpoint{5.014491in}{2.608572in}}%
\pgfpathlineto{\pgfqpoint{5.186289in}{2.504477in}}%
\pgfpathlineto{\pgfqpoint{5.358087in}{2.352965in}}%
\pgfpathlineto{\pgfqpoint{5.529885in}{2.466145in}}%
\pgfpathlineto{\pgfqpoint{5.701683in}{2.393144in}}%
\pgfpathlineto{\pgfqpoint{5.873481in}{2.534574in}}%
\pgfpathlineto{\pgfqpoint{6.045279in}{2.498484in}}%
\pgfpathlineto{\pgfqpoint{6.045279in}{2.827327in}}%
\pgfpathlineto{\pgfqpoint{6.045279in}{2.827327in}}%
\pgfpathlineto{\pgfqpoint{5.873481in}{2.823791in}}%
\pgfpathlineto{\pgfqpoint{5.701683in}{2.740325in}}%
\pgfpathlineto{\pgfqpoint{5.529885in}{2.839408in}}%
\pgfpathlineto{\pgfqpoint{5.358087in}{2.760401in}}%
\pgfpathlineto{\pgfqpoint{5.186289in}{2.801477in}}%
\pgfpathlineto{\pgfqpoint{5.014491in}{2.941918in}}%
\pgfpathlineto{\pgfqpoint{4.842693in}{2.838635in}}%
\pgfpathlineto{\pgfqpoint{4.670895in}{2.900060in}}%
\pgfpathlineto{\pgfqpoint{4.499097in}{3.011431in}}%
\pgfpathlineto{\pgfqpoint{4.327299in}{2.862757in}}%
\pgfpathlineto{\pgfqpoint{4.155501in}{2.886664in}}%
\pgfpathlineto{\pgfqpoint{3.983703in}{3.028363in}}%
\pgfpathlineto{\pgfqpoint{3.811905in}{3.056519in}}%
\pgfpathlineto{\pgfqpoint{3.640107in}{3.018143in}}%
\pgfpathlineto{\pgfqpoint{3.468308in}{3.025618in}}%
\pgfpathlineto{\pgfqpoint{3.296510in}{3.173367in}}%
\pgfpathlineto{\pgfqpoint{3.124712in}{3.065860in}}%
\pgfpathlineto{\pgfqpoint{2.952914in}{2.953679in}}%
\pgfpathlineto{\pgfqpoint{2.781116in}{2.940731in}}%
\pgfpathlineto{\pgfqpoint{2.609318in}{3.086845in}}%
\pgfpathlineto{\pgfqpoint{2.437520in}{3.046654in}}%
\pgfpathlineto{\pgfqpoint{2.265722in}{3.210194in}}%
\pgfpathlineto{\pgfqpoint{2.093924in}{3.155657in}}%
\pgfpathlineto{\pgfqpoint{1.922126in}{3.088035in}}%
\pgfpathlineto{\pgfqpoint{1.750328in}{3.280621in}}%
\pgfpathlineto{\pgfqpoint{1.578530in}{3.353610in}}%
\pgfpathlineto{\pgfqpoint{1.406732in}{3.267926in}}%
\pgfpathlineto{\pgfqpoint{1.234934in}{3.338194in}}%
\pgfpathlineto{\pgfqpoint{1.063136in}{3.728957in}}%
\pgfpathclose%
\pgfusepath{stroke,fill}%
\end{pgfscope}%
\begin{pgfscope}%
\pgfpathrectangle{\pgfqpoint{1.229207in}{0.915754in}}{\pgfqpoint{4.650000in}{3.020000in}}%
\pgfusepath{clip}%
\pgfsetroundcap%
\pgfsetroundjoin%
\pgfsetlinewidth{1.505625pt}%
\definecolor{currentstroke}{rgb}{0.298039,0.447059,0.690196}%
\pgfsetstrokecolor{currentstroke}%
\pgfsetdash{}{0pt}%
\pgfpathmoveto{\pgfqpoint{1.215319in}{1.962463in}}%
\pgfpathlineto{\pgfqpoint{1.264273in}{1.997756in}}%
\pgfpathlineto{\pgfqpoint{1.441156in}{2.004406in}}%
\pgfpathlineto{\pgfqpoint{1.616568in}{2.104117in}}%
\pgfpathlineto{\pgfqpoint{1.791150in}{2.117123in}}%
\pgfpathlineto{\pgfqpoint{1.968891in}{1.939232in}}%
\pgfpathlineto{\pgfqpoint{2.143118in}{2.017384in}}%
\pgfpathlineto{\pgfqpoint{2.322225in}{2.026065in}}%
\pgfpathlineto{\pgfqpoint{2.498742in}{1.940289in}}%
\pgfpathlineto{\pgfqpoint{2.674590in}{1.909201in}}%
\pgfpathlineto{\pgfqpoint{2.851517in}{1.950070in}}%
\pgfpathlineto{\pgfqpoint{3.026279in}{1.925738in}}%
\pgfpathlineto{\pgfqpoint{3.204726in}{1.874672in}}%
\pgfpathlineto{\pgfqpoint{3.381811in}{1.980488in}}%
\pgfpathlineto{\pgfqpoint{3.554796in}{2.017823in}}%
\pgfpathlineto{\pgfqpoint{3.731924in}{1.801472in}}%
\pgfpathlineto{\pgfqpoint{3.909831in}{1.940915in}}%
\pgfpathlineto{\pgfqpoint{4.083198in}{1.763152in}}%
\pgfpathlineto{\pgfqpoint{4.262040in}{1.983484in}}%
\pgfpathlineto{\pgfqpoint{4.437908in}{1.906745in}}%
\pgfpathlineto{\pgfqpoint{4.609306in}{1.864625in}}%
\pgfpathlineto{\pgfqpoint{4.789162in}{1.820223in}}%
\pgfpathlineto{\pgfqpoint{4.967437in}{1.844224in}}%
\pgfpathlineto{\pgfqpoint{5.140365in}{1.895054in}}%
\pgfpathlineto{\pgfqpoint{5.321320in}{1.855598in}}%
\pgfpathlineto{\pgfqpoint{5.493001in}{1.760076in}}%
\pgfpathlineto{\pgfqpoint{5.670989in}{1.619333in}}%
\pgfpathlineto{\pgfqpoint{5.847329in}{1.733697in}}%
\pgfpathlineto{\pgfqpoint{5.893096in}{1.724937in}}%
\pgfusepath{stroke}%
\end{pgfscope}%
\begin{pgfscope}%
\pgfpathrectangle{\pgfqpoint{1.229207in}{0.915754in}}{\pgfqpoint{4.650000in}{3.020000in}}%
\pgfusepath{clip}%
\pgfsetroundcap%
\pgfsetroundjoin%
\pgfsetlinewidth{1.505625pt}%
\definecolor{currentstroke}{rgb}{0.866667,0.517647,0.321569}%
\pgfsetstrokecolor{currentstroke}%
\pgfsetdash{}{0pt}%
\pgfpathmoveto{\pgfqpoint{1.215319in}{1.882883in}}%
\pgfpathlineto{\pgfqpoint{1.265689in}{1.938243in}}%
\pgfpathlineto{\pgfqpoint{1.441110in}{2.007066in}}%
\pgfpathlineto{\pgfqpoint{1.615478in}{1.945561in}}%
\pgfpathlineto{\pgfqpoint{1.794538in}{1.866544in}}%
\pgfpathlineto{\pgfqpoint{1.968346in}{1.727395in}}%
\pgfpathlineto{\pgfqpoint{2.144822in}{1.766437in}}%
\pgfpathlineto{\pgfqpoint{2.319359in}{1.799557in}}%
\pgfpathlineto{\pgfqpoint{2.495445in}{1.711518in}}%
\pgfpathlineto{\pgfqpoint{2.673755in}{2.003775in}}%
\pgfpathlineto{\pgfqpoint{2.849552in}{1.983446in}}%
\pgfpathlineto{\pgfqpoint{3.026042in}{1.986426in}}%
\pgfpathlineto{\pgfqpoint{3.205178in}{2.001970in}}%
\pgfpathlineto{\pgfqpoint{3.377738in}{1.739820in}}%
\pgfpathlineto{\pgfqpoint{3.556219in}{1.747338in}}%
\pgfpathlineto{\pgfqpoint{3.732015in}{1.670358in}}%
\pgfpathlineto{\pgfqpoint{3.911362in}{1.751459in}}%
\pgfpathlineto{\pgfqpoint{4.088755in}{1.790844in}}%
\pgfpathlineto{\pgfqpoint{4.260223in}{1.820706in}}%
\pgfpathlineto{\pgfqpoint{4.439105in}{1.803015in}}%
\pgfpathlineto{\pgfqpoint{4.614493in}{1.789442in}}%
\pgfpathlineto{\pgfqpoint{4.790769in}{1.638295in}}%
\pgfpathlineto{\pgfqpoint{4.964881in}{1.732664in}}%
\pgfpathlineto{\pgfqpoint{5.142310in}{1.676293in}}%
\pgfpathlineto{\pgfqpoint{5.316986in}{1.760915in}}%
\pgfpathlineto{\pgfqpoint{5.494555in}{1.692051in}}%
\pgfpathlineto{\pgfqpoint{5.671909in}{1.715836in}}%
\pgfpathlineto{\pgfqpoint{5.845348in}{1.822235in}}%
\pgfpathlineto{\pgfqpoint{5.893096in}{1.798729in}}%
\pgfusepath{stroke}%
\end{pgfscope}%
\begin{pgfscope}%
\pgfpathrectangle{\pgfqpoint{1.229207in}{0.915754in}}{\pgfqpoint{4.650000in}{3.020000in}}%
\pgfusepath{clip}%
\pgfsetroundcap%
\pgfsetroundjoin%
\pgfsetlinewidth{1.505625pt}%
\definecolor{currentstroke}{rgb}{0.333333,0.658824,0.407843}%
\pgfsetstrokecolor{currentstroke}%
\pgfsetdash{}{0pt}%
\pgfpathmoveto{\pgfqpoint{1.215319in}{3.182284in}}%
\pgfpathlineto{\pgfqpoint{1.322897in}{3.031678in}}%
\pgfpathlineto{\pgfqpoint{1.500065in}{3.091122in}}%
\pgfpathlineto{\pgfqpoint{1.673292in}{2.378157in}}%
\pgfpathlineto{\pgfqpoint{1.863605in}{2.220058in}}%
\pgfpathlineto{\pgfqpoint{2.058614in}{2.518056in}}%
\pgfpathlineto{\pgfqpoint{2.246967in}{2.168428in}}%
\pgfpathlineto{\pgfqpoint{2.419962in}{2.317646in}}%
\pgfpathlineto{\pgfqpoint{2.608677in}{2.057409in}}%
\pgfpathlineto{\pgfqpoint{2.788308in}{2.266771in}}%
\pgfpathlineto{\pgfqpoint{2.977959in}{2.029267in}}%
\pgfpathlineto{\pgfqpoint{3.171638in}{2.290654in}}%
\pgfpathlineto{\pgfqpoint{3.340789in}{2.090737in}}%
\pgfpathlineto{\pgfqpoint{3.519887in}{1.927750in}}%
\pgfpathlineto{\pgfqpoint{3.711443in}{2.037987in}}%
\pgfpathlineto{\pgfqpoint{3.885866in}{2.087616in}}%
\pgfpathlineto{\pgfqpoint{4.081932in}{1.975232in}}%
\pgfpathlineto{\pgfqpoint{4.264802in}{2.039228in}}%
\pgfpathlineto{\pgfqpoint{4.418574in}{2.183843in}}%
\pgfpathlineto{\pgfqpoint{4.627086in}{2.044427in}}%
\pgfpathlineto{\pgfqpoint{4.858007in}{2.073579in}}%
\pgfpathlineto{\pgfqpoint{4.977441in}{2.150522in}}%
\pgfpathlineto{\pgfqpoint{5.197900in}{1.964716in}}%
\pgfpathlineto{\pgfqpoint{5.333556in}{2.072406in}}%
\pgfpathlineto{\pgfqpoint{5.521432in}{1.922995in}}%
\pgfpathlineto{\pgfqpoint{5.723730in}{2.124549in}}%
\pgfpathlineto{\pgfqpoint{5.893096in}{1.903792in}}%
\pgfusepath{stroke}%
\end{pgfscope}%
\begin{pgfscope}%
\pgfpathrectangle{\pgfqpoint{1.229207in}{0.915754in}}{\pgfqpoint{4.650000in}{3.020000in}}%
\pgfusepath{clip}%
\pgfsetroundcap%
\pgfsetroundjoin%
\pgfsetlinewidth{1.505625pt}%
\definecolor{currentstroke}{rgb}{0.768627,0.305882,0.321569}%
\pgfsetstrokecolor{currentstroke}%
\pgfsetdash{}{0pt}%
\pgfpathmoveto{\pgfqpoint{1.215319in}{3.226764in}}%
\pgfpathlineto{\pgfqpoint{1.234934in}{3.183181in}}%
\pgfpathlineto{\pgfqpoint{1.406732in}{3.138680in}}%
\pgfpathlineto{\pgfqpoint{1.578530in}{3.239730in}}%
\pgfpathlineto{\pgfqpoint{1.750328in}{3.151295in}}%
\pgfpathlineto{\pgfqpoint{1.922126in}{2.951751in}}%
\pgfpathlineto{\pgfqpoint{2.093924in}{3.039832in}}%
\pgfpathlineto{\pgfqpoint{2.265722in}{3.088465in}}%
\pgfpathlineto{\pgfqpoint{2.437520in}{2.927656in}}%
\pgfpathlineto{\pgfqpoint{2.609318in}{2.931182in}}%
\pgfpathlineto{\pgfqpoint{2.781116in}{2.801763in}}%
\pgfpathlineto{\pgfqpoint{2.952914in}{2.772370in}}%
\pgfpathlineto{\pgfqpoint{3.124712in}{2.922458in}}%
\pgfpathlineto{\pgfqpoint{3.296510in}{3.019136in}}%
\pgfpathlineto{\pgfqpoint{3.468308in}{2.880482in}}%
\pgfpathlineto{\pgfqpoint{3.640107in}{2.883384in}}%
\pgfpathlineto{\pgfqpoint{3.811905in}{2.889326in}}%
\pgfpathlineto{\pgfqpoint{3.983703in}{2.921572in}}%
\pgfpathlineto{\pgfqpoint{4.155501in}{2.731158in}}%
\pgfpathlineto{\pgfqpoint{4.327299in}{2.716653in}}%
\pgfpathlineto{\pgfqpoint{4.499097in}{2.857440in}}%
\pgfpathlineto{\pgfqpoint{4.670895in}{2.737198in}}%
\pgfpathlineto{\pgfqpoint{4.842693in}{2.687855in}}%
\pgfpathlineto{\pgfqpoint{5.014491in}{2.796937in}}%
\pgfpathlineto{\pgfqpoint{5.186289in}{2.669206in}}%
\pgfpathlineto{\pgfqpoint{5.358087in}{2.589251in}}%
\pgfpathlineto{\pgfqpoint{5.529885in}{2.676818in}}%
\pgfpathlineto{\pgfqpoint{5.701683in}{2.588803in}}%
\pgfpathlineto{\pgfqpoint{5.873481in}{2.689946in}}%
\pgfpathlineto{\pgfqpoint{5.893096in}{2.687823in}}%
\pgfusepath{stroke}%
\end{pgfscope}%
\begin{pgfscope}%
\pgfsetrectcap%
\pgfsetmiterjoin%
\pgfsetlinewidth{1.254687pt}%
\definecolor{currentstroke}{rgb}{0.800000,0.800000,0.800000}%
\pgfsetstrokecolor{currentstroke}%
\pgfsetdash{}{0pt}%
\pgfpathmoveto{\pgfqpoint{1.229207in}{0.915754in}}%
\pgfpathlineto{\pgfqpoint{1.229207in}{3.935754in}}%
\pgfusepath{stroke}%
\end{pgfscope}%
\begin{pgfscope}%
\pgfsetrectcap%
\pgfsetmiterjoin%
\pgfsetlinewidth{1.254687pt}%
\definecolor{currentstroke}{rgb}{0.800000,0.800000,0.800000}%
\pgfsetstrokecolor{currentstroke}%
\pgfsetdash{}{0pt}%
\pgfpathmoveto{\pgfqpoint{5.879208in}{0.915754in}}%
\pgfpathlineto{\pgfqpoint{5.879208in}{3.935754in}}%
\pgfusepath{stroke}%
\end{pgfscope}%
\begin{pgfscope}%
\pgfsetrectcap%
\pgfsetmiterjoin%
\pgfsetlinewidth{1.254687pt}%
\definecolor{currentstroke}{rgb}{0.800000,0.800000,0.800000}%
\pgfsetstrokecolor{currentstroke}%
\pgfsetdash{}{0pt}%
\pgfpathmoveto{\pgfqpoint{1.229207in}{0.915754in}}%
\pgfpathlineto{\pgfqpoint{5.879208in}{0.915754in}}%
\pgfusepath{stroke}%
\end{pgfscope}%
\begin{pgfscope}%
\pgfsetrectcap%
\pgfsetmiterjoin%
\pgfsetlinewidth{1.254687pt}%
\definecolor{currentstroke}{rgb}{0.800000,0.800000,0.800000}%
\pgfsetstrokecolor{currentstroke}%
\pgfsetdash{}{0pt}%
\pgfpathmoveto{\pgfqpoint{1.229207in}{3.935754in}}%
\pgfpathlineto{\pgfqpoint{5.879208in}{3.935754in}}%
\pgfusepath{stroke}%
\end{pgfscope}%
\begin{pgfscope}%
\pgfsetbuttcap%
\pgfsetmiterjoin%
\definecolor{currentfill}{rgb}{1.000000,1.000000,1.000000}%
\pgfsetfillcolor{currentfill}%
\pgfsetfillopacity{0.800000}%
\pgfsetlinewidth{1.003750pt}%
\definecolor{currentstroke}{rgb}{0.800000,0.800000,0.800000}%
\pgfsetstrokecolor{currentstroke}%
\pgfsetstrokeopacity{0.800000}%
\pgfsetdash{}{0pt}%
\pgfpathmoveto{\pgfqpoint{1.966460in}{2.840139in}}%
\pgfpathlineto{\pgfqpoint{5.654624in}{2.840139in}}%
\pgfpathquadraticcurveto{\pgfqpoint{5.718791in}{2.840139in}}{\pgfqpoint{5.718791in}{2.904306in}}%
\pgfpathlineto{\pgfqpoint{5.718791in}{3.711171in}}%
\pgfpathquadraticcurveto{\pgfqpoint{5.718791in}{3.775338in}}{\pgfqpoint{5.654624in}{3.775338in}}%
\pgfpathlineto{\pgfqpoint{1.966460in}{3.775338in}}%
\pgfpathquadraticcurveto{\pgfqpoint{1.902294in}{3.775338in}}{\pgfqpoint{1.902294in}{3.711171in}}%
\pgfpathlineto{\pgfqpoint{1.902294in}{2.904306in}}%
\pgfpathquadraticcurveto{\pgfqpoint{1.902294in}{2.840139in}}{\pgfqpoint{1.966460in}{2.840139in}}%
\pgfpathclose%
\pgfusepath{stroke,fill}%
\end{pgfscope}%
\begin{pgfscope}%
\pgfsetroundcap%
\pgfsetroundjoin%
\pgfsetlinewidth{5.018750pt}%
\definecolor{currentstroke}{rgb}{0.298039,0.447059,0.690196}%
\pgfsetstrokecolor{currentstroke}%
\pgfsetdash{}{0pt}%
\pgfpathmoveto{\pgfqpoint{2.030627in}{3.519327in}}%
\pgfpathlineto{\pgfqpoint{2.672294in}{3.519327in}}%
\pgfusepath{stroke}%
\end{pgfscope}%
\begin{pgfscope}%
\definecolor{textcolor}{rgb}{0.150000,0.150000,0.150000}%
\pgfsetstrokecolor{textcolor}%
\pgfsetfillcolor{textcolor}%
\pgftext[x=2.928960in,y=3.407035in,left,base]{\color{textcolor}\rmfamily\fontsize{23.100000}{27.720000}\selectfont W-L}%
\end{pgfscope}%
\begin{pgfscope}%
\pgfsetroundcap%
\pgfsetroundjoin%
\pgfsetlinewidth{5.018750pt}%
\definecolor{currentstroke}{rgb}{0.866667,0.517647,0.321569}%
\pgfsetstrokecolor{currentstroke}%
\pgfsetdash{}{0pt}%
\pgfpathmoveto{\pgfqpoint{2.030627in}{3.147977in}}%
\pgfpathlineto{\pgfqpoint{2.672294in}{3.147977in}}%
\pgfusepath{stroke}%
\end{pgfscope}%
\begin{pgfscope}%
\definecolor{textcolor}{rgb}{0.150000,0.150000,0.150000}%
\pgfsetstrokecolor{textcolor}%
\pgfsetfillcolor{textcolor}%
\pgftext[x=2.928960in,y=3.035686in,left,base]{\color{textcolor}\rmfamily\fontsize{23.100000}{27.720000}\selectfont W-M}%
\end{pgfscope}%
\begin{pgfscope}%
\pgfsetroundcap%
\pgfsetroundjoin%
\pgfsetlinewidth{5.018750pt}%
\definecolor{currentstroke}{rgb}{0.333333,0.658824,0.407843}%
\pgfsetstrokecolor{currentstroke}%
\pgfsetdash{}{0pt}%
\pgfpathmoveto{\pgfqpoint{3.977694in}{3.519327in}}%
\pgfpathlineto{\pgfqpoint{4.619361in}{3.519327in}}%
\pgfusepath{stroke}%
\end{pgfscope}%
\begin{pgfscope}%
\definecolor{textcolor}{rgb}{0.150000,0.150000,0.150000}%
\pgfsetstrokecolor{textcolor}%
\pgfsetfillcolor{textcolor}%
\pgftext[x=4.876028in,y=3.407035in,left,base]{\color{textcolor}\rmfamily\fontsize{23.100000}{27.720000}\selectfont DCV}%
\end{pgfscope}%
\begin{pgfscope}%
\pgfsetroundcap%
\pgfsetroundjoin%
\pgfsetlinewidth{5.018750pt}%
\definecolor{currentstroke}{rgb}{0.768627,0.305882,0.321569}%
\pgfsetstrokecolor{currentstroke}%
\pgfsetdash{}{0pt}%
\pgfpathmoveto{\pgfqpoint{3.977694in}{3.147977in}}%
\pgfpathlineto{\pgfqpoint{4.619361in}{3.147977in}}%
\pgfusepath{stroke}%
\end{pgfscope}%
\begin{pgfscope}%
\definecolor{textcolor}{rgb}{0.150000,0.150000,0.150000}%
\pgfsetstrokecolor{textcolor}%
\pgfsetfillcolor{textcolor}%
\pgftext[x=4.876028in,y=3.035686in,left,base]{\color{textcolor}\rmfamily\fontsize{23.100000}{27.720000}\selectfont MC}%
\end{pgfscope}%
\end{pgfpicture}%
\makeatother%
\endgroup%

%% file: fig/continuous_circle-1000-4d_res.pgf
%% Creator: Matplotlib, PGF backend
%%
%% To include the figure in your LaTeX document, write
%%   \input{<filename>.pgf}
%%
%% Make sure the required packages are loaded in your preamble
%%   \usepackage{pgf}
%%
%% Figures using additional raster images can only be included by \input if
%% they are in the same directory as the main LaTeX file. For loading figures
%% from other directories you can use the `import` package
%% and then include the figures with
%%   \import{<path to file>}{<filename>.pgf}
%%
%% Matplotlib used the following preamble
%%
\begingroup%
\makeatletter%
\begin{pgfpicture}%
\pgfpathrectangle{\pgfpointorigin}{\pgfqpoint{4.850000in}{4.035754in}}%
\pgfusepath{use as bounding box, clip}%
\begin{pgfscope}%
\pgfsetbuttcap%
\pgfsetmiterjoin%
\definecolor{currentfill}{rgb}{1.000000,1.000000,1.000000}%
\pgfsetfillcolor{currentfill}%
\pgfsetlinewidth{0.000000pt}%
\definecolor{currentstroke}{rgb}{1.000000,1.000000,1.000000}%
\pgfsetstrokecolor{currentstroke}%
\pgfsetdash{}{0pt}%
\pgfpathmoveto{\pgfqpoint{0.000000in}{0.000000in}}%
\pgfpathlineto{\pgfqpoint{4.850000in}{0.000000in}}%
\pgfpathlineto{\pgfqpoint{4.850000in}{4.035754in}}%
\pgfpathlineto{\pgfqpoint{0.000000in}{4.035754in}}%
\pgfpathclose%
\pgfusepath{fill}%
\end{pgfscope}%
\begin{pgfscope}%
\pgfsetbuttcap%
\pgfsetmiterjoin%
\definecolor{currentfill}{rgb}{1.000000,1.000000,1.000000}%
\pgfsetfillcolor{currentfill}%
\pgfsetlinewidth{0.000000pt}%
\definecolor{currentstroke}{rgb}{0.000000,0.000000,0.000000}%
\pgfsetstrokecolor{currentstroke}%
\pgfsetstrokeopacity{0.000000}%
\pgfsetdash{}{0pt}%
\pgfpathmoveto{\pgfqpoint{0.100000in}{0.915754in}}%
\pgfpathlineto{\pgfqpoint{4.750000in}{0.915754in}}%
\pgfpathlineto{\pgfqpoint{4.750000in}{3.935754in}}%
\pgfpathlineto{\pgfqpoint{0.100000in}{3.935754in}}%
\pgfpathclose%
\pgfusepath{fill}%
\end{pgfscope}%
\begin{pgfscope}%
\pgfpathrectangle{\pgfqpoint{0.100000in}{0.915754in}}{\pgfqpoint{4.650000in}{3.020000in}}%
\pgfusepath{clip}%
\pgfsetroundcap%
\pgfsetroundjoin%
\pgfsetlinewidth{1.003750pt}%
\definecolor{currentstroke}{rgb}{0.800000,0.800000,0.800000}%
\pgfsetstrokecolor{currentstroke}%
\pgfsetdash{}{0pt}%
\pgfpathmoveto{\pgfqpoint{0.349107in}{0.915754in}}%
\pgfpathlineto{\pgfqpoint{0.349107in}{3.935754in}}%
\pgfusepath{stroke}%
\end{pgfscope}%
\begin{pgfscope}%
\definecolor{textcolor}{rgb}{0.150000,0.150000,0.150000}%
\pgfsetstrokecolor{textcolor}%
\pgfsetfillcolor{textcolor}%
\pgftext[x=0.349107in,y=0.783810in,,top]{\color{textcolor}\rmfamily\fontsize{23.100000}{27.720000}\selectfont 10}%
\end{pgfscope}%
\begin{pgfscope}%
\pgfpathrectangle{\pgfqpoint{0.100000in}{0.915754in}}{\pgfqpoint{4.650000in}{3.020000in}}%
\pgfusepath{clip}%
\pgfsetroundcap%
\pgfsetroundjoin%
\pgfsetlinewidth{1.003750pt}%
\definecolor{currentstroke}{rgb}{0.800000,0.800000,0.800000}%
\pgfsetstrokecolor{currentstroke}%
\pgfsetdash{}{0pt}%
\pgfpathmoveto{\pgfqpoint{1.179464in}{0.915754in}}%
\pgfpathlineto{\pgfqpoint{1.179464in}{3.935754in}}%
\pgfusepath{stroke}%
\end{pgfscope}%
\begin{pgfscope}%
\definecolor{textcolor}{rgb}{0.150000,0.150000,0.150000}%
\pgfsetstrokecolor{textcolor}%
\pgfsetfillcolor{textcolor}%
\pgftext[x=1.179464in,y=0.783810in,,top]{\color{textcolor}\rmfamily\fontsize{23.100000}{27.720000}\selectfont 20}%
\end{pgfscope}%
\begin{pgfscope}%
\pgfpathrectangle{\pgfqpoint{0.100000in}{0.915754in}}{\pgfqpoint{4.650000in}{3.020000in}}%
\pgfusepath{clip}%
\pgfsetroundcap%
\pgfsetroundjoin%
\pgfsetlinewidth{1.003750pt}%
\definecolor{currentstroke}{rgb}{0.800000,0.800000,0.800000}%
\pgfsetstrokecolor{currentstroke}%
\pgfsetdash{}{0pt}%
\pgfpathmoveto{\pgfqpoint{2.009821in}{0.915754in}}%
\pgfpathlineto{\pgfqpoint{2.009821in}{3.935754in}}%
\pgfusepath{stroke}%
\end{pgfscope}%
\begin{pgfscope}%
\definecolor{textcolor}{rgb}{0.150000,0.150000,0.150000}%
\pgfsetstrokecolor{textcolor}%
\pgfsetfillcolor{textcolor}%
\pgftext[x=2.009821in,y=0.783810in,,top]{\color{textcolor}\rmfamily\fontsize{23.100000}{27.720000}\selectfont 30}%
\end{pgfscope}%
\begin{pgfscope}%
\pgfpathrectangle{\pgfqpoint{0.100000in}{0.915754in}}{\pgfqpoint{4.650000in}{3.020000in}}%
\pgfusepath{clip}%
\pgfsetroundcap%
\pgfsetroundjoin%
\pgfsetlinewidth{1.003750pt}%
\definecolor{currentstroke}{rgb}{0.800000,0.800000,0.800000}%
\pgfsetstrokecolor{currentstroke}%
\pgfsetdash{}{0pt}%
\pgfpathmoveto{\pgfqpoint{2.840179in}{0.915754in}}%
\pgfpathlineto{\pgfqpoint{2.840179in}{3.935754in}}%
\pgfusepath{stroke}%
\end{pgfscope}%
\begin{pgfscope}%
\definecolor{textcolor}{rgb}{0.150000,0.150000,0.150000}%
\pgfsetstrokecolor{textcolor}%
\pgfsetfillcolor{textcolor}%
\pgftext[x=2.840179in,y=0.783810in,,top]{\color{textcolor}\rmfamily\fontsize{23.100000}{27.720000}\selectfont 40}%
\end{pgfscope}%
\begin{pgfscope}%
\pgfpathrectangle{\pgfqpoint{0.100000in}{0.915754in}}{\pgfqpoint{4.650000in}{3.020000in}}%
\pgfusepath{clip}%
\pgfsetroundcap%
\pgfsetroundjoin%
\pgfsetlinewidth{1.003750pt}%
\definecolor{currentstroke}{rgb}{0.800000,0.800000,0.800000}%
\pgfsetstrokecolor{currentstroke}%
\pgfsetdash{}{0pt}%
\pgfpathmoveto{\pgfqpoint{3.670536in}{0.915754in}}%
\pgfpathlineto{\pgfqpoint{3.670536in}{3.935754in}}%
\pgfusepath{stroke}%
\end{pgfscope}%
\begin{pgfscope}%
\definecolor{textcolor}{rgb}{0.150000,0.150000,0.150000}%
\pgfsetstrokecolor{textcolor}%
\pgfsetfillcolor{textcolor}%
\pgftext[x=3.670536in,y=0.783810in,,top]{\color{textcolor}\rmfamily\fontsize{23.100000}{27.720000}\selectfont 50}%
\end{pgfscope}%
\begin{pgfscope}%
\pgfpathrectangle{\pgfqpoint{0.100000in}{0.915754in}}{\pgfqpoint{4.650000in}{3.020000in}}%
\pgfusepath{clip}%
\pgfsetroundcap%
\pgfsetroundjoin%
\pgfsetlinewidth{1.003750pt}%
\definecolor{currentstroke}{rgb}{0.800000,0.800000,0.800000}%
\pgfsetstrokecolor{currentstroke}%
\pgfsetdash{}{0pt}%
\pgfpathmoveto{\pgfqpoint{4.500893in}{0.915754in}}%
\pgfpathlineto{\pgfqpoint{4.500893in}{3.935754in}}%
\pgfusepath{stroke}%
\end{pgfscope}%
\begin{pgfscope}%
\definecolor{textcolor}{rgb}{0.150000,0.150000,0.150000}%
\pgfsetstrokecolor{textcolor}%
\pgfsetfillcolor{textcolor}%
\pgftext[x=4.500893in,y=0.783810in,,top]{\color{textcolor}\rmfamily\fontsize{23.100000}{27.720000}\selectfont 60}%
\end{pgfscope}%
\begin{pgfscope}%
\definecolor{textcolor}{rgb}{0.150000,0.150000,0.150000}%
\pgfsetstrokecolor{textcolor}%
\pgfsetfillcolor{textcolor}%
\pgftext[x=2.425000in,y=0.407183in,,top]{\color{textcolor}\rmfamily\fontsize{25.200000}{30.240000}\selectfont runtime in seconds}%
\end{pgfscope}%
\begin{pgfscope}%
\pgfpathrectangle{\pgfqpoint{0.100000in}{0.915754in}}{\pgfqpoint{4.650000in}{3.020000in}}%
\pgfusepath{clip}%
\pgfsetroundcap%
\pgfsetroundjoin%
\pgfsetlinewidth{1.003750pt}%
\definecolor{currentstroke}{rgb}{0.800000,0.800000,0.800000}%
\pgfsetstrokecolor{currentstroke}%
\pgfsetdash{}{0pt}%
\pgfpathmoveto{\pgfqpoint{0.100000in}{1.328214in}}%
\pgfpathlineto{\pgfqpoint{4.750000in}{1.328214in}}%
\pgfusepath{stroke}%
\end{pgfscope}%
\begin{pgfscope}%
\pgfpathrectangle{\pgfqpoint{0.100000in}{0.915754in}}{\pgfqpoint{4.650000in}{3.020000in}}%
\pgfusepath{clip}%
\pgfsetroundcap%
\pgfsetroundjoin%
\pgfsetlinewidth{1.003750pt}%
\definecolor{currentstroke}{rgb}{0.800000,0.800000,0.800000}%
\pgfsetstrokecolor{currentstroke}%
\pgfsetdash{}{0pt}%
\pgfpathmoveto{\pgfqpoint{0.100000in}{2.698375in}}%
\pgfpathlineto{\pgfqpoint{4.750000in}{2.698375in}}%
\pgfusepath{stroke}%
\end{pgfscope}%
\begin{pgfscope}%
\pgfpathrectangle{\pgfqpoint{0.100000in}{0.915754in}}{\pgfqpoint{4.650000in}{3.020000in}}%
\pgfusepath{clip}%
\pgfsetbuttcap%
\pgfsetroundjoin%
\definecolor{currentfill}{rgb}{0.298039,0.447059,0.690196}%
\pgfsetfillcolor{currentfill}%
\pgfsetfillopacity{0.200000}%
\pgfsetlinewidth{1.003750pt}%
\definecolor{currentstroke}{rgb}{0.298039,0.447059,0.690196}%
\pgfsetstrokecolor{currentstroke}%
\pgfsetstrokeopacity{0.200000}%
\pgfsetdash{}{0pt}%
\pgfpathmoveto{\pgfqpoint{-0.023532in}{2.696291in}}%
\pgfpathlineto{\pgfqpoint{-0.023532in}{2.462493in}}%
\pgfpathlineto{\pgfqpoint{0.152068in}{2.171637in}}%
\pgfpathlineto{\pgfqpoint{0.329072in}{2.156547in}}%
\pgfpathlineto{\pgfqpoint{0.499467in}{1.650357in}}%
\pgfpathlineto{\pgfqpoint{0.673754in}{1.892405in}}%
\pgfpathlineto{\pgfqpoint{0.846819in}{1.508689in}}%
\pgfpathlineto{\pgfqpoint{1.022331in}{1.759430in}}%
\pgfpathlineto{\pgfqpoint{1.202483in}{1.705051in}}%
\pgfpathlineto{\pgfqpoint{1.377281in}{1.848266in}}%
\pgfpathlineto{\pgfqpoint{1.550563in}{1.707281in}}%
\pgfpathlineto{\pgfqpoint{1.726299in}{1.913482in}}%
\pgfpathlineto{\pgfqpoint{1.901176in}{1.842561in}}%
\pgfpathlineto{\pgfqpoint{2.073976in}{2.083888in}}%
\pgfpathlineto{\pgfqpoint{2.253660in}{1.788057in}}%
\pgfpathlineto{\pgfqpoint{2.425081in}{1.836695in}}%
\pgfpathlineto{\pgfqpoint{2.598336in}{1.950597in}}%
\pgfpathlineto{\pgfqpoint{2.777216in}{1.877382in}}%
\pgfpathlineto{\pgfqpoint{2.950069in}{1.998396in}}%
\pgfpathlineto{\pgfqpoint{3.127870in}{1.853674in}}%
\pgfpathlineto{\pgfqpoint{3.303124in}{2.015674in}}%
\pgfpathlineto{\pgfqpoint{3.472183in}{1.810592in}}%
\pgfpathlineto{\pgfqpoint{3.650796in}{2.004460in}}%
\pgfpathlineto{\pgfqpoint{3.832813in}{2.033510in}}%
\pgfpathlineto{\pgfqpoint{4.002021in}{1.921711in}}%
\pgfpathlineto{\pgfqpoint{4.175909in}{1.947825in}}%
\pgfpathlineto{\pgfqpoint{4.353019in}{1.931481in}}%
\pgfpathlineto{\pgfqpoint{4.531220in}{2.047470in}}%
\pgfpathlineto{\pgfqpoint{4.705425in}{1.968630in}}%
\pgfpathlineto{\pgfqpoint{4.880827in}{1.970677in}}%
\pgfpathlineto{\pgfqpoint{5.058331in}{1.935664in}}%
\pgfpathlineto{\pgfqpoint{5.058331in}{2.251059in}}%
\pgfpathlineto{\pgfqpoint{5.058331in}{2.251059in}}%
\pgfpathlineto{\pgfqpoint{4.880827in}{2.272218in}}%
\pgfpathlineto{\pgfqpoint{4.705425in}{2.275354in}}%
\pgfpathlineto{\pgfqpoint{4.531220in}{2.313529in}}%
\pgfpathlineto{\pgfqpoint{4.353019in}{2.241179in}}%
\pgfpathlineto{\pgfqpoint{4.175909in}{2.203766in}}%
\pgfpathlineto{\pgfqpoint{4.002021in}{2.166804in}}%
\pgfpathlineto{\pgfqpoint{3.832813in}{2.318534in}}%
\pgfpathlineto{\pgfqpoint{3.650796in}{2.296871in}}%
\pgfpathlineto{\pgfqpoint{3.472183in}{2.164368in}}%
\pgfpathlineto{\pgfqpoint{3.303124in}{2.297251in}}%
\pgfpathlineto{\pgfqpoint{3.127870in}{2.166177in}}%
\pgfpathlineto{\pgfqpoint{2.950069in}{2.282081in}}%
\pgfpathlineto{\pgfqpoint{2.777216in}{2.129048in}}%
\pgfpathlineto{\pgfqpoint{2.598336in}{2.221826in}}%
\pgfpathlineto{\pgfqpoint{2.425081in}{2.170138in}}%
\pgfpathlineto{\pgfqpoint{2.253660in}{2.089651in}}%
\pgfpathlineto{\pgfqpoint{2.073976in}{2.369665in}}%
\pgfpathlineto{\pgfqpoint{1.901176in}{2.209163in}}%
\pgfpathlineto{\pgfqpoint{1.726299in}{2.201160in}}%
\pgfpathlineto{\pgfqpoint{1.550563in}{2.087285in}}%
\pgfpathlineto{\pgfqpoint{1.377281in}{2.197202in}}%
\pgfpathlineto{\pgfqpoint{1.202483in}{2.000812in}}%
\pgfpathlineto{\pgfqpoint{1.022331in}{2.118799in}}%
\pgfpathlineto{\pgfqpoint{0.846819in}{1.832262in}}%
\pgfpathlineto{\pgfqpoint{0.673754in}{2.178554in}}%
\pgfpathlineto{\pgfqpoint{0.499467in}{1.999698in}}%
\pgfpathlineto{\pgfqpoint{0.329072in}{2.434118in}}%
\pgfpathlineto{\pgfqpoint{0.152068in}{2.527226in}}%
\pgfpathlineto{\pgfqpoint{-0.023532in}{2.696291in}}%
\pgfpathclose%
\pgfusepath{stroke,fill}%
\end{pgfscope}%
\begin{pgfscope}%
\pgfpathrectangle{\pgfqpoint{0.100000in}{0.915754in}}{\pgfqpoint{4.650000in}{3.020000in}}%
\pgfusepath{clip}%
\pgfsetbuttcap%
\pgfsetroundjoin%
\definecolor{currentfill}{rgb}{0.866667,0.517647,0.321569}%
\pgfsetfillcolor{currentfill}%
\pgfsetfillopacity{0.200000}%
\pgfsetlinewidth{1.003750pt}%
\definecolor{currentstroke}{rgb}{0.866667,0.517647,0.321569}%
\pgfsetstrokecolor{currentstroke}%
\pgfsetstrokeopacity{0.200000}%
\pgfsetdash{}{0pt}%
\pgfpathmoveto{\pgfqpoint{-0.020636in}{3.383314in}}%
\pgfpathlineto{\pgfqpoint{-0.020636in}{3.277222in}}%
\pgfpathlineto{\pgfqpoint{0.154184in}{3.082491in}}%
\pgfpathlineto{\pgfqpoint{0.322702in}{3.039017in}}%
\pgfpathlineto{\pgfqpoint{0.501250in}{2.908262in}}%
\pgfpathlineto{\pgfqpoint{0.677433in}{2.785757in}}%
\pgfpathlineto{\pgfqpoint{0.853174in}{2.649325in}}%
\pgfpathlineto{\pgfqpoint{1.022286in}{2.550422in}}%
\pgfpathlineto{\pgfqpoint{1.200355in}{2.442567in}}%
\pgfpathlineto{\pgfqpoint{1.372670in}{2.414896in}}%
\pgfpathlineto{\pgfqpoint{1.547231in}{2.575477in}}%
\pgfpathlineto{\pgfqpoint{1.721311in}{2.517014in}}%
\pgfpathlineto{\pgfqpoint{1.903084in}{2.733269in}}%
\pgfpathlineto{\pgfqpoint{2.073757in}{2.630979in}}%
\pgfpathlineto{\pgfqpoint{2.251539in}{2.585948in}}%
\pgfpathlineto{\pgfqpoint{2.426400in}{2.602105in}}%
\pgfpathlineto{\pgfqpoint{2.603936in}{2.785447in}}%
\pgfpathlineto{\pgfqpoint{2.773922in}{2.691651in}}%
\pgfpathlineto{\pgfqpoint{2.950697in}{2.608791in}}%
\pgfpathlineto{\pgfqpoint{3.132595in}{2.558418in}}%
\pgfpathlineto{\pgfqpoint{3.299308in}{2.576426in}}%
\pgfpathlineto{\pgfqpoint{3.478575in}{2.616258in}}%
\pgfpathlineto{\pgfqpoint{3.653441in}{2.584040in}}%
\pgfpathlineto{\pgfqpoint{3.828244in}{2.556533in}}%
\pgfpathlineto{\pgfqpoint{4.005932in}{2.329141in}}%
\pgfpathlineto{\pgfqpoint{4.176743in}{2.646798in}}%
\pgfpathlineto{\pgfqpoint{4.356597in}{2.658842in}}%
\pgfpathlineto{\pgfqpoint{4.530369in}{2.508268in}}%
\pgfpathlineto{\pgfqpoint{4.710803in}{2.622046in}}%
\pgfpathlineto{\pgfqpoint{4.883189in}{2.610811in}}%
\pgfpathlineto{\pgfqpoint{5.063738in}{2.678938in}}%
\pgfpathlineto{\pgfqpoint{5.063738in}{2.878759in}}%
\pgfpathlineto{\pgfqpoint{5.063738in}{2.878759in}}%
\pgfpathlineto{\pgfqpoint{4.883189in}{2.846036in}}%
\pgfpathlineto{\pgfqpoint{4.710803in}{2.849297in}}%
\pgfpathlineto{\pgfqpoint{4.530369in}{2.781445in}}%
\pgfpathlineto{\pgfqpoint{4.356597in}{2.847897in}}%
\pgfpathlineto{\pgfqpoint{4.176743in}{2.861364in}}%
\pgfpathlineto{\pgfqpoint{4.005932in}{2.624998in}}%
\pgfpathlineto{\pgfqpoint{3.828244in}{2.795137in}}%
\pgfpathlineto{\pgfqpoint{3.653441in}{2.815358in}}%
\pgfpathlineto{\pgfqpoint{3.478575in}{2.788591in}}%
\pgfpathlineto{\pgfqpoint{3.299308in}{2.759664in}}%
\pgfpathlineto{\pgfqpoint{3.132595in}{2.760287in}}%
\pgfpathlineto{\pgfqpoint{2.950697in}{2.823172in}}%
\pgfpathlineto{\pgfqpoint{2.773922in}{2.858736in}}%
\pgfpathlineto{\pgfqpoint{2.603936in}{2.876727in}}%
\pgfpathlineto{\pgfqpoint{2.426400in}{2.830233in}}%
\pgfpathlineto{\pgfqpoint{2.251539in}{2.818256in}}%
\pgfpathlineto{\pgfqpoint{2.073757in}{2.871456in}}%
\pgfpathlineto{\pgfqpoint{1.903084in}{2.916318in}}%
\pgfpathlineto{\pgfqpoint{1.721311in}{2.746518in}}%
\pgfpathlineto{\pgfqpoint{1.547231in}{2.780147in}}%
\pgfpathlineto{\pgfqpoint{1.372670in}{2.638896in}}%
\pgfpathlineto{\pgfqpoint{1.200355in}{2.615683in}}%
\pgfpathlineto{\pgfqpoint{1.022286in}{2.734182in}}%
\pgfpathlineto{\pgfqpoint{0.853174in}{2.792776in}}%
\pgfpathlineto{\pgfqpoint{0.677433in}{2.923677in}}%
\pgfpathlineto{\pgfqpoint{0.501250in}{3.040409in}}%
\pgfpathlineto{\pgfqpoint{0.322702in}{3.153009in}}%
\pgfpathlineto{\pgfqpoint{0.154184in}{3.233270in}}%
\pgfpathlineto{\pgfqpoint{-0.020636in}{3.383314in}}%
\pgfpathclose%
\pgfusepath{stroke,fill}%
\end{pgfscope}%
\begin{pgfscope}%
\pgfpathrectangle{\pgfqpoint{0.100000in}{0.915754in}}{\pgfqpoint{4.650000in}{3.020000in}}%
\pgfusepath{clip}%
\pgfsetbuttcap%
\pgfsetroundjoin%
\definecolor{currentfill}{rgb}{0.333333,0.658824,0.407843}%
\pgfsetfillcolor{currentfill}%
\pgfsetfillopacity{0.200000}%
\pgfsetlinewidth{1.003750pt}%
\definecolor{currentstroke}{rgb}{0.333333,0.658824,0.407843}%
\pgfsetstrokecolor{currentstroke}%
\pgfsetstrokeopacity{0.200000}%
\pgfsetdash{}{0pt}%
\pgfpathmoveto{\pgfqpoint{0.006571in}{3.309852in}}%
\pgfpathlineto{\pgfqpoint{0.006571in}{2.981476in}}%
\pgfpathlineto{\pgfqpoint{0.191726in}{2.997782in}}%
\pgfpathlineto{\pgfqpoint{0.375005in}{2.790730in}}%
\pgfpathlineto{\pgfqpoint{0.555898in}{2.665327in}}%
\pgfpathlineto{\pgfqpoint{0.732045in}{2.645966in}}%
\pgfpathlineto{\pgfqpoint{0.919014in}{2.477171in}}%
\pgfpathlineto{\pgfqpoint{1.105442in}{2.440022in}}%
\pgfpathlineto{\pgfqpoint{1.295550in}{2.467091in}}%
\pgfpathlineto{\pgfqpoint{1.483373in}{2.472604in}}%
\pgfpathlineto{\pgfqpoint{1.673437in}{2.473456in}}%
\pgfpathlineto{\pgfqpoint{1.856948in}{2.197927in}}%
\pgfpathlineto{\pgfqpoint{2.031981in}{2.281294in}}%
\pgfpathlineto{\pgfqpoint{2.239791in}{2.166154in}}%
\pgfpathlineto{\pgfqpoint{2.413910in}{2.358337in}}%
\pgfpathlineto{\pgfqpoint{2.601416in}{2.309839in}}%
\pgfpathlineto{\pgfqpoint{2.795668in}{2.406885in}}%
\pgfpathlineto{\pgfqpoint{2.983274in}{2.396537in}}%
\pgfpathlineto{\pgfqpoint{3.152252in}{2.094079in}}%
\pgfpathlineto{\pgfqpoint{3.341217in}{2.415363in}}%
\pgfpathlineto{\pgfqpoint{3.528941in}{2.475141in}}%
\pgfpathlineto{\pgfqpoint{3.722401in}{2.209268in}}%
\pgfpathlineto{\pgfqpoint{3.903859in}{2.210830in}}%
\pgfpathlineto{\pgfqpoint{4.105288in}{2.215037in}}%
\pgfpathlineto{\pgfqpoint{4.265376in}{2.218920in}}%
\pgfpathlineto{\pgfqpoint{4.443106in}{2.314660in}}%
\pgfpathlineto{\pgfqpoint{4.644395in}{2.420906in}}%
\pgfpathlineto{\pgfqpoint{4.833283in}{2.144238in}}%
\pgfpathlineto{\pgfqpoint{5.023197in}{2.278407in}}%
\pgfpathlineto{\pgfqpoint{5.200241in}{2.336296in}}%
\pgfpathlineto{\pgfqpoint{5.395308in}{2.382063in}}%
\pgfpathlineto{\pgfqpoint{5.395308in}{2.710896in}}%
\pgfpathlineto{\pgfqpoint{5.395308in}{2.710896in}}%
\pgfpathlineto{\pgfqpoint{5.200241in}{2.614818in}}%
\pgfpathlineto{\pgfqpoint{5.023197in}{2.632308in}}%
\pgfpathlineto{\pgfqpoint{4.833283in}{2.444771in}}%
\pgfpathlineto{\pgfqpoint{4.644395in}{2.643084in}}%
\pgfpathlineto{\pgfqpoint{4.443106in}{2.683496in}}%
\pgfpathlineto{\pgfqpoint{4.265376in}{2.625229in}}%
\pgfpathlineto{\pgfqpoint{4.105288in}{2.590418in}}%
\pgfpathlineto{\pgfqpoint{3.903859in}{2.480499in}}%
\pgfpathlineto{\pgfqpoint{3.722401in}{2.513080in}}%
\pgfpathlineto{\pgfqpoint{3.528941in}{2.811685in}}%
\pgfpathlineto{\pgfqpoint{3.341217in}{2.705547in}}%
\pgfpathlineto{\pgfqpoint{3.152252in}{2.543045in}}%
\pgfpathlineto{\pgfqpoint{2.983274in}{2.774698in}}%
\pgfpathlineto{\pgfqpoint{2.795668in}{2.712428in}}%
\pgfpathlineto{\pgfqpoint{2.601416in}{2.670480in}}%
\pgfpathlineto{\pgfqpoint{2.413910in}{2.600628in}}%
\pgfpathlineto{\pgfqpoint{2.239791in}{2.618806in}}%
\pgfpathlineto{\pgfqpoint{2.031981in}{2.683222in}}%
\pgfpathlineto{\pgfqpoint{1.856948in}{2.798667in}}%
\pgfpathlineto{\pgfqpoint{1.673437in}{2.823277in}}%
\pgfpathlineto{\pgfqpoint{1.483373in}{2.835177in}}%
\pgfpathlineto{\pgfqpoint{1.295550in}{2.807298in}}%
\pgfpathlineto{\pgfqpoint{1.105442in}{2.886795in}}%
\pgfpathlineto{\pgfqpoint{0.919014in}{2.903385in}}%
\pgfpathlineto{\pgfqpoint{0.732045in}{3.161673in}}%
\pgfpathlineto{\pgfqpoint{0.555898in}{2.973607in}}%
\pgfpathlineto{\pgfqpoint{0.375005in}{3.143542in}}%
\pgfpathlineto{\pgfqpoint{0.191726in}{3.355724in}}%
\pgfpathlineto{\pgfqpoint{0.006571in}{3.309852in}}%
\pgfpathclose%
\pgfusepath{stroke,fill}%
\end{pgfscope}%
\begin{pgfscope}%
\pgfpathrectangle{\pgfqpoint{0.100000in}{0.915754in}}{\pgfqpoint{4.650000in}{3.020000in}}%
\pgfusepath{clip}%
\pgfsetbuttcap%
\pgfsetroundjoin%
\definecolor{currentfill}{rgb}{0.768627,0.305882,0.321569}%
\pgfsetfillcolor{currentfill}%
\pgfsetfillopacity{0.200000}%
\pgfsetlinewidth{1.003750pt}%
\definecolor{currentstroke}{rgb}{0.768627,0.305882,0.321569}%
\pgfsetstrokecolor{currentstroke}%
\pgfsetstrokeopacity{0.200000}%
\pgfsetdash{}{0pt}%
\pgfpathmoveto{\pgfqpoint{-0.066071in}{3.426090in}}%
\pgfpathlineto{\pgfqpoint{-0.066071in}{3.131282in}}%
\pgfpathlineto{\pgfqpoint{0.105727in}{2.914417in}}%
\pgfpathlineto{\pgfqpoint{0.277525in}{3.023832in}}%
\pgfpathlineto{\pgfqpoint{0.449323in}{2.828997in}}%
\pgfpathlineto{\pgfqpoint{0.621121in}{2.630710in}}%
\pgfpathlineto{\pgfqpoint{0.792919in}{2.644318in}}%
\pgfpathlineto{\pgfqpoint{0.964717in}{2.495613in}}%
\pgfpathlineto{\pgfqpoint{1.136515in}{2.687121in}}%
\pgfpathlineto{\pgfqpoint{1.308313in}{2.587852in}}%
\pgfpathlineto{\pgfqpoint{1.480111in}{2.534819in}}%
\pgfpathlineto{\pgfqpoint{1.651909in}{2.562126in}}%
\pgfpathlineto{\pgfqpoint{1.823707in}{2.683220in}}%
\pgfpathlineto{\pgfqpoint{1.995505in}{2.631646in}}%
\pgfpathlineto{\pgfqpoint{2.167303in}{2.552572in}}%
\pgfpathlineto{\pgfqpoint{2.339101in}{2.604886in}}%
\pgfpathlineto{\pgfqpoint{2.510899in}{2.495308in}}%
\pgfpathlineto{\pgfqpoint{2.682697in}{2.348963in}}%
\pgfpathlineto{\pgfqpoint{2.854495in}{2.456847in}}%
\pgfpathlineto{\pgfqpoint{3.026293in}{2.408429in}}%
\pgfpathlineto{\pgfqpoint{3.198091in}{2.374931in}}%
\pgfpathlineto{\pgfqpoint{3.369889in}{2.431059in}}%
\pgfpathlineto{\pgfqpoint{3.541687in}{2.339329in}}%
\pgfpathlineto{\pgfqpoint{3.713485in}{2.459013in}}%
\pgfpathlineto{\pgfqpoint{3.885283in}{2.387396in}}%
\pgfpathlineto{\pgfqpoint{4.057081in}{2.409360in}}%
\pgfpathlineto{\pgfqpoint{4.228879in}{2.386493in}}%
\pgfpathlineto{\pgfqpoint{4.400677in}{2.370523in}}%
\pgfpathlineto{\pgfqpoint{4.572475in}{2.425861in}}%
\pgfpathlineto{\pgfqpoint{4.744273in}{2.170576in}}%
\pgfpathlineto{\pgfqpoint{4.916071in}{2.351362in}}%
\pgfpathlineto{\pgfqpoint{4.916071in}{2.647400in}}%
\pgfpathlineto{\pgfqpoint{4.916071in}{2.647400in}}%
\pgfpathlineto{\pgfqpoint{4.744273in}{2.581410in}}%
\pgfpathlineto{\pgfqpoint{4.572475in}{2.726159in}}%
\pgfpathlineto{\pgfqpoint{4.400677in}{2.675507in}}%
\pgfpathlineto{\pgfqpoint{4.228879in}{2.672398in}}%
\pgfpathlineto{\pgfqpoint{4.057081in}{2.691925in}}%
\pgfpathlineto{\pgfqpoint{3.885283in}{2.700372in}}%
\pgfpathlineto{\pgfqpoint{3.713485in}{2.748130in}}%
\pgfpathlineto{\pgfqpoint{3.541687in}{2.665988in}}%
\pgfpathlineto{\pgfqpoint{3.369889in}{2.705318in}}%
\pgfpathlineto{\pgfqpoint{3.198091in}{2.628613in}}%
\pgfpathlineto{\pgfqpoint{3.026293in}{2.762288in}}%
\pgfpathlineto{\pgfqpoint{2.854495in}{2.695846in}}%
\pgfpathlineto{\pgfqpoint{2.682697in}{2.694476in}}%
\pgfpathlineto{\pgfqpoint{2.510899in}{2.758575in}}%
\pgfpathlineto{\pgfqpoint{2.339101in}{2.875573in}}%
\pgfpathlineto{\pgfqpoint{2.167303in}{2.844777in}}%
\pgfpathlineto{\pgfqpoint{1.995505in}{2.937858in}}%
\pgfpathlineto{\pgfqpoint{1.823707in}{2.976900in}}%
\pgfpathlineto{\pgfqpoint{1.651909in}{2.820777in}}%
\pgfpathlineto{\pgfqpoint{1.480111in}{2.846893in}}%
\pgfpathlineto{\pgfqpoint{1.308313in}{2.927622in}}%
\pgfpathlineto{\pgfqpoint{1.136515in}{2.990489in}}%
\pgfpathlineto{\pgfqpoint{0.964717in}{2.893490in}}%
\pgfpathlineto{\pgfqpoint{0.792919in}{3.072179in}}%
\pgfpathlineto{\pgfqpoint{0.621121in}{2.983546in}}%
\pgfpathlineto{\pgfqpoint{0.449323in}{3.157061in}}%
\pgfpathlineto{\pgfqpoint{0.277525in}{3.333995in}}%
\pgfpathlineto{\pgfqpoint{0.105727in}{3.257742in}}%
\pgfpathlineto{\pgfqpoint{-0.066071in}{3.426090in}}%
\pgfpathclose%
\pgfusepath{stroke,fill}%
\end{pgfscope}%
\begin{pgfscope}%
\pgfpathrectangle{\pgfqpoint{0.100000in}{0.915754in}}{\pgfqpoint{4.650000in}{3.020000in}}%
\pgfusepath{clip}%
\pgfsetroundcap%
\pgfsetroundjoin%
\pgfsetlinewidth{1.505625pt}%
\definecolor{currentstroke}{rgb}{0.298039,0.447059,0.690196}%
\pgfsetstrokecolor{currentstroke}%
\pgfsetdash{}{0pt}%
\pgfpathmoveto{\pgfqpoint{0.086111in}{2.447527in}}%
\pgfpathlineto{\pgfqpoint{0.152068in}{2.361683in}}%
\pgfpathlineto{\pgfqpoint{0.329072in}{2.308640in}}%
\pgfpathlineto{\pgfqpoint{0.499467in}{1.840037in}}%
\pgfpathlineto{\pgfqpoint{0.673754in}{2.045905in}}%
\pgfpathlineto{\pgfqpoint{0.846819in}{1.691915in}}%
\pgfpathlineto{\pgfqpoint{1.022331in}{1.959353in}}%
\pgfpathlineto{\pgfqpoint{1.202483in}{1.868456in}}%
\pgfpathlineto{\pgfqpoint{1.377281in}{2.039786in}}%
\pgfpathlineto{\pgfqpoint{1.550563in}{1.914550in}}%
\pgfpathlineto{\pgfqpoint{1.726299in}{2.072508in}}%
\pgfpathlineto{\pgfqpoint{1.901176in}{2.046542in}}%
\pgfpathlineto{\pgfqpoint{2.073976in}{2.243933in}}%
\pgfpathlineto{\pgfqpoint{2.253660in}{1.951417in}}%
\pgfpathlineto{\pgfqpoint{2.425081in}{2.027348in}}%
\pgfpathlineto{\pgfqpoint{2.598336in}{2.095502in}}%
\pgfpathlineto{\pgfqpoint{2.777216in}{2.012368in}}%
\pgfpathlineto{\pgfqpoint{2.950069in}{2.146722in}}%
\pgfpathlineto{\pgfqpoint{3.127870in}{2.026527in}}%
\pgfpathlineto{\pgfqpoint{3.303124in}{2.164756in}}%
\pgfpathlineto{\pgfqpoint{3.472183in}{2.007331in}}%
\pgfpathlineto{\pgfqpoint{3.650796in}{2.161434in}}%
\pgfpathlineto{\pgfqpoint{3.832813in}{2.189432in}}%
\pgfpathlineto{\pgfqpoint{4.002021in}{2.059500in}}%
\pgfpathlineto{\pgfqpoint{4.175909in}{2.091765in}}%
\pgfpathlineto{\pgfqpoint{4.353019in}{2.113704in}}%
\pgfpathlineto{\pgfqpoint{4.531220in}{2.194754in}}%
\pgfpathlineto{\pgfqpoint{4.705425in}{2.140315in}}%
\pgfpathlineto{\pgfqpoint{4.763889in}{2.141844in}}%
\pgfusepath{stroke}%
\end{pgfscope}%
\begin{pgfscope}%
\pgfpathrectangle{\pgfqpoint{0.100000in}{0.915754in}}{\pgfqpoint{4.650000in}{3.020000in}}%
\pgfusepath{clip}%
\pgfsetroundcap%
\pgfsetroundjoin%
\pgfsetlinewidth{1.505625pt}%
\definecolor{currentstroke}{rgb}{0.866667,0.517647,0.321569}%
\pgfsetstrokecolor{currentstroke}%
\pgfsetdash{}{0pt}%
\pgfpathmoveto{\pgfqpoint{0.086111in}{3.227151in}}%
\pgfpathlineto{\pgfqpoint{0.154184in}{3.160656in}}%
\pgfpathlineto{\pgfqpoint{0.322702in}{3.098437in}}%
\pgfpathlineto{\pgfqpoint{0.501250in}{2.978404in}}%
\pgfpathlineto{\pgfqpoint{0.677433in}{2.854510in}}%
\pgfpathlineto{\pgfqpoint{0.853174in}{2.723624in}}%
\pgfpathlineto{\pgfqpoint{1.022286in}{2.651068in}}%
\pgfpathlineto{\pgfqpoint{1.200355in}{2.538983in}}%
\pgfpathlineto{\pgfqpoint{1.372670in}{2.534383in}}%
\pgfpathlineto{\pgfqpoint{1.547231in}{2.688981in}}%
\pgfpathlineto{\pgfqpoint{1.721311in}{2.645557in}}%
\pgfpathlineto{\pgfqpoint{1.903084in}{2.830734in}}%
\pgfpathlineto{\pgfqpoint{2.073757in}{2.768625in}}%
\pgfpathlineto{\pgfqpoint{2.251539in}{2.714859in}}%
\pgfpathlineto{\pgfqpoint{2.426400in}{2.724203in}}%
\pgfpathlineto{\pgfqpoint{2.603936in}{2.834726in}}%
\pgfpathlineto{\pgfqpoint{2.773922in}{2.781786in}}%
\pgfpathlineto{\pgfqpoint{2.950697in}{2.716790in}}%
\pgfpathlineto{\pgfqpoint{3.132595in}{2.662114in}}%
\pgfpathlineto{\pgfqpoint{3.299308in}{2.672564in}}%
\pgfpathlineto{\pgfqpoint{3.478575in}{2.710945in}}%
\pgfpathlineto{\pgfqpoint{3.653441in}{2.706988in}}%
\pgfpathlineto{\pgfqpoint{3.828244in}{2.681380in}}%
\pgfpathlineto{\pgfqpoint{4.005932in}{2.492240in}}%
\pgfpathlineto{\pgfqpoint{4.176743in}{2.761490in}}%
\pgfpathlineto{\pgfqpoint{4.356597in}{2.755569in}}%
\pgfpathlineto{\pgfqpoint{4.530369in}{2.655701in}}%
\pgfpathlineto{\pgfqpoint{4.710803in}{2.743725in}}%
\pgfpathlineto{\pgfqpoint{4.763889in}{2.740661in}}%
\pgfusepath{stroke}%
\end{pgfscope}%
\begin{pgfscope}%
\pgfpathrectangle{\pgfqpoint{0.100000in}{0.915754in}}{\pgfqpoint{4.650000in}{3.020000in}}%
\pgfusepath{clip}%
\pgfsetroundcap%
\pgfsetroundjoin%
\pgfsetlinewidth{1.505625pt}%
\definecolor{currentstroke}{rgb}{0.333333,0.658824,0.407843}%
\pgfsetstrokecolor{currentstroke}%
\pgfsetdash{}{0pt}%
\pgfpathmoveto{\pgfqpoint{0.086111in}{3.176504in}}%
\pgfpathlineto{\pgfqpoint{0.191726in}{3.196502in}}%
\pgfpathlineto{\pgfqpoint{0.375005in}{2.973931in}}%
\pgfpathlineto{\pgfqpoint{0.555898in}{2.831018in}}%
\pgfpathlineto{\pgfqpoint{0.732045in}{2.931346in}}%
\pgfpathlineto{\pgfqpoint{0.919014in}{2.715821in}}%
\pgfpathlineto{\pgfqpoint{1.105442in}{2.689850in}}%
\pgfpathlineto{\pgfqpoint{1.295550in}{2.658871in}}%
\pgfpathlineto{\pgfqpoint{1.483373in}{2.660101in}}%
\pgfpathlineto{\pgfqpoint{1.673437in}{2.663258in}}%
\pgfpathlineto{\pgfqpoint{1.856948in}{2.525402in}}%
\pgfpathlineto{\pgfqpoint{2.031981in}{2.512118in}}%
\pgfpathlineto{\pgfqpoint{2.239791in}{2.419003in}}%
\pgfpathlineto{\pgfqpoint{2.413910in}{2.484468in}}%
\pgfpathlineto{\pgfqpoint{2.601416in}{2.502549in}}%
\pgfpathlineto{\pgfqpoint{2.795668in}{2.568647in}}%
\pgfpathlineto{\pgfqpoint{2.983274in}{2.591409in}}%
\pgfpathlineto{\pgfqpoint{3.152252in}{2.328619in}}%
\pgfpathlineto{\pgfqpoint{3.341217in}{2.568566in}}%
\pgfpathlineto{\pgfqpoint{3.528941in}{2.646950in}}%
\pgfpathlineto{\pgfqpoint{3.722401in}{2.370476in}}%
\pgfpathlineto{\pgfqpoint{3.903859in}{2.365203in}}%
\pgfpathlineto{\pgfqpoint{4.105288in}{2.418005in}}%
\pgfpathlineto{\pgfqpoint{4.265376in}{2.426014in}}%
\pgfpathlineto{\pgfqpoint{4.443106in}{2.513997in}}%
\pgfpathlineto{\pgfqpoint{4.644395in}{2.540613in}}%
\pgfpathlineto{\pgfqpoint{4.763889in}{2.398456in}}%
\pgfusepath{stroke}%
\end{pgfscope}%
\begin{pgfscope}%
\pgfpathrectangle{\pgfqpoint{0.100000in}{0.915754in}}{\pgfqpoint{4.650000in}{3.020000in}}%
\pgfusepath{clip}%
\pgfsetroundcap%
\pgfsetroundjoin%
\pgfsetlinewidth{1.505625pt}%
\definecolor{currentstroke}{rgb}{0.768627,0.305882,0.321569}%
\pgfsetstrokecolor{currentstroke}%
\pgfsetdash{}{0pt}%
\pgfpathmoveto{\pgfqpoint{0.086111in}{3.114023in}}%
\pgfpathlineto{\pgfqpoint{0.105727in}{3.091196in}}%
\pgfpathlineto{\pgfqpoint{0.277525in}{3.202111in}}%
\pgfpathlineto{\pgfqpoint{0.449323in}{3.012835in}}%
\pgfpathlineto{\pgfqpoint{0.621121in}{2.821599in}}%
\pgfpathlineto{\pgfqpoint{0.792919in}{2.891194in}}%
\pgfpathlineto{\pgfqpoint{0.964717in}{2.721064in}}%
\pgfpathlineto{\pgfqpoint{1.136515in}{2.848649in}}%
\pgfpathlineto{\pgfqpoint{1.308313in}{2.762805in}}%
\pgfpathlineto{\pgfqpoint{1.480111in}{2.710982in}}%
\pgfpathlineto{\pgfqpoint{1.651909in}{2.698035in}}%
\pgfpathlineto{\pgfqpoint{1.823707in}{2.847618in}}%
\pgfpathlineto{\pgfqpoint{1.995505in}{2.805541in}}%
\pgfpathlineto{\pgfqpoint{2.167303in}{2.705454in}}%
\pgfpathlineto{\pgfqpoint{2.339101in}{2.752357in}}%
\pgfpathlineto{\pgfqpoint{2.510899in}{2.638139in}}%
\pgfpathlineto{\pgfqpoint{2.682697in}{2.535167in}}%
\pgfpathlineto{\pgfqpoint{2.854495in}{2.584785in}}%
\pgfpathlineto{\pgfqpoint{3.026293in}{2.592460in}}%
\pgfpathlineto{\pgfqpoint{3.198091in}{2.519769in}}%
\pgfpathlineto{\pgfqpoint{3.369889in}{2.586665in}}%
\pgfpathlineto{\pgfqpoint{3.541687in}{2.518938in}}%
\pgfpathlineto{\pgfqpoint{3.713485in}{2.619256in}}%
\pgfpathlineto{\pgfqpoint{3.885283in}{2.560980in}}%
\pgfpathlineto{\pgfqpoint{4.057081in}{2.557873in}}%
\pgfpathlineto{\pgfqpoint{4.228879in}{2.537689in}}%
\pgfpathlineto{\pgfqpoint{4.400677in}{2.538031in}}%
\pgfpathlineto{\pgfqpoint{4.572475in}{2.582175in}}%
\pgfpathlineto{\pgfqpoint{4.744273in}{2.402249in}}%
\pgfpathlineto{\pgfqpoint{4.763889in}{2.414461in}}%
\pgfusepath{stroke}%
\end{pgfscope}%
\begin{pgfscope}%
\pgfsetrectcap%
\pgfsetmiterjoin%
\pgfsetlinewidth{1.254687pt}%
\definecolor{currentstroke}{rgb}{0.800000,0.800000,0.800000}%
\pgfsetstrokecolor{currentstroke}%
\pgfsetdash{}{0pt}%
\pgfpathmoveto{\pgfqpoint{0.100000in}{0.915754in}}%
\pgfpathlineto{\pgfqpoint{0.100000in}{3.935754in}}%
\pgfusepath{stroke}%
\end{pgfscope}%
\begin{pgfscope}%
\pgfsetrectcap%
\pgfsetmiterjoin%
\pgfsetlinewidth{1.254687pt}%
\definecolor{currentstroke}{rgb}{0.800000,0.800000,0.800000}%
\pgfsetstrokecolor{currentstroke}%
\pgfsetdash{}{0pt}%
\pgfpathmoveto{\pgfqpoint{4.750000in}{0.915754in}}%
\pgfpathlineto{\pgfqpoint{4.750000in}{3.935754in}}%
\pgfusepath{stroke}%
\end{pgfscope}%
\begin{pgfscope}%
\pgfsetrectcap%
\pgfsetmiterjoin%
\pgfsetlinewidth{1.254687pt}%
\definecolor{currentstroke}{rgb}{0.800000,0.800000,0.800000}%
\pgfsetstrokecolor{currentstroke}%
\pgfsetdash{}{0pt}%
\pgfpathmoveto{\pgfqpoint{0.100000in}{0.915754in}}%
\pgfpathlineto{\pgfqpoint{4.750000in}{0.915754in}}%
\pgfusepath{stroke}%
\end{pgfscope}%
\begin{pgfscope}%
\pgfsetrectcap%
\pgfsetmiterjoin%
\pgfsetlinewidth{1.254687pt}%
\definecolor{currentstroke}{rgb}{0.800000,0.800000,0.800000}%
\pgfsetstrokecolor{currentstroke}%
\pgfsetdash{}{0pt}%
\pgfpathmoveto{\pgfqpoint{0.100000in}{3.935754in}}%
\pgfpathlineto{\pgfqpoint{4.750000in}{3.935754in}}%
\pgfusepath{stroke}%
\end{pgfscope}%
\end{pgfpicture}%
\makeatother%
\endgroup%

%% file: fig/continuous_sparse_moon_res.pgf
%% Creator: Matplotlib, PGF backend
%%
%% To include the figure in your LaTeX document, write
%%   \input{<filename>.pgf}
%%
%% Make sure the required packages are loaded in your preamble
%%   \usepackage{pgf}
%%
%% Figures using additional raster images can only be included by \input if
%% they are in the same directory as the main LaTeX file. For loading figures
%% from other directories you can use the `import` package
%% and then include the figures with
%%   \import{<path to file>}{<filename>.pgf}
%%
%% Matplotlib used the following preamble
%%
\begingroup%
\makeatletter%
\begin{pgfpicture}%
\pgfpathrectangle{\pgfpointorigin}{\pgfqpoint{4.850000in}{4.035754in}}%
\pgfusepath{use as bounding box, clip}%
\begin{pgfscope}%
\pgfsetbuttcap%
\pgfsetmiterjoin%
\definecolor{currentfill}{rgb}{1.000000,1.000000,1.000000}%
\pgfsetfillcolor{currentfill}%
\pgfsetlinewidth{0.000000pt}%
\definecolor{currentstroke}{rgb}{1.000000,1.000000,1.000000}%
\pgfsetstrokecolor{currentstroke}%
\pgfsetdash{}{0pt}%
\pgfpathmoveto{\pgfqpoint{0.000000in}{0.000000in}}%
\pgfpathlineto{\pgfqpoint{4.850000in}{0.000000in}}%
\pgfpathlineto{\pgfqpoint{4.850000in}{4.035754in}}%
\pgfpathlineto{\pgfqpoint{0.000000in}{4.035754in}}%
\pgfpathclose%
\pgfusepath{fill}%
\end{pgfscope}%
\begin{pgfscope}%
\pgfsetbuttcap%
\pgfsetmiterjoin%
\definecolor{currentfill}{rgb}{1.000000,1.000000,1.000000}%
\pgfsetfillcolor{currentfill}%
\pgfsetlinewidth{0.000000pt}%
\definecolor{currentstroke}{rgb}{0.000000,0.000000,0.000000}%
\pgfsetstrokecolor{currentstroke}%
\pgfsetstrokeopacity{0.000000}%
\pgfsetdash{}{0pt}%
\pgfpathmoveto{\pgfqpoint{0.100000in}{0.915754in}}%
\pgfpathlineto{\pgfqpoint{4.750000in}{0.915754in}}%
\pgfpathlineto{\pgfqpoint{4.750000in}{3.935754in}}%
\pgfpathlineto{\pgfqpoint{0.100000in}{3.935754in}}%
\pgfpathclose%
\pgfusepath{fill}%
\end{pgfscope}%
\begin{pgfscope}%
\pgfpathrectangle{\pgfqpoint{0.100000in}{0.915754in}}{\pgfqpoint{4.650000in}{3.020000in}}%
\pgfusepath{clip}%
\pgfsetroundcap%
\pgfsetroundjoin%
\pgfsetlinewidth{1.003750pt}%
\definecolor{currentstroke}{rgb}{0.800000,0.800000,0.800000}%
\pgfsetstrokecolor{currentstroke}%
\pgfsetdash{}{0pt}%
\pgfpathmoveto{\pgfqpoint{0.349107in}{0.915754in}}%
\pgfpathlineto{\pgfqpoint{0.349107in}{3.935754in}}%
\pgfusepath{stroke}%
\end{pgfscope}%
\begin{pgfscope}%
\definecolor{textcolor}{rgb}{0.150000,0.150000,0.150000}%
\pgfsetstrokecolor{textcolor}%
\pgfsetfillcolor{textcolor}%
\pgftext[x=0.349107in,y=0.783810in,,top]{\color{textcolor}\rmfamily\fontsize{23.100000}{27.720000}\selectfont 10}%
\end{pgfscope}%
\begin{pgfscope}%
\pgfpathrectangle{\pgfqpoint{0.100000in}{0.915754in}}{\pgfqpoint{4.650000in}{3.020000in}}%
\pgfusepath{clip}%
\pgfsetroundcap%
\pgfsetroundjoin%
\pgfsetlinewidth{1.003750pt}%
\definecolor{currentstroke}{rgb}{0.800000,0.800000,0.800000}%
\pgfsetstrokecolor{currentstroke}%
\pgfsetdash{}{0pt}%
\pgfpathmoveto{\pgfqpoint{1.179464in}{0.915754in}}%
\pgfpathlineto{\pgfqpoint{1.179464in}{3.935754in}}%
\pgfusepath{stroke}%
\end{pgfscope}%
\begin{pgfscope}%
\definecolor{textcolor}{rgb}{0.150000,0.150000,0.150000}%
\pgfsetstrokecolor{textcolor}%
\pgfsetfillcolor{textcolor}%
\pgftext[x=1.179464in,y=0.783810in,,top]{\color{textcolor}\rmfamily\fontsize{23.100000}{27.720000}\selectfont 20}%
\end{pgfscope}%
\begin{pgfscope}%
\pgfpathrectangle{\pgfqpoint{0.100000in}{0.915754in}}{\pgfqpoint{4.650000in}{3.020000in}}%
\pgfusepath{clip}%
\pgfsetroundcap%
\pgfsetroundjoin%
\pgfsetlinewidth{1.003750pt}%
\definecolor{currentstroke}{rgb}{0.800000,0.800000,0.800000}%
\pgfsetstrokecolor{currentstroke}%
\pgfsetdash{}{0pt}%
\pgfpathmoveto{\pgfqpoint{2.009821in}{0.915754in}}%
\pgfpathlineto{\pgfqpoint{2.009821in}{3.935754in}}%
\pgfusepath{stroke}%
\end{pgfscope}%
\begin{pgfscope}%
\definecolor{textcolor}{rgb}{0.150000,0.150000,0.150000}%
\pgfsetstrokecolor{textcolor}%
\pgfsetfillcolor{textcolor}%
\pgftext[x=2.009821in,y=0.783810in,,top]{\color{textcolor}\rmfamily\fontsize{23.100000}{27.720000}\selectfont 30}%
\end{pgfscope}%
\begin{pgfscope}%
\pgfpathrectangle{\pgfqpoint{0.100000in}{0.915754in}}{\pgfqpoint{4.650000in}{3.020000in}}%
\pgfusepath{clip}%
\pgfsetroundcap%
\pgfsetroundjoin%
\pgfsetlinewidth{1.003750pt}%
\definecolor{currentstroke}{rgb}{0.800000,0.800000,0.800000}%
\pgfsetstrokecolor{currentstroke}%
\pgfsetdash{}{0pt}%
\pgfpathmoveto{\pgfqpoint{2.840179in}{0.915754in}}%
\pgfpathlineto{\pgfqpoint{2.840179in}{3.935754in}}%
\pgfusepath{stroke}%
\end{pgfscope}%
\begin{pgfscope}%
\definecolor{textcolor}{rgb}{0.150000,0.150000,0.150000}%
\pgfsetstrokecolor{textcolor}%
\pgfsetfillcolor{textcolor}%
\pgftext[x=2.840179in,y=0.783810in,,top]{\color{textcolor}\rmfamily\fontsize{23.100000}{27.720000}\selectfont 40}%
\end{pgfscope}%
\begin{pgfscope}%
\pgfpathrectangle{\pgfqpoint{0.100000in}{0.915754in}}{\pgfqpoint{4.650000in}{3.020000in}}%
\pgfusepath{clip}%
\pgfsetroundcap%
\pgfsetroundjoin%
\pgfsetlinewidth{1.003750pt}%
\definecolor{currentstroke}{rgb}{0.800000,0.800000,0.800000}%
\pgfsetstrokecolor{currentstroke}%
\pgfsetdash{}{0pt}%
\pgfpathmoveto{\pgfqpoint{3.670536in}{0.915754in}}%
\pgfpathlineto{\pgfqpoint{3.670536in}{3.935754in}}%
\pgfusepath{stroke}%
\end{pgfscope}%
\begin{pgfscope}%
\definecolor{textcolor}{rgb}{0.150000,0.150000,0.150000}%
\pgfsetstrokecolor{textcolor}%
\pgfsetfillcolor{textcolor}%
\pgftext[x=3.670536in,y=0.783810in,,top]{\color{textcolor}\rmfamily\fontsize{23.100000}{27.720000}\selectfont 50}%
\end{pgfscope}%
\begin{pgfscope}%
\pgfpathrectangle{\pgfqpoint{0.100000in}{0.915754in}}{\pgfqpoint{4.650000in}{3.020000in}}%
\pgfusepath{clip}%
\pgfsetroundcap%
\pgfsetroundjoin%
\pgfsetlinewidth{1.003750pt}%
\definecolor{currentstroke}{rgb}{0.800000,0.800000,0.800000}%
\pgfsetstrokecolor{currentstroke}%
\pgfsetdash{}{0pt}%
\pgfpathmoveto{\pgfqpoint{4.500893in}{0.915754in}}%
\pgfpathlineto{\pgfqpoint{4.500893in}{3.935754in}}%
\pgfusepath{stroke}%
\end{pgfscope}%
\begin{pgfscope}%
\definecolor{textcolor}{rgb}{0.150000,0.150000,0.150000}%
\pgfsetstrokecolor{textcolor}%
\pgfsetfillcolor{textcolor}%
\pgftext[x=4.500893in,y=0.783810in,,top]{\color{textcolor}\rmfamily\fontsize{23.100000}{27.720000}\selectfont 60}%
\end{pgfscope}%
\begin{pgfscope}%
\definecolor{textcolor}{rgb}{0.150000,0.150000,0.150000}%
\pgfsetstrokecolor{textcolor}%
\pgfsetfillcolor{textcolor}%
\pgftext[x=2.425000in,y=0.407183in,,top]{\color{textcolor}\rmfamily\fontsize{25.200000}{30.240000}\selectfont runtime in seconds}%
\end{pgfscope}%
\begin{pgfscope}%
\pgfpathrectangle{\pgfqpoint{0.100000in}{0.915754in}}{\pgfqpoint{4.650000in}{3.020000in}}%
\pgfusepath{clip}%
\pgfsetroundcap%
\pgfsetroundjoin%
\pgfsetlinewidth{1.003750pt}%
\definecolor{currentstroke}{rgb}{0.800000,0.800000,0.800000}%
\pgfsetstrokecolor{currentstroke}%
\pgfsetdash{}{0pt}%
\pgfpathmoveto{\pgfqpoint{0.100000in}{1.328214in}}%
\pgfpathlineto{\pgfqpoint{4.750000in}{1.328214in}}%
\pgfusepath{stroke}%
\end{pgfscope}%
\begin{pgfscope}%
\pgfpathrectangle{\pgfqpoint{0.100000in}{0.915754in}}{\pgfqpoint{4.650000in}{3.020000in}}%
\pgfusepath{clip}%
\pgfsetroundcap%
\pgfsetroundjoin%
\pgfsetlinewidth{1.003750pt}%
\definecolor{currentstroke}{rgb}{0.800000,0.800000,0.800000}%
\pgfsetstrokecolor{currentstroke}%
\pgfsetdash{}{0pt}%
\pgfpathmoveto{\pgfqpoint{0.100000in}{2.698375in}}%
\pgfpathlineto{\pgfqpoint{4.750000in}{2.698375in}}%
\pgfusepath{stroke}%
\end{pgfscope}%
\begin{pgfscope}%
\pgfpathrectangle{\pgfqpoint{0.100000in}{0.915754in}}{\pgfqpoint{4.650000in}{3.020000in}}%
\pgfusepath{clip}%
\pgfsetbuttcap%
\pgfsetroundjoin%
\definecolor{currentfill}{rgb}{0.298039,0.447059,0.690196}%
\pgfsetfillcolor{currentfill}%
\pgfsetfillopacity{0.200000}%
\pgfsetlinewidth{1.003750pt}%
\definecolor{currentstroke}{rgb}{0.298039,0.447059,0.690196}%
\pgfsetstrokecolor{currentstroke}%
\pgfsetstrokeopacity{0.200000}%
\pgfsetdash{}{0pt}%
\pgfpathmoveto{\pgfqpoint{-0.025493in}{2.084141in}}%
\pgfpathlineto{\pgfqpoint{-0.025493in}{1.757663in}}%
\pgfpathlineto{\pgfqpoint{0.150965in}{1.557861in}}%
\pgfpathlineto{\pgfqpoint{0.326351in}{1.156125in}}%
\pgfpathlineto{\pgfqpoint{0.502927in}{1.158114in}}%
\pgfpathlineto{\pgfqpoint{0.681370in}{1.242708in}}%
\pgfpathlineto{\pgfqpoint{0.858342in}{1.202684in}}%
\pgfpathlineto{\pgfqpoint{1.033456in}{1.134421in}}%
\pgfpathlineto{\pgfqpoint{1.210785in}{1.534985in}}%
\pgfpathlineto{\pgfqpoint{1.386047in}{1.519923in}}%
\pgfpathlineto{\pgfqpoint{1.562614in}{1.475881in}}%
\pgfpathlineto{\pgfqpoint{1.743052in}{1.547796in}}%
\pgfpathlineto{\pgfqpoint{1.919028in}{1.552420in}}%
\pgfpathlineto{\pgfqpoint{2.095804in}{1.517242in}}%
\pgfpathlineto{\pgfqpoint{2.275178in}{1.429938in}}%
\pgfpathlineto{\pgfqpoint{2.453482in}{1.428992in}}%
\pgfpathlineto{\pgfqpoint{2.631141in}{1.197218in}}%
\pgfpathlineto{\pgfqpoint{2.808740in}{1.109783in}}%
\pgfpathlineto{\pgfqpoint{2.985017in}{1.038189in}}%
\pgfpathlineto{\pgfqpoint{3.164524in}{1.064787in}}%
\pgfpathlineto{\pgfqpoint{3.347387in}{1.050850in}}%
\pgfpathlineto{\pgfqpoint{3.527911in}{1.059091in}}%
\pgfpathlineto{\pgfqpoint{3.703697in}{1.040082in}}%
\pgfpathlineto{\pgfqpoint{3.883857in}{1.062621in}}%
\pgfpathlineto{\pgfqpoint{4.061730in}{0.860852in}}%
\pgfpathlineto{\pgfqpoint{4.243113in}{1.241874in}}%
\pgfpathlineto{\pgfqpoint{4.420719in}{1.159389in}}%
\pgfpathlineto{\pgfqpoint{4.595779in}{1.209838in}}%
\pgfpathlineto{\pgfqpoint{4.774787in}{1.315400in}}%
\pgfpathlineto{\pgfqpoint{4.954184in}{1.433114in}}%
\pgfpathlineto{\pgfqpoint{5.133992in}{1.172359in}}%
\pgfpathlineto{\pgfqpoint{5.133992in}{1.439903in}}%
\pgfpathlineto{\pgfqpoint{5.133992in}{1.439903in}}%
\pgfpathlineto{\pgfqpoint{4.954184in}{1.622478in}}%
\pgfpathlineto{\pgfqpoint{4.774787in}{1.511067in}}%
\pgfpathlineto{\pgfqpoint{4.595779in}{1.462586in}}%
\pgfpathlineto{\pgfqpoint{4.420719in}{1.480401in}}%
\pgfpathlineto{\pgfqpoint{4.243113in}{1.490366in}}%
\pgfpathlineto{\pgfqpoint{4.061730in}{1.222961in}}%
\pgfpathlineto{\pgfqpoint{3.883857in}{1.341130in}}%
\pgfpathlineto{\pgfqpoint{3.703697in}{1.356352in}}%
\pgfpathlineto{\pgfqpoint{3.527911in}{1.338326in}}%
\pgfpathlineto{\pgfqpoint{3.347387in}{1.382163in}}%
\pgfpathlineto{\pgfqpoint{3.164524in}{1.409517in}}%
\pgfpathlineto{\pgfqpoint{2.985017in}{1.337193in}}%
\pgfpathlineto{\pgfqpoint{2.808740in}{1.375081in}}%
\pgfpathlineto{\pgfqpoint{2.631141in}{1.496992in}}%
\pgfpathlineto{\pgfqpoint{2.453482in}{1.648635in}}%
\pgfpathlineto{\pgfqpoint{2.275178in}{1.655922in}}%
\pgfpathlineto{\pgfqpoint{2.095804in}{1.694933in}}%
\pgfpathlineto{\pgfqpoint{1.919028in}{1.731597in}}%
\pgfpathlineto{\pgfqpoint{1.743052in}{1.766819in}}%
\pgfpathlineto{\pgfqpoint{1.562614in}{1.700656in}}%
\pgfpathlineto{\pgfqpoint{1.386047in}{1.710285in}}%
\pgfpathlineto{\pgfqpoint{1.210785in}{1.792189in}}%
\pgfpathlineto{\pgfqpoint{1.033456in}{1.452317in}}%
\pgfpathlineto{\pgfqpoint{0.858342in}{1.465303in}}%
\pgfpathlineto{\pgfqpoint{0.681370in}{1.572539in}}%
\pgfpathlineto{\pgfqpoint{0.502927in}{1.452437in}}%
\pgfpathlineto{\pgfqpoint{0.326351in}{1.495038in}}%
\pgfpathlineto{\pgfqpoint{0.150965in}{1.843682in}}%
\pgfpathlineto{\pgfqpoint{-0.025493in}{2.084141in}}%
\pgfpathclose%
\pgfusepath{stroke,fill}%
\end{pgfscope}%
\begin{pgfscope}%
\pgfpathrectangle{\pgfqpoint{0.100000in}{0.915754in}}{\pgfqpoint{4.650000in}{3.020000in}}%
\pgfusepath{clip}%
\pgfsetbuttcap%
\pgfsetroundjoin%
\definecolor{currentfill}{rgb}{0.866667,0.517647,0.321569}%
\pgfsetfillcolor{currentfill}%
\pgfsetfillopacity{0.200000}%
\pgfsetlinewidth{1.003750pt}%
\definecolor{currentstroke}{rgb}{0.866667,0.517647,0.321569}%
\pgfsetstrokecolor{currentstroke}%
\pgfsetstrokeopacity{0.200000}%
\pgfsetdash{}{0pt}%
\pgfpathmoveto{\pgfqpoint{-0.024330in}{1.912459in}}%
\pgfpathlineto{\pgfqpoint{-0.024330in}{1.590981in}}%
\pgfpathlineto{\pgfqpoint{0.148830in}{1.828118in}}%
\pgfpathlineto{\pgfqpoint{0.325222in}{1.799141in}}%
\pgfpathlineto{\pgfqpoint{0.502094in}{1.753186in}}%
\pgfpathlineto{\pgfqpoint{0.677239in}{1.786132in}}%
\pgfpathlineto{\pgfqpoint{0.855709in}{1.727401in}}%
\pgfpathlineto{\pgfqpoint{1.037563in}{1.676978in}}%
\pgfpathlineto{\pgfqpoint{1.212270in}{1.770891in}}%
\pgfpathlineto{\pgfqpoint{1.386035in}{1.759647in}}%
\pgfpathlineto{\pgfqpoint{1.564757in}{1.784639in}}%
\pgfpathlineto{\pgfqpoint{1.741664in}{1.679904in}}%
\pgfpathlineto{\pgfqpoint{1.923918in}{1.693820in}}%
\pgfpathlineto{\pgfqpoint{2.096537in}{1.544956in}}%
\pgfpathlineto{\pgfqpoint{2.275391in}{1.542133in}}%
\pgfpathlineto{\pgfqpoint{2.450890in}{1.391387in}}%
\pgfpathlineto{\pgfqpoint{2.634845in}{1.407765in}}%
\pgfpathlineto{\pgfqpoint{2.809588in}{1.310301in}}%
\pgfpathlineto{\pgfqpoint{2.988139in}{1.299651in}}%
\pgfpathlineto{\pgfqpoint{3.171196in}{1.341113in}}%
\pgfpathlineto{\pgfqpoint{3.357004in}{1.341780in}}%
\pgfpathlineto{\pgfqpoint{3.529757in}{1.325945in}}%
\pgfpathlineto{\pgfqpoint{3.710504in}{1.420498in}}%
\pgfpathlineto{\pgfqpoint{3.884907in}{1.382599in}}%
\pgfpathlineto{\pgfqpoint{4.066524in}{1.361338in}}%
\pgfpathlineto{\pgfqpoint{4.245874in}{1.482511in}}%
\pgfpathlineto{\pgfqpoint{4.421112in}{1.354043in}}%
\pgfpathlineto{\pgfqpoint{4.600265in}{1.410769in}}%
\pgfpathlineto{\pgfqpoint{4.779067in}{1.551039in}}%
\pgfpathlineto{\pgfqpoint{4.950395in}{1.484479in}}%
\pgfpathlineto{\pgfqpoint{5.131923in}{1.505326in}}%
\pgfpathlineto{\pgfqpoint{5.131923in}{1.725476in}}%
\pgfpathlineto{\pgfqpoint{5.131923in}{1.725476in}}%
\pgfpathlineto{\pgfqpoint{4.950395in}{1.723097in}}%
\pgfpathlineto{\pgfqpoint{4.779067in}{1.700477in}}%
\pgfpathlineto{\pgfqpoint{4.600265in}{1.625971in}}%
\pgfpathlineto{\pgfqpoint{4.421112in}{1.621965in}}%
\pgfpathlineto{\pgfqpoint{4.245874in}{1.673756in}}%
\pgfpathlineto{\pgfqpoint{4.066524in}{1.610932in}}%
\pgfpathlineto{\pgfqpoint{3.884907in}{1.602225in}}%
\pgfpathlineto{\pgfqpoint{3.710504in}{1.624220in}}%
\pgfpathlineto{\pgfqpoint{3.529757in}{1.536592in}}%
\pgfpathlineto{\pgfqpoint{3.357004in}{1.593968in}}%
\pgfpathlineto{\pgfqpoint{3.171196in}{1.582577in}}%
\pgfpathlineto{\pgfqpoint{2.988139in}{1.482873in}}%
\pgfpathlineto{\pgfqpoint{2.809588in}{1.559908in}}%
\pgfpathlineto{\pgfqpoint{2.634845in}{1.605649in}}%
\pgfpathlineto{\pgfqpoint{2.450890in}{1.625697in}}%
\pgfpathlineto{\pgfqpoint{2.275391in}{1.763876in}}%
\pgfpathlineto{\pgfqpoint{2.096537in}{1.759067in}}%
\pgfpathlineto{\pgfqpoint{1.923918in}{1.810618in}}%
\pgfpathlineto{\pgfqpoint{1.741664in}{1.835673in}}%
\pgfpathlineto{\pgfqpoint{1.564757in}{1.914421in}}%
\pgfpathlineto{\pgfqpoint{1.386035in}{1.899147in}}%
\pgfpathlineto{\pgfqpoint{1.212270in}{1.920242in}}%
\pgfpathlineto{\pgfqpoint{1.037563in}{1.849199in}}%
\pgfpathlineto{\pgfqpoint{0.855709in}{1.913068in}}%
\pgfpathlineto{\pgfqpoint{0.677239in}{1.969634in}}%
\pgfpathlineto{\pgfqpoint{0.502094in}{1.944682in}}%
\pgfpathlineto{\pgfqpoint{0.325222in}{2.030360in}}%
\pgfpathlineto{\pgfqpoint{0.148830in}{2.036910in}}%
\pgfpathlineto{\pgfqpoint{-0.024330in}{1.912459in}}%
\pgfpathclose%
\pgfusepath{stroke,fill}%
\end{pgfscope}%
\begin{pgfscope}%
\pgfpathrectangle{\pgfqpoint{0.100000in}{0.915754in}}{\pgfqpoint{4.650000in}{3.020000in}}%
\pgfusepath{clip}%
\pgfsetbuttcap%
\pgfsetroundjoin%
\definecolor{currentfill}{rgb}{0.333333,0.658824,0.407843}%
\pgfsetfillcolor{currentfill}%
\pgfsetfillopacity{0.200000}%
\pgfsetlinewidth{1.003750pt}%
\definecolor{currentstroke}{rgb}{0.333333,0.658824,0.407843}%
\pgfsetstrokecolor{currentstroke}%
\pgfsetstrokeopacity{0.200000}%
\pgfsetdash{}{0pt}%
\pgfpathmoveto{\pgfqpoint{0.018881in}{2.346814in}}%
\pgfpathlineto{\pgfqpoint{0.018881in}{1.998080in}}%
\pgfpathlineto{\pgfqpoint{0.221315in}{1.820160in}}%
\pgfpathlineto{\pgfqpoint{0.382941in}{1.736270in}}%
\pgfpathlineto{\pgfqpoint{0.561725in}{1.722336in}}%
\pgfpathlineto{\pgfqpoint{0.759263in}{1.584109in}}%
\pgfpathlineto{\pgfqpoint{0.940426in}{1.529700in}}%
\pgfpathlineto{\pgfqpoint{1.110746in}{1.472389in}}%
\pgfpathlineto{\pgfqpoint{1.301967in}{1.477452in}}%
\pgfpathlineto{\pgfqpoint{1.457643in}{1.414211in}}%
\pgfpathlineto{\pgfqpoint{1.659453in}{1.221835in}}%
\pgfpathlineto{\pgfqpoint{1.859748in}{1.273308in}}%
\pgfpathlineto{\pgfqpoint{2.006086in}{1.390467in}}%
\pgfpathlineto{\pgfqpoint{2.203336in}{1.383800in}}%
\pgfpathlineto{\pgfqpoint{2.385605in}{1.182994in}}%
\pgfpathlineto{\pgfqpoint{2.589079in}{1.176485in}}%
\pgfpathlineto{\pgfqpoint{2.723881in}{1.151362in}}%
\pgfpathlineto{\pgfqpoint{2.940385in}{1.155810in}}%
\pgfpathlineto{\pgfqpoint{3.100758in}{1.006931in}}%
\pgfpathlineto{\pgfqpoint{3.278805in}{1.184210in}}%
\pgfpathlineto{\pgfqpoint{3.460600in}{1.150123in}}%
\pgfpathlineto{\pgfqpoint{3.651117in}{1.176160in}}%
\pgfpathlineto{\pgfqpoint{3.831658in}{0.993851in}}%
\pgfpathlineto{\pgfqpoint{4.000892in}{0.978885in}}%
\pgfpathlineto{\pgfqpoint{4.198249in}{1.181683in}}%
\pgfpathlineto{\pgfqpoint{4.394496in}{0.918947in}}%
\pgfpathlineto{\pgfqpoint{4.563022in}{1.161161in}}%
\pgfpathlineto{\pgfqpoint{4.761398in}{1.039212in}}%
\pgfpathlineto{\pgfqpoint{4.947457in}{1.229795in}}%
\pgfpathlineto{\pgfqpoint{5.081962in}{0.893519in}}%
\pgfpathlineto{\pgfqpoint{5.290417in}{0.994147in}}%
\pgfpathlineto{\pgfqpoint{5.290417in}{1.352119in}}%
\pgfpathlineto{\pgfqpoint{5.290417in}{1.352119in}}%
\pgfpathlineto{\pgfqpoint{5.081962in}{1.185946in}}%
\pgfpathlineto{\pgfqpoint{4.947457in}{1.487983in}}%
\pgfpathlineto{\pgfqpoint{4.761398in}{1.417599in}}%
\pgfpathlineto{\pgfqpoint{4.563022in}{1.494574in}}%
\pgfpathlineto{\pgfqpoint{4.394496in}{1.299688in}}%
\pgfpathlineto{\pgfqpoint{4.198249in}{1.420789in}}%
\pgfpathlineto{\pgfqpoint{4.000892in}{1.340743in}}%
\pgfpathlineto{\pgfqpoint{3.831658in}{1.396235in}}%
\pgfpathlineto{\pgfqpoint{3.651117in}{1.495485in}}%
\pgfpathlineto{\pgfqpoint{3.460600in}{1.470224in}}%
\pgfpathlineto{\pgfqpoint{3.278805in}{1.473733in}}%
\pgfpathlineto{\pgfqpoint{3.100758in}{1.393709in}}%
\pgfpathlineto{\pgfqpoint{2.940385in}{1.431369in}}%
\pgfpathlineto{\pgfqpoint{2.723881in}{1.462908in}}%
\pgfpathlineto{\pgfqpoint{2.589079in}{1.612861in}}%
\pgfpathlineto{\pgfqpoint{2.385605in}{1.505559in}}%
\pgfpathlineto{\pgfqpoint{2.203336in}{1.739405in}}%
\pgfpathlineto{\pgfqpoint{2.006086in}{1.734346in}}%
\pgfpathlineto{\pgfqpoint{1.859748in}{1.714459in}}%
\pgfpathlineto{\pgfqpoint{1.659453in}{1.538569in}}%
\pgfpathlineto{\pgfqpoint{1.457643in}{1.753478in}}%
\pgfpathlineto{\pgfqpoint{1.301967in}{1.753307in}}%
\pgfpathlineto{\pgfqpoint{1.110746in}{1.838094in}}%
\pgfpathlineto{\pgfqpoint{0.940426in}{1.820049in}}%
\pgfpathlineto{\pgfqpoint{0.759263in}{1.944306in}}%
\pgfpathlineto{\pgfqpoint{0.561725in}{1.972324in}}%
\pgfpathlineto{\pgfqpoint{0.382941in}{2.132683in}}%
\pgfpathlineto{\pgfqpoint{0.221315in}{2.180789in}}%
\pgfpathlineto{\pgfqpoint{0.018881in}{2.346814in}}%
\pgfpathclose%
\pgfusepath{stroke,fill}%
\end{pgfscope}%
\begin{pgfscope}%
\pgfpathrectangle{\pgfqpoint{0.100000in}{0.915754in}}{\pgfqpoint{4.650000in}{3.020000in}}%
\pgfusepath{clip}%
\pgfsetbuttcap%
\pgfsetroundjoin%
\definecolor{currentfill}{rgb}{0.768627,0.305882,0.321569}%
\pgfsetfillcolor{currentfill}%
\pgfsetfillopacity{0.200000}%
\pgfsetlinewidth{1.003750pt}%
\definecolor{currentstroke}{rgb}{0.768627,0.305882,0.321569}%
\pgfsetstrokecolor{currentstroke}%
\pgfsetstrokeopacity{0.200000}%
\pgfsetdash{}{0pt}%
\pgfpathmoveto{\pgfqpoint{-0.066071in}{2.900595in}}%
\pgfpathlineto{\pgfqpoint{-0.066071in}{2.602980in}}%
\pgfpathlineto{\pgfqpoint{0.105727in}{2.511748in}}%
\pgfpathlineto{\pgfqpoint{0.277525in}{2.497752in}}%
\pgfpathlineto{\pgfqpoint{0.449323in}{2.411268in}}%
\pgfpathlineto{\pgfqpoint{0.621121in}{2.202492in}}%
\pgfpathlineto{\pgfqpoint{0.792919in}{2.208457in}}%
\pgfpathlineto{\pgfqpoint{0.964717in}{2.105379in}}%
\pgfpathlineto{\pgfqpoint{1.136515in}{2.102008in}}%
\pgfpathlineto{\pgfqpoint{1.308313in}{2.104718in}}%
\pgfpathlineto{\pgfqpoint{1.480111in}{2.132939in}}%
\pgfpathlineto{\pgfqpoint{1.651909in}{2.088703in}}%
\pgfpathlineto{\pgfqpoint{1.823707in}{2.087820in}}%
\pgfpathlineto{\pgfqpoint{1.995505in}{1.933053in}}%
\pgfpathlineto{\pgfqpoint{2.167303in}{2.105063in}}%
\pgfpathlineto{\pgfqpoint{2.339101in}{1.906935in}}%
\pgfpathlineto{\pgfqpoint{2.510899in}{1.958859in}}%
\pgfpathlineto{\pgfqpoint{2.682697in}{1.871984in}}%
\pgfpathlineto{\pgfqpoint{2.854495in}{1.976218in}}%
\pgfpathlineto{\pgfqpoint{3.026293in}{1.890654in}}%
\pgfpathlineto{\pgfqpoint{3.198091in}{1.968852in}}%
\pgfpathlineto{\pgfqpoint{3.369889in}{1.900848in}}%
\pgfpathlineto{\pgfqpoint{3.541687in}{1.765585in}}%
\pgfpathlineto{\pgfqpoint{3.713485in}{1.646362in}}%
\pgfpathlineto{\pgfqpoint{3.885283in}{1.907639in}}%
\pgfpathlineto{\pgfqpoint{4.057081in}{1.832764in}}%
\pgfpathlineto{\pgfqpoint{4.228879in}{1.804445in}}%
\pgfpathlineto{\pgfqpoint{4.400677in}{1.789739in}}%
\pgfpathlineto{\pgfqpoint{4.572475in}{1.917479in}}%
\pgfpathlineto{\pgfqpoint{4.744273in}{1.774828in}}%
\pgfpathlineto{\pgfqpoint{4.916071in}{1.693162in}}%
\pgfpathlineto{\pgfqpoint{4.916071in}{1.987473in}}%
\pgfpathlineto{\pgfqpoint{4.916071in}{1.987473in}}%
\pgfpathlineto{\pgfqpoint{4.744273in}{2.073347in}}%
\pgfpathlineto{\pgfqpoint{4.572475in}{2.208746in}}%
\pgfpathlineto{\pgfqpoint{4.400677in}{2.144924in}}%
\pgfpathlineto{\pgfqpoint{4.228879in}{2.111981in}}%
\pgfpathlineto{\pgfqpoint{4.057081in}{2.192165in}}%
\pgfpathlineto{\pgfqpoint{3.885283in}{2.207491in}}%
\pgfpathlineto{\pgfqpoint{3.713485in}{1.971886in}}%
\pgfpathlineto{\pgfqpoint{3.541687in}{2.146813in}}%
\pgfpathlineto{\pgfqpoint{3.369889in}{2.228193in}}%
\pgfpathlineto{\pgfqpoint{3.198091in}{2.243663in}}%
\pgfpathlineto{\pgfqpoint{3.026293in}{2.214208in}}%
\pgfpathlineto{\pgfqpoint{2.854495in}{2.330105in}}%
\pgfpathlineto{\pgfqpoint{2.682697in}{2.167778in}}%
\pgfpathlineto{\pgfqpoint{2.510899in}{2.255458in}}%
\pgfpathlineto{\pgfqpoint{2.339101in}{2.271608in}}%
\pgfpathlineto{\pgfqpoint{2.167303in}{2.379748in}}%
\pgfpathlineto{\pgfqpoint{1.995505in}{2.280235in}}%
\pgfpathlineto{\pgfqpoint{1.823707in}{2.428017in}}%
\pgfpathlineto{\pgfqpoint{1.651909in}{2.375847in}}%
\pgfpathlineto{\pgfqpoint{1.480111in}{2.479792in}}%
\pgfpathlineto{\pgfqpoint{1.308313in}{2.457950in}}%
\pgfpathlineto{\pgfqpoint{1.136515in}{2.383893in}}%
\pgfpathlineto{\pgfqpoint{0.964717in}{2.421705in}}%
\pgfpathlineto{\pgfqpoint{0.792919in}{2.553817in}}%
\pgfpathlineto{\pgfqpoint{0.621121in}{2.538396in}}%
\pgfpathlineto{\pgfqpoint{0.449323in}{2.710186in}}%
\pgfpathlineto{\pgfqpoint{0.277525in}{2.885351in}}%
\pgfpathlineto{\pgfqpoint{0.105727in}{2.778854in}}%
\pgfpathlineto{\pgfqpoint{-0.066071in}{2.900595in}}%
\pgfpathclose%
\pgfusepath{stroke,fill}%
\end{pgfscope}%
\begin{pgfscope}%
\pgfpathrectangle{\pgfqpoint{0.100000in}{0.915754in}}{\pgfqpoint{4.650000in}{3.020000in}}%
\pgfusepath{clip}%
\pgfsetroundcap%
\pgfsetroundjoin%
\pgfsetlinewidth{1.505625pt}%
\definecolor{currentstroke}{rgb}{0.298039,0.447059,0.690196}%
\pgfsetstrokecolor{currentstroke}%
\pgfsetdash{}{0pt}%
\pgfpathmoveto{\pgfqpoint{0.086111in}{1.799090in}}%
\pgfpathlineto{\pgfqpoint{0.150965in}{1.717270in}}%
\pgfpathlineto{\pgfqpoint{0.326351in}{1.334989in}}%
\pgfpathlineto{\pgfqpoint{0.502927in}{1.316619in}}%
\pgfpathlineto{\pgfqpoint{0.681370in}{1.425767in}}%
\pgfpathlineto{\pgfqpoint{0.858342in}{1.346534in}}%
\pgfpathlineto{\pgfqpoint{1.033456in}{1.307210in}}%
\pgfpathlineto{\pgfqpoint{1.210785in}{1.671624in}}%
\pgfpathlineto{\pgfqpoint{1.386047in}{1.622239in}}%
\pgfpathlineto{\pgfqpoint{1.562614in}{1.599482in}}%
\pgfpathlineto{\pgfqpoint{1.743052in}{1.666687in}}%
\pgfpathlineto{\pgfqpoint{1.919028in}{1.644850in}}%
\pgfpathlineto{\pgfqpoint{2.095804in}{1.614859in}}%
\pgfpathlineto{\pgfqpoint{2.275178in}{1.555134in}}%
\pgfpathlineto{\pgfqpoint{2.453482in}{1.543852in}}%
\pgfpathlineto{\pgfqpoint{2.631141in}{1.363045in}}%
\pgfpathlineto{\pgfqpoint{2.808740in}{1.255034in}}%
\pgfpathlineto{\pgfqpoint{2.985017in}{1.204738in}}%
\pgfpathlineto{\pgfqpoint{3.164524in}{1.252279in}}%
\pgfpathlineto{\pgfqpoint{3.347387in}{1.230273in}}%
\pgfpathlineto{\pgfqpoint{3.527911in}{1.203392in}}%
\pgfpathlineto{\pgfqpoint{3.703697in}{1.210742in}}%
\pgfpathlineto{\pgfqpoint{3.883857in}{1.216084in}}%
\pgfpathlineto{\pgfqpoint{4.061730in}{1.058731in}}%
\pgfpathlineto{\pgfqpoint{4.243113in}{1.373883in}}%
\pgfpathlineto{\pgfqpoint{4.420719in}{1.348164in}}%
\pgfpathlineto{\pgfqpoint{4.595779in}{1.342784in}}%
\pgfpathlineto{\pgfqpoint{4.763889in}{1.414585in}}%
\pgfusepath{stroke}%
\end{pgfscope}%
\begin{pgfscope}%
\pgfpathrectangle{\pgfqpoint{0.100000in}{0.915754in}}{\pgfqpoint{4.650000in}{3.020000in}}%
\pgfusepath{clip}%
\pgfsetroundcap%
\pgfsetroundjoin%
\pgfsetlinewidth{1.505625pt}%
\definecolor{currentstroke}{rgb}{0.866667,0.517647,0.321569}%
\pgfsetstrokecolor{currentstroke}%
\pgfsetdash{}{0pt}%
\pgfpathmoveto{\pgfqpoint{0.086111in}{1.877423in}}%
\pgfpathlineto{\pgfqpoint{0.148830in}{1.942224in}}%
\pgfpathlineto{\pgfqpoint{0.325222in}{1.921984in}}%
\pgfpathlineto{\pgfqpoint{0.502094in}{1.856705in}}%
\pgfpathlineto{\pgfqpoint{0.677239in}{1.883092in}}%
\pgfpathlineto{\pgfqpoint{0.855709in}{1.823107in}}%
\pgfpathlineto{\pgfqpoint{1.037563in}{1.768924in}}%
\pgfpathlineto{\pgfqpoint{1.212270in}{1.847449in}}%
\pgfpathlineto{\pgfqpoint{1.386035in}{1.832392in}}%
\pgfpathlineto{\pgfqpoint{1.564757in}{1.847251in}}%
\pgfpathlineto{\pgfqpoint{1.741664in}{1.761217in}}%
\pgfpathlineto{\pgfqpoint{1.923918in}{1.751073in}}%
\pgfpathlineto{\pgfqpoint{2.096537in}{1.657609in}}%
\pgfpathlineto{\pgfqpoint{2.275391in}{1.659272in}}%
\pgfpathlineto{\pgfqpoint{2.450890in}{1.520711in}}%
\pgfpathlineto{\pgfqpoint{2.634845in}{1.518861in}}%
\pgfpathlineto{\pgfqpoint{2.809588in}{1.445772in}}%
\pgfpathlineto{\pgfqpoint{2.988139in}{1.394377in}}%
\pgfpathlineto{\pgfqpoint{3.171196in}{1.479070in}}%
\pgfpathlineto{\pgfqpoint{3.357004in}{1.475323in}}%
\pgfpathlineto{\pgfqpoint{3.529757in}{1.440029in}}%
\pgfpathlineto{\pgfqpoint{3.710504in}{1.528548in}}%
\pgfpathlineto{\pgfqpoint{3.884907in}{1.497463in}}%
\pgfpathlineto{\pgfqpoint{4.066524in}{1.501646in}}%
\pgfpathlineto{\pgfqpoint{4.245874in}{1.587206in}}%
\pgfpathlineto{\pgfqpoint{4.421112in}{1.503211in}}%
\pgfpathlineto{\pgfqpoint{4.600265in}{1.513163in}}%
\pgfpathlineto{\pgfqpoint{4.763889in}{1.619445in}}%
\pgfusepath{stroke}%
\end{pgfscope}%
\begin{pgfscope}%
\pgfpathrectangle{\pgfqpoint{0.100000in}{0.915754in}}{\pgfqpoint{4.650000in}{3.020000in}}%
\pgfusepath{clip}%
\pgfsetroundcap%
\pgfsetroundjoin%
\pgfsetlinewidth{1.505625pt}%
\definecolor{currentstroke}{rgb}{0.333333,0.658824,0.407843}%
\pgfsetstrokecolor{currentstroke}%
\pgfsetdash{}{0pt}%
\pgfpathmoveto{\pgfqpoint{0.086111in}{2.132542in}}%
\pgfpathlineto{\pgfqpoint{0.221315in}{2.022375in}}%
\pgfpathlineto{\pgfqpoint{0.382941in}{1.959378in}}%
\pgfpathlineto{\pgfqpoint{0.561725in}{1.857402in}}%
\pgfpathlineto{\pgfqpoint{0.759263in}{1.781845in}}%
\pgfpathlineto{\pgfqpoint{0.940426in}{1.685371in}}%
\pgfpathlineto{\pgfqpoint{1.110746in}{1.672154in}}%
\pgfpathlineto{\pgfqpoint{1.301967in}{1.629219in}}%
\pgfpathlineto{\pgfqpoint{1.457643in}{1.599882in}}%
\pgfpathlineto{\pgfqpoint{1.659453in}{1.400993in}}%
\pgfpathlineto{\pgfqpoint{1.859748in}{1.509138in}}%
\pgfpathlineto{\pgfqpoint{2.006086in}{1.577683in}}%
\pgfpathlineto{\pgfqpoint{2.203336in}{1.581901in}}%
\pgfpathlineto{\pgfqpoint{2.385605in}{1.356984in}}%
\pgfpathlineto{\pgfqpoint{2.589079in}{1.416261in}}%
\pgfpathlineto{\pgfqpoint{2.723881in}{1.323665in}}%
\pgfpathlineto{\pgfqpoint{2.940385in}{1.307997in}}%
\pgfpathlineto{\pgfqpoint{3.100758in}{1.225465in}}%
\pgfpathlineto{\pgfqpoint{3.278805in}{1.340971in}}%
\pgfpathlineto{\pgfqpoint{3.460600in}{1.325494in}}%
\pgfpathlineto{\pgfqpoint{3.651117in}{1.349043in}}%
\pgfpathlineto{\pgfqpoint{3.831658in}{1.222456in}}%
\pgfpathlineto{\pgfqpoint{4.000892in}{1.174725in}}%
\pgfpathlineto{\pgfqpoint{4.198249in}{1.306363in}}%
\pgfpathlineto{\pgfqpoint{4.394496in}{1.137899in}}%
\pgfpathlineto{\pgfqpoint{4.563022in}{1.347980in}}%
\pgfpathlineto{\pgfqpoint{4.761398in}{1.253397in}}%
\pgfpathlineto{\pgfqpoint{4.763889in}{1.254975in}}%
\pgfusepath{stroke}%
\end{pgfscope}%
\begin{pgfscope}%
\pgfpathrectangle{\pgfqpoint{0.100000in}{0.915754in}}{\pgfqpoint{4.650000in}{3.020000in}}%
\pgfusepath{clip}%
\pgfsetroundcap%
\pgfsetroundjoin%
\pgfsetlinewidth{1.505625pt}%
\definecolor{currentstroke}{rgb}{0.768627,0.305882,0.321569}%
\pgfsetstrokecolor{currentstroke}%
\pgfsetdash{}{0pt}%
\pgfpathmoveto{\pgfqpoint{0.086111in}{2.676346in}}%
\pgfpathlineto{\pgfqpoint{0.105727in}{2.664447in}}%
\pgfpathlineto{\pgfqpoint{0.277525in}{2.712321in}}%
\pgfpathlineto{\pgfqpoint{0.449323in}{2.566513in}}%
\pgfpathlineto{\pgfqpoint{0.621121in}{2.374816in}}%
\pgfpathlineto{\pgfqpoint{0.792919in}{2.391077in}}%
\pgfpathlineto{\pgfqpoint{0.964717in}{2.281455in}}%
\pgfpathlineto{\pgfqpoint{1.136515in}{2.248241in}}%
\pgfpathlineto{\pgfqpoint{1.308313in}{2.308902in}}%
\pgfpathlineto{\pgfqpoint{1.480111in}{2.324019in}}%
\pgfpathlineto{\pgfqpoint{1.651909in}{2.242877in}}%
\pgfpathlineto{\pgfqpoint{1.823707in}{2.279620in}}%
\pgfpathlineto{\pgfqpoint{1.995505in}{2.119370in}}%
\pgfpathlineto{\pgfqpoint{2.167303in}{2.258228in}}%
\pgfpathlineto{\pgfqpoint{2.339101in}{2.107369in}}%
\pgfpathlineto{\pgfqpoint{2.510899in}{2.116998in}}%
\pgfpathlineto{\pgfqpoint{2.682697in}{2.033594in}}%
\pgfpathlineto{\pgfqpoint{2.854495in}{2.159284in}}%
\pgfpathlineto{\pgfqpoint{3.026293in}{2.064167in}}%
\pgfpathlineto{\pgfqpoint{3.198091in}{2.106955in}}%
\pgfpathlineto{\pgfqpoint{3.369889in}{2.079765in}}%
\pgfpathlineto{\pgfqpoint{3.541687in}{1.987117in}}%
\pgfpathlineto{\pgfqpoint{3.713485in}{1.830356in}}%
\pgfpathlineto{\pgfqpoint{3.885283in}{2.060244in}}%
\pgfpathlineto{\pgfqpoint{4.057081in}{2.026652in}}%
\pgfpathlineto{\pgfqpoint{4.228879in}{1.967132in}}%
\pgfpathlineto{\pgfqpoint{4.400677in}{1.997319in}}%
\pgfpathlineto{\pgfqpoint{4.572475in}{2.071927in}}%
\pgfpathlineto{\pgfqpoint{4.744273in}{1.929660in}}%
\pgfpathlineto{\pgfqpoint{4.763889in}{1.920599in}}%
\pgfusepath{stroke}%
\end{pgfscope}%
\begin{pgfscope}%
\pgfsetrectcap%
\pgfsetmiterjoin%
\pgfsetlinewidth{1.254687pt}%
\definecolor{currentstroke}{rgb}{0.800000,0.800000,0.800000}%
\pgfsetstrokecolor{currentstroke}%
\pgfsetdash{}{0pt}%
\pgfpathmoveto{\pgfqpoint{0.100000in}{0.915754in}}%
\pgfpathlineto{\pgfqpoint{0.100000in}{3.935754in}}%
\pgfusepath{stroke}%
\end{pgfscope}%
\begin{pgfscope}%
\pgfsetrectcap%
\pgfsetmiterjoin%
\pgfsetlinewidth{1.254687pt}%
\definecolor{currentstroke}{rgb}{0.800000,0.800000,0.800000}%
\pgfsetstrokecolor{currentstroke}%
\pgfsetdash{}{0pt}%
\pgfpathmoveto{\pgfqpoint{4.750000in}{0.915754in}}%
\pgfpathlineto{\pgfqpoint{4.750000in}{3.935754in}}%
\pgfusepath{stroke}%
\end{pgfscope}%
\begin{pgfscope}%
\pgfsetrectcap%
\pgfsetmiterjoin%
\pgfsetlinewidth{1.254687pt}%
\definecolor{currentstroke}{rgb}{0.800000,0.800000,0.800000}%
\pgfsetstrokecolor{currentstroke}%
\pgfsetdash{}{0pt}%
\pgfpathmoveto{\pgfqpoint{0.100000in}{0.915754in}}%
\pgfpathlineto{\pgfqpoint{4.750000in}{0.915754in}}%
\pgfusepath{stroke}%
\end{pgfscope}%
\begin{pgfscope}%
\pgfsetrectcap%
\pgfsetmiterjoin%
\pgfsetlinewidth{1.254687pt}%
\definecolor{currentstroke}{rgb}{0.800000,0.800000,0.800000}%
\pgfsetstrokecolor{currentstroke}%
\pgfsetdash{}{0pt}%
\pgfpathmoveto{\pgfqpoint{0.100000in}{3.935754in}}%
\pgfpathlineto{\pgfqpoint{4.750000in}{3.935754in}}%
\pgfusepath{stroke}%
\end{pgfscope}%
\end{pgfpicture}%
\makeatother%
\endgroup%

%% file: fig/continuous_two_circles_res.pgf
%% Creator: Matplotlib, PGF backend
%%
%% To include the figure in your LaTeX document, write
%%   \input{<filename>.pgf}
%%
%% Make sure the required packages are loaded in your preamble
%%   \usepackage{pgf}
%%
%% Figures using additional raster images can only be included by \input if
%% they are in the same directory as the main LaTeX file. For loading figures
%% from other directories you can use the `import` package
%% and then include the figures with
%%   \import{<path to file>}{<filename>.pgf}
%%
%% Matplotlib used the following preamble
%%
\begingroup%
\makeatletter%
\begin{pgfpicture}%
\pgfpathrectangle{\pgfpointorigin}{\pgfqpoint{4.850000in}{4.035754in}}%
\pgfusepath{use as bounding box, clip}%
\begin{pgfscope}%
\pgfsetbuttcap%
\pgfsetmiterjoin%
\definecolor{currentfill}{rgb}{1.000000,1.000000,1.000000}%
\pgfsetfillcolor{currentfill}%
\pgfsetlinewidth{0.000000pt}%
\definecolor{currentstroke}{rgb}{1.000000,1.000000,1.000000}%
\pgfsetstrokecolor{currentstroke}%
\pgfsetdash{}{0pt}%
\pgfpathmoveto{\pgfqpoint{0.000000in}{0.000000in}}%
\pgfpathlineto{\pgfqpoint{4.850000in}{0.000000in}}%
\pgfpathlineto{\pgfqpoint{4.850000in}{4.035754in}}%
\pgfpathlineto{\pgfqpoint{0.000000in}{4.035754in}}%
\pgfpathclose%
\pgfusepath{fill}%
\end{pgfscope}%
\begin{pgfscope}%
\pgfsetbuttcap%
\pgfsetmiterjoin%
\definecolor{currentfill}{rgb}{1.000000,1.000000,1.000000}%
\pgfsetfillcolor{currentfill}%
\pgfsetlinewidth{0.000000pt}%
\definecolor{currentstroke}{rgb}{0.000000,0.000000,0.000000}%
\pgfsetstrokecolor{currentstroke}%
\pgfsetstrokeopacity{0.000000}%
\pgfsetdash{}{0pt}%
\pgfpathmoveto{\pgfqpoint{0.100000in}{0.915754in}}%
\pgfpathlineto{\pgfqpoint{4.750000in}{0.915754in}}%
\pgfpathlineto{\pgfqpoint{4.750000in}{3.935754in}}%
\pgfpathlineto{\pgfqpoint{0.100000in}{3.935754in}}%
\pgfpathclose%
\pgfusepath{fill}%
\end{pgfscope}%
\begin{pgfscope}%
\pgfpathrectangle{\pgfqpoint{0.100000in}{0.915754in}}{\pgfqpoint{4.650000in}{3.020000in}}%
\pgfusepath{clip}%
\pgfsetroundcap%
\pgfsetroundjoin%
\pgfsetlinewidth{1.003750pt}%
\definecolor{currentstroke}{rgb}{0.800000,0.800000,0.800000}%
\pgfsetstrokecolor{currentstroke}%
\pgfsetdash{}{0pt}%
\pgfpathmoveto{\pgfqpoint{0.349107in}{0.915754in}}%
\pgfpathlineto{\pgfqpoint{0.349107in}{3.935754in}}%
\pgfusepath{stroke}%
\end{pgfscope}%
\begin{pgfscope}%
\definecolor{textcolor}{rgb}{0.150000,0.150000,0.150000}%
\pgfsetstrokecolor{textcolor}%
\pgfsetfillcolor{textcolor}%
\pgftext[x=0.349107in,y=0.783810in,,top]{\color{textcolor}\rmfamily\fontsize{23.100000}{27.720000}\selectfont 10}%
\end{pgfscope}%
\begin{pgfscope}%
\pgfpathrectangle{\pgfqpoint{0.100000in}{0.915754in}}{\pgfqpoint{4.650000in}{3.020000in}}%
\pgfusepath{clip}%
\pgfsetroundcap%
\pgfsetroundjoin%
\pgfsetlinewidth{1.003750pt}%
\definecolor{currentstroke}{rgb}{0.800000,0.800000,0.800000}%
\pgfsetstrokecolor{currentstroke}%
\pgfsetdash{}{0pt}%
\pgfpathmoveto{\pgfqpoint{1.179464in}{0.915754in}}%
\pgfpathlineto{\pgfqpoint{1.179464in}{3.935754in}}%
\pgfusepath{stroke}%
\end{pgfscope}%
\begin{pgfscope}%
\definecolor{textcolor}{rgb}{0.150000,0.150000,0.150000}%
\pgfsetstrokecolor{textcolor}%
\pgfsetfillcolor{textcolor}%
\pgftext[x=1.179464in,y=0.783810in,,top]{\color{textcolor}\rmfamily\fontsize{23.100000}{27.720000}\selectfont 20}%
\end{pgfscope}%
\begin{pgfscope}%
\pgfpathrectangle{\pgfqpoint{0.100000in}{0.915754in}}{\pgfqpoint{4.650000in}{3.020000in}}%
\pgfusepath{clip}%
\pgfsetroundcap%
\pgfsetroundjoin%
\pgfsetlinewidth{1.003750pt}%
\definecolor{currentstroke}{rgb}{0.800000,0.800000,0.800000}%
\pgfsetstrokecolor{currentstroke}%
\pgfsetdash{}{0pt}%
\pgfpathmoveto{\pgfqpoint{2.009821in}{0.915754in}}%
\pgfpathlineto{\pgfqpoint{2.009821in}{3.935754in}}%
\pgfusepath{stroke}%
\end{pgfscope}%
\begin{pgfscope}%
\definecolor{textcolor}{rgb}{0.150000,0.150000,0.150000}%
\pgfsetstrokecolor{textcolor}%
\pgfsetfillcolor{textcolor}%
\pgftext[x=2.009821in,y=0.783810in,,top]{\color{textcolor}\rmfamily\fontsize{23.100000}{27.720000}\selectfont 30}%
\end{pgfscope}%
\begin{pgfscope}%
\pgfpathrectangle{\pgfqpoint{0.100000in}{0.915754in}}{\pgfqpoint{4.650000in}{3.020000in}}%
\pgfusepath{clip}%
\pgfsetroundcap%
\pgfsetroundjoin%
\pgfsetlinewidth{1.003750pt}%
\definecolor{currentstroke}{rgb}{0.800000,0.800000,0.800000}%
\pgfsetstrokecolor{currentstroke}%
\pgfsetdash{}{0pt}%
\pgfpathmoveto{\pgfqpoint{2.840179in}{0.915754in}}%
\pgfpathlineto{\pgfqpoint{2.840179in}{3.935754in}}%
\pgfusepath{stroke}%
\end{pgfscope}%
\begin{pgfscope}%
\definecolor{textcolor}{rgb}{0.150000,0.150000,0.150000}%
\pgfsetstrokecolor{textcolor}%
\pgfsetfillcolor{textcolor}%
\pgftext[x=2.840179in,y=0.783810in,,top]{\color{textcolor}\rmfamily\fontsize{23.100000}{27.720000}\selectfont 40}%
\end{pgfscope}%
\begin{pgfscope}%
\pgfpathrectangle{\pgfqpoint{0.100000in}{0.915754in}}{\pgfqpoint{4.650000in}{3.020000in}}%
\pgfusepath{clip}%
\pgfsetroundcap%
\pgfsetroundjoin%
\pgfsetlinewidth{1.003750pt}%
\definecolor{currentstroke}{rgb}{0.800000,0.800000,0.800000}%
\pgfsetstrokecolor{currentstroke}%
\pgfsetdash{}{0pt}%
\pgfpathmoveto{\pgfqpoint{3.670536in}{0.915754in}}%
\pgfpathlineto{\pgfqpoint{3.670536in}{3.935754in}}%
\pgfusepath{stroke}%
\end{pgfscope}%
\begin{pgfscope}%
\definecolor{textcolor}{rgb}{0.150000,0.150000,0.150000}%
\pgfsetstrokecolor{textcolor}%
\pgfsetfillcolor{textcolor}%
\pgftext[x=3.670536in,y=0.783810in,,top]{\color{textcolor}\rmfamily\fontsize{23.100000}{27.720000}\selectfont 50}%
\end{pgfscope}%
\begin{pgfscope}%
\pgfpathrectangle{\pgfqpoint{0.100000in}{0.915754in}}{\pgfqpoint{4.650000in}{3.020000in}}%
\pgfusepath{clip}%
\pgfsetroundcap%
\pgfsetroundjoin%
\pgfsetlinewidth{1.003750pt}%
\definecolor{currentstroke}{rgb}{0.800000,0.800000,0.800000}%
\pgfsetstrokecolor{currentstroke}%
\pgfsetdash{}{0pt}%
\pgfpathmoveto{\pgfqpoint{4.500893in}{0.915754in}}%
\pgfpathlineto{\pgfqpoint{4.500893in}{3.935754in}}%
\pgfusepath{stroke}%
\end{pgfscope}%
\begin{pgfscope}%
\definecolor{textcolor}{rgb}{0.150000,0.150000,0.150000}%
\pgfsetstrokecolor{textcolor}%
\pgfsetfillcolor{textcolor}%
\pgftext[x=4.500893in,y=0.783810in,,top]{\color{textcolor}\rmfamily\fontsize{23.100000}{27.720000}\selectfont 60}%
\end{pgfscope}%
\begin{pgfscope}%
\definecolor{textcolor}{rgb}{0.150000,0.150000,0.150000}%
\pgfsetstrokecolor{textcolor}%
\pgfsetfillcolor{textcolor}%
\pgftext[x=2.425000in,y=0.407183in,,top]{\color{textcolor}\rmfamily\fontsize{25.200000}{30.240000}\selectfont runtime in seconds}%
\end{pgfscope}%
\begin{pgfscope}%
\pgfpathrectangle{\pgfqpoint{0.100000in}{0.915754in}}{\pgfqpoint{4.650000in}{3.020000in}}%
\pgfusepath{clip}%
\pgfsetroundcap%
\pgfsetroundjoin%
\pgfsetlinewidth{1.003750pt}%
\definecolor{currentstroke}{rgb}{0.800000,0.800000,0.800000}%
\pgfsetstrokecolor{currentstroke}%
\pgfsetdash{}{0pt}%
\pgfpathmoveto{\pgfqpoint{0.100000in}{1.328214in}}%
\pgfpathlineto{\pgfqpoint{4.750000in}{1.328214in}}%
\pgfusepath{stroke}%
\end{pgfscope}%
\begin{pgfscope}%
\pgfpathrectangle{\pgfqpoint{0.100000in}{0.915754in}}{\pgfqpoint{4.650000in}{3.020000in}}%
\pgfusepath{clip}%
\pgfsetroundcap%
\pgfsetroundjoin%
\pgfsetlinewidth{1.003750pt}%
\definecolor{currentstroke}{rgb}{0.800000,0.800000,0.800000}%
\pgfsetstrokecolor{currentstroke}%
\pgfsetdash{}{0pt}%
\pgfpathmoveto{\pgfqpoint{0.100000in}{2.698375in}}%
\pgfpathlineto{\pgfqpoint{4.750000in}{2.698375in}}%
\pgfusepath{stroke}%
\end{pgfscope}%
\begin{pgfscope}%
\pgfpathrectangle{\pgfqpoint{0.100000in}{0.915754in}}{\pgfqpoint{4.650000in}{3.020000in}}%
\pgfusepath{clip}%
\pgfsetbuttcap%
\pgfsetroundjoin%
\definecolor{currentfill}{rgb}{0.298039,0.447059,0.690196}%
\pgfsetfillcolor{currentfill}%
\pgfsetfillopacity{0.200000}%
\pgfsetlinewidth{1.003750pt}%
\definecolor{currentstroke}{rgb}{0.298039,0.447059,0.690196}%
\pgfsetstrokecolor{currentstroke}%
\pgfsetstrokeopacity{0.200000}%
\pgfsetdash{}{0pt}%
\pgfpathmoveto{\pgfqpoint{-0.023216in}{2.681364in}}%
\pgfpathlineto{\pgfqpoint{-0.023216in}{2.618661in}}%
\pgfpathlineto{\pgfqpoint{0.150317in}{2.462761in}}%
\pgfpathlineto{\pgfqpoint{0.330263in}{2.365129in}}%
\pgfpathlineto{\pgfqpoint{0.504962in}{2.279616in}}%
\pgfpathlineto{\pgfqpoint{0.680121in}{2.140924in}}%
\pgfpathlineto{\pgfqpoint{0.858966in}{1.970155in}}%
\pgfpathlineto{\pgfqpoint{1.038283in}{2.013635in}}%
\pgfpathlineto{\pgfqpoint{1.218997in}{1.999786in}}%
\pgfpathlineto{\pgfqpoint{1.393196in}{1.870737in}}%
\pgfpathlineto{\pgfqpoint{1.566335in}{1.927385in}}%
\pgfpathlineto{\pgfqpoint{1.749073in}{1.881159in}}%
\pgfpathlineto{\pgfqpoint{1.929706in}{1.828480in}}%
\pgfpathlineto{\pgfqpoint{2.104929in}{1.694078in}}%
\pgfpathlineto{\pgfqpoint{2.281316in}{1.784915in}}%
\pgfpathlineto{\pgfqpoint{2.462692in}{1.731257in}}%
\pgfpathlineto{\pgfqpoint{2.634181in}{1.886018in}}%
\pgfpathlineto{\pgfqpoint{2.816873in}{1.782230in}}%
\pgfpathlineto{\pgfqpoint{2.996362in}{1.713723in}}%
\pgfpathlineto{\pgfqpoint{3.182860in}{1.705060in}}%
\pgfpathlineto{\pgfqpoint{3.357751in}{1.619491in}}%
\pgfpathlineto{\pgfqpoint{3.539658in}{1.560447in}}%
\pgfpathlineto{\pgfqpoint{3.719660in}{1.648855in}}%
\pgfpathlineto{\pgfqpoint{3.895151in}{1.632361in}}%
\pgfpathlineto{\pgfqpoint{4.079217in}{1.712719in}}%
\pgfpathlineto{\pgfqpoint{4.251720in}{1.539214in}}%
\pgfpathlineto{\pgfqpoint{4.426993in}{1.553950in}}%
\pgfpathlineto{\pgfqpoint{4.608779in}{1.578609in}}%
\pgfpathlineto{\pgfqpoint{4.792328in}{1.454520in}}%
\pgfpathlineto{\pgfqpoint{4.971522in}{1.566509in}}%
\pgfpathlineto{\pgfqpoint{5.144714in}{1.594595in}}%
\pgfpathlineto{\pgfqpoint{5.144714in}{1.806850in}}%
\pgfpathlineto{\pgfqpoint{5.144714in}{1.806850in}}%
\pgfpathlineto{\pgfqpoint{4.971522in}{1.756005in}}%
\pgfpathlineto{\pgfqpoint{4.792328in}{1.690940in}}%
\pgfpathlineto{\pgfqpoint{4.608779in}{1.788468in}}%
\pgfpathlineto{\pgfqpoint{4.426993in}{1.767435in}}%
\pgfpathlineto{\pgfqpoint{4.251720in}{1.772949in}}%
\pgfpathlineto{\pgfqpoint{4.079217in}{1.876383in}}%
\pgfpathlineto{\pgfqpoint{3.895151in}{1.792213in}}%
\pgfpathlineto{\pgfqpoint{3.719660in}{1.845365in}}%
\pgfpathlineto{\pgfqpoint{3.539658in}{1.810218in}}%
\pgfpathlineto{\pgfqpoint{3.357751in}{1.831148in}}%
\pgfpathlineto{\pgfqpoint{3.182860in}{1.863175in}}%
\pgfpathlineto{\pgfqpoint{2.996362in}{1.875748in}}%
\pgfpathlineto{\pgfqpoint{2.816873in}{1.923042in}}%
\pgfpathlineto{\pgfqpoint{2.634181in}{1.991747in}}%
\pgfpathlineto{\pgfqpoint{2.462692in}{1.914577in}}%
\pgfpathlineto{\pgfqpoint{2.281316in}{1.981065in}}%
\pgfpathlineto{\pgfqpoint{2.104929in}{1.940718in}}%
\pgfpathlineto{\pgfqpoint{1.929706in}{2.011502in}}%
\pgfpathlineto{\pgfqpoint{1.749073in}{2.059188in}}%
\pgfpathlineto{\pgfqpoint{1.566335in}{2.088433in}}%
\pgfpathlineto{\pgfqpoint{1.393196in}{2.024278in}}%
\pgfpathlineto{\pgfqpoint{1.218997in}{2.159993in}}%
\pgfpathlineto{\pgfqpoint{1.038283in}{2.213729in}}%
\pgfpathlineto{\pgfqpoint{0.858966in}{2.185788in}}%
\pgfpathlineto{\pgfqpoint{0.680121in}{2.338392in}}%
\pgfpathlineto{\pgfqpoint{0.504962in}{2.456329in}}%
\pgfpathlineto{\pgfqpoint{0.330263in}{2.509937in}}%
\pgfpathlineto{\pgfqpoint{0.150317in}{2.603907in}}%
\pgfpathlineto{\pgfqpoint{-0.023216in}{2.681364in}}%
\pgfpathclose%
\pgfusepath{stroke,fill}%
\end{pgfscope}%
\begin{pgfscope}%
\pgfpathrectangle{\pgfqpoint{0.100000in}{0.915754in}}{\pgfqpoint{4.650000in}{3.020000in}}%
\pgfusepath{clip}%
\pgfsetbuttcap%
\pgfsetroundjoin%
\definecolor{currentfill}{rgb}{0.866667,0.517647,0.321569}%
\pgfsetfillcolor{currentfill}%
\pgfsetfillopacity{0.200000}%
\pgfsetlinewidth{1.003750pt}%
\definecolor{currentstroke}{rgb}{0.866667,0.517647,0.321569}%
\pgfsetstrokecolor{currentstroke}%
\pgfsetstrokeopacity{0.200000}%
\pgfsetdash{}{0pt}%
\pgfpathmoveto{\pgfqpoint{-0.023703in}{2.444870in}}%
\pgfpathlineto{\pgfqpoint{-0.023703in}{2.270984in}}%
\pgfpathlineto{\pgfqpoint{0.150508in}{2.245443in}}%
\pgfpathlineto{\pgfqpoint{0.325594in}{1.958852in}}%
\pgfpathlineto{\pgfqpoint{0.503074in}{2.016406in}}%
\pgfpathlineto{\pgfqpoint{0.678648in}{1.738252in}}%
\pgfpathlineto{\pgfqpoint{0.858059in}{1.517742in}}%
\pgfpathlineto{\pgfqpoint{1.035538in}{1.478758in}}%
\pgfpathlineto{\pgfqpoint{1.212460in}{1.512598in}}%
\pgfpathlineto{\pgfqpoint{1.391935in}{1.556721in}}%
\pgfpathlineto{\pgfqpoint{1.567118in}{1.423712in}}%
\pgfpathlineto{\pgfqpoint{1.746357in}{1.440928in}}%
\pgfpathlineto{\pgfqpoint{1.923051in}{1.460100in}}%
\pgfpathlineto{\pgfqpoint{2.100660in}{1.517890in}}%
\pgfpathlineto{\pgfqpoint{2.281891in}{1.542567in}}%
\pgfpathlineto{\pgfqpoint{2.458390in}{1.555871in}}%
\pgfpathlineto{\pgfqpoint{2.638727in}{1.464444in}}%
\pgfpathlineto{\pgfqpoint{2.813287in}{1.423486in}}%
\pgfpathlineto{\pgfqpoint{2.996901in}{1.527649in}}%
\pgfpathlineto{\pgfqpoint{3.173109in}{1.143757in}}%
\pgfpathlineto{\pgfqpoint{3.353023in}{1.410019in}}%
\pgfpathlineto{\pgfqpoint{3.532967in}{1.274828in}}%
\pgfpathlineto{\pgfqpoint{3.713024in}{1.310896in}}%
\pgfpathlineto{\pgfqpoint{3.887521in}{1.192775in}}%
\pgfpathlineto{\pgfqpoint{4.064139in}{1.387977in}}%
\pgfpathlineto{\pgfqpoint{4.248630in}{1.310987in}}%
\pgfpathlineto{\pgfqpoint{4.423648in}{1.354173in}}%
\pgfpathlineto{\pgfqpoint{4.607255in}{1.341267in}}%
\pgfpathlineto{\pgfqpoint{4.783147in}{1.377142in}}%
\pgfpathlineto{\pgfqpoint{4.958584in}{1.254973in}}%
\pgfpathlineto{\pgfqpoint{5.141997in}{1.354634in}}%
\pgfpathlineto{\pgfqpoint{5.141997in}{1.620537in}}%
\pgfpathlineto{\pgfqpoint{5.141997in}{1.620537in}}%
\pgfpathlineto{\pgfqpoint{4.958584in}{1.526431in}}%
\pgfpathlineto{\pgfqpoint{4.783147in}{1.639307in}}%
\pgfpathlineto{\pgfqpoint{4.607255in}{1.566048in}}%
\pgfpathlineto{\pgfqpoint{4.423648in}{1.631341in}}%
\pgfpathlineto{\pgfqpoint{4.248630in}{1.566554in}}%
\pgfpathlineto{\pgfqpoint{4.064139in}{1.666453in}}%
\pgfpathlineto{\pgfqpoint{3.887521in}{1.500612in}}%
\pgfpathlineto{\pgfqpoint{3.713024in}{1.538510in}}%
\pgfpathlineto{\pgfqpoint{3.532967in}{1.563579in}}%
\pgfpathlineto{\pgfqpoint{3.353023in}{1.673305in}}%
\pgfpathlineto{\pgfqpoint{3.173109in}{1.458990in}}%
\pgfpathlineto{\pgfqpoint{2.996901in}{1.715680in}}%
\pgfpathlineto{\pgfqpoint{2.813287in}{1.649319in}}%
\pgfpathlineto{\pgfqpoint{2.638727in}{1.700822in}}%
\pgfpathlineto{\pgfqpoint{2.458390in}{1.791514in}}%
\pgfpathlineto{\pgfqpoint{2.281891in}{1.726665in}}%
\pgfpathlineto{\pgfqpoint{2.100660in}{1.782128in}}%
\pgfpathlineto{\pgfqpoint{1.923051in}{1.751695in}}%
\pgfpathlineto{\pgfqpoint{1.746357in}{1.678560in}}%
\pgfpathlineto{\pgfqpoint{1.567118in}{1.672572in}}%
\pgfpathlineto{\pgfqpoint{1.391935in}{1.849518in}}%
\pgfpathlineto{\pgfqpoint{1.212460in}{1.769060in}}%
\pgfpathlineto{\pgfqpoint{1.035538in}{1.745279in}}%
\pgfpathlineto{\pgfqpoint{0.858059in}{1.802211in}}%
\pgfpathlineto{\pgfqpoint{0.678648in}{2.014988in}}%
\pgfpathlineto{\pgfqpoint{0.503074in}{2.261731in}}%
\pgfpathlineto{\pgfqpoint{0.325594in}{2.260811in}}%
\pgfpathlineto{\pgfqpoint{0.150508in}{2.411857in}}%
\pgfpathlineto{\pgfqpoint{-0.023703in}{2.444870in}}%
\pgfpathclose%
\pgfusepath{stroke,fill}%
\end{pgfscope}%
\begin{pgfscope}%
\pgfpathrectangle{\pgfqpoint{0.100000in}{0.915754in}}{\pgfqpoint{4.650000in}{3.020000in}}%
\pgfusepath{clip}%
\pgfsetbuttcap%
\pgfsetroundjoin%
\definecolor{currentfill}{rgb}{0.333333,0.658824,0.407843}%
\pgfsetfillcolor{currentfill}%
\pgfsetfillopacity{0.200000}%
\pgfsetlinewidth{1.003750pt}%
\definecolor{currentstroke}{rgb}{0.333333,0.658824,0.407843}%
\pgfsetstrokecolor{currentstroke}%
\pgfsetstrokeopacity{0.200000}%
\pgfsetdash{}{0pt}%
\pgfpathmoveto{\pgfqpoint{0.019732in}{3.114565in}}%
\pgfpathlineto{\pgfqpoint{0.019732in}{2.845888in}}%
\pgfpathlineto{\pgfqpoint{0.202666in}{2.497038in}}%
\pgfpathlineto{\pgfqpoint{0.411550in}{2.294851in}}%
\pgfpathlineto{\pgfqpoint{0.573073in}{2.184369in}}%
\pgfpathlineto{\pgfqpoint{0.744032in}{1.910695in}}%
\pgfpathlineto{\pgfqpoint{0.935057in}{1.920400in}}%
\pgfpathlineto{\pgfqpoint{1.109339in}{1.905007in}}%
\pgfpathlineto{\pgfqpoint{1.295440in}{1.814514in}}%
\pgfpathlineto{\pgfqpoint{1.465631in}{1.702730in}}%
\pgfpathlineto{\pgfqpoint{1.657170in}{1.666729in}}%
\pgfpathlineto{\pgfqpoint{1.848154in}{1.668575in}}%
\pgfpathlineto{\pgfqpoint{2.022230in}{1.745347in}}%
\pgfpathlineto{\pgfqpoint{2.201303in}{1.609488in}}%
\pgfpathlineto{\pgfqpoint{2.380178in}{1.463357in}}%
\pgfpathlineto{\pgfqpoint{2.580704in}{1.617785in}}%
\pgfpathlineto{\pgfqpoint{2.754383in}{1.526979in}}%
\pgfpathlineto{\pgfqpoint{2.926043in}{1.674506in}}%
\pgfpathlineto{\pgfqpoint{3.124081in}{1.440639in}}%
\pgfpathlineto{\pgfqpoint{3.295549in}{1.516180in}}%
\pgfpathlineto{\pgfqpoint{3.499097in}{1.505872in}}%
\pgfpathlineto{\pgfqpoint{3.667629in}{1.225431in}}%
\pgfpathlineto{\pgfqpoint{3.867661in}{1.540191in}}%
\pgfpathlineto{\pgfqpoint{4.047629in}{1.524559in}}%
\pgfpathlineto{\pgfqpoint{4.201929in}{1.510970in}}%
\pgfpathlineto{\pgfqpoint{4.382966in}{1.414407in}}%
\pgfpathlineto{\pgfqpoint{4.563303in}{1.471307in}}%
\pgfpathlineto{\pgfqpoint{4.760119in}{1.552852in}}%
\pgfpathlineto{\pgfqpoint{4.935133in}{1.460880in}}%
\pgfpathlineto{\pgfqpoint{5.116337in}{1.341289in}}%
\pgfpathlineto{\pgfqpoint{5.317853in}{1.398312in}}%
\pgfpathlineto{\pgfqpoint{5.317853in}{1.651341in}}%
\pgfpathlineto{\pgfqpoint{5.317853in}{1.651341in}}%
\pgfpathlineto{\pgfqpoint{5.116337in}{1.622618in}}%
\pgfpathlineto{\pgfqpoint{4.935133in}{1.719596in}}%
\pgfpathlineto{\pgfqpoint{4.760119in}{1.820330in}}%
\pgfpathlineto{\pgfqpoint{4.563303in}{1.744089in}}%
\pgfpathlineto{\pgfqpoint{4.382966in}{1.689754in}}%
\pgfpathlineto{\pgfqpoint{4.201929in}{1.759318in}}%
\pgfpathlineto{\pgfqpoint{4.047629in}{1.782905in}}%
\pgfpathlineto{\pgfqpoint{3.867661in}{1.744124in}}%
\pgfpathlineto{\pgfqpoint{3.667629in}{1.514084in}}%
\pgfpathlineto{\pgfqpoint{3.499097in}{1.732808in}}%
\pgfpathlineto{\pgfqpoint{3.295549in}{1.865007in}}%
\pgfpathlineto{\pgfqpoint{3.124081in}{1.707643in}}%
\pgfpathlineto{\pgfqpoint{2.926043in}{2.037929in}}%
\pgfpathlineto{\pgfqpoint{2.754383in}{1.948070in}}%
\pgfpathlineto{\pgfqpoint{2.580704in}{1.926375in}}%
\pgfpathlineto{\pgfqpoint{2.380178in}{1.755012in}}%
\pgfpathlineto{\pgfqpoint{2.201303in}{1.883544in}}%
\pgfpathlineto{\pgfqpoint{2.022230in}{2.062942in}}%
\pgfpathlineto{\pgfqpoint{1.848154in}{2.063810in}}%
\pgfpathlineto{\pgfqpoint{1.657170in}{1.957935in}}%
\pgfpathlineto{\pgfqpoint{1.465631in}{2.002361in}}%
\pgfpathlineto{\pgfqpoint{1.295440in}{2.153510in}}%
\pgfpathlineto{\pgfqpoint{1.109339in}{2.294371in}}%
\pgfpathlineto{\pgfqpoint{0.935057in}{2.216847in}}%
\pgfpathlineto{\pgfqpoint{0.744032in}{2.251420in}}%
\pgfpathlineto{\pgfqpoint{0.573073in}{2.520507in}}%
\pgfpathlineto{\pgfqpoint{0.411550in}{2.684320in}}%
\pgfpathlineto{\pgfqpoint{0.202666in}{2.875892in}}%
\pgfpathlineto{\pgfqpoint{0.019732in}{3.114565in}}%
\pgfpathclose%
\pgfusepath{stroke,fill}%
\end{pgfscope}%
\begin{pgfscope}%
\pgfpathrectangle{\pgfqpoint{0.100000in}{0.915754in}}{\pgfqpoint{4.650000in}{3.020000in}}%
\pgfusepath{clip}%
\pgfsetbuttcap%
\pgfsetroundjoin%
\definecolor{currentfill}{rgb}{0.768627,0.305882,0.321569}%
\pgfsetfillcolor{currentfill}%
\pgfsetfillopacity{0.200000}%
\pgfsetlinewidth{1.003750pt}%
\definecolor{currentstroke}{rgb}{0.768627,0.305882,0.321569}%
\pgfsetstrokecolor{currentstroke}%
\pgfsetstrokeopacity{0.200000}%
\pgfsetdash{}{0pt}%
\pgfpathmoveto{\pgfqpoint{-0.066071in}{2.636370in}}%
\pgfpathlineto{\pgfqpoint{-0.066071in}{2.280019in}}%
\pgfpathlineto{\pgfqpoint{0.105727in}{2.391516in}}%
\pgfpathlineto{\pgfqpoint{0.277525in}{2.495244in}}%
\pgfpathlineto{\pgfqpoint{0.449323in}{2.252005in}}%
\pgfpathlineto{\pgfqpoint{0.621121in}{2.121116in}}%
\pgfpathlineto{\pgfqpoint{0.792919in}{2.317447in}}%
\pgfpathlineto{\pgfqpoint{0.964717in}{2.004304in}}%
\pgfpathlineto{\pgfqpoint{1.136515in}{2.156962in}}%
\pgfpathlineto{\pgfqpoint{1.308313in}{2.122430in}}%
\pgfpathlineto{\pgfqpoint{1.480111in}{2.071344in}}%
\pgfpathlineto{\pgfqpoint{1.651909in}{2.017551in}}%
\pgfpathlineto{\pgfqpoint{1.823707in}{1.998112in}}%
\pgfpathlineto{\pgfqpoint{1.995505in}{1.861131in}}%
\pgfpathlineto{\pgfqpoint{2.167303in}{1.786220in}}%
\pgfpathlineto{\pgfqpoint{2.339101in}{2.062991in}}%
\pgfpathlineto{\pgfqpoint{2.510899in}{1.942075in}}%
\pgfpathlineto{\pgfqpoint{2.682697in}{1.883175in}}%
\pgfpathlineto{\pgfqpoint{2.854495in}{1.928380in}}%
\pgfpathlineto{\pgfqpoint{3.026293in}{1.964007in}}%
\pgfpathlineto{\pgfqpoint{3.198091in}{1.978929in}}%
\pgfpathlineto{\pgfqpoint{3.369889in}{1.909236in}}%
\pgfpathlineto{\pgfqpoint{3.541687in}{1.852799in}}%
\pgfpathlineto{\pgfqpoint{3.713485in}{1.981998in}}%
\pgfpathlineto{\pgfqpoint{3.885283in}{1.839208in}}%
\pgfpathlineto{\pgfqpoint{4.057081in}{1.965641in}}%
\pgfpathlineto{\pgfqpoint{4.228879in}{1.648178in}}%
\pgfpathlineto{\pgfqpoint{4.400677in}{1.630176in}}%
\pgfpathlineto{\pgfqpoint{4.572475in}{1.755882in}}%
\pgfpathlineto{\pgfqpoint{4.744273in}{1.893326in}}%
\pgfpathlineto{\pgfqpoint{4.916071in}{1.738325in}}%
\pgfpathlineto{\pgfqpoint{4.916071in}{2.024987in}}%
\pgfpathlineto{\pgfqpoint{4.916071in}{2.024987in}}%
\pgfpathlineto{\pgfqpoint{4.744273in}{2.165633in}}%
\pgfpathlineto{\pgfqpoint{4.572475in}{2.096489in}}%
\pgfpathlineto{\pgfqpoint{4.400677in}{2.014152in}}%
\pgfpathlineto{\pgfqpoint{4.228879in}{2.027235in}}%
\pgfpathlineto{\pgfqpoint{4.057081in}{2.227000in}}%
\pgfpathlineto{\pgfqpoint{3.885283in}{2.120764in}}%
\pgfpathlineto{\pgfqpoint{3.713485in}{2.266368in}}%
\pgfpathlineto{\pgfqpoint{3.541687in}{2.207636in}}%
\pgfpathlineto{\pgfqpoint{3.369889in}{2.203792in}}%
\pgfpathlineto{\pgfqpoint{3.198091in}{2.280039in}}%
\pgfpathlineto{\pgfqpoint{3.026293in}{2.224131in}}%
\pgfpathlineto{\pgfqpoint{2.854495in}{2.252031in}}%
\pgfpathlineto{\pgfqpoint{2.682697in}{2.229853in}}%
\pgfpathlineto{\pgfqpoint{2.510899in}{2.254722in}}%
\pgfpathlineto{\pgfqpoint{2.339101in}{2.329166in}}%
\pgfpathlineto{\pgfqpoint{2.167303in}{2.130846in}}%
\pgfpathlineto{\pgfqpoint{1.995505in}{2.279721in}}%
\pgfpathlineto{\pgfqpoint{1.823707in}{2.364146in}}%
\pgfpathlineto{\pgfqpoint{1.651909in}{2.370832in}}%
\pgfpathlineto{\pgfqpoint{1.480111in}{2.418990in}}%
\pgfpathlineto{\pgfqpoint{1.308313in}{2.429529in}}%
\pgfpathlineto{\pgfqpoint{1.136515in}{2.430470in}}%
\pgfpathlineto{\pgfqpoint{0.964717in}{2.359829in}}%
\pgfpathlineto{\pgfqpoint{0.792919in}{2.601170in}}%
\pgfpathlineto{\pgfqpoint{0.621121in}{2.434544in}}%
\pgfpathlineto{\pgfqpoint{0.449323in}{2.581409in}}%
\pgfpathlineto{\pgfqpoint{0.277525in}{2.841838in}}%
\pgfpathlineto{\pgfqpoint{0.105727in}{2.750420in}}%
\pgfpathlineto{\pgfqpoint{-0.066071in}{2.636370in}}%
\pgfpathclose%
\pgfusepath{stroke,fill}%
\end{pgfscope}%
\begin{pgfscope}%
\pgfpathrectangle{\pgfqpoint{0.100000in}{0.915754in}}{\pgfqpoint{4.650000in}{3.020000in}}%
\pgfusepath{clip}%
\pgfsetroundcap%
\pgfsetroundjoin%
\pgfsetlinewidth{1.505625pt}%
\definecolor{currentstroke}{rgb}{0.298039,0.447059,0.690196}%
\pgfsetstrokecolor{currentstroke}%
\pgfsetdash{}{0pt}%
\pgfpathmoveto{\pgfqpoint{0.086111in}{2.583010in}}%
\pgfpathlineto{\pgfqpoint{0.150317in}{2.543857in}}%
\pgfpathlineto{\pgfqpoint{0.330263in}{2.443865in}}%
\pgfpathlineto{\pgfqpoint{0.504962in}{2.368433in}}%
\pgfpathlineto{\pgfqpoint{0.680121in}{2.251366in}}%
\pgfpathlineto{\pgfqpoint{0.858966in}{2.081737in}}%
\pgfpathlineto{\pgfqpoint{1.038283in}{2.119906in}}%
\pgfpathlineto{\pgfqpoint{1.218997in}{2.082959in}}%
\pgfpathlineto{\pgfqpoint{1.393196in}{1.950796in}}%
\pgfpathlineto{\pgfqpoint{1.566335in}{2.012466in}}%
\pgfpathlineto{\pgfqpoint{1.749073in}{1.978249in}}%
\pgfpathlineto{\pgfqpoint{1.929706in}{1.928856in}}%
\pgfpathlineto{\pgfqpoint{2.104929in}{1.827470in}}%
\pgfpathlineto{\pgfqpoint{2.281316in}{1.893115in}}%
\pgfpathlineto{\pgfqpoint{2.462692in}{1.830523in}}%
\pgfpathlineto{\pgfqpoint{2.634181in}{1.940343in}}%
\pgfpathlineto{\pgfqpoint{2.816873in}{1.860263in}}%
\pgfpathlineto{\pgfqpoint{2.996362in}{1.798214in}}%
\pgfpathlineto{\pgfqpoint{3.182860in}{1.789262in}}%
\pgfpathlineto{\pgfqpoint{3.357751in}{1.735077in}}%
\pgfpathlineto{\pgfqpoint{3.539658in}{1.697074in}}%
\pgfpathlineto{\pgfqpoint{3.719660in}{1.755632in}}%
\pgfpathlineto{\pgfqpoint{3.895151in}{1.717278in}}%
\pgfpathlineto{\pgfqpoint{4.079217in}{1.798315in}}%
\pgfpathlineto{\pgfqpoint{4.251720in}{1.670095in}}%
\pgfpathlineto{\pgfqpoint{4.426993in}{1.667281in}}%
\pgfpathlineto{\pgfqpoint{4.608779in}{1.694084in}}%
\pgfpathlineto{\pgfqpoint{4.763889in}{1.600562in}}%
\pgfusepath{stroke}%
\end{pgfscope}%
\begin{pgfscope}%
\pgfpathrectangle{\pgfqpoint{0.100000in}{0.915754in}}{\pgfqpoint{4.650000in}{3.020000in}}%
\pgfusepath{clip}%
\pgfsetroundcap%
\pgfsetroundjoin%
\pgfsetlinewidth{1.505625pt}%
\definecolor{currentstroke}{rgb}{0.866667,0.517647,0.321569}%
\pgfsetstrokecolor{currentstroke}%
\pgfsetdash{}{0pt}%
\pgfpathmoveto{\pgfqpoint{0.086111in}{2.346988in}}%
\pgfpathlineto{\pgfqpoint{0.150508in}{2.334586in}}%
\pgfpathlineto{\pgfqpoint{0.325594in}{2.128887in}}%
\pgfpathlineto{\pgfqpoint{0.503074in}{2.145865in}}%
\pgfpathlineto{\pgfqpoint{0.678648in}{1.892939in}}%
\pgfpathlineto{\pgfqpoint{0.858059in}{1.666599in}}%
\pgfpathlineto{\pgfqpoint{1.035538in}{1.623053in}}%
\pgfpathlineto{\pgfqpoint{1.212460in}{1.648744in}}%
\pgfpathlineto{\pgfqpoint{1.391935in}{1.706179in}}%
\pgfpathlineto{\pgfqpoint{1.567118in}{1.563046in}}%
\pgfpathlineto{\pgfqpoint{1.746357in}{1.572394in}}%
\pgfpathlineto{\pgfqpoint{1.923051in}{1.613472in}}%
\pgfpathlineto{\pgfqpoint{2.100660in}{1.665254in}}%
\pgfpathlineto{\pgfqpoint{2.281891in}{1.645324in}}%
\pgfpathlineto{\pgfqpoint{2.458390in}{1.691456in}}%
\pgfpathlineto{\pgfqpoint{2.638727in}{1.588168in}}%
\pgfpathlineto{\pgfqpoint{2.813287in}{1.542035in}}%
\pgfpathlineto{\pgfqpoint{2.996901in}{1.625655in}}%
\pgfpathlineto{\pgfqpoint{3.173109in}{1.317382in}}%
\pgfpathlineto{\pgfqpoint{3.353023in}{1.554374in}}%
\pgfpathlineto{\pgfqpoint{3.532967in}{1.428432in}}%
\pgfpathlineto{\pgfqpoint{3.713024in}{1.434963in}}%
\pgfpathlineto{\pgfqpoint{3.887521in}{1.367895in}}%
\pgfpathlineto{\pgfqpoint{4.064139in}{1.532133in}}%
\pgfpathlineto{\pgfqpoint{4.248630in}{1.453008in}}%
\pgfpathlineto{\pgfqpoint{4.423648in}{1.502582in}}%
\pgfpathlineto{\pgfqpoint{4.607255in}{1.461669in}}%
\pgfpathlineto{\pgfqpoint{4.763889in}{1.516345in}}%
\pgfusepath{stroke}%
\end{pgfscope}%
\begin{pgfscope}%
\pgfpathrectangle{\pgfqpoint{0.100000in}{0.915754in}}{\pgfqpoint{4.650000in}{3.020000in}}%
\pgfusepath{clip}%
\pgfsetroundcap%
\pgfsetroundjoin%
\pgfsetlinewidth{1.505625pt}%
\definecolor{currentstroke}{rgb}{0.333333,0.658824,0.407843}%
\pgfsetstrokecolor{currentstroke}%
\pgfsetdash{}{0pt}%
\pgfpathmoveto{\pgfqpoint{0.086111in}{2.881140in}}%
\pgfpathlineto{\pgfqpoint{0.202666in}{2.699101in}}%
\pgfpathlineto{\pgfqpoint{0.411550in}{2.504833in}}%
\pgfpathlineto{\pgfqpoint{0.573073in}{2.366020in}}%
\pgfpathlineto{\pgfqpoint{0.744032in}{2.099922in}}%
\pgfpathlineto{\pgfqpoint{0.935057in}{2.084122in}}%
\pgfpathlineto{\pgfqpoint{1.109339in}{2.120055in}}%
\pgfpathlineto{\pgfqpoint{1.295440in}{2.008562in}}%
\pgfpathlineto{\pgfqpoint{1.465631in}{1.864847in}}%
\pgfpathlineto{\pgfqpoint{1.657170in}{1.821448in}}%
\pgfpathlineto{\pgfqpoint{1.848154in}{1.876812in}}%
\pgfpathlineto{\pgfqpoint{2.022230in}{1.919131in}}%
\pgfpathlineto{\pgfqpoint{2.201303in}{1.761556in}}%
\pgfpathlineto{\pgfqpoint{2.380178in}{1.628556in}}%
\pgfpathlineto{\pgfqpoint{2.580704in}{1.781993in}}%
\pgfpathlineto{\pgfqpoint{2.754383in}{1.748040in}}%
\pgfpathlineto{\pgfqpoint{2.926043in}{1.862547in}}%
\pgfpathlineto{\pgfqpoint{3.124081in}{1.587666in}}%
\pgfpathlineto{\pgfqpoint{3.295549in}{1.711151in}}%
\pgfpathlineto{\pgfqpoint{3.499097in}{1.637342in}}%
\pgfpathlineto{\pgfqpoint{3.667629in}{1.386486in}}%
\pgfpathlineto{\pgfqpoint{3.867661in}{1.650304in}}%
\pgfpathlineto{\pgfqpoint{4.047629in}{1.657759in}}%
\pgfpathlineto{\pgfqpoint{4.201929in}{1.651501in}}%
\pgfpathlineto{\pgfqpoint{4.382966in}{1.561472in}}%
\pgfpathlineto{\pgfqpoint{4.563303in}{1.615510in}}%
\pgfpathlineto{\pgfqpoint{4.760119in}{1.702750in}}%
\pgfpathlineto{\pgfqpoint{4.763889in}{1.700798in}}%
\pgfusepath{stroke}%
\end{pgfscope}%
\begin{pgfscope}%
\pgfpathrectangle{\pgfqpoint{0.100000in}{0.915754in}}{\pgfqpoint{4.650000in}{3.020000in}}%
\pgfusepath{clip}%
\pgfsetroundcap%
\pgfsetroundjoin%
\pgfsetlinewidth{1.505625pt}%
\definecolor{currentstroke}{rgb}{0.768627,0.305882,0.321569}%
\pgfsetstrokecolor{currentstroke}%
\pgfsetdash{}{0pt}%
\pgfpathmoveto{\pgfqpoint{0.086111in}{2.576302in}}%
\pgfpathlineto{\pgfqpoint{0.105727in}{2.589522in}}%
\pgfpathlineto{\pgfqpoint{0.277525in}{2.686062in}}%
\pgfpathlineto{\pgfqpoint{0.449323in}{2.434677in}}%
\pgfpathlineto{\pgfqpoint{0.621121in}{2.308725in}}%
\pgfpathlineto{\pgfqpoint{0.792919in}{2.466746in}}%
\pgfpathlineto{\pgfqpoint{0.964717in}{2.203037in}}%
\pgfpathlineto{\pgfqpoint{1.136515in}{2.299172in}}%
\pgfpathlineto{\pgfqpoint{1.308313in}{2.295897in}}%
\pgfpathlineto{\pgfqpoint{1.480111in}{2.274032in}}%
\pgfpathlineto{\pgfqpoint{1.651909in}{2.219240in}}%
\pgfpathlineto{\pgfqpoint{1.823707in}{2.205287in}}%
\pgfpathlineto{\pgfqpoint{1.995505in}{2.088876in}}%
\pgfpathlineto{\pgfqpoint{2.167303in}{1.976806in}}%
\pgfpathlineto{\pgfqpoint{2.339101in}{2.202683in}}%
\pgfpathlineto{\pgfqpoint{2.510899in}{2.110058in}}%
\pgfpathlineto{\pgfqpoint{2.682697in}{2.071051in}}%
\pgfpathlineto{\pgfqpoint{2.854495in}{2.106327in}}%
\pgfpathlineto{\pgfqpoint{3.026293in}{2.109310in}}%
\pgfpathlineto{\pgfqpoint{3.198091in}{2.146078in}}%
\pgfpathlineto{\pgfqpoint{3.369889in}{2.073896in}}%
\pgfpathlineto{\pgfqpoint{3.541687in}{2.054718in}}%
\pgfpathlineto{\pgfqpoint{3.713485in}{2.137257in}}%
\pgfpathlineto{\pgfqpoint{3.885283in}{2.006494in}}%
\pgfpathlineto{\pgfqpoint{4.057081in}{2.112163in}}%
\pgfpathlineto{\pgfqpoint{4.228879in}{1.853323in}}%
\pgfpathlineto{\pgfqpoint{4.400677in}{1.832335in}}%
\pgfpathlineto{\pgfqpoint{4.572475in}{1.949566in}}%
\pgfpathlineto{\pgfqpoint{4.744273in}{2.040374in}}%
\pgfpathlineto{\pgfqpoint{4.763889in}{2.023611in}}%
\pgfusepath{stroke}%
\end{pgfscope}%
\begin{pgfscope}%
\pgfsetrectcap%
\pgfsetmiterjoin%
\pgfsetlinewidth{1.254687pt}%
\definecolor{currentstroke}{rgb}{0.800000,0.800000,0.800000}%
\pgfsetstrokecolor{currentstroke}%
\pgfsetdash{}{0pt}%
\pgfpathmoveto{\pgfqpoint{0.100000in}{0.915754in}}%
\pgfpathlineto{\pgfqpoint{0.100000in}{3.935754in}}%
\pgfusepath{stroke}%
\end{pgfscope}%
\begin{pgfscope}%
\pgfsetrectcap%
\pgfsetmiterjoin%
\pgfsetlinewidth{1.254687pt}%
\definecolor{currentstroke}{rgb}{0.800000,0.800000,0.800000}%
\pgfsetstrokecolor{currentstroke}%
\pgfsetdash{}{0pt}%
\pgfpathmoveto{\pgfqpoint{4.750000in}{0.915754in}}%
\pgfpathlineto{\pgfqpoint{4.750000in}{3.935754in}}%
\pgfusepath{stroke}%
\end{pgfscope}%
\begin{pgfscope}%
\pgfsetrectcap%
\pgfsetmiterjoin%
\pgfsetlinewidth{1.254687pt}%
\definecolor{currentstroke}{rgb}{0.800000,0.800000,0.800000}%
\pgfsetstrokecolor{currentstroke}%
\pgfsetdash{}{0pt}%
\pgfpathmoveto{\pgfqpoint{0.100000in}{0.915754in}}%
\pgfpathlineto{\pgfqpoint{4.750000in}{0.915754in}}%
\pgfusepath{stroke}%
\end{pgfscope}%
\begin{pgfscope}%
\pgfsetrectcap%
\pgfsetmiterjoin%
\pgfsetlinewidth{1.254687pt}%
\definecolor{currentstroke}{rgb}{0.800000,0.800000,0.800000}%
\pgfsetstrokecolor{currentstroke}%
\pgfsetdash{}{0pt}%
\pgfpathmoveto{\pgfqpoint{0.100000in}{3.935754in}}%
\pgfpathlineto{\pgfqpoint{4.750000in}{3.935754in}}%
\pgfusepath{stroke}%
\end{pgfscope}%
\end{pgfpicture}%
\makeatother%
\endgroup%